\pdfoutput=1

\documentclass[runningheads]{llncs}
\usepackage{graphicx}

\usepackage{tikz}
\usepackage{comment}
\usepackage{amsmath,amssymb} 
\usepackage{color}

\usepackage[accsupp]{axessibility}  

\usepackage[width=122mm,left=12mm,paperwidth=146mm,height=193mm,top=12mm,paperheight=217mm]{geometry}

\usepackage{booktabs, multirow} 
\usepackage{soul}
\usepackage{changepage,threeparttable} 
\usepackage{adjustbox}
\usepackage{appendix}
\usepackage{hyperref}

\usepackage[ruled,linesnumbered]{algorithm2e}
\usepackage{rotating}

\newcommand{\etal}{\textit{et al. }}

\begin{document}
\pagestyle{headings}
\mainmatter
\def\ECCVSubNumber{3181}  

\title{
Discovering Transferable Forensic Features\\
for CNN-generated Images Detection
} 


\titlerunning{Discovering Transferable Forensic Features}
%
\author{
Keshigeyan Chandrasegaran\inst{1} \and
Ngoc-Trung Tran\inst{1} \and
Alexander Binder\inst{2, 3}\and
Ngai-Man Cheung\inst{1}
}
\authorrunning{K. Chandrasegaran et al.}
%
\institute{Singapore University of Technology and Design (SUTD) \\ \email{\{keshigeyan, ngoctrung\_tran, ngaiman\_cheung\}@sutd.edu.sg} \and
{Singapore Institute of Technology (SIT) \and
University of Oslo (UIO)} \\
\email{alexander.binder@singaporetech.edu.sg\quad alexabin@uio.no}
}
\maketitle

\RestyleAlgo{ruled}
\SetKwComment{Comment}{/*}{*/}
\SetKwInput{KwInput}{Input}                
\SetKwInput{KwOutput}{Output}              

\vspace{-0.5cm}
\begin{abstract}
Visual counterfeits \footnote{We refer to CNN-generated images as counterfeits throughout this paper} are increasingly causing an existential conundrum in mainstream media with rapid evolution in neural image synthesis methods.
Though detection of such counterfeits has been a taxing problem in the image forensics community, a recent class of forensic detectors -- \textit{universal detectors} -- are able to surprisingly spot counterfeit images regardless of generator architectures, loss functions, training datasets, and resolutions \cite{Wang_2020_CVPR}.
This intriguing property suggests the possible existence of \textit{transferable forensic features (T-FF)} in \textit{universal detectors}. 
In this work, we conduct the first analytical study to discover and understand \textit{T-FF} in \textit{universal detectors}. 
Our contributions are 2-fold:
1) We propose a novel \textit{forensic feature relevance statistic (FF-RS)} to quantify and discover \textit{T-FF} in \textit{universal detectors} and,
2) Our qualitative and quantitative investigations uncover an unexpected finding: \textit{color} is a critical \textit{T-FF} in \textit{universal detectors}.
Code and models are available at \url{https://keshik6.github.io/transferable-forensic-features/}
\end{abstract}

\begin{figure}[!h]
\centering
\begin{tabular}{ccccccc}
    \multicolumn{1}{p{0.125\linewidth}}{\tiny \enskip ProGAN \cite{karras2018progressive}} &
    \multicolumn{1}{p{0.15\linewidth}}{\tiny  \enskip StyleGAN2 \cite{Karras_2020_CVPR}} &
    \multicolumn{1}{p{0.14\linewidth}}{\tiny StyleGAN \cite{Karras_2019_CVPR}} &
    \multicolumn{1}{p{0.125\linewidth}}{\tiny BigGAN \cite{brock2018large}} &
    \multicolumn{1}{p{0.132\linewidth}}{\tiny CycleGAN \cite{zhu2017unpaired}} &
    \multicolumn{1}{p{0.135\linewidth}}{\tiny \enskip StarGAN \cite{choi2018stargan}} &
    {\tiny GauGAN \cite{park2019semantic}} \\

    \multicolumn{7}{c}{\includegraphics[width=0.99\linewidth]{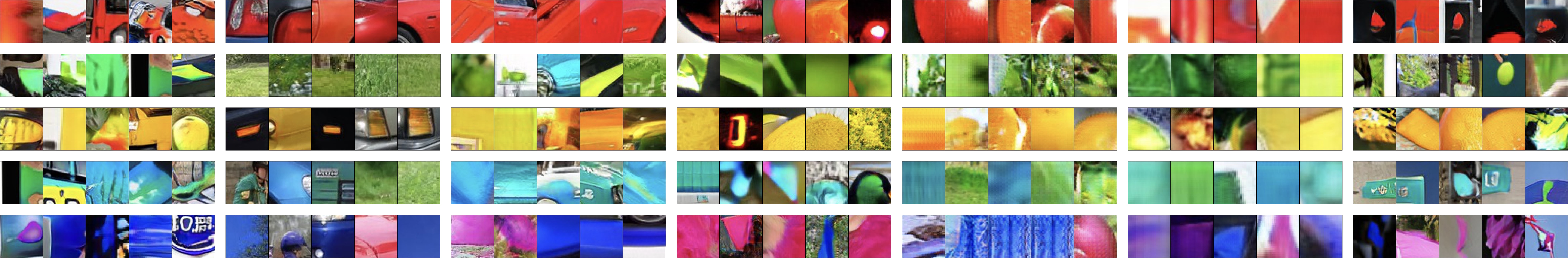}}

\end{tabular}
\vspace{-0.3cm}
\caption{
Color is a critical \textit{transferable forensic feature (T-FF)} in universal detectors:
Large-scale study on visual interpretability of \textit{T-FF} discovered through our proposed \textit{forensic feature relevance statistic (FF-RS)
,} reveal that color information is critical for cross-model forensic transfer.
Each row represents a color-conditional \textit{T-FF} and
we show the LRP-max response regions for different GAN 
counterfeits 
for the publicly released ResNet-50 universal detector by Wang \etal \cite{Wang_2020_CVPR}.
This detector is trained with ProGAN
\cite{karras2018progressive} 
counterfeits \cite{Wang_2020_CVPR} and cross-model forensic transfer is evaluated on unseen GANs.
All counterfeits are obtained from the ForenSynths dataset 
\cite{Wang_2020_CVPR}.
The consistent color-conditional LRP-max response across all GANs for these \textit{T-FF} clearly indicate that \textit{color} is critical for cross-model forensic transfer in universal detectors.
We further observe similar results using an EfficientNet-B0-based \cite{tan2019efficientnet} universal detector following the exact training / test strategy proposed by 
Wang \etal 
\cite{Wang_2020_CVPR}
in Fig. \ref{fig:lrp_patches_efb0}.
More visualizations are included in Supplementary
\ref{sec_supp:color_tff}.
}
\label{fig:lrp_patches_r50}
\vspace{-0.1cm}
\end{figure}

\section{Introduction}
\label{sec:intro}
Visual counterfeits are increasingly causing an existential conundrum in mainstream media \cite{doi:10.1177/10776990211035395sage,schick_2021,cbs_news_deepfake_july_2022,foley_2022,aqil_haziq_mahmud,cbs_news_2021,harrison_2021,hao_heaven_2020,simonite_2020}.
With rapid improvements in CNN-based generative modelling 
\cite{NIPS2014_5ca3e9b1,karras2021alias,NEURIPS2020_8d30aa96,zhao2020diffaugment,NEURIPS2019_5f8e2fa1,Choi_2020_CVPR,brock2018large,park2019semantic,zhu2017unpaired,lee2021infomax,arjovsky2017wasserstein,Tran2018DistGANAI,zhao2022closer,NEURIPS2019_d04cb95b,tran2021data,lim2018doping,10.1145/3528223.3530065_rewriting_gan_geometry,Koh_2021_ICCV},\newline
detection of such counterfeits is increasingly becoming challenging and critical.
Nevertheless, a recent class of forensic detectors known as \textit{universal detectors} are able to surprisingly spot counterfeits regardless of generator architectures, loss functions, datasets and resolutions without any extensive adaptation \cite{Wang_2020_CVPR}.
i.e.: Publicly released ResNet-50 \cite{He_2016_CVPR} universal detector by Wang \etal \cite{Wang_2020_CVPR} trained only on ProGAN \cite{karras2018progressive} counterfeits, surprisingly generalizes well to detect counterfeits from unseen GANs including StyleGAN2 \cite{Karras_2020_CVPR}, StyleGAN \cite{Karras_2019_CVPR}, BigGAN \cite{brock2018large}, CycleGAN \cite{zhu2017unpaired}, StarGAN \cite{choi2018stargan} and GauGAN \cite{park2019semantic}.
This intriguing cross-model forensic transfer property suggests the existence of \textit{transferable forensic features (T-FF)} in universal detectors.

\subsection{Transferable Forensic Features (T-FF) in Universal Detectors}
This work is motivated by a profound and challenging thesis statement:
\textit{What transferable forensic features (T-FF) are used by universal detectors for counterfeit detection?}
A more elemental representation of this thesis statement would be: given an image of a real car and a high fidelity synthetic car generated by an unseen GAN (i.e.: StyleGAN2 \cite{Karras_2020_CVPR}), what \textit{T-FF} are used by the universal detector, such that it detects the synthetic car as counterfeit accurately? 
Though Wang \etal \cite{Wang_2020_CVPR} hypothesize that universal detectors may learn low-level CNN artifacts for detection, no qualitative / quantitative evidence is available in contemporary literature to understand \textit{T-FF} in universal detectors. 
\textit{Our work takes the first step towards discovering and understanding \textit{T-FF} in universal detectors for counterfeit detection.}
A foundational understanding on \textit{T-FF} and their properties are of paramount importance to both image forensics research and image synthesis research.
Understanding \textit{T-FF} will allow to build robust forensic detectors and to devise techniques to improve image synthesis methods to avoid generation of forensic footprints.

\subsection{Our contributions}

Our work conducts the \textit{first analytical study to discover and understand \textit{T-FF} in universal detectors for counterfeit detection.}
We begin our study by comprehensively demonstrating that input-space attribution -- using 2 popular algorithms namely Guided-GradCAM \cite{selvaraju2017grad} and LRP \cite{bach2015pixel} -- of universal detector decisions are 
not informative
to discover \textit{T-FF}.
Next, we study the forensic feature space of universal detectors to discover \textit{T-FF}.  
But investigating the feature space is an extremely daunting task due to the sheer amount of feature maps present. 
i.e.: ResNet-50 \cite{He_2016_CVPR} architecture contains approximately 27K feature maps.
To tackle this challenging task, \textit{we propose a novel \textit{forensic feature relevance statistic (FF-RS),}
to quantify and discover \textit{T-FF} in universal detectors.}
Our proposed FF-RS ($\omega$) 
is a scalar which 
quantifies the ratio between positive forensic relevance of the feature map and the total unsigned 
relevance of the entire layer that contains the particular feature map. 
Using our proposed
FF-RS
($\omega$), we successfully discover \textit{T-FF} in the publicly released ResNet-50 universal detector \cite{Wang_2020_CVPR}. 

Next, to understand the discovered \textit{T-FF}, \textit{we introduce a novel  pixel-wise  explanation method based on maximum spatial Layer-wise Relevance Propagation response (LRP-max)}. 
Particularly we visualize the pixel-wise explanations of each discovered \textit{T-FF} in universal detectors independently using 
LRP-max
visualization method.
Large-scale study on visual interpretability of \textit{T-FF} reveal that color information is critical for cross-model forensic transfer.  
Further large-scale quantitative investigations using median counterfeits probability analysis and statistical tests on maximum spatial activation distributions based on color ablation show that $color$ is a critical \textit{T-FF} in universal detectors.
Our findings are intriguing and new to the research community,
as many contemporary image forensics works focus on  frequency discrepancies between real and counterfeit images \cite{Durall_2020_CVPR,dzanic2020fourier,GAN_Artifacts,Chandrasegaran_2021_CVPR,schwarz2021frequency,khayatkhoei2020spatial}.
In summary, our contributions are as follows:
\begin{itemize}
    \item We propose a novel \textit{forensic feature relevance statistic (FF-RS) }to quantify and discover \textit{transferable forensic features (T-FF)} in universal detectors for counterfeit detection.
    
    \item We qualitatively -- using our proposed LRP-max visualization for feature map activations -- and quantitatively -- using median counterfeits probability analysis and statistical tests on maximum spatial activation distributions based on color ablation -- show that \textit{color} is a critical \textit{transferable forensic feature (T-FF)} in universal detectors for counterfeit detection.
\end{itemize}

\section{Related Work}
\label{sec:related_work}

\textbf{Counterfeit detection.}
Recent works have studied counterfeit detection both
in the RGB domain
\cite{rossler2019faceforensics++,gan_social_networks,cozzolino2018forensictransfer,GAN_Artifacts,nataraj2019detecting,wang2021fakespotter,Wang_2020_CVPR}
and 
frequency domain
\cite{dzanic2020fourier,Durall_2020_CVPR,Chandrasegaran_2021_CVPR,frank2020leveraging,luo2021generalizing}.
Particularly, notable number of works have proposed to use hand-crafted features for counterfeit detection \cite{dzanic2020fourier,Durall_2020_CVPR,Chandrasegaran_2021_CVPR,nataraj2019detecting}.
Using simple experiments, Mccloskey \etal \cite{mccloskey2019detecting} showed that detection based on the frequency of over-exposed pixels can provide good discrimination between real images and counterfeits. 
Li \etal observed disparities between GAN images and real images in the residual domain of the chrominance color components \cite{li2020identification}.
Some recent works have also proposed methods to detect and attribute counterfeits to the generating architectures
\cite{yu2019attributing,marra2019gans}.
Anomaly detection techniques leveraging on pre-trained face recognition models have also been proposed \cite{wang2021fakespotter}.\\

\noindent
\textbf{Cross-model forensic transfer.}
Most counterfeit detection works do not focus on cross-model forensic transfer.
Among the works that study forensic transfer, Cozzolino \etal \cite{cozzolino2018forensictransfer} and Zhang \etal \cite{GAN_Artifacts} observed that counterfeit detectors generalized poorly during cross-model forensic transfer.
In order to solve poor forensic transfer performance, Cozzolino \etal \cite{cozzolino2018forensictransfer} proposed an autoencoder based adaptation framework to improve cross-model forensic transfer.
The work by Wang \etal \cite{Wang_2020_CVPR} was the first work to show that counterfeit detectors -- universal detectors -- can generalize well during cross-model forensic transfer without any re-training / fine-tuning / adaptation on the target samples suggesting the possible existence of \textit{transferable forensic features}.
Furthermore, Chai \etal \cite{chai2020makes} showed that patch-based detectors with limited receptive fields
often perform better at detecting unseen counterfeits compared to full-image based detectors.\\

\noindent
\textbf{Interpretability methods.}
A number of interpretability methods in machine learning aim to summarize the relations which a model has learnt as a whole, such as PCA and t-SNE \cite{kpearsonpca,JMLR:v9:vandermaaten08a}, or to explain single decisions of a neural network. The latter may follow  different lines of questioning, such as identifying similar training samples in k-NN and prototype CNNs \cite{lloydknn,chenrudin}, finding modified samples such as pertinent negatives \cite{dhurandhar2018explanations}, or model-based uncertainty estimates \cite{pmlr-v48-gal16}.
One class of algorithms aims at computing input space attributions. This includes Shapley values \cite{strumbelj,lundberglee,chen2018lshapley} suitable for tabular data types, and methods for data types for which dropping a feature is not well defined, relying on modified gradients such as Guided Backprop \cite{DBLP:journals/corr/SpringenbergDBR14}, Layer-wise Relevance Propagation (LRP) \cite{bach2015pixel}, Guided-GradCAM \cite{selvaraju2017grad}, Full-Grad \cite{srinivas2019full}, and class-attention-mapping inspired research \cite{Desai2020AblationCAMVE,wang2020score,jiang2021layercam,fu2020axiom,muhammad2020eigen}.
Bau \etal proposed
frameworks for interpreting representations at the feature map level for classifiers \cite{bau2017network} and GANs \cite{bau2018gan}.
\vspace{-0.3cm}

\begin{figure}[!t]
\centering
\begin{tabular}{c @{\hskip 0.03in} 
                c @{\hskip 0.0in} c @{\hskip 0.02in}
                c @{\hskip 0.0in} c @{\hskip 0.02in}
                c @{\hskip 0.0in} c @{\hskip 0.02in} 
                c @{\hskip 0.0in} c @{\hskip 0.02in} 
                c @{\hskip 0.0in} c @{\hskip 0.0in}}
    
    \multicolumn{1}{c}{} &
    \multicolumn{2}{p{2cm}}{\tiny \quad \quad ProGAN \cite{karras2018progressive}} &
    \multicolumn{2}{p{2.2cm}}{\tiny \quad \quad StyleGAN2 \cite{Karras_2020_CVPR}} &
    \multicolumn{2}{p{2.2cm}}{\tiny \quad \quad StyleGAN \cite{Karras_2019_CVPR}} &
    \multicolumn{2}{p{2.2cm}}{\tiny \quad \quad  BigGAN \cite{brock2018large}} &
    \multicolumn{2}{c}{\tiny CycleGAN \cite{zhu2017unpaired}}
     \\
     
    \begin{turn}{90} 
    \textbf{\tiny Image}
    \end{turn} &
    
    \multicolumn{10}{c}{\includegraphics[width=0.92\linewidth]{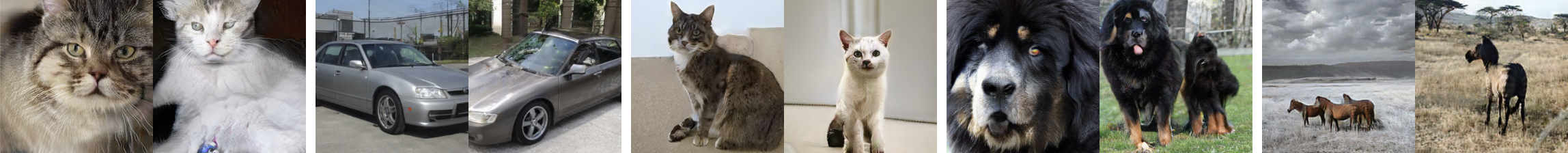}}

    \\
    
    \toprule
    \multicolumn{11}{c}{\tiny Pixel-wise explanations  of universal detector decisions \cite{Wang_2020_CVPR} using Guided-GradCAM (GGC) \cite{selvaraju2017grad} and LRP \cite{bach2015pixel} }\\
    
    \begin{turn}{90} 
    \textbf{\tiny GGC \cite{selvaraju2017grad}}
    \end{turn} &
    
    \multicolumn{10}{c}{\includegraphics[width=0.92\linewidth]{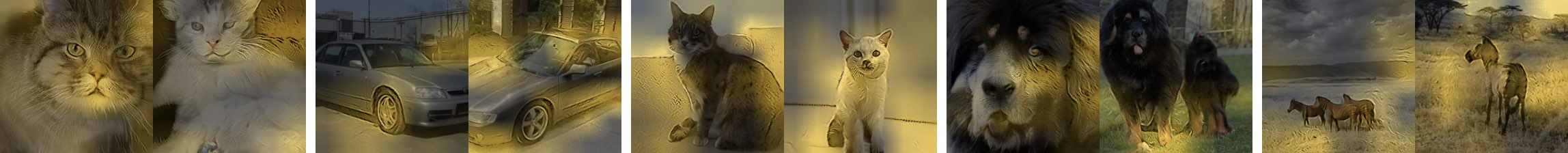}}
    
    \\
    
     \begin{turn}{90} 
    \textbf{\tiny LRP \cite{bach2015pixel}}
    \end{turn} &
    
    \multicolumn{10}{c}{\includegraphics[width=0.92\linewidth]{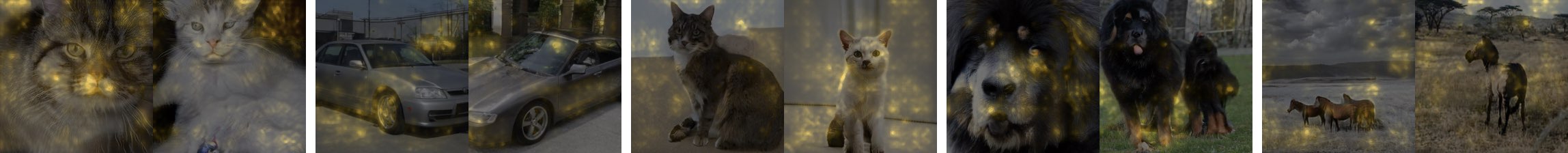}}
      
    \\
    
    \toprule
    \multicolumn{11}{c}{\tiny Pixel-wise explanations of ImageNet classifier decisions using Guided-GradCAM (GGC) \cite{selvaraju2017grad} and LRP \cite{bach2015pixel} }\\
    
    \begin{turn}{90} 
    \textbf{\tiny GGC \cite{selvaraju2017grad}}
    \end{turn} &

    \multicolumn{10}{c}{\includegraphics[width=0.92\linewidth]{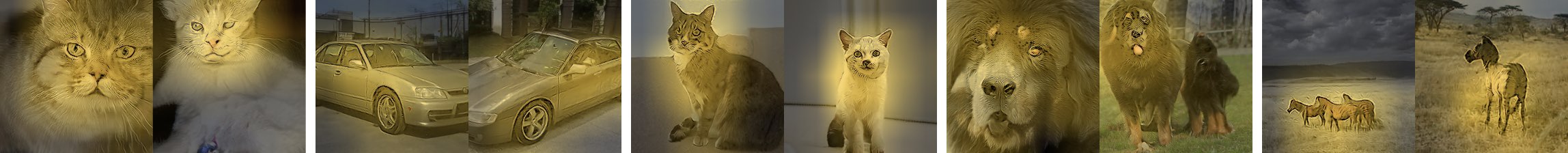}}
   
    \\
    
    \begin{turn}{90} 
    \textbf{\tiny LRP \cite{bach2015pixel}}
    \end{turn} &

    \multicolumn{10}{c}{\includegraphics[width=0.92\linewidth]{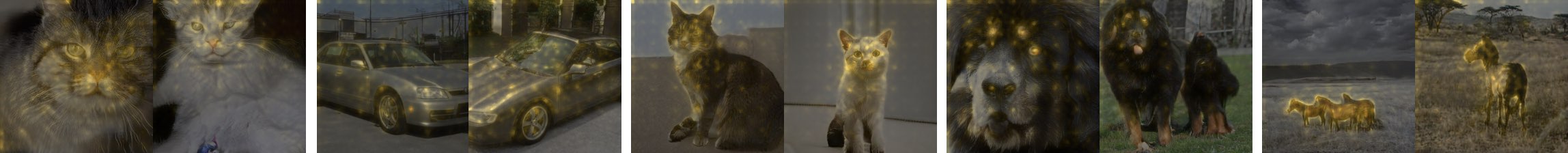}}
    
    \\

\end{tabular}
\vspace{-0.3cm}
\caption{
Pixel-wise explanations of universal detector decisions are
not informative
to discover \textit{T-FF}:
We show pixel-wise explanations using Guided-GradCAM (GGC) (row 2) \cite{selvaraju2017grad} and LRP (row 3) \cite{bach2015pixel} for the 
ResNet-50 universal detector 
\cite{Wang_2020_CVPR} for ProGAN \cite{karras2018progressive}, CycleGAN \cite{zhu2017unpaired}, StarGAN \cite{choi2018stargan}, BigGAN \cite{brock2018large} and StyleGAN2 \cite{Karras_2020_CVPR}.
The universal detector predicts probability $p>=95\%$ for all counterfeit images shown above. 
All these counterfeits are obtained from ForenSynths dataset 
\cite{Wang_2020_CVPR}.
For LRP \cite{bach2015pixel}, we only show positive relevances.
We also show pixel-wise explanations of ImageNet classifier decisions for the exact counterfeits using GGC (row 4) and LRP (row 5). This is shown as a control experiment to emphasize the significance of our observations.
As one can clearly observe, pixel-wise explanations of universal detector decisions are 
not informative
to discover \textit{T-FF} (rows 2, 3) as the explanations appear to be random and not reveal any meaningful visual features used for counterfeit detection.
Particularly, it remains unknown as to why the universal detector outputs high detection probability ($p>=95\%$) for these counterfeits.
On the other hand, pixel-wise explanations of ImageNet classifier decisions produce meaningful results. 
i.e.: GGC (row 4) and LRP (row 5) explanation results for cat samples (columns 1, 2, 5, 6) show that ImageNet uses features such as eyes and whiskers to classify cats.
This shows that interpretability techniques such as GGC and LRP are 
not informative
to discover \textit{T-FF} in universal detectors.
In other words, we are unable to discover any forensic footprints based on pixel-wise explanations of universal detectors.
More examples in Supplementary
\ref{sec_supp:pixel-wise_explanations}.
\vspace{-0.7cm}
}
\label{fig_main:pixel_wise_explanations}
\end{figure}

\section{Dataset / Metrics}
We use the ForenSynths dataset proposed by Wang \etal \cite{Wang_2020_CVPR}. ForenSynths is the largest counterfeit benchmark dataset containing CNN-generated images from multiple generator architectures, datasets, loss functions and resolutions. 
In addition to ProGAN \cite{karras2018progressive}, we select 6 candidate GANs to comprehensively study cross-model forensic transfer in this work namely,
StyleGAN2 \cite{Karras_2020_CVPR}, StyleGAN \cite{Karras_2019_CVPR}, BigGAN \cite{brock2018large}, CycleGAN \cite{zhu2017unpaired}, StarGAN \cite{choi2018stargan} and GauGAN \cite{park2019semantic}.
Following Wang \etal \cite{Wang_2020_CVPR}, we use AP (Average Precision) to measure cross-model forensic transfer of universal detectors. 
Particularly, we also show the accuracies for real and counterfeit images as we intend to understand counterfeit detection.
For detector calibration, we follow \cite{Wang_2020_CVPR} and use the oracle threshold obtained using geometric mean of sensitivity and specificity. 
\vspace{-0.3cm}

\section{Discovering Transferable Forensic Features (T-FF)}
\vspace{-0.2cm}
\subsection{Input-space attribution methods}
\label{sec_main:input_space_attribution_methods}
Interpretable machine learning algorithms are useful exploratory tools to visualize neural networks' decisions by input-space attribution 
\cite{lrp,selvaraju2017grad,srinivas2019full,Desai2020AblationCAMVE,wang2020score,jiang2021layercam,fu2020axiom,muhammad2020eigen}.
We start from the following question: \textit{Are interpretability methods suitable to discover 
\textit{T-FF}
in universal detectors? }

We use 2 popular interpretability methods namely Guided-GradCAM \cite{selvaraju2017grad}
and LRP \cite{bach2015pixel}
to analyse the pixel-wise explanations of universal detector decisions. These methods were chosen due to their relatively low amount of gradient shattering noise \cite{DBLP:conf/icml/BalduzziFLLMM17}.
We show the pixel-wise explanation results of
ResNet-50 universal detector \cite{Wang_2020_CVPR} decisions for ProGAN \cite{karras2018progressive} and 4 GANs not used for training -- CycleGAN \cite{zhu2017unpaired}, StarGAN \cite{choi2018stargan}, BigGAN \cite{brock2018large} and StyleGAN2 \cite{Karras_2020_CVPR}-- in Fig.~\ref{fig_main:pixel_wise_explanations}.
As one can observe in Fig.~\ref{fig_main:pixel_wise_explanations}, pixel-wise explanations of universal detector decisions are not informative to discover 
\textit{T-FF} due to their focus on spatial localization.
Particularly, we are unable to discover any forensic footprints based on pixel-wise explanations of universal detector decisions. 
This is consistently seen across both Guided-GradCAM \cite{selvaraju2017grad}
and LRP \cite{bach2015pixel} methods.
We remark that these observations do not indicate failure modes of Guided-GradCAM \cite{selvaraju2017grad} or LRP \cite{bach2015pixel} methods, but rather suggest that universal detectors are learning more complex \textit{T-FF} that are not easily human-parsable.

\subsection{Forensic Feature Space}
\label{sec_main:forensic_feature_space}

Given that input-space attribution methods are 
not informative
to discover 
\textit{T-FF}, 
we study the feature space to discover \textit{T-FF} in universal detectors for counterfeit detection.
Particularly, we ask the question: which feature maps in universal detectors are responsible for cross-model forensic transfer?
This is a challenging problem as it requires quantifying the importance of every feature map in universal detectors for counterfeit detection. The ResNet-50 universal detector \cite{Wang_2020_CVPR} consists of approximately 27K intermediate feature maps.
\\

\begin{table}[t]
\caption{
Sensitivity assessments using feature map dropout showing that our proposed \textit{FF-RS ($\omega$)} successfully quantifies and discovers \textit{T-FF}:
We show the results for the publicly released ResNet-50 universal detector 
\cite{Wang_2020_CVPR} 
(top) and 
our own version of EfficientNet-B0 \cite{tan2019efficientnet}
universal detector (bottom) following the exact training and test strategy proposed in
\cite{Wang_2020_CVPR}. 
We show the AP, real and GAN image detection accuracies for baseline \cite{Wang_2020_CVPR}, top-$k$, random-$k$ and low-$k$ forensic feature dropout. 
The random-$k$ experiments are repeated 5 times and
average results are reported.
Feature map dropout is performed by suppressing (zeroing out) the resulting activations of target feature maps (i.e.: top-$k$). 
We can clearly observe that feature map dropout of top-$k$ corresponding to \textit{T-FF} results in substantial drop in AP and GAN detection accuracies across ProGAN and all 6 unseen GANs 
\cite{Karras_2020_CVPR,Karras_2019_CVPR,brock2018large,zhu2017unpaired,choi2018stargan,park2019semantic}
compared to baseline, random-$k$ and low-$k$ results.
This is consistently seen in both ResNet-50 and EfficientNet-B0 universal detectors.
This shows that our proposed \textit{FF-RS ($\omega$)} can successfully quantify and discover the \textit{T-FF} in universal detectors. 
$k \approx 0.5 \%$ of total feature maps. 
More details included in Supplementary
\ref{sec_supp:k_hyper-parameter}.
}
\begin{center}
\begin{adjustbox}{width=1.0\columnwidth,center}
\begin{tabular}{c|ccc|ccc|ccc|ccc|ccc|ccc|ccc}
\multicolumn{22}{c}{\bf \Large ResNet-50} \\ \toprule
&\multicolumn{3}{c}{\textbf{ProGAN} \cite{karras2018progressive}} 
&\multicolumn{3}{c}{\textbf{StyleGAN2} \cite{Karras_2020_CVPR}} 
&\multicolumn{3}{c}{\textbf{StyleGAN} \cite{Karras_2019_CVPR}} &\multicolumn{3}{c}{\textbf{BigGAN} \cite{brock2018large}} &\multicolumn{3}{c}{\textbf{CycleGAN} \cite{zhu2017unpaired} } &\multicolumn{3}{c}{\textbf{StarGAN} \cite{choi2018stargan}} &\multicolumn{3}{c}{\textbf{GauGAN} \cite{park2019semantic}} \\

\cmidrule{2-22}

\textbf{$k=114$} &\textbf{AP} &\textbf{Real} &\textbf{GAN} &\textbf{AP} &\textbf{Real} &\textbf{GAN} &\textbf{AP} &\textbf{Real} &\textbf{GAN} &\textbf{AP} &\textbf{Real} &\textbf{GAN} &\textbf{AP} &\textbf{Real} &\textbf{GAN} &\textbf{AP} &\textbf{Real} &\textbf{GAN} &\textbf{AP} &\textbf{Real} &\textbf{GAN} \\
\midrule

baseline \cite{Wang_2020_CVPR} &100. &100.0 &100. &99.1 &95.5 &95.0 &99.3 &96.0 &95.6 &90.4 &83.9 &85.1 &97.9 &93.4 &92.6 &97.5 &94.0 &89.3 &98.8 &93.9 &96.4 \\
\textbf{top-k} &\textbf{69.8} &\textbf{99.4} &\textbf{3.2} &\textbf{55.3} &\textbf{89.4} &\textbf{11.3} &\textbf{56.6} &\textbf{90.6} &\textbf{13.7} &\textbf{55.4} &\textbf{86.3} &\textbf{18.3} &\textbf{61.2} &\textbf{91.4} &\textbf{17.4} &\textbf{72.6} &\textbf{89.4} &\textbf{35.9} &\textbf{71.0} &\textbf{95.0} &\textbf{18.8} \\
random-k &100. &99.9 &96.1 &98.6 &89.4 &96.9 &98.7 &91.4 &96.1 &88.0 &79.4 &85.0 &96.6 &81.0 &96.2 &97.0 &88.0 &91.7 &98.7 &91.9 &97.1 \\
low-k &100. &100. &100. &99.1 &95.6 &95.0 &99.3 &96.0 &95.6 &90.4 &83.9 &85.1 &97.9 &93.4 &92.6 &97.5 &94.0 &89.3 &98.8 &93.9 &96.4 \\
\bottomrule
\end{tabular}
\end{adjustbox}
\end{center}

\begin{center}
\begin{adjustbox}{width=1.0\columnwidth,center}
\begin{tabular}{c|ccc|ccc|ccc|ccc|ccc|ccc|ccc}
\multicolumn{22}{c}{\bf \Large EfficientNet-B0} \\ \toprule
&\multicolumn{3}{c}{\textbf{ProGAN} \cite{karras2018progressive}} 
&\multicolumn{3}{c}{\textbf{StyleGAN2} \cite{Karras_2020_CVPR}} 
&\multicolumn{3}{c}{\textbf{StyleGAN} \cite{Karras_2019_CVPR}} &\multicolumn{3}{c}{\textbf{BigGAN} \cite{brock2018large}} &\multicolumn{3}{c}{\textbf{CycleGAN} \cite{zhu2017unpaired} } &\multicolumn{3}{c}{\textbf{StarGAN} \cite{choi2018stargan}} &\multicolumn{3}{c}{\textbf{GauGAN} \cite{park2019semantic}} \\

\cmidrule{2-22}

\textbf{$k=27$} &\textbf{AP} &\textbf{Real} &\textbf{GAN} &\textbf{AP} &\textbf{Real} &\textbf{GAN} &\textbf{AP} &\textbf{Real} &\textbf{GAN} &\textbf{AP} &\textbf{Real} &\textbf{GAN} &\textbf{AP} &\textbf{Real} &\textbf{GAN} &\textbf{AP} &\textbf{Real} &\textbf{GAN} &\textbf{AP} &\textbf{Real} &\textbf{GAN} \\
\midrule

baseline \cite{Wang_2020_CVPR} &100. &100. &100. &95.9 &95.2 &85.4 &99.0 &96.1 &94.3 &84.4 &79.7 &75.9 &97.3 &89.6 &93.0 &96.0 &92.8 &85.5 &98.3 &94.1 &94.4 \\

\textbf{top-k} &\textbf{50.0} &\textbf{100.} &\textbf{0.0} &\textbf{54.5} &\textbf{94.3} &\textbf{7.0} &\textbf{52.1} &\textbf{97.3} &\textbf{2.6} &\textbf{53.5} &\textbf{97.4} &\textbf{3.8} &\textbf{47.5} &\textbf{100.} &\textbf{0.0} &\textbf{50.0} &\textbf{100.} &\textbf{0.0} &\textbf{46.2} &\textbf{100.} &\textbf{0.0} \\

random-k &100. &99.9 &100. &96.5 &91.9 &89.8 &99.2 &91.2 &97.5 &84.5 &59.4 &89.1 &96.9 &82.6 &95.8 &96.7 &82.5 &93.3 &98.1 &87.8 &96.2 \\
low-k &100. &100. &100. &95.3 &88.7 &88.3 &98.9 &90.8 &96.1 &83.5 &70.8 &80.8 &96.6 &85.2 &94.1 &95.4 &91.0 &85.4 &98.1 &91.2 &96.4 \\
\bottomrule

\end{tabular}
\end{adjustbox}
\end{center}

\label{table_main:sensitivity_assessments}
\vspace{-0.9cm}
\end{table}

\noindent
\textbf{Forensic feature relevance statistic (FF-RS).} 
We propose a novel
\textit{FF-RS ($\omega$)}
to quantify the relevance of every feature map in universal detectors for counterfeit detection.
Specifically, for feature map at layer $l$ and channel $c$, 
$\omega(l_c)$ computes the forensic relevance of this feature map for counterfeit detection.
We describe the important design considerations and intuitions behind our proposed 
\textit{FF-RS ($\omega$)}
below and include the pseudocode in Algorithm \ref{alg:omega_algorithm}:
\begin{itemize}
    \item We postulate the existence of a set of feature maps in universal detectors that are responsible for cross-model forensic transfer. In particular, we hypothesize that there is a set of \textit{common transferable forensic feature maps} that mostly gets activated when passing counterfeits from ProGAN \cite{karras2018progressive} and unseen GANs.
    
    \item Our proposed 
    \textit{FF-RS ($\omega$)}
    is a scalar that quantifies the forensic relevance of every feature map. In particular, $\omega$ for a feature map quantifies the ratio between positive forensic relevance of the feature map and the total unsigned forensic relevance of the entire layer that contains the particular feature map. This is shown in Line 8 in Algorithm \ref{alg:omega_algorithm}. 
    For the numerator we are only interested in positive relevance, therefore use a max operation to select only positive relevance (identical to a ReLU operation).
    
    \item The relevance scores are calculated using LRP \cite{bach2015pixel} (More details on LRP \cite{bach2015pixel} in Supplementary \ref{sec_supp:lrp_base}). This is shown in Line 5 in Algorithm \ref{alg:omega_algorithm} where $r_i(l, c, h, w)$ is the estimated relevance of the feature map at layer $l$, channel $c$ at the spatial location $h, w$
    
    \item $\omega$ is calculated over large number of counterfeit images and is bounded between $[0, 1]$. i.e.: $\omega = 1$ indicates that the particular feature map is the most relevant forensic feature and $\omega = 0$ indicates vice versa.
    
    \item Finally we use $\omega$ to rank all the feature maps and identify the set of \textit{T-FF}. We refer to this set as top-k in our experiments.
    
\end{itemize}

\noindent
\textbf{Experiments : Sensitivity assessments of discovered T-FF using algorithm \ref{alg:omega_algorithm}.}
We perform rigorous sensitivity assessments using feature map dropout experiments to demonstrate that our proposed
\textit{FF-RS ($\omega$)}
is able to quantify and discover 
\textit{T-FF}.
Feature map dropout suppresses (zeroing out) the resulting activations of the target feature maps. 
Particularly, feature map dropout of \textit{T-FF} should satisfy the following sensitivity conditions:
\begin{enumerate}
    \item Significant reduction in overall AP across ProGAN \cite{karras2018progressive} and all unseen GANs
    \cite{Karras_2020_CVPR,Karras_2019_CVPR,brock2018large,zhu2017unpaired,choi2018stargan,park2019semantic}
    indicating poor cross-model forensic transfer.
    
    \item Significant reduction in GAN /counterfeit detection accuracies across  ProGAN \cite{karras2018progressive} and all unseen GANs
    \cite{Karras_2020_CVPR,Karras_2019_CVPR,brock2018large,zhu2017unpaired,choi2018stargan,park2019semantic} compared to real image detection accuracies as $\omega$ is calculated for counterfeits. 
\end{enumerate}

\noindent
\textbf{Test bed details. }
We use the ForenSynths test set \cite{Wang_2020_CVPR}.
$\omega$ is calculated using 1000 ProGAN \cite{karras2018progressive} counterfeits (validation set).
We use the following experiment codes:
\begin{itemize}
    \item top-$k$ : Set of T-FF discovered using \textit{FF-RS $(\omega)$}
    
    \item random-$k$ : Set of random feature maps used as a control experiment.
    
    \item low-$k$ : Set of low-ranked feature maps corresponding to extremely small values of $\omega$, i.e.: $\omega \approx 0$.
\end{itemize}
    
\noindent
\textbf{Results.}
We show the results in Table \ref{table_main:sensitivity_assessments} for ResNet-50 and EfficientNet-B0 universal detectors.
We clearly observe that feature map dropout of
top-$k$ features corresponding to \textit{T-FF} 
satisfies both sensitivity conditions above indicating that our proposed 
\textit{FF-RS ($\omega$)}
is able to quantify and discover \textit{transferable forensic features}.
We also observe that feature map dropout of low-$k$ (low-ranked) forensic features has little / no effect on cross-model forensic transfer which further adds merit to our proposed \textit{FF-RS ($\omega$)}.

\begin{algorithm*}[tb]
\caption{Calculate FF-RS ($\omega$) (Non-vectorized)}\label{alg:omega_algorithm}

\KwInput{\\
forensics detector $M$, \\
data $D=\{x\}_{i=1}^n$, $D$ is a large counterfeit dataset where $x_i$ indicates the $i^{th}$ counterfeit image.
}

\KwOutput{\\
$\omega(l_c)$ where $l, c$ indicates the layer and channel index of forensic feature maps. Every forensic feature map can be characterized by a unique set of $l, c$. 
}

$R \gets [\;]$\ \Comment*[r]{List to store feature map relevances}
Set $M$ to evaluation mode

\For{$i\;in\;\{0, 1, , ...., n\}$}{
    $f(x_i) \gets M(x_i)$ \Comment*[r]{logit output}
    
    $r_i \gets LRP(M, x_i, f(x_i))$\ \Comment*[r]{calculate LRP scores for counterfeits}
    
    \For{$l'\;in\;r_i.size(0)$}{
        \For{$c'\;in\;r_i.size(1)$}{
            ${r}_i(l', c', h, w) \gets \frac{max(0, {r}_i(l', c', h, w))} {\sum_{c,h,w}^{} \vert \vert {r}_i(l', c, h, w) \vert \vert } $\
            
            $R.append(r_i)$ \Comment*[r]{$r_i$.size():(layer, channel, height, width)}
        }
    }
}
$\omega(l_c) \gets \sum_{h,w}^{} \frac{1}{N} \sum_{i}^{n} R_i (l, c, h, w)$ \Comment*[r]{forensic feature relevance}


\Return $\omega(l_c)$ 
\end{algorithm*}

\section{Understanding Transferable Forensic Features (T-FF)}

\begin{figure}
\centering
\begin{tabular}{ccccccc}
    \multicolumn{1}{p{0.125\linewidth}}{\tiny \enskip ProGAN \cite{karras2018progressive}} &
    \multicolumn{1}{p{0.15\linewidth}}{\tiny  \enskip StyleGAN2 \cite{Karras_2020_CVPR}} &
    \multicolumn{1}{p{0.14\linewidth}}{\tiny StyleGAN \cite{Karras_2019_CVPR}} &
    \multicolumn{1}{p{0.125\linewidth}}{\tiny BigGAN \cite{brock2018large}} &
    \multicolumn{1}{p{0.132\linewidth}}{\tiny CycleGAN \cite{zhu2017unpaired}} &
    \multicolumn{1}{p{0.135\linewidth}}{\tiny \enskip StarGAN \cite{choi2018stargan}} &
    {\tiny GauGAN \cite{park2019semantic}} \\

    \multicolumn{7}{c}{\includegraphics[width=0.99\linewidth]{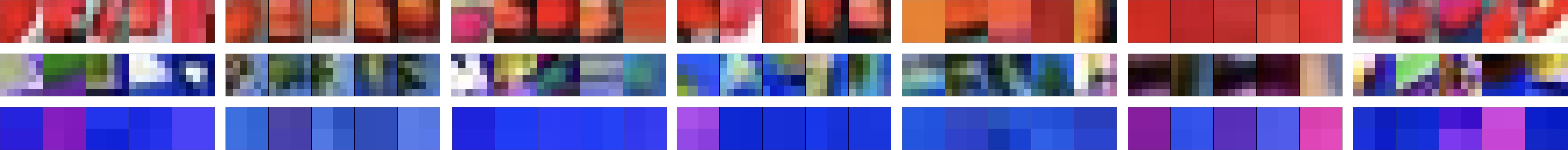}}
    
\end{tabular}
\vspace{-0.3cm}
\caption{
Color is a critical \textit{T-FF} in universal detectors:
Large-scale study on visual interpretability of \textit{T-FF} discovered through our proposed \textit{FF-RS ($\omega$)} reveal that color information is critical for cross-model forensic transfer.
Each row represents a color-based \textit{T-FF} and
we show the LRP-max response regions for ProGAN
and all 6 unseen GAN 
\cite{Karras_2020_CVPR,Karras_2019_CVPR,brock2018large,zhu2017unpaired,choi2018stargan,park2019semantic}
counterfeits 
for our own version of EfficientNet-B0 \cite{tan2019efficientnet} universal detector following the exact training / test strategy proposed
by Wang \etal 
\cite{Wang_2020_CVPR}.
This detector is trained with ProGAN
\cite{karras2018progressive} 
counterfeits \cite{Wang_2020_CVPR} and cross-model forensic transfer is evaluated on other unseen GANs.
All counterfeits are obtained from the ForenSynths dataset
\cite{Wang_2020_CVPR}.
The consistent color-conditional LRP-max response across all GANs for these \textit{T-FF} clearly indicate that \textit{color} is critical for cross-model forensic transfer in universal detectors.
More visualizations are included in Supplementary
\ref{sec_supp:color_tff}.
\vspace{-0.6cm}
}
\label{fig:lrp_patches_efb0}
\end{figure}

Given the successful discovery of \textit{T-FF} using our proposed \textit{FF-RS ($\omega$)}, 
in this section, 
we ask the following question: 
what counterfeit properties are detected by this set of \textit{T-FF}?
Though Wang \etal \cite{Wang_2020_CVPR} hypothesize that universal detectors may learn low-level CNN artifacts for cross-model forensic transfer, 
no 
evidence is available to understand as to what features in counterfeits are being detected during cross-model forensic transfer.

\subsection{LRP-max explanations of T-FF}
\label{sec_main:lrp_max}
We approach this problem from a visual interpretability perspective. 
In this section, we introduce a novel pixel-wise explanation method for feature map activations based on maximum spatial Layer-wise Relevance Propagation response (LRP-max). 
The idea behind LRP-max is to independently visualize which pixels in the input space correspond to maximum spatial relevance scores for each \textit{T-FF}. Particularly, instead of back-propagating using the detector logits, we back-propagate from the maximum spatial relevance neuron of each \textit{T-FF} independently.
LRP-max automatically extracts image regions for every T-FF and does not depend on external modules such as segmentation used in \cite{bau2017network,bau2018gan}.
The pseudocode is included in \ref{alg:lrp_max_algorithm}.

\textit{Color is a critical T-FF in universal detectors: }
LRP-max visualizations of \textit{T-FF} uncover the
unexpected
observation that a substantial amount of \textit{T-FF} exhibits color-conditional activations. 
We show the LRP-max regions for ProGAN \cite{karras2018progressive} and all unseen GANs
\cite{Karras_2020_CVPR,Karras_2019_CVPR,brock2018large,zhu2017unpaired,choi2018stargan,park2019semantic} for ResNet-50 and EfficientNet-B0 universal detectors in Fig. \ref{fig:lrp_patches_r50} and \ref{fig:lrp_patches_efb0} respectively.
As one can observe, the consistent color-conditional LRP-max response across all GANs for these T-FF clearly indicate that color is critical for cross-model forensic transfer in universal detectors. 
This is notably surprising and observed for the first time in transferable image forensics research. 
In the next section, we conduct quantitative studies to rigorously verify that color is a critical \textit{T-FF} in universal detectors.

\begin{algorithm*}
\caption{Obtain LRP-max pixel-wise explanations ( For a single feature map, for a single sample )}\label{alg:lrp_max_algorithm}

\KwInput{\\
forensics detector $M$, \\
counterfeit image $x$ where $x.size() = (3, x_{height}, x_{width})$, \\
forensic feature map $l,c$ where $l, c$ indicate layer and channel index respectively.
}

\KwOutput{\\
$\hat{E}_{l_c}(x)$ where E indicates the LRP-max pixel-wise explanations for sample $x$ corresponding to forensic feature map at layer index $l$ and channel index $c$. \\
Do note that $\hat{E}_{l_c}(x).size()$ is $(x_{height}, x_{width})$.\\
Every forensic feature map can be characterized by a unique set of $l, c$. 
}

$z_{l_c}(x) \gets LRP-FORWARD(M_{l_c}(x_i))$ \Comment*[r]{(h, w) relevance scores}

$h^{*}, w^{*} \gets argmax(z_{l_c}(x))$ \Comment*[r]{find index of max relevance}

$z_{l_c}^{max}(x) \gets z_{l_c}(x)[h^{*}, w^{*}]$ \Comment*[r]{LRP-max response neuron}

$E_{l_c}(x) \gets LRP-BACKWARD(z_{l_c}^{max}(x))$ \Comment*[r]{explain LRP-max neuron}

$\hat{E}_{l_c}(x) \gets \sum_{k=0}^{3}(E_{l_c}(x)(k, x_{height}, x_{width})$ \Comment*[r]{spatial LRP-max}

\Return $\hat{E}_{l_c}(x)$ 
\end{algorithm*}

\subsection{Color Ablation Studies}
\label{sec_main:color_ablation}
In this section, we conduct 2 quantitative studies to show that \textit{color} is a critical \textit{transferable forensic feature} in universal detectors. 
Our studies measure the sensitivity of universal detectors before and after color ablation.

\begin{algorithm*}[t]
\caption{Statistical test over maximum spatial activations for \textit{T-FF} (Non-vectorized)}\label{alg_main:median_test}

\KwInput{\\
forensics detector $M$, \\
data $D=\{x\}_{i=1}^n$, $D$ is a large counterfeit dataset where $x_i$ indicates the $i^{th}$ counterfeit image. \\
T-FF set $S$
}

\KwOutput{\\
$p(l_c)$ where $l, c$ indicates the layer and channel index of forensic feature maps. 
$p$ indicates $p$-value of the statistical test.

Every forensic feature map can be characterized by a unique set of $l, c$. 
}

Set $M$ to evaluation mode

\For{$l', c'\;in\;S$}{
    $A_{b} \gets [\;] $\ \Comment*[r]{store baseline counterfeits activations}
    
    $A_{g} \gets [\;] $\ \Comment*[r]{store grayscale counterfeits activations}
    
    \For{$i\;in\;\{0, 1, , ...., n\}$}{
        $a_b \gets GLOBAL\_MAXPOOL(M_{l_c}(x_i))$\ \Comment*[r]{baseline}
        
        $a_g \gets GLOBAL\_MAXPOOL(M_{l_c}(grayscale(x_i)))$\ \Comment*[r]{grayscale}
        
        $A_{b}.append(a_b)$
        
        $A_{g}.append(a_g)$
        
    }
    
    $p(l'_{c'}) \gets MEDIAN-TEST(A_b, A_g)$ \Comment*[r]{median test}
}

\Return $p(l_c)$ 
\end{algorithm*}

\begin{figure}[!t]
\centering
\begin{tabular}{ccccccc}
    {\tiny ProGAN \cite{karras2018progressive}} &
    {\tiny StyleGAN2 \cite{Karras_2020_CVPR}} &
    {\tiny StyleGAN \cite{Karras_2019_CVPR}} &
    {\tiny BigGAN \cite{brock2018large}} &
    {\tiny CycleGAN \cite{zhu2017unpaired}} &
    {\tiny StarGAN \cite{choi2018stargan}} &
    {\tiny GauGAN \cite{park2019semantic}} \\

    \includegraphics[width=0.13\linewidth]{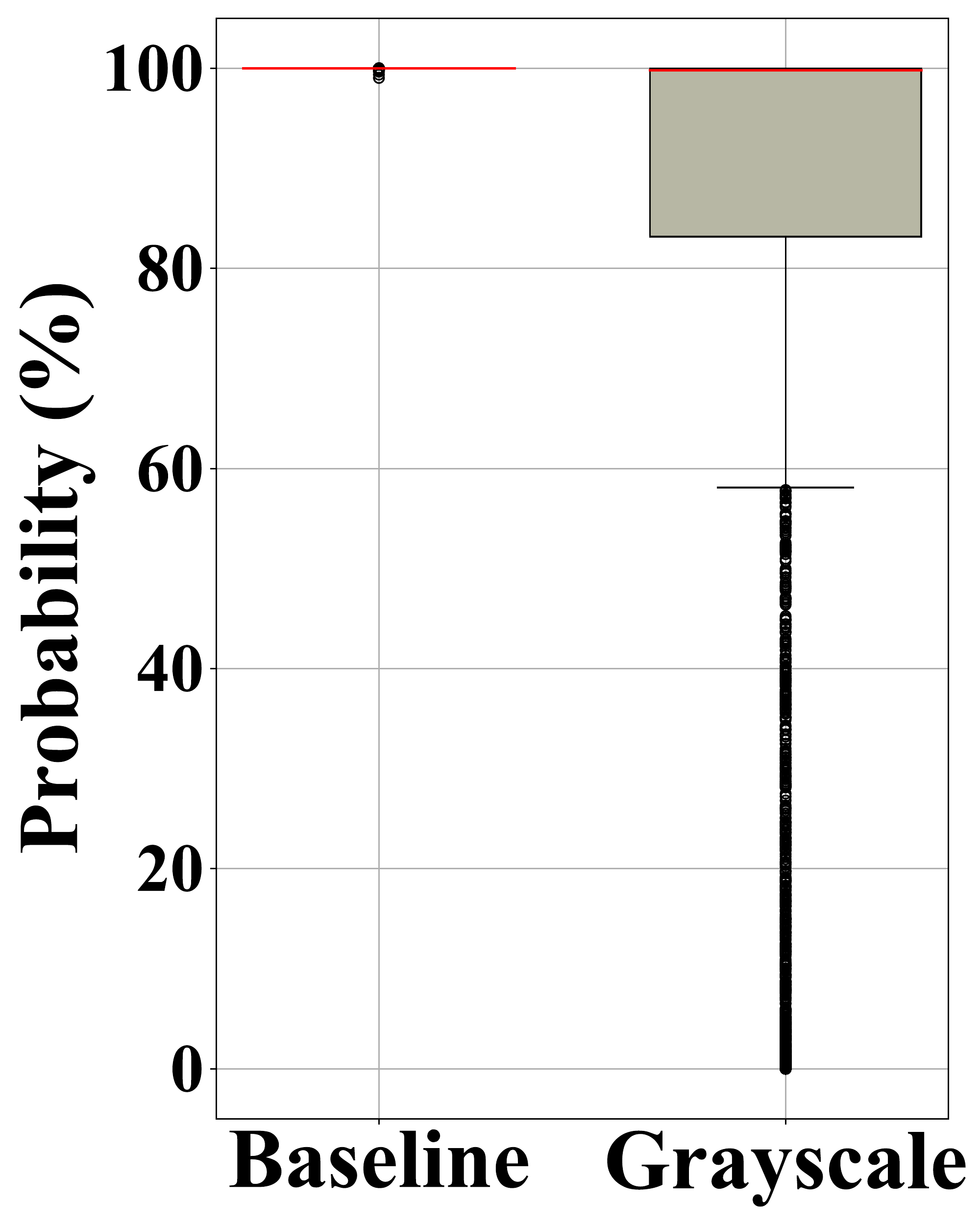} &
    \includegraphics[width=0.13\linewidth]{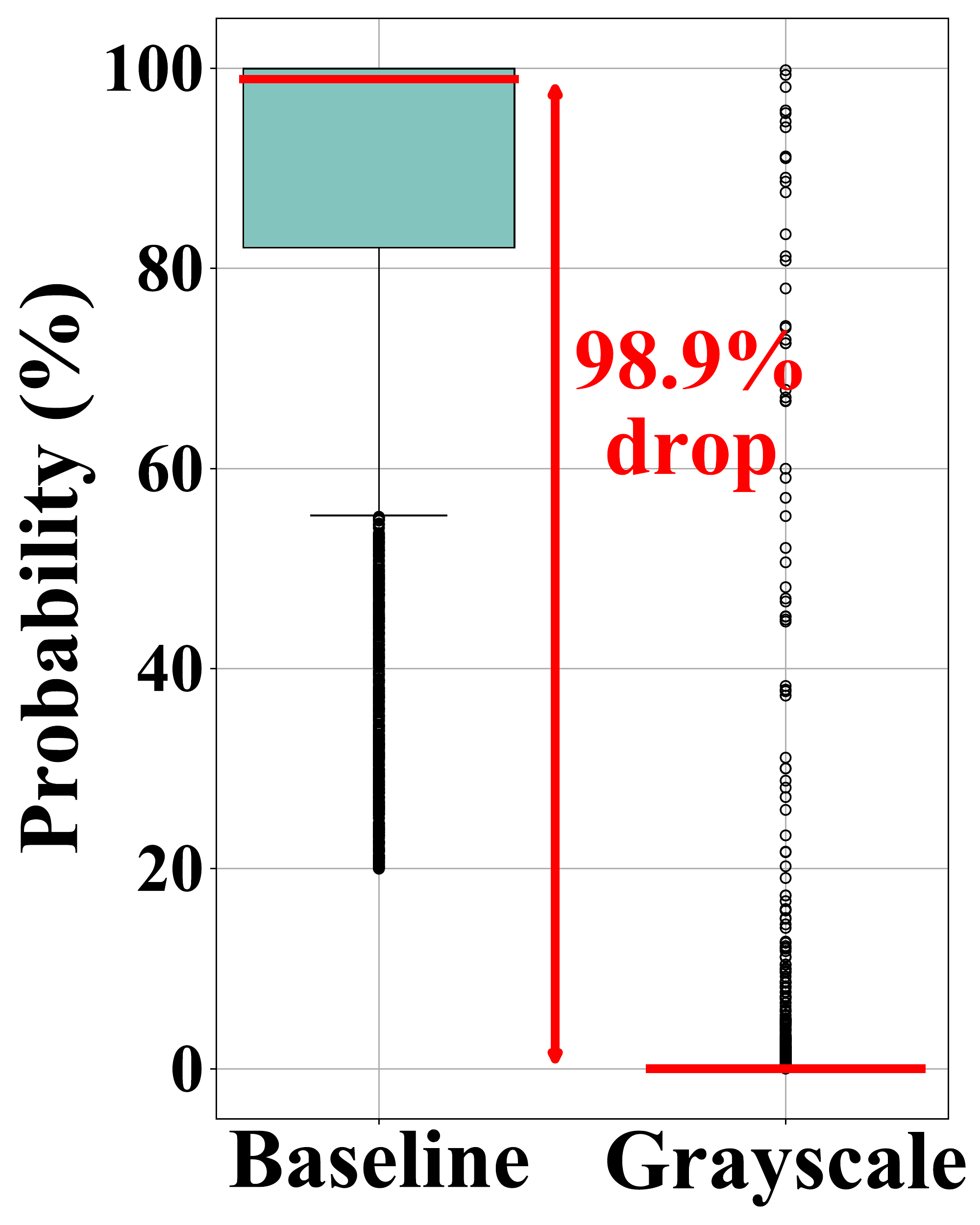} &
    \includegraphics[width=0.13\linewidth]{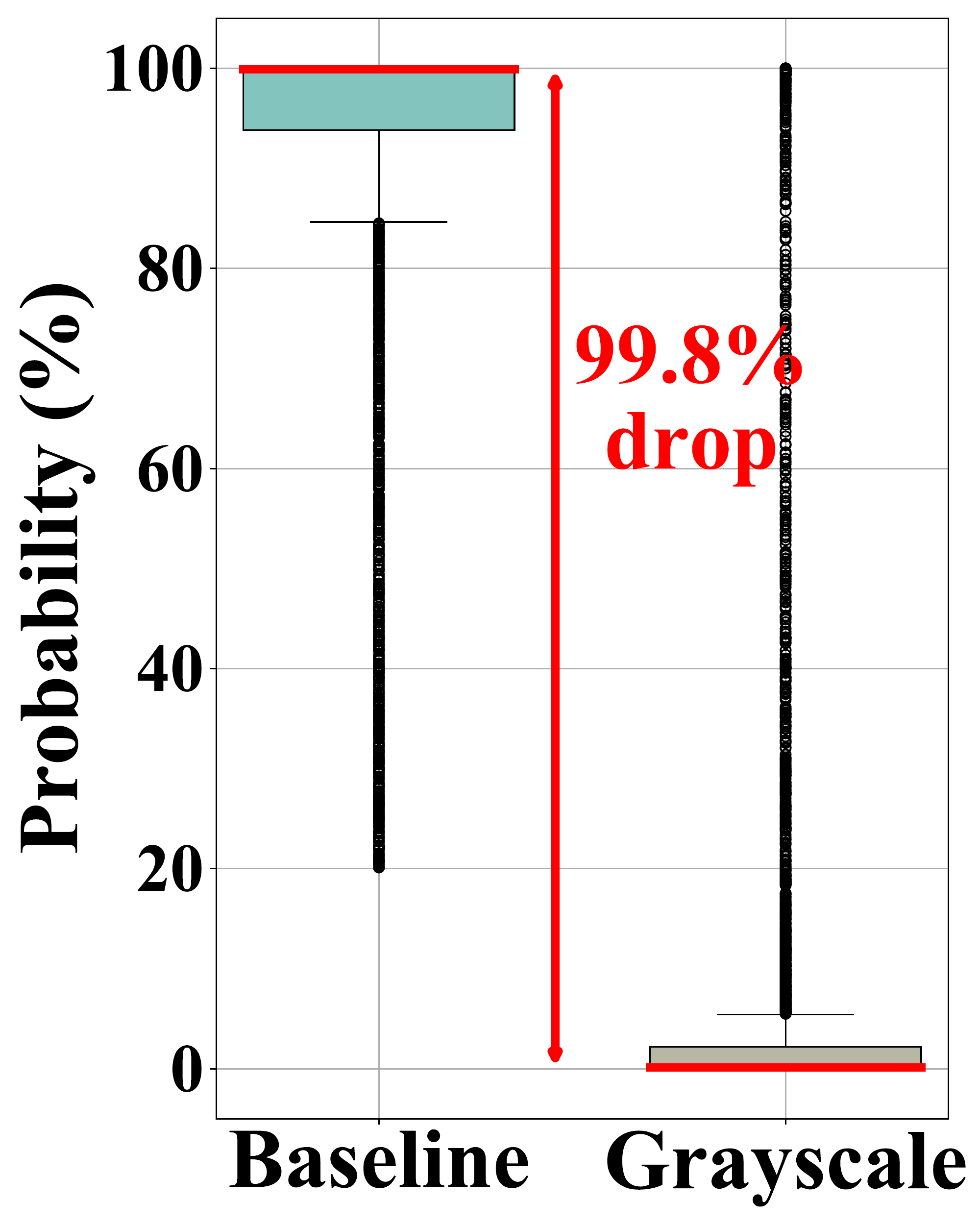} &
    \includegraphics[width=0.13\linewidth]{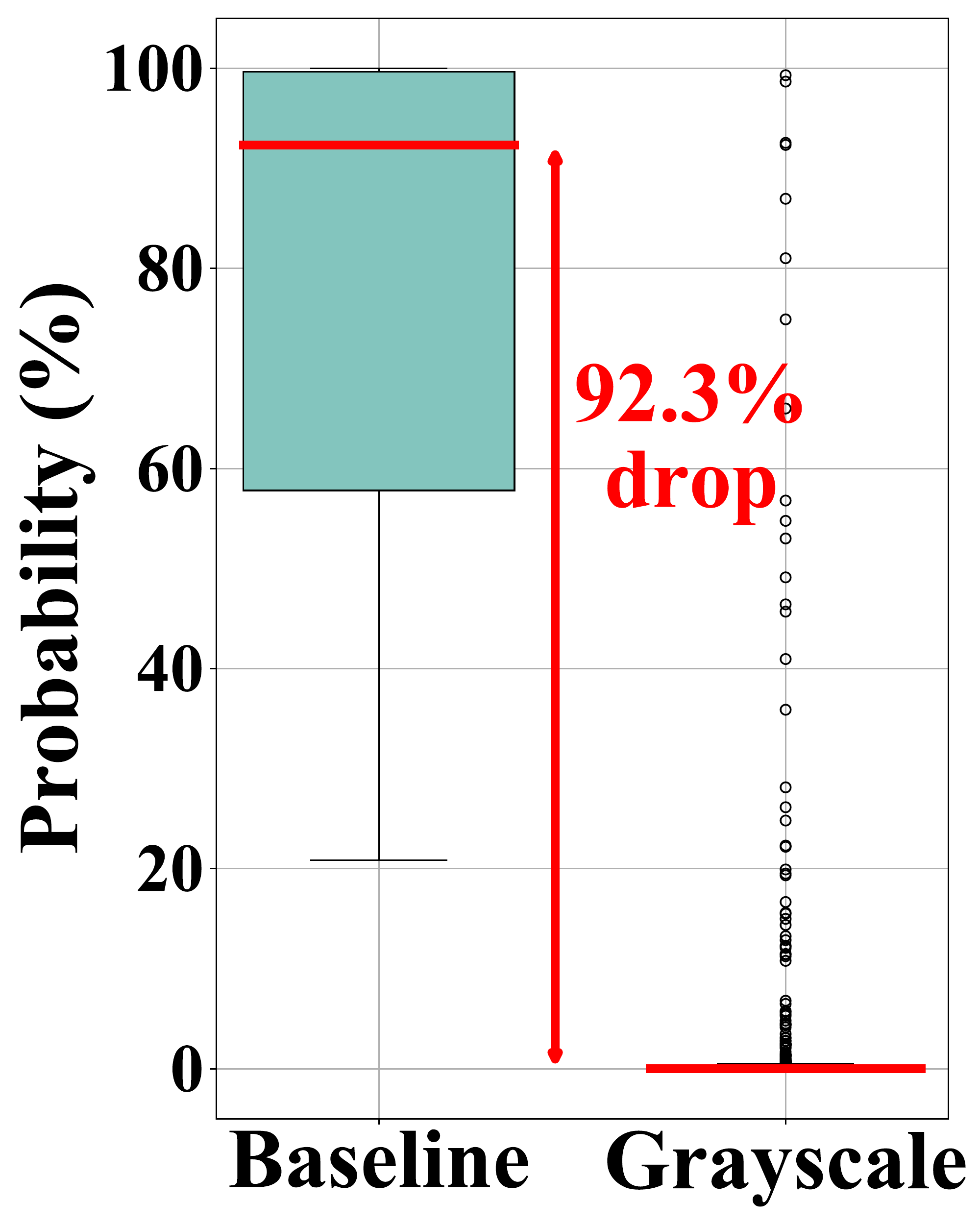} &
    \includegraphics[width=0.13\linewidth]{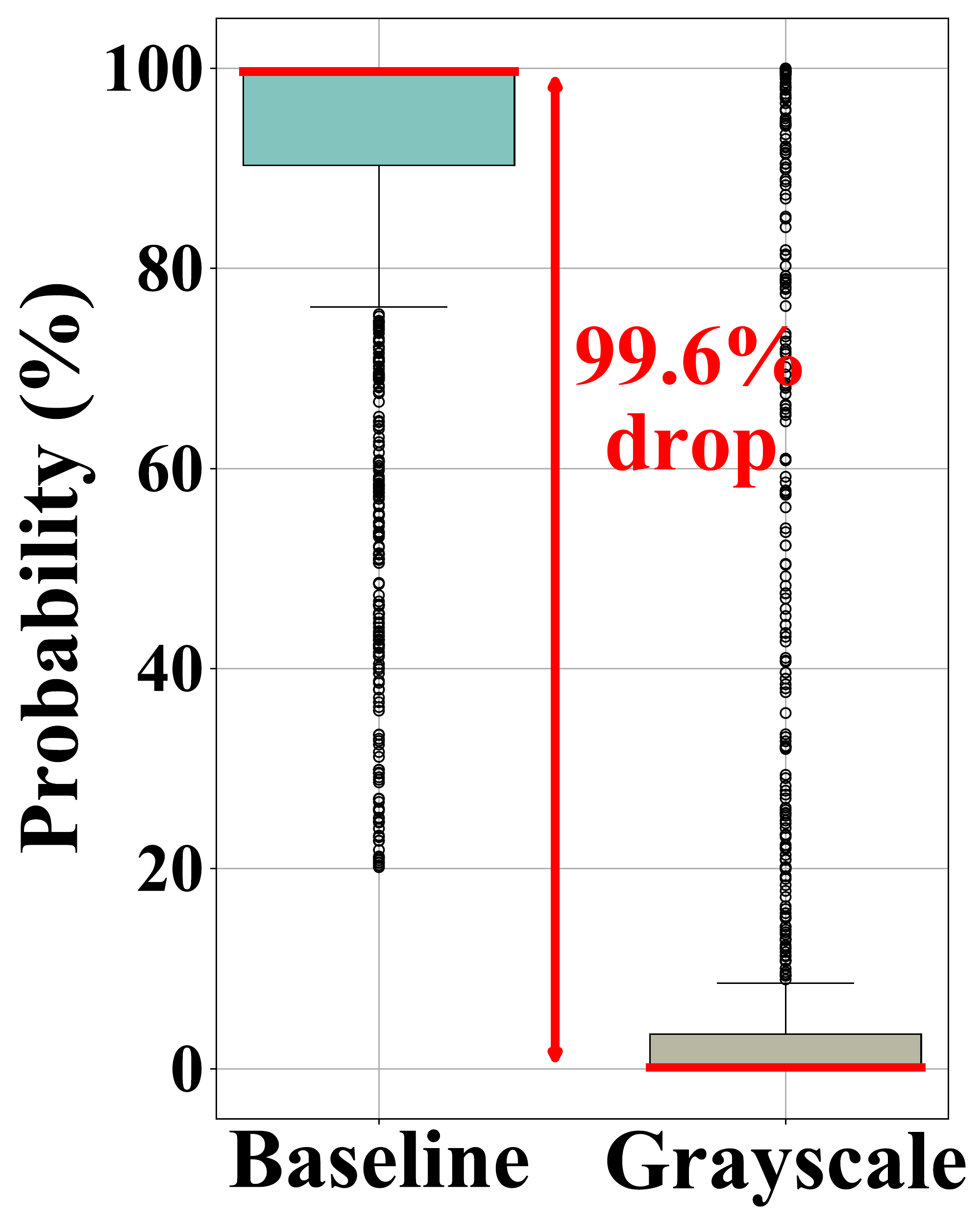} &
    \includegraphics[width=0.13\linewidth]{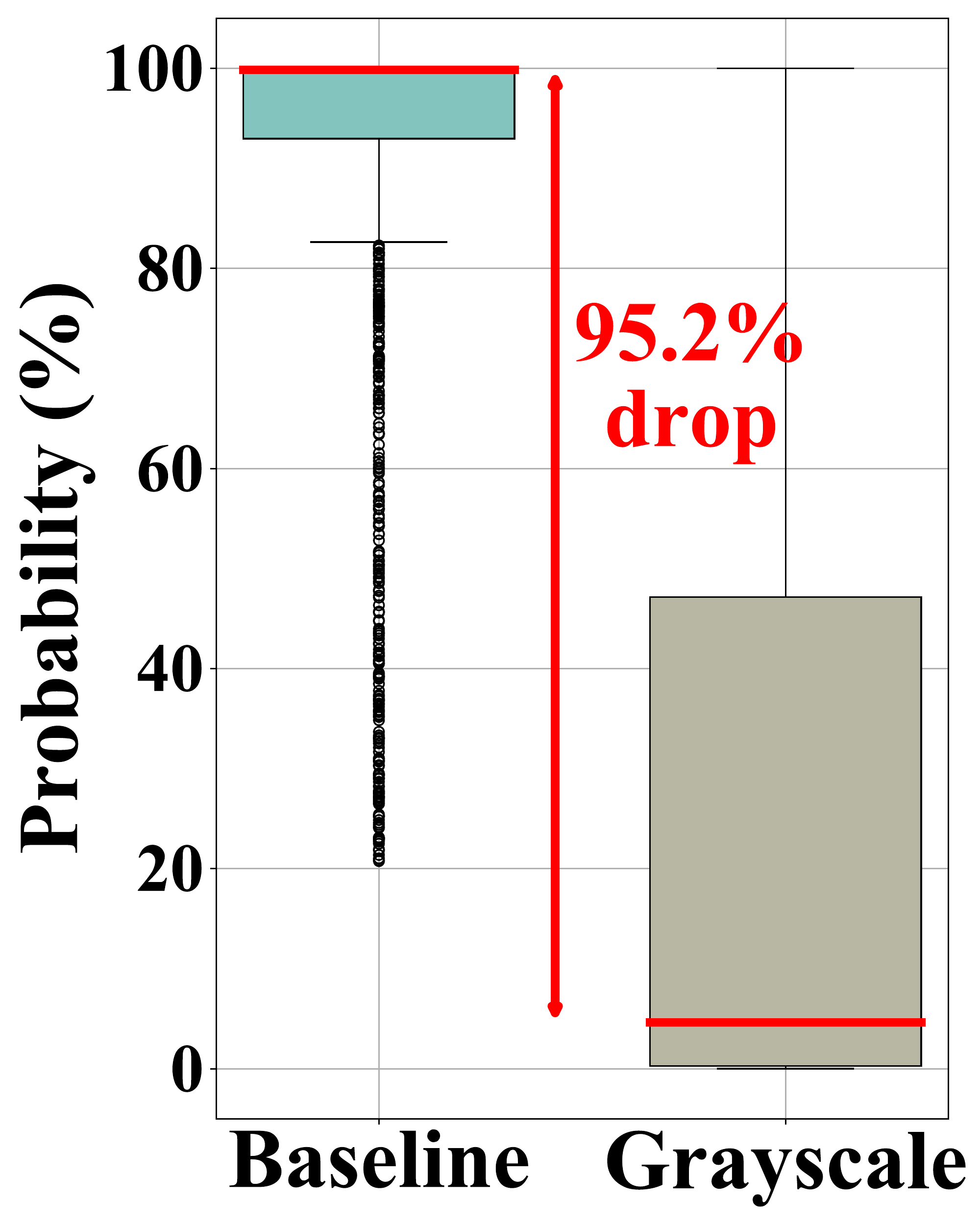} &
    \includegraphics[width=0.13\linewidth]{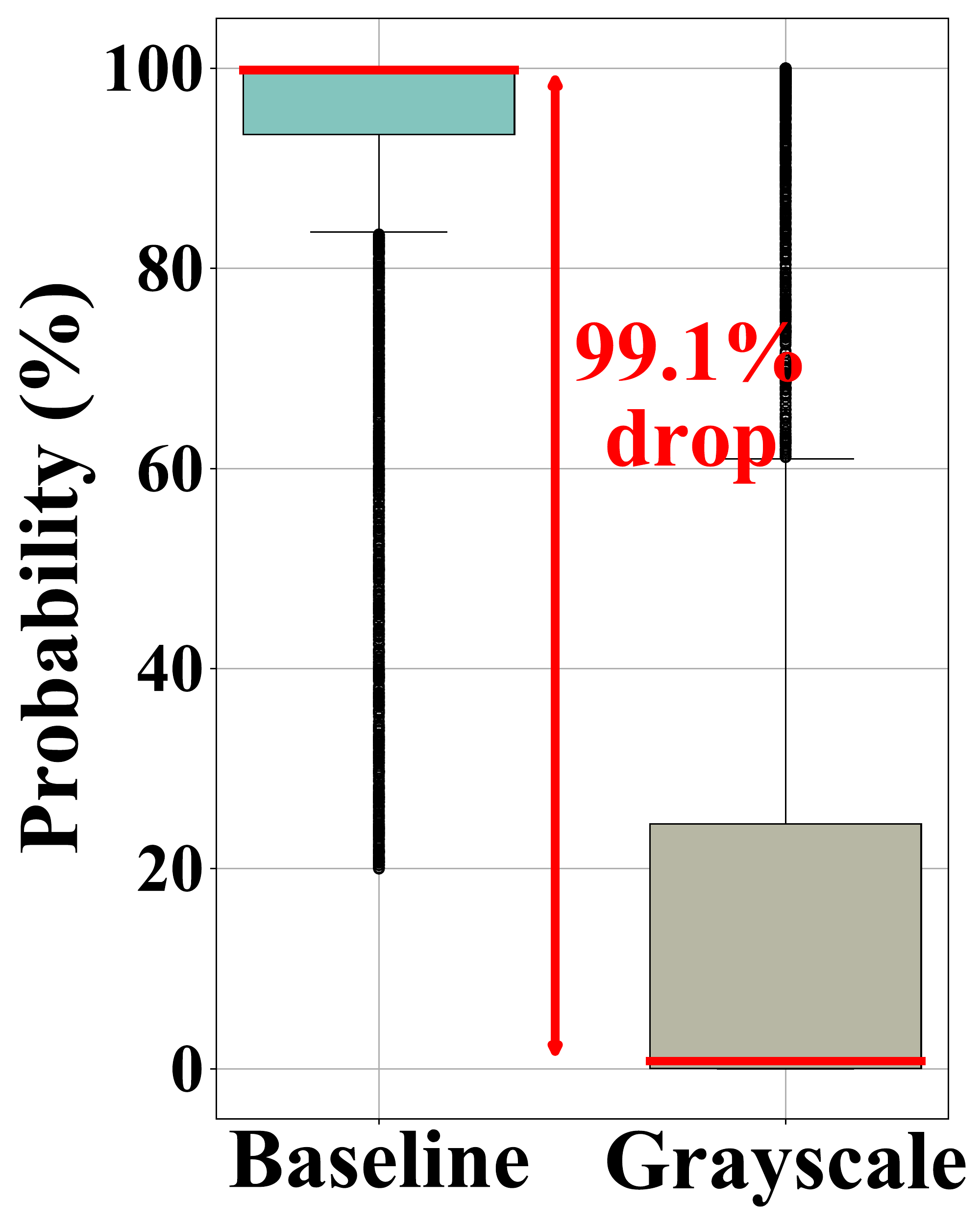}
    \\
    
    \includegraphics[width=0.13\linewidth]{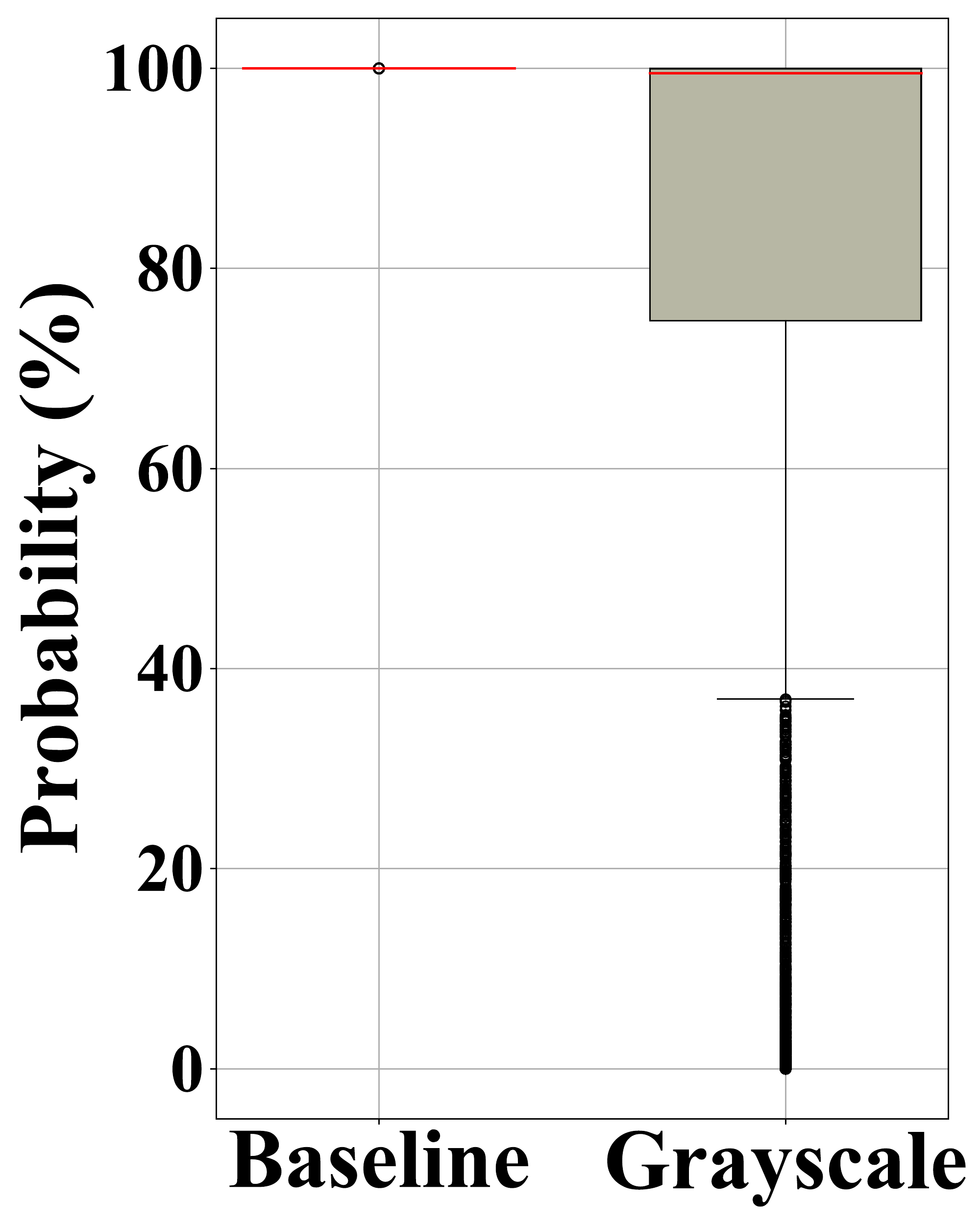} &
    \includegraphics[width=0.13\linewidth]{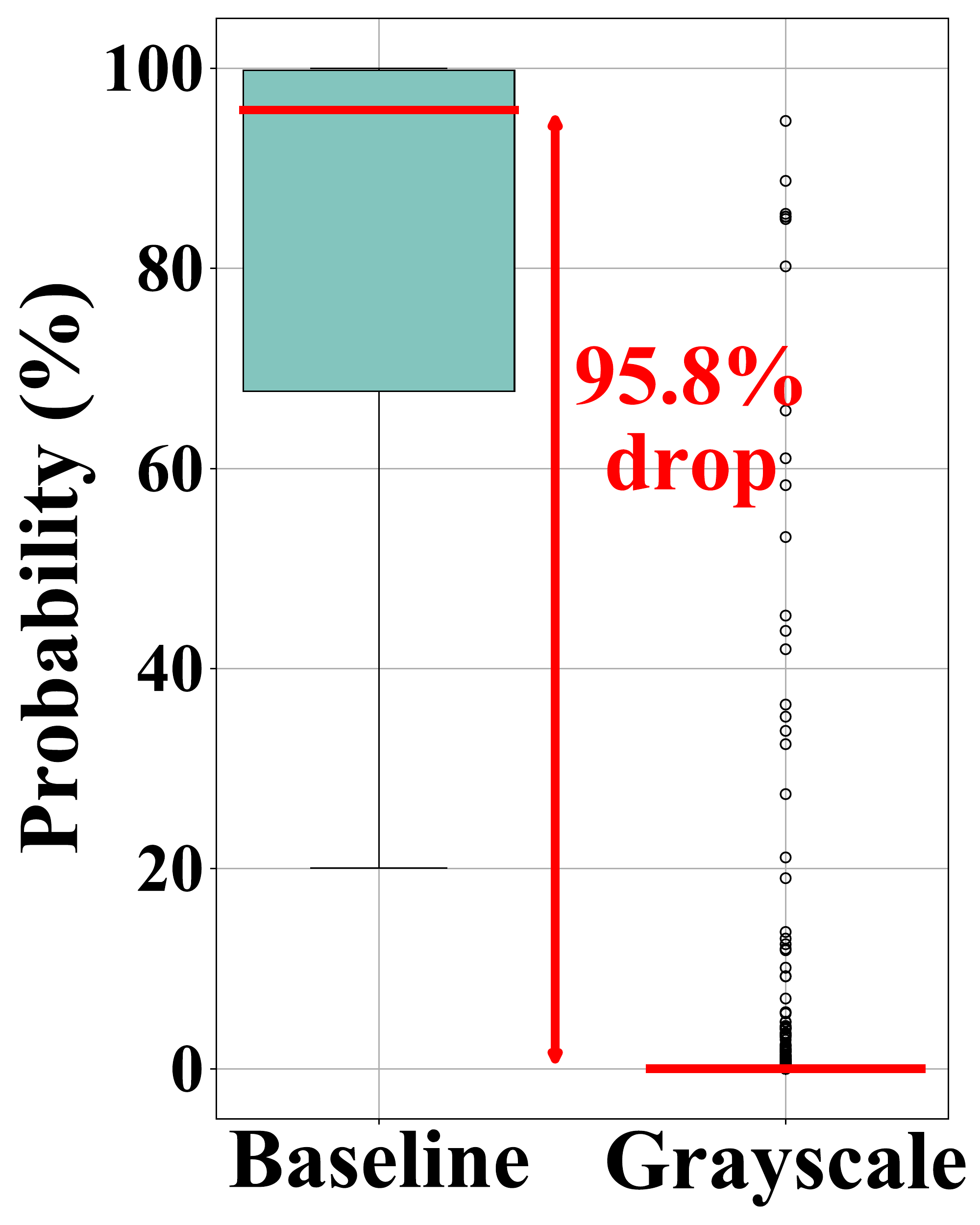} &
    \includegraphics[width=0.13\linewidth]{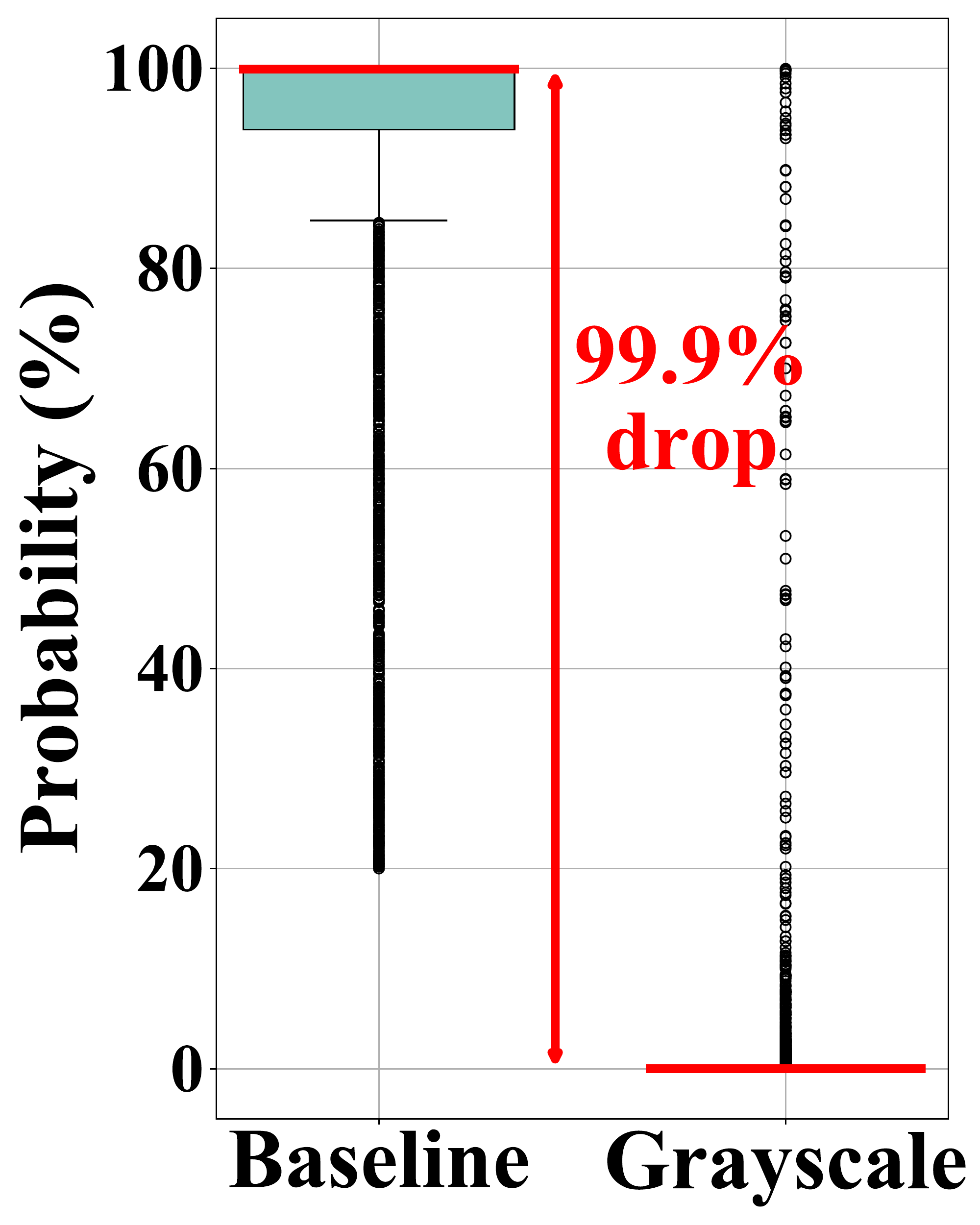} &
    \includegraphics[width=0.13\linewidth]{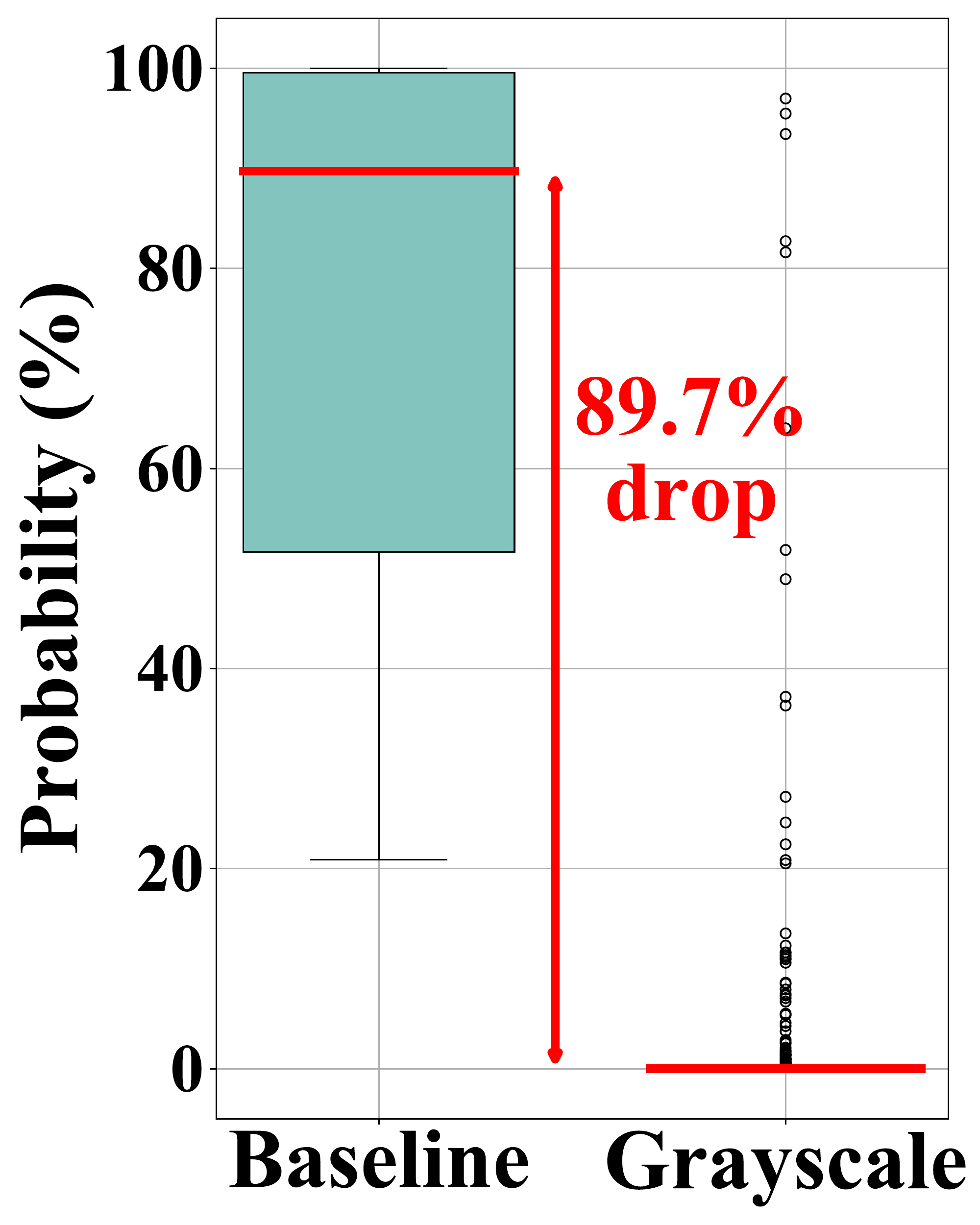} &
    \includegraphics[width=0.13\linewidth]{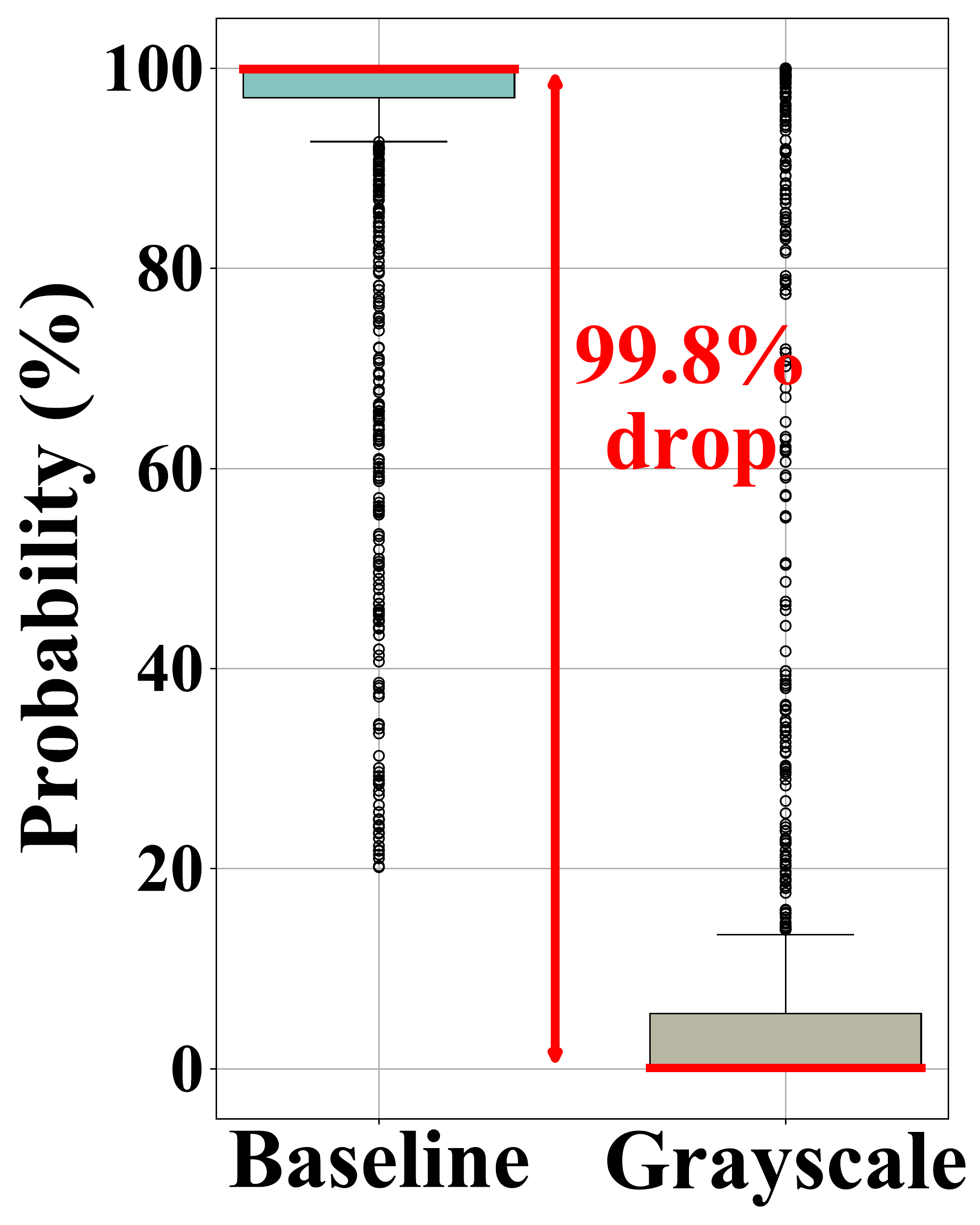} &
    \includegraphics[width=0.13\linewidth]{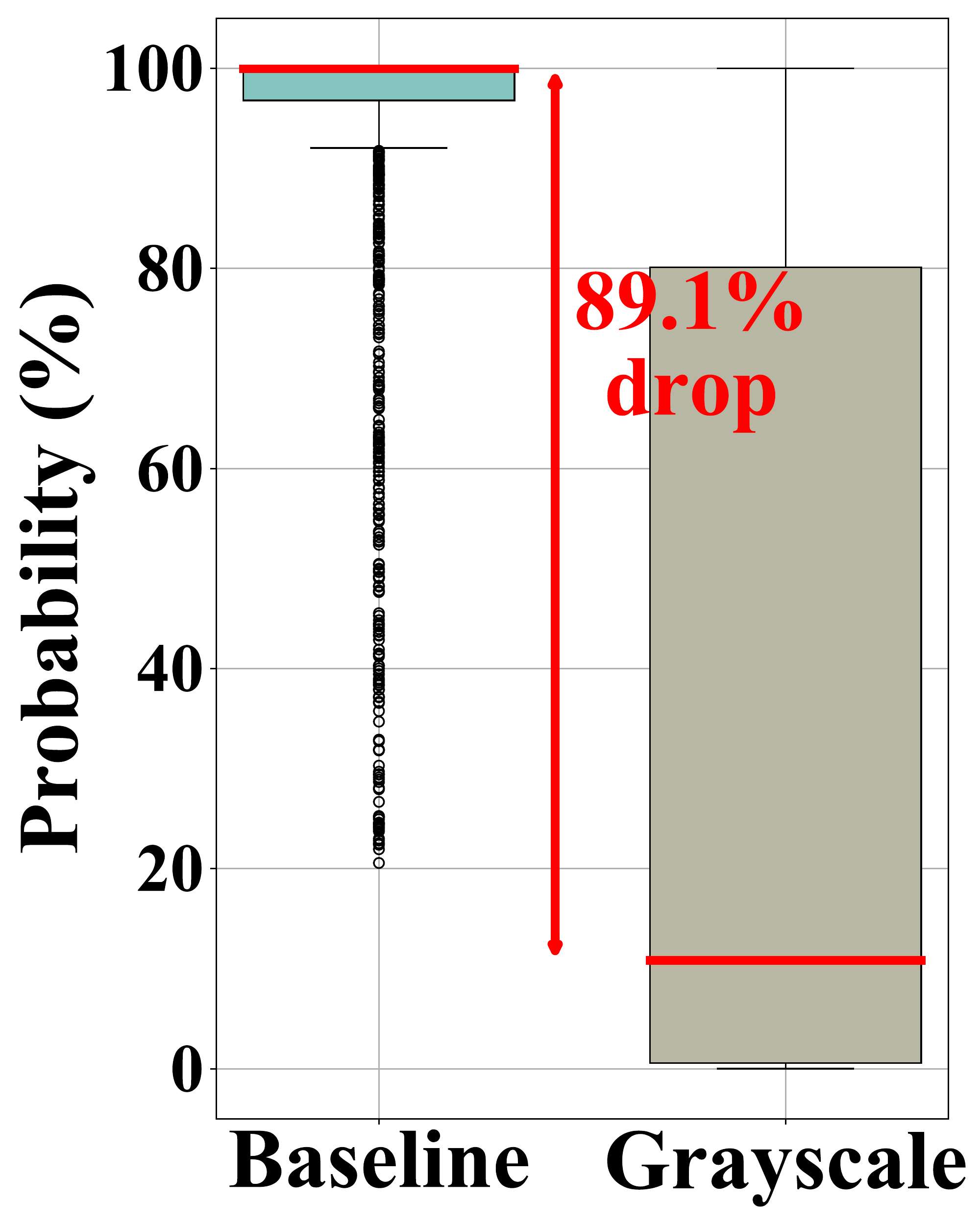} &
    \includegraphics[width=0.13\linewidth]{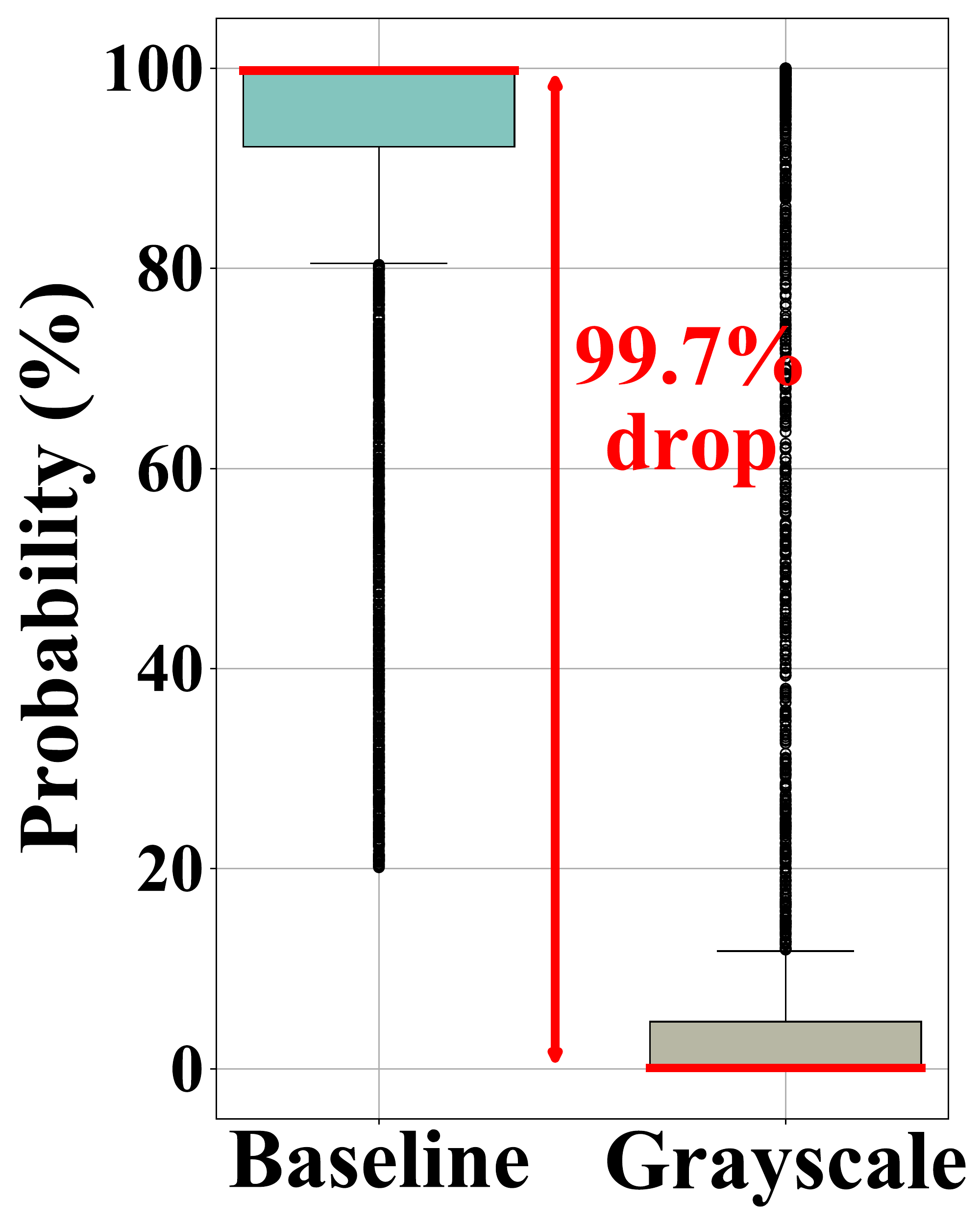}
    \\

\end{tabular}
\vspace{-0.3cm}
\caption{
\textit{Color} is a critical \textit{T-FF} in \textit{universal detectors}:
We show the box-whisker plots of probability (\%) predicted by the universal detector for counterfeits before (Baseline) and after \textit{color ablation} (Grayscale) for 7 GAN models. 
The red line in each box-plot shows the median probability. 
We show the results for the 
ResNet-50
universal detector 
\cite{Wang_2020_CVPR} (top row) and 
our version of EfficientNet-B0 \cite{tan2019efficientnet} universal detector following the exact training / test strategy proposed in
\cite{Wang_2020_CVPR} 
(bottom row).
These detectors are trained with ProGAN
counterfeits and cross-model forensic transfer is evaluated on other unseen GANs.
All counterfeits are obtained from the ForenSynths dataset 
\cite{Wang_2020_CVPR}.
We clearly show that \textit{color ablation} causes the median probability for counterfeits to drop by more than 89\% across all unseen GANs. 
This is consistently seen across both universal detectors.
These observations quantitatively show that \textit{color} is a critical \textit{T-FF} in universal detectors.
AP and accuracies shown in Supplementary
\ref{table_supp:original}.
\vspace{-0.5cm}
}
\label{fig_main:median_color_ablation}
\end{figure}

\textbf{Study 1.} We investigate the change in probability distribution of universal detectors when removing color information in counterfeits during cross-model forensic transfer. 
We specifically study the change in median counterfeit probability when removing color information 
(median is not sensitive to outliers). 
The results for both ResNet-50 and EfficientNet-B0 universal detectors are shown in Fig.  \ref{fig_main:median_color_ablation}.
As one can clearly observe, color ablation causes the median probability predicted by the universal detector to drop by more than 89\% across all unseen GANs showing that \textit{color} is a critical \textit{T-FF} in universal detectors. This is observed in both ResNet-50 and EfficientNet-B0 universal detectors.

\begin{figure}[!t]
\centering
 
\begin{tabular}{ccccccc}
    {\tiny ProGAN \cite{karras2018progressive}} &
    {\tiny StyleGAN2 \cite{Karras_2020_CVPR}} &
    {\tiny StyleGAN \cite{Karras_2019_CVPR}} &
    {\tiny BigGAN \cite{brock2018large}} &
    {\tiny CycleGAN \cite{zhu2017unpaired}} &
    {\tiny StarGAN \cite{choi2018stargan}} &
    {\tiny GauGAN \cite{park2019semantic}} \\
    
    \multicolumn{7}{c}{\bf ResNet-50}\\
    \includegraphics[width=0.13\linewidth]{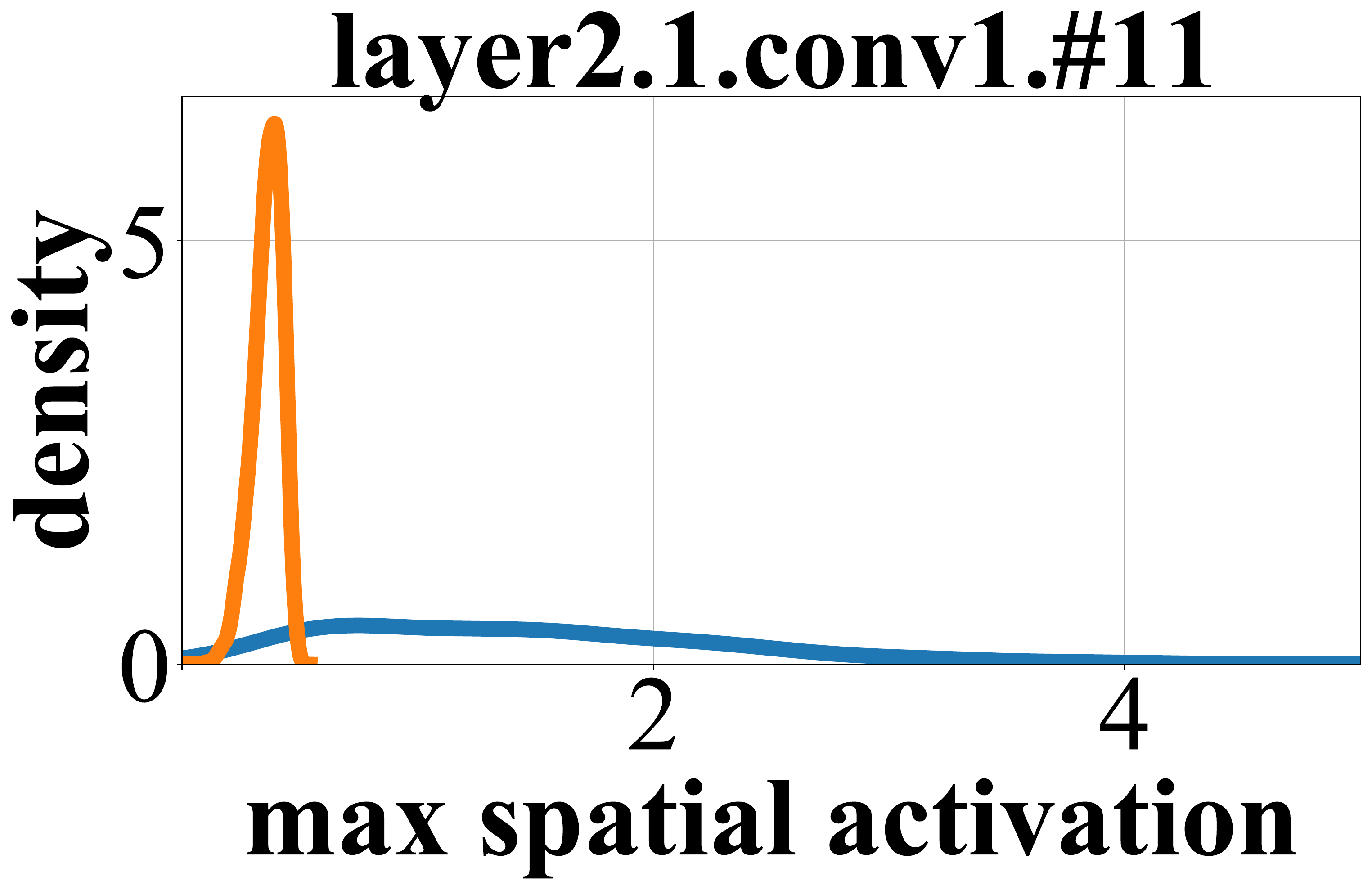} &
    \includegraphics[width=0.13\linewidth]{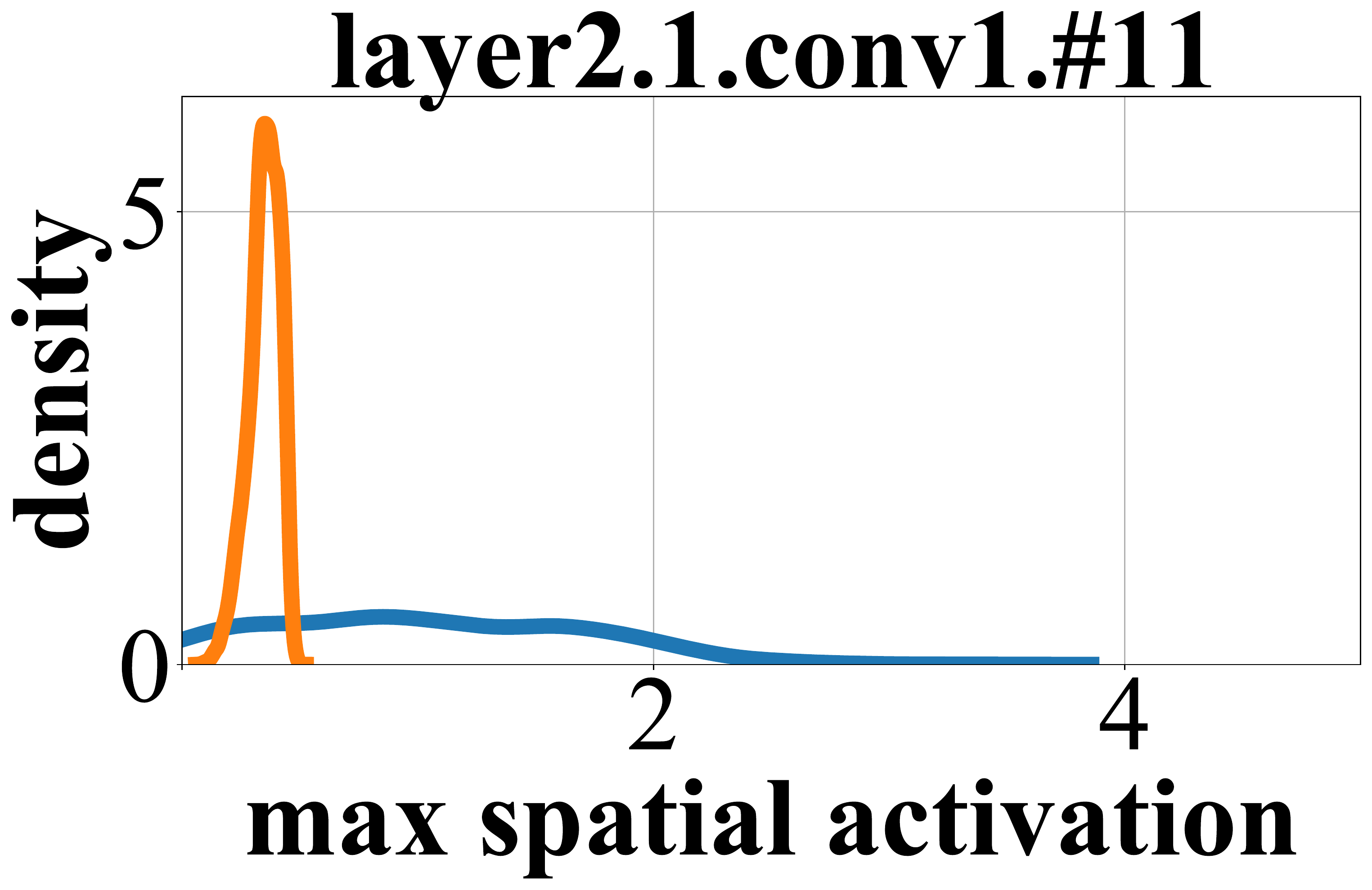} &
    \includegraphics[width=0.13\linewidth]{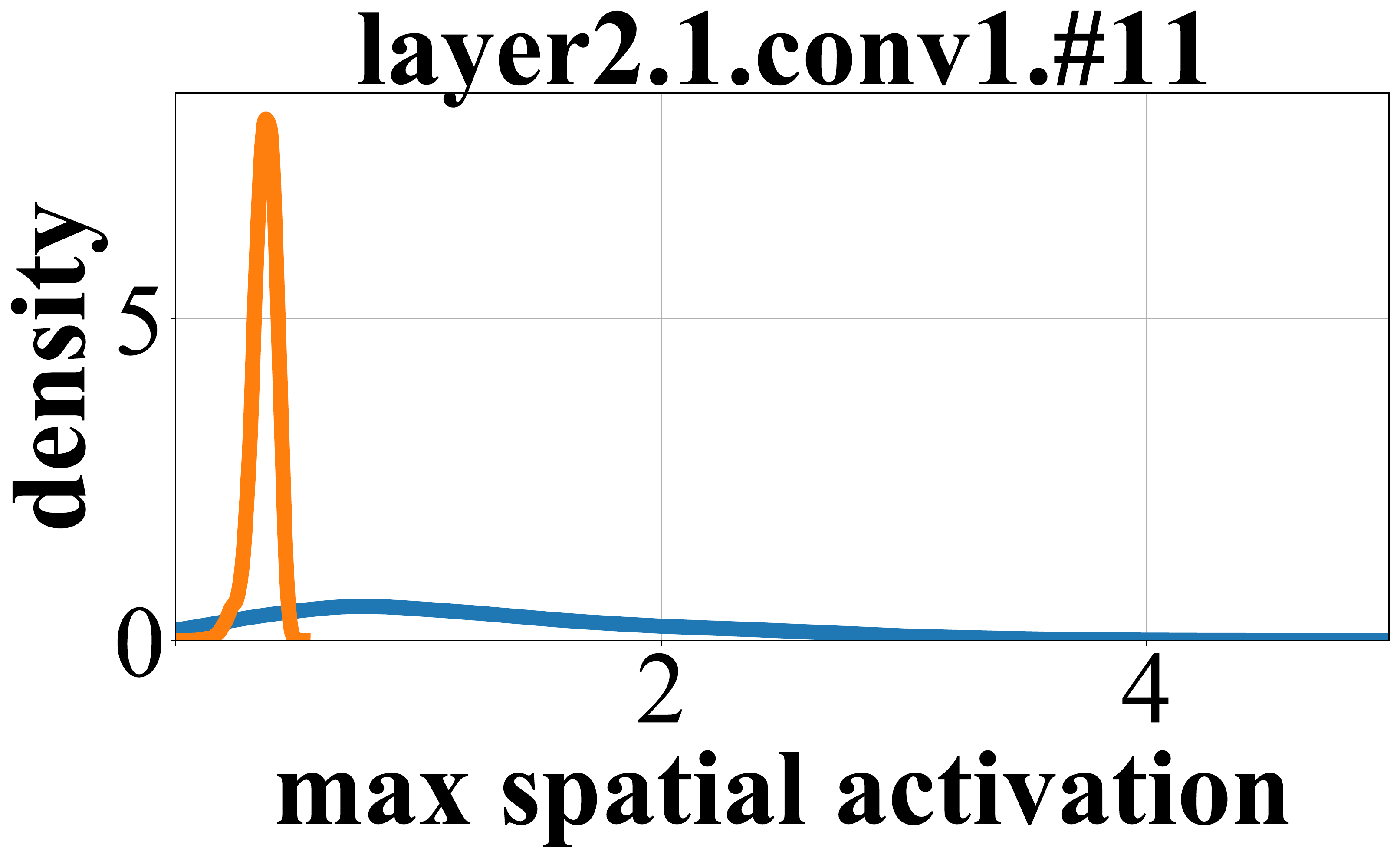} &
     \includegraphics[width=0.13\linewidth]{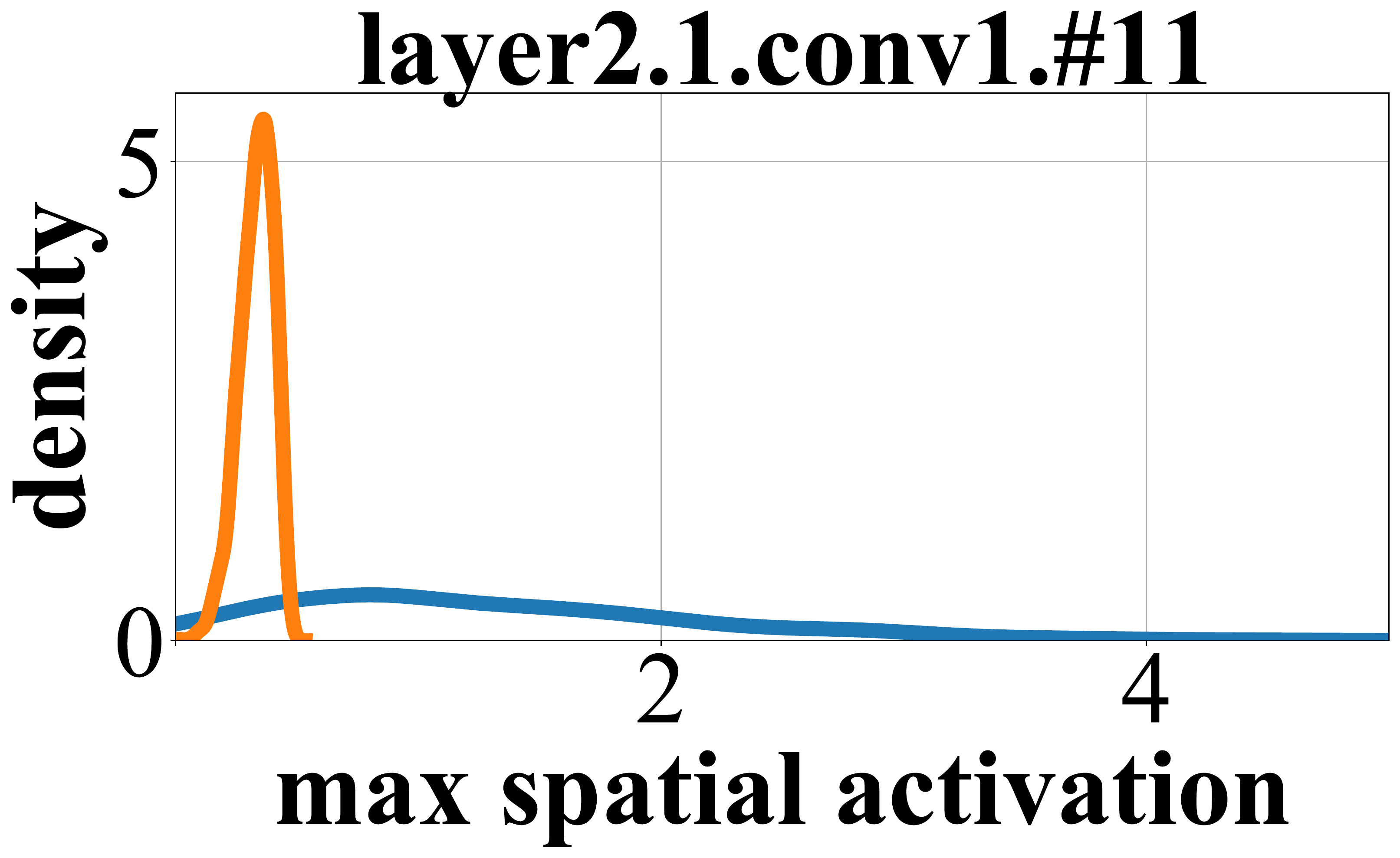} &
    \includegraphics[width=0.13\linewidth]{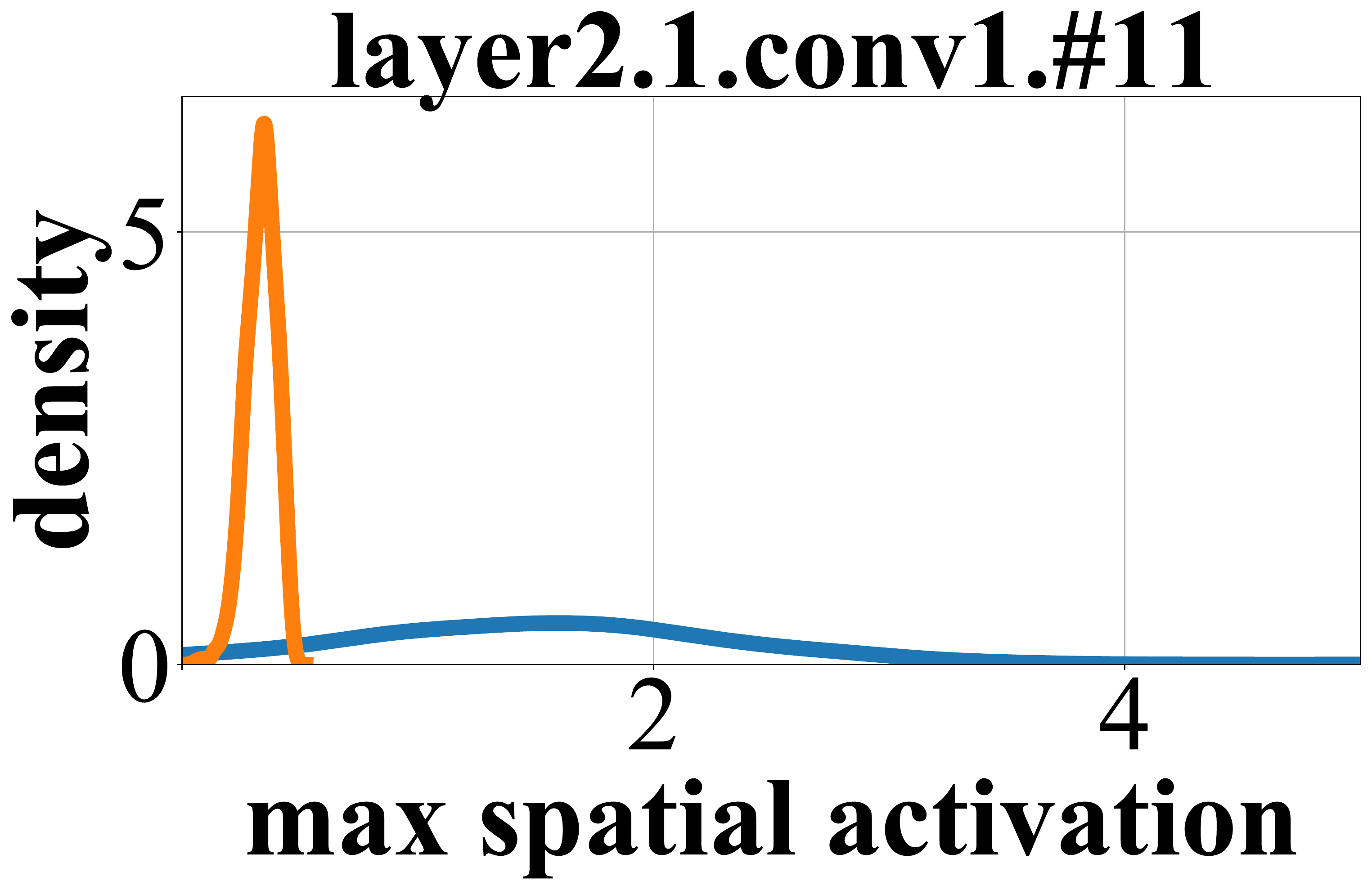} &
     \includegraphics[width=0.13\linewidth]{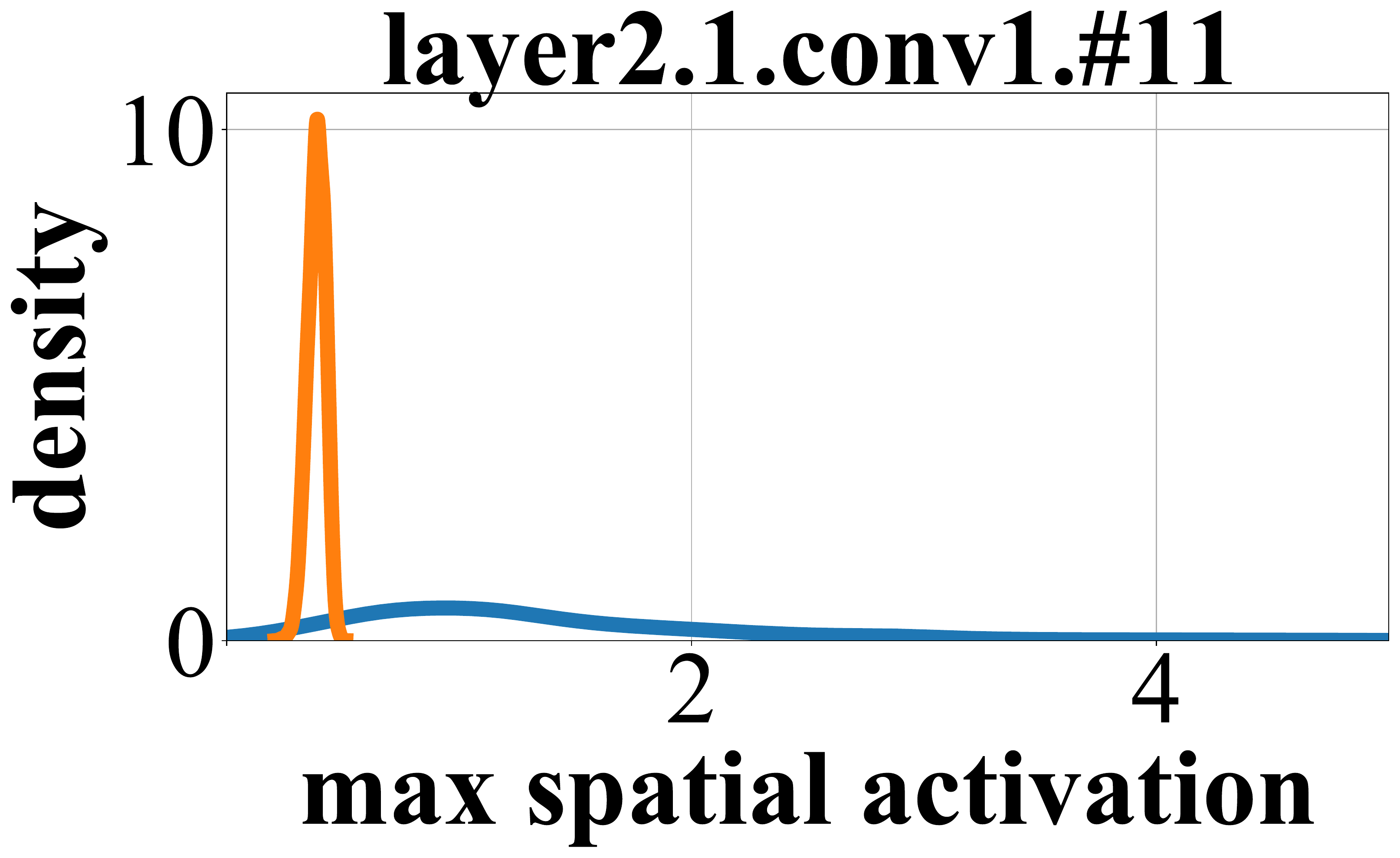} &
     \includegraphics[width=0.13\linewidth]{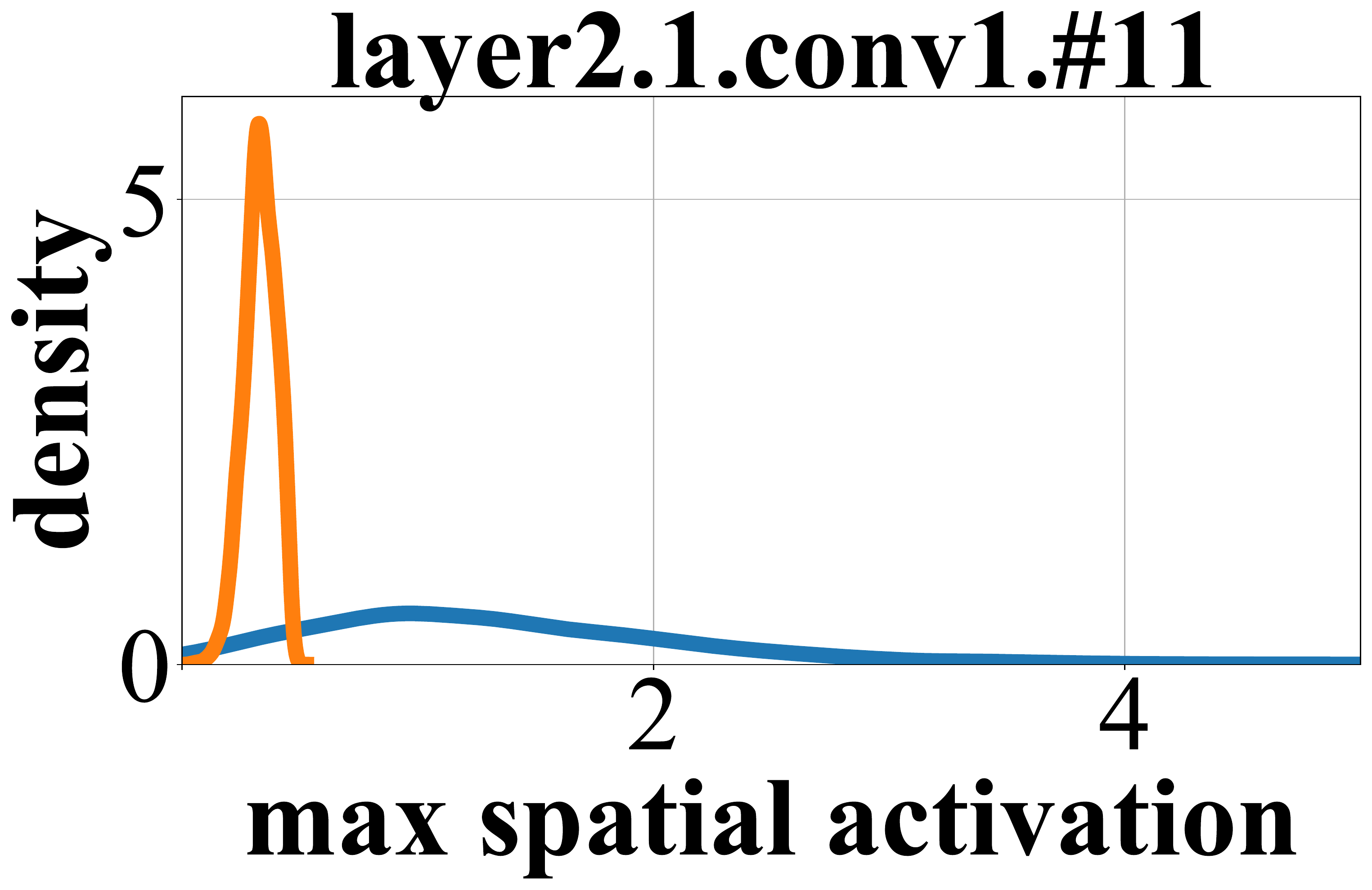}
    \\
    
    \includegraphics[width=0.13\linewidth]{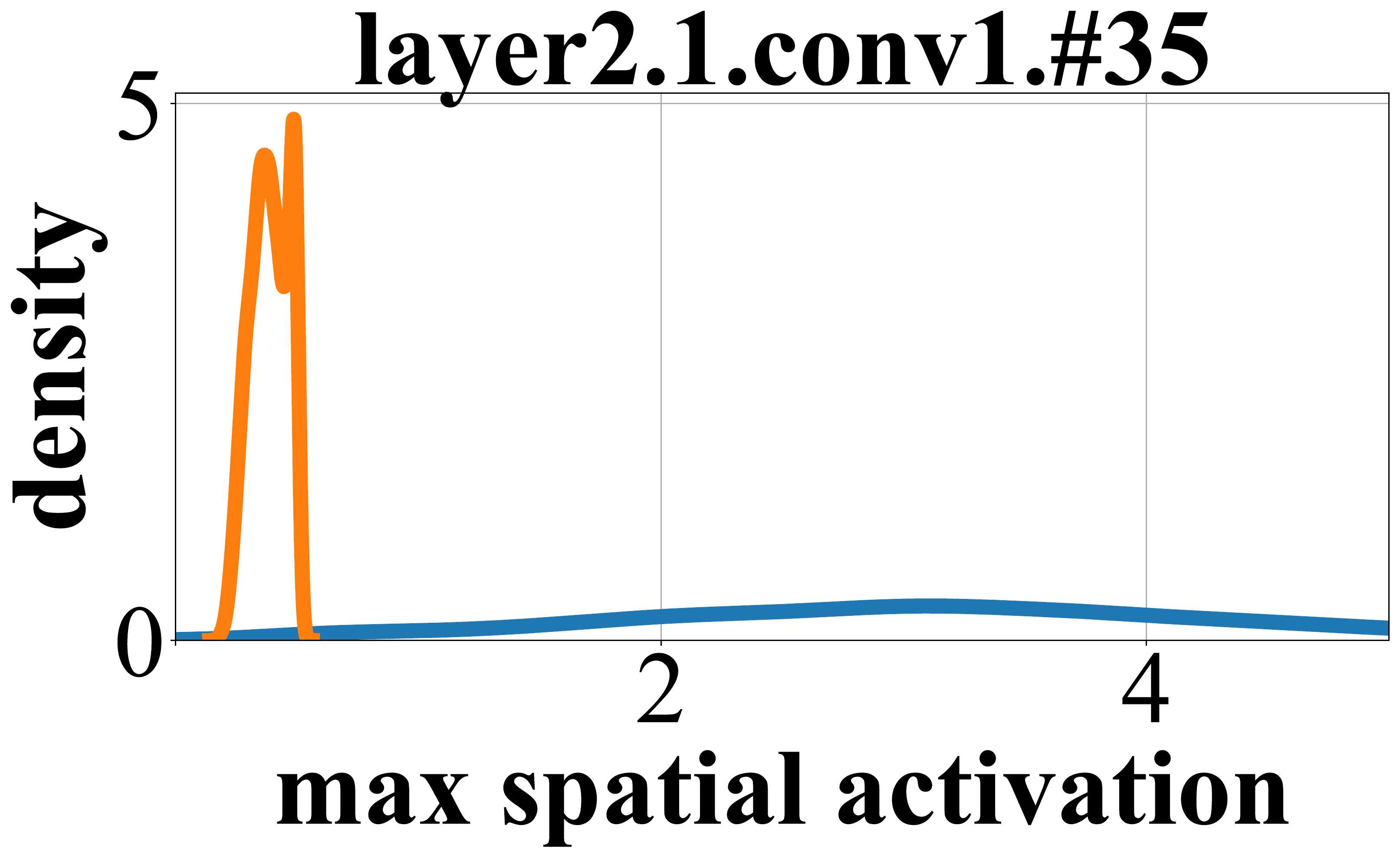} &
    \includegraphics[width=0.13\linewidth]{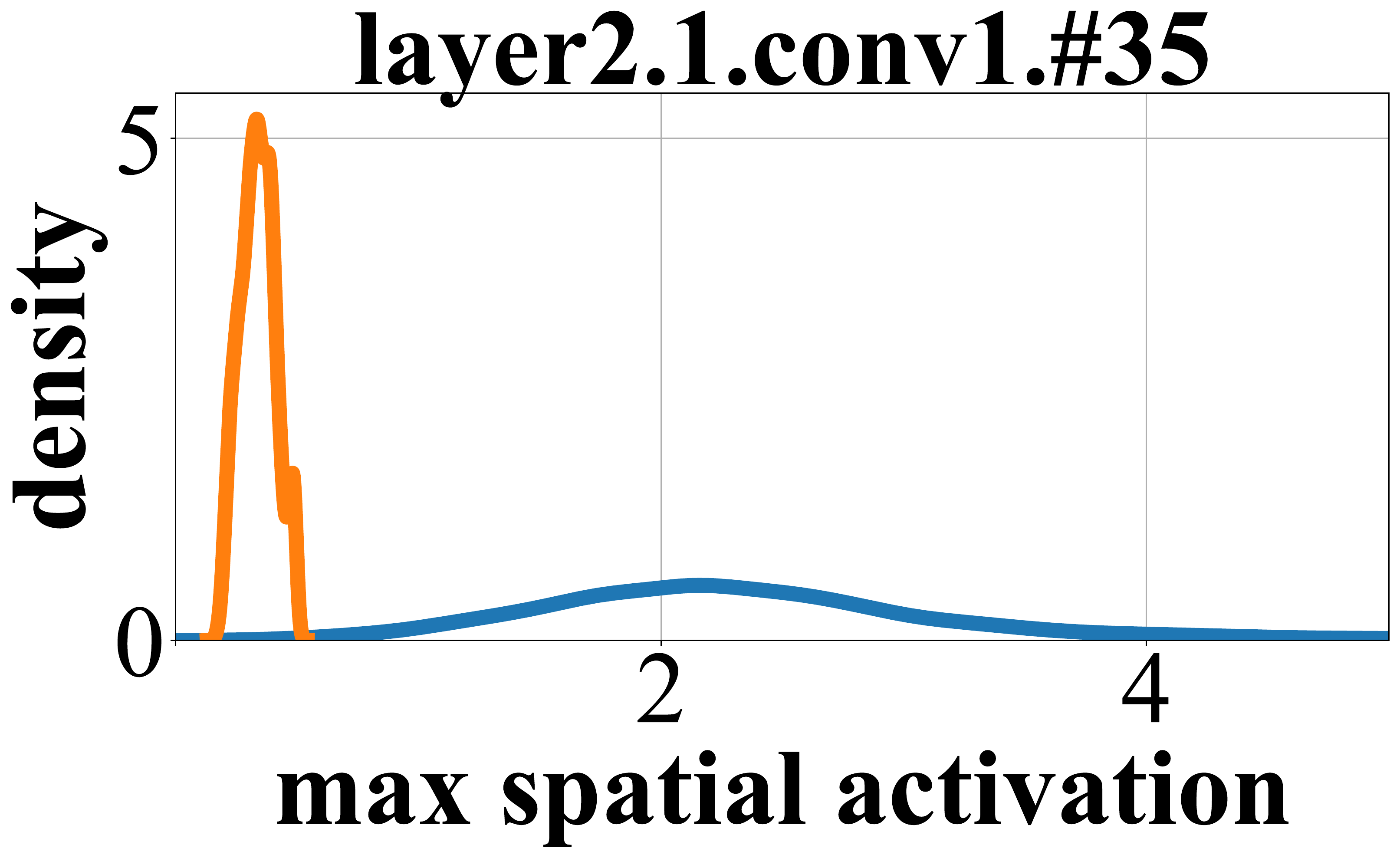} &
    \includegraphics[width=0.13\linewidth]{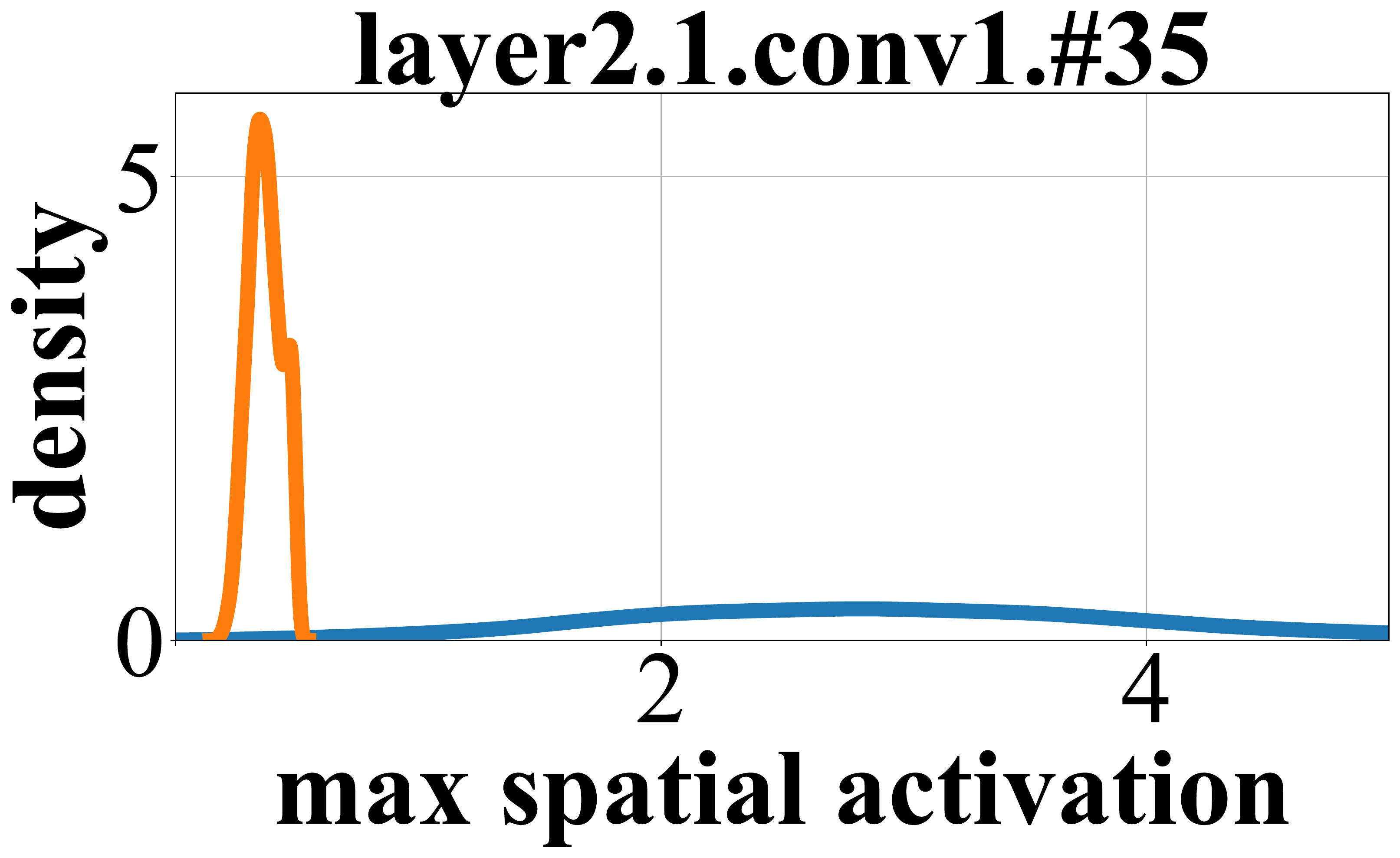} &
     \includegraphics[width=0.13\linewidth]{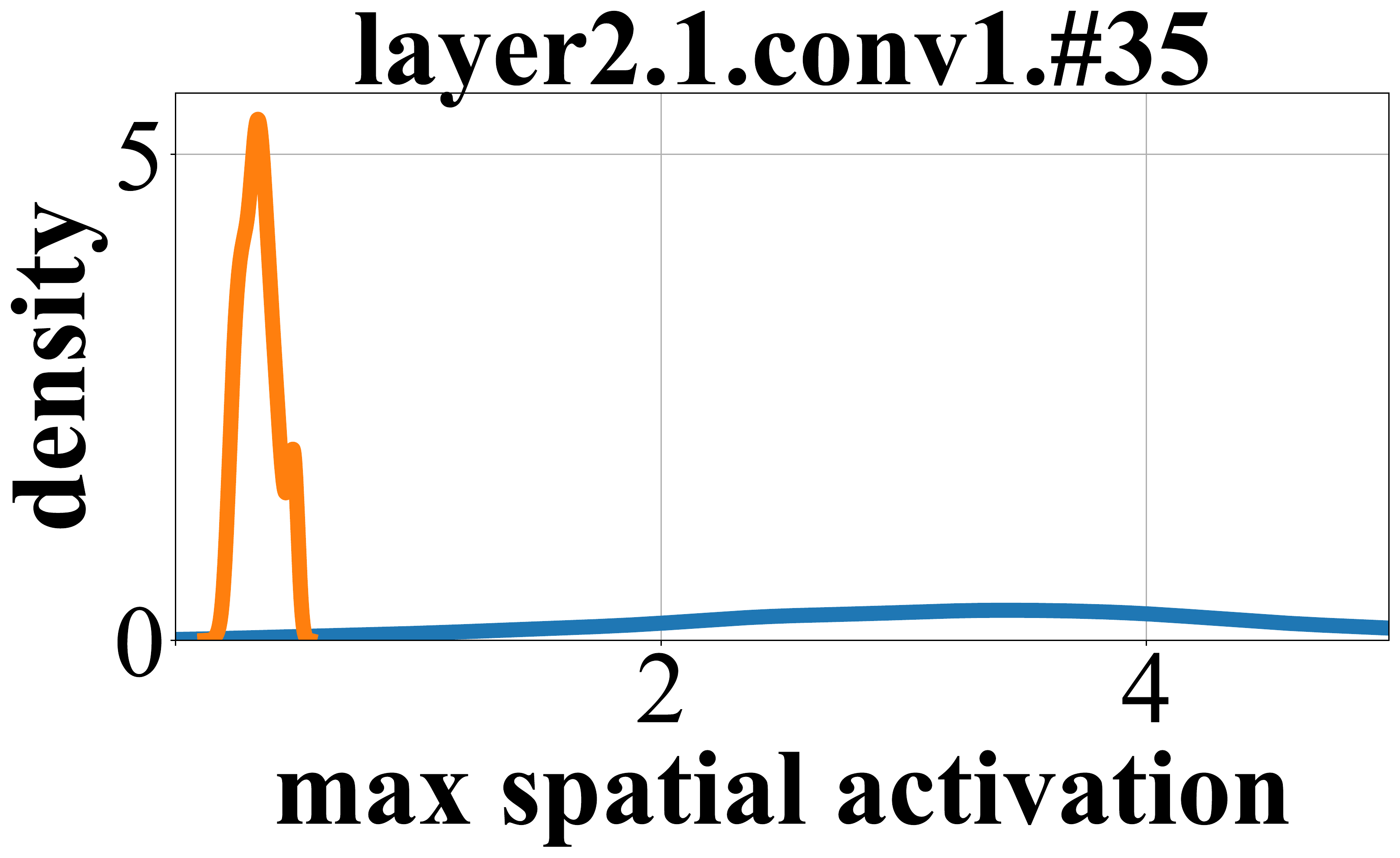} &
    \includegraphics[width=0.13\linewidth]{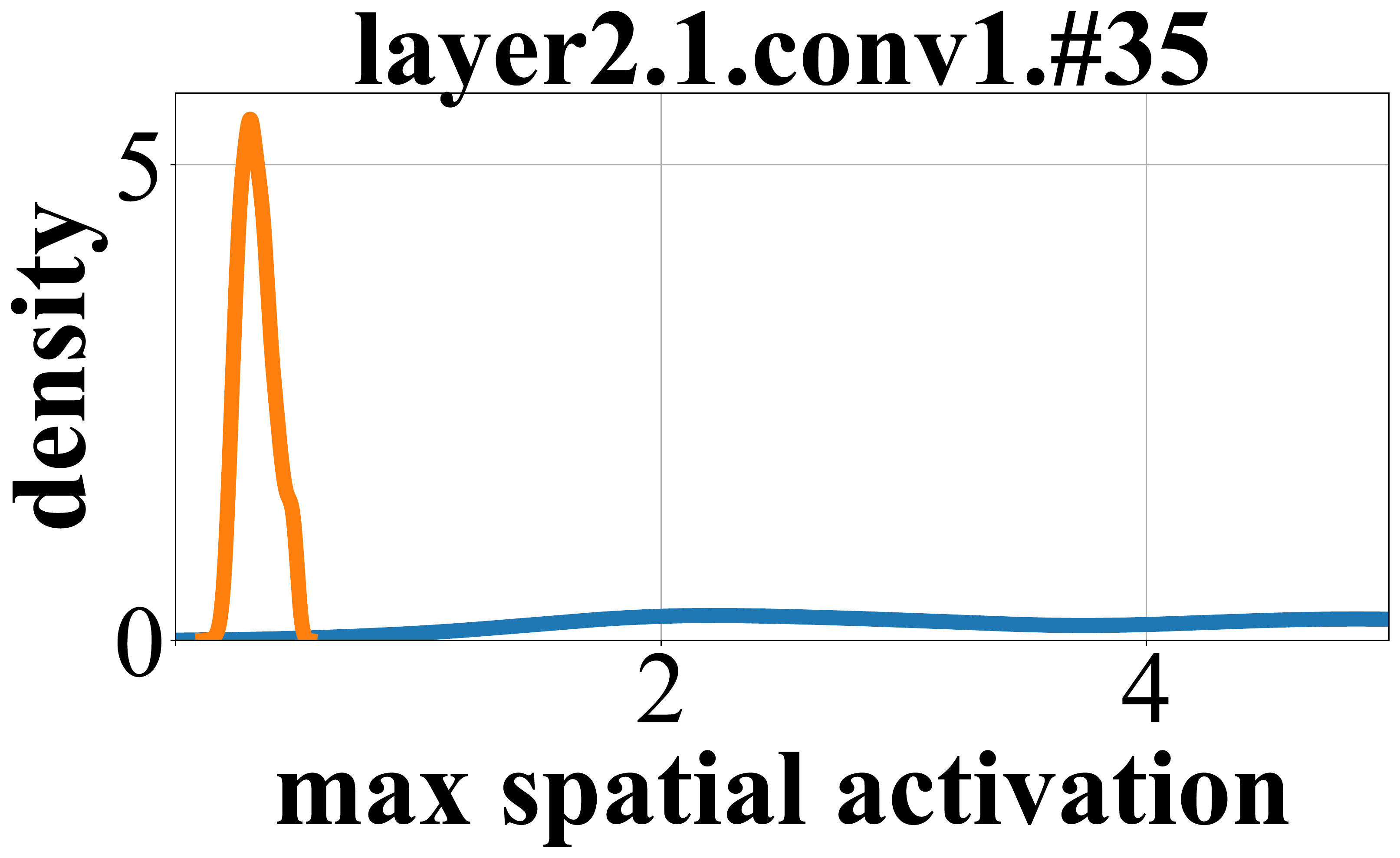} &
     \includegraphics[width=0.13\linewidth]{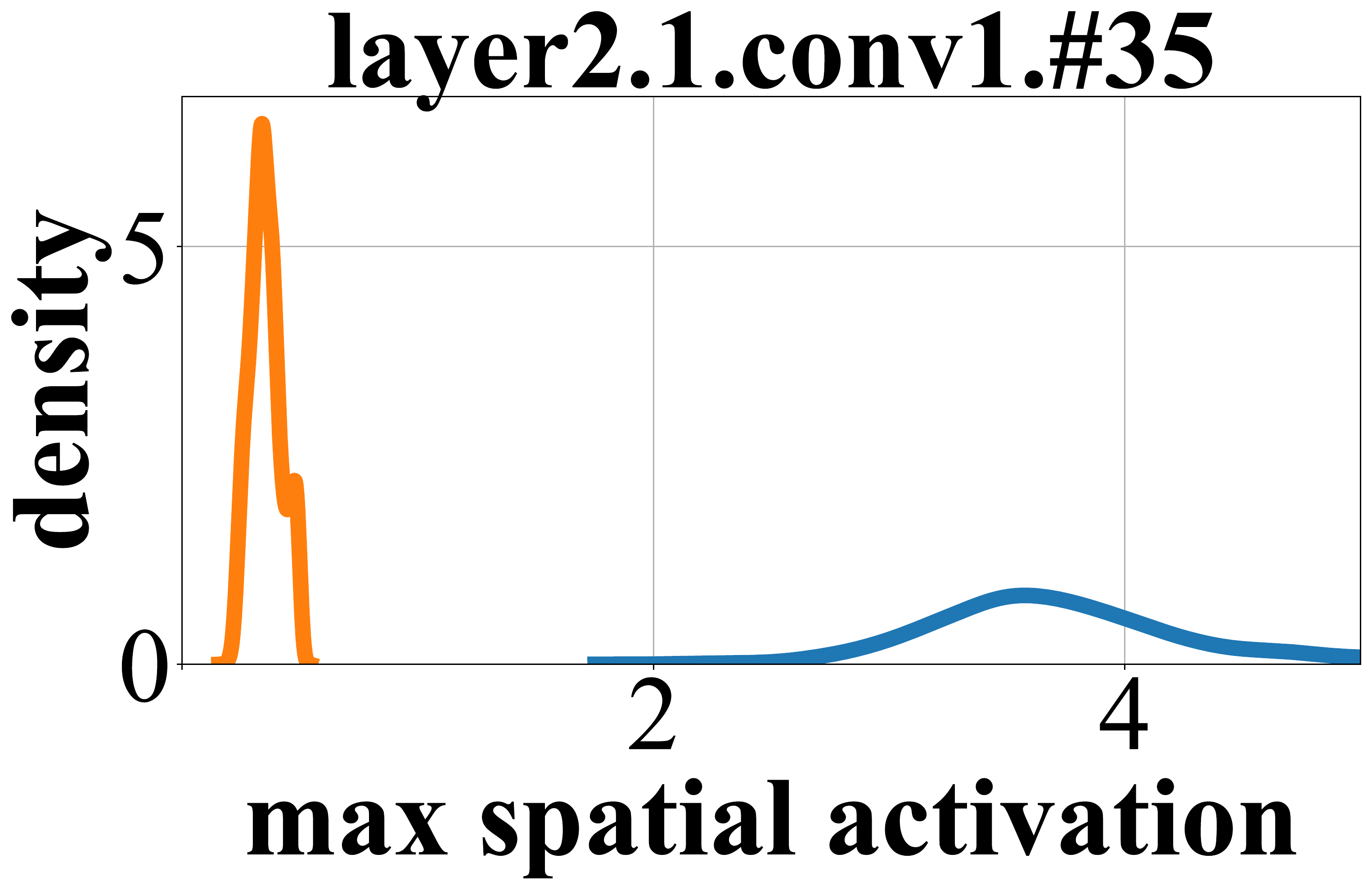} &
     \includegraphics[width=0.13\linewidth]{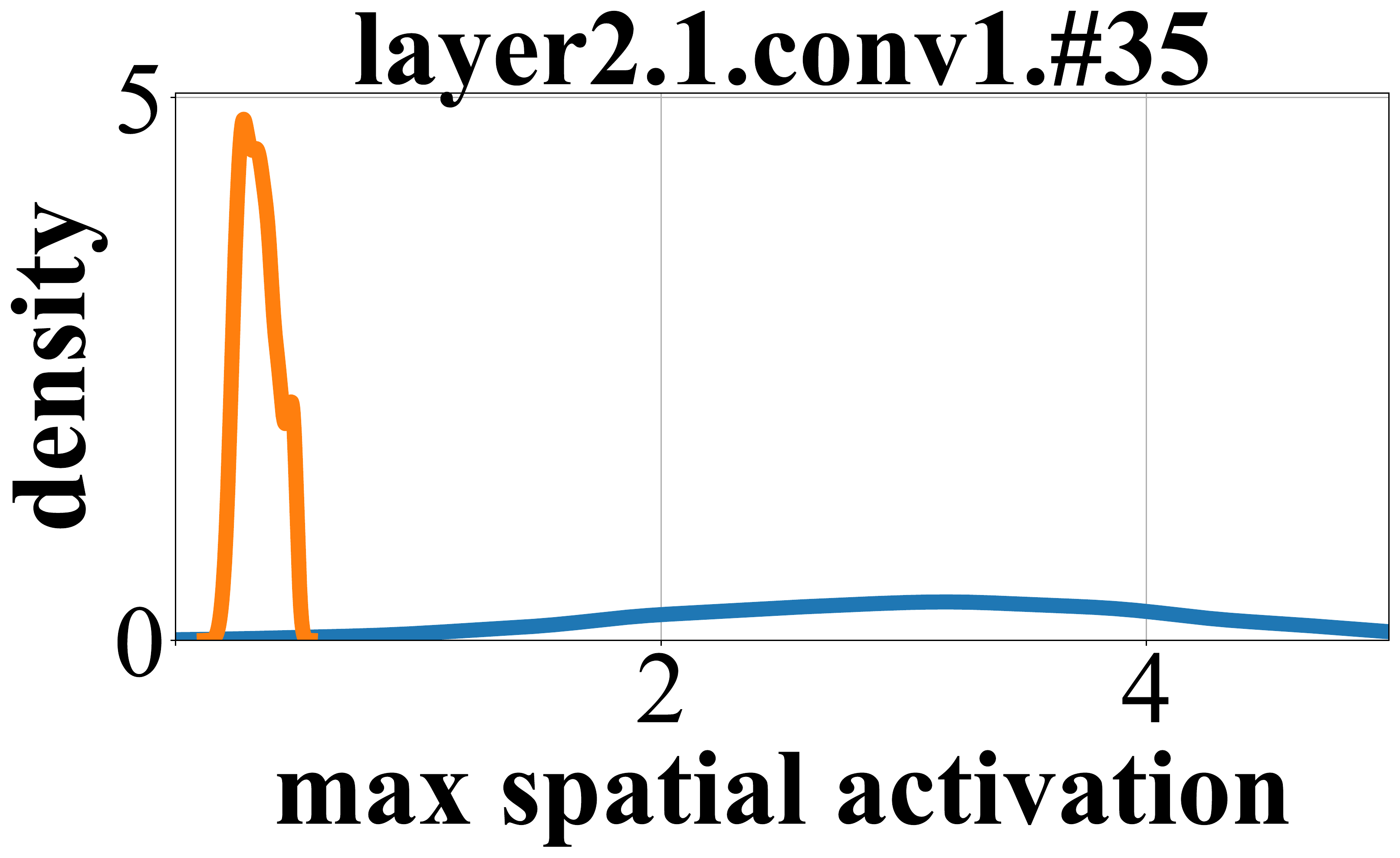}
    \\
    
     \includegraphics[width=0.13\linewidth]{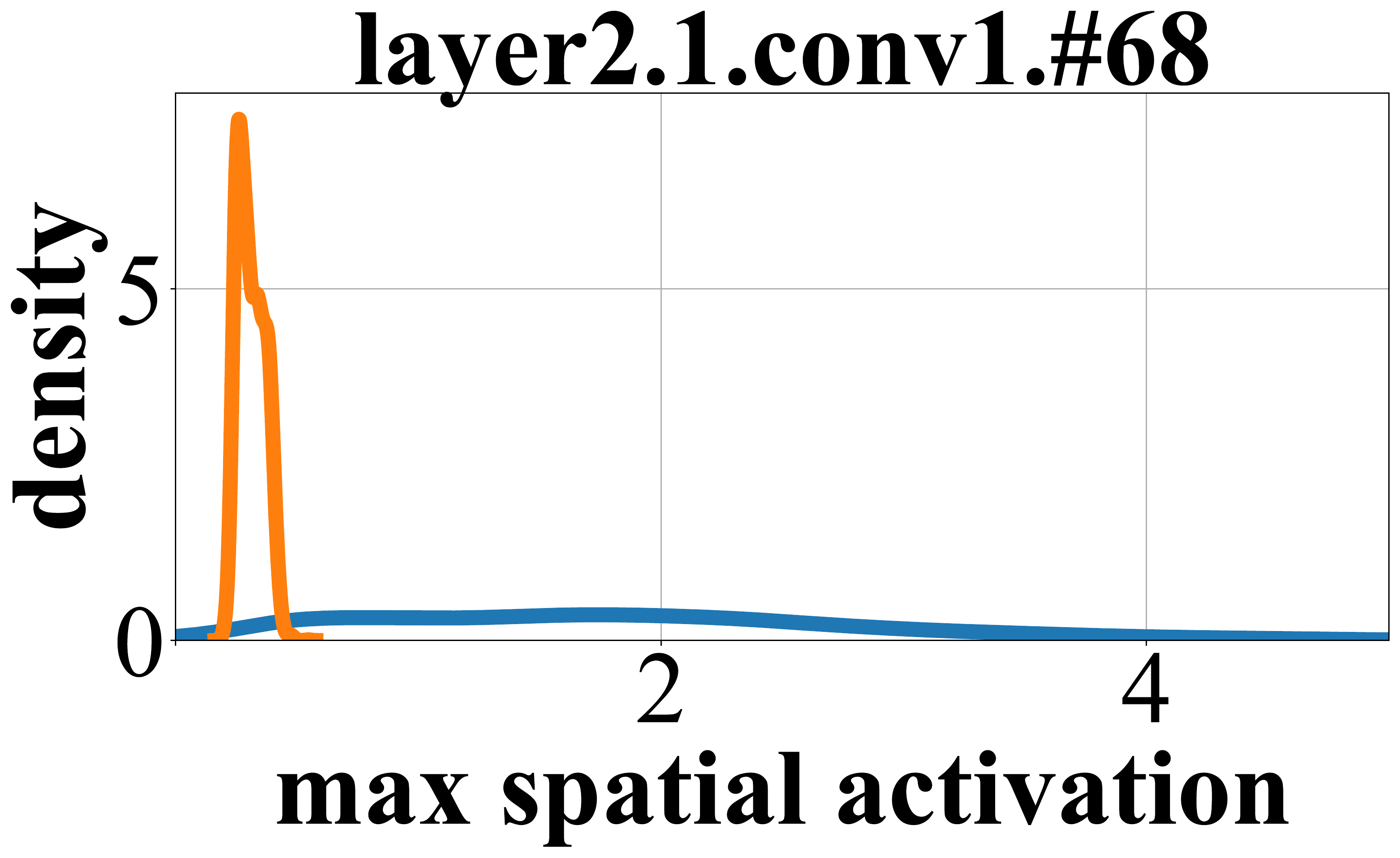} &
    \includegraphics[width=0.13\linewidth]{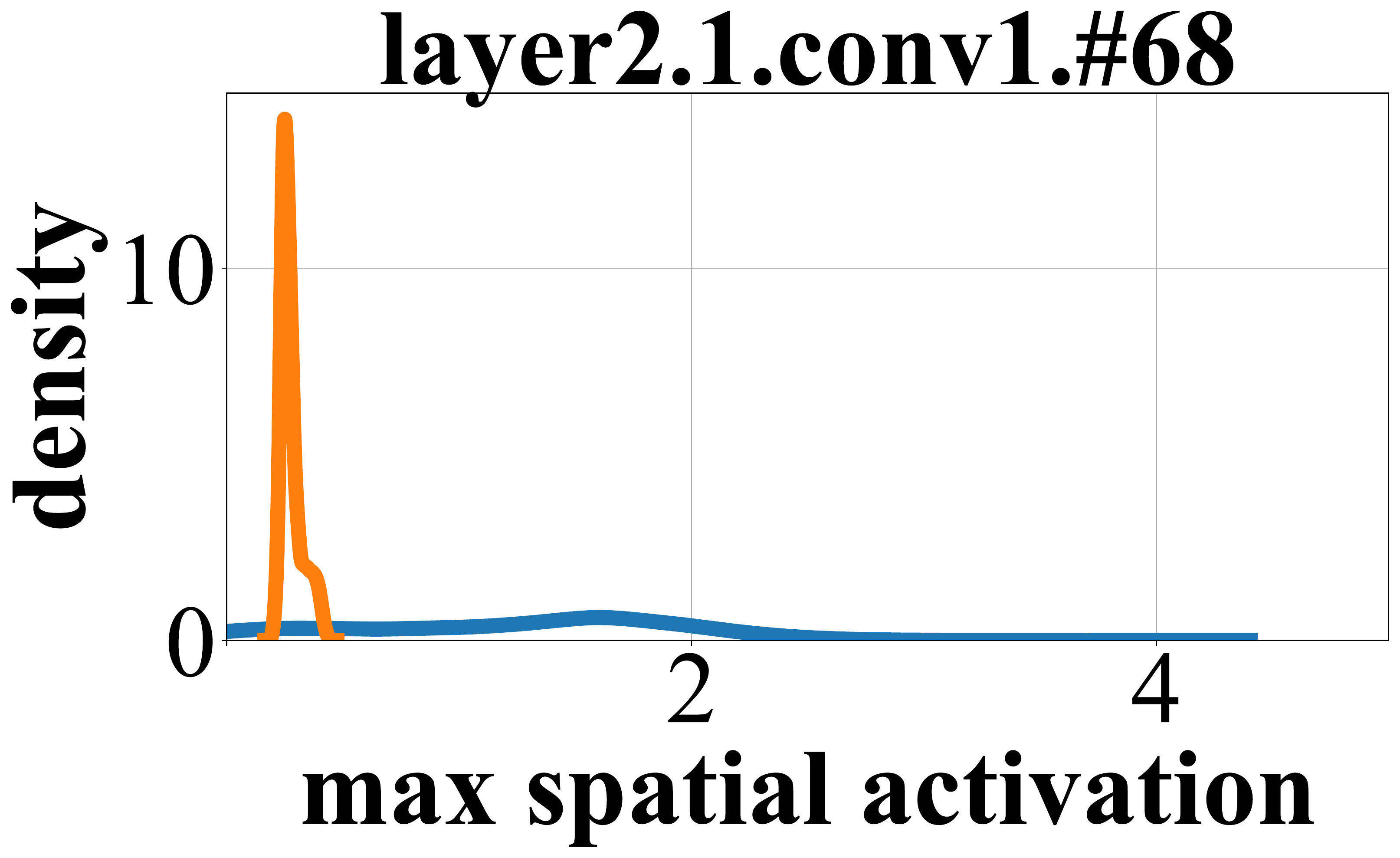} &
    \includegraphics[width=0.13\linewidth]{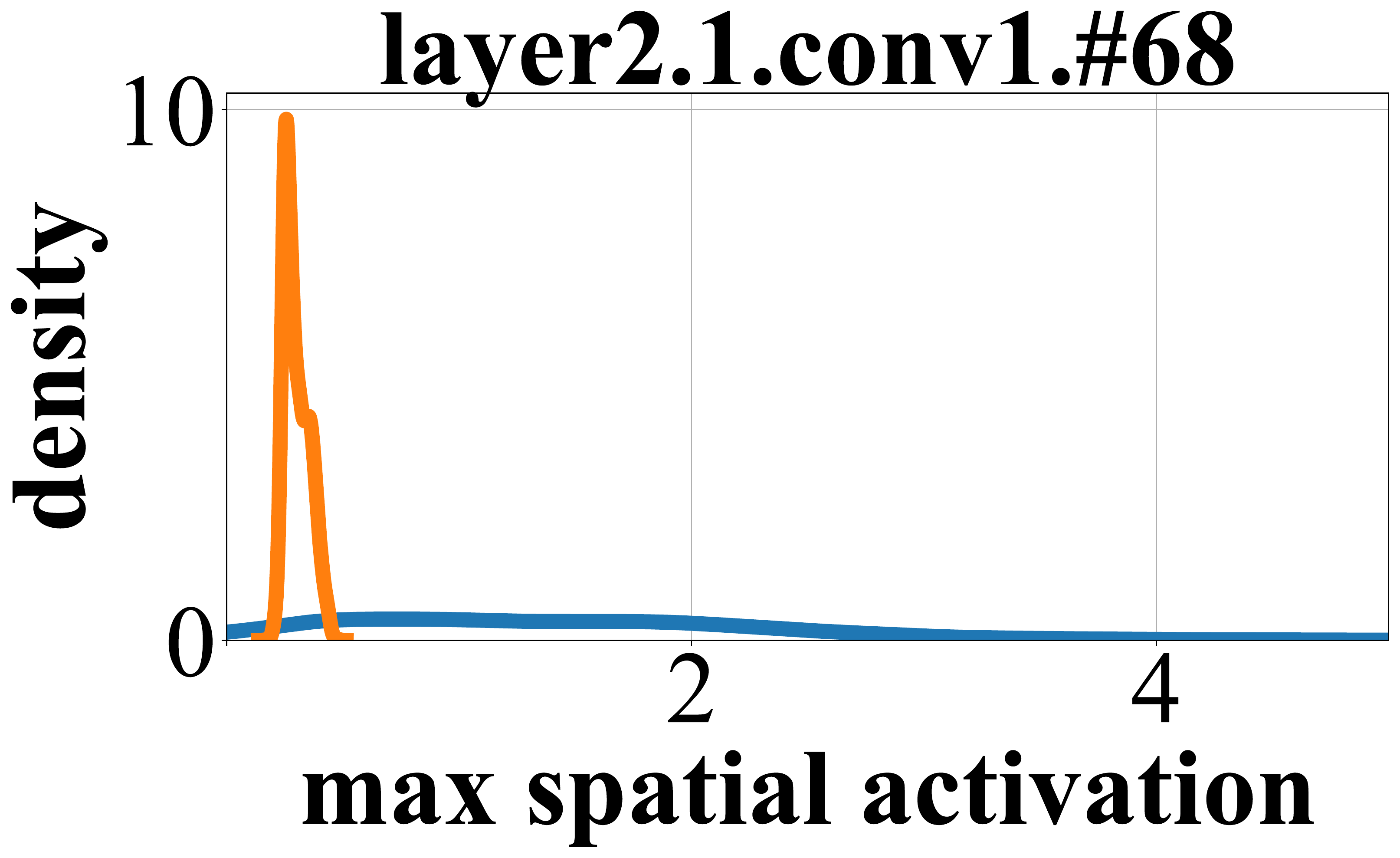} &
     \includegraphics[width=0.13\linewidth]{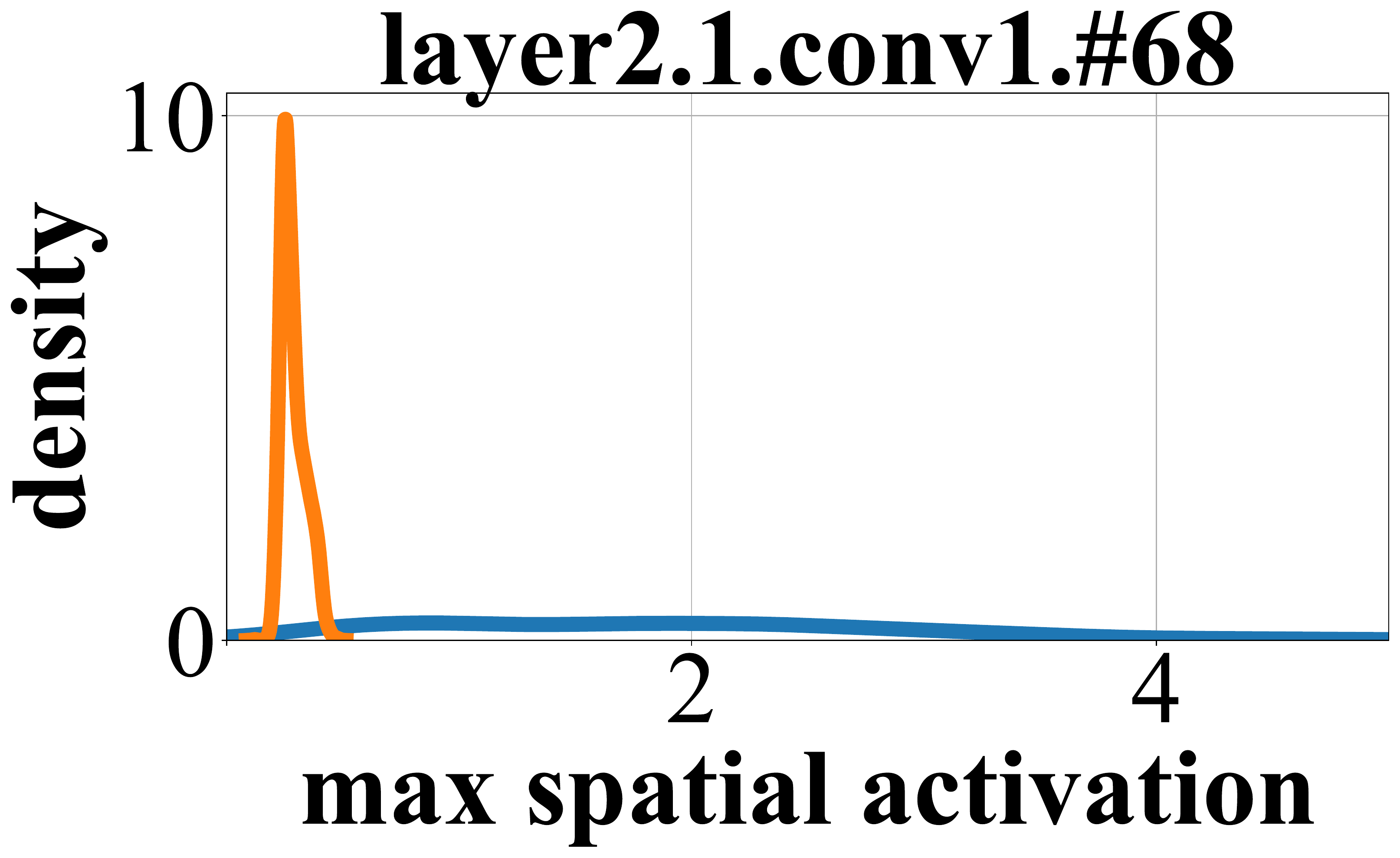} &
    \includegraphics[width=0.13\linewidth]{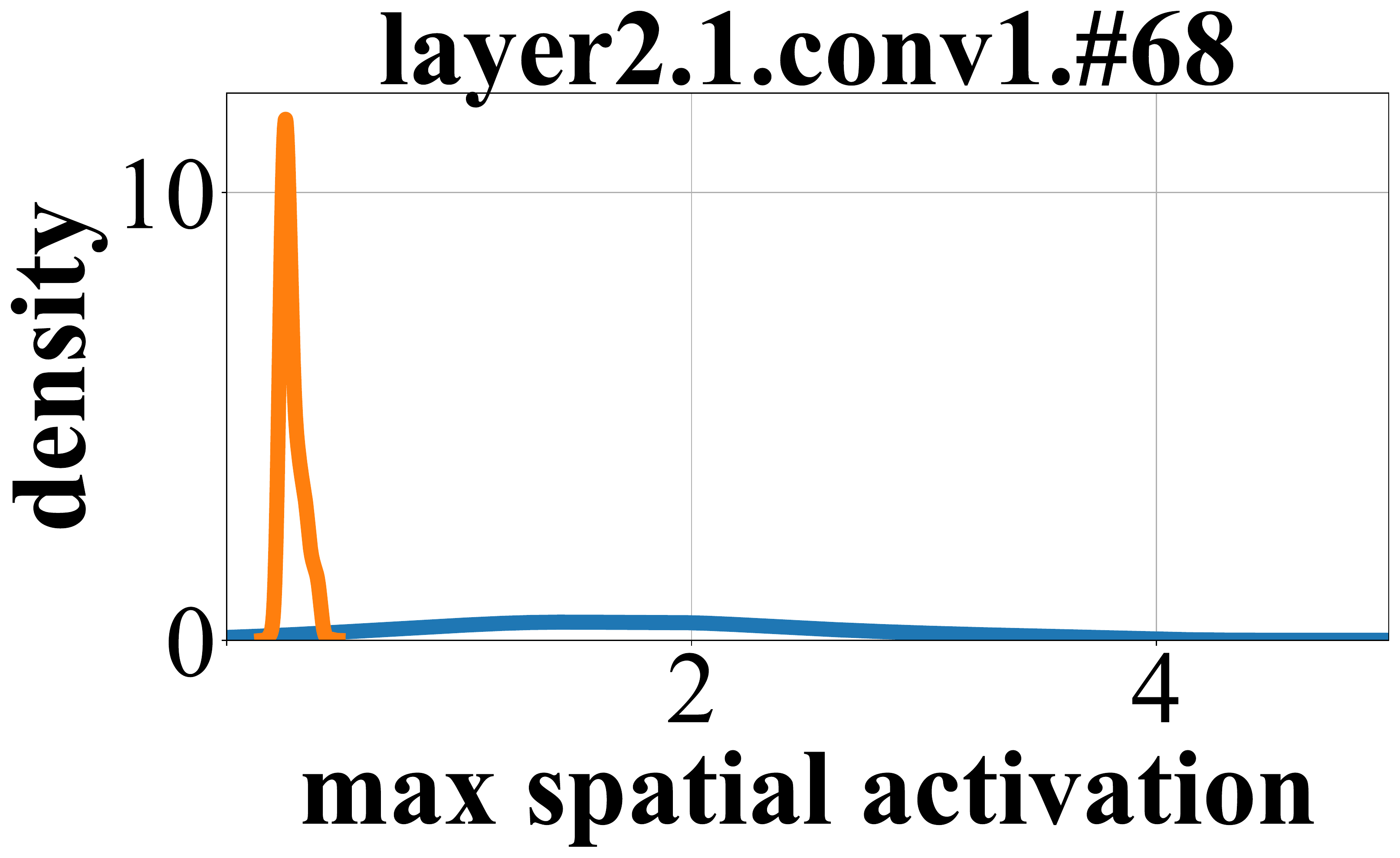} &
     \includegraphics[width=0.13\linewidth]{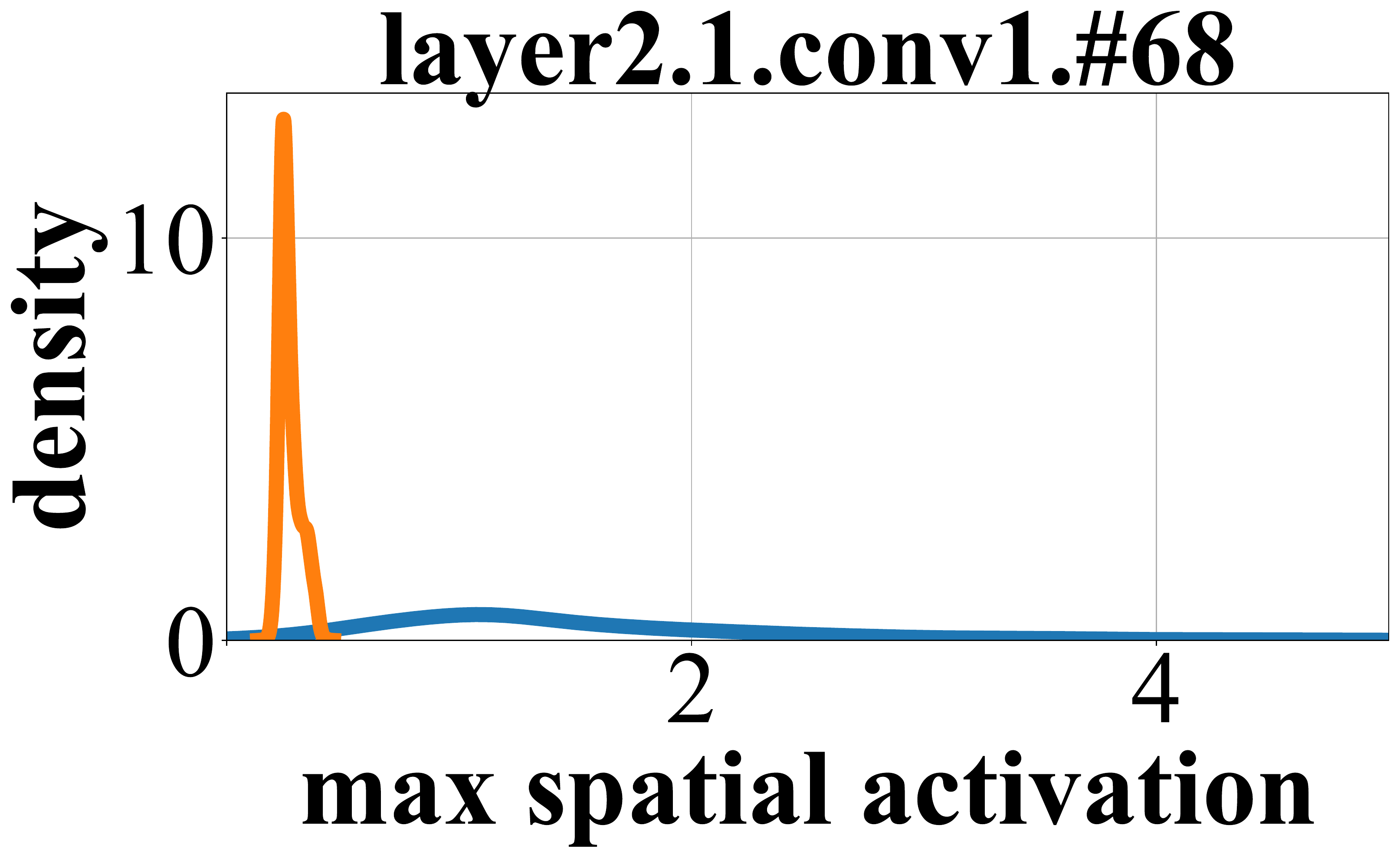} &
     \includegraphics[width=0.13\linewidth]{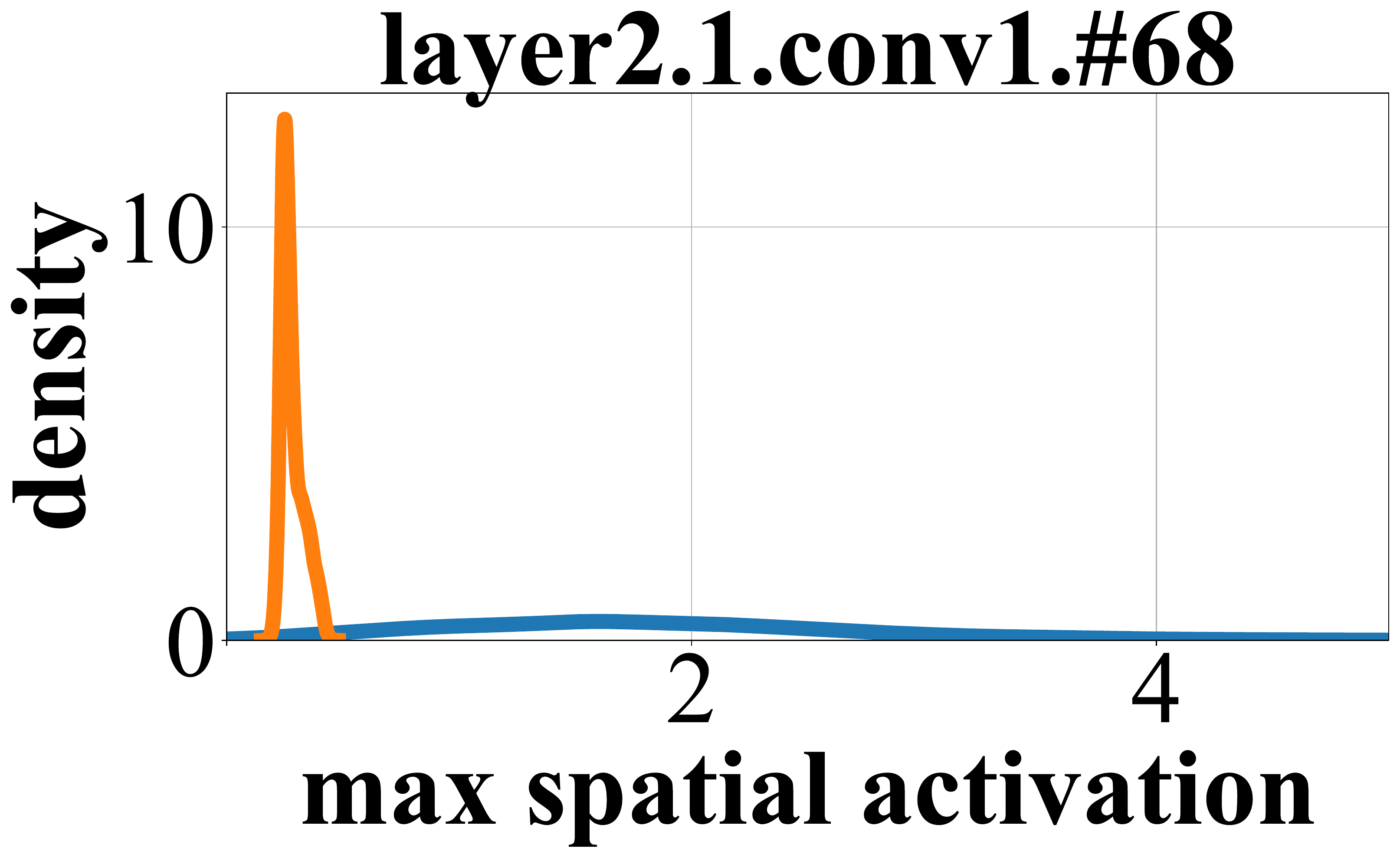}
    \\
    
    \includegraphics[width=0.13\linewidth]{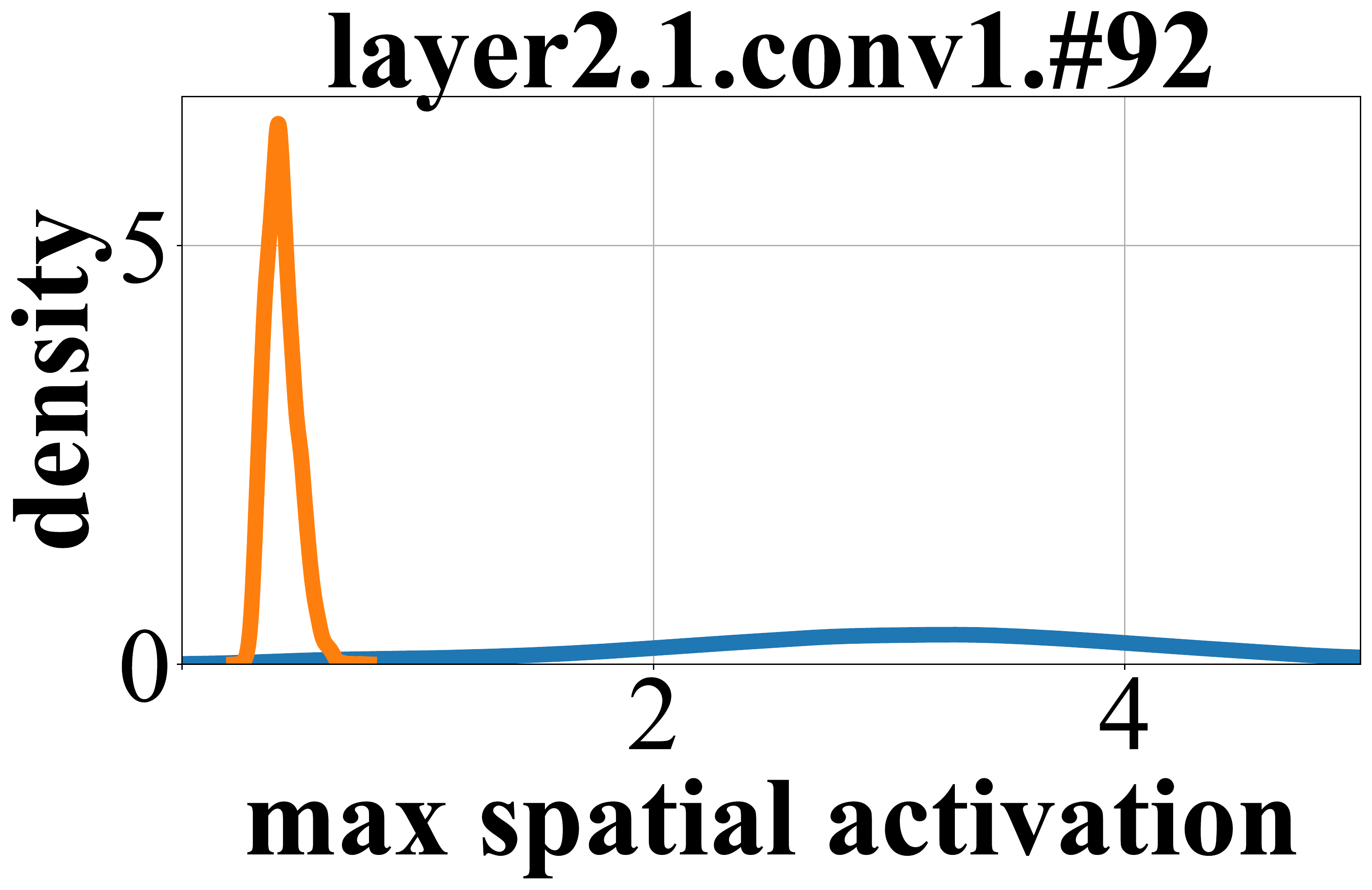} &
    \includegraphics[width=0.13\linewidth]{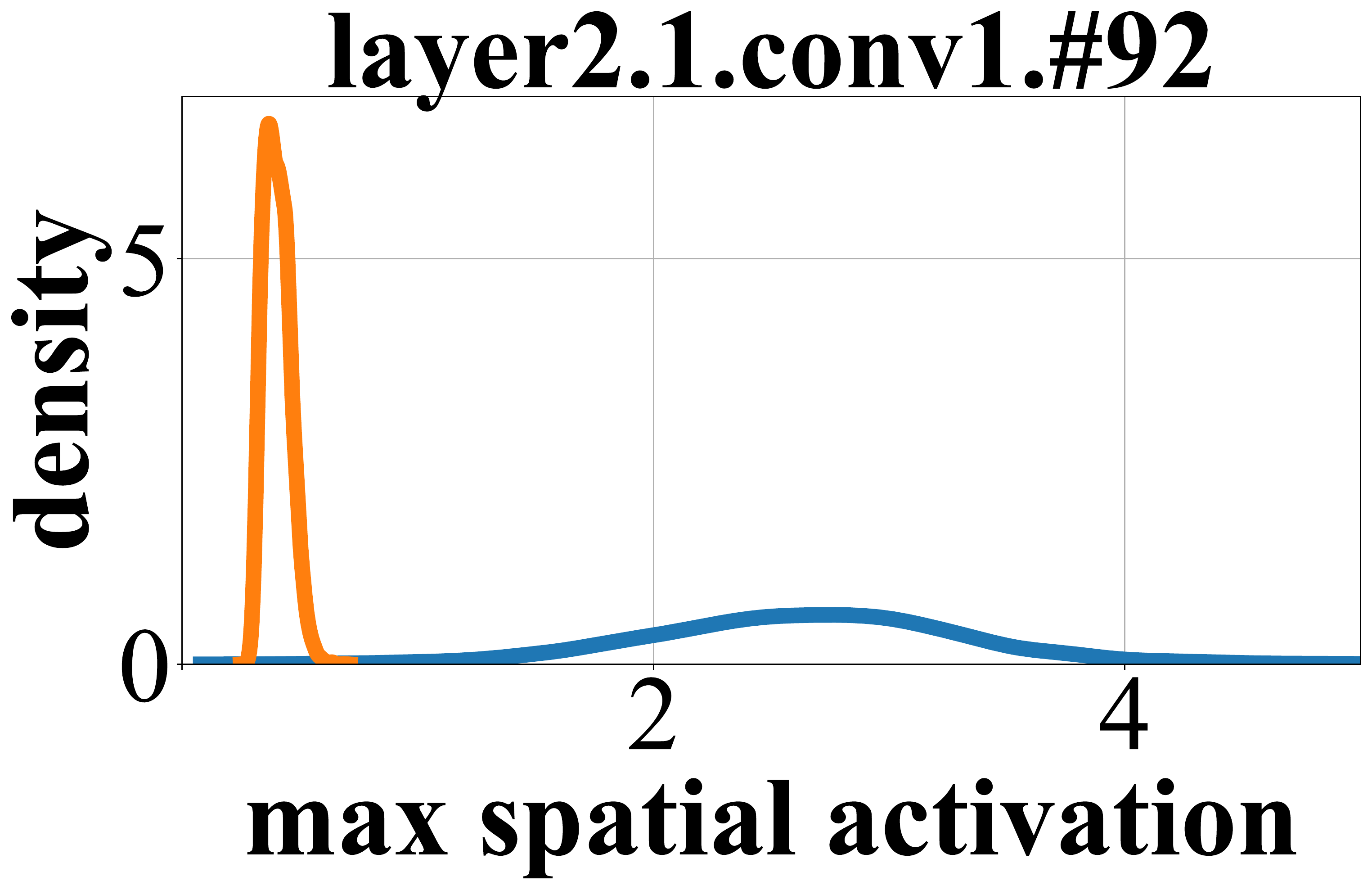} &
    \includegraphics[width=0.13\linewidth]{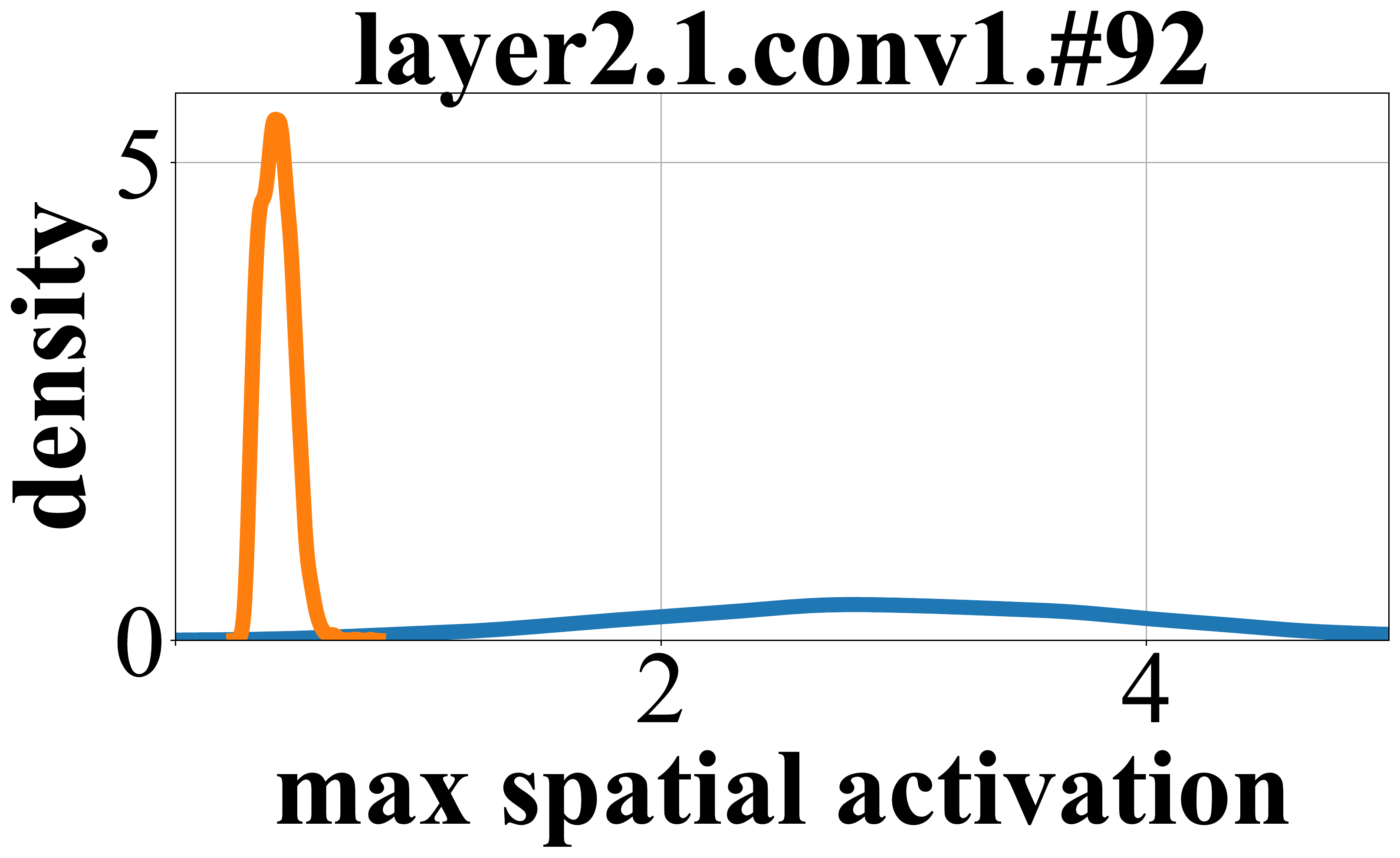} &
     \includegraphics[width=0.13\linewidth]{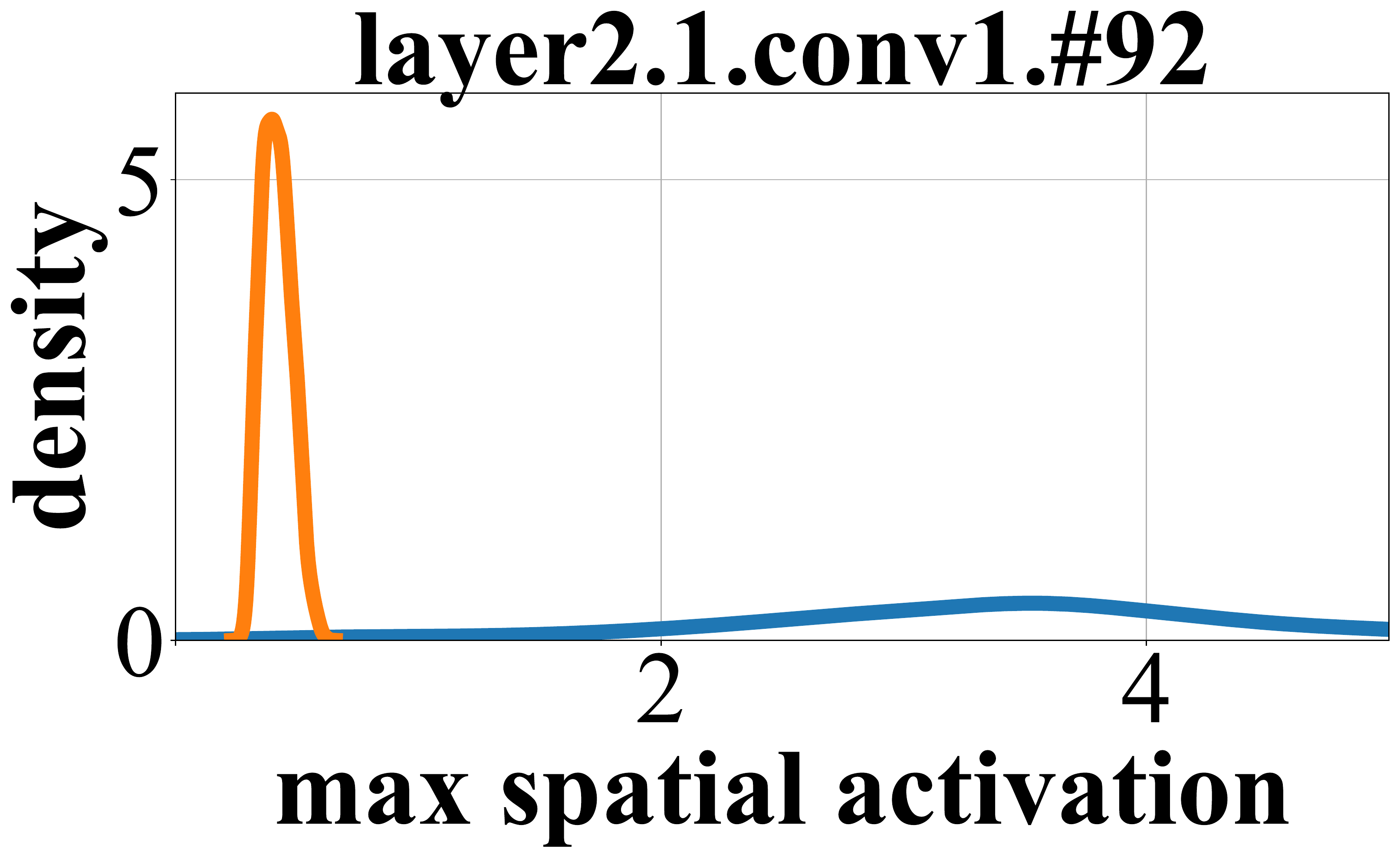} &
    \includegraphics[width=0.13\linewidth]{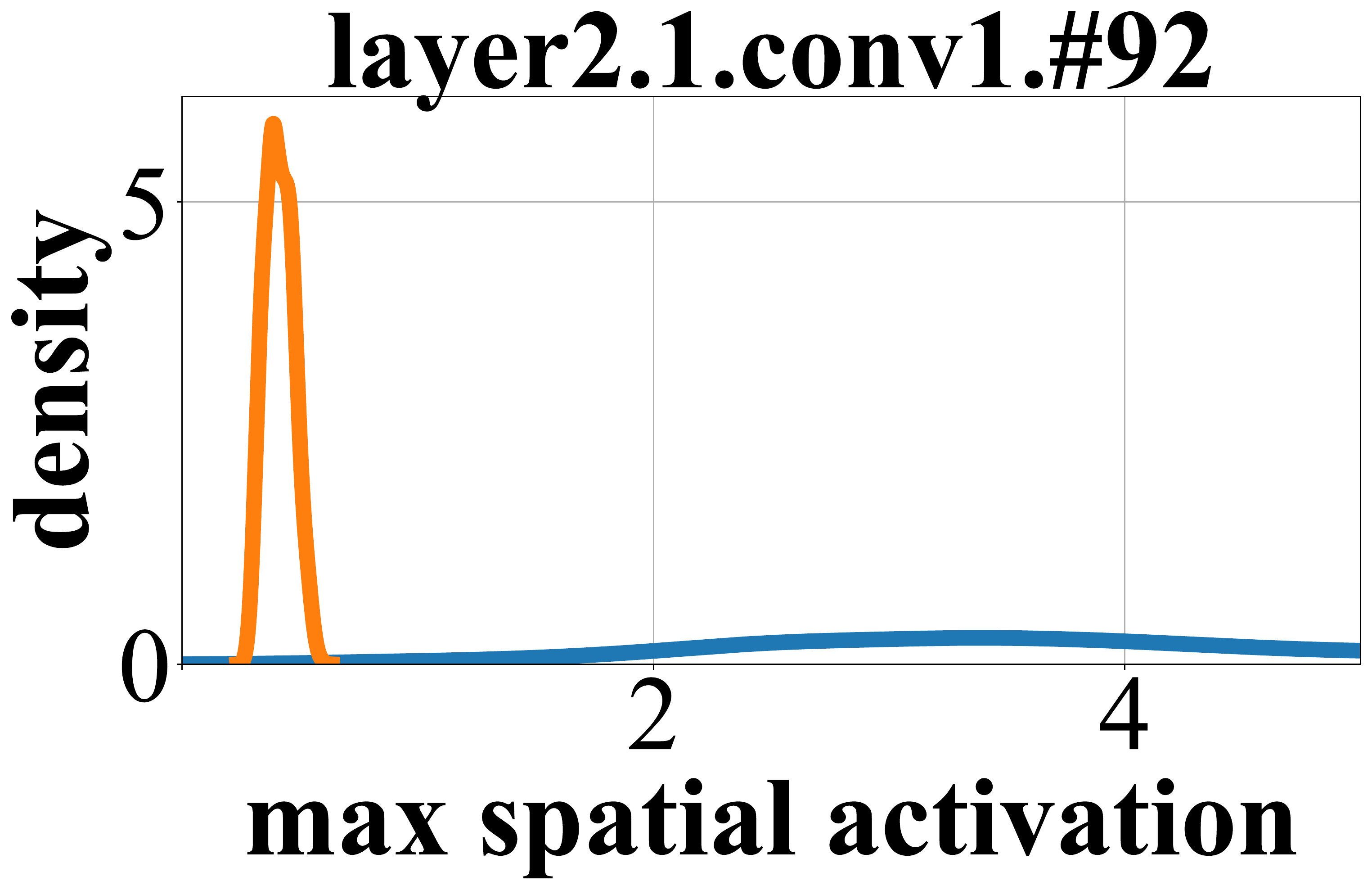} &
     \includegraphics[width=0.13\linewidth]{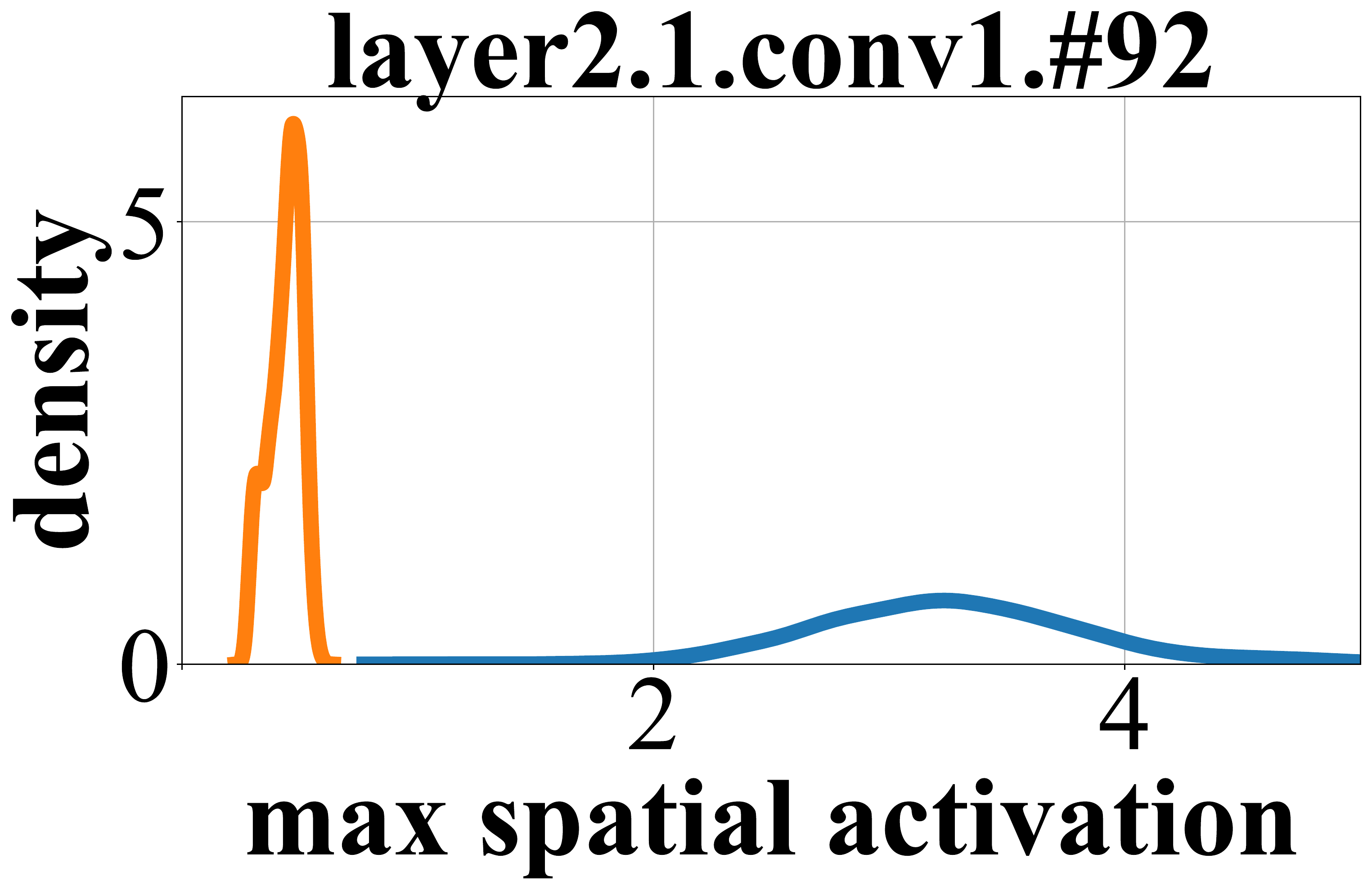} &
     \includegraphics[width=0.13\linewidth]{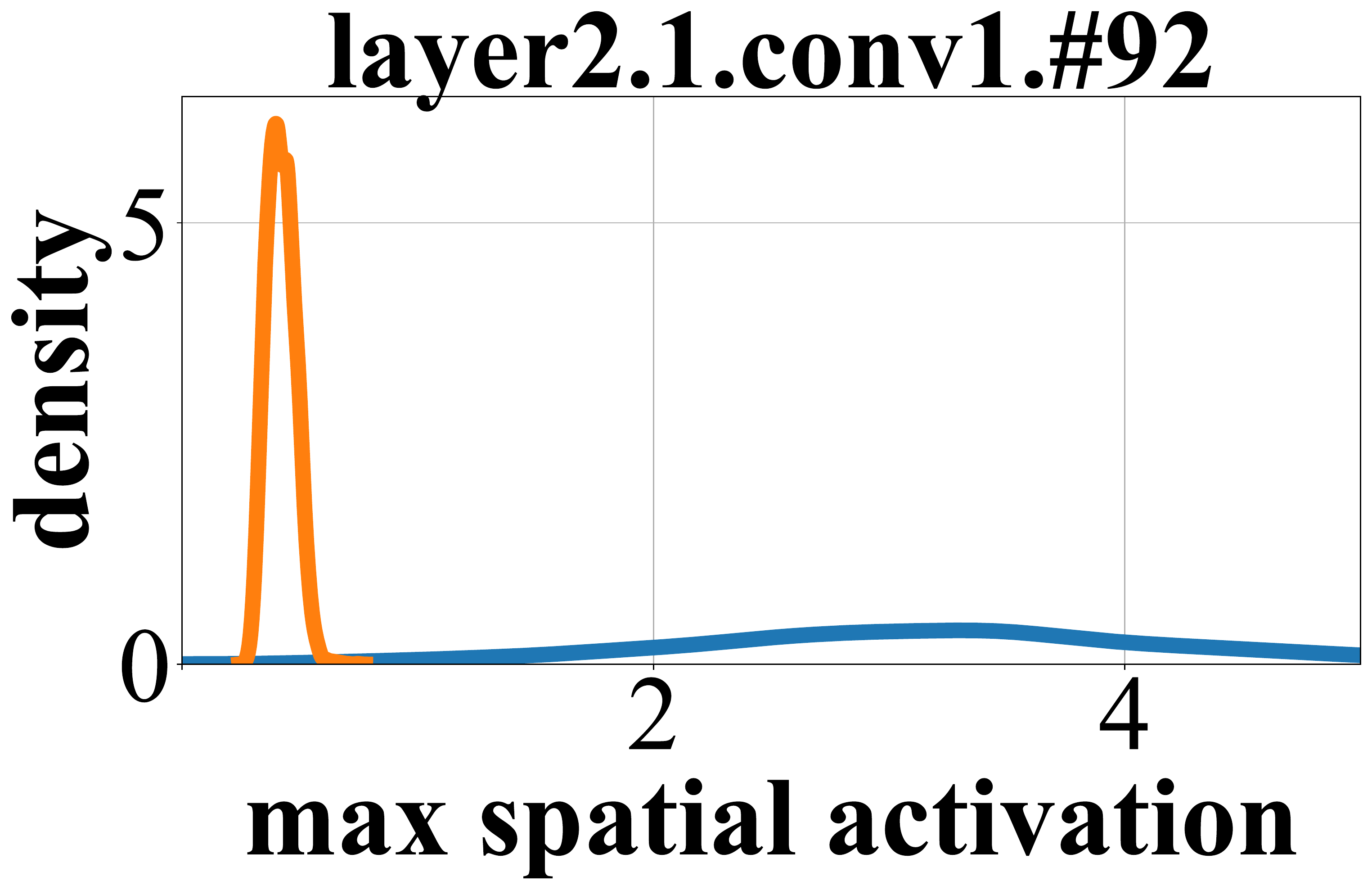}
    \\
    
     \includegraphics[width=0.13\linewidth]{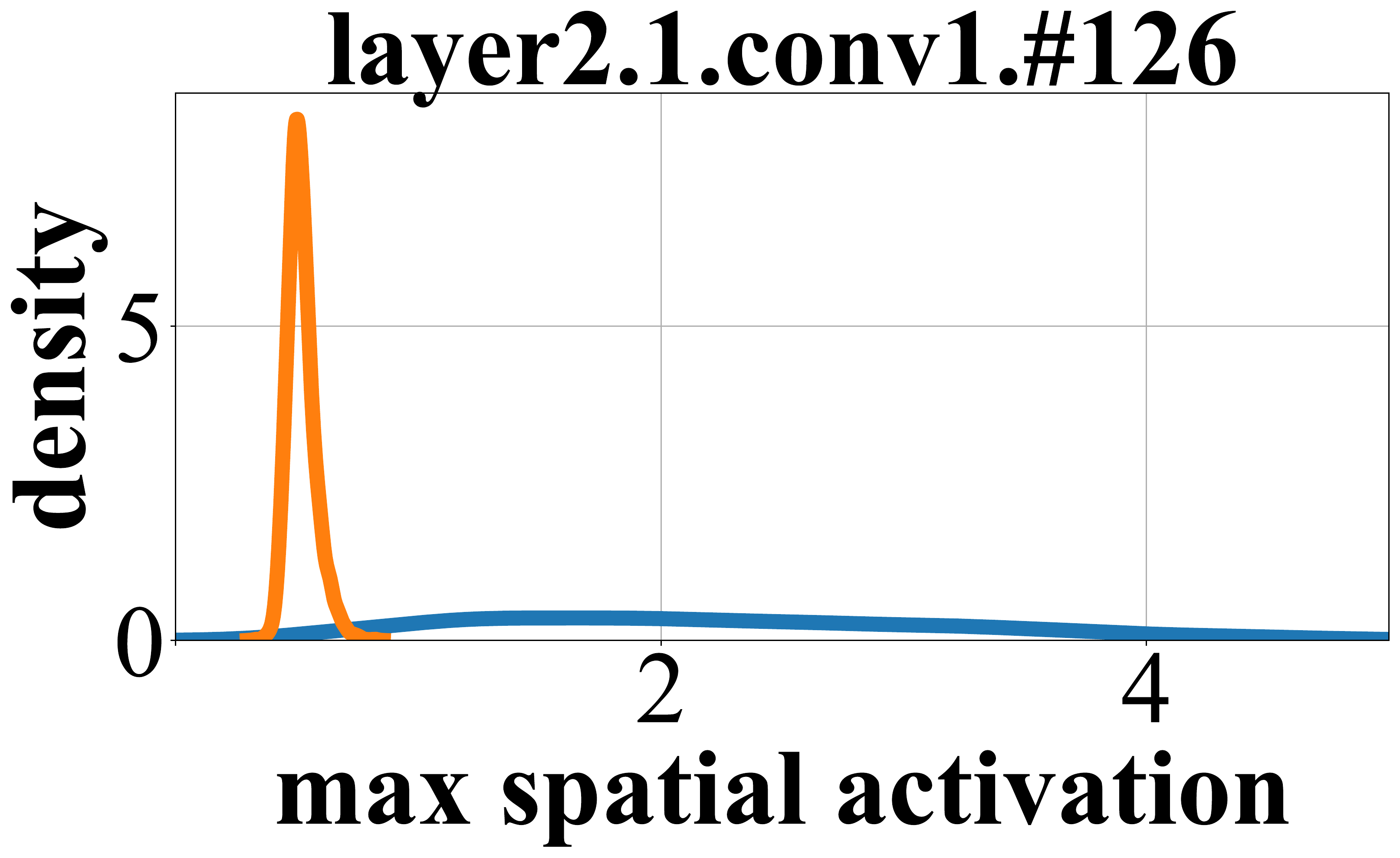} &
    \includegraphics[width=0.13\linewidth]{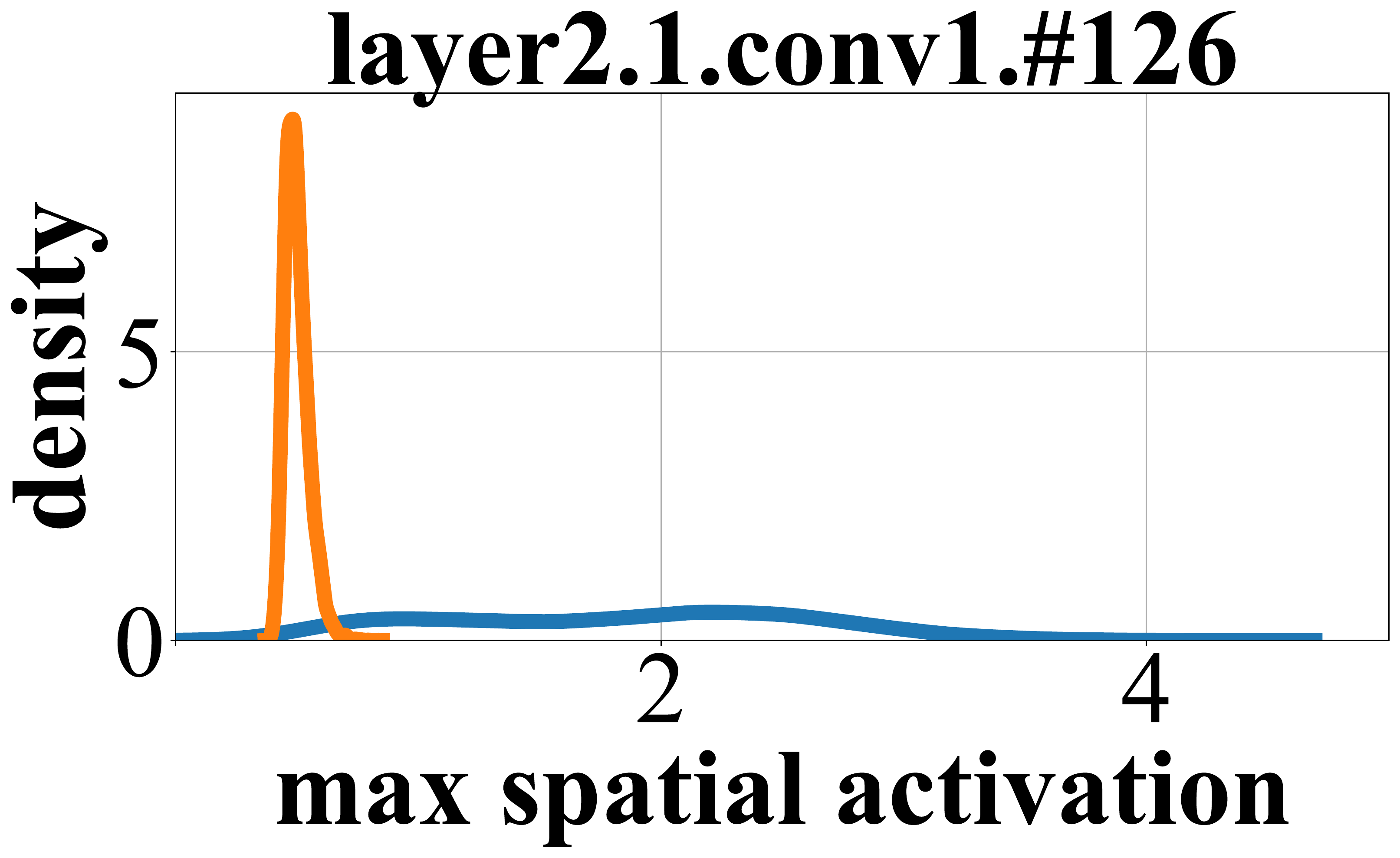} &
    \includegraphics[width=0.13\linewidth]{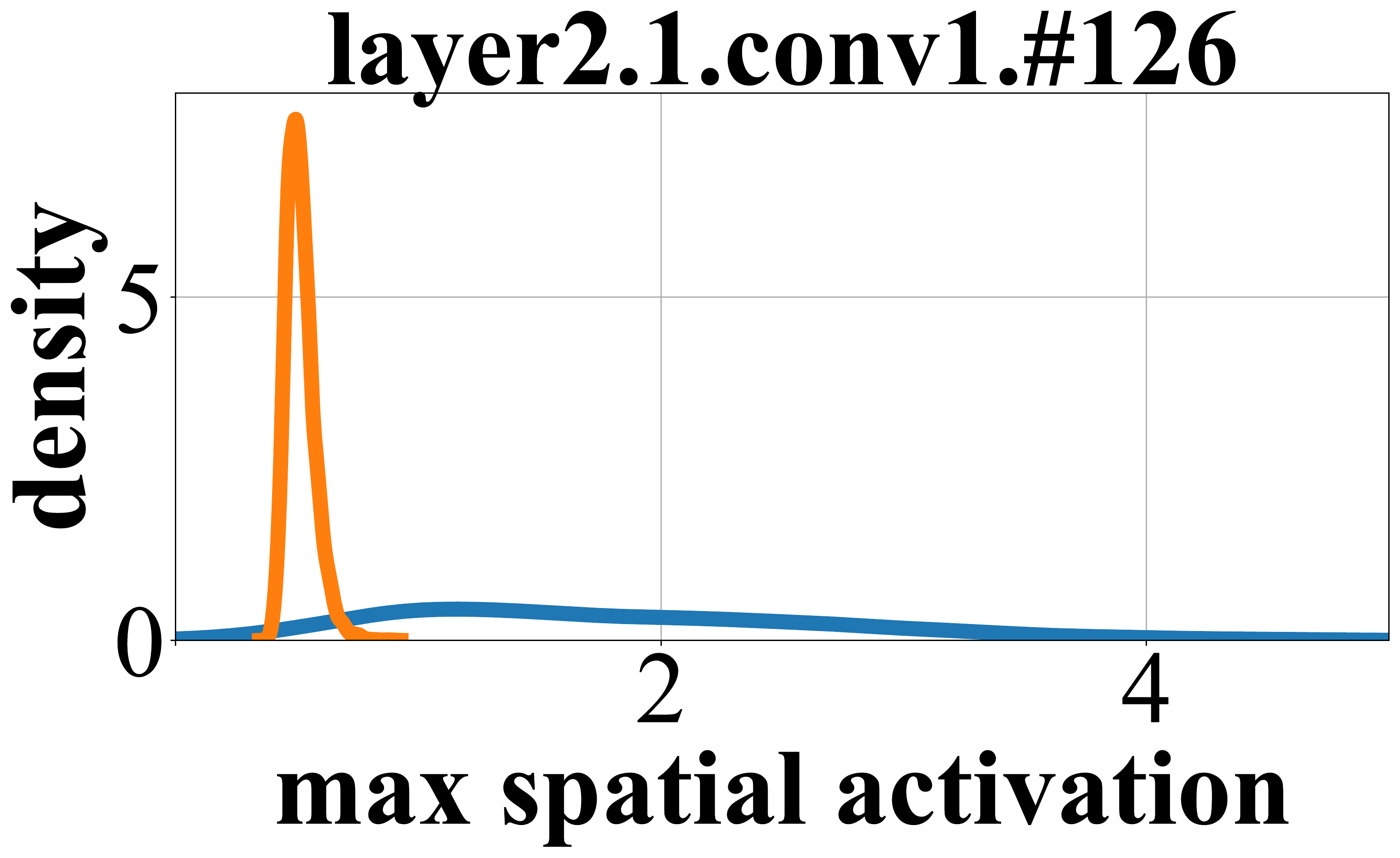} &
     \includegraphics[width=0.13\linewidth]{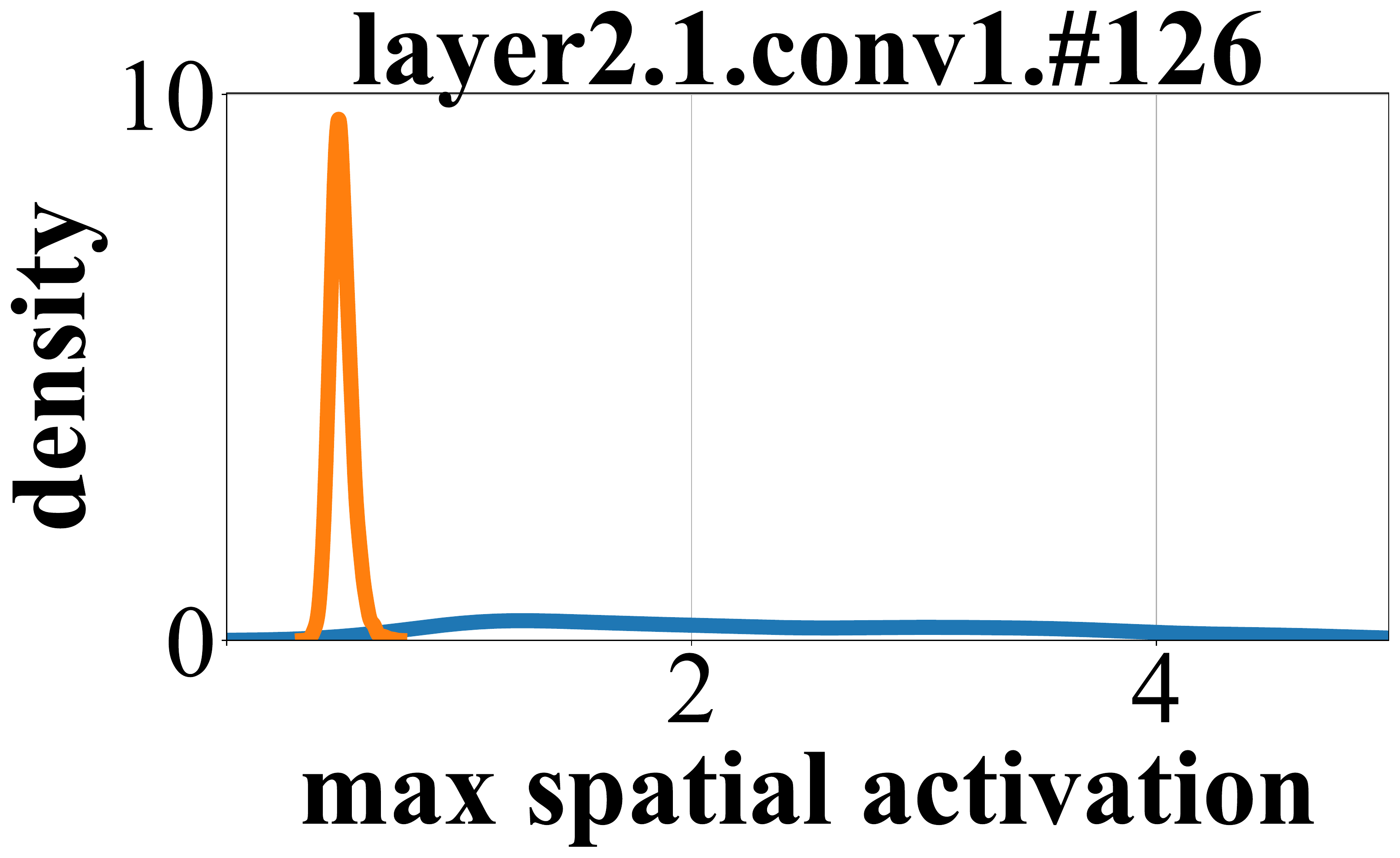} &
    \includegraphics[width=0.13\linewidth]{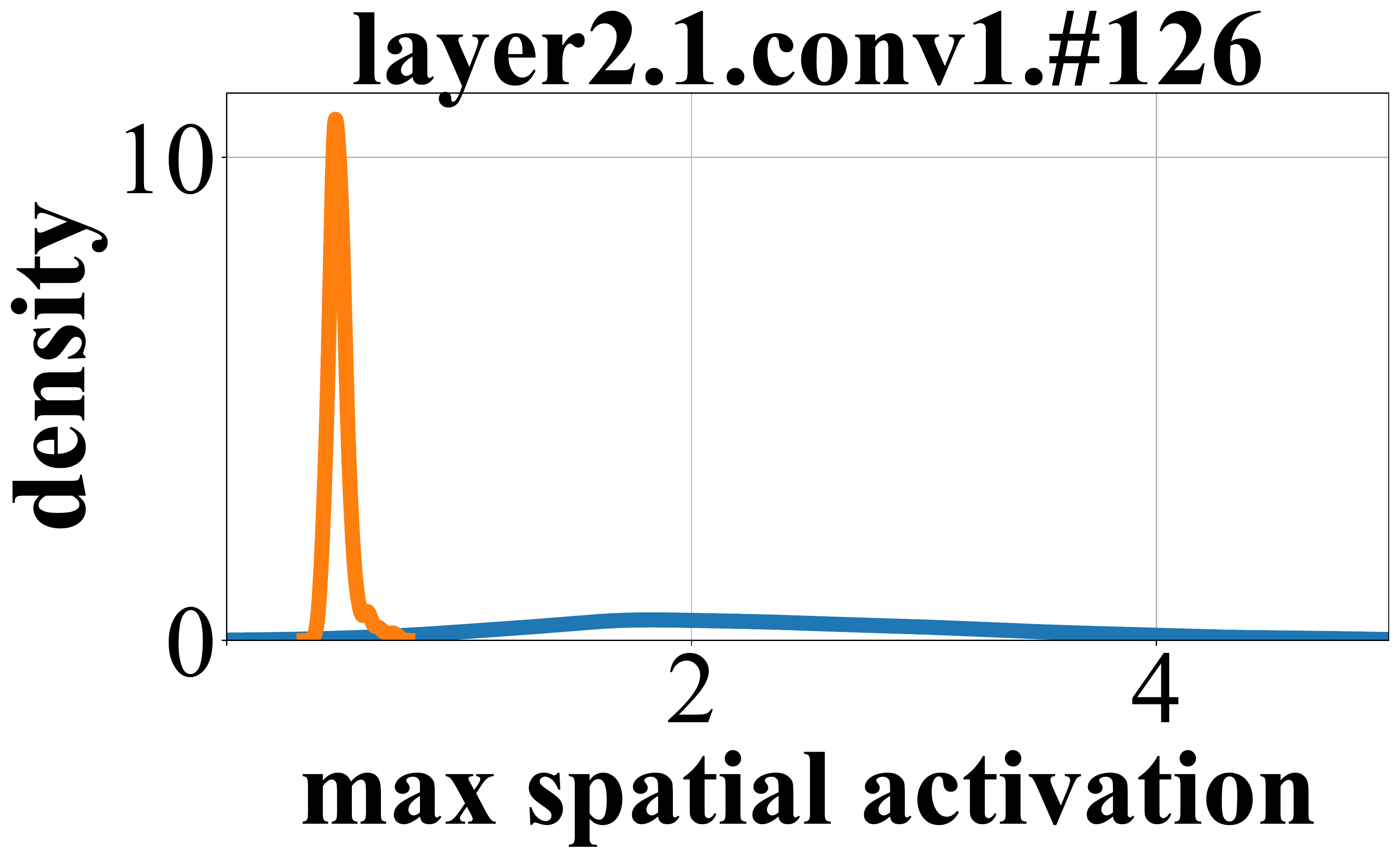} &
     \includegraphics[width=0.13\linewidth]{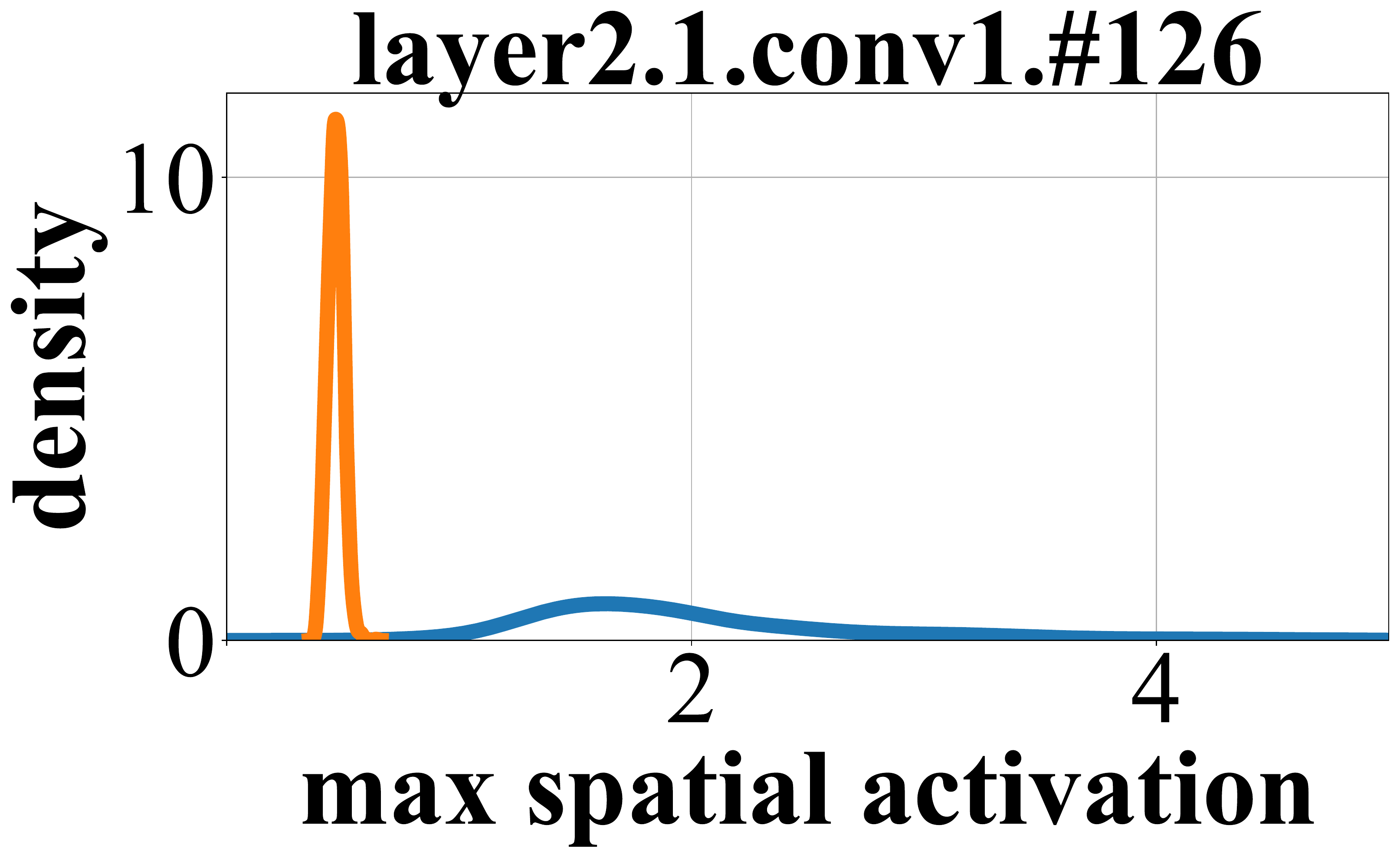} &
     \includegraphics[width=0.13\linewidth]{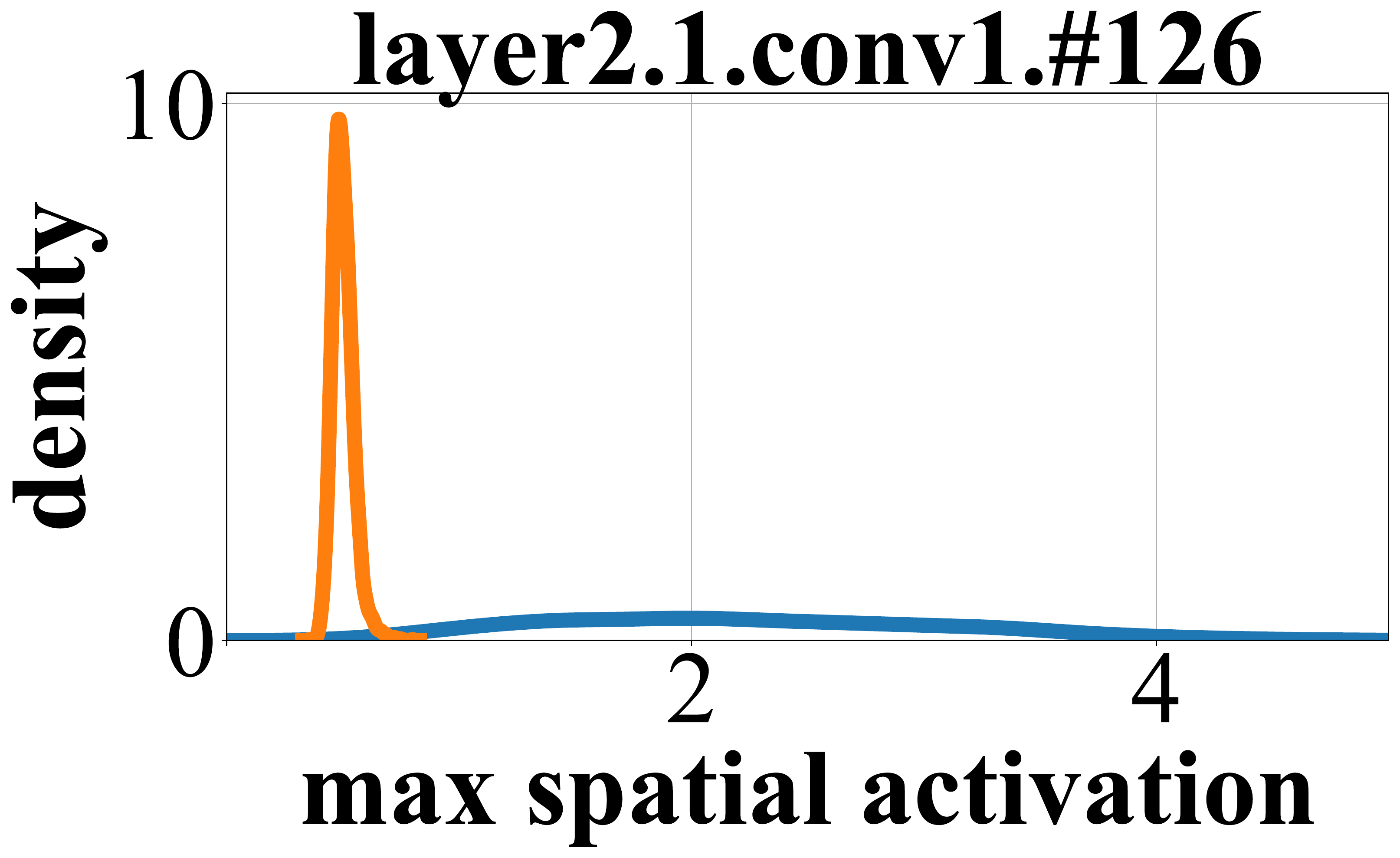}
    \\
    
    \hline

    \multicolumn{7}{c}{\bf EfficientNet-B0}\\
    \includegraphics[width=0.13\linewidth]{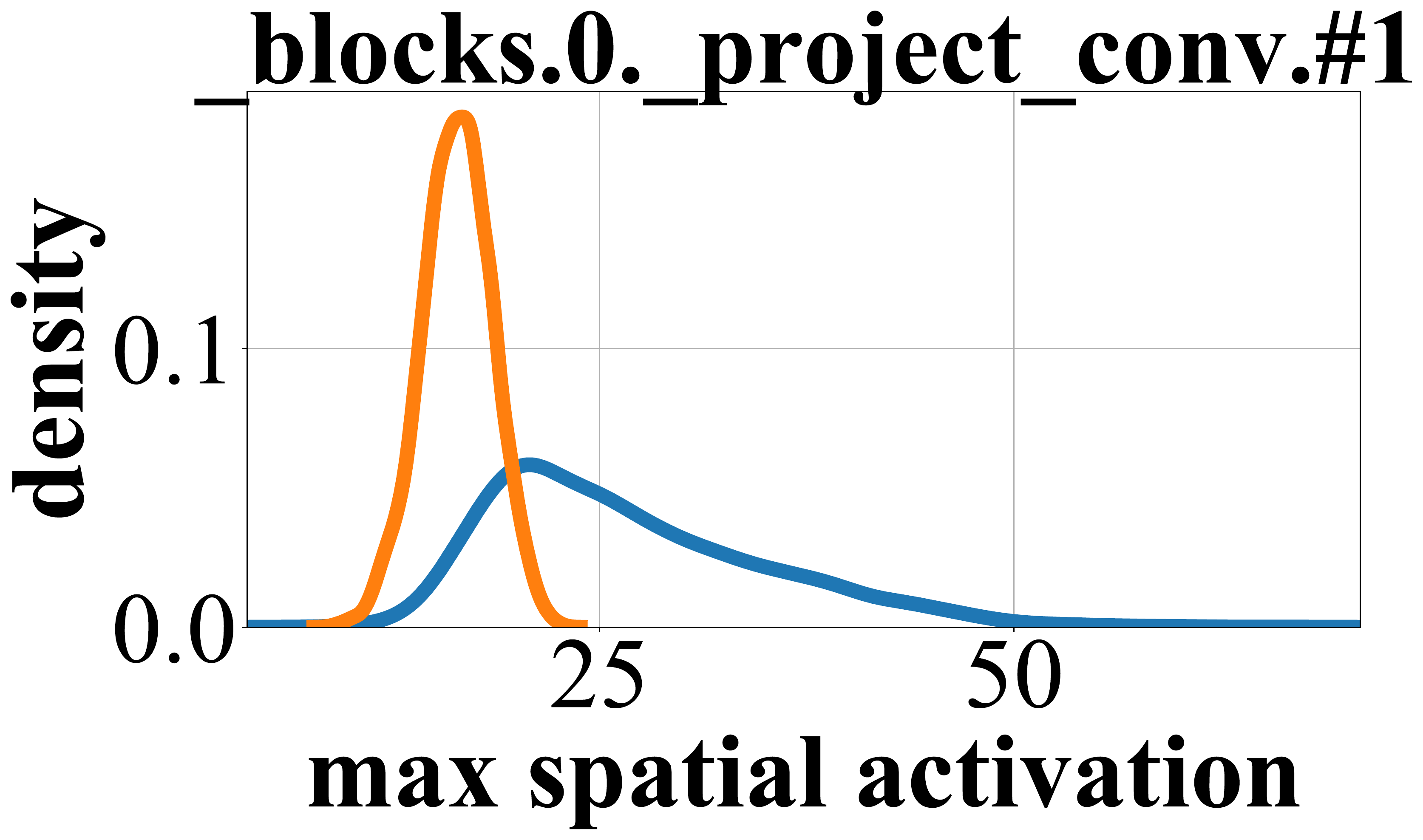} &
    \includegraphics[width=0.13\linewidth]{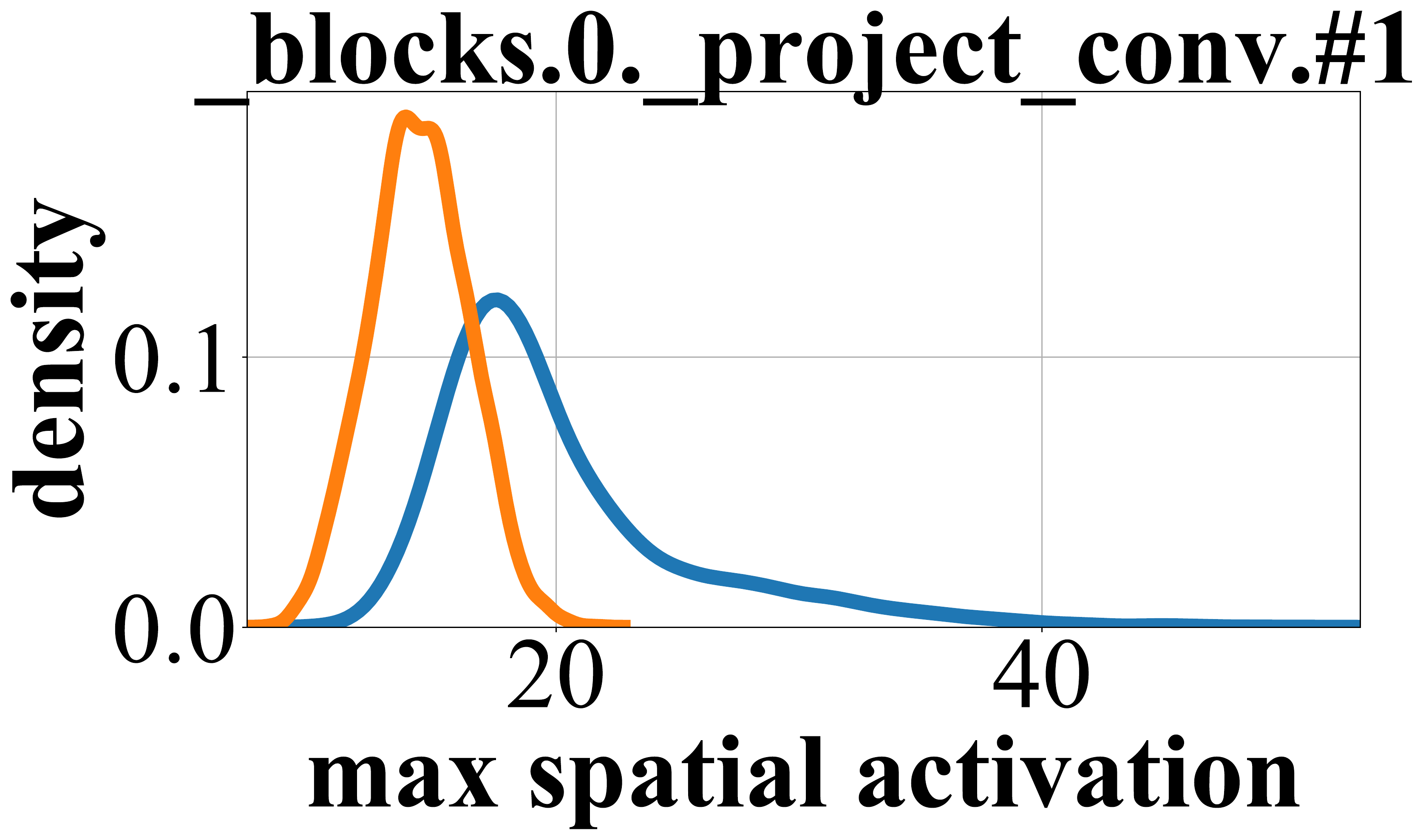} &
    \includegraphics[width=0.13\linewidth]{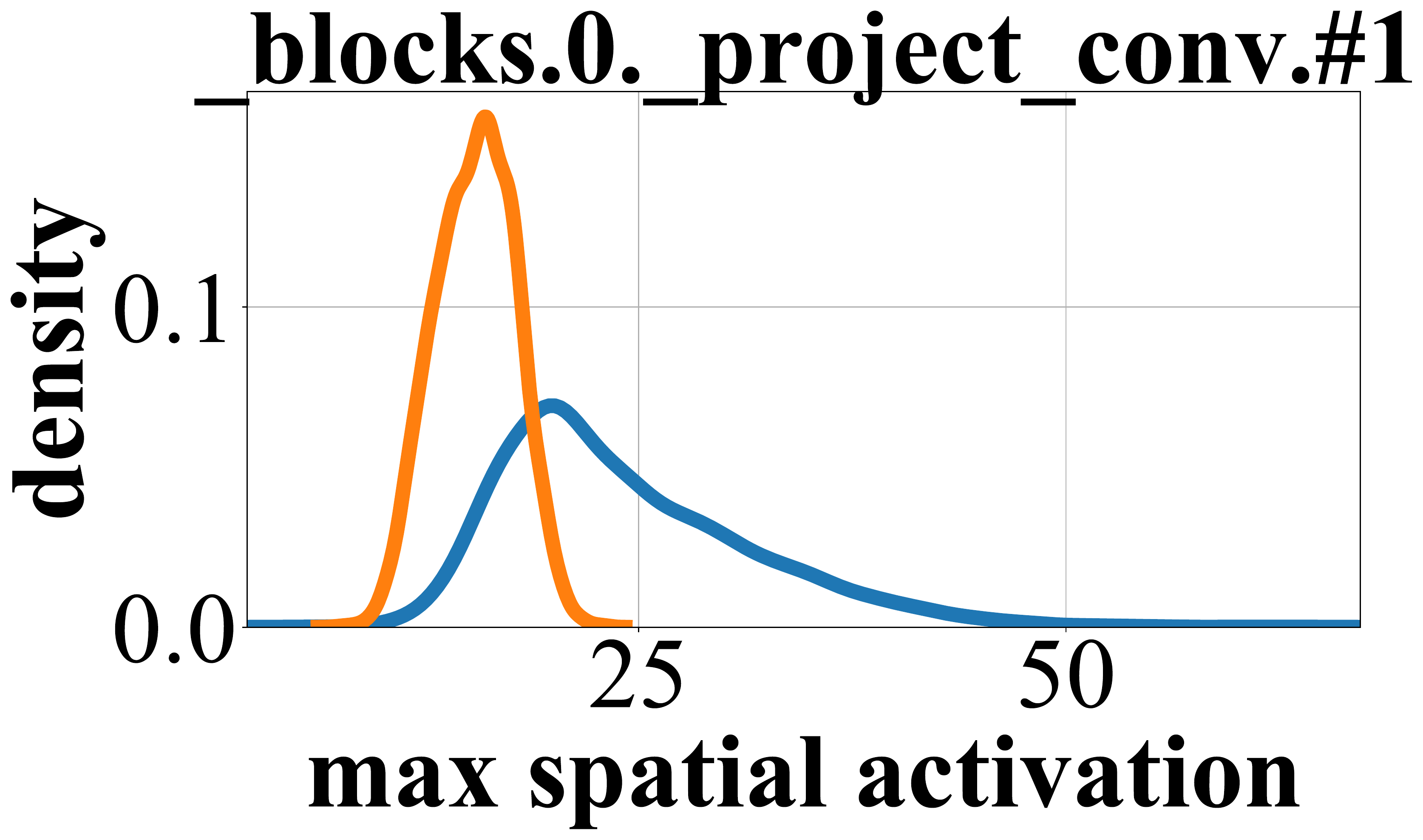} &
    \includegraphics[width=0.13\linewidth]{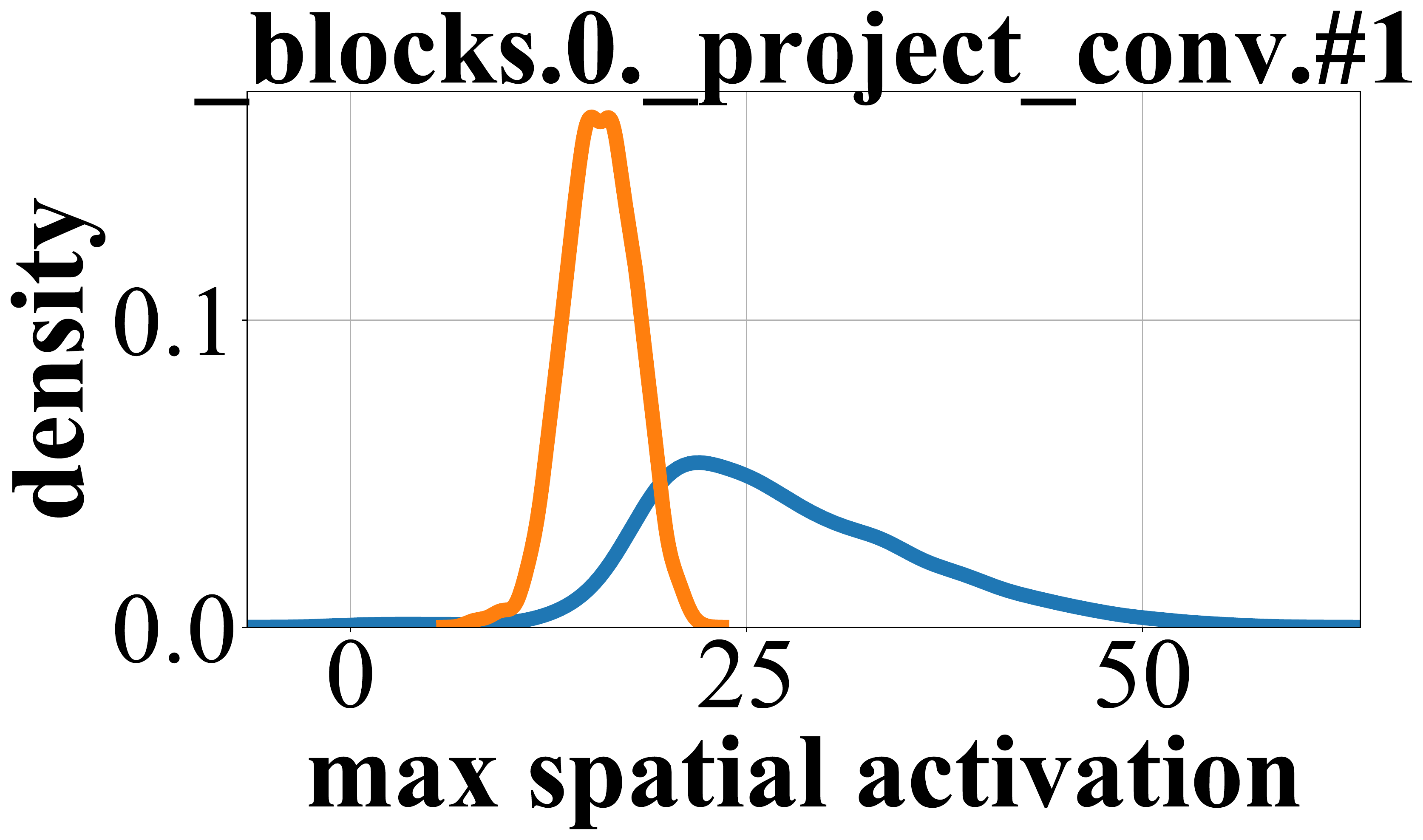} &
    \includegraphics[width=0.13\linewidth]{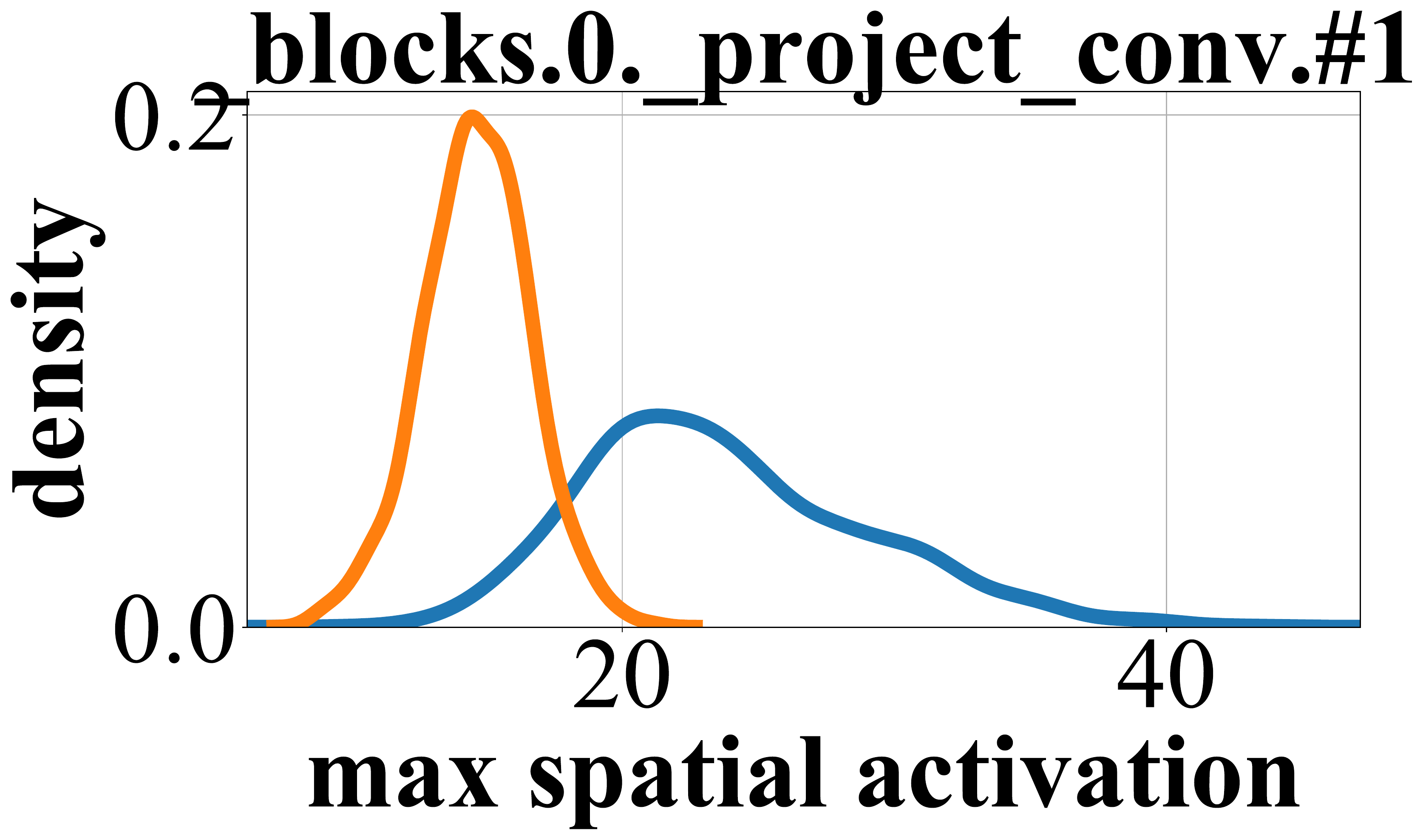} &
     \includegraphics[width=0.13\linewidth]{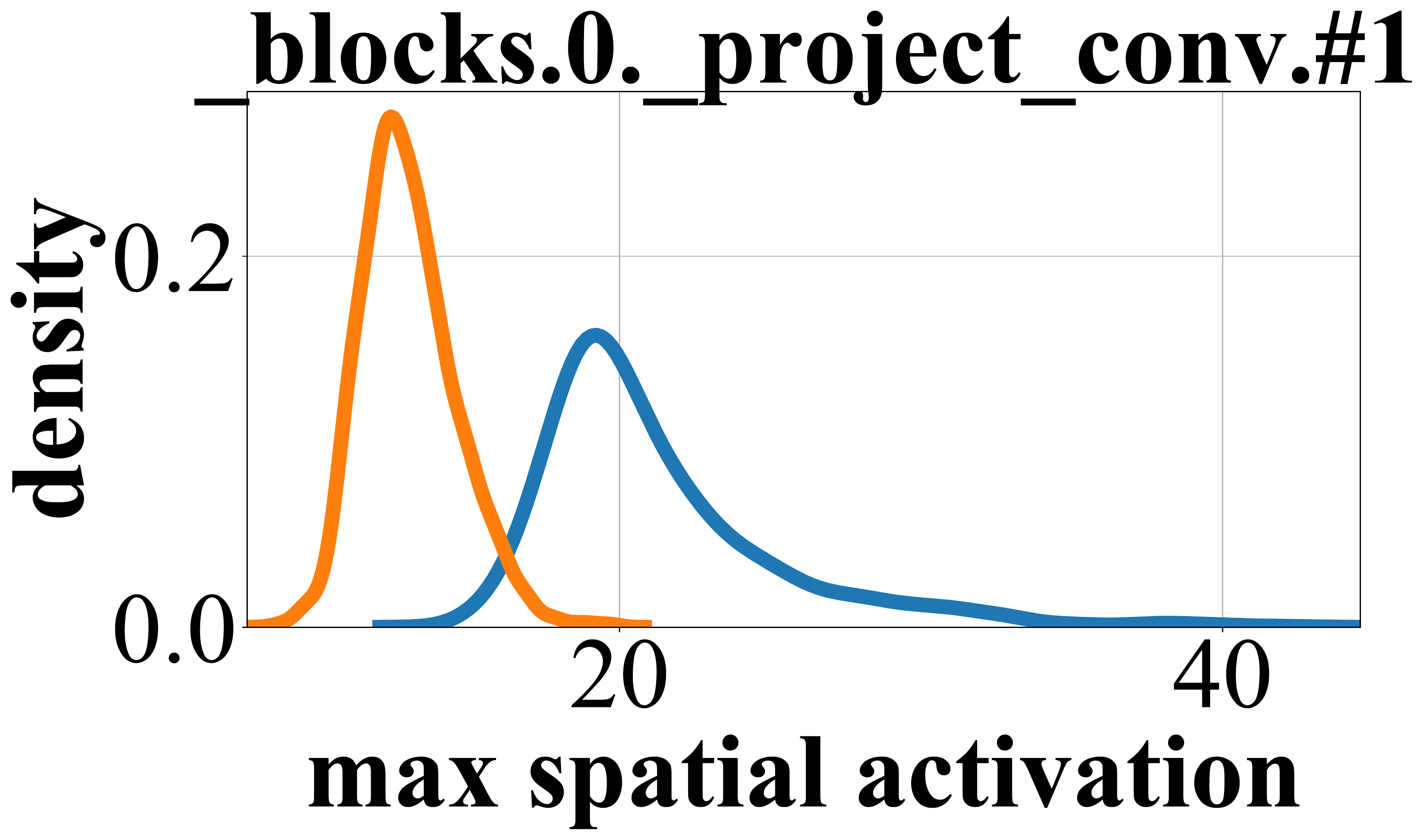} &
    \includegraphics[width=0.13\linewidth]{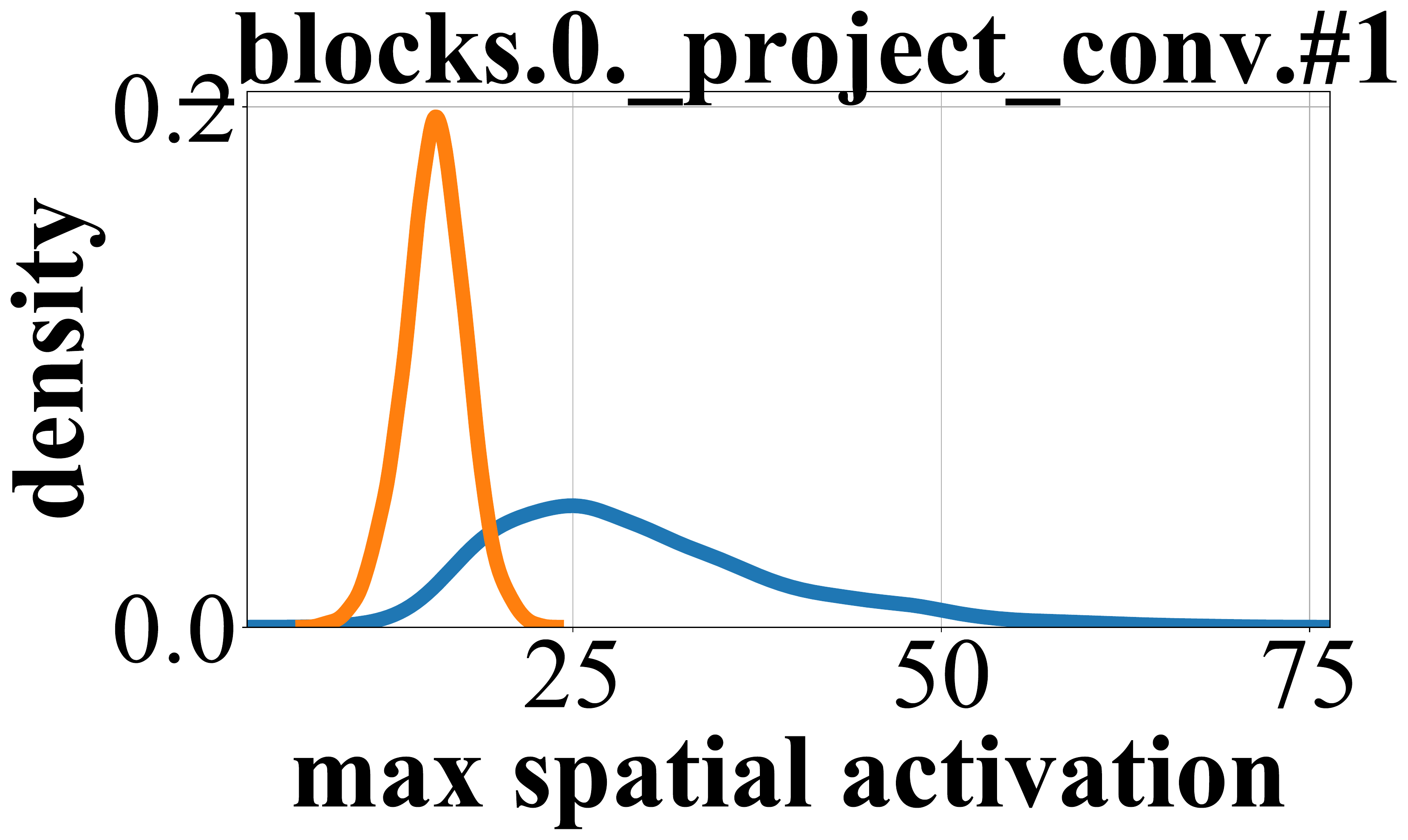}
    \\
    
    \includegraphics[width=0.13\linewidth]{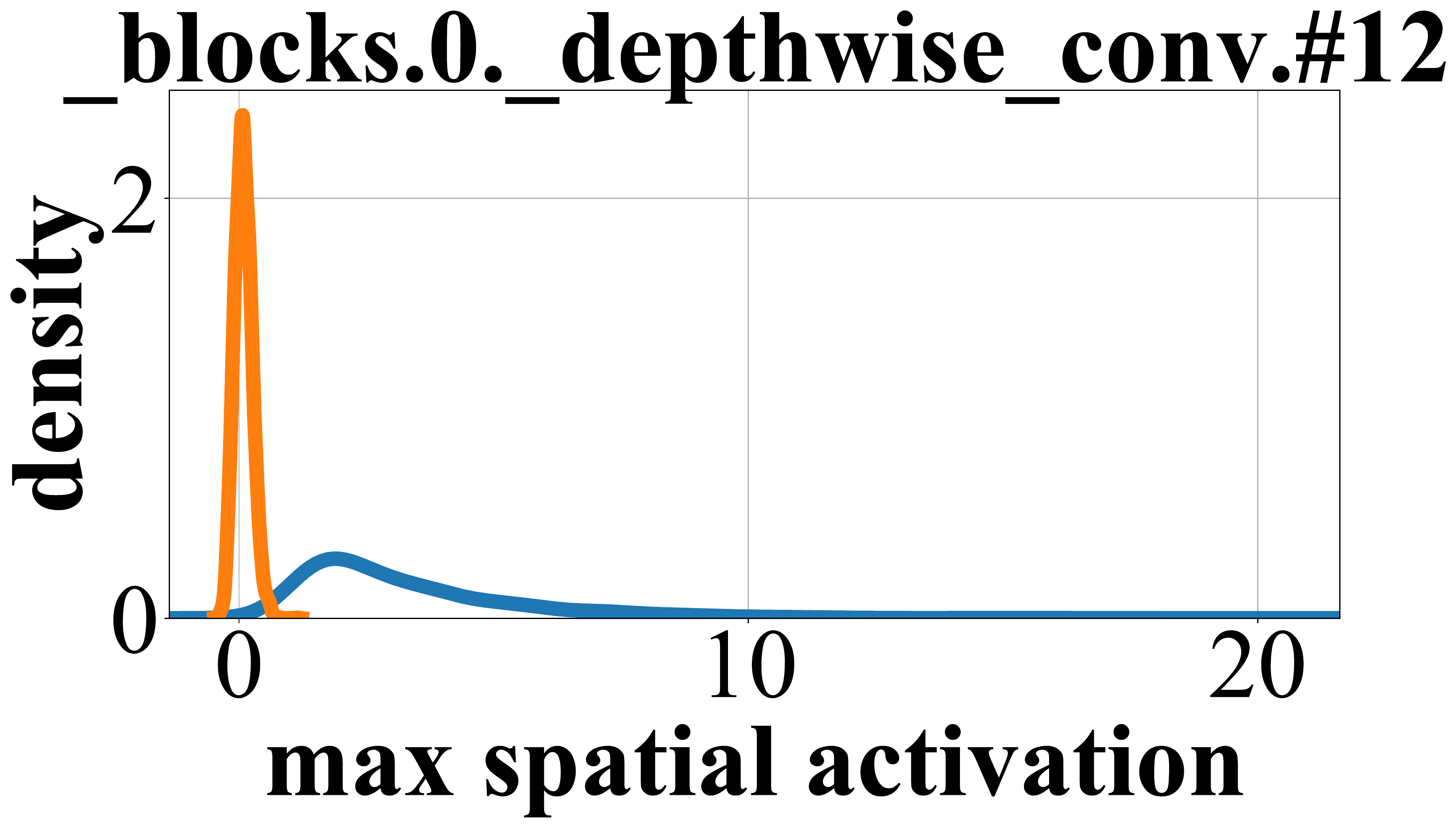} &
    \includegraphics[width=0.13\linewidth]{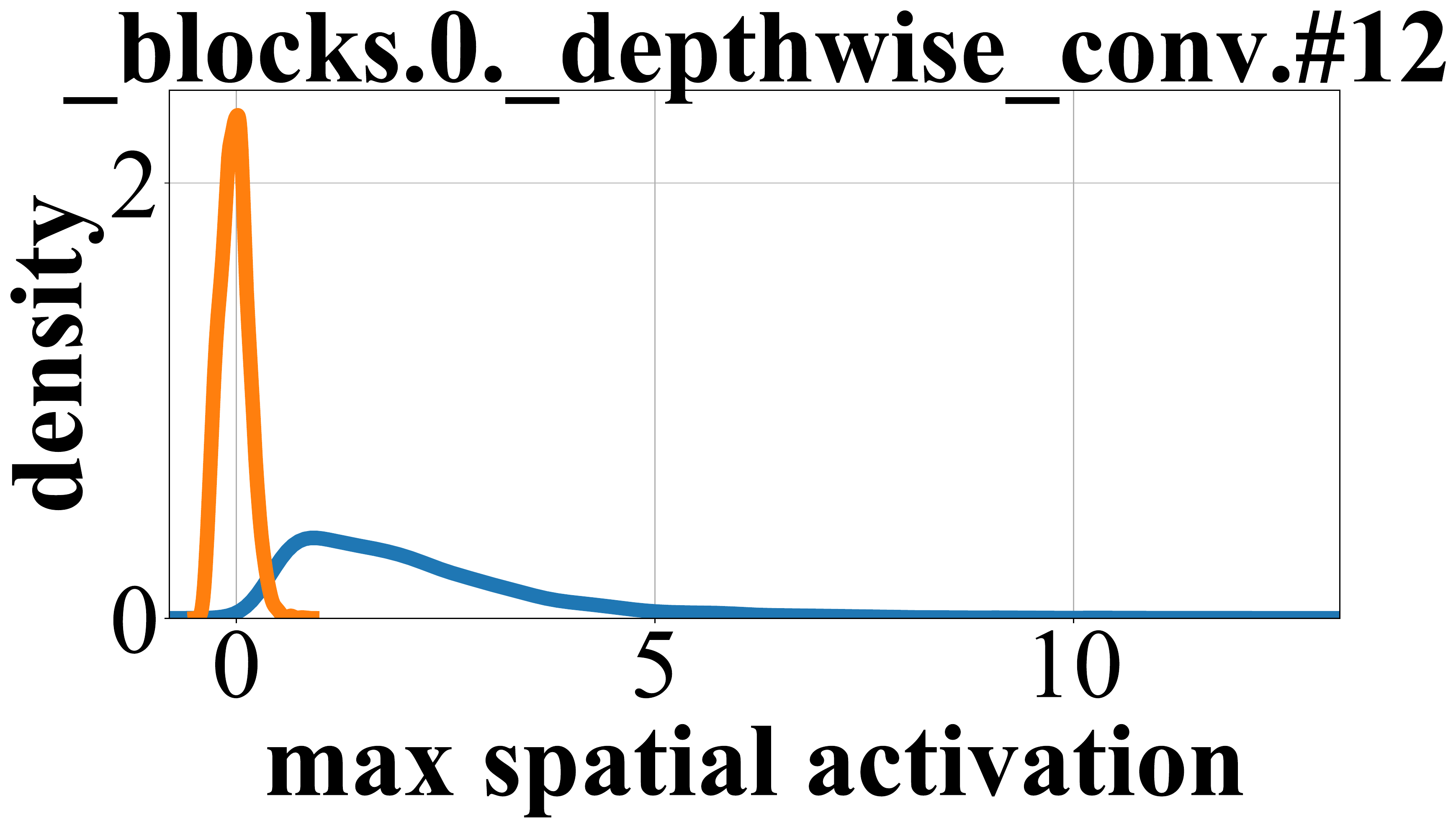} &
    \includegraphics[width=0.13\linewidth]{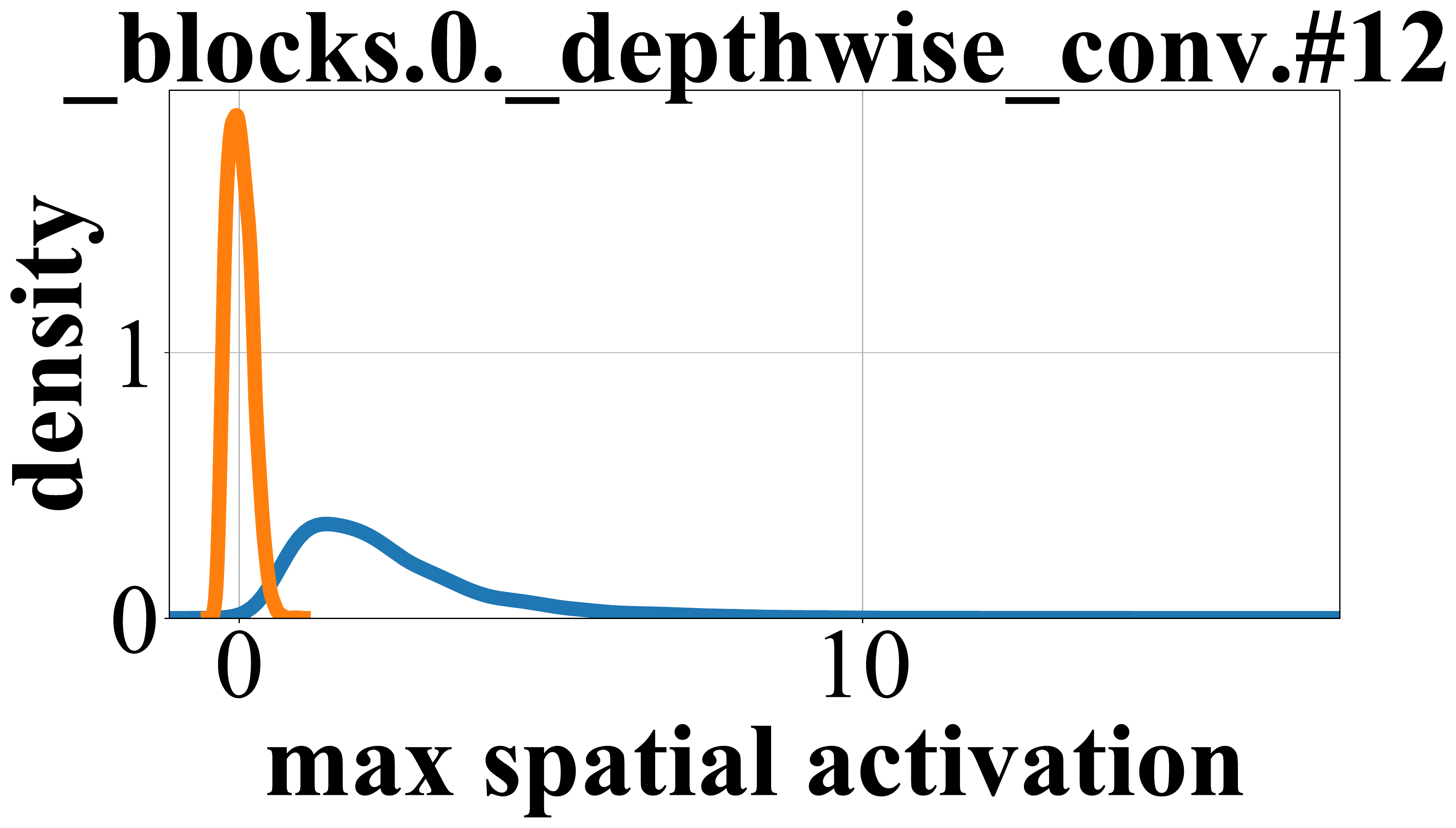} &
    \includegraphics[width=0.13\linewidth]{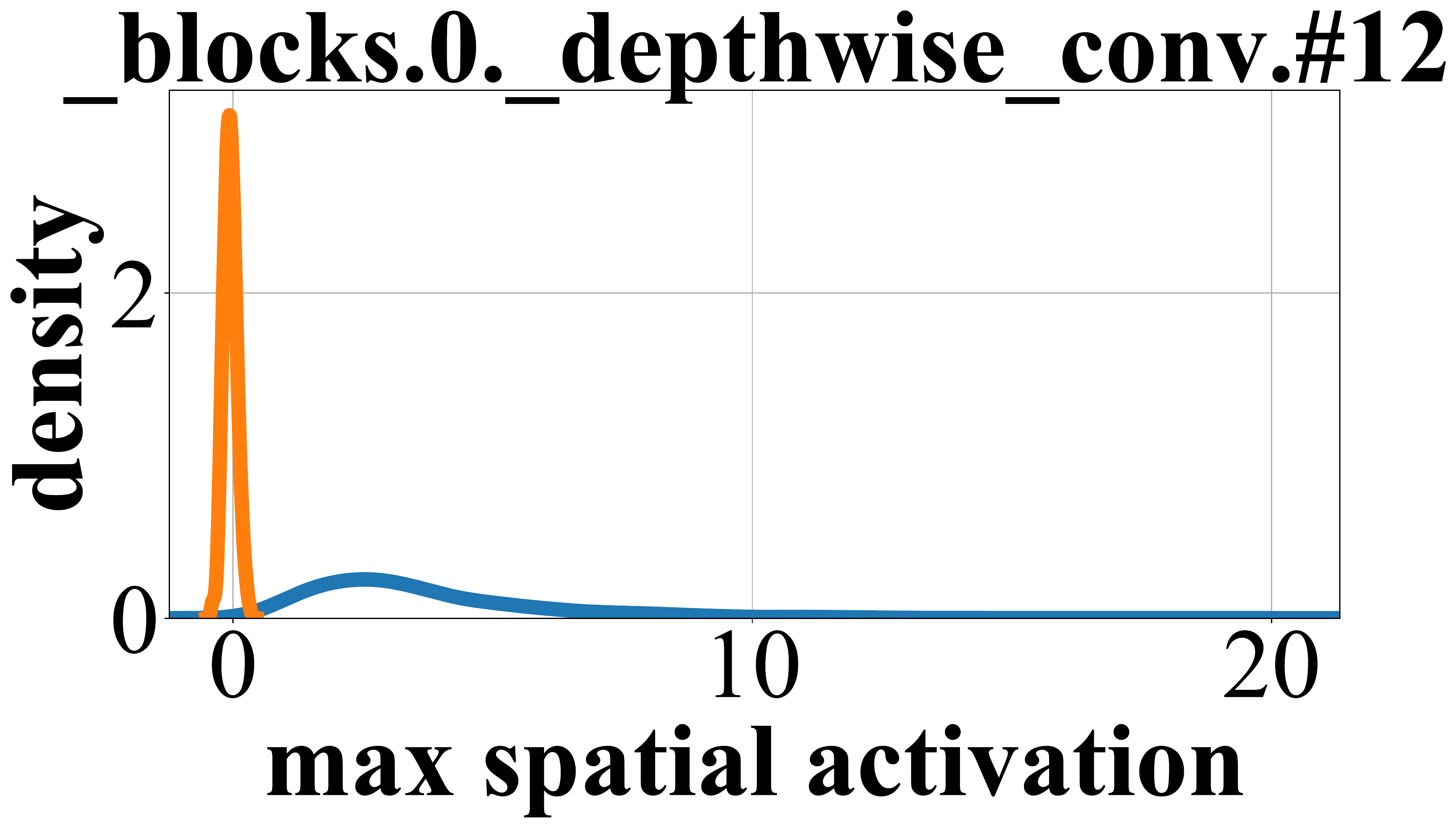} &
    \includegraphics[width=0.13\linewidth]{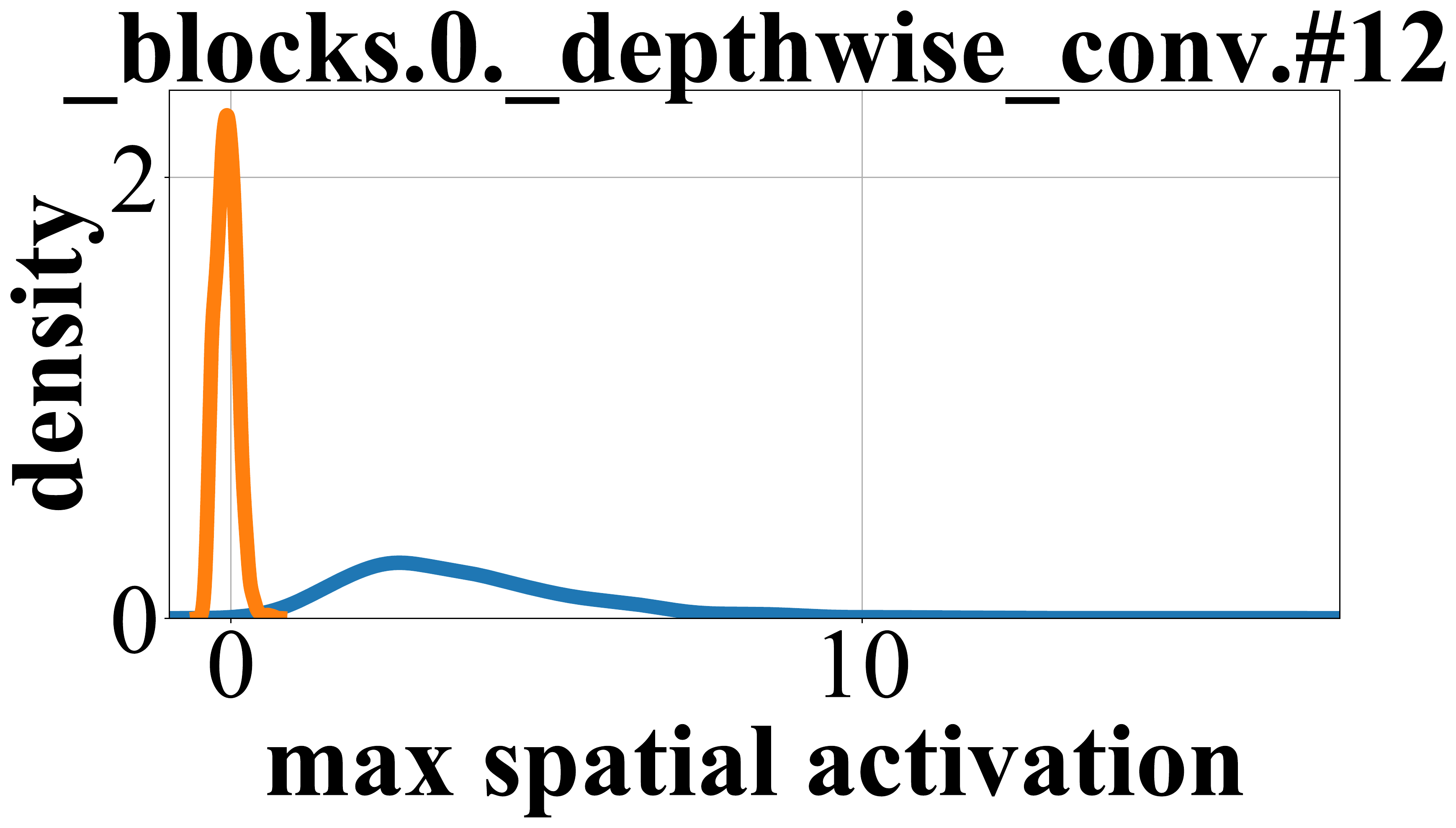} &
     \includegraphics[width=0.13\linewidth]{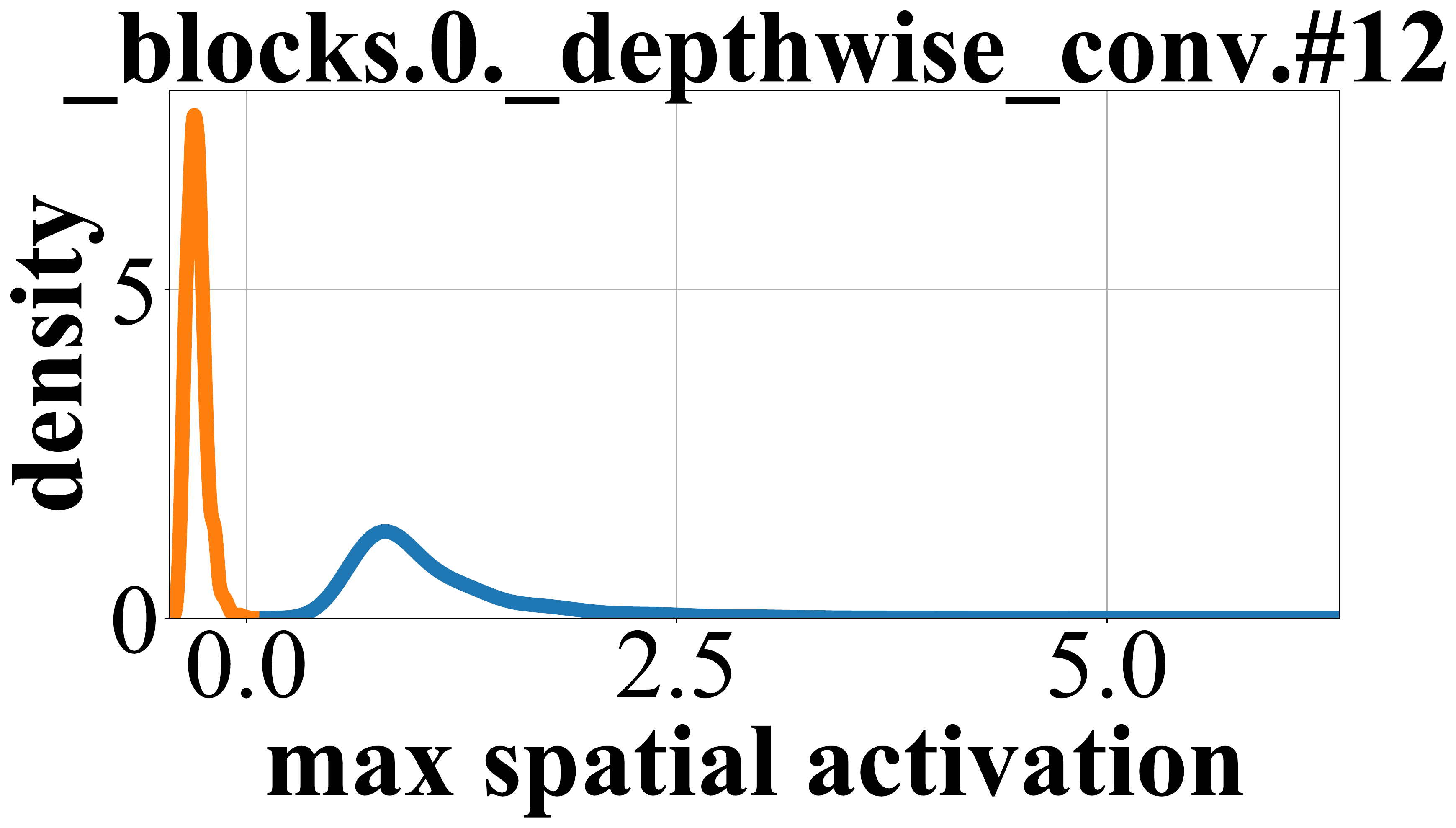} &
    \includegraphics[width=0.13\linewidth]{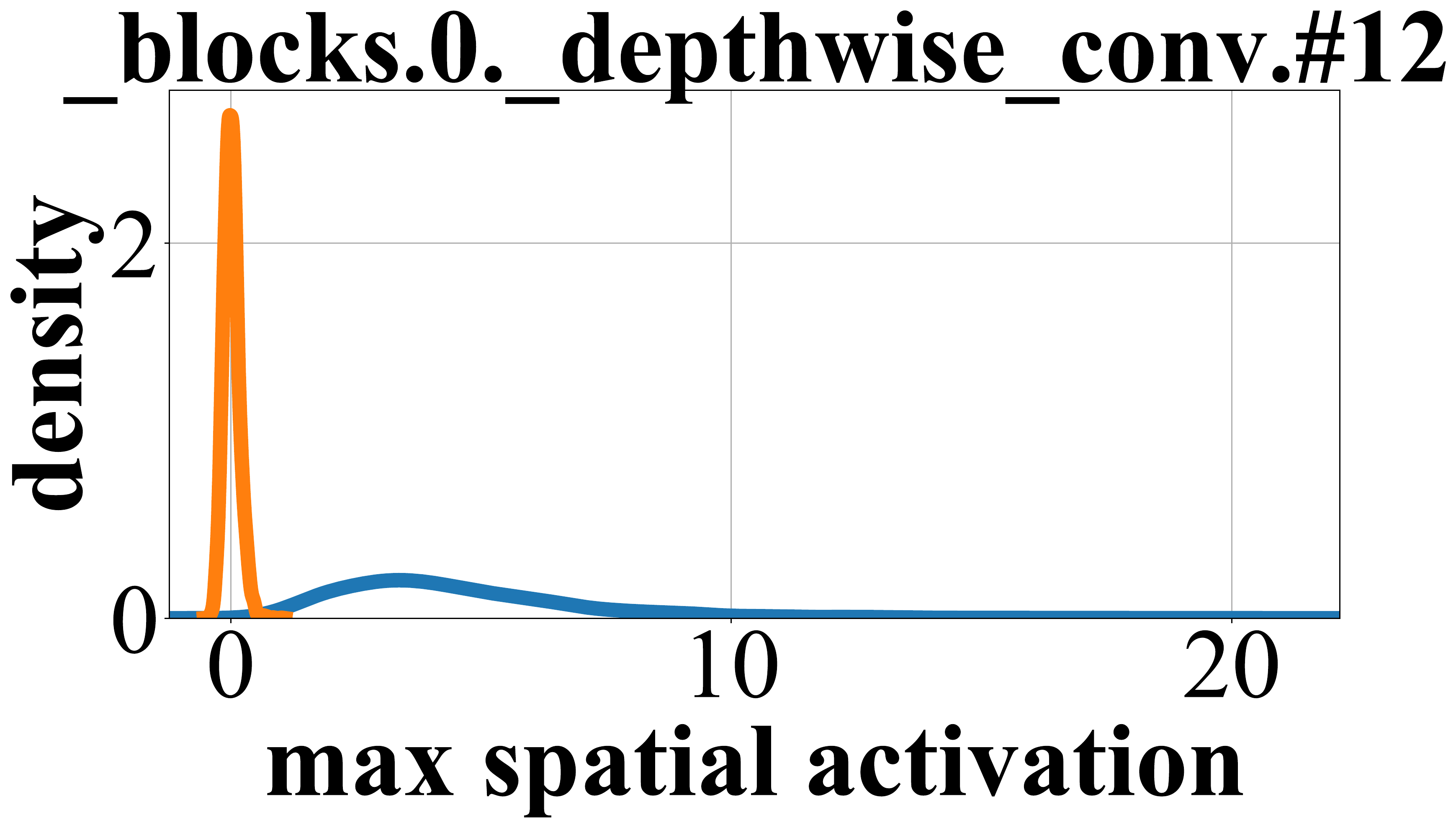}
    \\
    
    \includegraphics[width=0.13\linewidth]{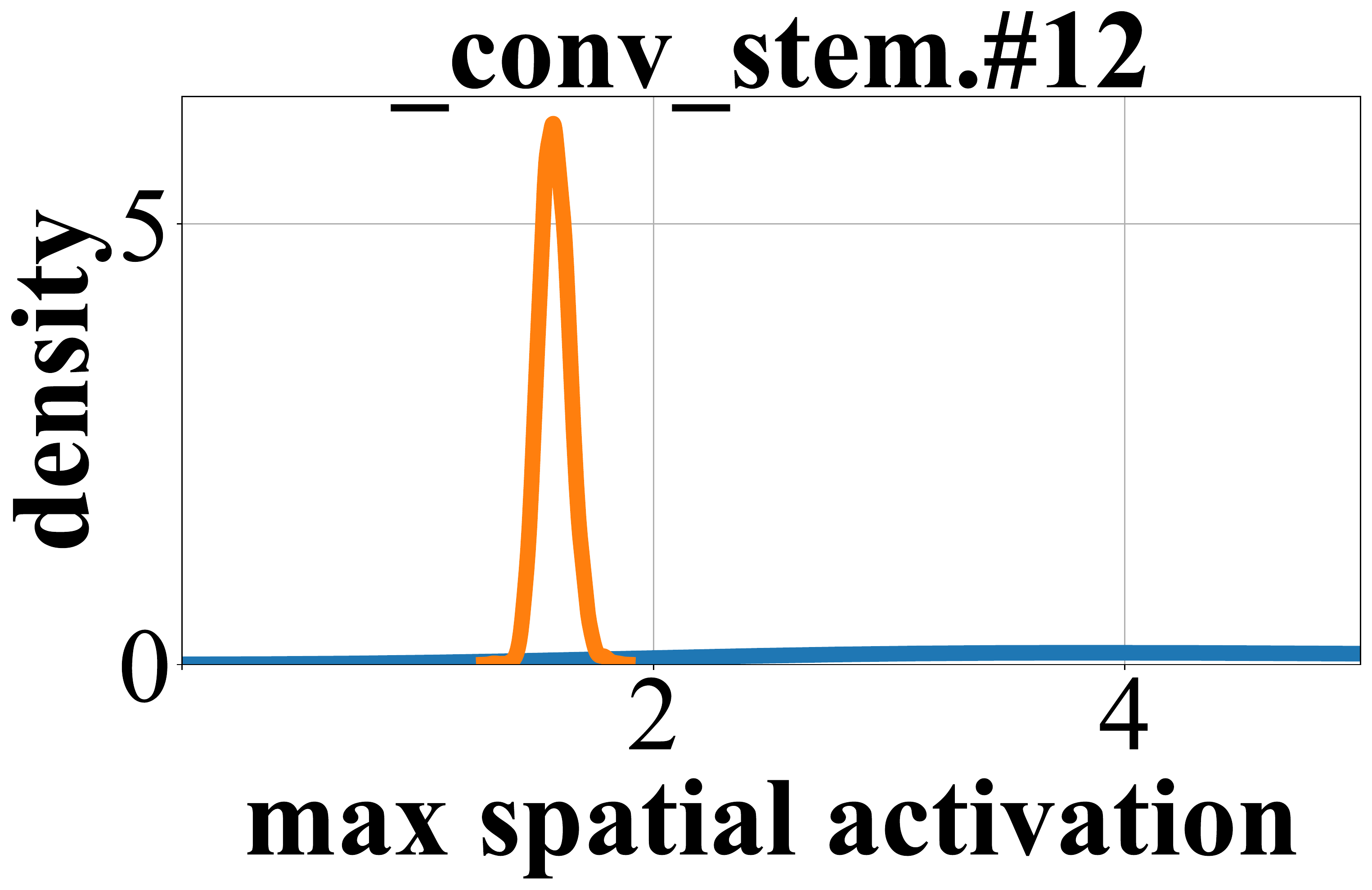} &
    \includegraphics[width=0.13\linewidth]{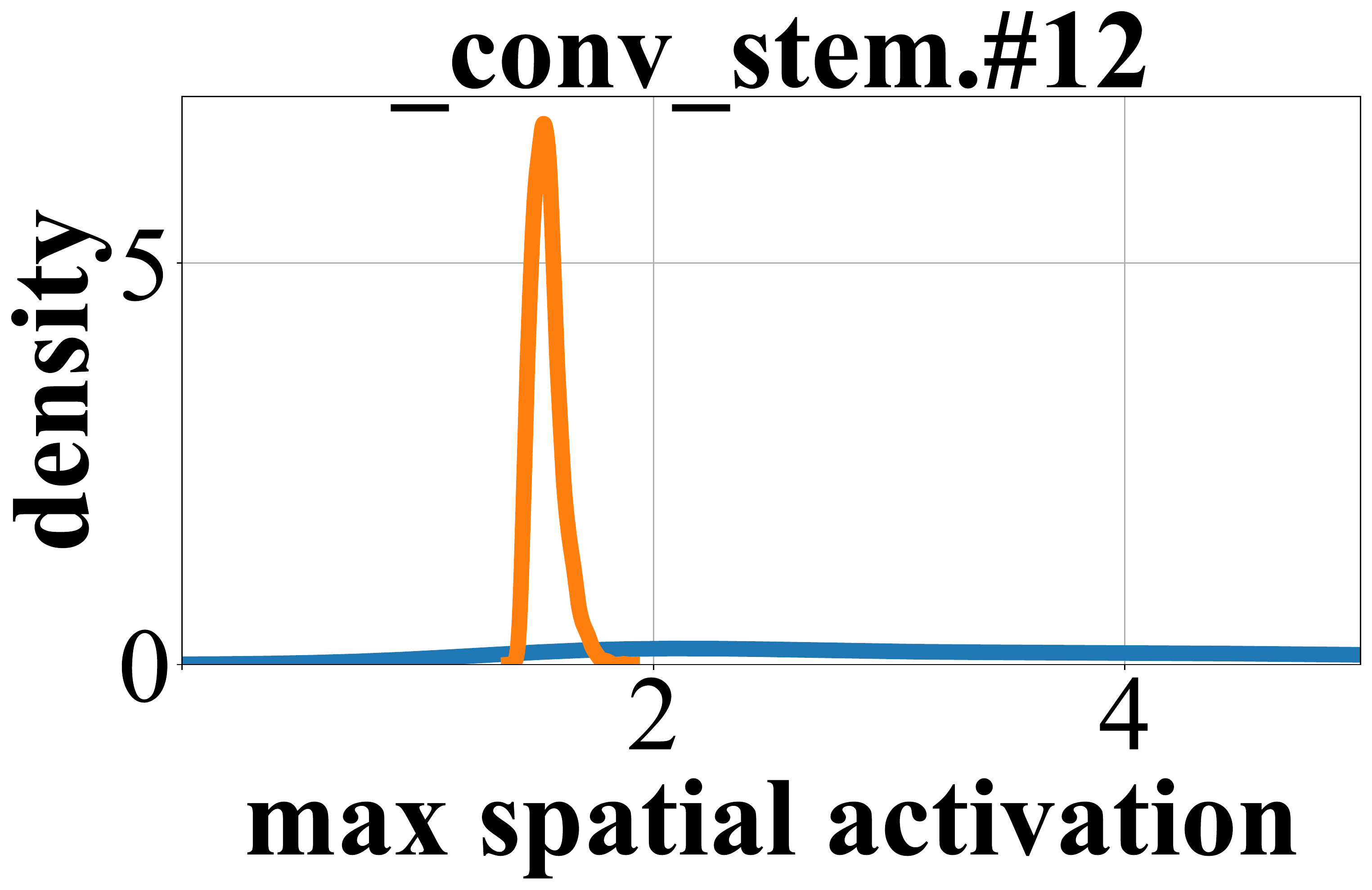} &
    \includegraphics[width=0.13\linewidth]{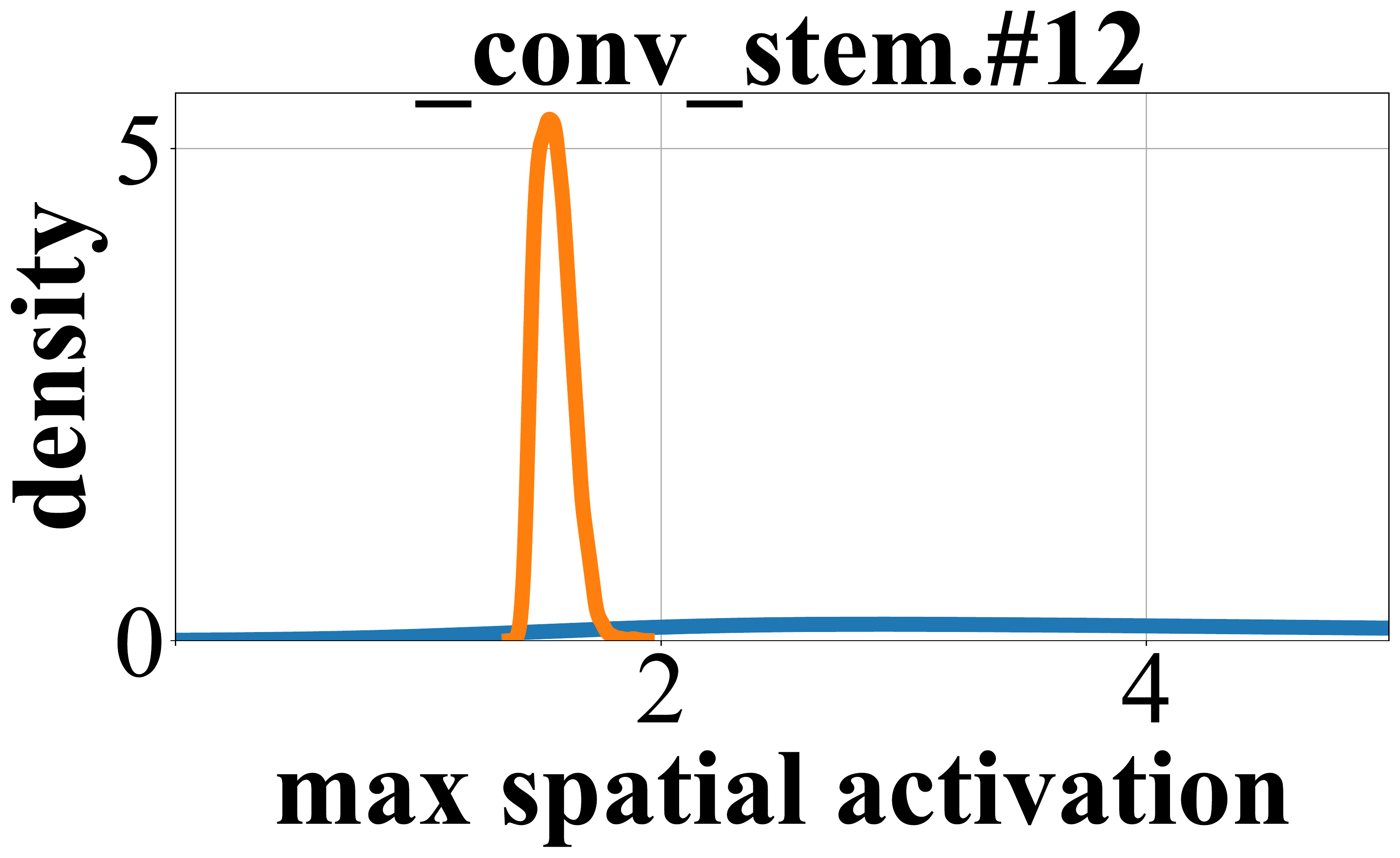} &
    \includegraphics[width=0.13\linewidth]{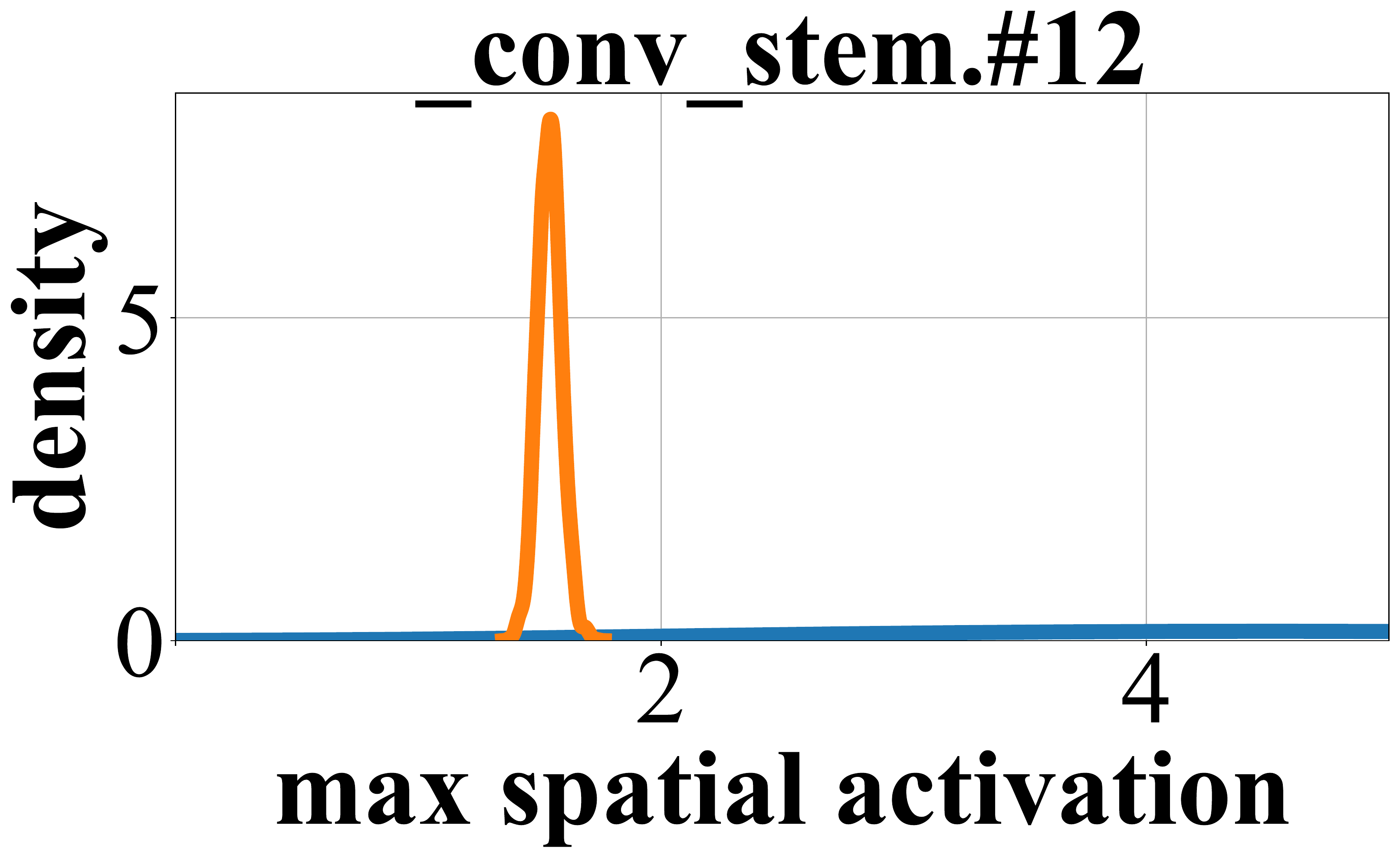} &
    \includegraphics[width=0.13\linewidth]{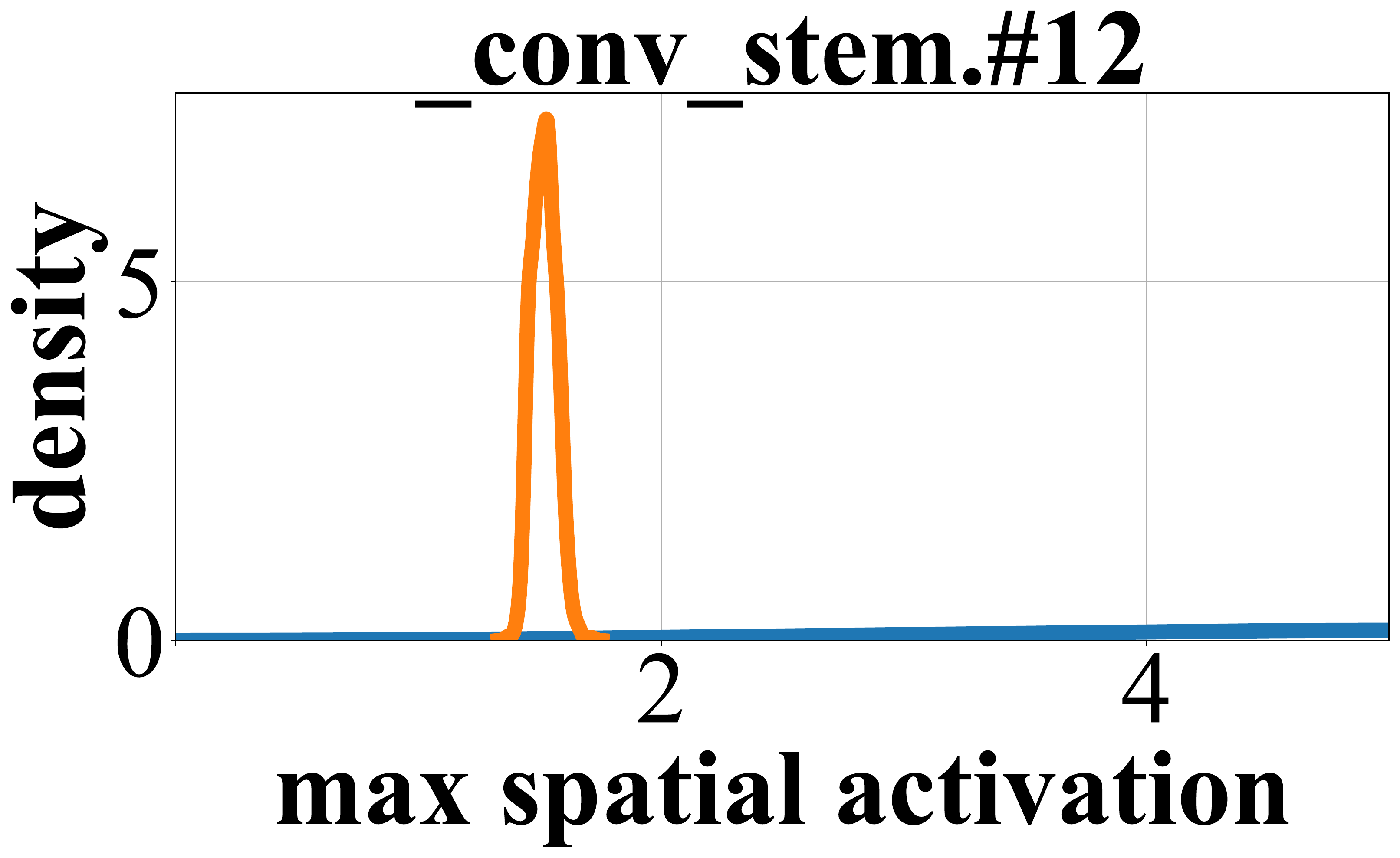} &
     \includegraphics[width=0.13\linewidth]{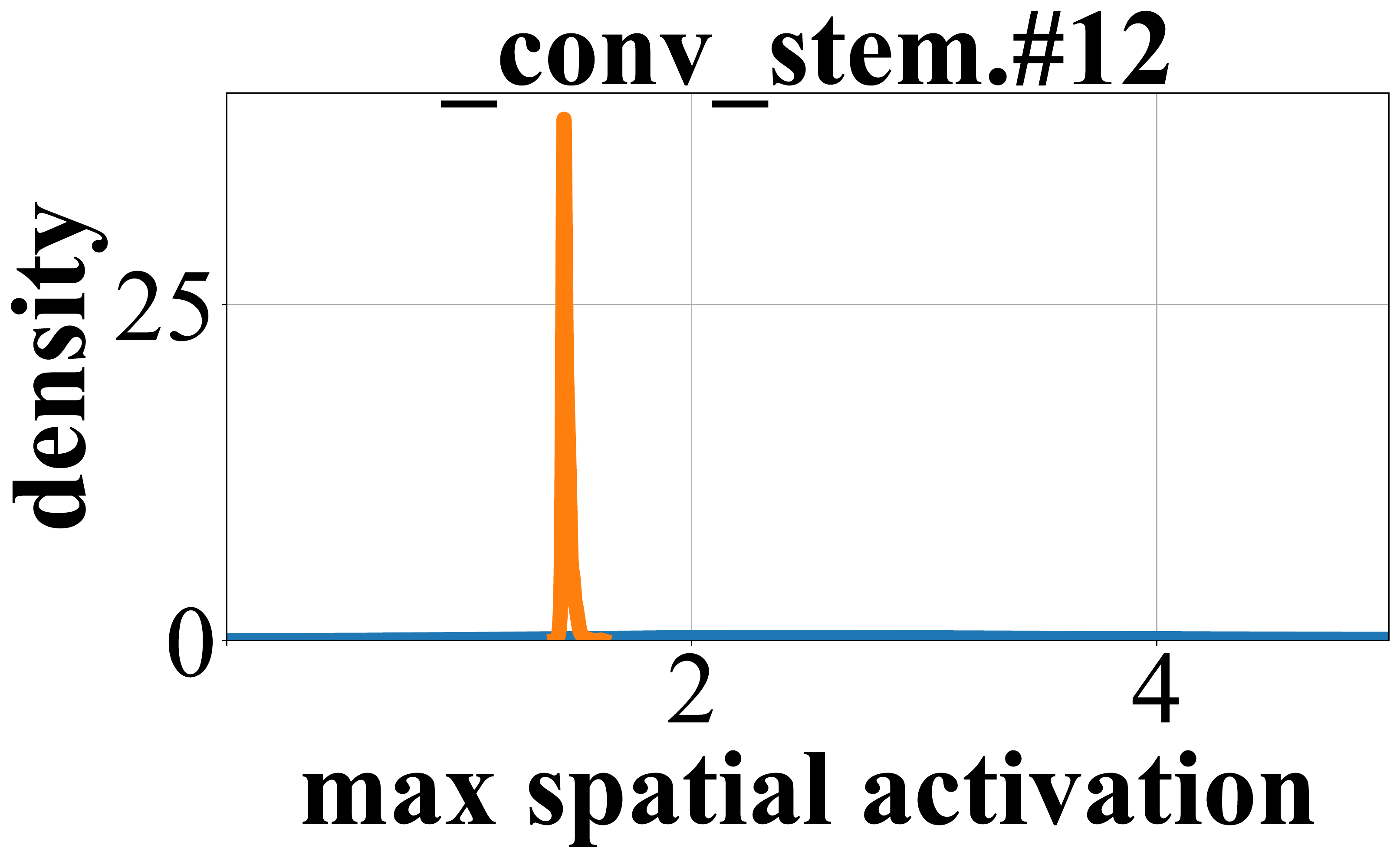} &
    \includegraphics[width=0.13\linewidth]{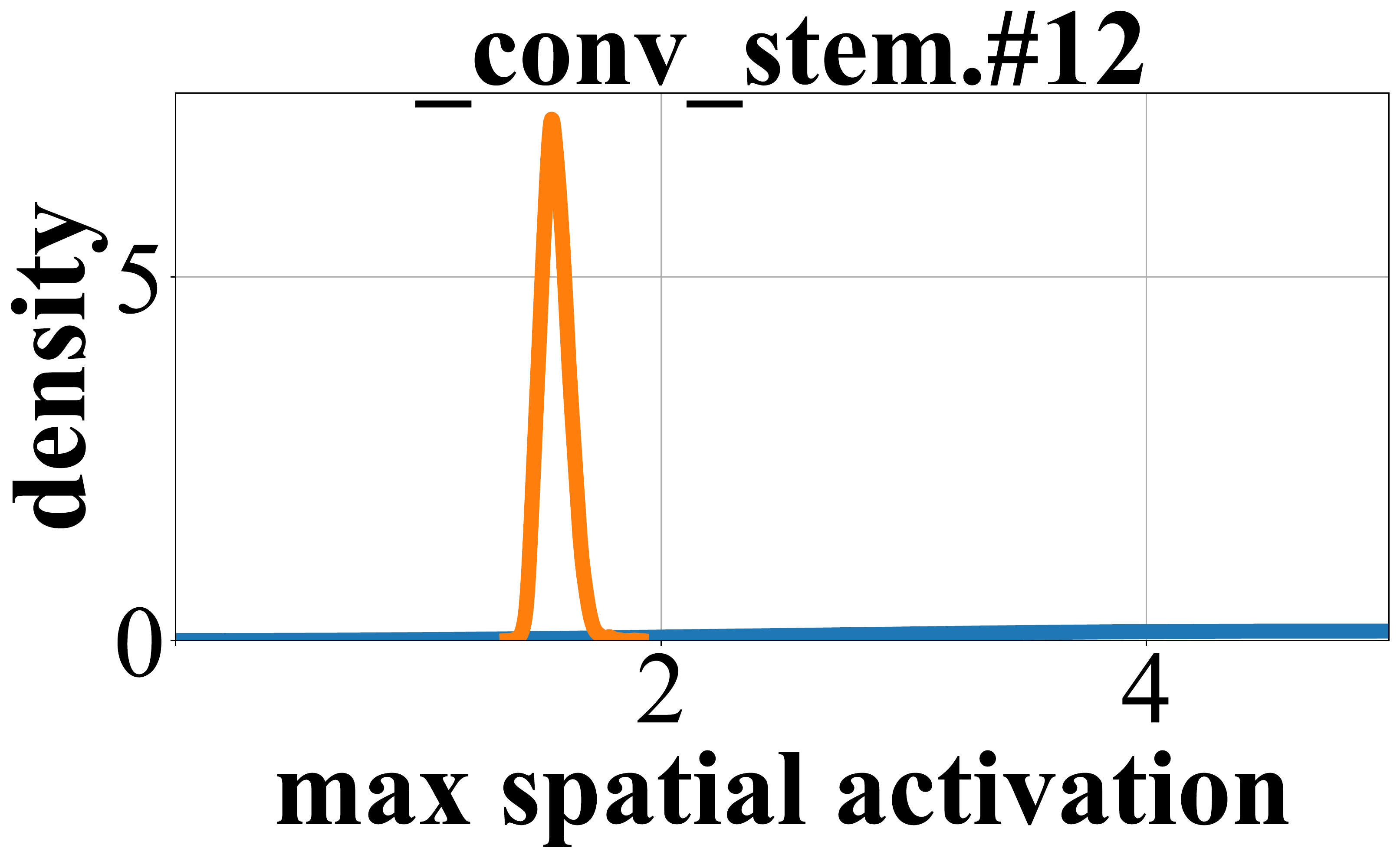}

\end{tabular}
\includegraphics[width=0.45\linewidth]{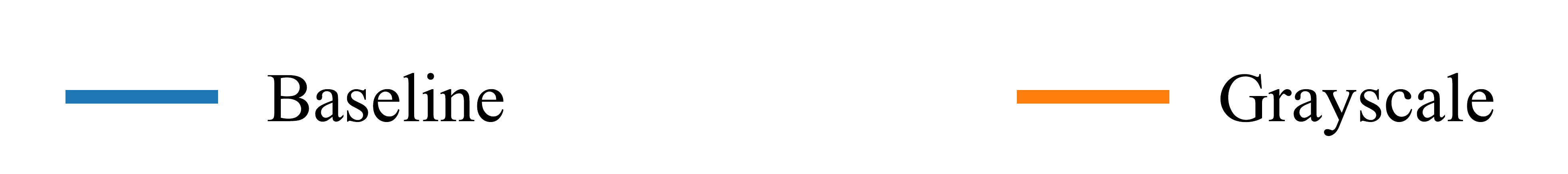}
\vspace{-0.4cm}
\caption{
\textit{Color-conditional T-FF in ResNet-50 and EfficientNet-B0:}
Each row represents a color-conditional \textit{T-FF}. These are the exact same T-FF shown in Fig. \ref{fig:lrp_patches_r50} (ResNet-50) and Fig. \ref{fig:lrp_patches_efb0}(EfficientNet-B0).
We show the maximum spatial activation distributions
for 7 GAN models 
before (Baseline) and after color ablation (Grayscale).
We remark that for each counterfeit in the ForenSynths dataset \cite{Wang_2020_CVPR}, we apply global max pooling to the specific T-FF to obtain a {\em maximum spatial activation} value (scalar).
We can clearly observe that these \textit{T-FF} are producing noticeably lower spatial activations (max) for the same set of counterfeits after removing color information. 
This clearly indicates that these \textit{T-FF} are color-conditional.
}
\label{fig_main:activation_hist_r50}
\vspace{-0.6cm}
\end{figure}

\textbf{Study 2.} In this study, we measure the percentage of \textit{T-FF} that are color-conditional. 
Particularly, we conduct a statistical test to compare the maximum globally pooled spatial activation distributions of each \textit{T-FF} before and after color ablation.
The intuition is that with color ablation, color-conditional \textit{T-FF} will produce lower amount of activations for the same sample and we perform a hypothesis test to measure whether the maximum spatial activation distributions are statistically different before (Baseline) and after color ablation (Grayscale).
Particularly, we use Mood's median test (non-parametric, low-power) with a significance level of $\alpha=0.05$ in our study.
The pseudocode is shown in Algorithm \ref{alg_main:median_test}.
The results for ResNet-50 and EfficientNet-B0 universal detectors are shown in Table \ref{table_main:color_conditional_percentage} (rows 1, 2).
Our results show that substantial amount of \textit{T-FF} in universal detectors are color-conditional indicating that color is a critical \textit{T-FF}.
We also show the maximum spatial activation distributions for several color-conditional \textit{T-FF} for ResNet-50 and EfficientNet-B0 universal detectors in Fig. \ref{fig_main:activation_hist_r50}.
As one can observe, maximum spatial activations are suppressed for these \textit{T-FF} across ProGAN \cite{karras2018progressive} and all unseen GANs \cite{Karras_2020_CVPR,Karras_2019_CVPR,brock2018large,zhu2017unpaired,choi2018stargan,park2019semantic} when removing color information. This clearly suggests that these \textit{T-FF} are color-conditional.

\begin{figure}[!t]
\centering
\begin{tabular}{ccccccc}
    {\tiny ProGAN \cite{karras2018progressive}} &
    {\tiny StyleGAN2 \cite{Karras_2020_CVPR}} &
    {\tiny StyleGAN \cite{Karras_2019_CVPR}} &
    {\tiny BigGAN \cite{brock2018large}} &
    {\tiny CycleGAN \cite{zhu2017unpaired}} &
    {\tiny StarGAN \cite{choi2018stargan}} &
    {\tiny GauGAN \cite{park2019semantic}} \\

    \includegraphics[width=0.13\linewidth]{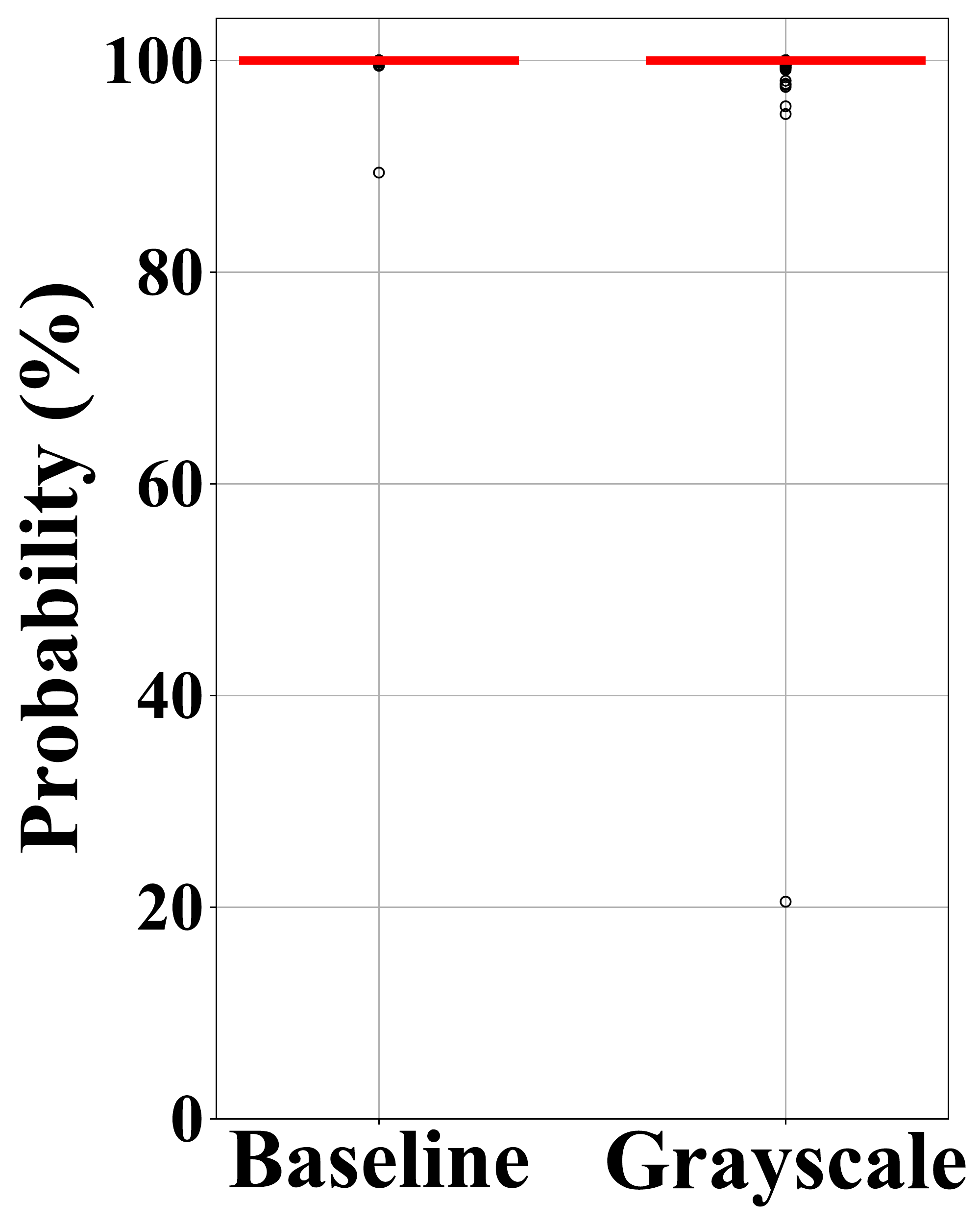} &
    \includegraphics[width=0.13\linewidth]{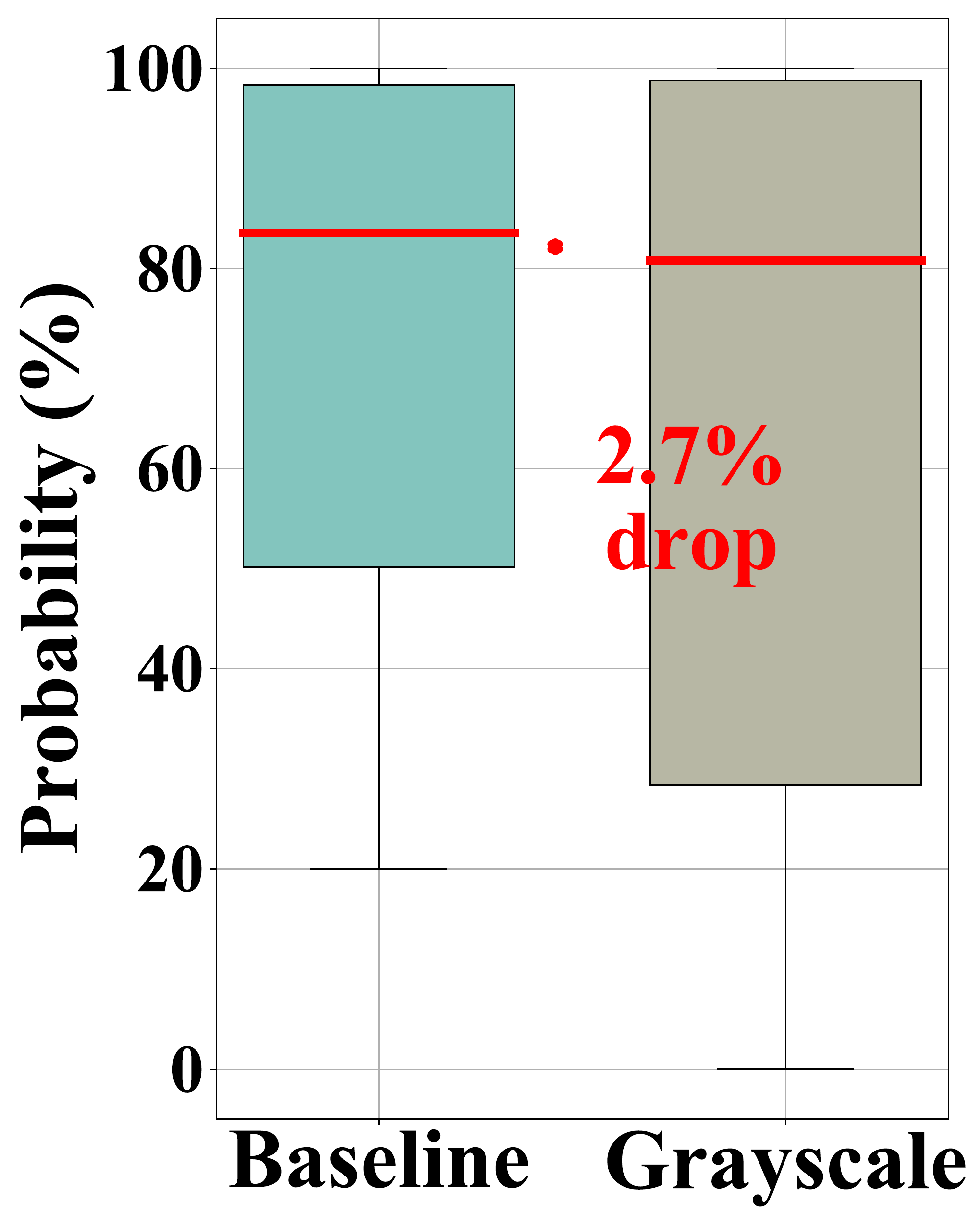} &
    \includegraphics[width=0.13\linewidth]{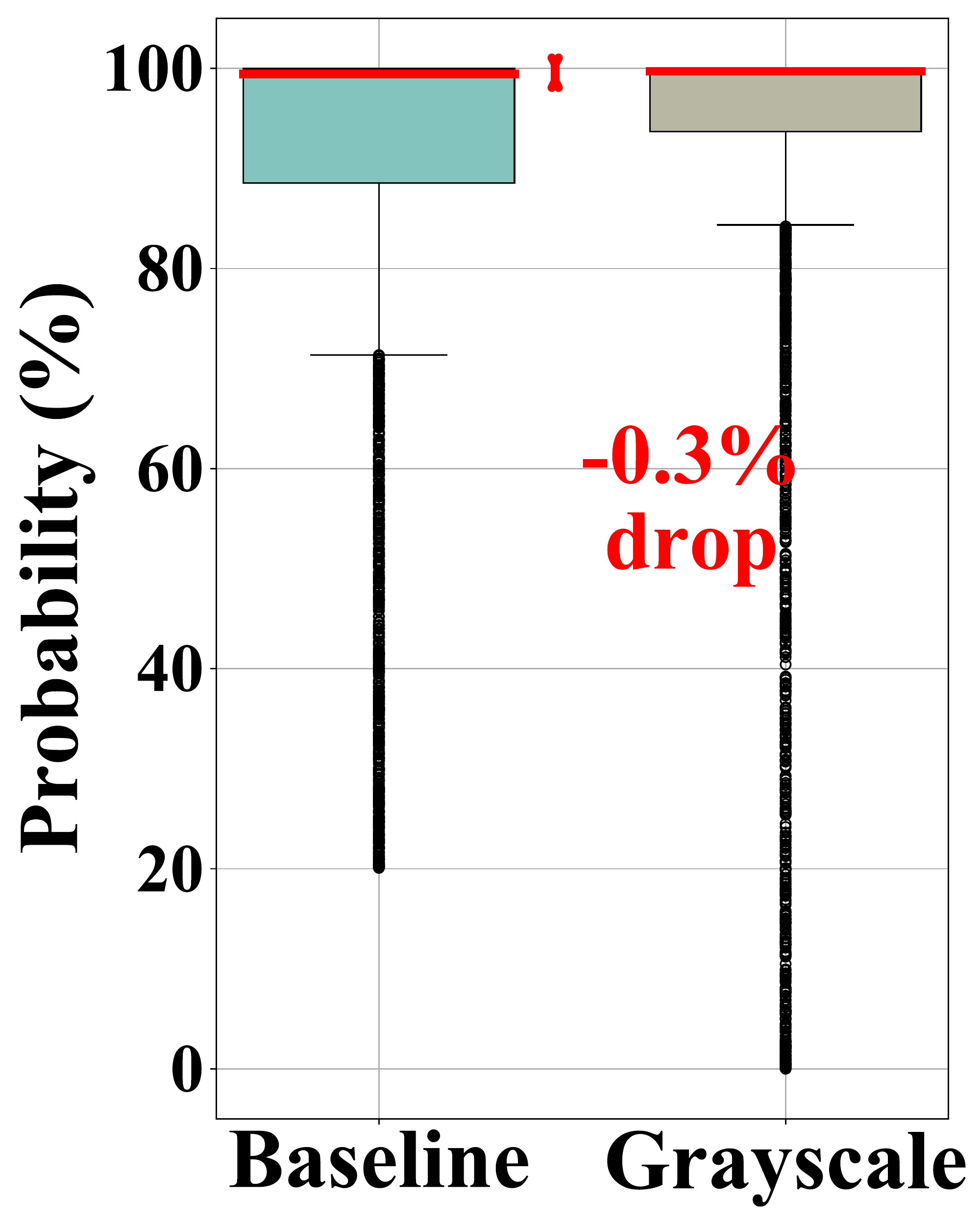} &
    \includegraphics[width=0.13\linewidth]{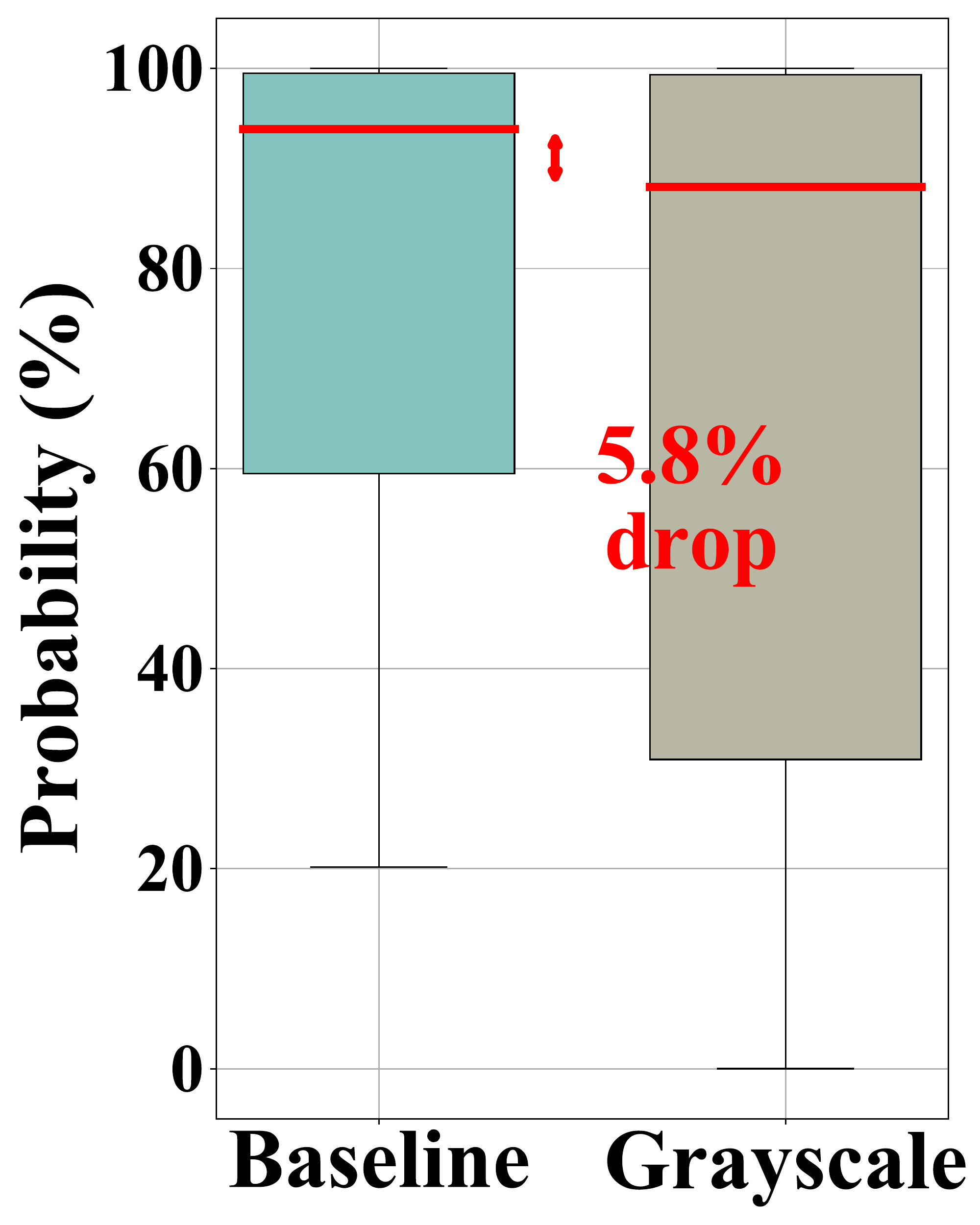} &
    \includegraphics[width=0.13\linewidth]{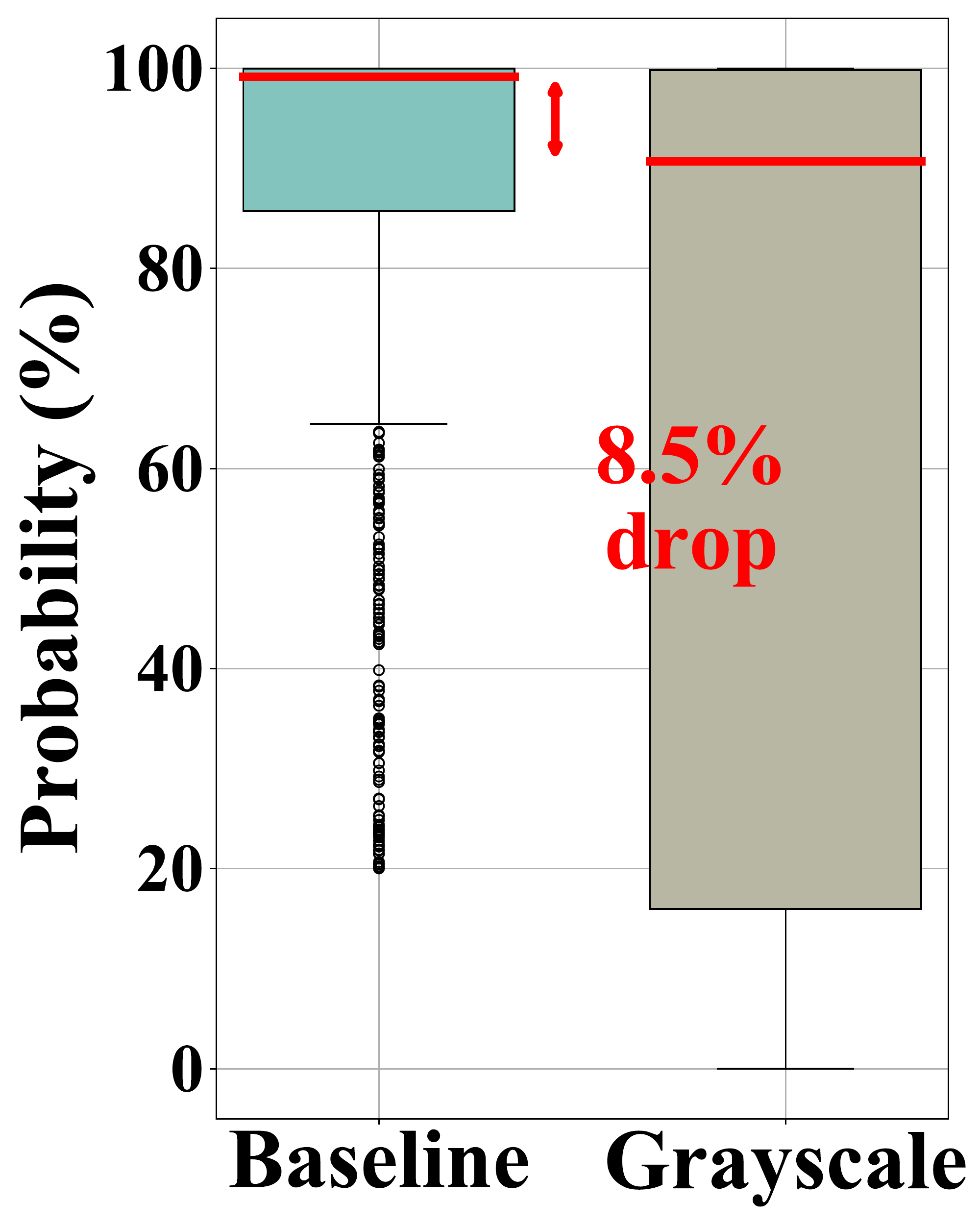} &
    \includegraphics[width=0.13\linewidth]{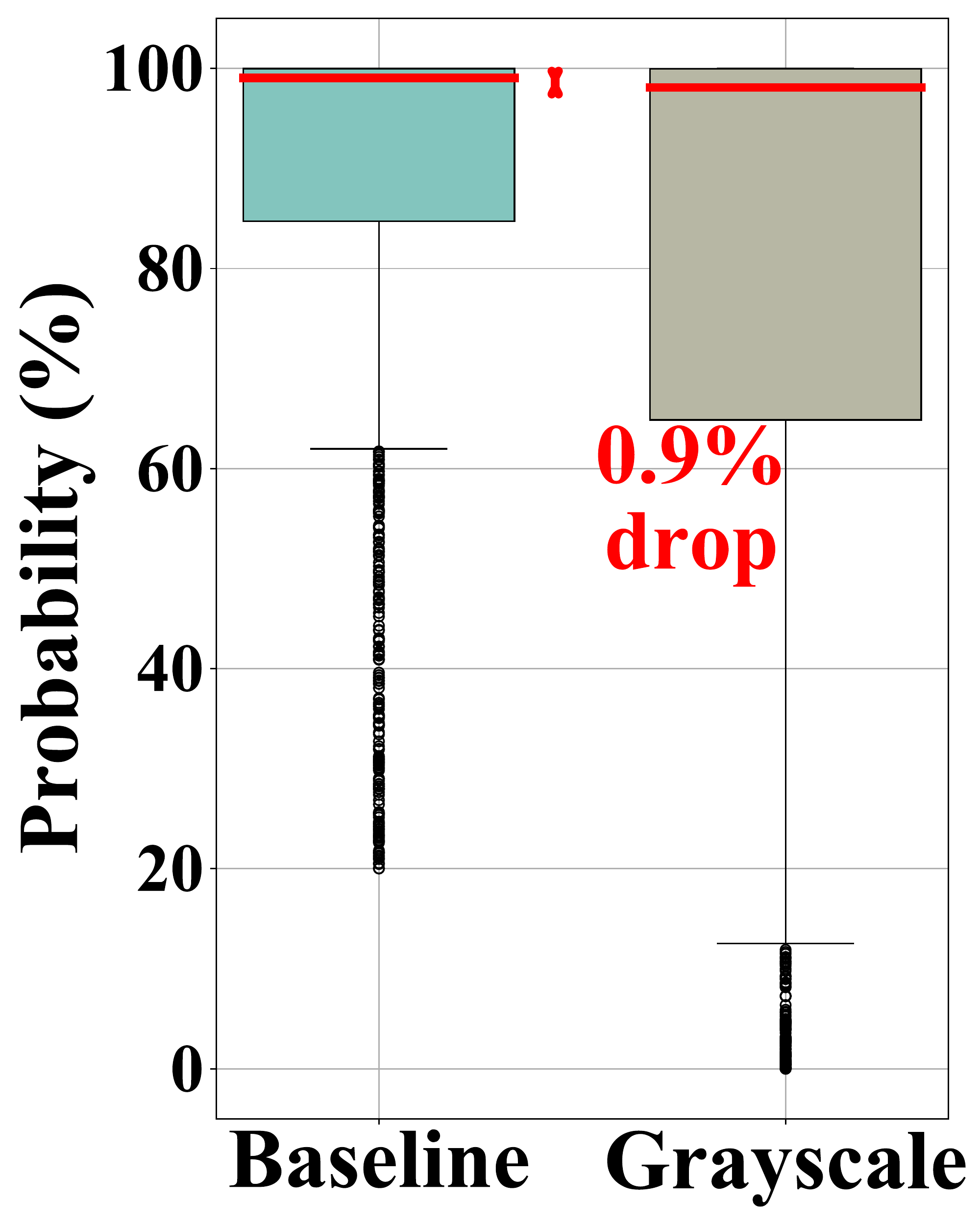} &
    \includegraphics[width=0.13\linewidth]{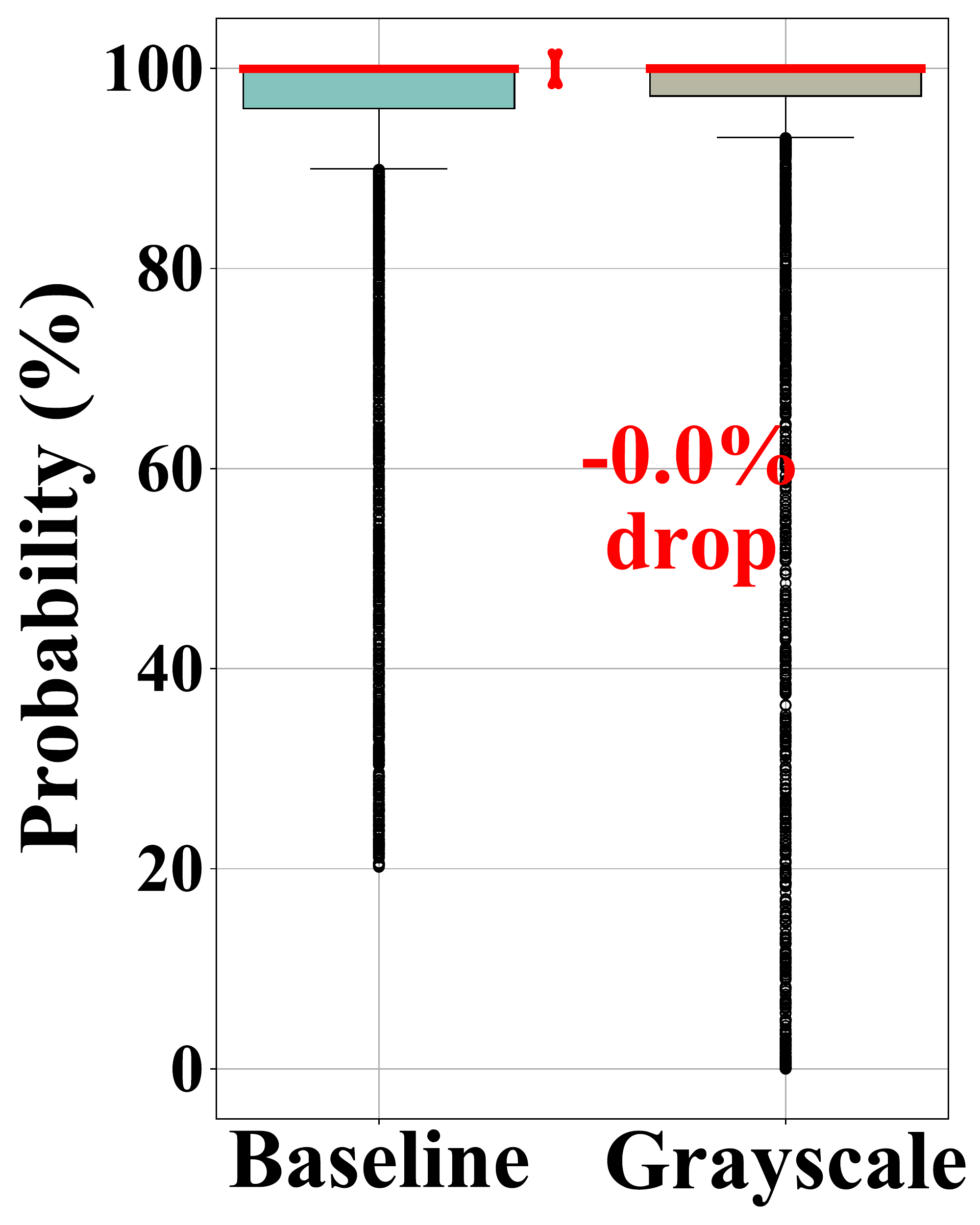}
    \\
    
    \includegraphics[width=0.13\linewidth]{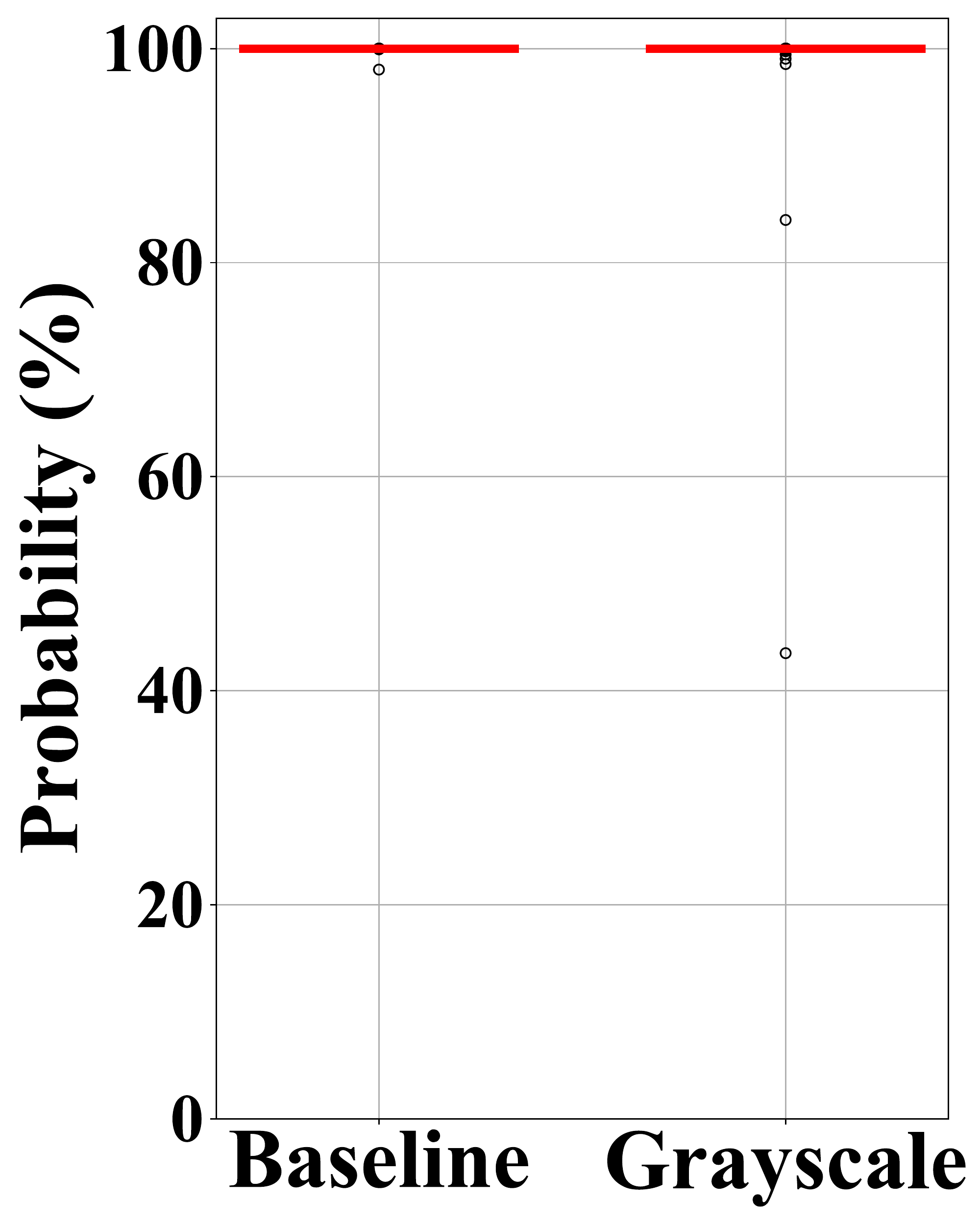} &
    \includegraphics[width=0.13\linewidth]{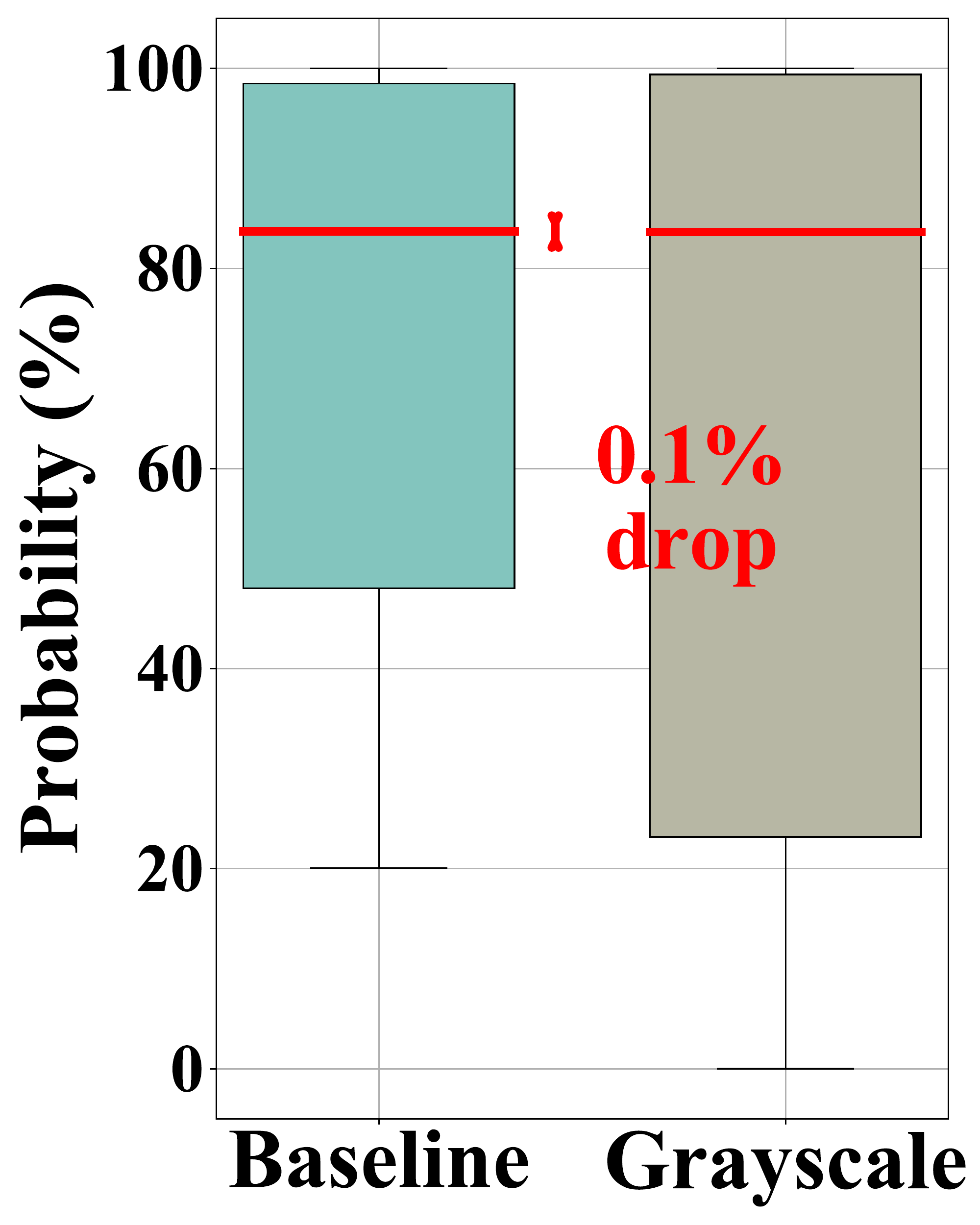} &
    \includegraphics[width=0.13\linewidth]{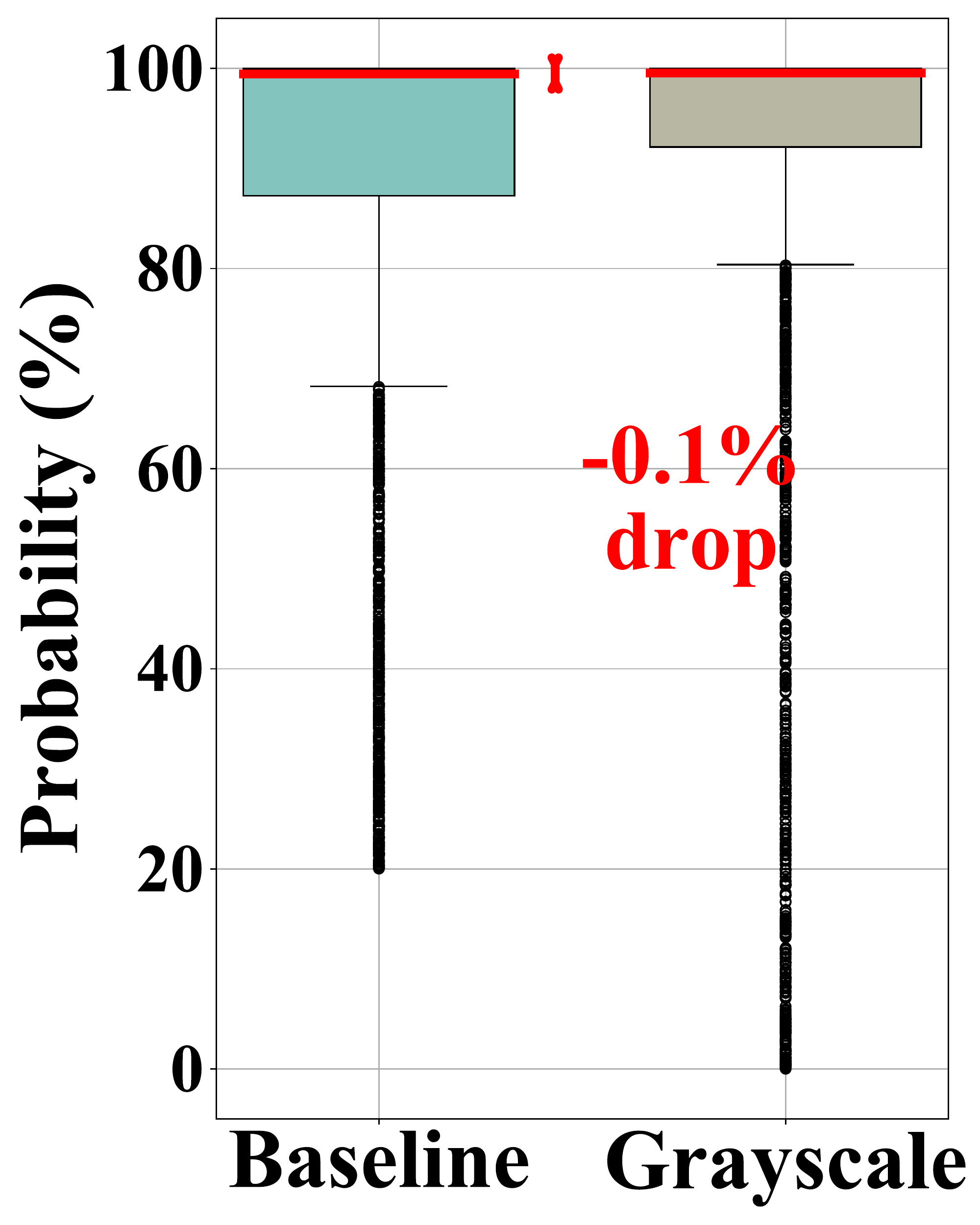} &
    \includegraphics[width=0.13\linewidth]{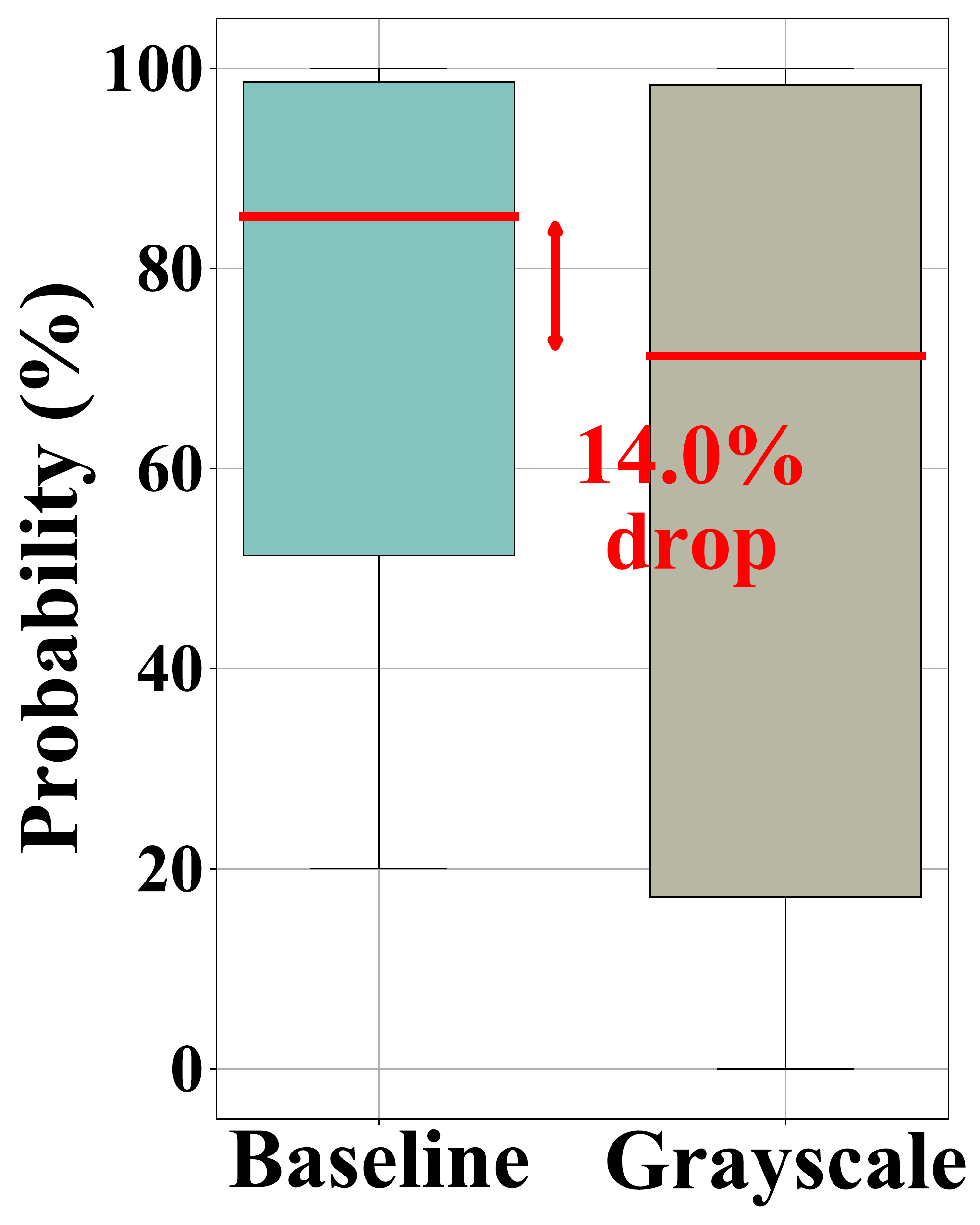} &
    \includegraphics[width=0.13\linewidth]{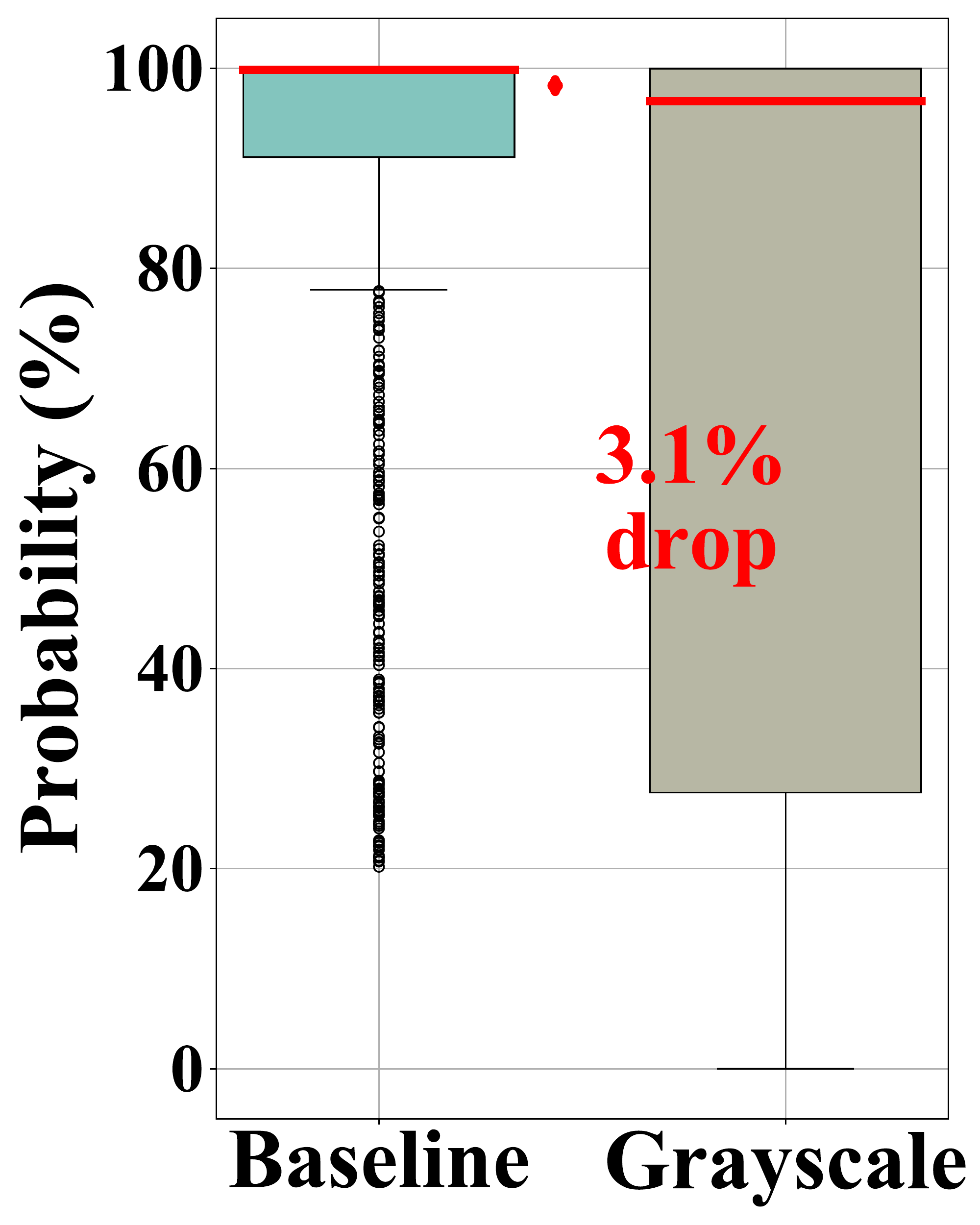} &
    \includegraphics[width=0.13\linewidth]{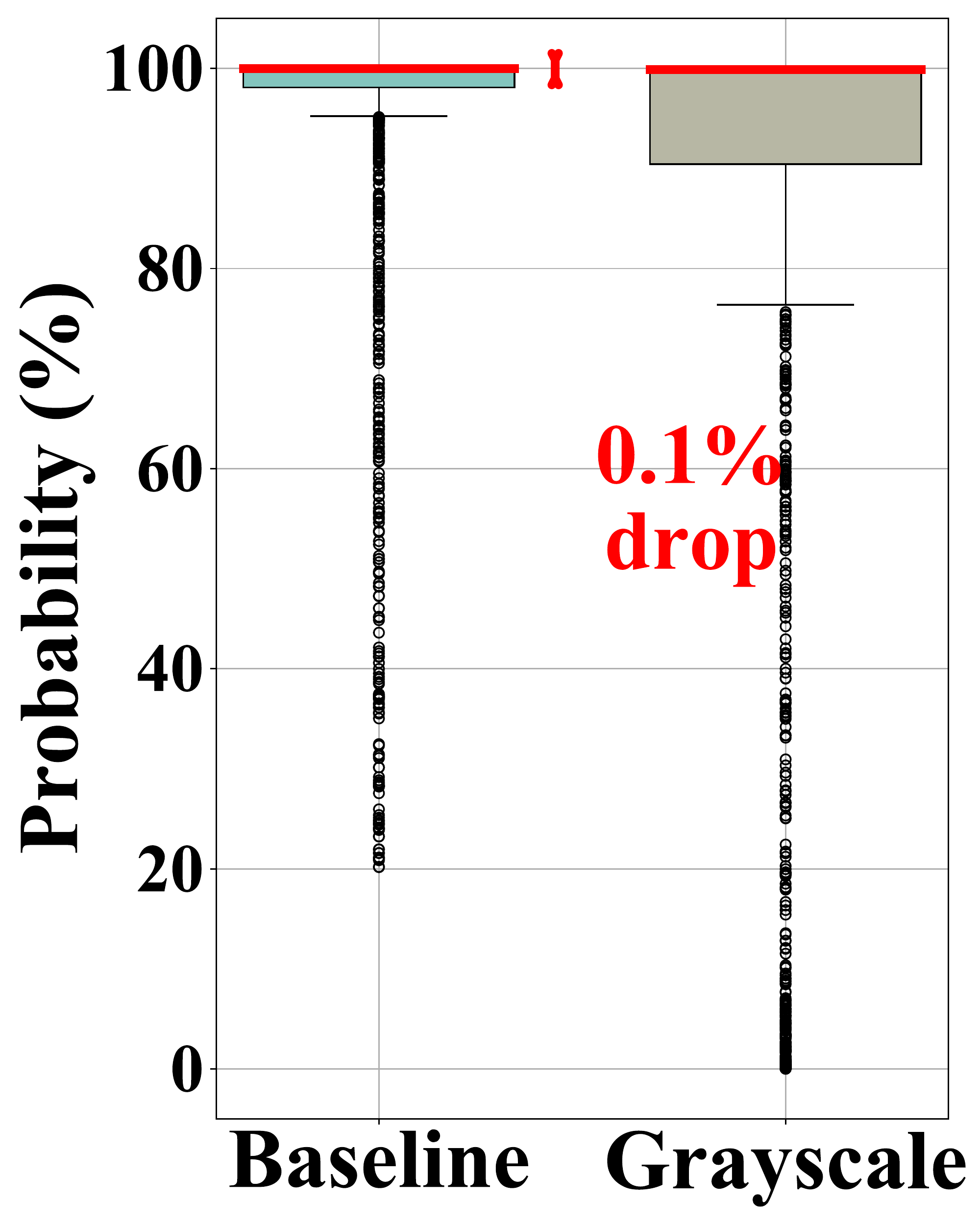} &
    \includegraphics[width=0.13\linewidth]{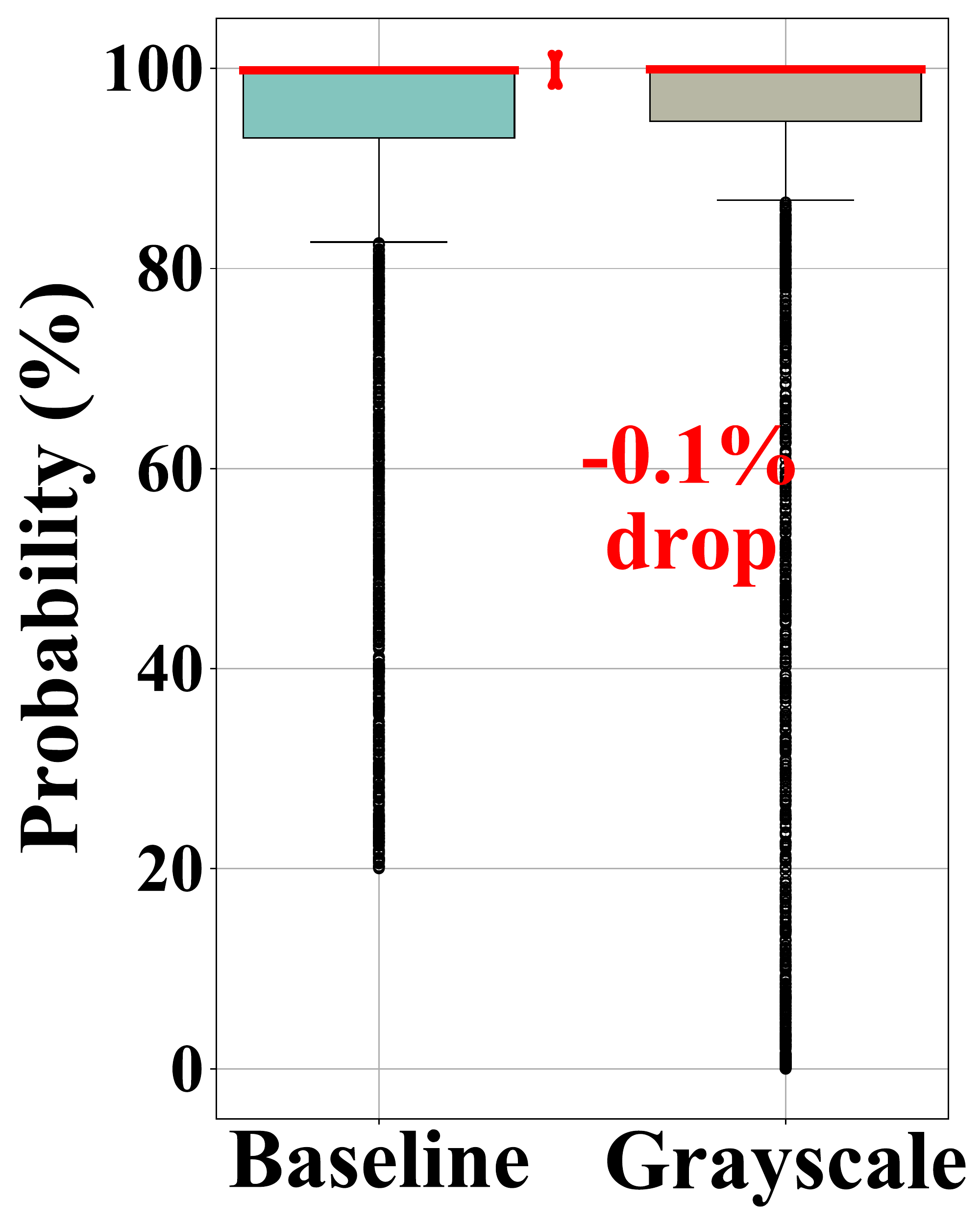}
    \\
    
\end{tabular}
\vspace{-0.3cm}
\caption{
\textit{CR-universal detectors trained using our proposed data augmentation scheme (Sec. \ref{sec_main:cr_universal_detectors}) are more robust to color ablation during cross-model forensic transfer:}
These universal detectors are trained with data augmentation where color 
is ablated 50\% of the time during training. 
This ensures that \textit{T-FF} do not substantially rely on color information.
We show the box-whisker plots of probability (\%) predicted by the CR-universal detectors for counterfeits before (Baseline) and after \textit{color ablation} (Grayscale) for 7 GAN models.
The red line in each box-plot shows the median probability.
We show the results for the 
ResNet-50
CR-universal detector 
\cite{Wang_2020_CVPR} (top row) and 
EfficientNet-B0 \cite{tan2019efficientnet} CR-universal detector (bottom row).
We clearly observe that the median probability for counterfeits have similar values (compared to Fig. \ref{fig_main:median_color_ablation}) before and after color ablation indicating CR-universal detectors are more robust to color-ablated counterfeit attacks. 
AP and accuracies shown in Supplementary
\ref{table_supp:cr}.
\vspace{-0.2cm}
}
\label{fig_main:median_color_ablation_robust_detector}
\end{figure}

\begin{table}[!t]
\begin{center}
\caption{
Median Test Results.
\textcircled{\raisebox{-0.8pt}{1}} \textit{Significant amount of T-FF are color-conditional (rows 1, 2):}
We show the percentage(\%) of color-conditional T-FF in ResNet-50 and EfficientNet-B0 universal detectors measured using Mood's median test. 
We show the results for ProGAN \cite{karras2018progressive} and 
all 6 unseen GANs 
\cite{Karras_2020_CVPR,Karras_2019_CVPR,brock2018large,zhu2017unpaired,choi2018stargan,park2019semantic}.
Particularly, we consider a \textit{T-FF} to be color conditional if the $p$-value of the median test is less than the significance level of $\alpha=0.05$. 
As one can clearly observe, significant amount of \textit{T-FF} are color-conditional. 
This quantitatively shows that color is a critical \textit{T-FF} in universal detectors.
\textcircled{\raisebox{-0.8pt}{2}} \textit{CR-universal detectors have lower amount of color-conditional T-FF (rows 3,4)}: 
We clearly observe that
training universal detectors using our proposed data augmentation scheme (Sec \ref{sec_main:cr_universal_detectors}) results in detectors that contain noticeably lower amount of color-conditional \textit{T-FF}.
\vspace{-0.2cm}
}
\begin{adjustbox}{width=1.0\columnwidth,center}
\begin{tabular}{l|c|c|c|c|c|c|c}\toprule
\textbf{\% Color-conditional} &
\textbf{ ProGAN \cite{karras2018progressive}} &
\textbf{ StyleGAN2 \cite{Karras_2020_CVPR}} &
\textbf{ StyleGAN \cite{Karras_2019_CVPR}} &
\textbf{ BigGAN \cite{brock2018large}} &
\textbf{ CycleGAN \cite{zhu2017unpaired}} &
\textbf{ StarGAN \cite{choi2018stargan}} &
\textbf{ GauGAN \cite{park2019semantic}} \\
\toprule

ResNet-50 &85.1 &74.6 &73.7 &68.4 &86.8 &71.1 &70.2 \\ \midrule
Efficient-B0 &51.9 &48.1 &40.7 &40.7 &44.4 &44.4 &37.0 \\ \midrule
CR-ResNet-50 &55.3 &33.3 &48.2 &31.6 &56.1 &48.2 &39.5 \\ \midrule
CR-EfficientNet-B0 &20.0 &30.0 &20.0 &10.0 &20.0 &20.0 &10.0 \\ \bottomrule
\end{tabular}
\end{adjustbox}
\label{table_main:color_conditional_percentage}
\end{center}
\vspace{-0.9cm}
\end{table}

\vspace{-0.6cm}
\section{Applications : Color-Robust (CR) Universal Detectors}
\label{sec_main:cr_universal_detectors}

Reliance on substantial amount of color information for cross-model forensic transfer exposes universal detectors to attacks via color-ablated counterfeits.
This is particularly unfavourable. 
In this section, we 
propose a data augmentation scheme to
build Color-Robust (CR) universal detectors that do not substantially rely on color information for cross-model forensic transfer.
The crux of the idea is to randomly remove color information from samples during training (both for real and counterfeit images).
Particularly, we perform random Grayscaling during training with 50\% probability to maneuver universal detectors to learn \textit{T-FF} that do not substantially rely on color information.

\noindent
\textbf{Results.} Median probability analysis results for ResNet-50 and EfficientNet-B0 CR-universal detectors are shown in Fig. \ref{fig_main:median_color_ablation_robust_detector}.
We clearly observe that with our proposed data augmentation scheme, CR-universal detectors are more robust to color ablation during cross-model forensic transfer indicating that they learn \textit{T-FF} that do not substantially rely on color information.
We further show the percentage of color-conditional \textit{T-FF} 
in CR-ResNet-50 and CR-EfficientNet-B0
in Table \ref{table_main:color_conditional_percentage} (rows 3, 4),
quantitatively showing that CR-universal detectors learn substantially lower amount of color-conditional \textit{T-FF}.

{
\textbf{T-FF in CR-Universal Detectors.}
We further discover T-FF in CR-universal detectors using our proposed \textit{FF-RS ($\omega$)}. We show LRP-max visualization of T-FF in CR-ResNet-50 in Supplementary Fig. \ref{fig_supp:lrp_patches_cr_r50}. These T-FF largely correspond to patterns / artifacts (i.e.: wheels).
We emphasize that our proposed method can identify different types of \textit{T-FF} in addition to color.
}

\section{Discussion and Conclusion}

We conducted the 
\textit{first analytical study to discover and understand \textit{transferable forensic features (T-FF)} in universal detectors.}
Our first set of investigations demonstrated that input-space attribution methods such as Guided-GradCAM \cite{selvaraju2017grad} and LRP \cite{bach2015pixel} are not informative to discover \textit{T-FF} (Sec \ref{sec_main:input_space_attribution_methods}). 
In light of these observations, we study the forensic feature space of universal detectors.
Particularly, we propose a novel \textit{forensic feature relevance statistic (FF-RS)} to quantify and discover \textit{T-FF} in universal detectors.
Rigorous sensitivity assessments using feature map dropout convincingly show that our proposed FF-RS ($\omega$) is able to successfully quantify and discover \textit{T-FF} (Sec \ref{sec_main:forensic_feature_space}).

Further investigations on \textit{T-FF} uncover an unexpected finding: $color$ is a critical \textit{T-FF} in universal detectors. 
We show this critical finding qualitatively using our proposed LRP-max visualization of discovered \textit{T-FF} (Sec \ref{sec_main:lrp_max}).
Further we validate this finding quantitatively using median counterfeit probability analysis and statistical tests on maximum spatial activation distributions of \textit{T-FF} based on color ablation (Sec \ref{sec_main:color_ablation}). 
i.e.: We showed that $\approx 85\%$ of \textit{T-FF} are color-conditional in the publicly released ResNet-50 universal detector \cite{Wang_2020_CVPR}.  
Finally, we propose a simple data augmentation scheme to train 
Color-Robust (CR) universal detectors (Sec \ref{sec_main:cr_universal_detectors}). We remark that color is not the only \textit{T-FF}, but it is a critical \textit{T-FF} in universal detectors.
We also discuss computational complexity of FF-RS ($\omega$) and LRP-max in Supplementary \ref{sec_supp:computational_complexity}.
A natural question would be why is color a critical \textit{T-FF}. Though this is not a straight-forward question to answer, we provide our perspective: 
Color distribution of real images is non-uniform, and
we hypothesize that most GANs 
struggle to capture the diverse,
multi-modal color distribution of real images. i.e.: low-density color regions.
This may result in noticeable discrepancies between real and GAN images (counterfeits) in the color space
which
can be used as \textit{T-FF} to 
detect counterfeits.
To conclude, through this work we discover and understand \textit{T-FF} in universal detectors for counterfeit detection, and hope that our contributions will inspire further research in image forensics and image synthesis methods.
\newline

\noindent
\textbf{Limitations / Broader Impact.}
With deepfakes-in-the-wild being generated using diverse techniques in addition to GAN-based methods including shallow methods (i.e.: Photoshop) and face-swapping frameworks (i.e.: DeepFaceLab \cite{perov2020deepfacelab}), studying transferable forensic features in such synthesis methods are essential to build robust general-purpose image forensics detectors.
With increasing usage of machine learning methods in proliferating mis- and disinformation, we hope that our discovery on transferable forensic features can open-up more plausible research directions to combat the fight against visual disinformation.
\newline

\noindent
{\small
\textbf{Acknowledgements.}
This research is supported by the National Research Foundation, Singapore under its AI Singapore Programmes (AISG Award No.: AISG2-RP-2021-021; AISG Award No.: AISG-100E2018-005). 
This project is also supported by SUTD project PIE-SGP-AI-2018-01.
Alexander Binder was supported by the SFI Visual Intelligence, project no. 309439 of the Research Council of Norway.
}

\clearpage
%
%

\clearpage

\RestyleAlgo{ruled}
\SetKwComment{Comment}{/*}{*/}
\SetKwInput{KwInput}{Input}                
\SetKwInput{KwOutput}{Output}              

\begin{center}
{\Large\bfseries Supplementary Materials}

\end{center}

\section*{Contents}
This Supplementary provides additional experiments, analysis, discussion and code / reproducibility details to further support our findings. 
The Supplementary materials are organized as follows:

\begin{itemize}

        \item Section \ref{sec_supp:lrp_base}: A brief overview of the LRP-algorithm used
    
    \item Section \ref{sec_supp:computational_complexity}: Computational complexity of \textit{FF-RS ($\omega$)} / LRP-max.
       
    \item Section \ref{sec_supp:color_tff_non_color}: Non Color-conditional \textit{T-FF}

    \item Section \ref{sec_supp:k_hyper-parameter}: $k$ hyper-parameter in top-$k$ for \textit{T-FF}
    
    \item Section \ref{sec_supp:biggan_transfer}: Cross-model forensic transfer using BigGAN \cite{brock2018large} pre-training dataset
    
    \item Section \ref{sec_supp:performance_degrade}: Is the performance degrade in universal detectors due to unseen corruptions (OOD)?
    
    \item Section \ref{sec_supp:color_tff}: Color-conditional \textit{T-FF} (Additional Results)
    
    \item Section \ref{sec_supp:cr_universal_detectors}: CR-Universal Detectors (Additional Results)

    \item Section \ref{sec_supp:pixel-wise_explanations}: Pixel-wise explanations are not informative to discover \textit{T-FF} (Additional Results)
 
    \item Section \ref{sec_supp:reproducibility}: Research Reproducibility / Code Details
    
    \item Section \ref{sec_supp:future_work}: Future Work: Can we identify globally relevant channels for counterfeit detection in a Generator?

\end{itemize}

\appendix
\setcounter{figure}{0} 
\setcounter{table}{0} 
\renewcommand\thefigure{\thesection.\arabic{figure}}
\renewcommand\theHfigure{\thesection.\arabic{figure}}
\renewcommand\thetable{\thesection.\arabic{table}}  
\renewcommand\theHtable{\thesection.\arabic{table}}  


\section{A brief overview of the LRP-algorithm used}
\label{sec_supp:lrp_base}

\setcounter{figure}{0} 
\setcounter{table}{0} 


Layer-wise relevance propagation (LRP) \cite{bach2015pixel} is a modified-gradient type algorithm for backward passes in neural networks and other models. LRP is based on the idea of replacing the partial derivatives, which are usually flowing back along the edges of a graph, by terms derived from Taylor decompositions for single layers
\cite{MONTAVON2017211}
of the network. While the $\epsilon$-LRP-rule is similar to gradient-times-input, other rules such as the $\beta$-rule
\cite{Montavon2019}
result in explanations which exhibit visually low noise and are robust to gradient shattering effects
\cite{DBLP:conf/icml/BalduzziFLLMM17}
common in deep neural networks due to its normalization properties. 
Consider a neuron $y$ with inputs $x_i$, weights $w_i$, and a relevance score being already computed for its output being $R_y$. The relevance score $R_y$ is the analogue for the total derivative $\frac{dz}{dy}$ in conventional backpropagation started at output logits, however computed using LRP. Then the relevance score for the input $x_i$ according to the $\beta=0$-rule is given as
\begin{align}
R_i =  R_y  \frac{ (w_ix_i)_+ }{ \sum_k (w_kx_k)_+} 
\end{align}
where $(\cdot)_+$ is the positive part.
This measures the proportion of the positive part of the weighted input $(w_ix_i)_+$ for the input neuron $i$ relative to the positive weighted inputs from all inputs used to compute the value of neuron $y$. Therefore it redistributes relevance from an output to the inputs proportional to this fraction and proportional to the relevance $R_y$ of the output neuron.
We used the $\beta=0$-rule for all convolution layers and the $\epsilon$-rule for the top-most fully connected layer. Before applying LRP, we fuse batchnorm layers into convolution layers and reset the batchnorm layers. The backpropagation in the resetted batchnorm layers uses the identity. Technically the base LRP algorithm is implemented in PyTorch as custom static autograd functions. This results for convolution layers in relevance scores having a shape of $(1,C,H,W)$ in the gradient field.

LRP scores computed in the input space of neural networks have been shown to perform well on metrics regarding the ordering of input space regions according to the computed explanation scores and the correlation of this ordering to changes in model output logits
\cite{DBLP:journals/tnn/SamekBMLM17,DBLP:conf/acl/SchutzeRP18,DBLP:conf/blackboxnlp/ArrasOMS19}
when modifying the highest scoring regions.

\section{Computational Complexity of \textit{FF-RS ($\omega$)} / LRP-max}
\label{sec_supp:computational_complexity}

Both \textit{FF-RS ($\omega$)} and LRP-max require an additional forward and backward pass during computation. We emphasize that \textit{FF-RS ($\omega$)} and LRP-max are not used during training, and are only used for analysis / interpretability. Therefore, computational overhead is not substantial. All our experiments were performed using a single Nvidia RTX 3090 GPU.

\section{Non Color-conditional T-FF}
\label{sec_supp:color_tff_non_color}
\setcounter{figure}{0} 
\setcounter{table}{0} 

\begin{figure}
\centering
\begin{tabular}{ccccccc}
    \multicolumn{1}{p{0.125\linewidth}}{\tiny \enskip ProGAN \cite{karras2018progressive}} &
    \multicolumn{1}{p{0.15\linewidth}}{\tiny  \enskip StyleGAN2 \cite{Karras_2020_CVPR}} &
    \multicolumn{1}{p{0.14\linewidth}}{\tiny StyleGAN \cite{Karras_2019_CVPR}} &
    \multicolumn{1}{p{0.125\linewidth}}{\tiny BigGAN \cite{brock2018large}} &
    \multicolumn{1}{p{0.132\linewidth}}{\tiny CycleGAN \cite{zhu2017unpaired}} &
    \multicolumn{1}{p{0.135\linewidth}}{\tiny \enskip StarGAN \cite{choi2018stargan}} &
    {\tiny GauGAN \cite{park2019semantic}} \\
    
    \multicolumn{7}{c}{\includegraphics[width=0.99\linewidth]{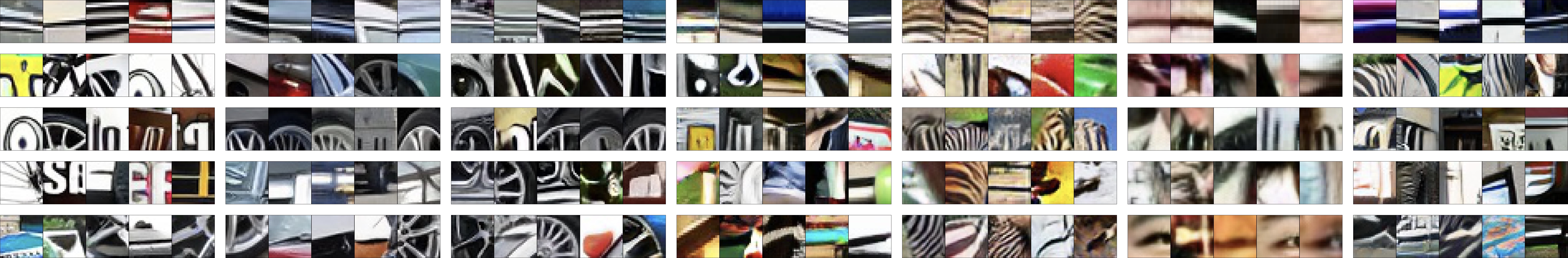}}

\end{tabular}
\caption{
\textit{T-FF} that are \textit{not} color-conditional in ResNet-50 Universal detector:
We show the LRP-max response regions for 5 \textit{non} color-conditional T-FF for ProGAN 
\cite{karras2018progressive}
and all 6 unseen GANs 
\cite{Karras_2020_CVPR,Karras_2019_CVPR,brock2018large,zhu2017unpaired,choi2018stargan,park2019semantic}.
Each row represents a \textit{non} color-conditional T-FF.
We emphasize that \textit{T-FF} are discovered using our proposed \textit{forensic feature relevance statistic (FF-RS)}.
This detector is trained with ProGAN
\cite{karras2018progressive} 
counterfeits \cite{Wang_2020_CVPR} and cross-model forensic transfer is evaluated on other unseen GANs.
All counterfeits are obtained from the ForenSynths dataset 
\cite{Wang_2020_CVPR}.
Visual inspection of LRP-max regions of \textit{non} color-conditional \textit{T-FF} shows frequency / texture artifacts.
i.e.: rapid changes in pixel intensities.
This shows that the universal detector also uses frequency / texture artifacts for cross-model transfer although \textit{color is a critical T-FF} as $\approx 85\%$ of \textit{T-FF} are color-conditional.
We emphasize that our proposed method is capable of identifying different types of \textit{T-FF} (i.e.: frequency / texture artifacts).
}
\label{fig_supp:lrp_patches_r50_non_color}
\end{figure}

\begin{figure}
\centering
\begin{tabular}{ccccccc}
    {\tiny ProGAN \cite{karras2018progressive}} &
    {\tiny StyleGAN2 \cite{Karras_2020_CVPR}} &
    {\tiny StyleGAN \cite{Karras_2019_CVPR}} &
    {\tiny BigGAN \cite{brock2018large}} &
    {\tiny CycleGAN \cite{zhu2017unpaired}} &
    {\tiny StarGAN \cite{choi2018stargan}} &
    {\tiny GauGAN \cite{park2019semantic}} \\

   \includegraphics[width=0.13\linewidth]{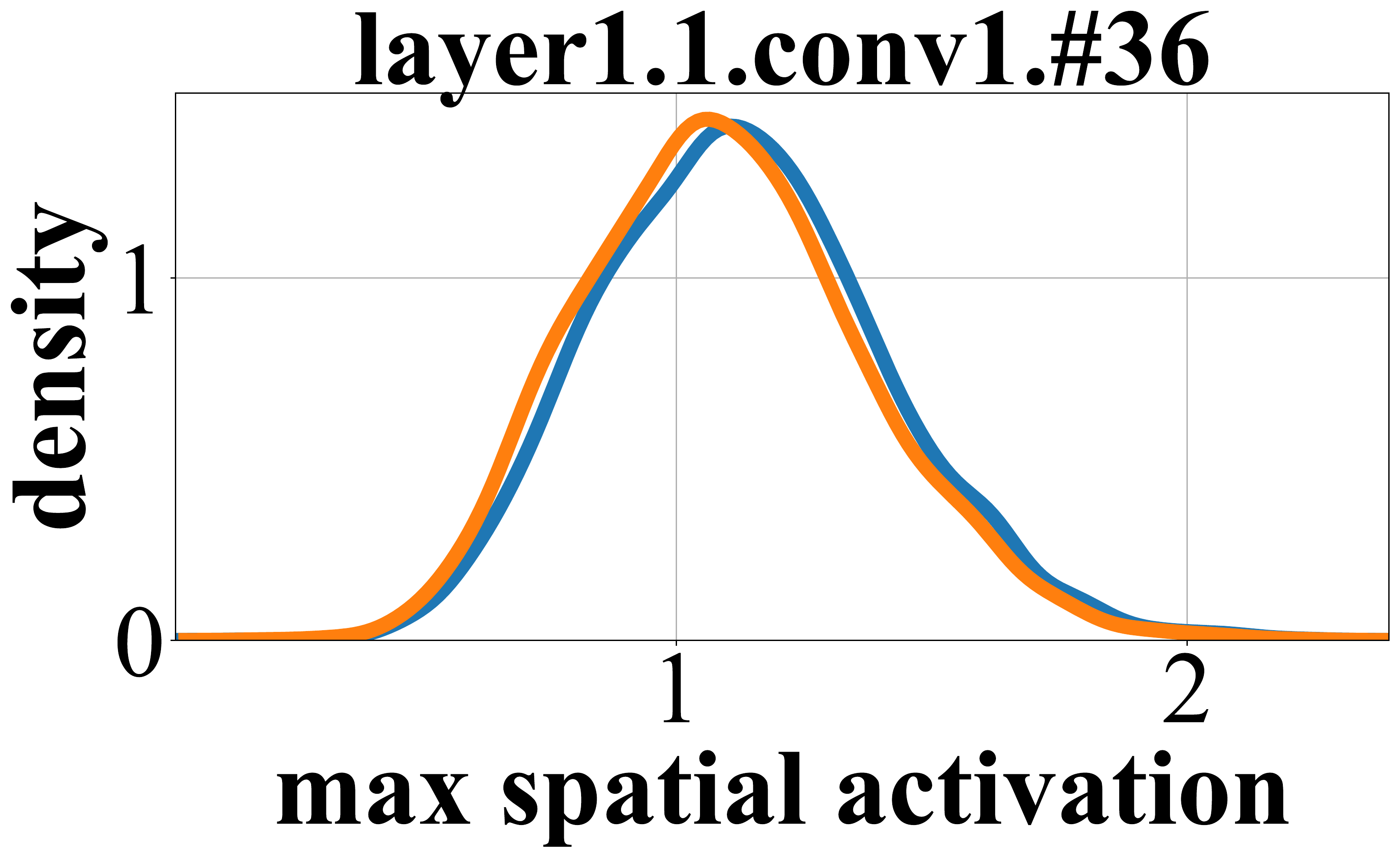} &
    \includegraphics[width=0.13\linewidth]{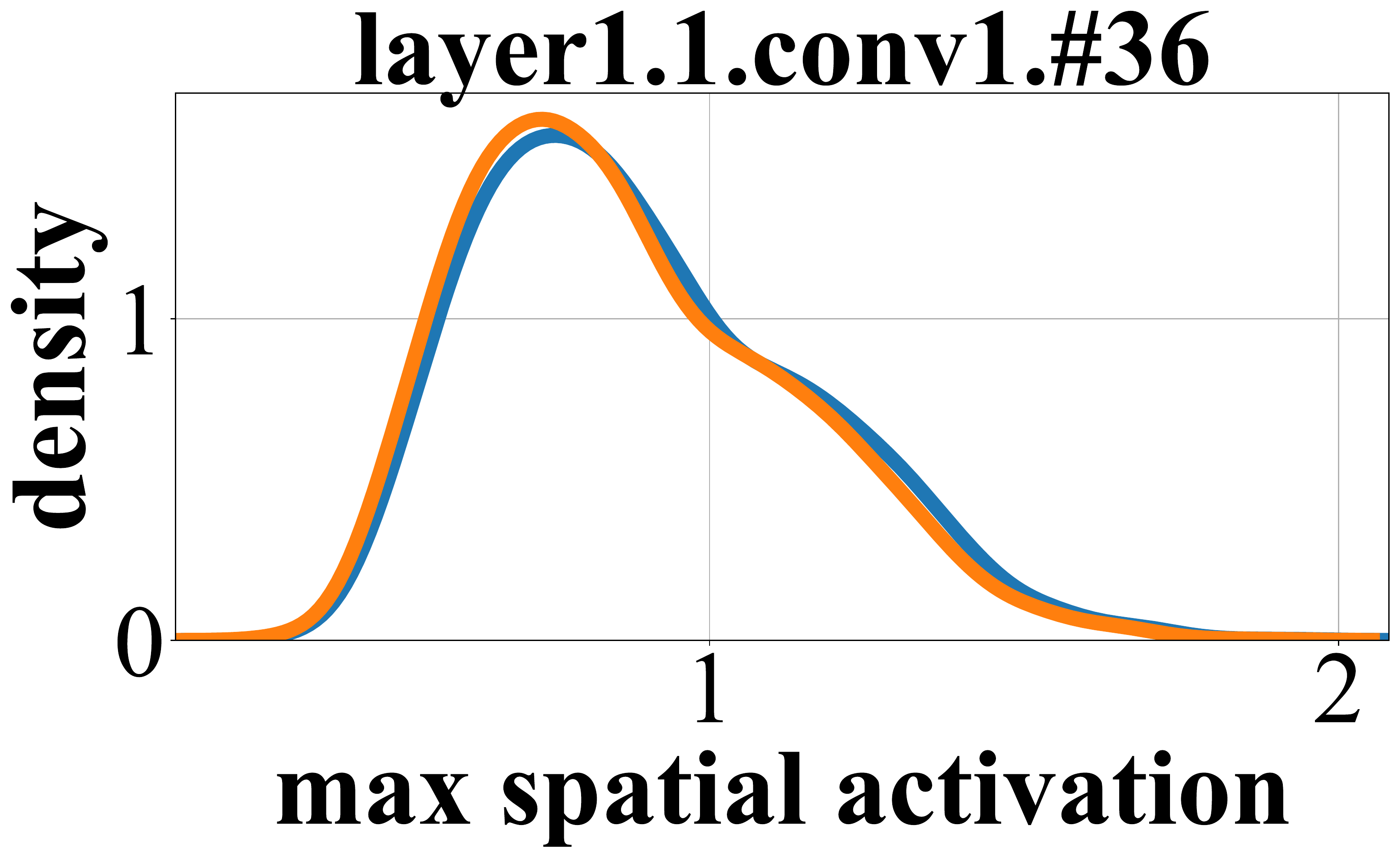} &
    \includegraphics[width=0.13\linewidth]{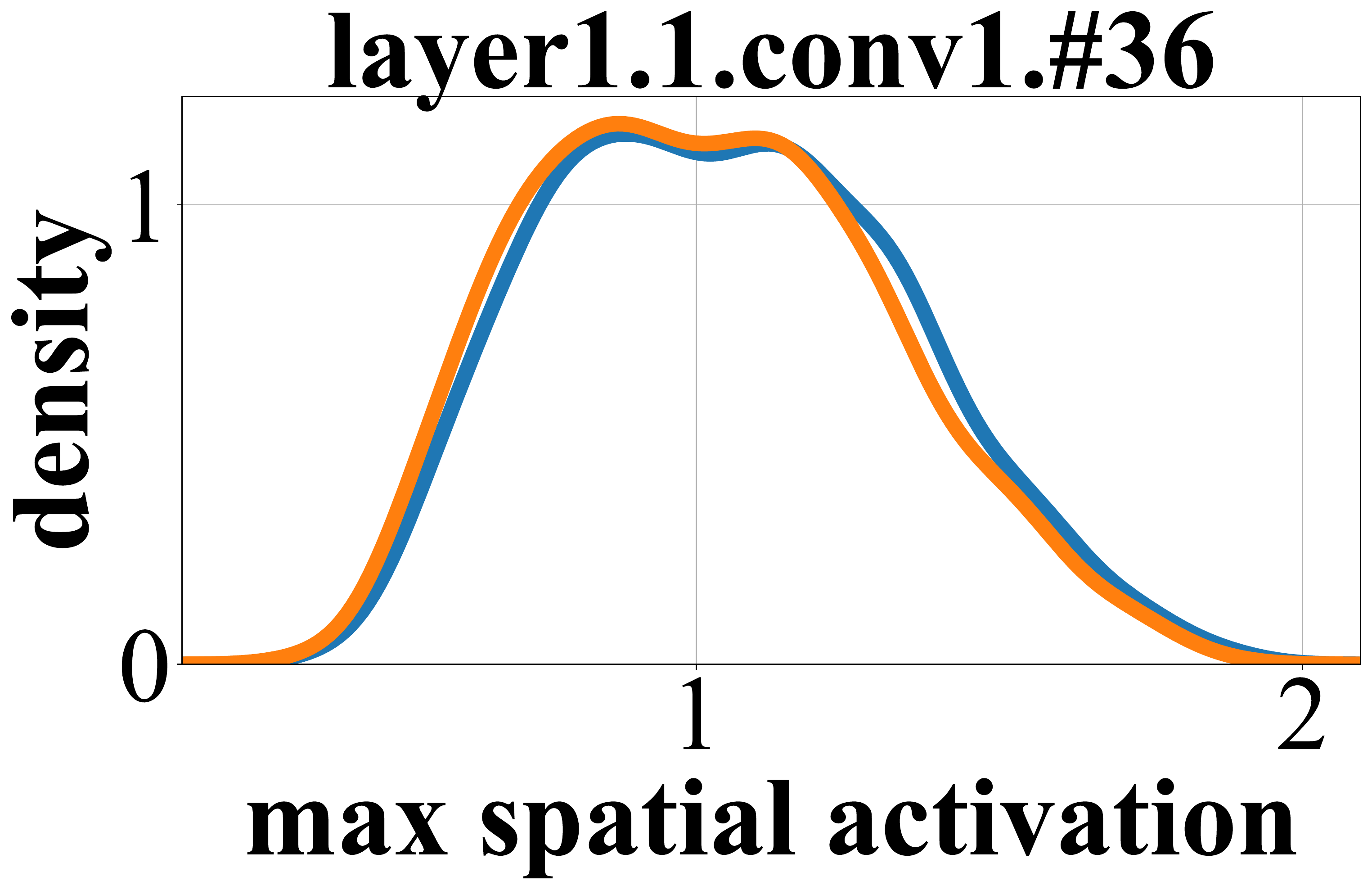} &
    \includegraphics[width=0.13\linewidth]{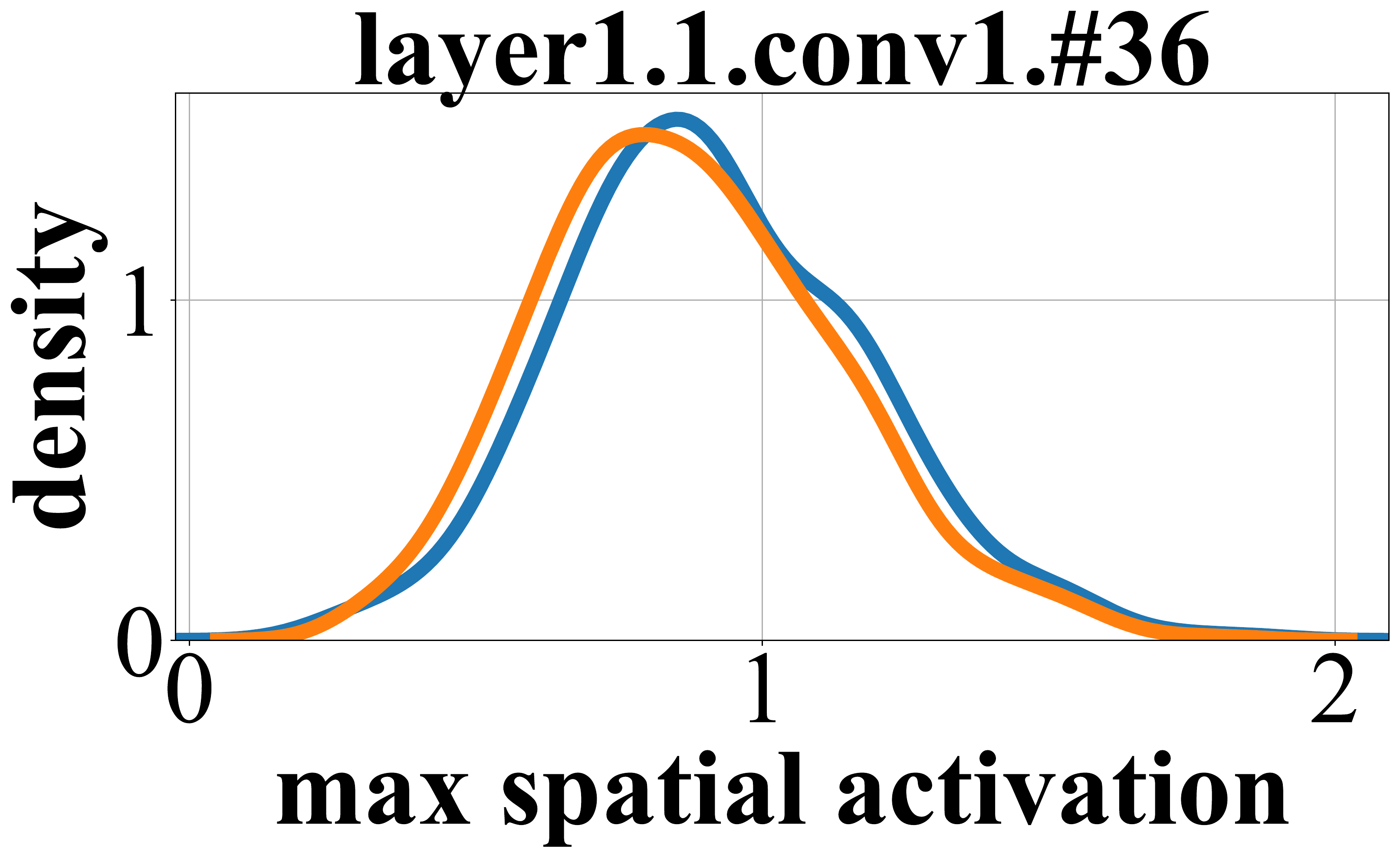} &
    \includegraphics[width=0.13\linewidth]{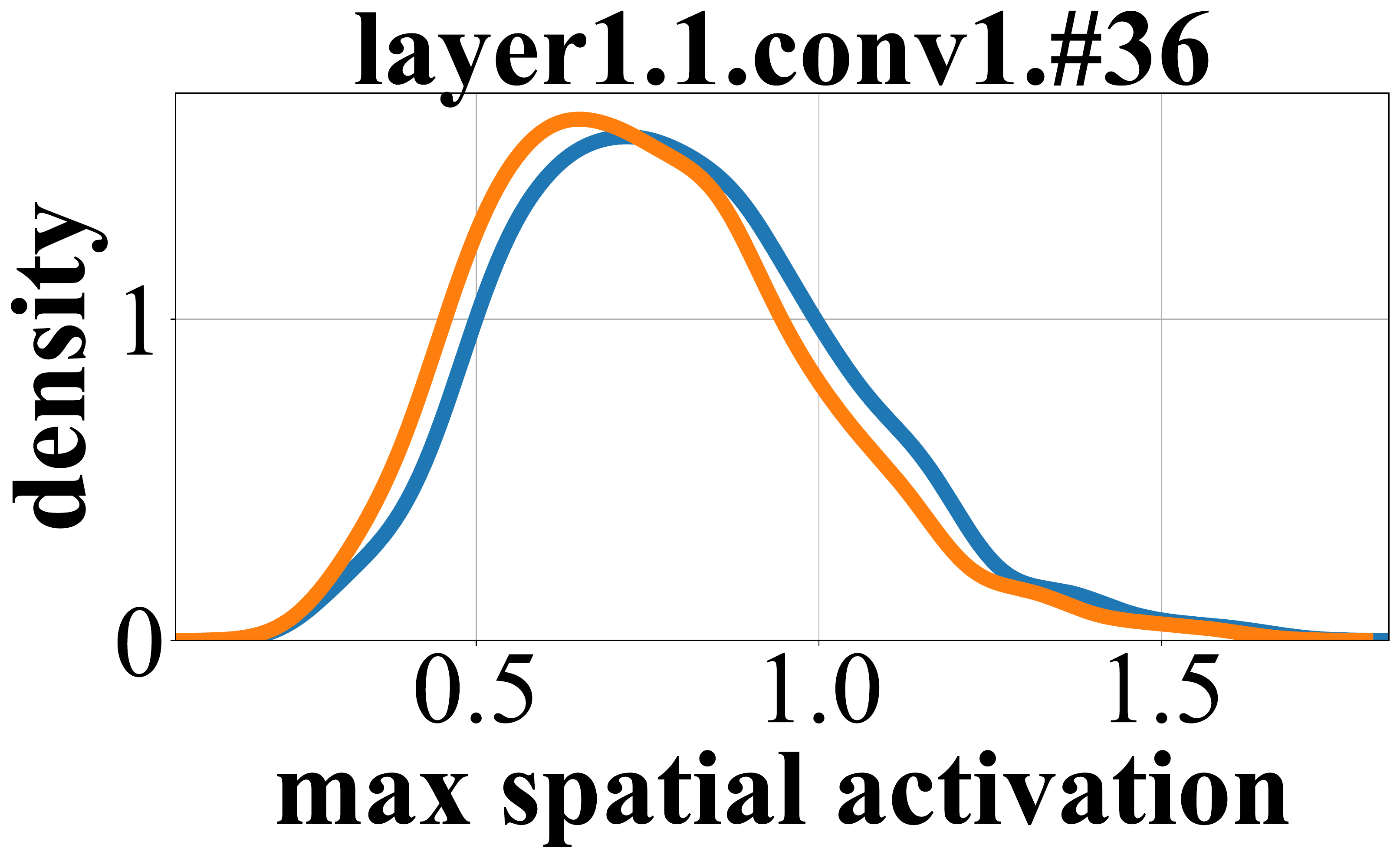} &
    \includegraphics[width=0.13\linewidth]{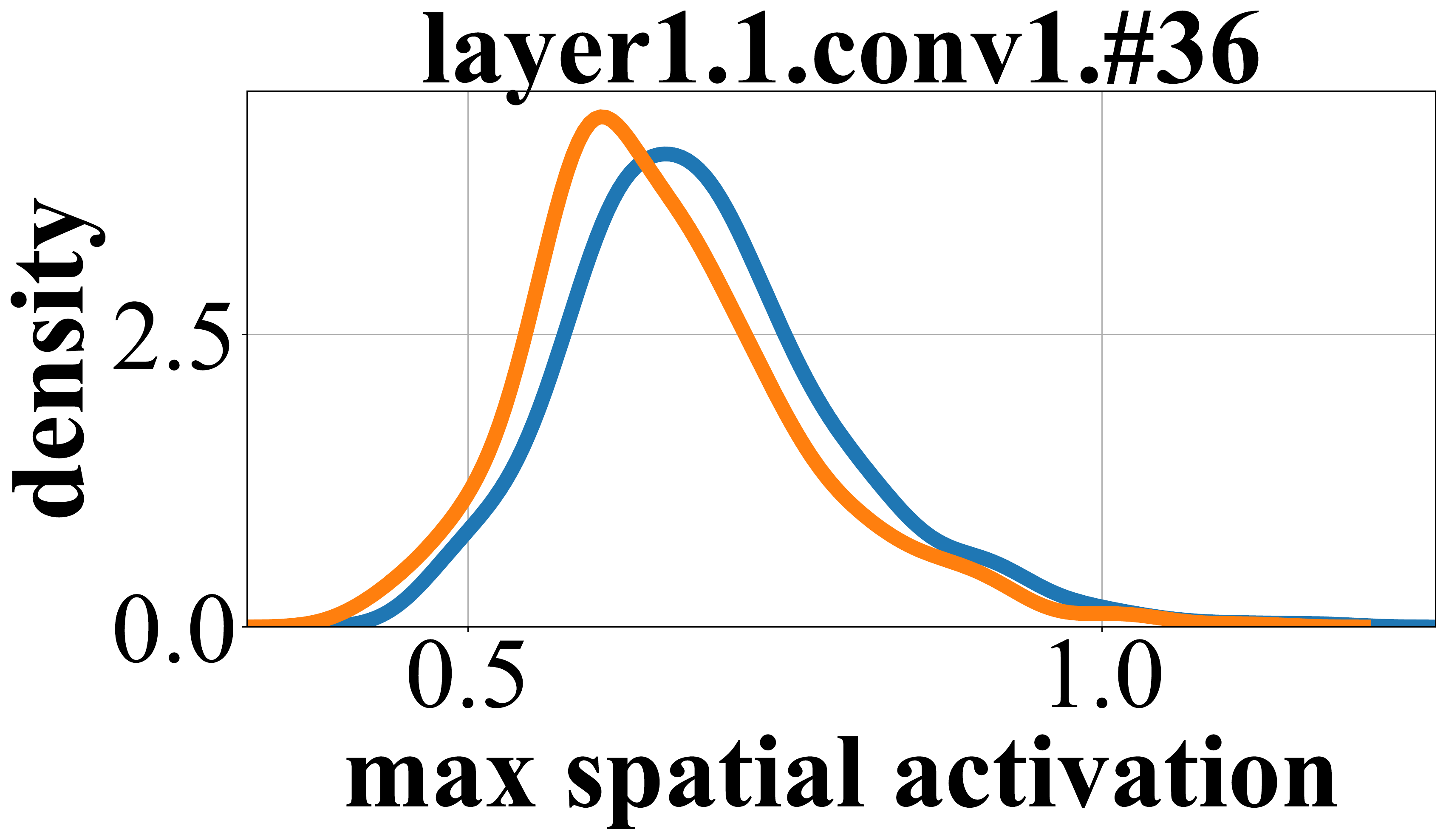} &
    \includegraphics[width=0.13\linewidth]{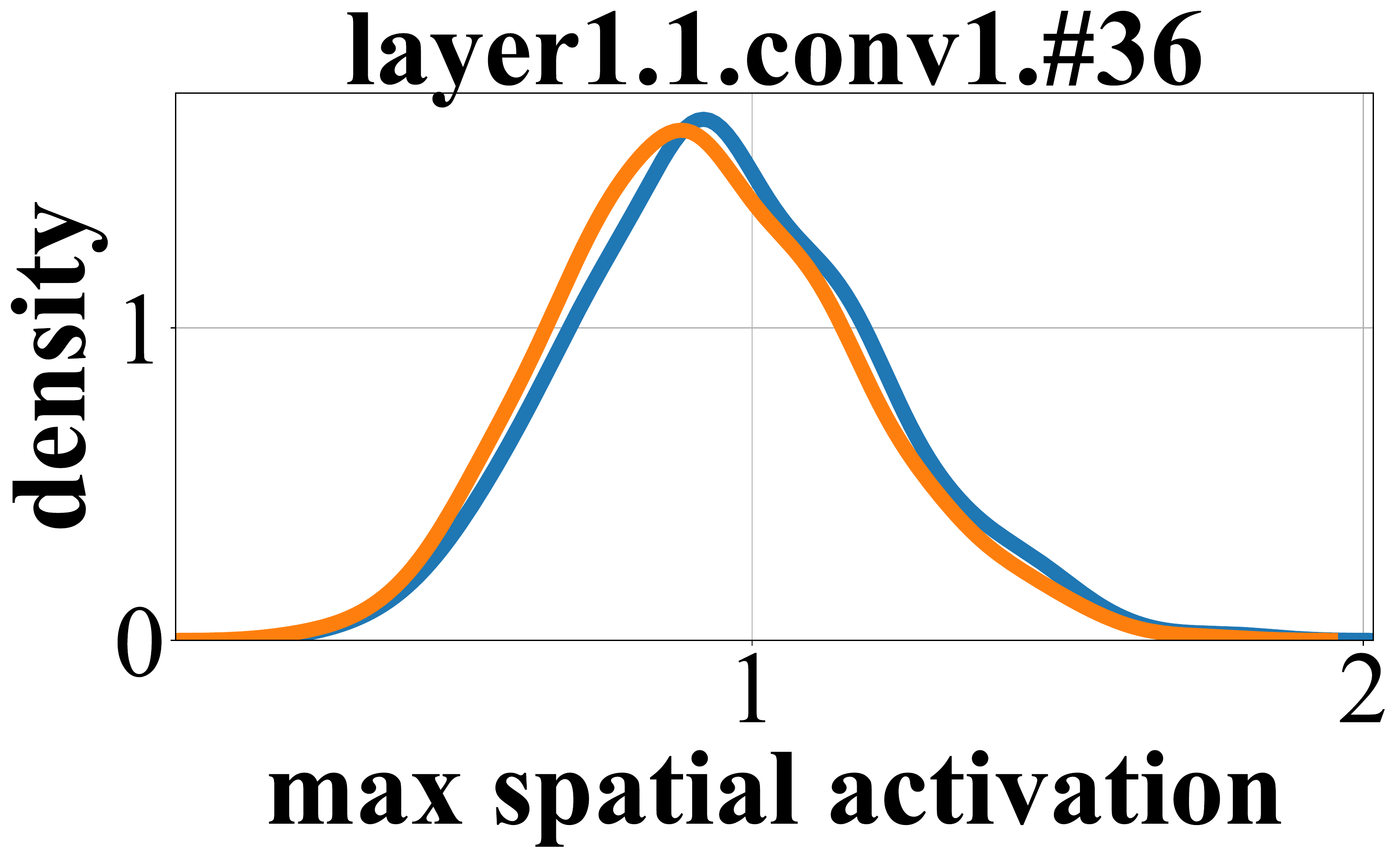} 
    \\
    
     \includegraphics[width=0.13\linewidth]{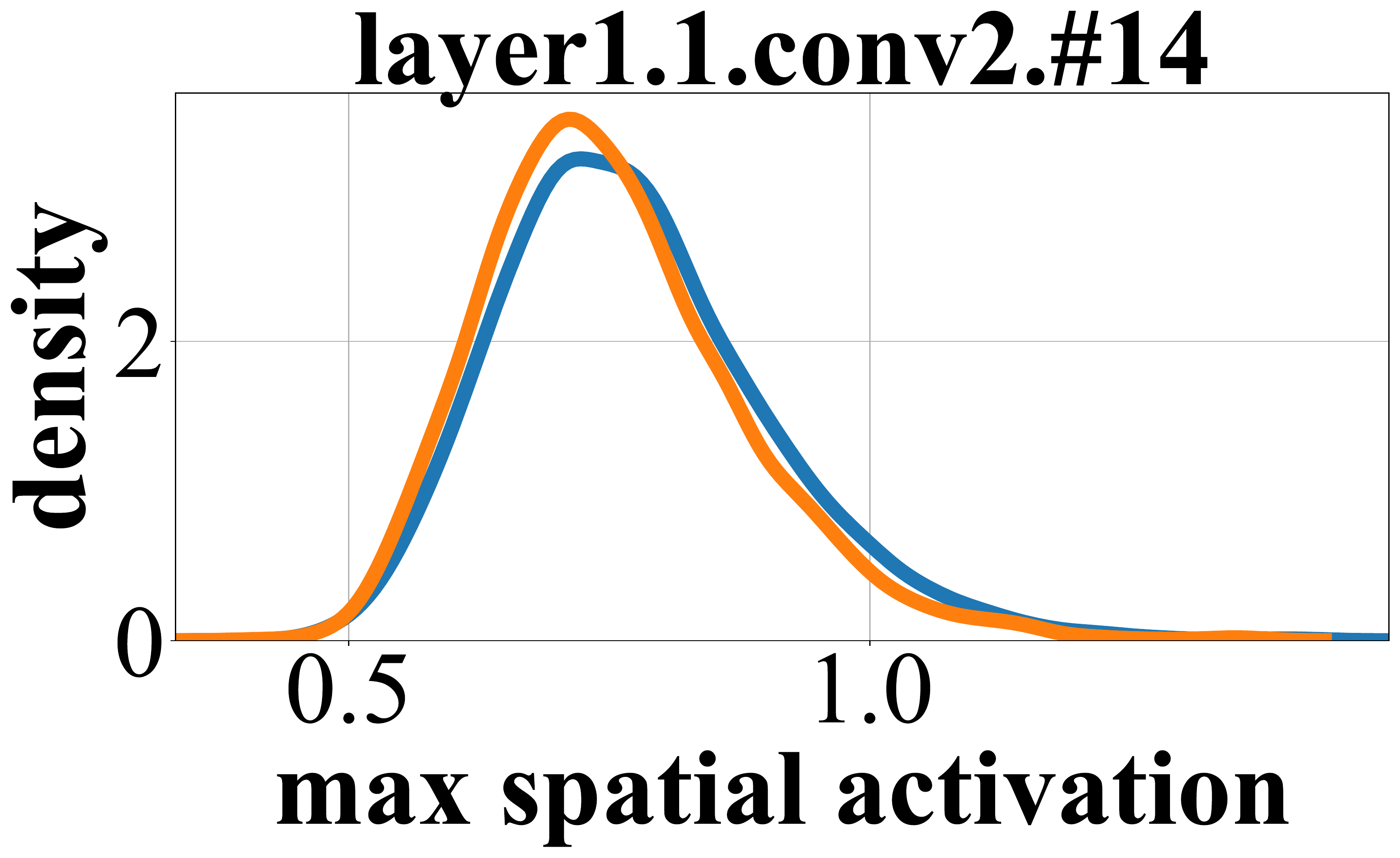} &
    \includegraphics[width=0.13\linewidth]{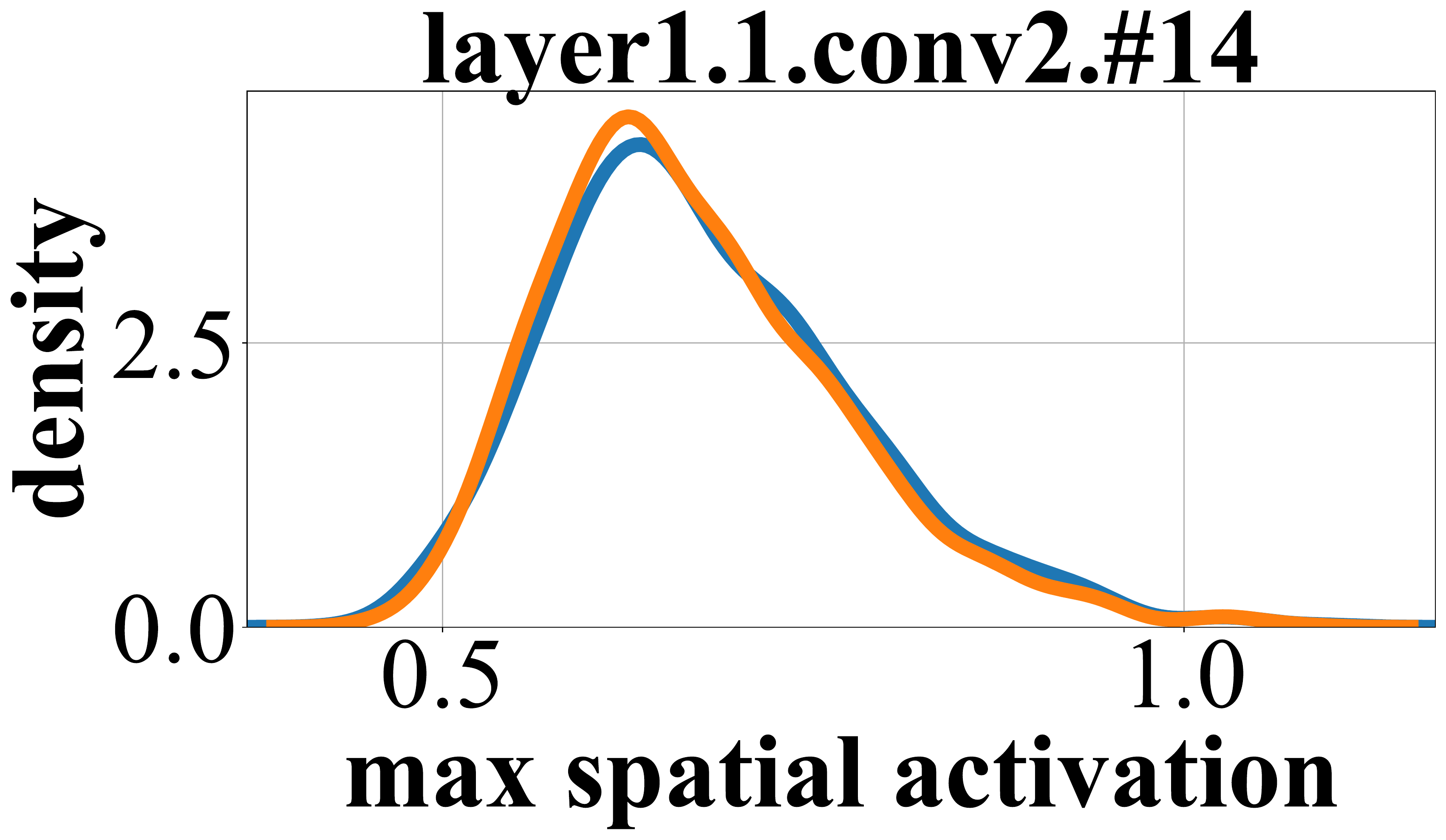} &
    \includegraphics[width=0.13\linewidth]{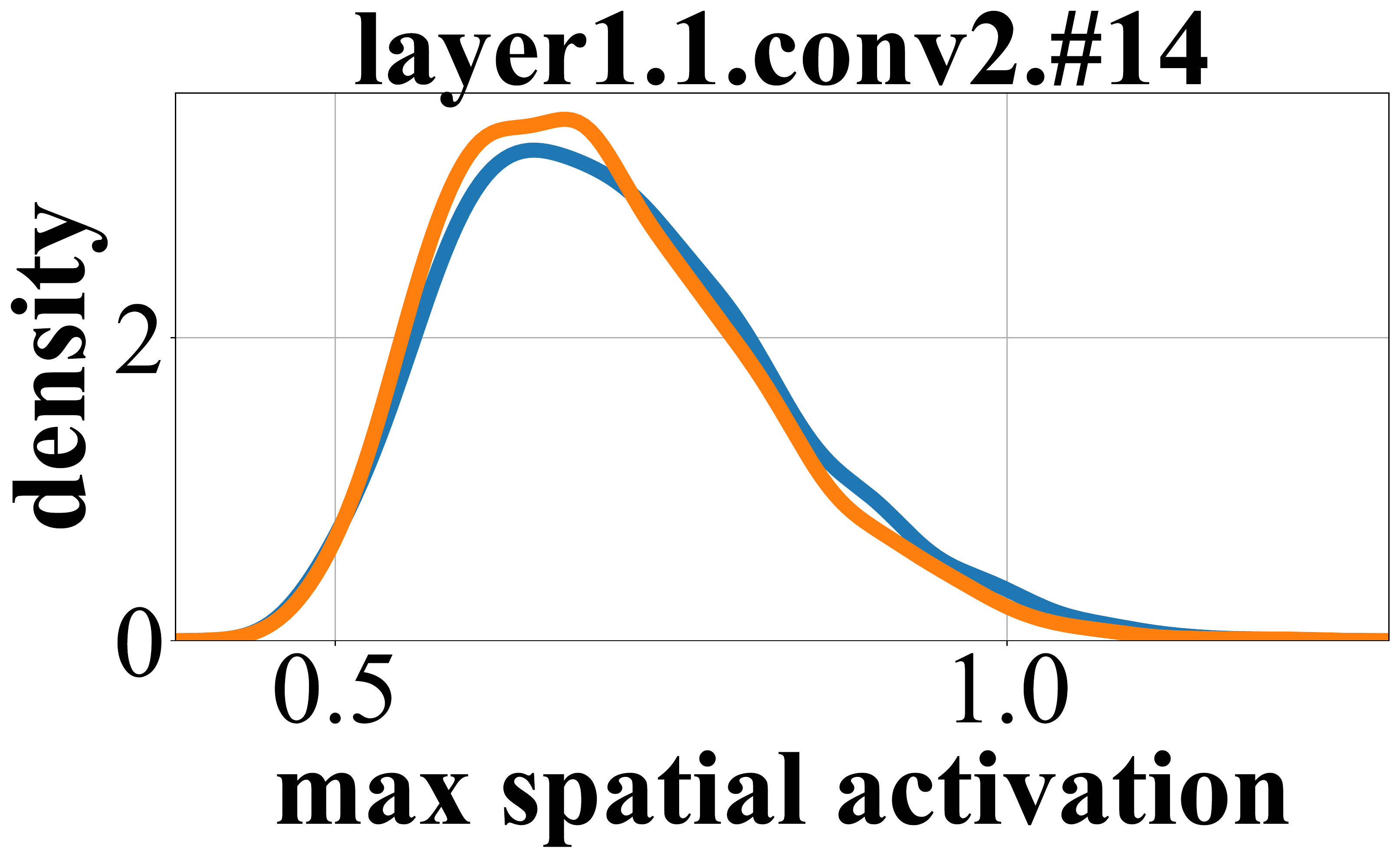} &
    \includegraphics[width=0.13\linewidth]{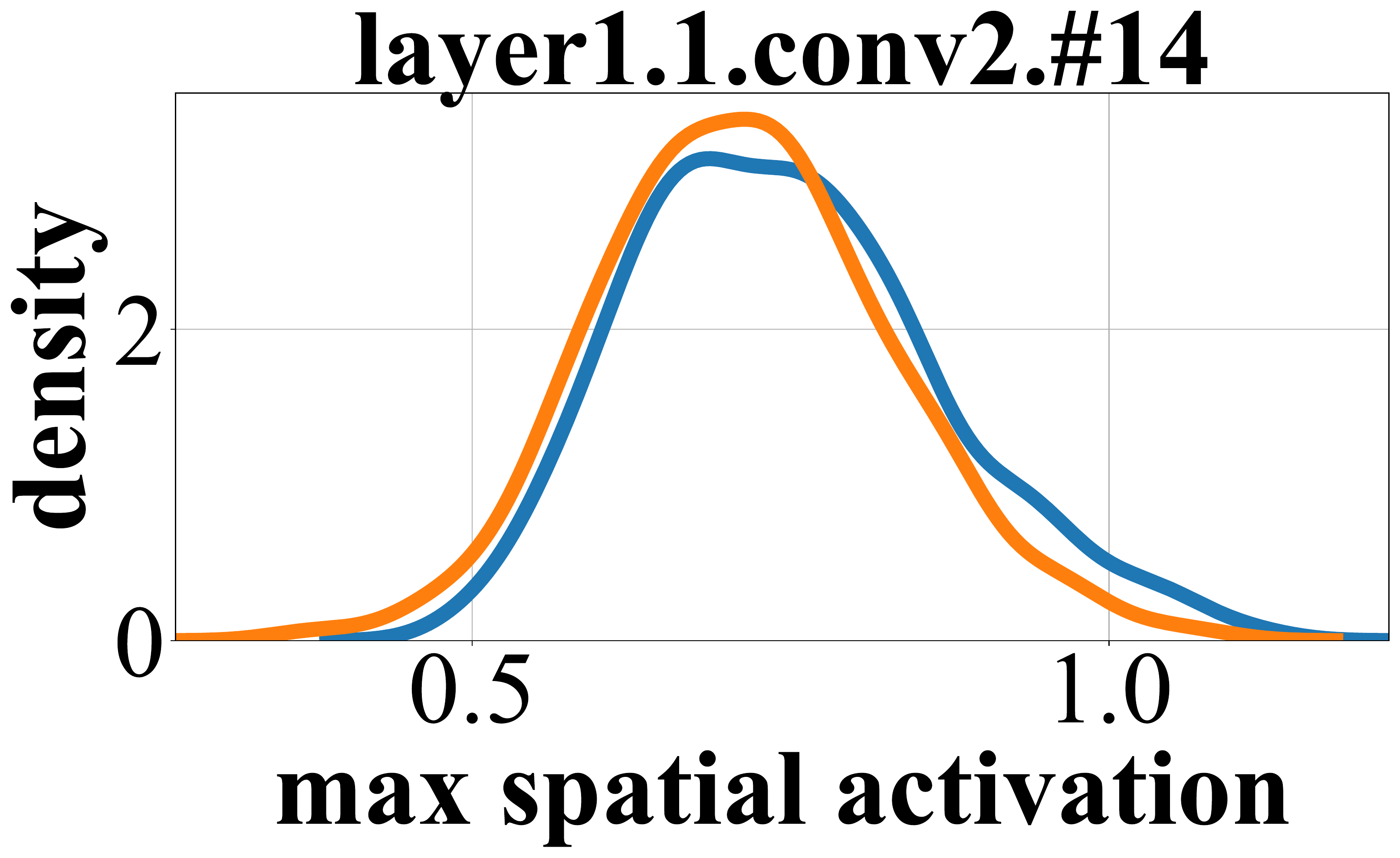} &
    \includegraphics[width=0.13\linewidth]{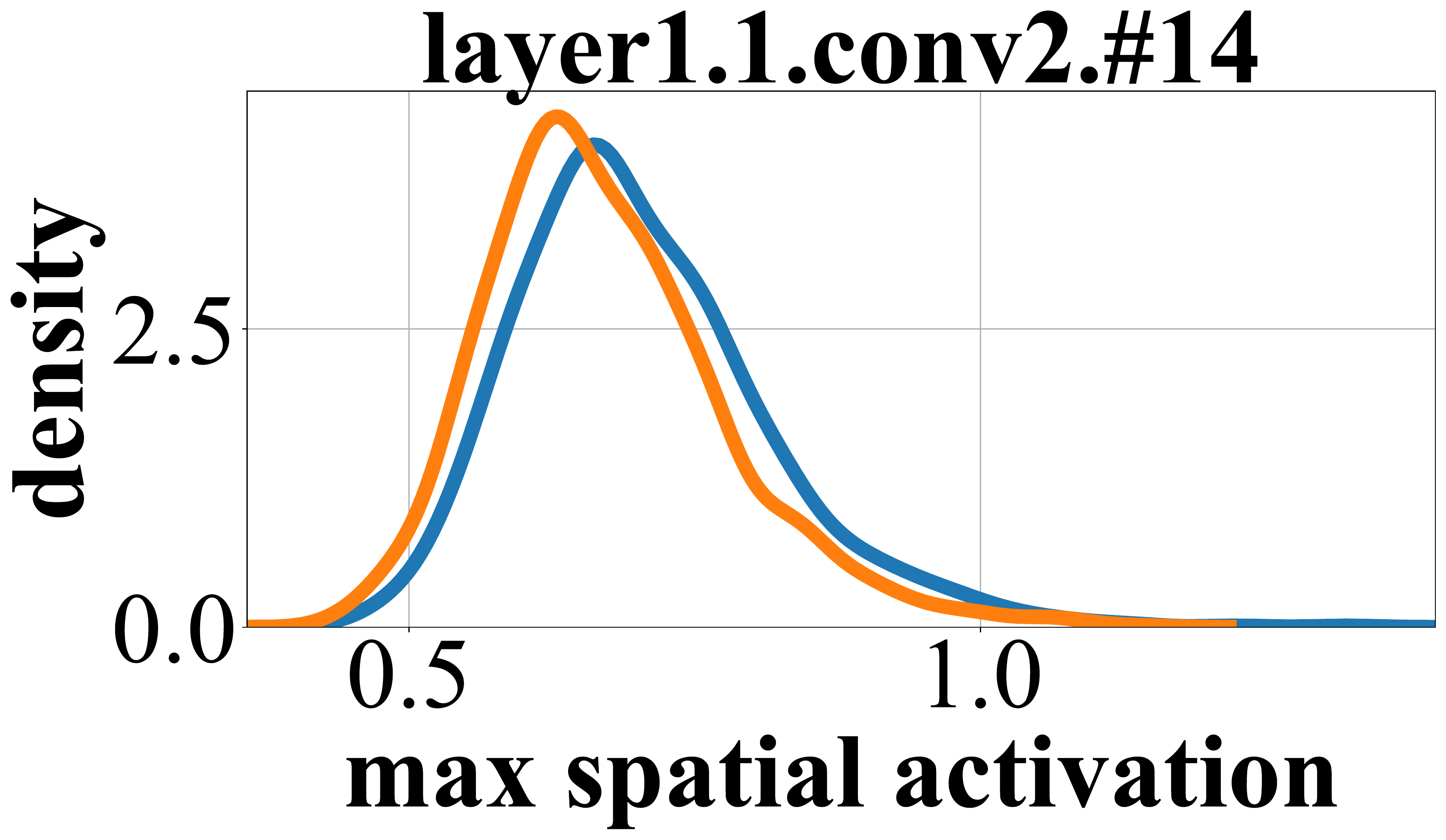} &
    \includegraphics[width=0.13\linewidth]{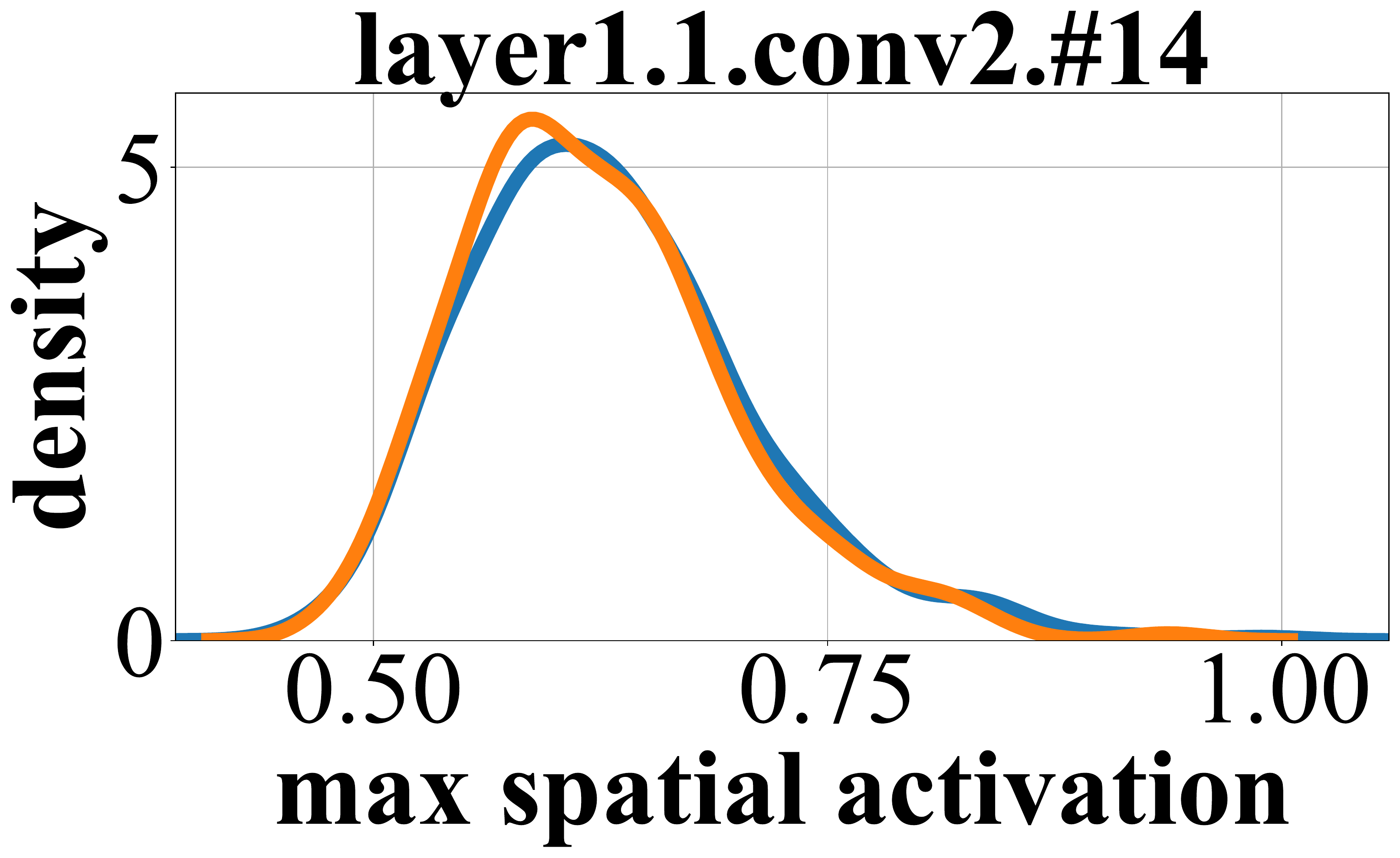} &
    \includegraphics[width=0.13\linewidth]{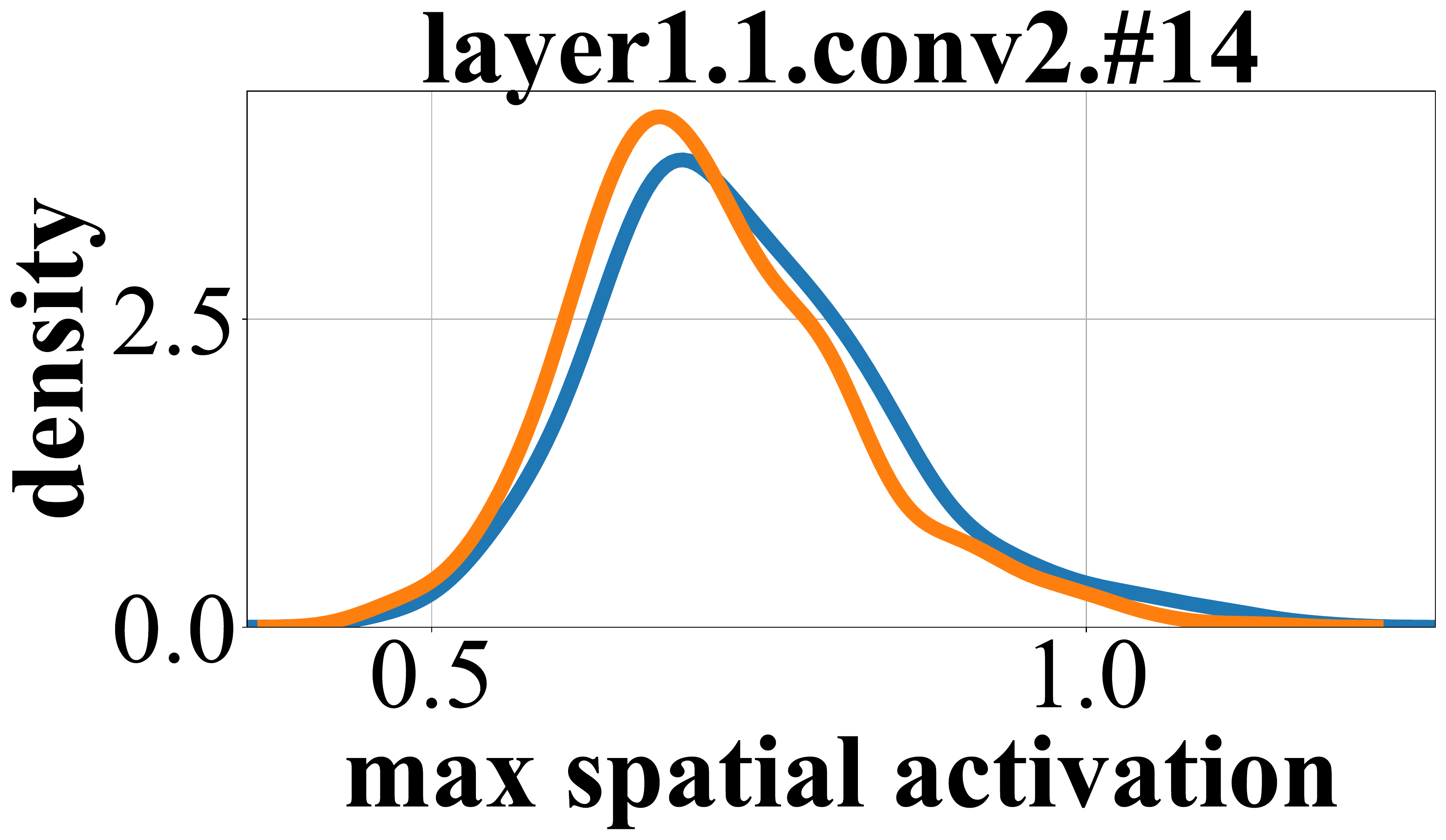} 
    \\
    
    \includegraphics[width=0.13\linewidth]{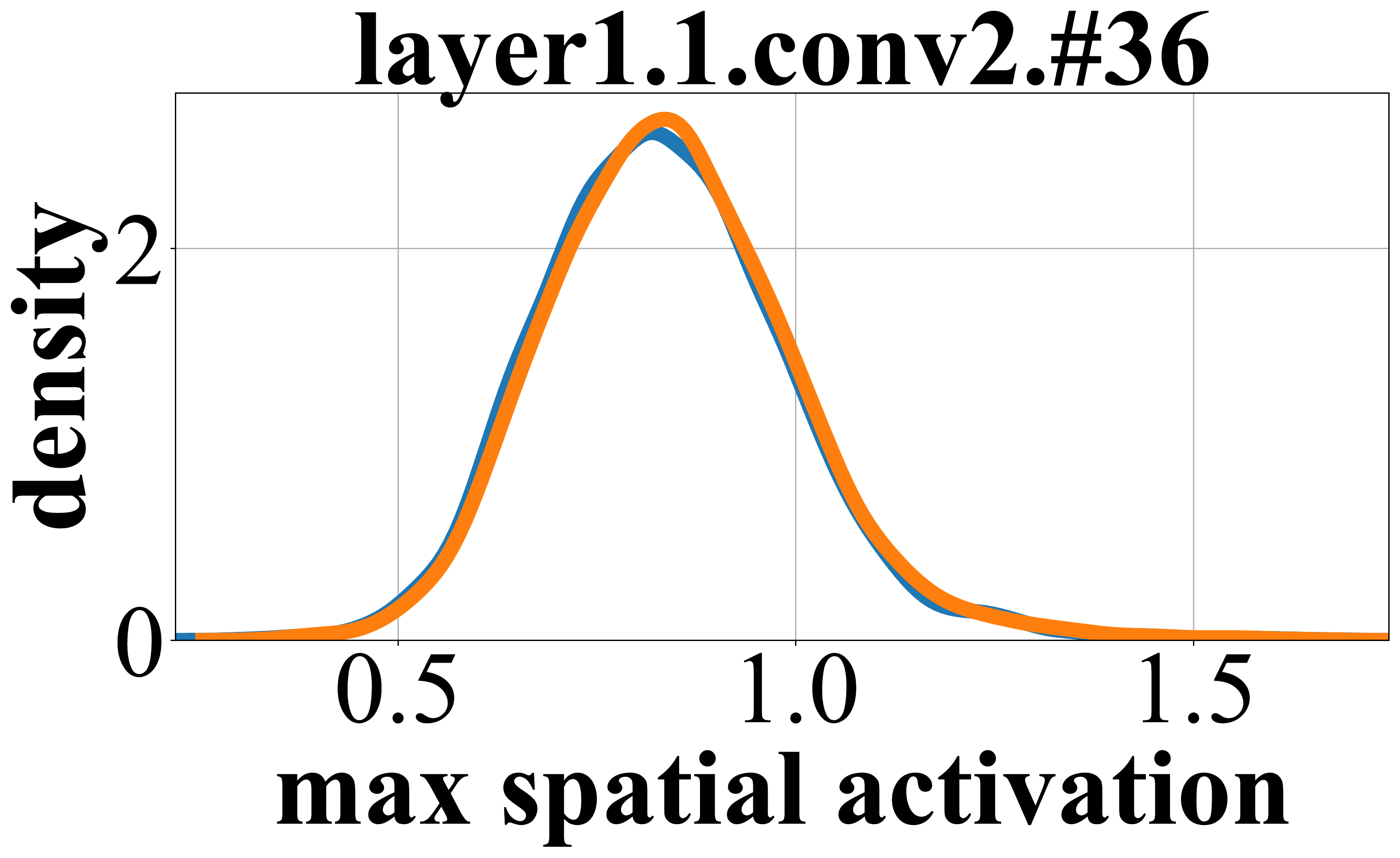} &
    \includegraphics[width=0.13\linewidth]{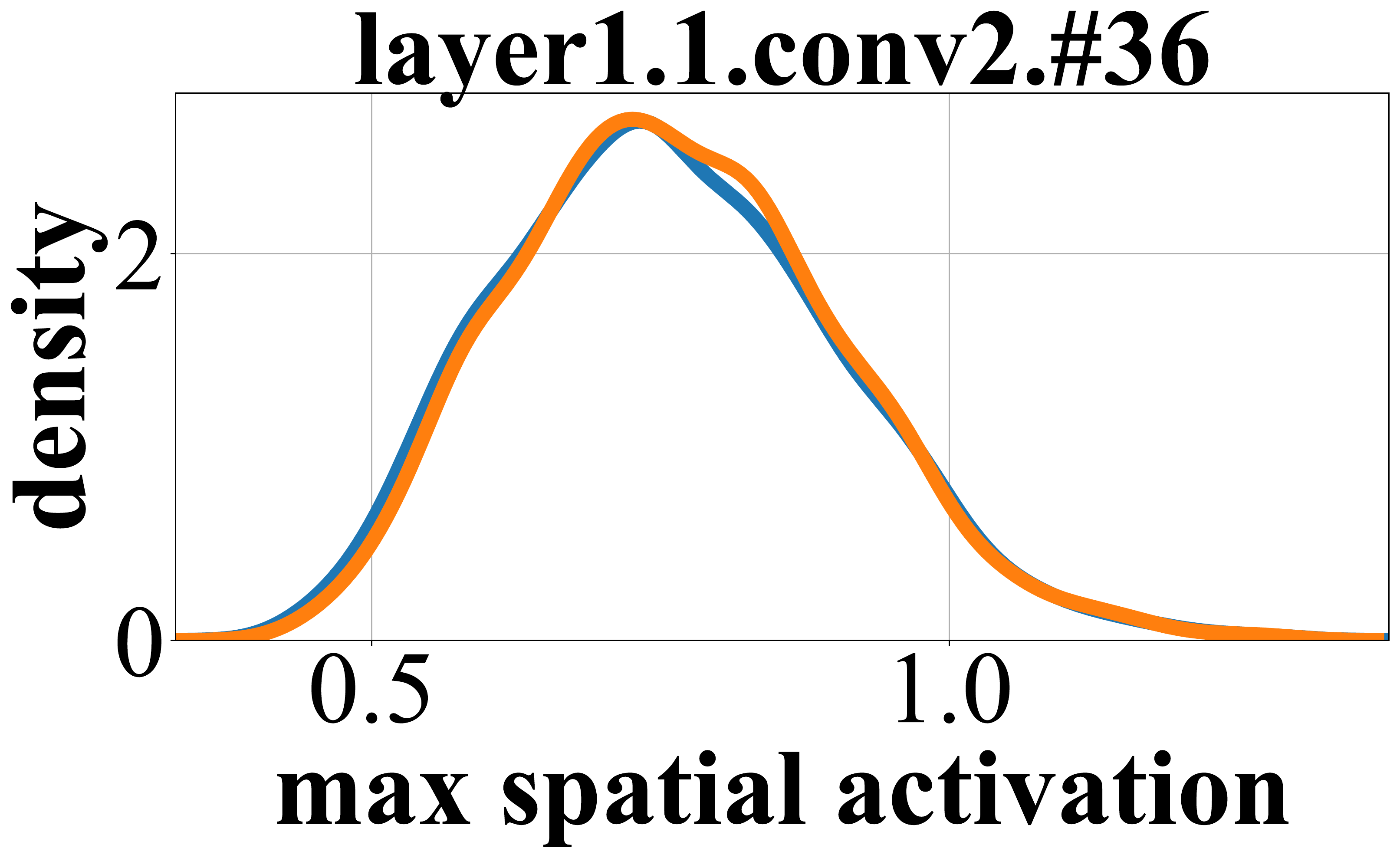} &
    \includegraphics[width=0.13\linewidth]{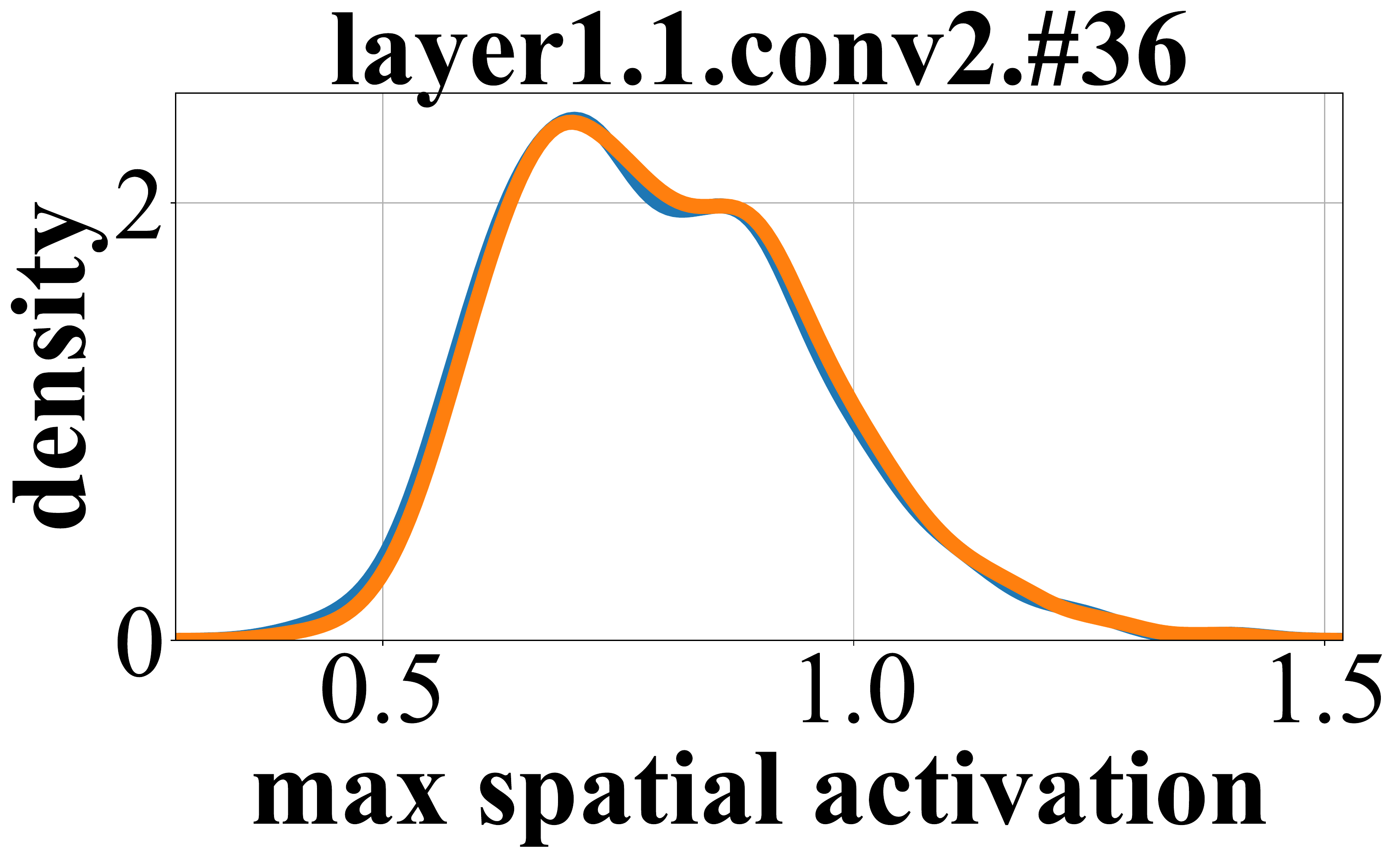} &
    \includegraphics[width=0.13\linewidth]{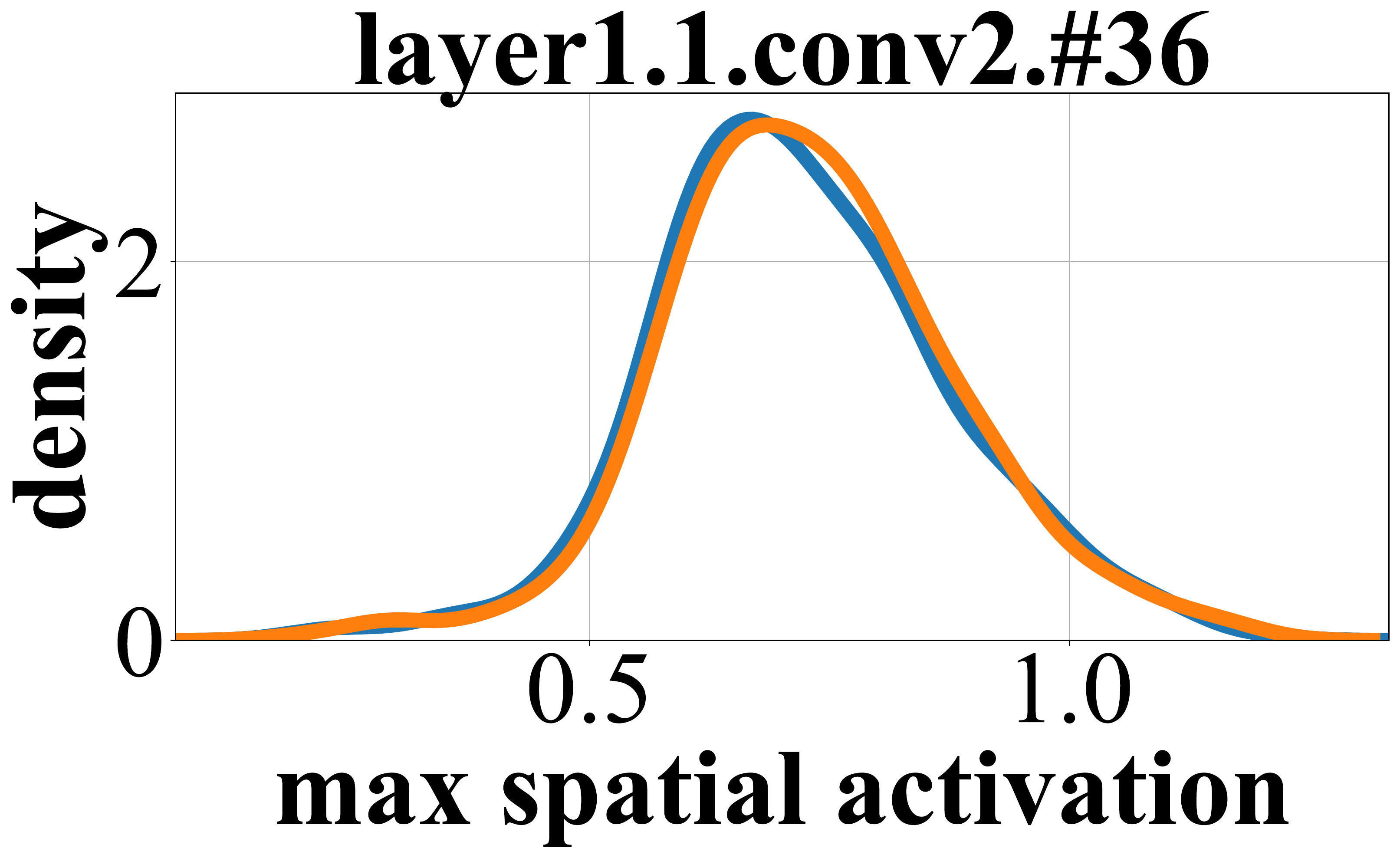} &
    \includegraphics[width=0.13\linewidth]{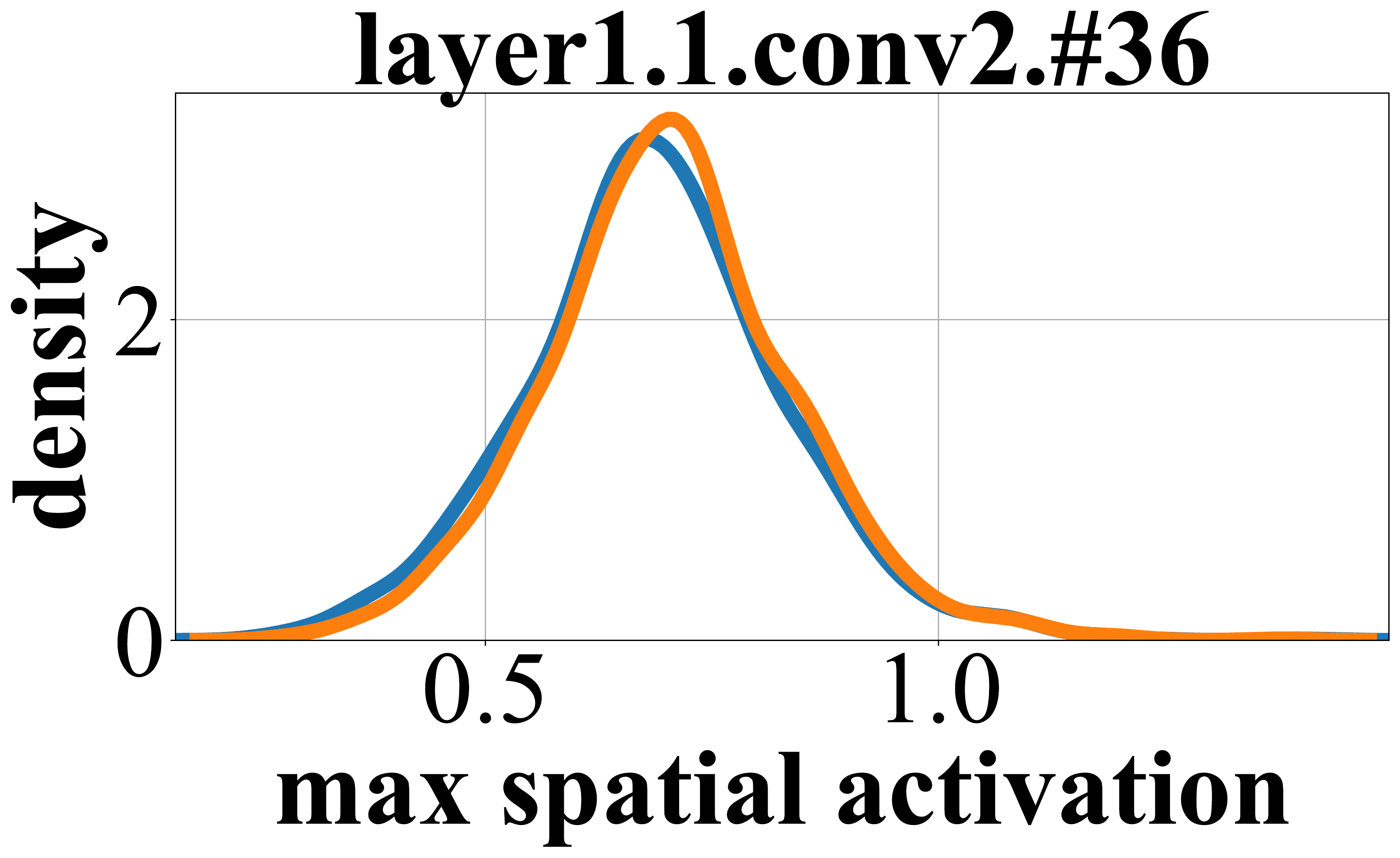} &
    \includegraphics[width=0.13\linewidth]{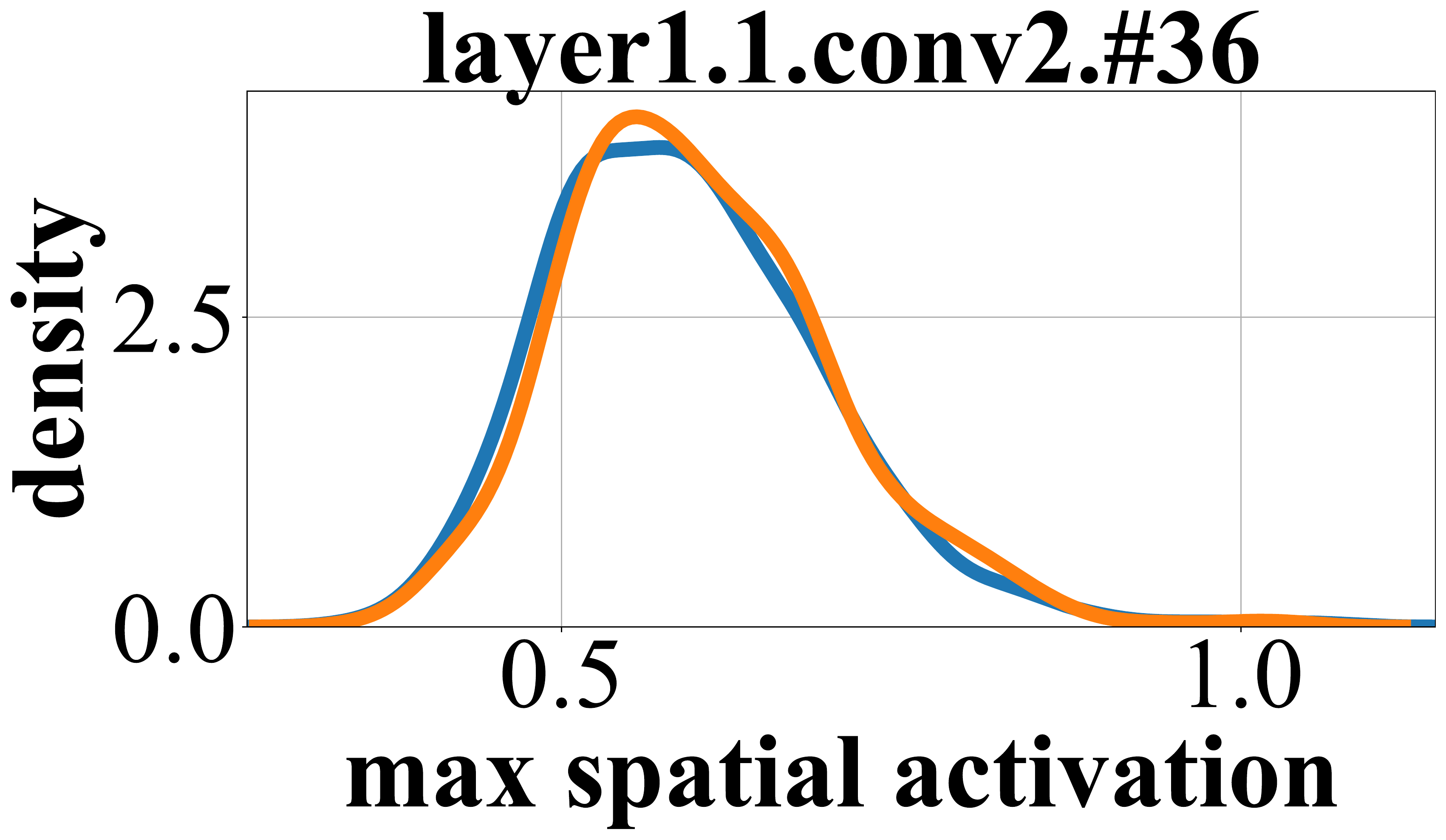} &
    \includegraphics[width=0.13\linewidth]{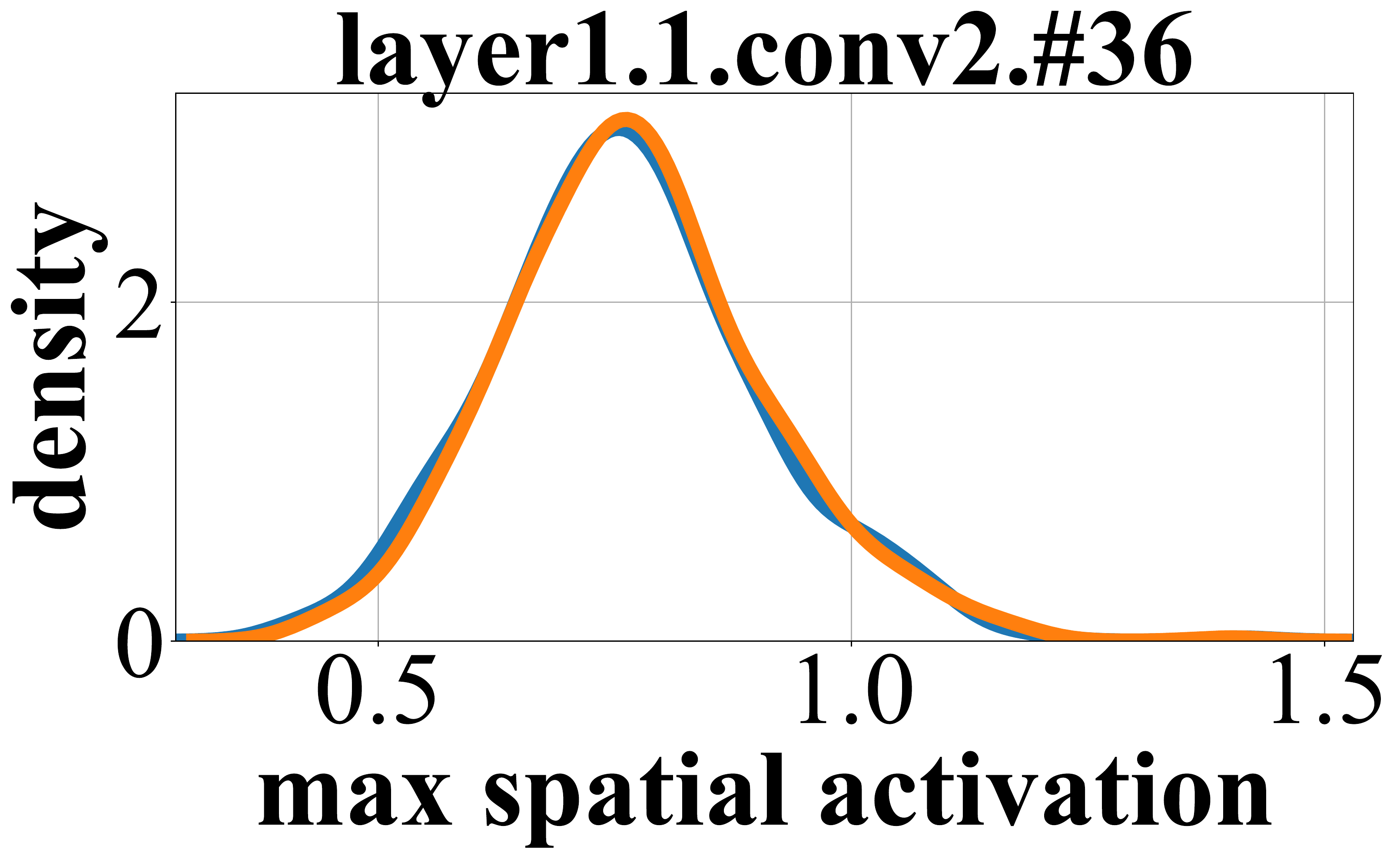} 
    \\
    
    \includegraphics[width=0.13\linewidth]{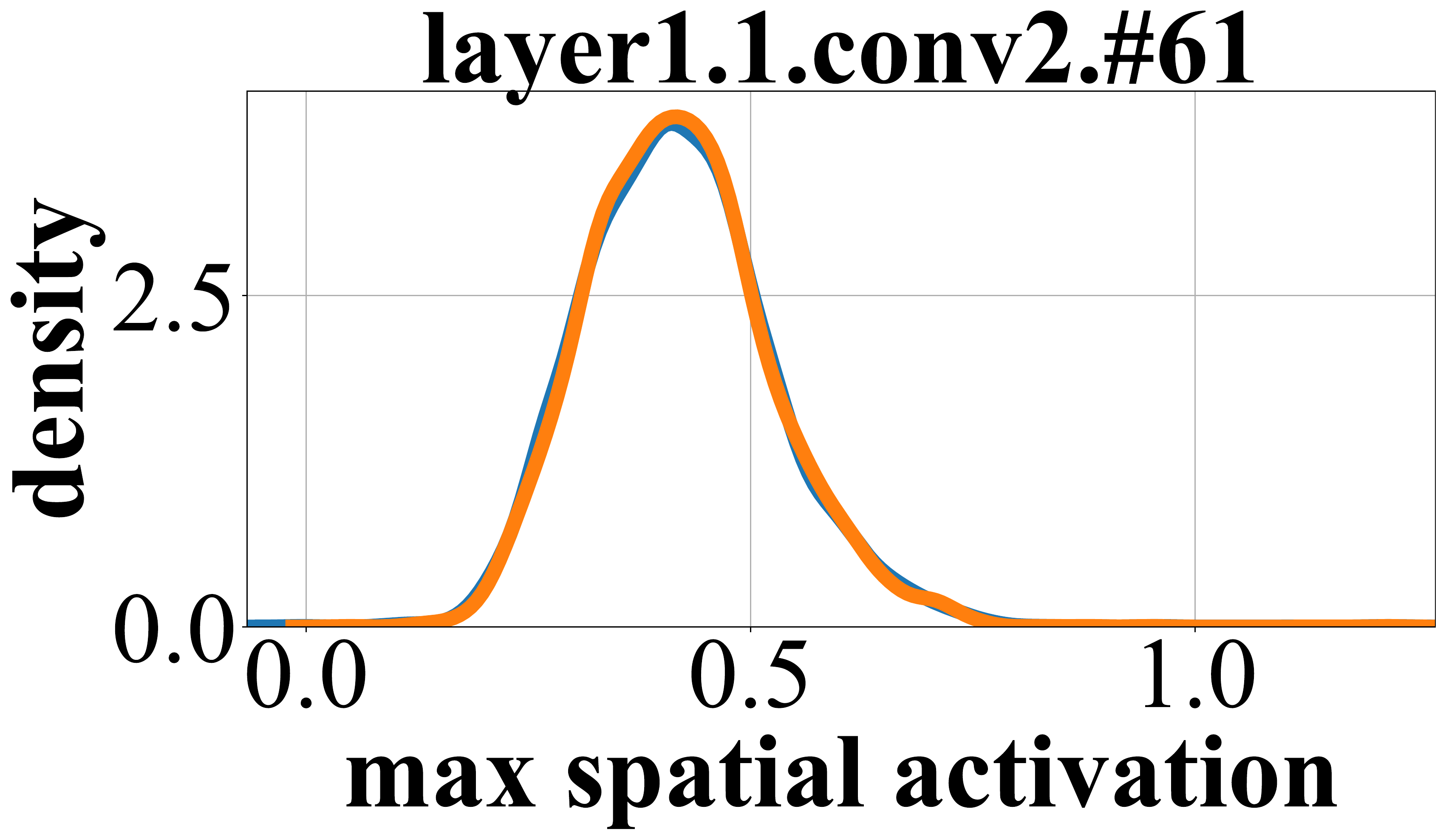} &
    \includegraphics[width=0.13\linewidth]{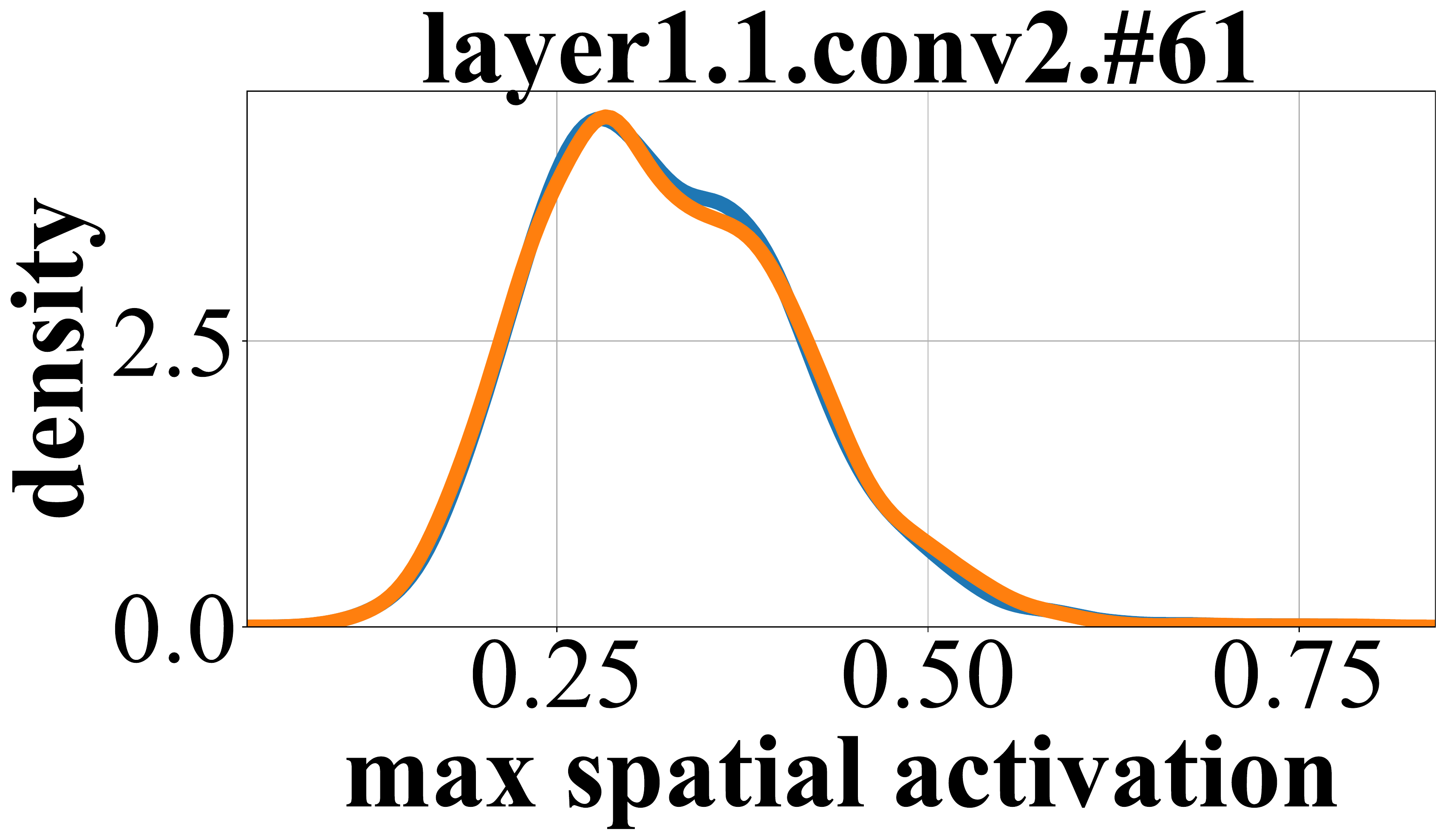} &
    \includegraphics[width=0.13\linewidth]{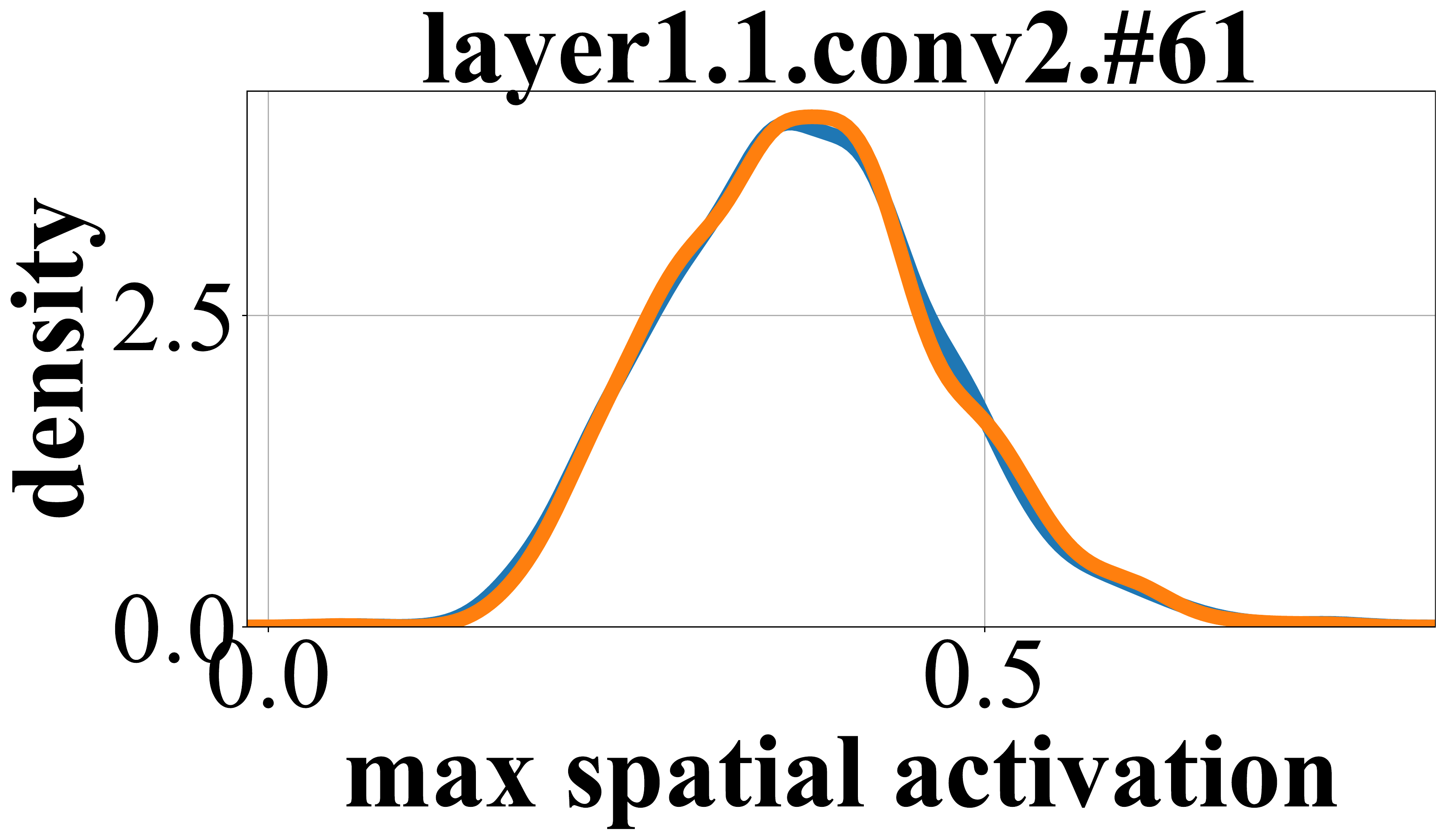} &
    \includegraphics[width=0.13\linewidth]{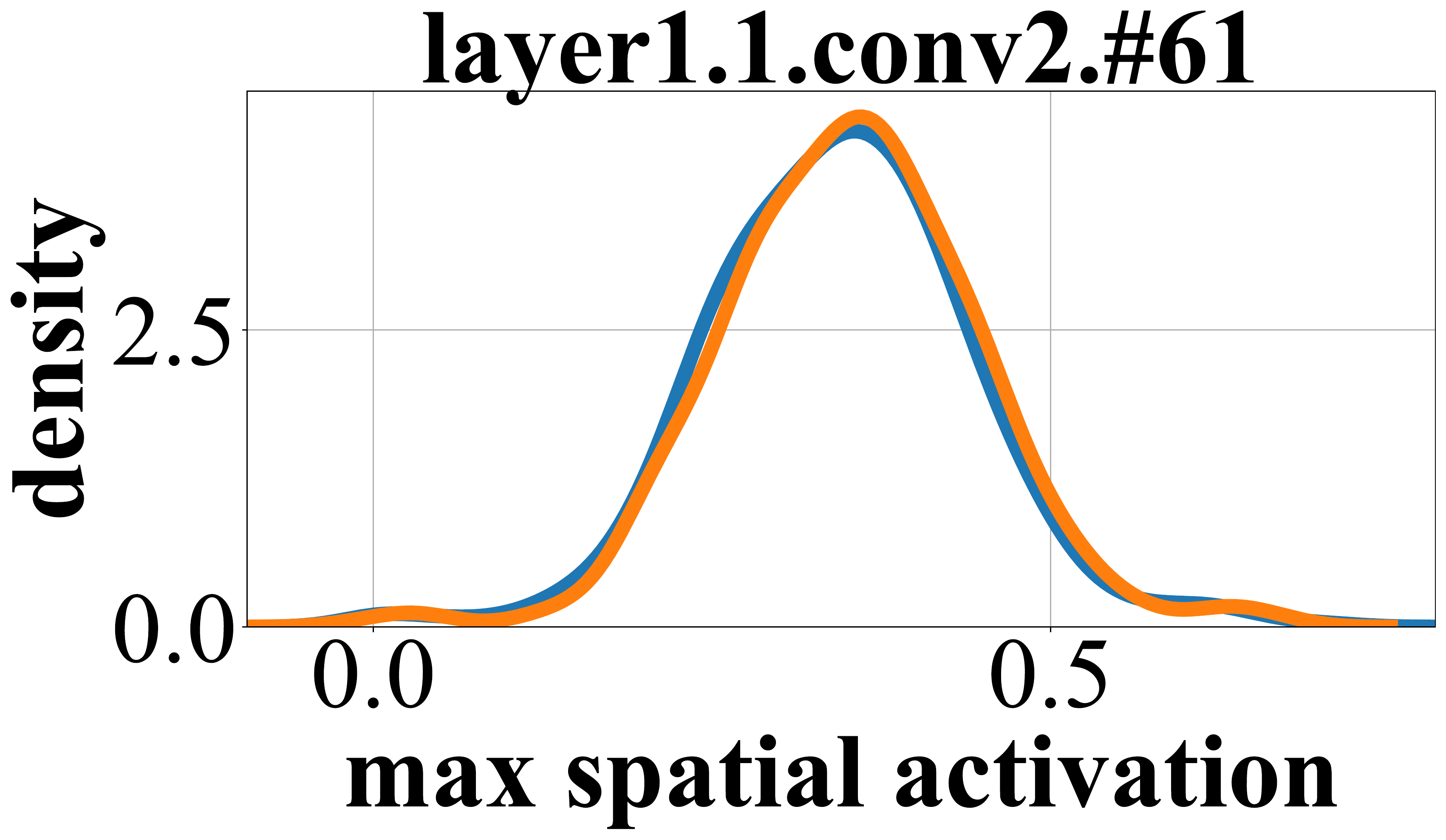} &
    \includegraphics[width=0.13\linewidth]{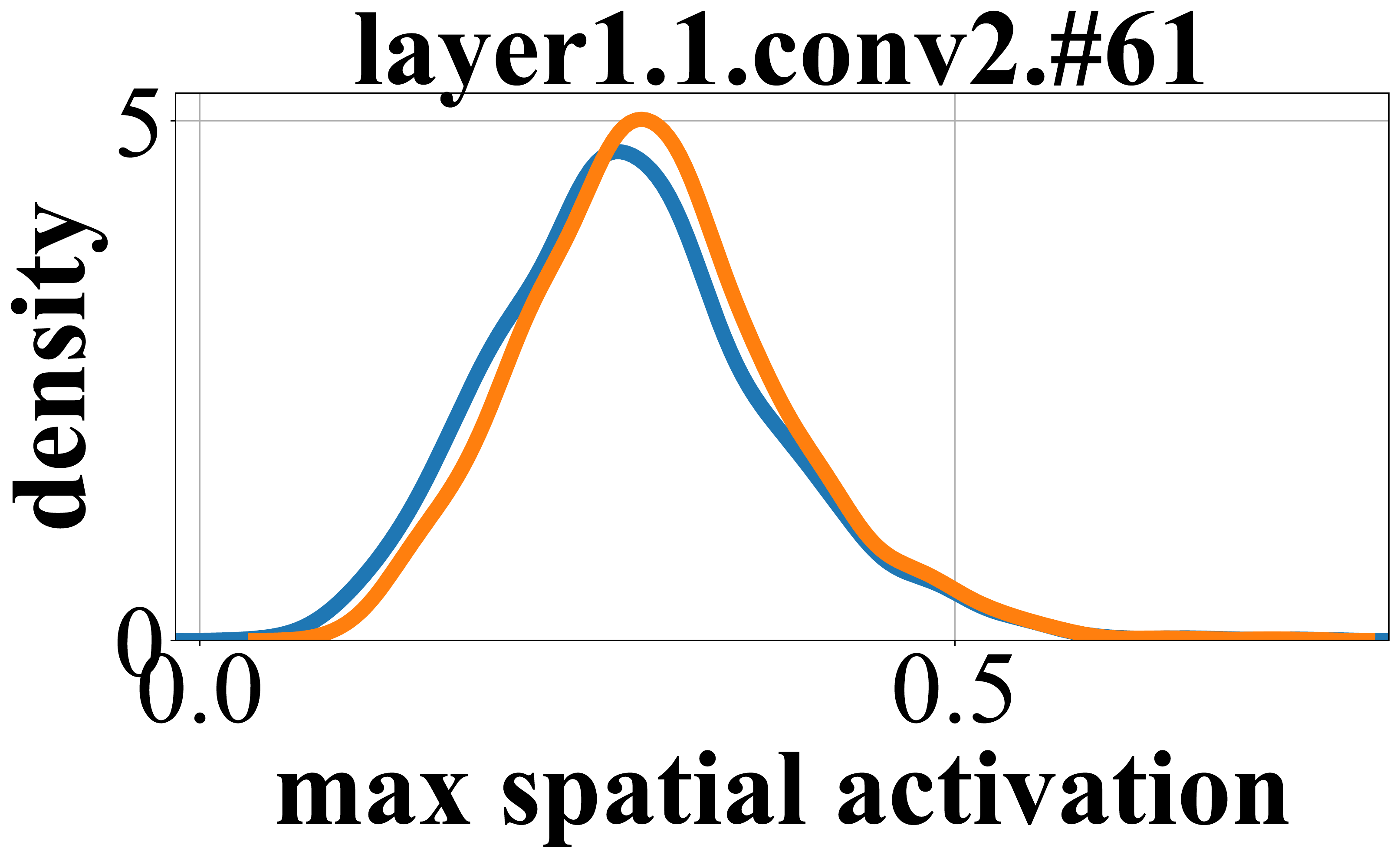} &
    \includegraphics[width=0.13\linewidth]{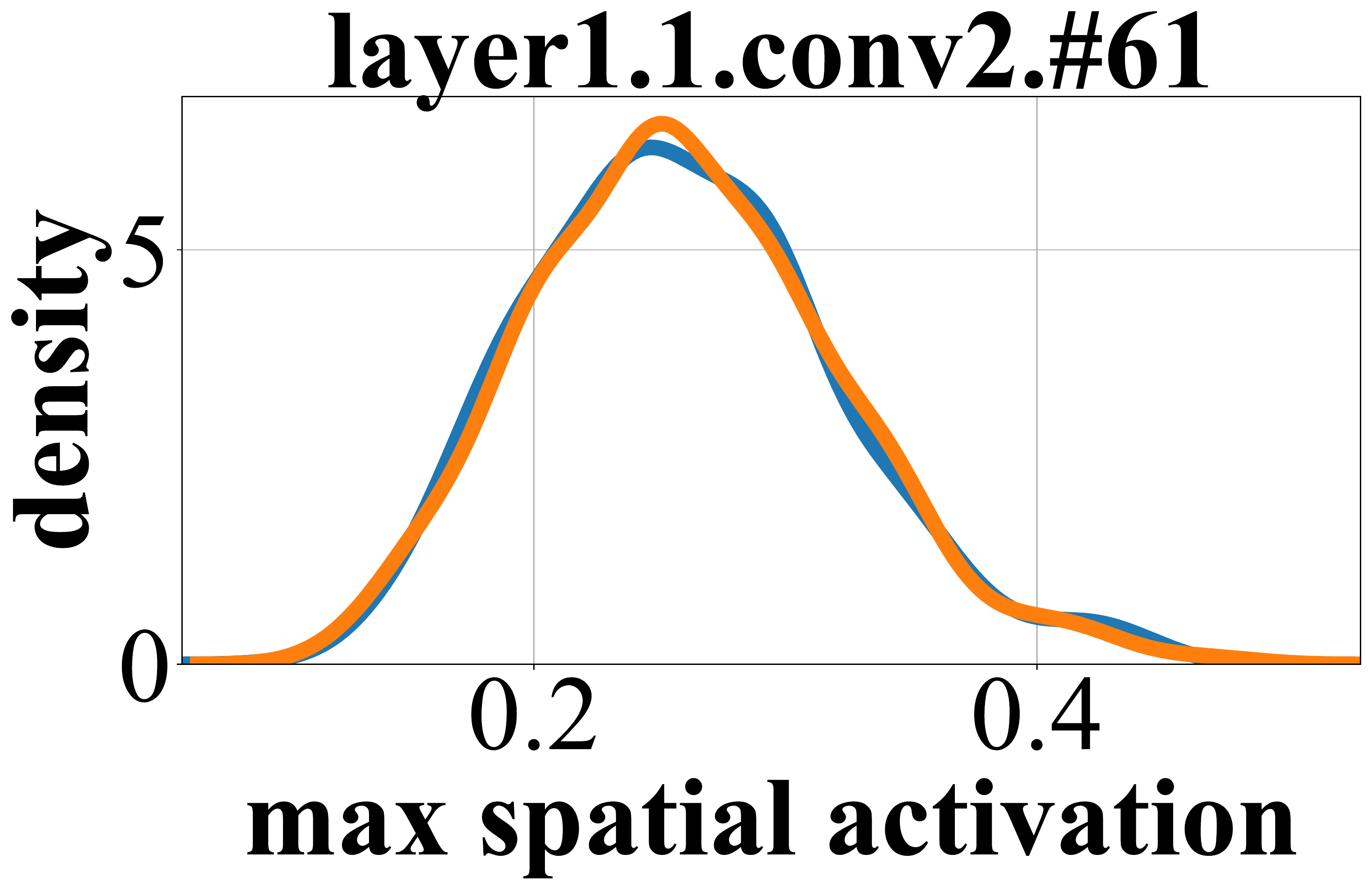} &
    \includegraphics[width=0.13\linewidth]{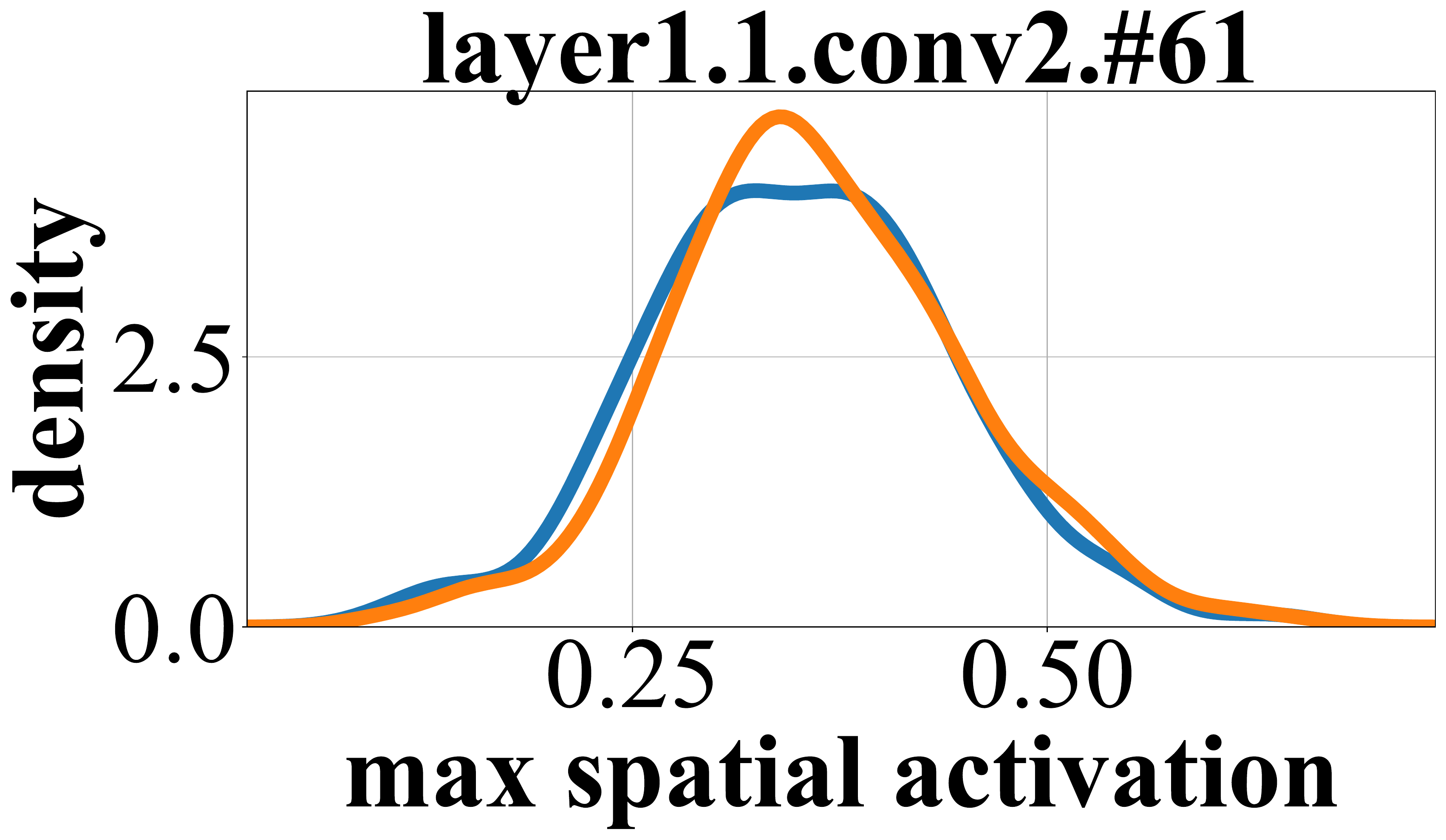} 
    \\
    
    \includegraphics[width=0.13\linewidth]{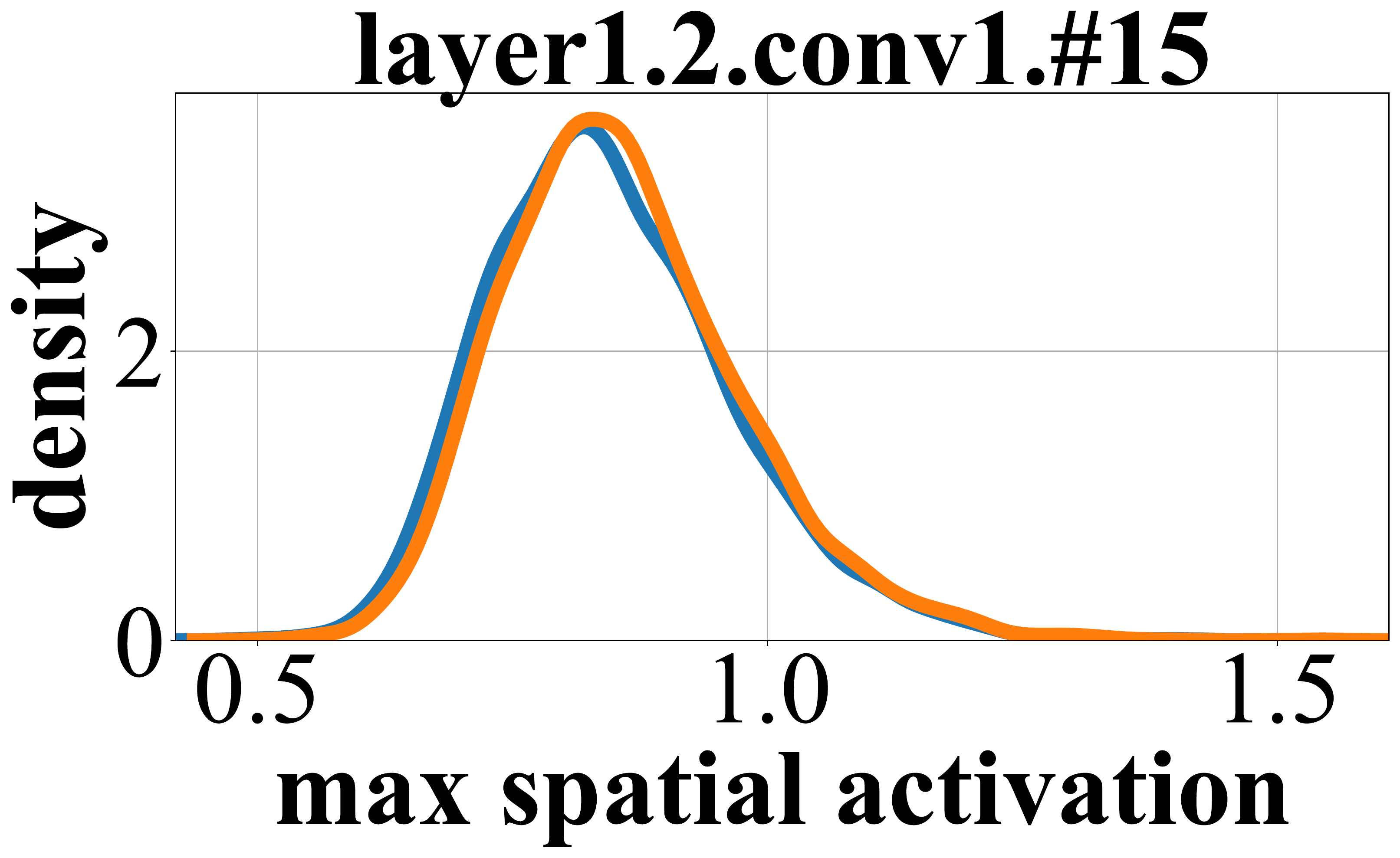} &
    \includegraphics[width=0.13\linewidth]{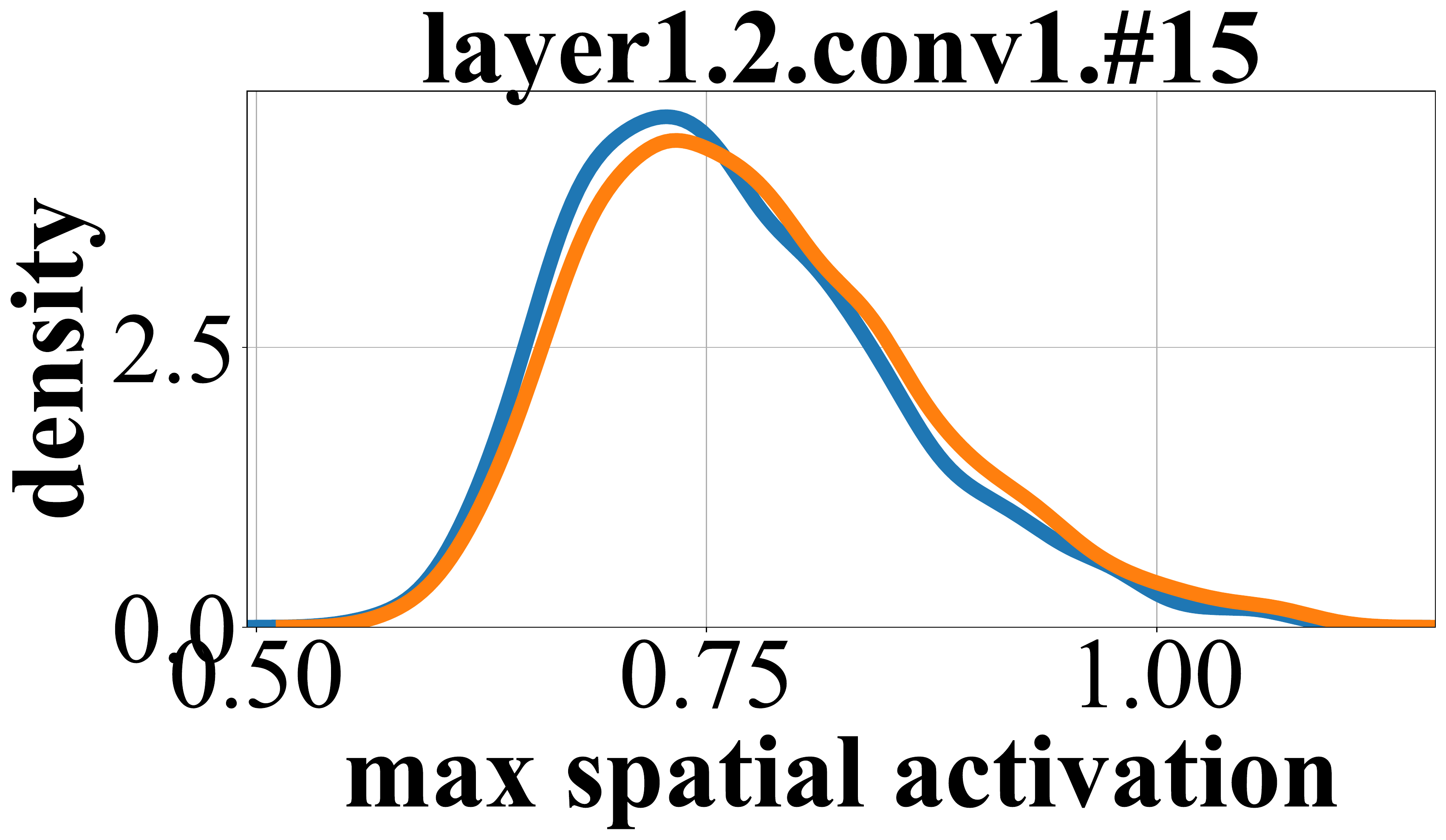} &
    \includegraphics[width=0.13\linewidth]{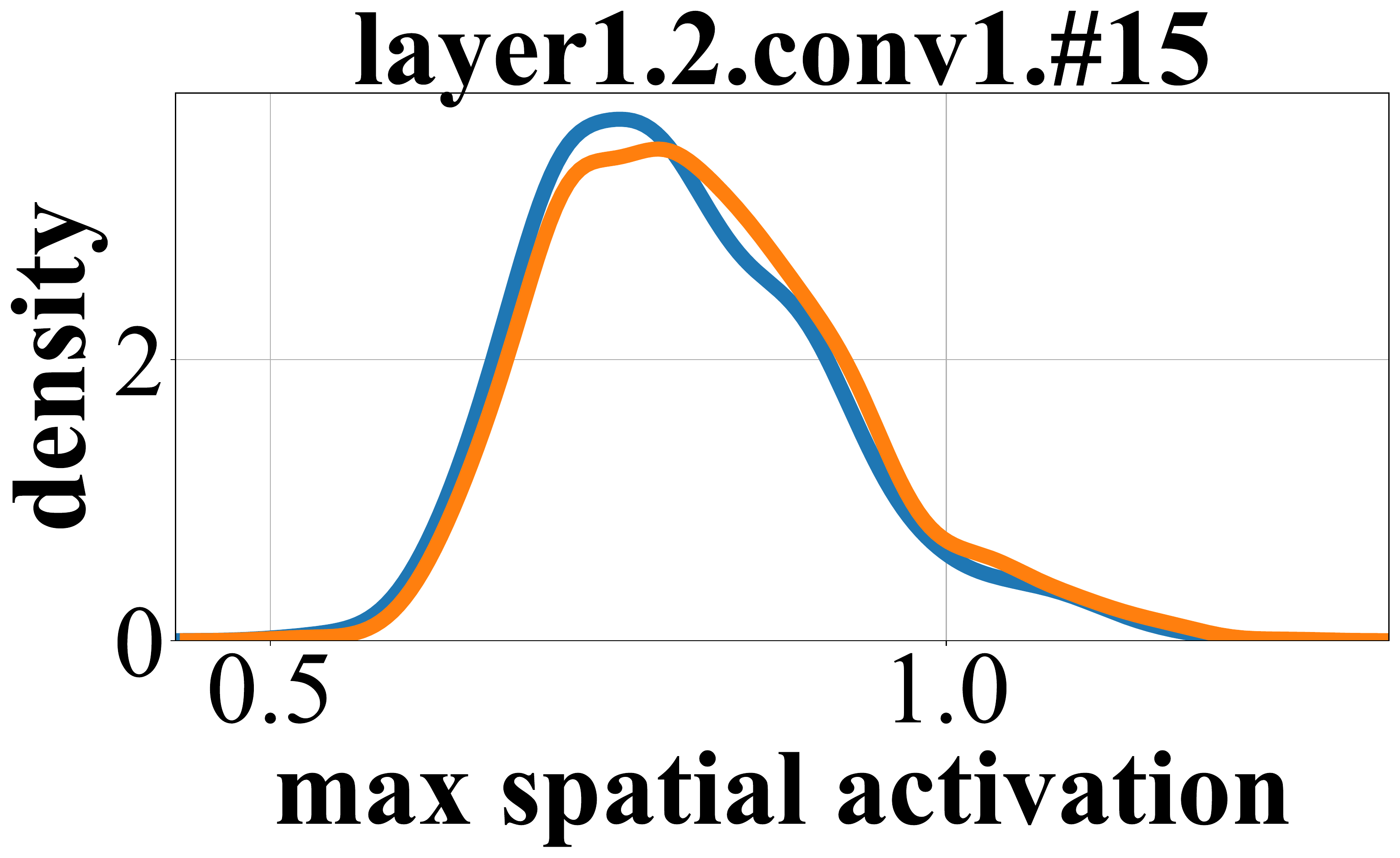} &
    \includegraphics[width=0.13\linewidth]{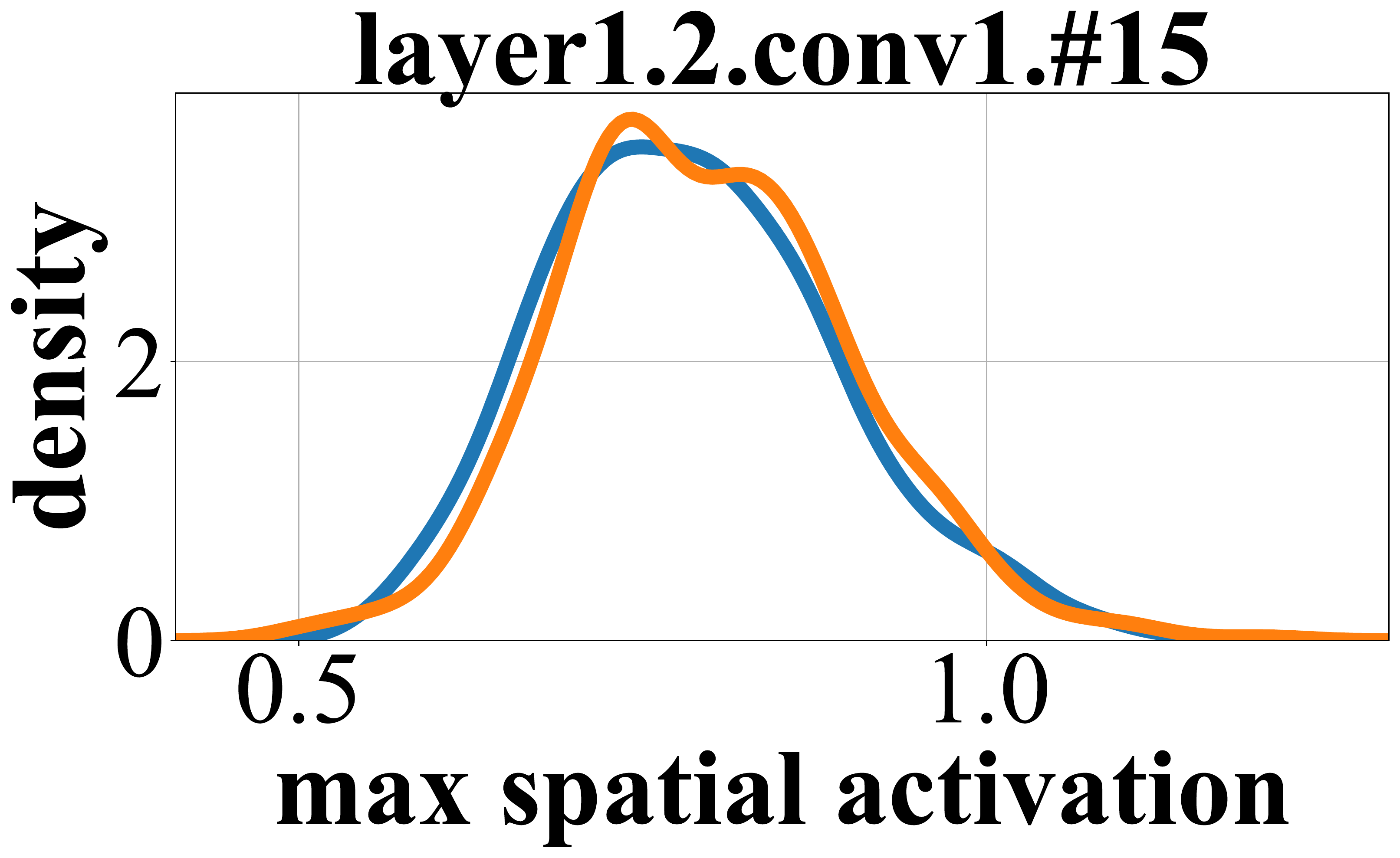} &
    \includegraphics[width=0.13\linewidth]{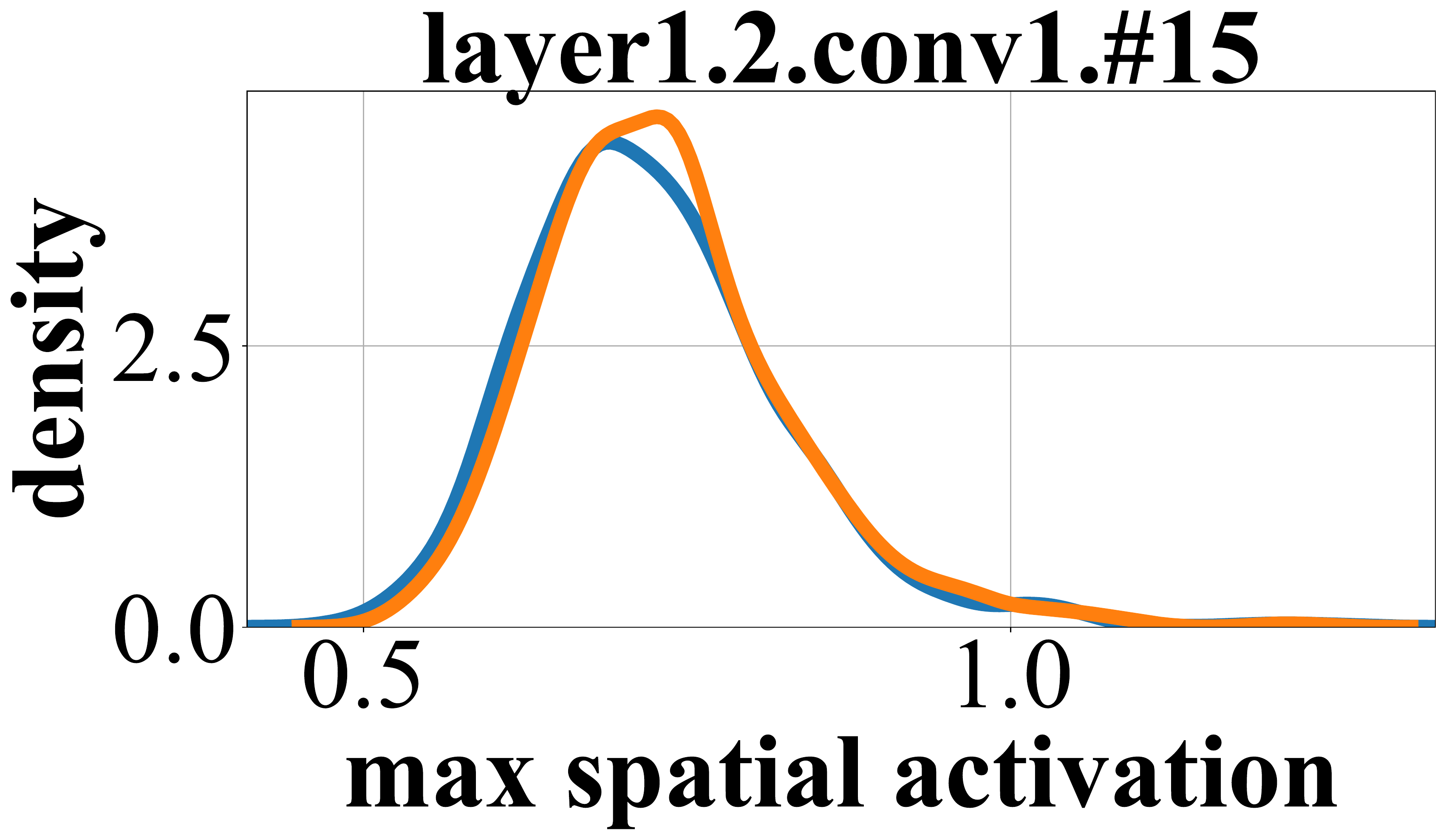} &
    \includegraphics[width=0.13\linewidth]{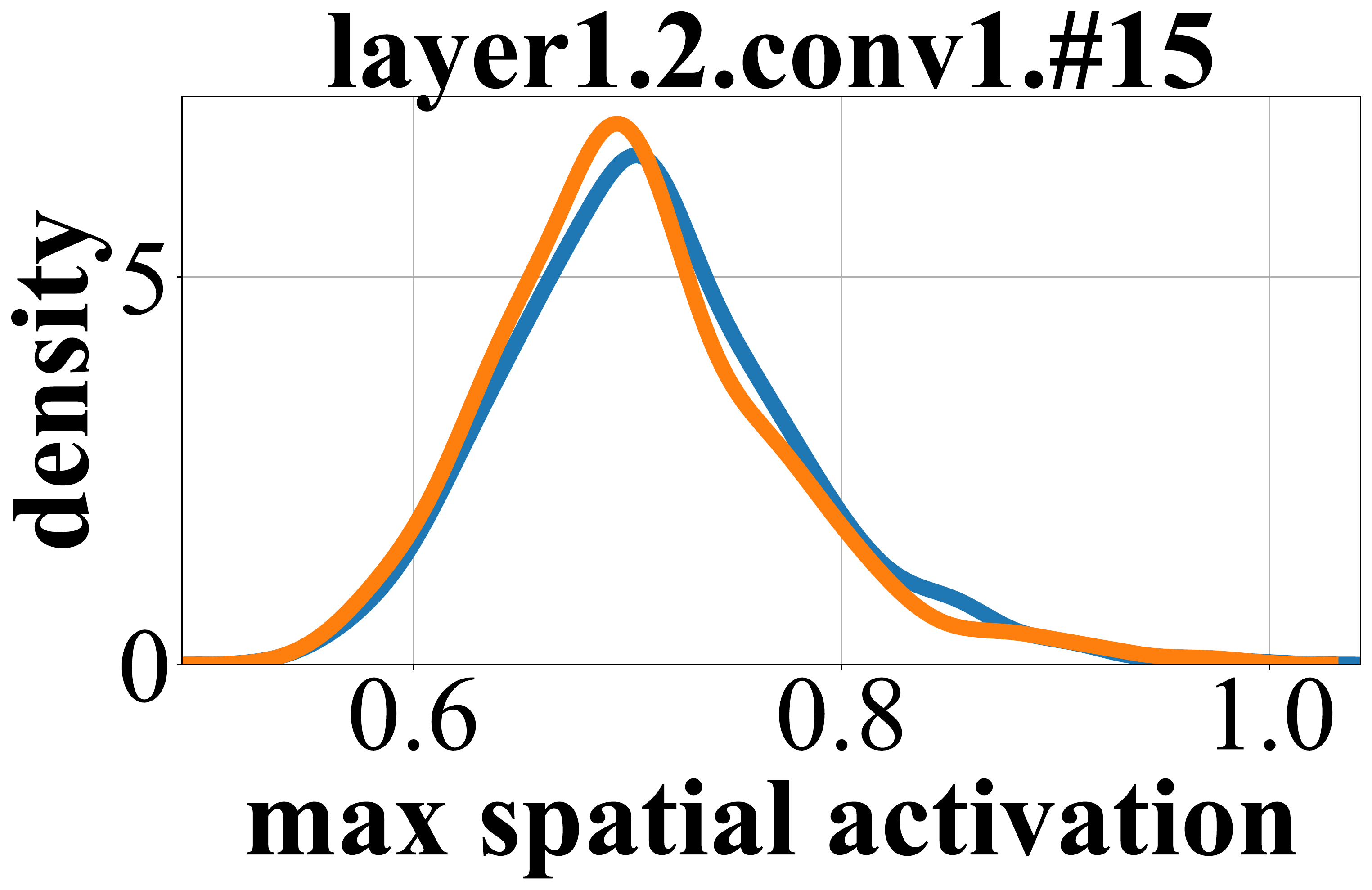} &
    \includegraphics[width=0.13\linewidth]{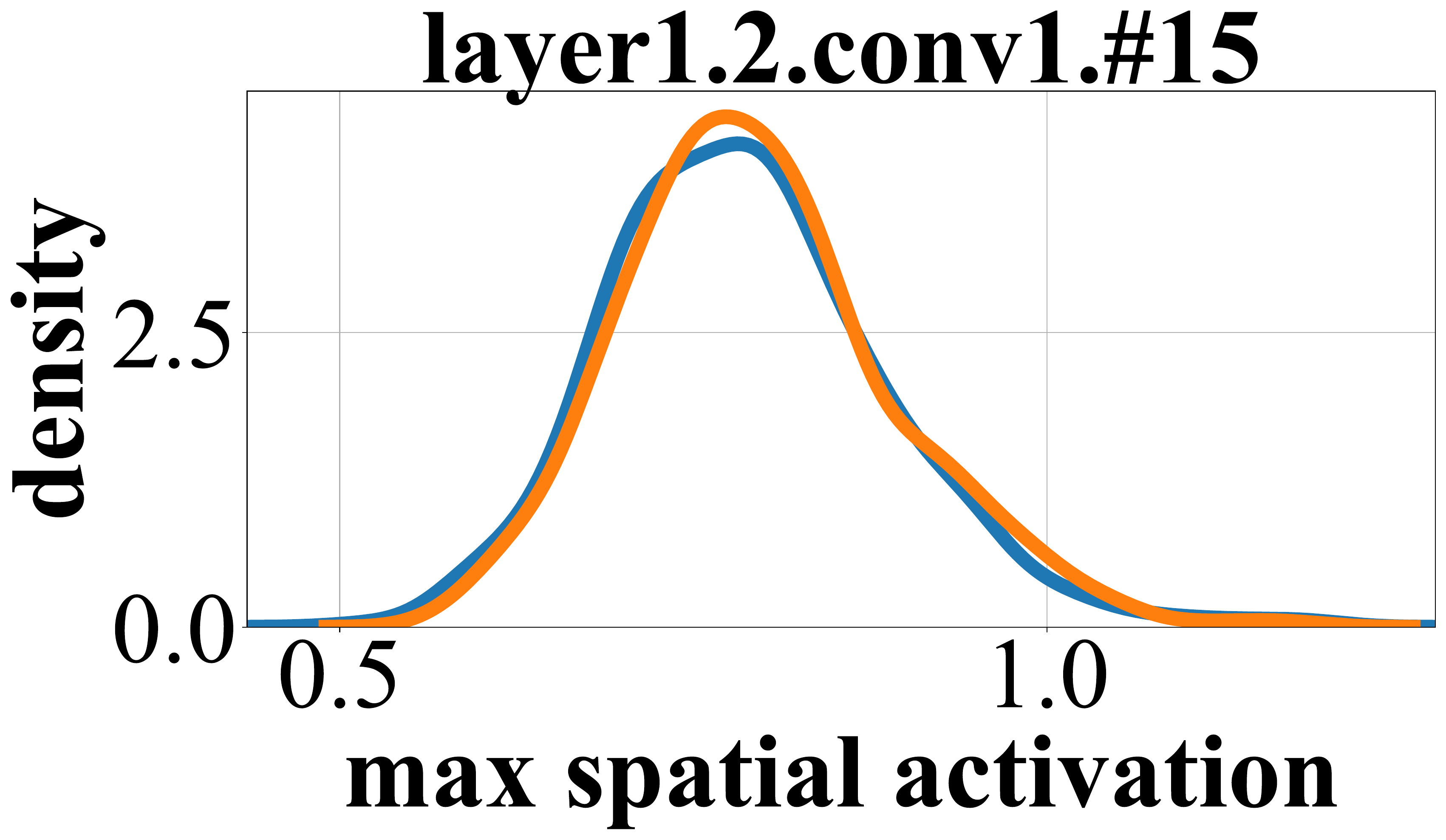} 
    \\

\end{tabular}
\includegraphics[width=0.50\linewidth]{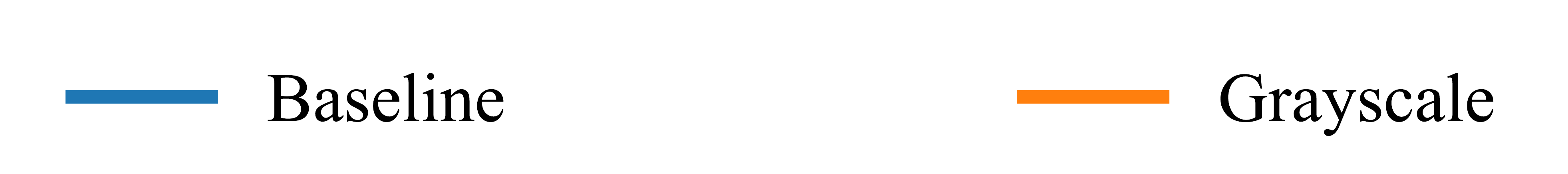}
\vspace{-0.5cm}
\caption{
\textit{Non Color-conditional T-FF in ResNet-50:}
Each row represents a \textit{non} color-conditional \textit{T-FF} (exact same T-FF as shown in Fig. \ref{fig_supp:lrp_patches_r50_non_color}), and 
we show the maximum spatial activation distributions
for ProGAN \cite{karras2018progressive}, StyleGAN2 \cite{Karras_2020_CVPR}, StyleGAN \cite{Karras_2019_CVPR}, BigGAN \cite{brock2018large}, CycleGAN \cite{zhu2017unpaired}, StarGAN \cite{choi2018stargan} and GauGAN \cite{park2019semantic} counterfeits
before (Baseline) and after color ablation (Grayscale).
We remark that for each counterfeit in the ForenSynths dataset \cite{Wang_2020_CVPR}, we apply global max pooling to the specific T-FF to obtain a {\em maximum spatial activation} value (scalar).
We can clearly observe that these \textit{T-FF} are producing identical / similar spatial activations (max) for the same set of counterfeits after removing color information which demonstrates that these \textit{T-FF} do not respond to color information.
This clearly indicates that these \textit{T-FF} are \textit{not} color-conditional
(Confirmed by Mood's median test).
}
\label{fig_supp:activation_hist_r50_non_color}
\end{figure}

\begin{figure}
\centering
\begin{tabular}{ccccccc}
    \multicolumn{1}{p{0.125\linewidth}}{\tiny \enskip ProGAN \cite{karras2018progressive}} &
    \multicolumn{1}{p{0.15\linewidth}}{\tiny  \enskip StyleGAN2 \cite{Karras_2020_CVPR}} &
    \multicolumn{1}{p{0.14\linewidth}}{\tiny StyleGAN \cite{Karras_2019_CVPR}} &
    \multicolumn{1}{p{0.125\linewidth}}{\tiny BigGAN \cite{brock2018large}} &
    \multicolumn{1}{p{0.132\linewidth}}{\tiny CycleGAN \cite{zhu2017unpaired}} &
    \multicolumn{1}{p{0.135\linewidth}}{\tiny \enskip StarGAN \cite{choi2018stargan}} &
    {\tiny GauGAN \cite{park2019semantic}} \\
    
    \multicolumn{7}{c}{\includegraphics[width=0.99\linewidth]{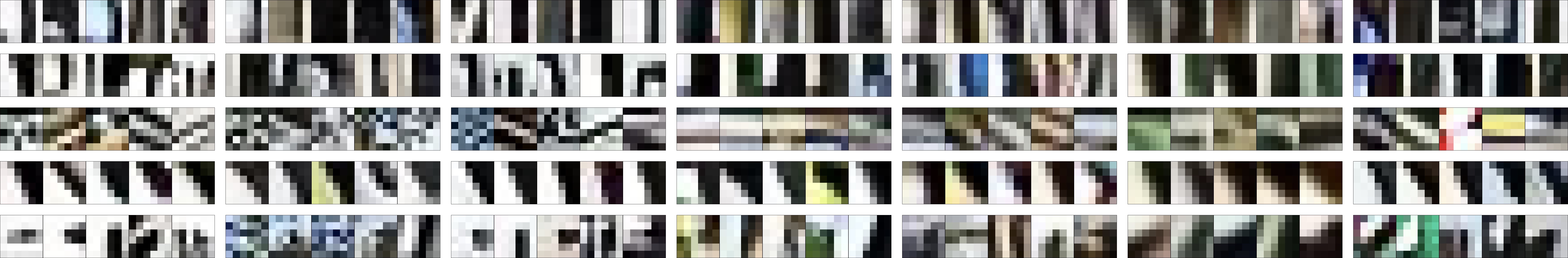}}
    
\end{tabular}
\caption{
\textit{T-FF} that are \textit{not} color-conditional in EfficientNet-B0 Universal detector:
We show the LRP-max response regions for 5 \textit{non} color-conditional T-FF for ProGAN 
\cite{karras2018progressive}
and all 6 unseen GANs 
\cite{Karras_2020_CVPR,Karras_2019_CVPR,brock2018large,zhu2017unpaired,choi2018stargan,park2019semantic}.
Each row represents a \textit{non} color-conditional T-FF.
We emphasize that \textit{T-FF} are discovered using our proposed \textit{forensic feature relevance statistic (FF-RS)}.
This detector is trained with ProGAN
\cite{karras2018progressive} 
counterfeits \cite{Wang_2020_CVPR} and cross-model forensic transfer is evaluated on other unseen GANs.
All counterfeits are obtained from the ForenSynths dataset 
\cite{Wang_2020_CVPR}.
Visual inspection of LRP-max regions of \textit{non} color-conditional \textit{T-FF} shows frequency / texture artifacts.
i.e.: rapid changes in pixel intensities.
This shows that the universal detector also uses frequency / texture artifacts for cross-model transfer although \textit{color is a critical T-FF} as $\approx 52\%$ of \textit{T-FF} are color-conditional.
We emphasize that our proposed method is capable of identifying different types of \textit{T-FF} (i.e.: frequency / texture artifacts).
}
\label{fig_supp:lrp_patches_efb0_non_color}
\end{figure}

\begin{figure}
\centering
\begin{tabular}{ccccccc}
    {\tiny ProGAN \cite{karras2018progressive}} &
    {\tiny StyleGAN2 \cite{Karras_2020_CVPR}} &
    {\tiny StyleGAN \cite{Karras_2019_CVPR}} &
    {\tiny BigGAN \cite{brock2018large}} &
    {\tiny CycleGAN \cite{zhu2017unpaired}} &
    {\tiny StarGAN \cite{choi2018stargan}} &
    {\tiny GauGAN \cite{park2019semantic}} \\

  \includegraphics[width=0.13\linewidth]{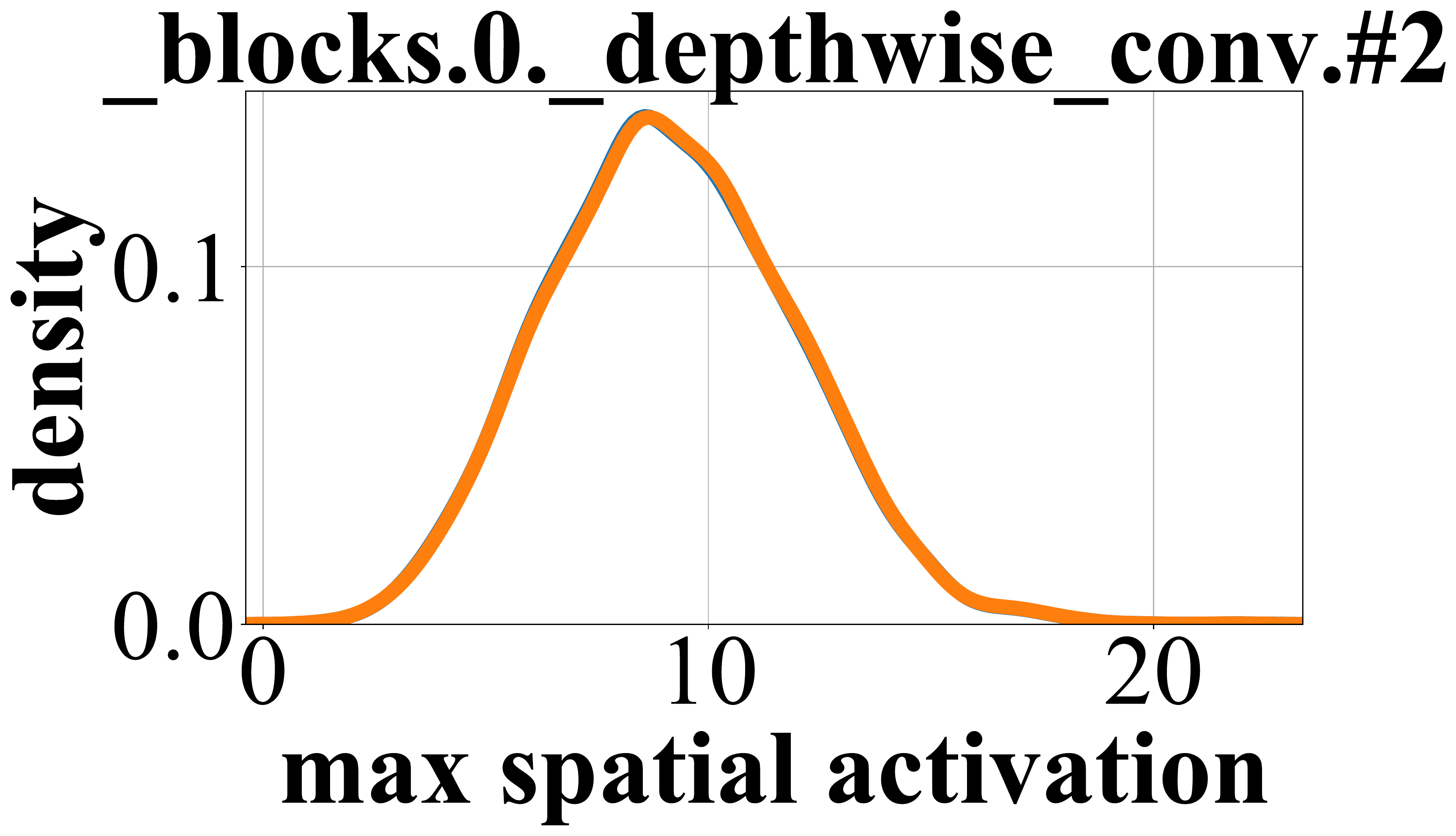} &
    \includegraphics[width=0.13\linewidth]{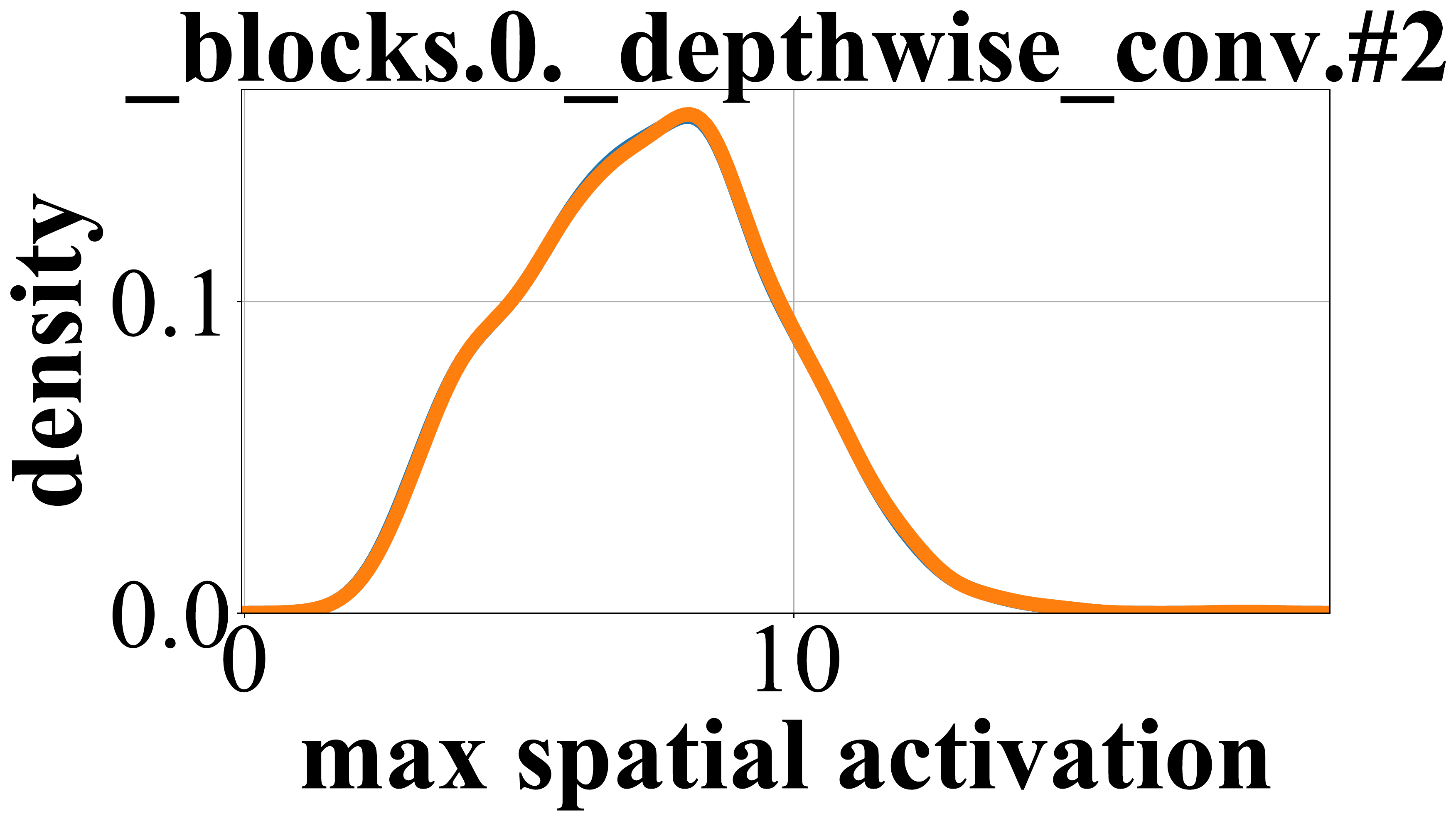} &
    \includegraphics[width=0.13\linewidth]{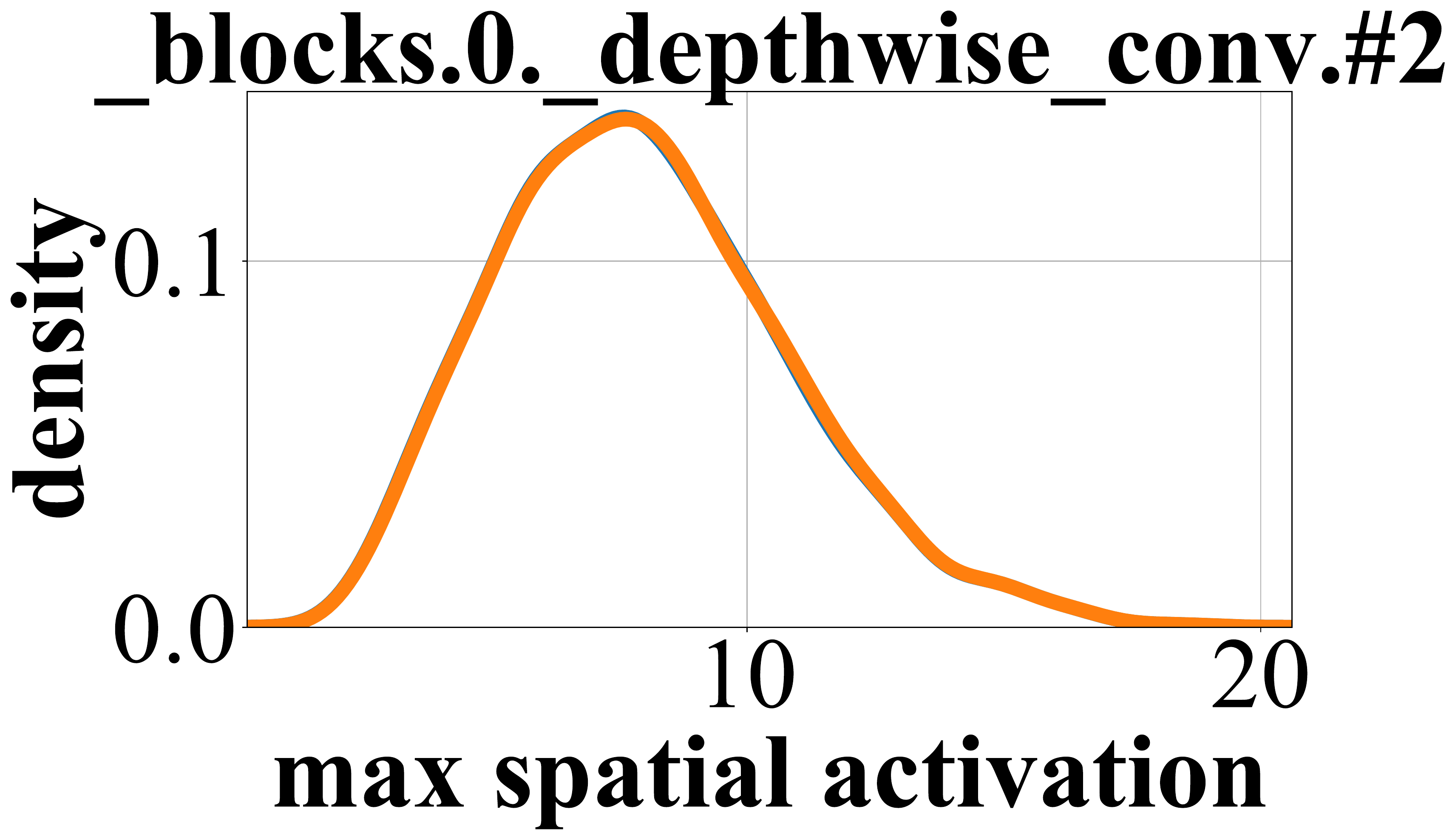} &
    \includegraphics[width=0.13\linewidth]{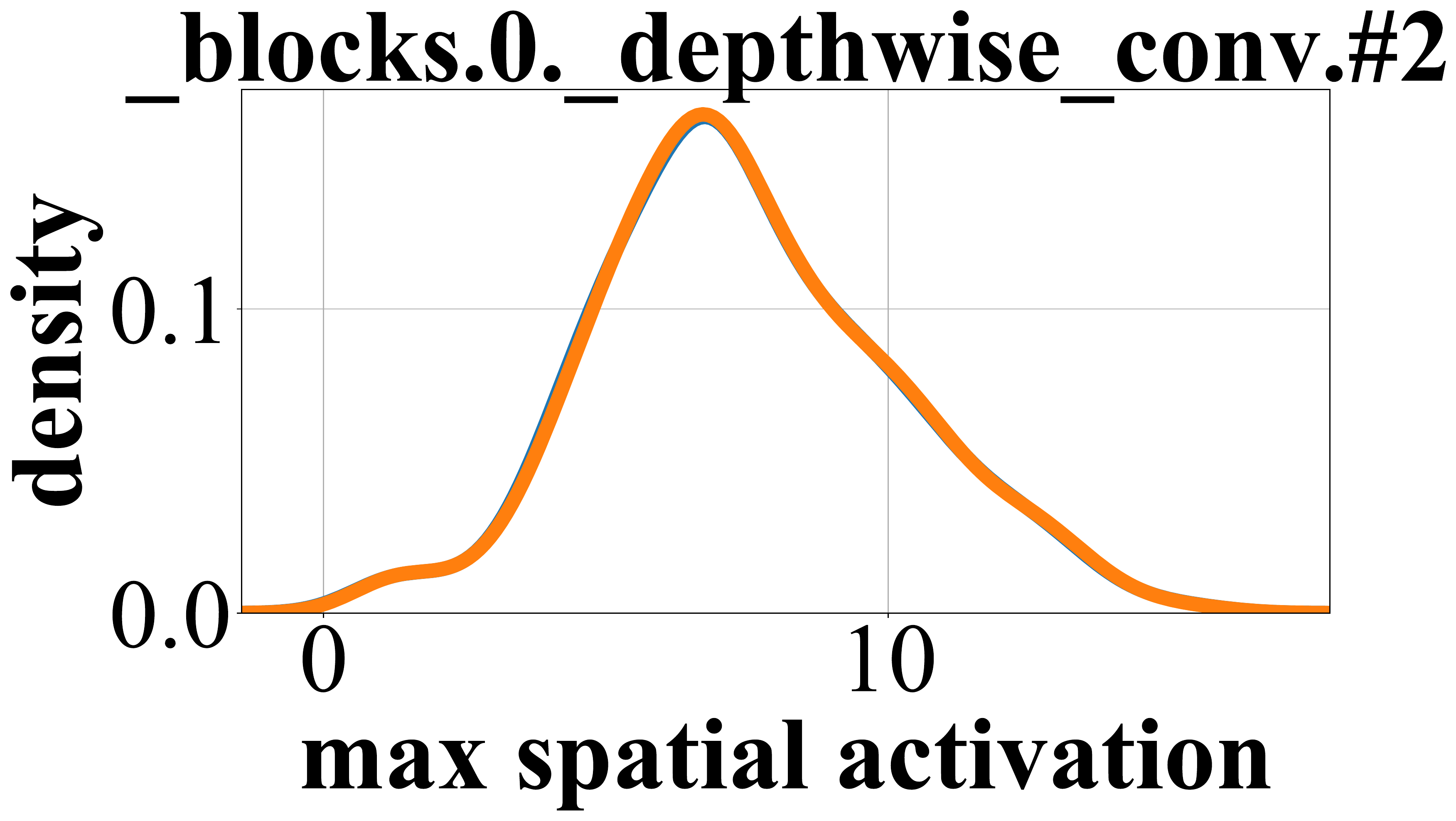} &
    \includegraphics[width=0.13\linewidth]{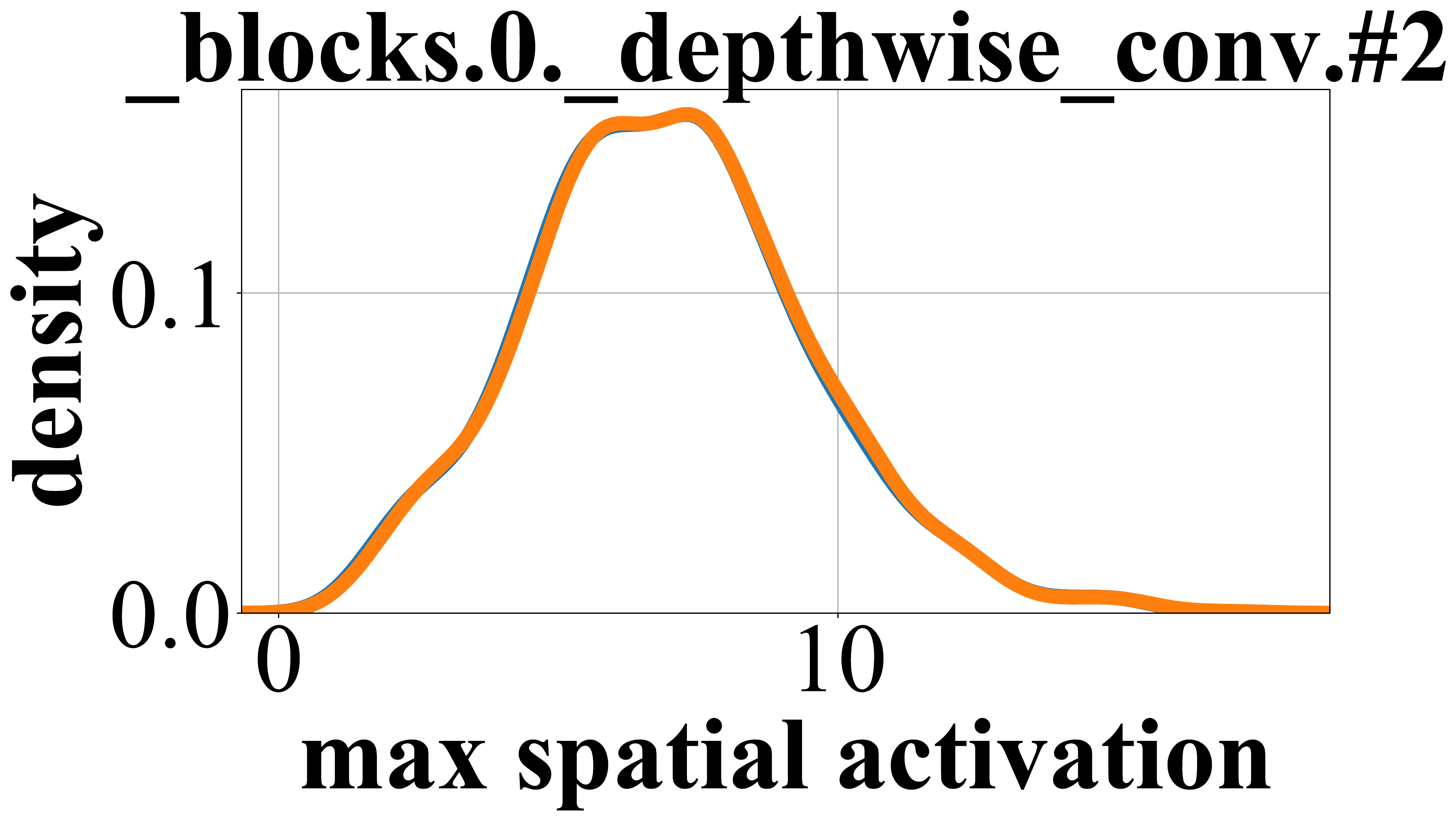} &
     \includegraphics[width=0.13\linewidth]{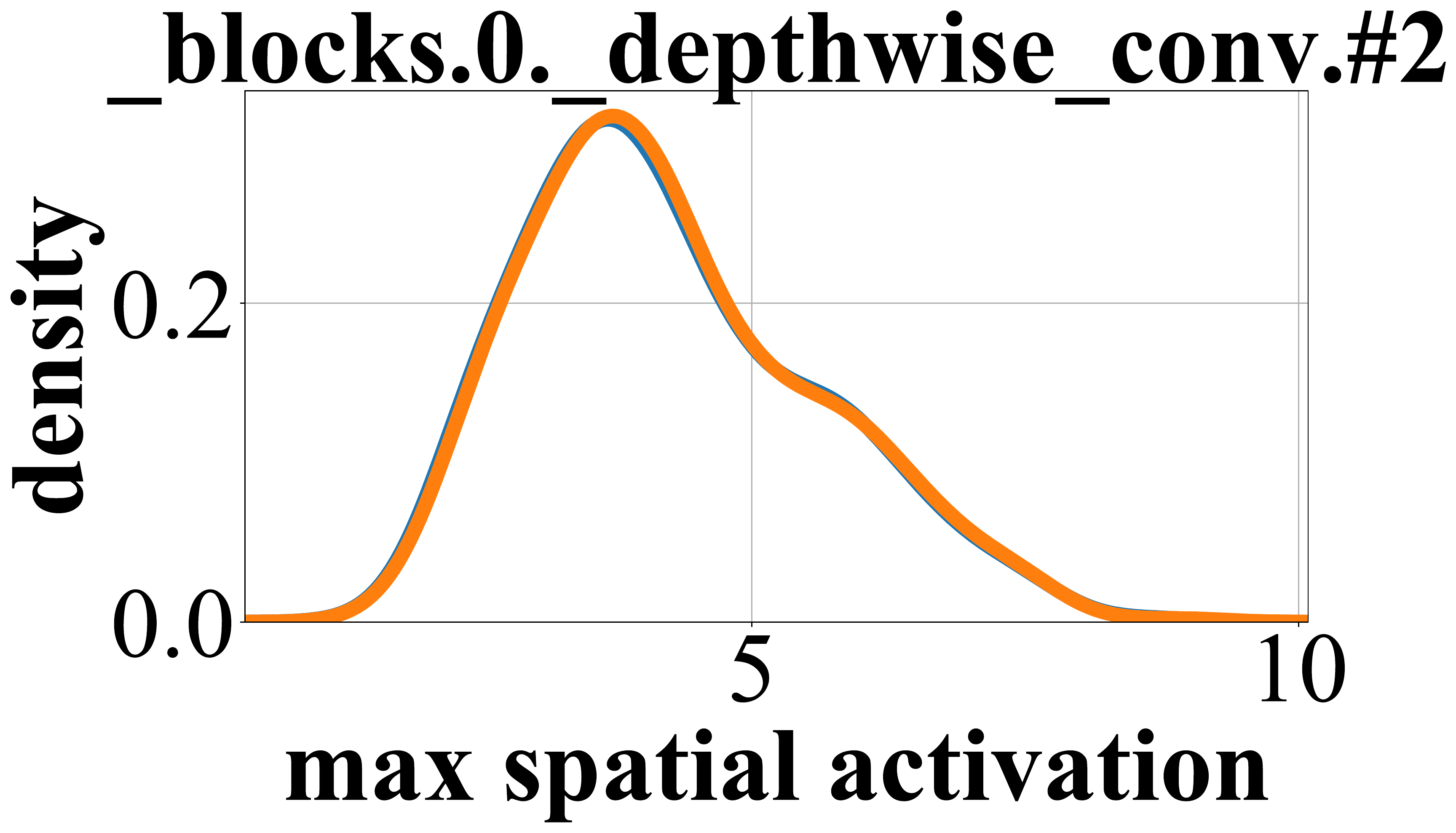} &
    \includegraphics[width=0.13\linewidth]{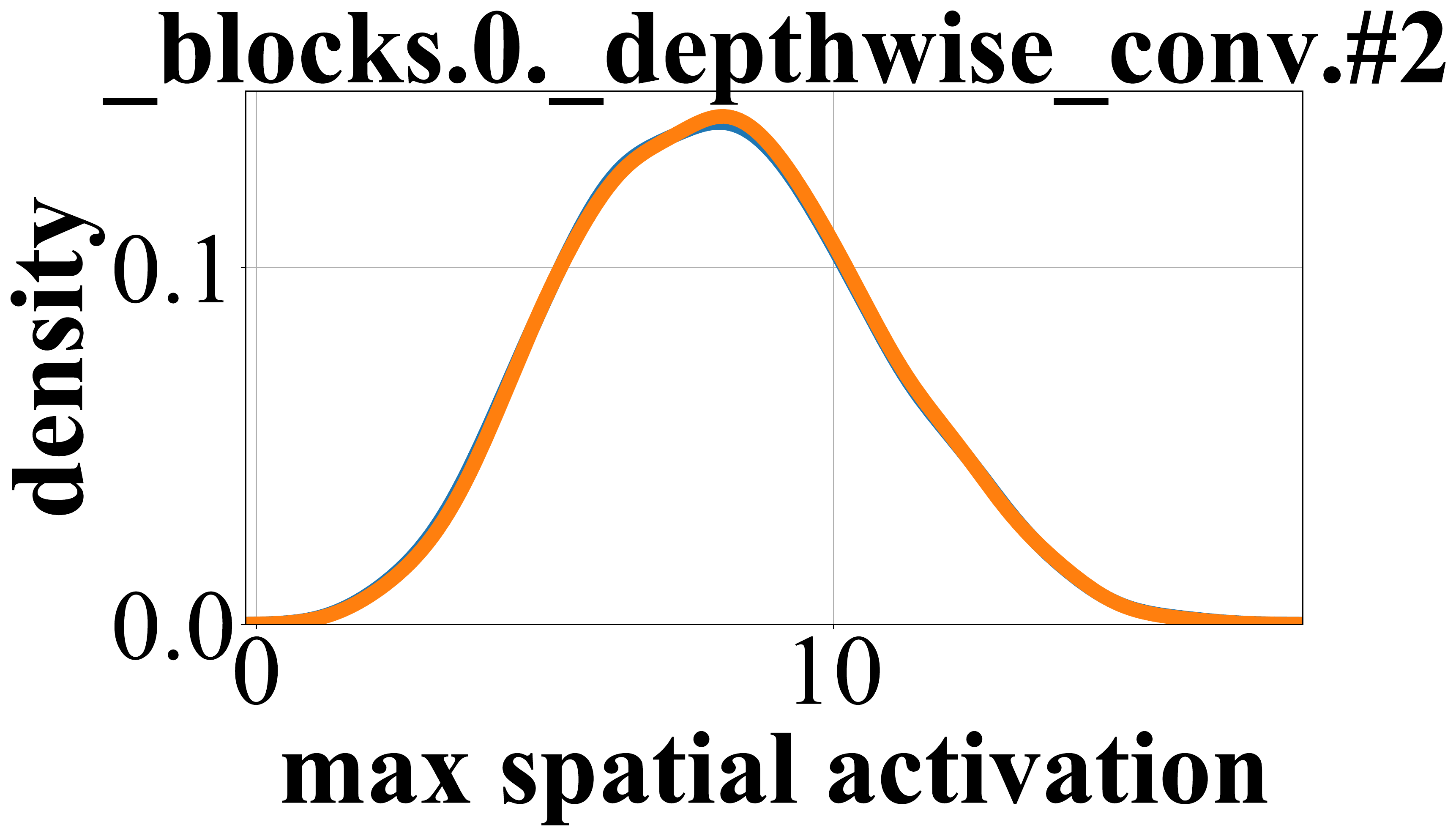}
    \\
    
     \includegraphics[width=0.13\linewidth]{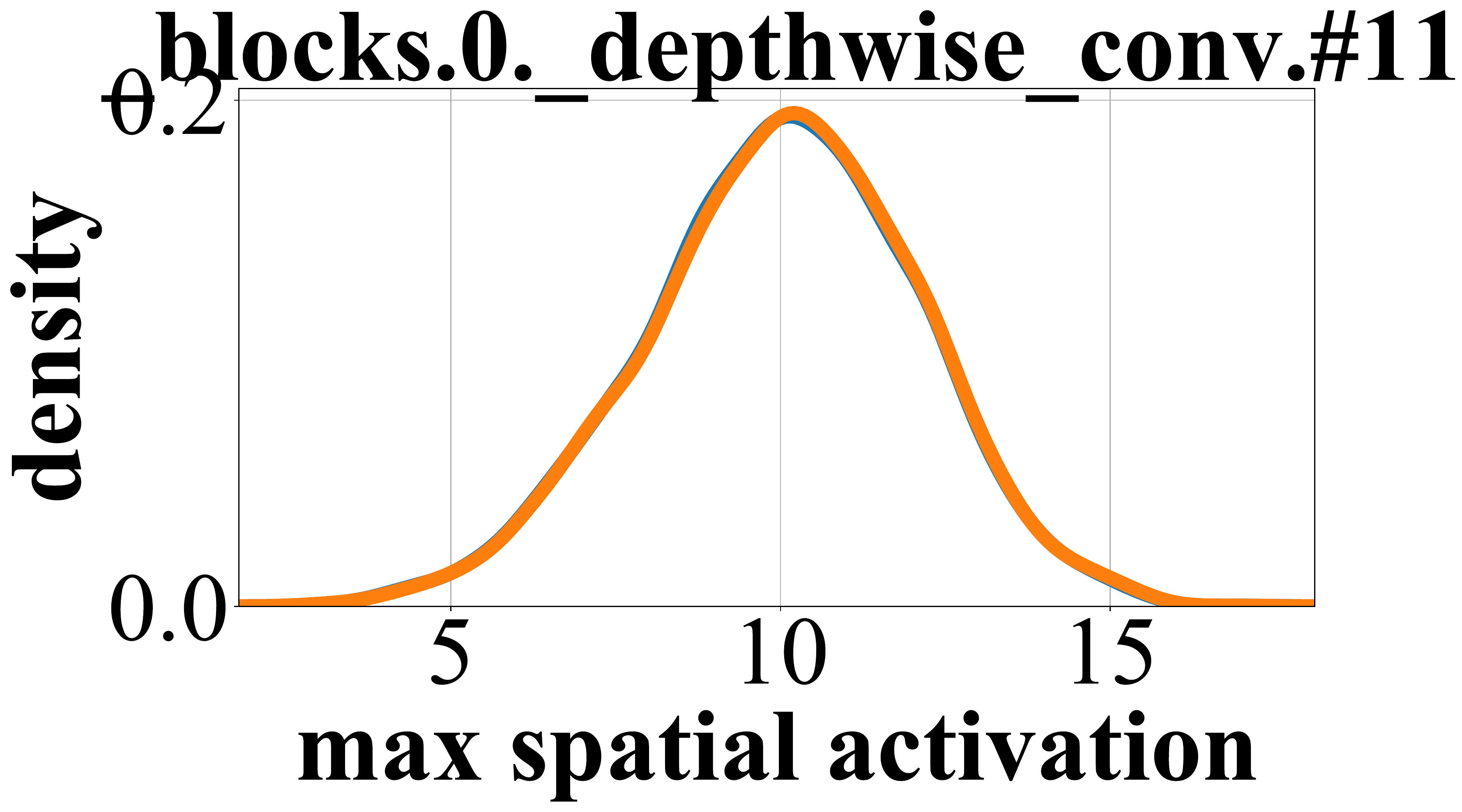} &
    \includegraphics[width=0.13\linewidth]{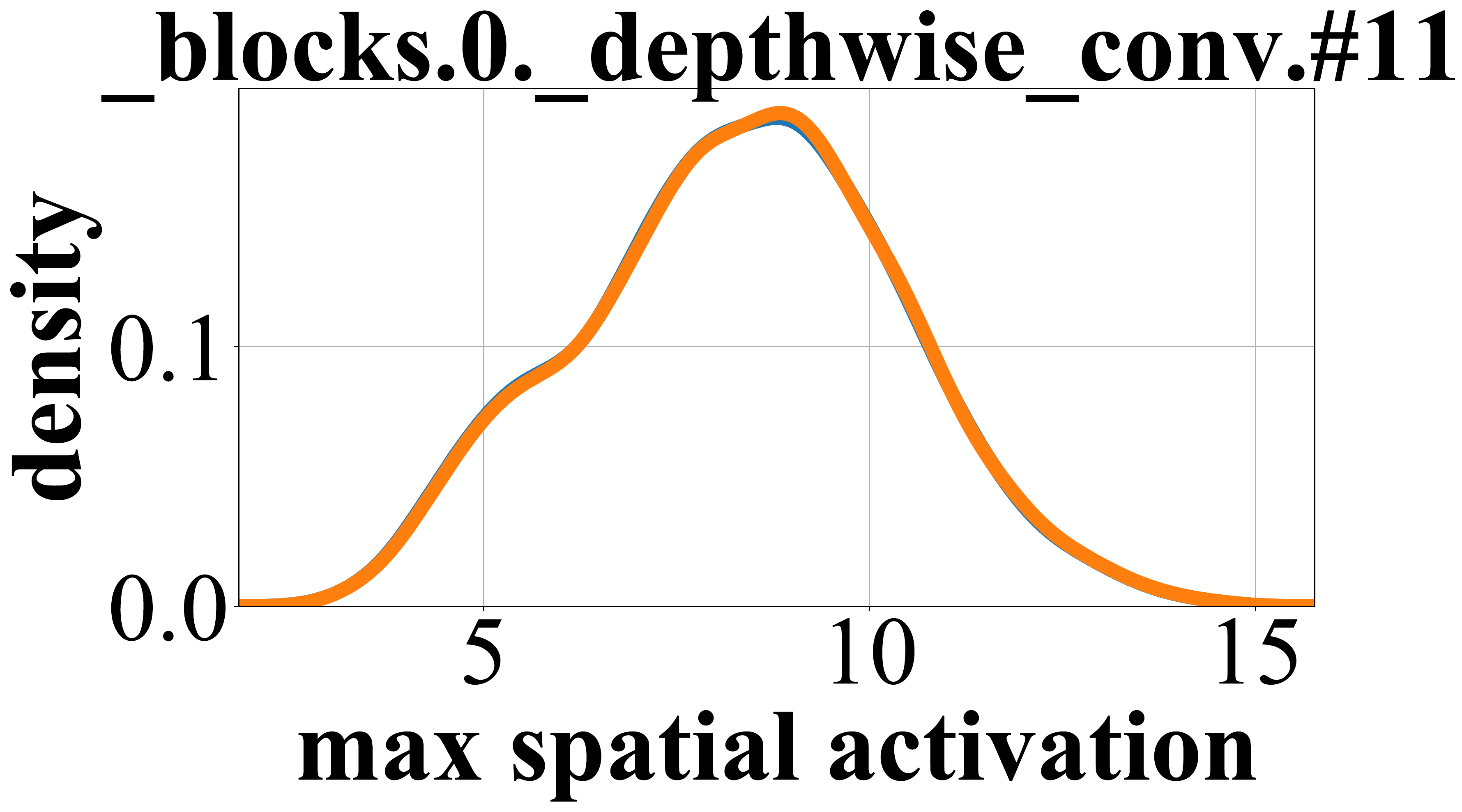} &
    \includegraphics[width=0.13\linewidth]{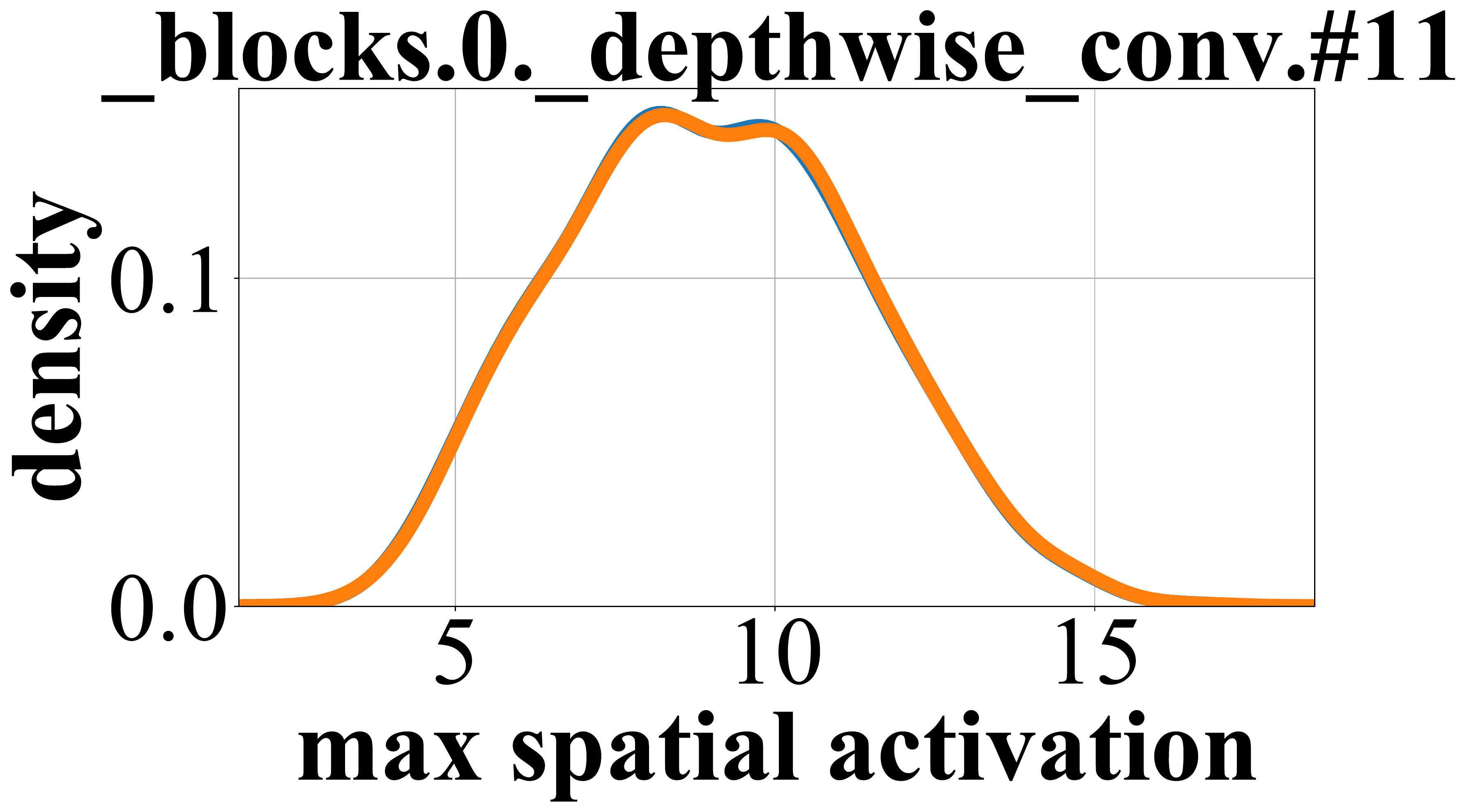} &
    \includegraphics[width=0.13\linewidth]{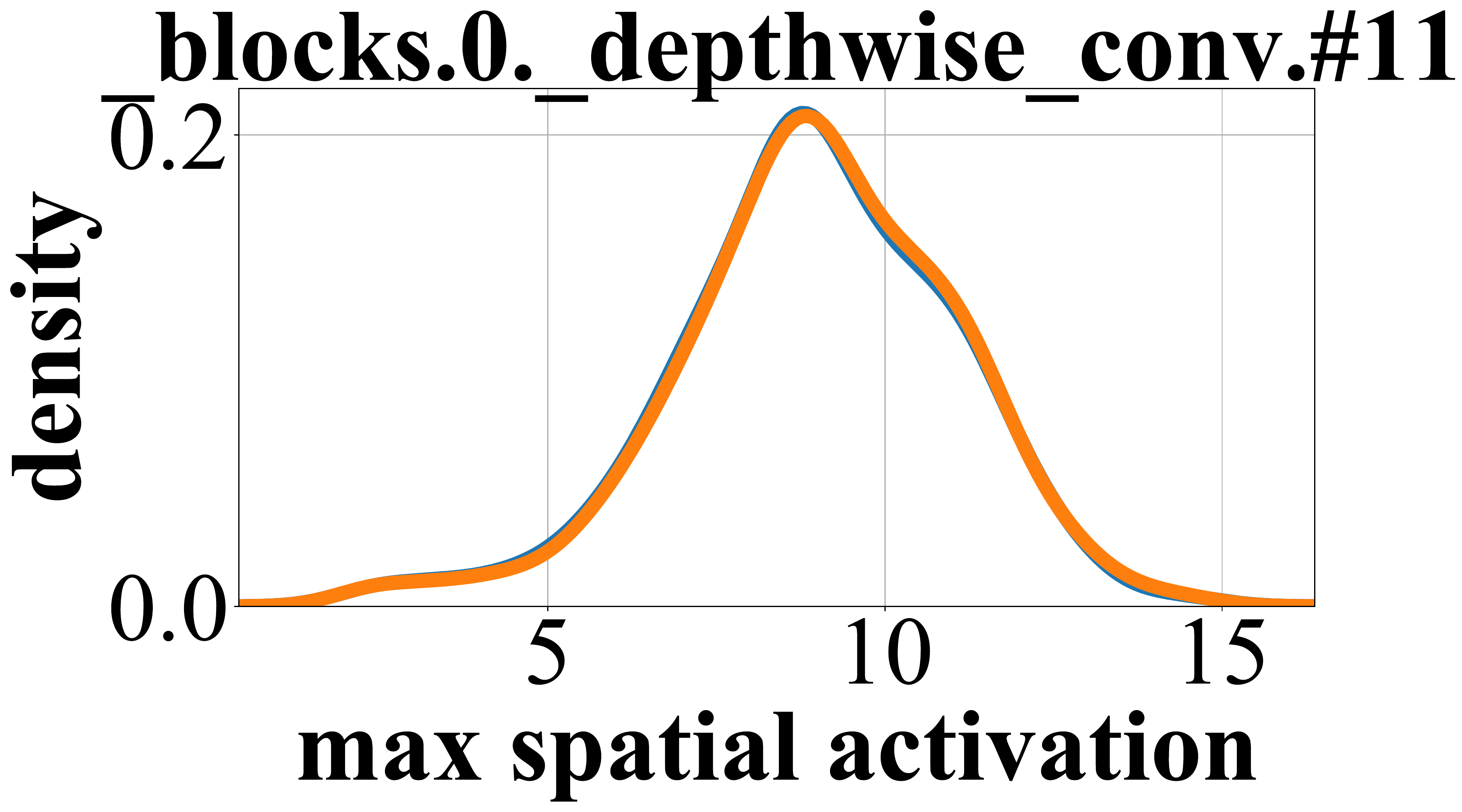} &
    \includegraphics[width=0.13\linewidth]{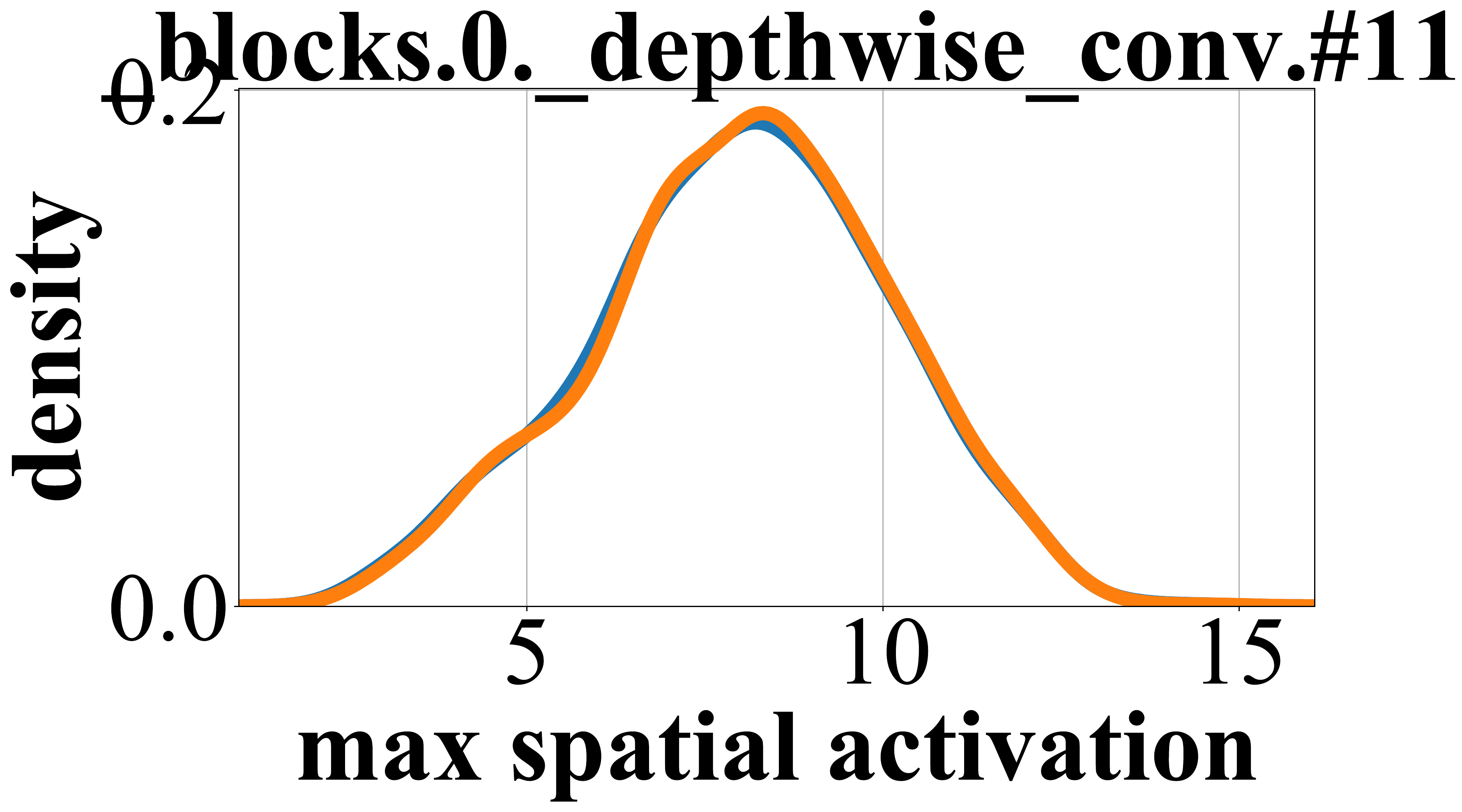} &
     \includegraphics[width=0.13\linewidth]{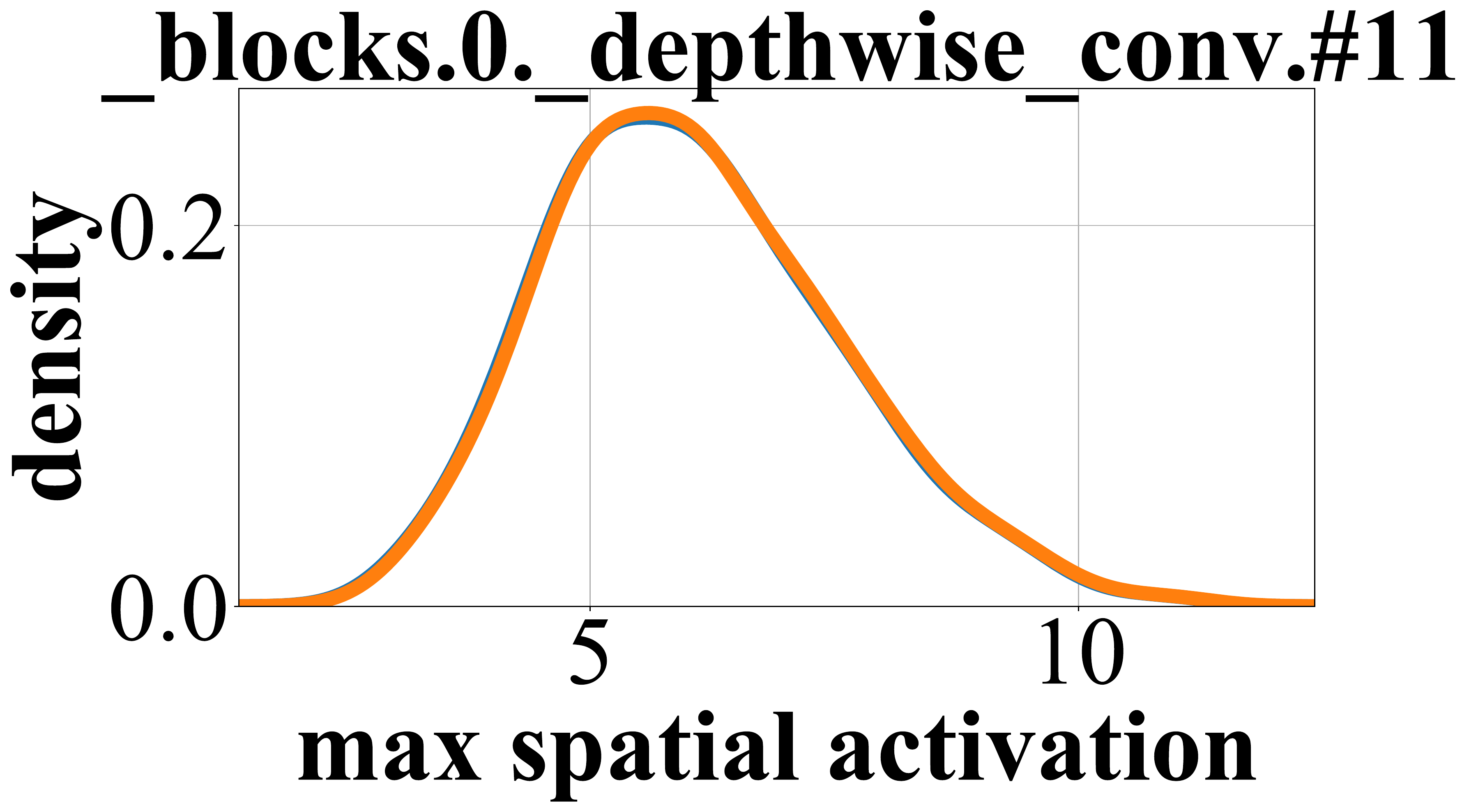} &
    \includegraphics[width=0.13\linewidth]{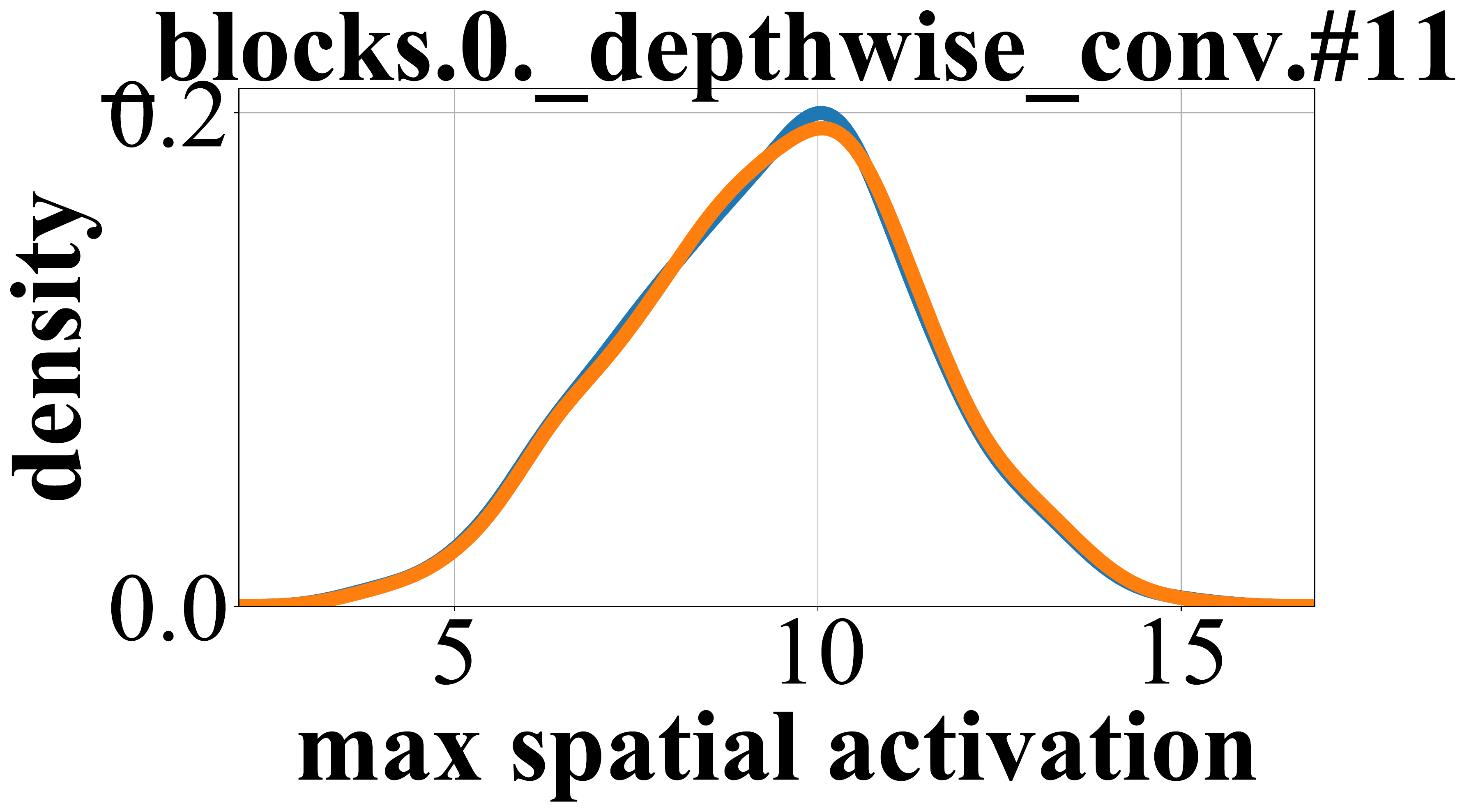}
    \\
    
     \includegraphics[width=0.13\linewidth]{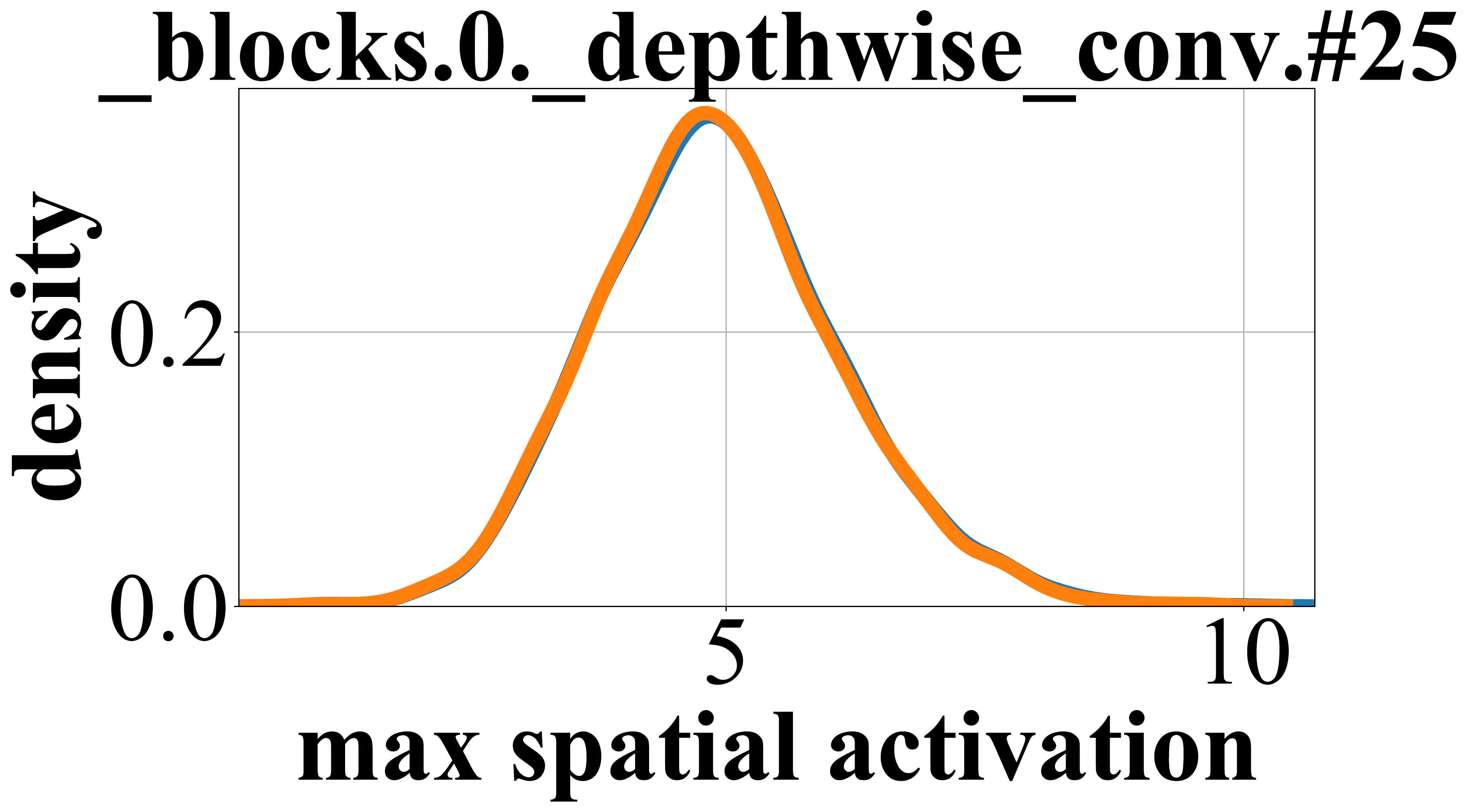} &
    \includegraphics[width=0.13\linewidth]{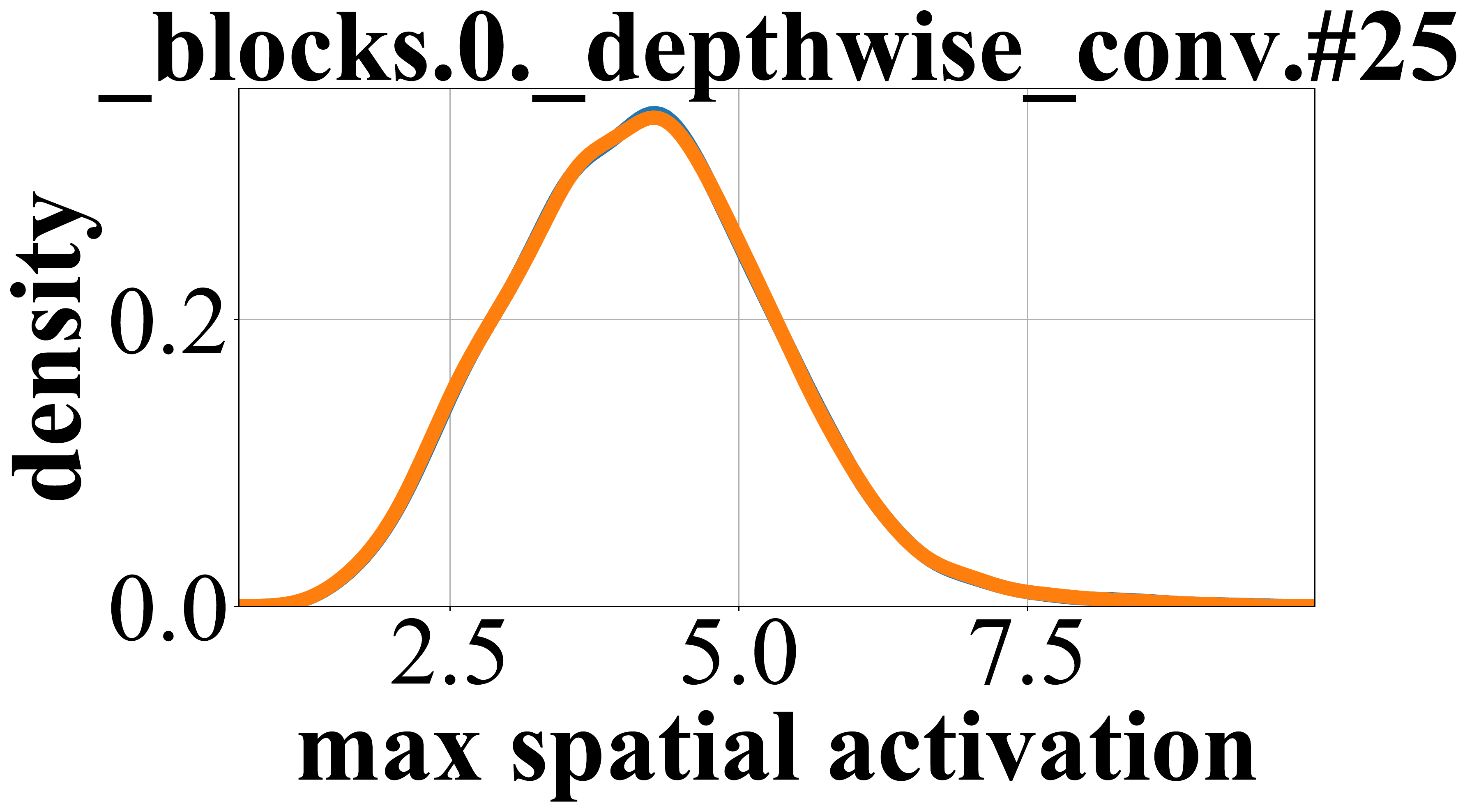} &
    \includegraphics[width=0.13\linewidth]{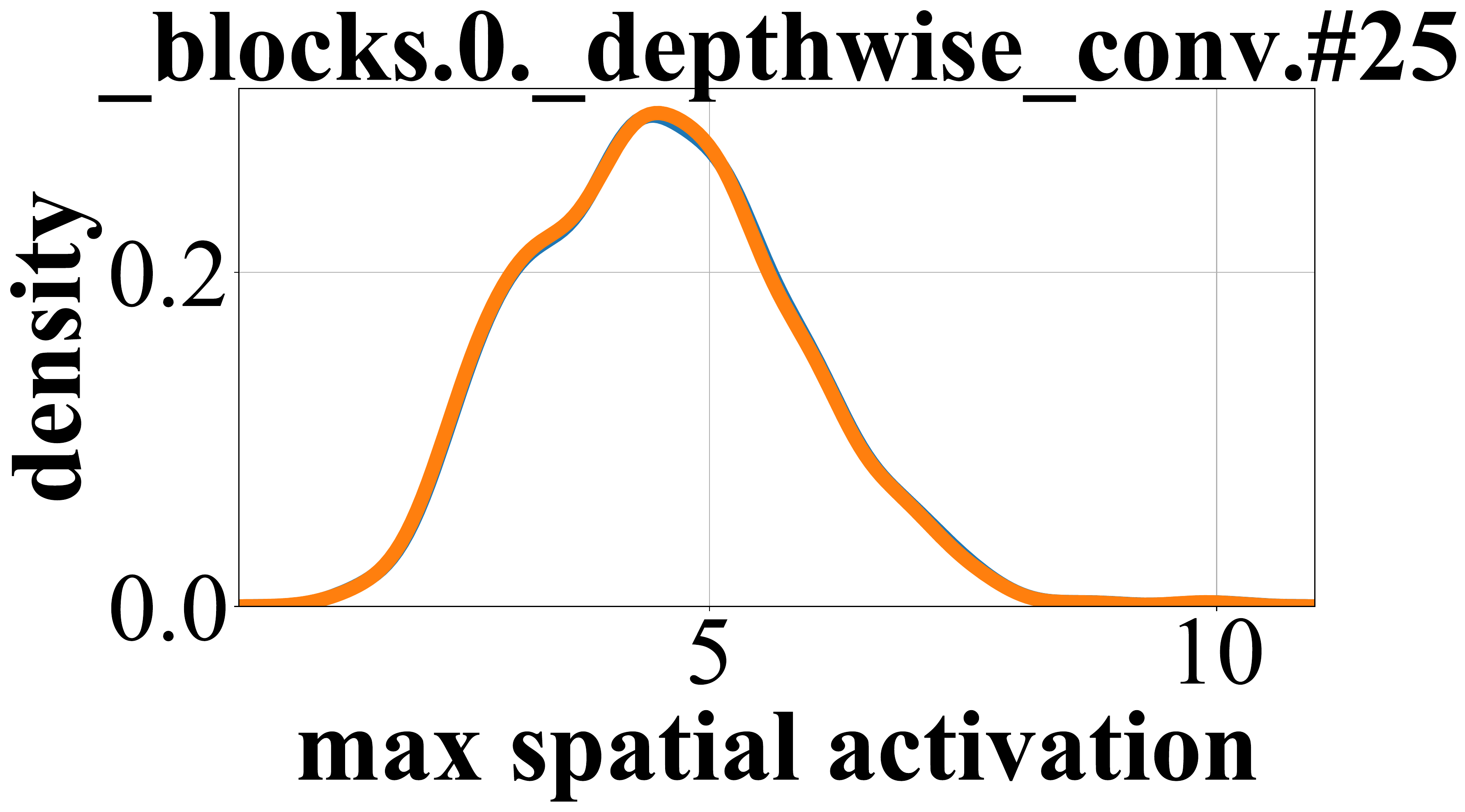} &
    \includegraphics[width=0.13\linewidth]{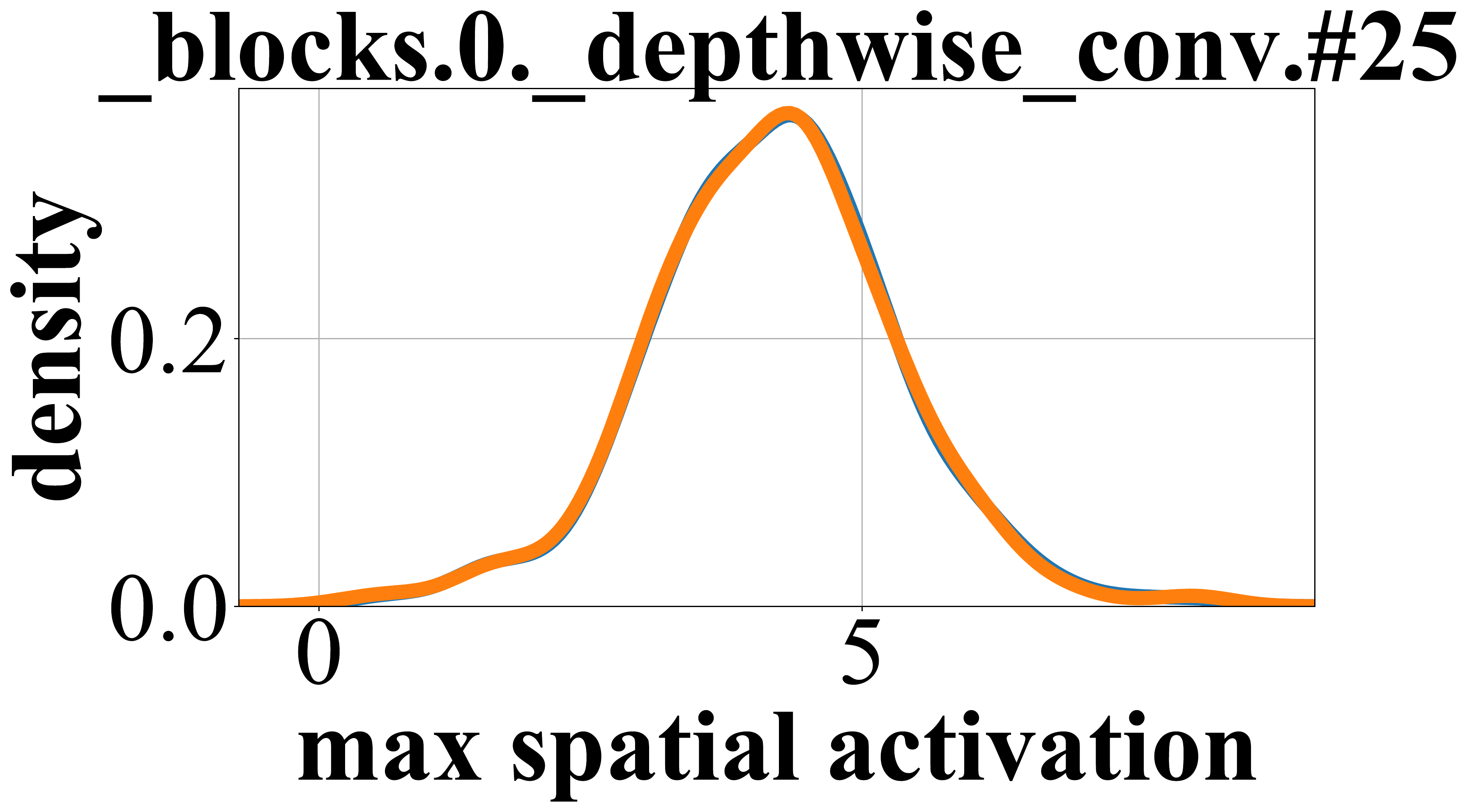} &
    \includegraphics[width=0.13\linewidth]{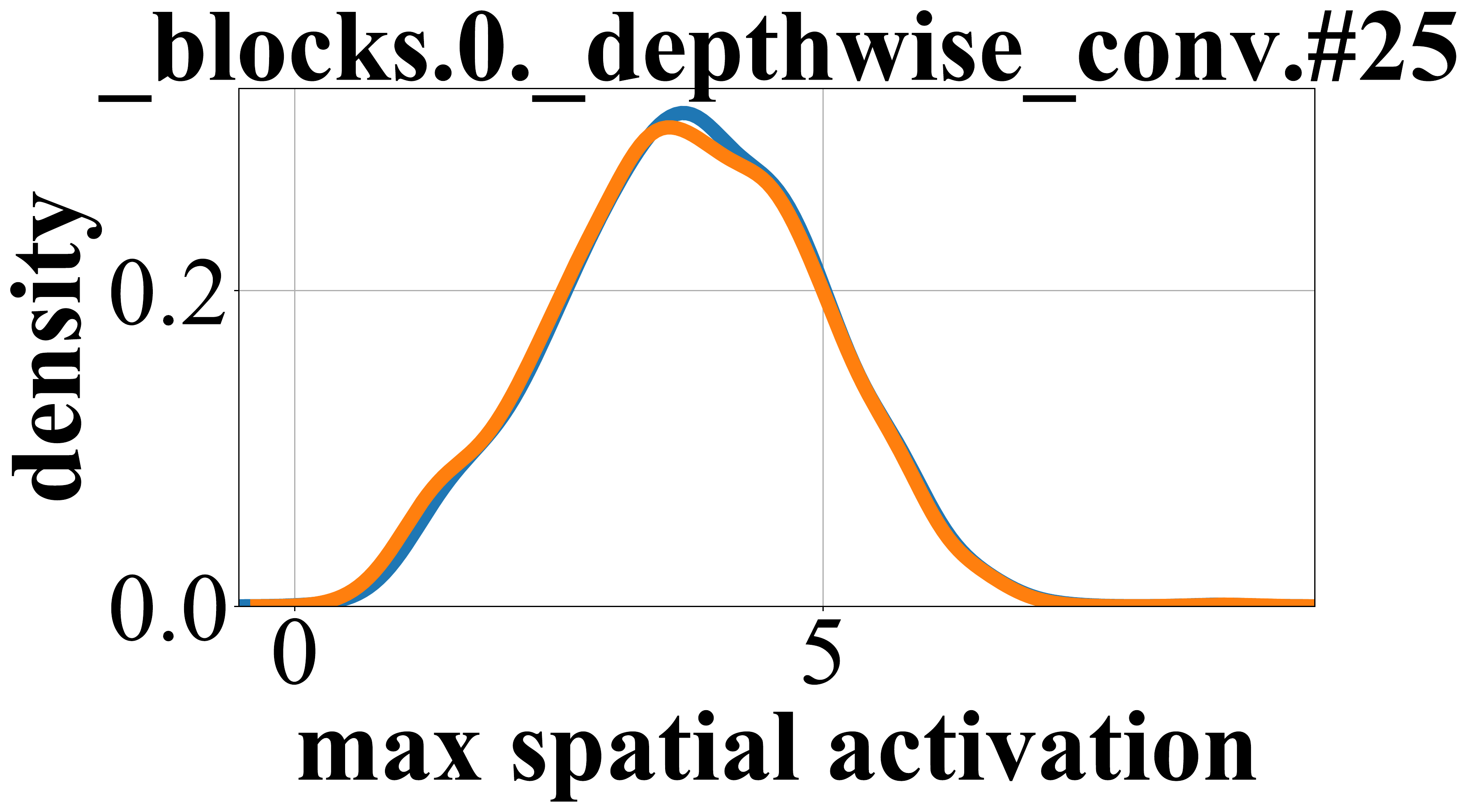} &
     \includegraphics[width=0.13\linewidth]{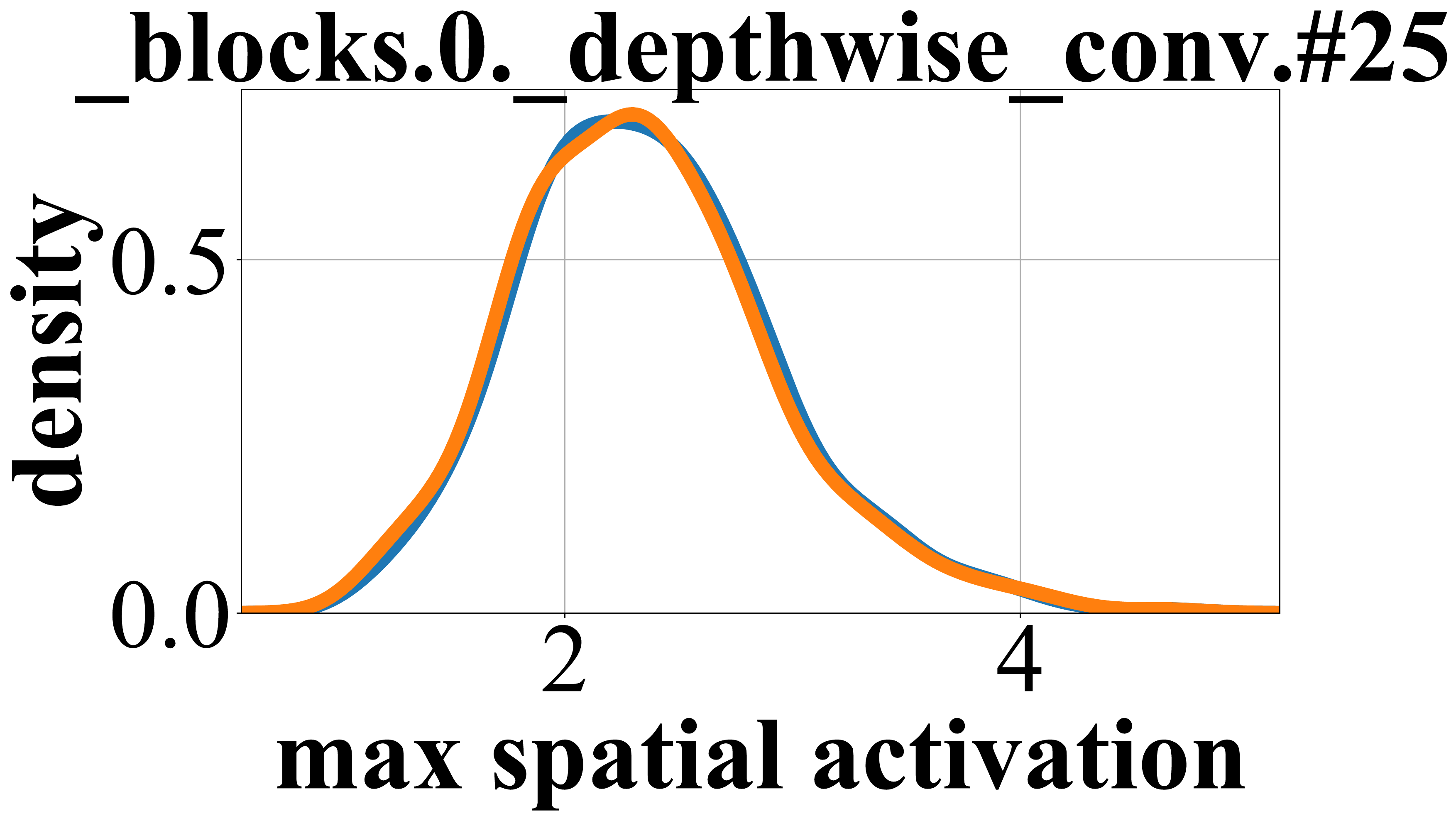} &
    \includegraphics[width=0.13\linewidth]{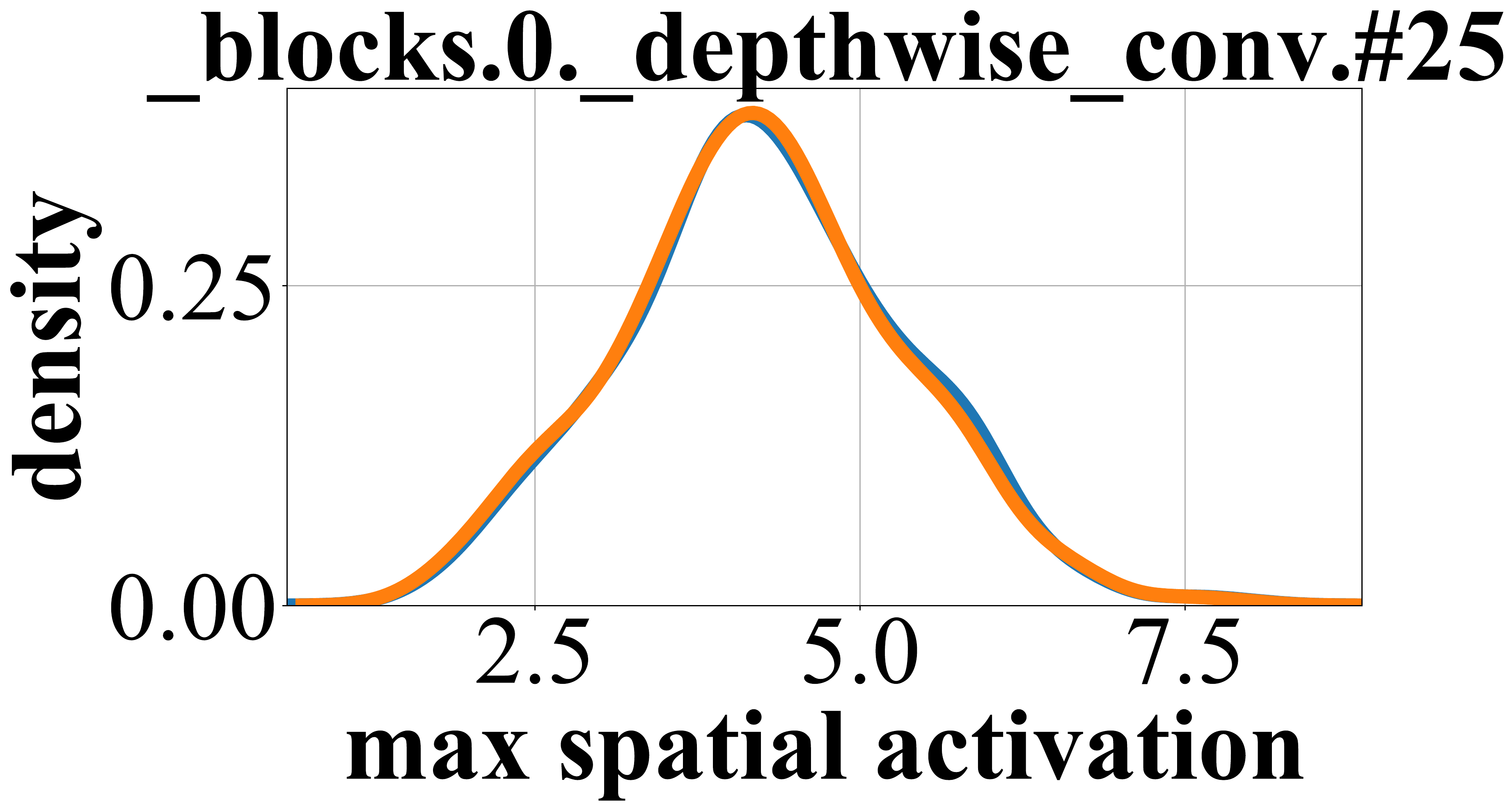}
    \\
    
     \includegraphics[width=0.13\linewidth]{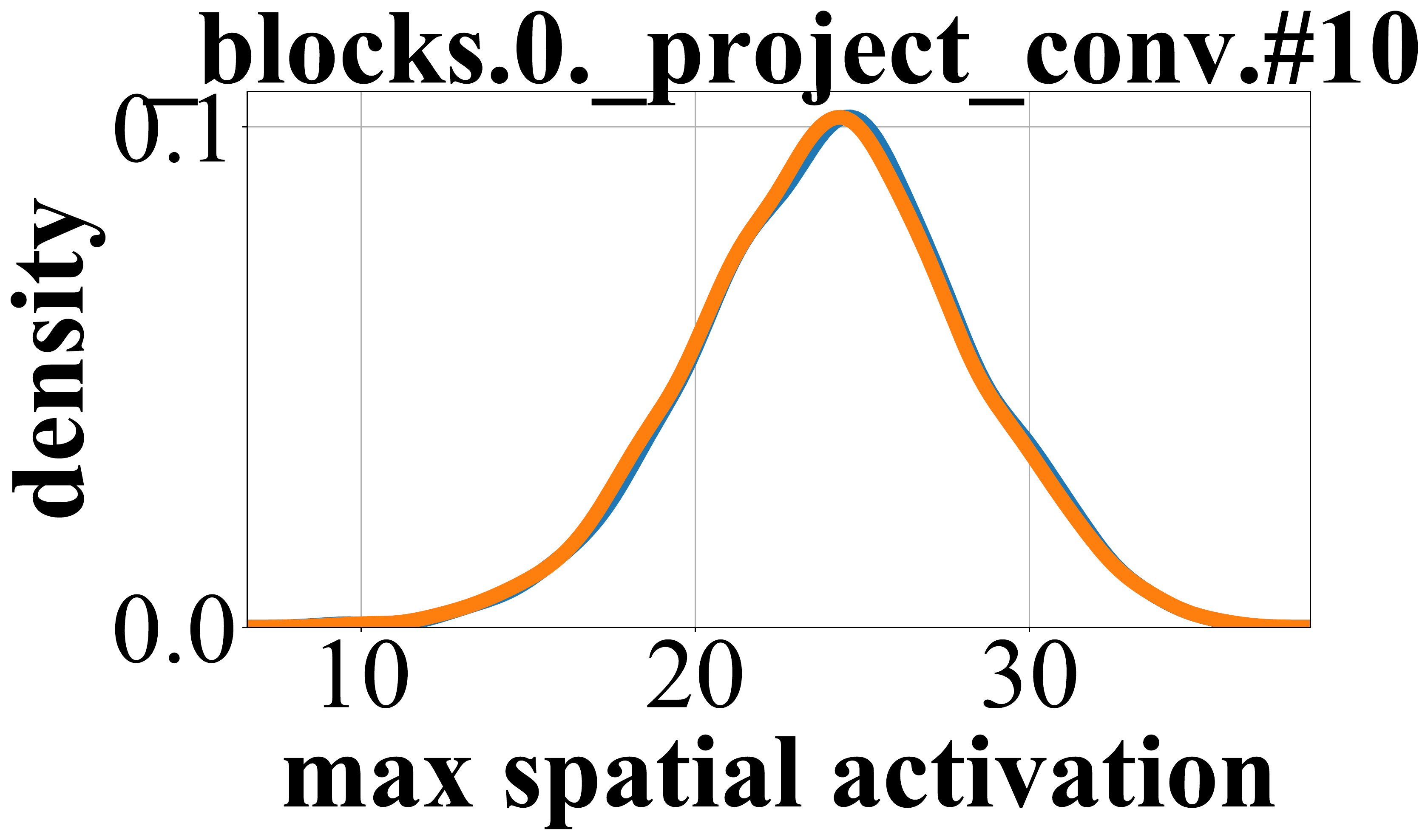} &
    \includegraphics[width=0.13\linewidth]{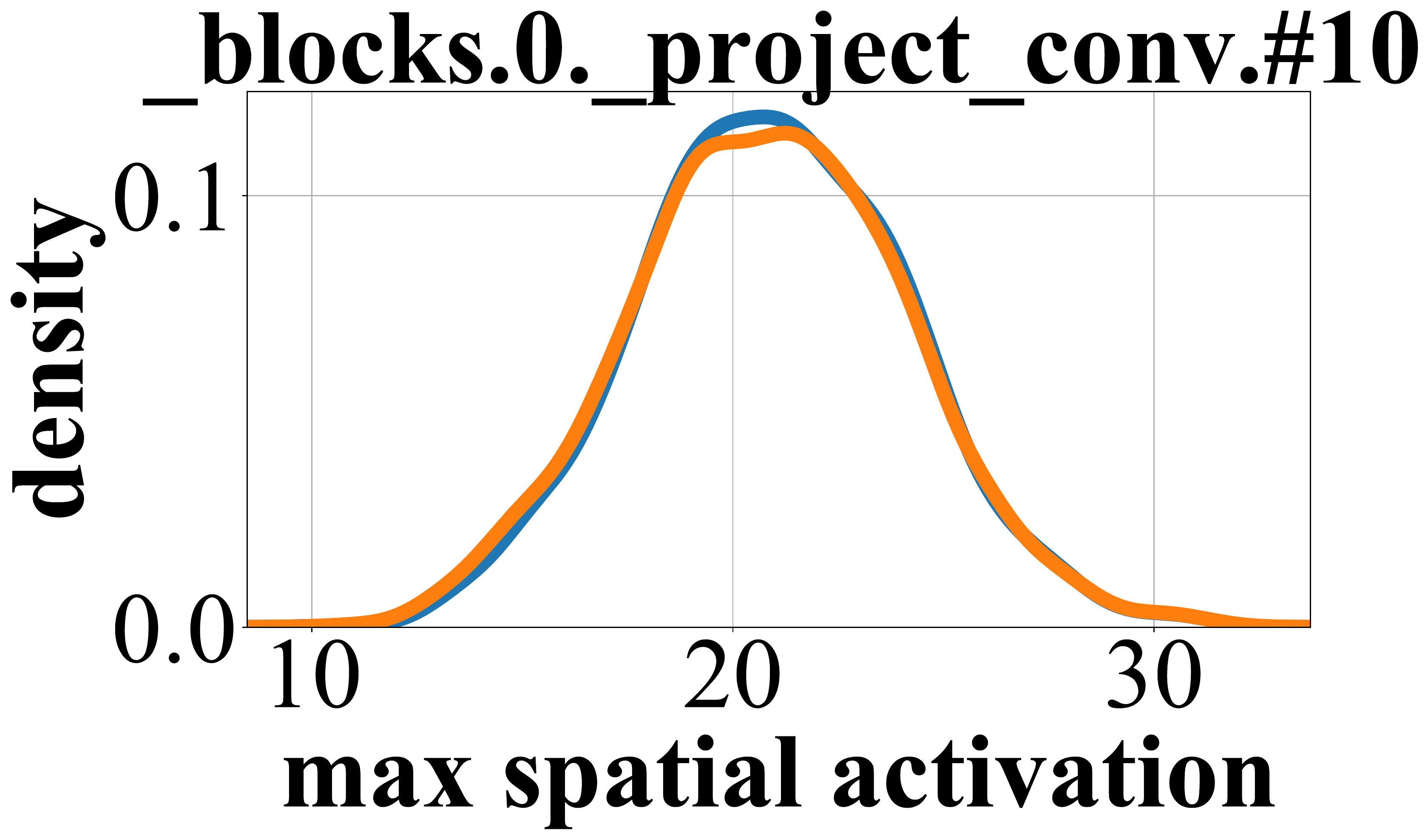} &
    \includegraphics[width=0.13\linewidth]{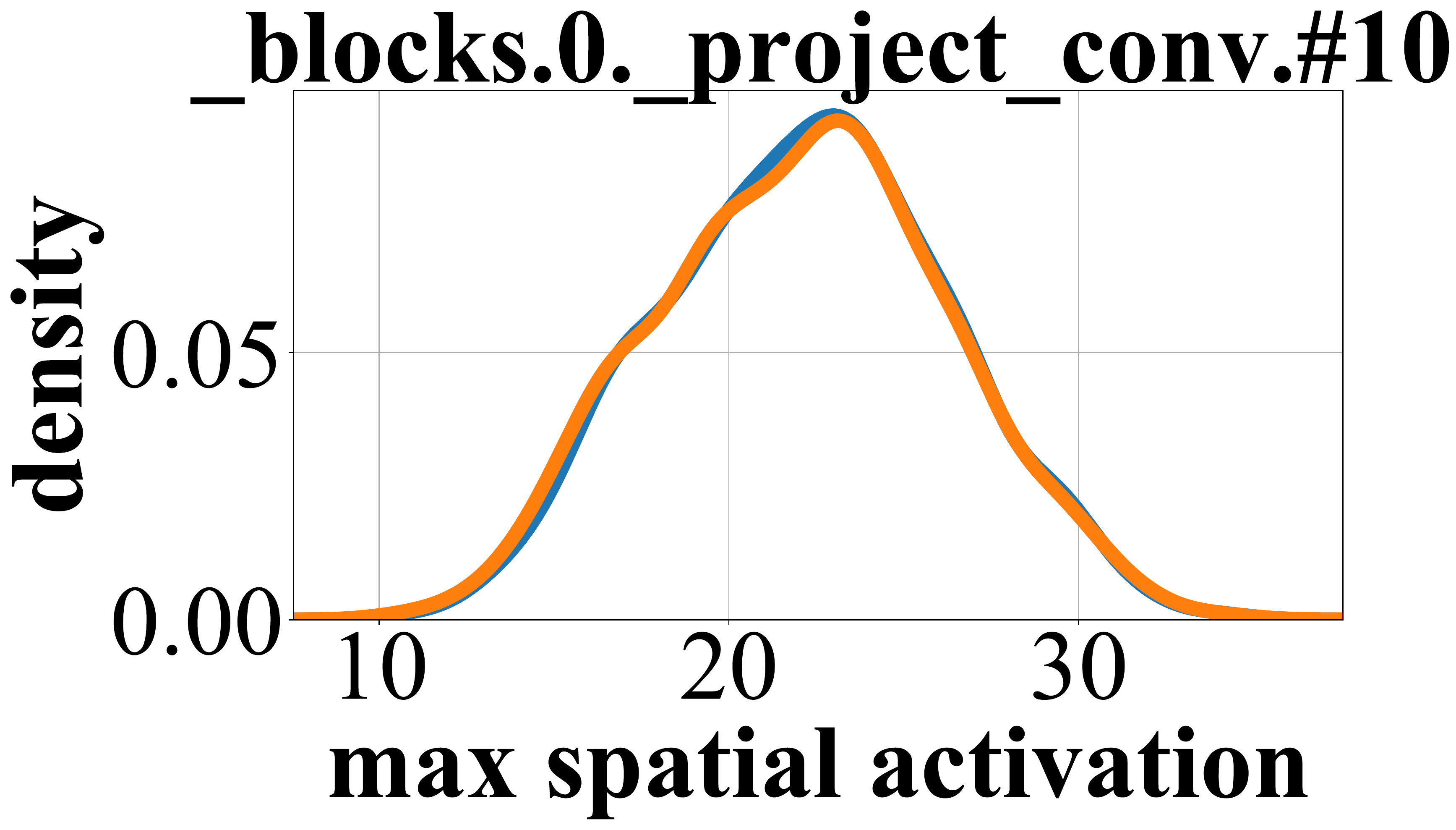} &
    \includegraphics[width=0.13\linewidth]{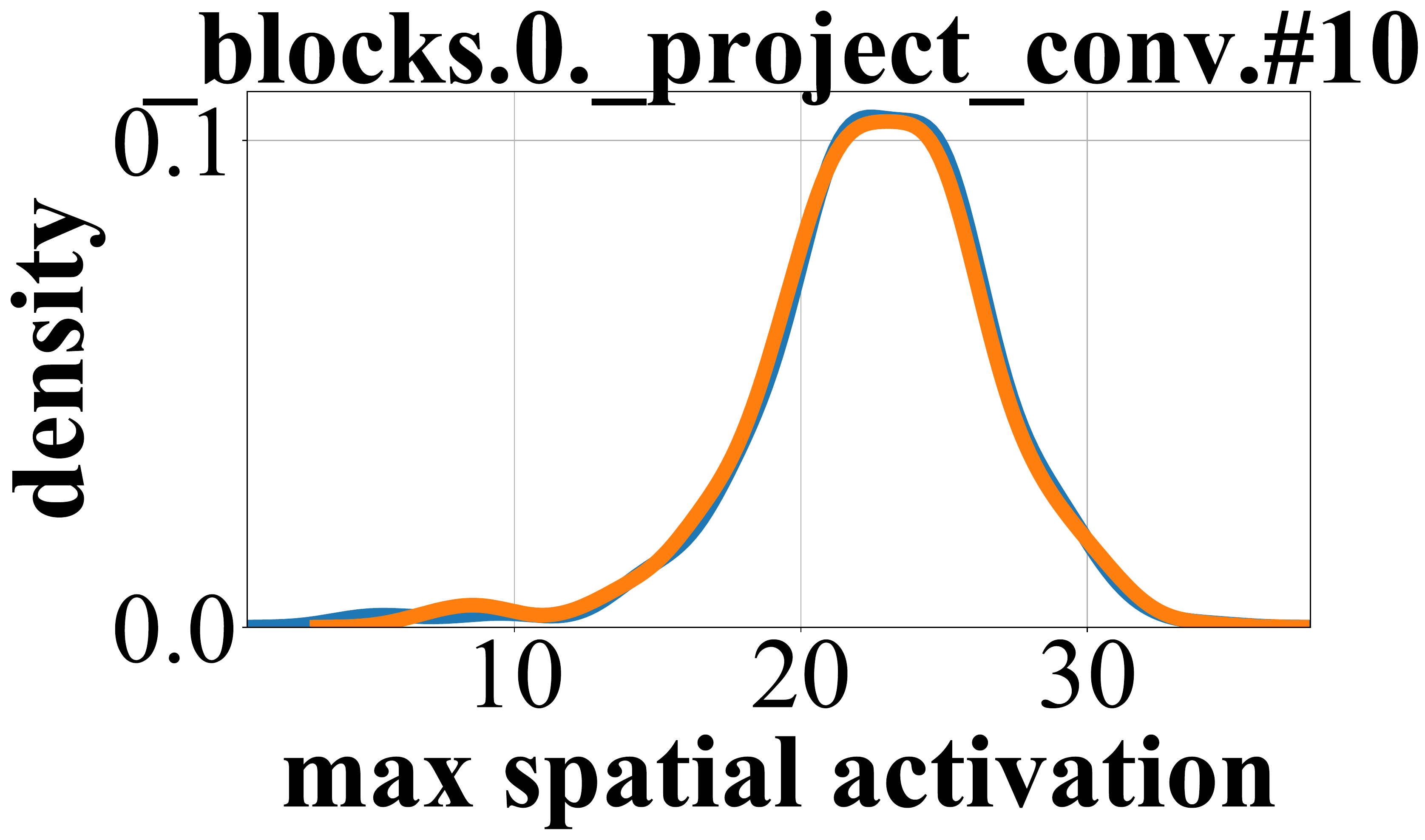} &
    \includegraphics[width=0.13\linewidth]{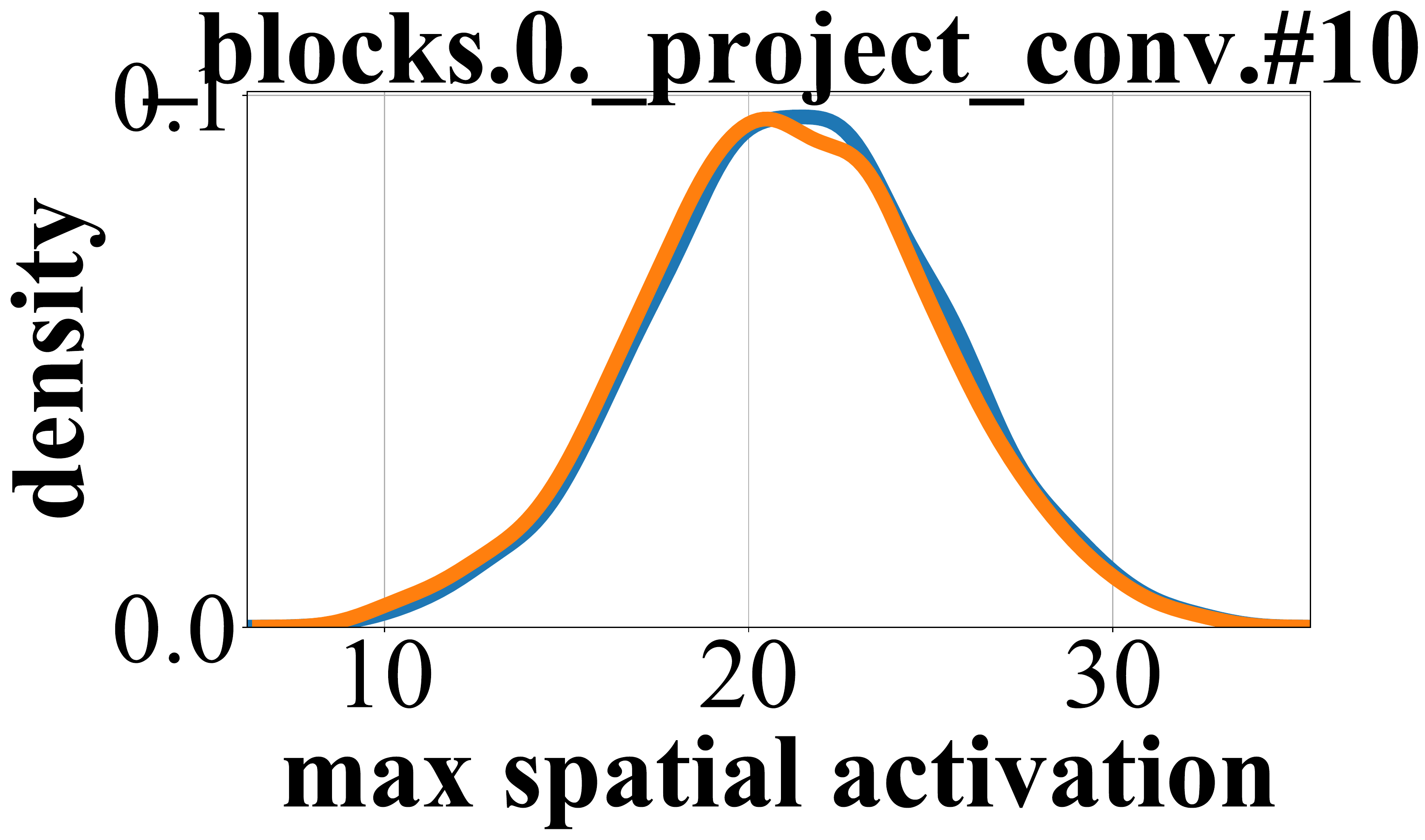} &
     \includegraphics[width=0.13\linewidth]{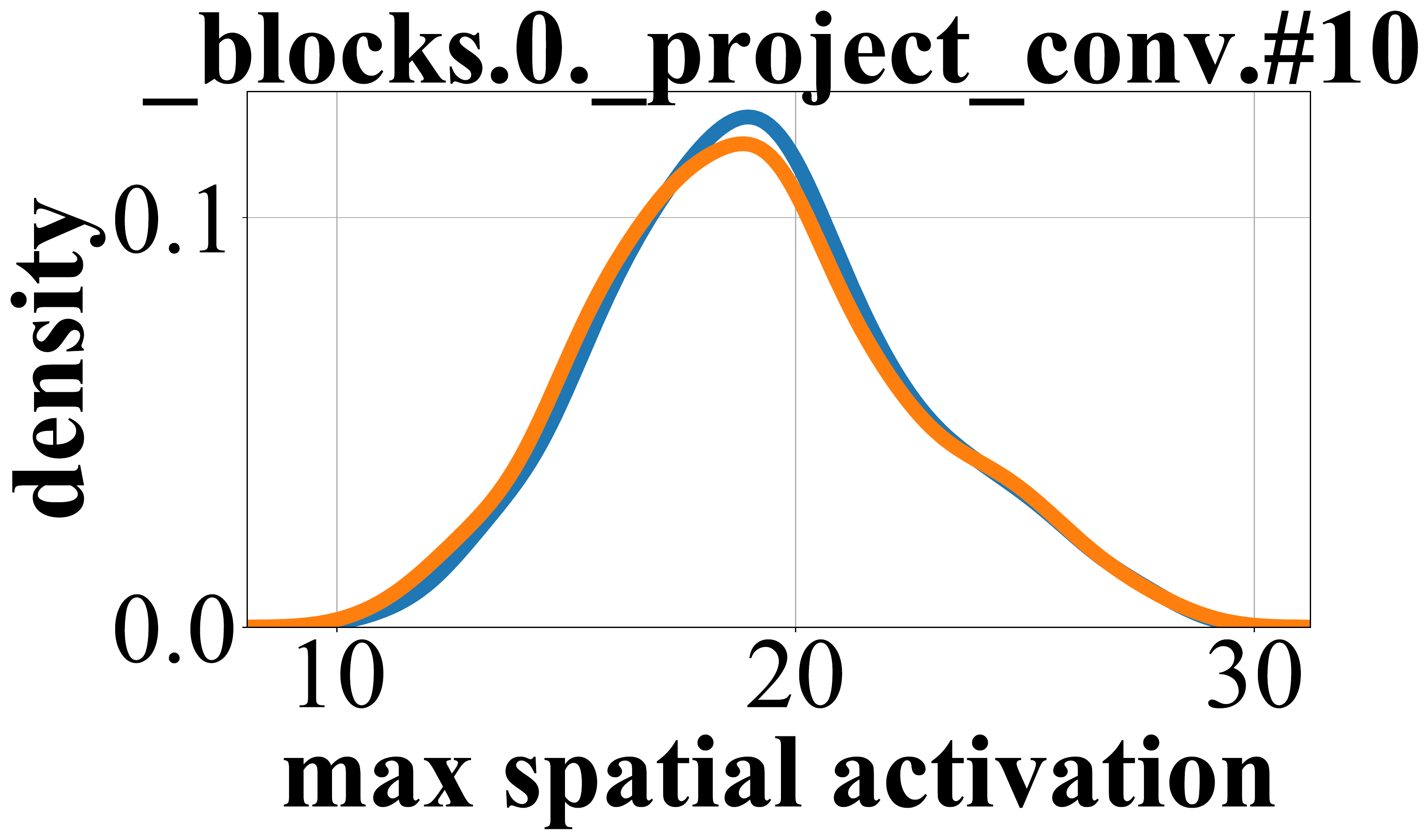} &
    \includegraphics[width=0.13\linewidth]{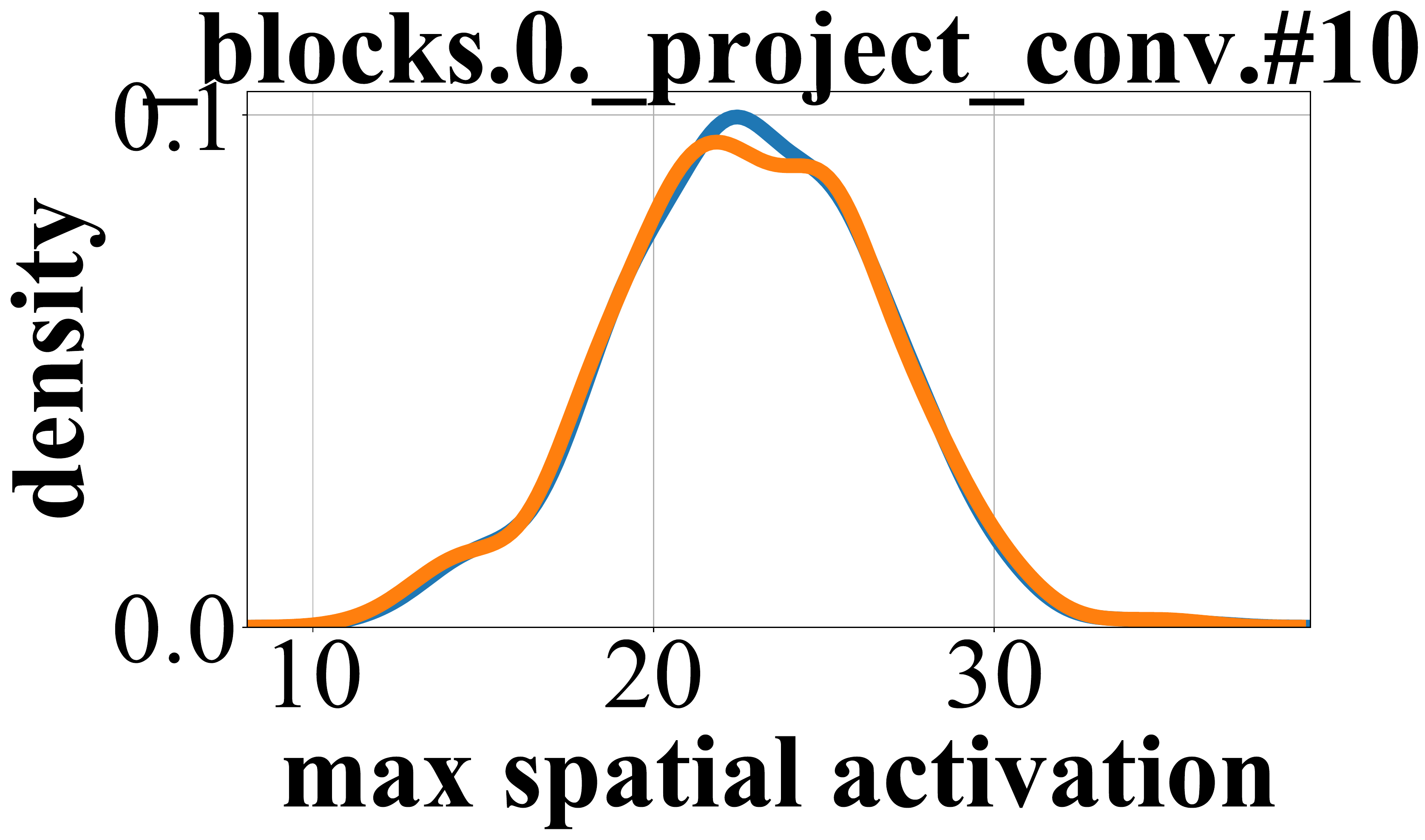}
    \\

     \includegraphics[width=0.13\linewidth]{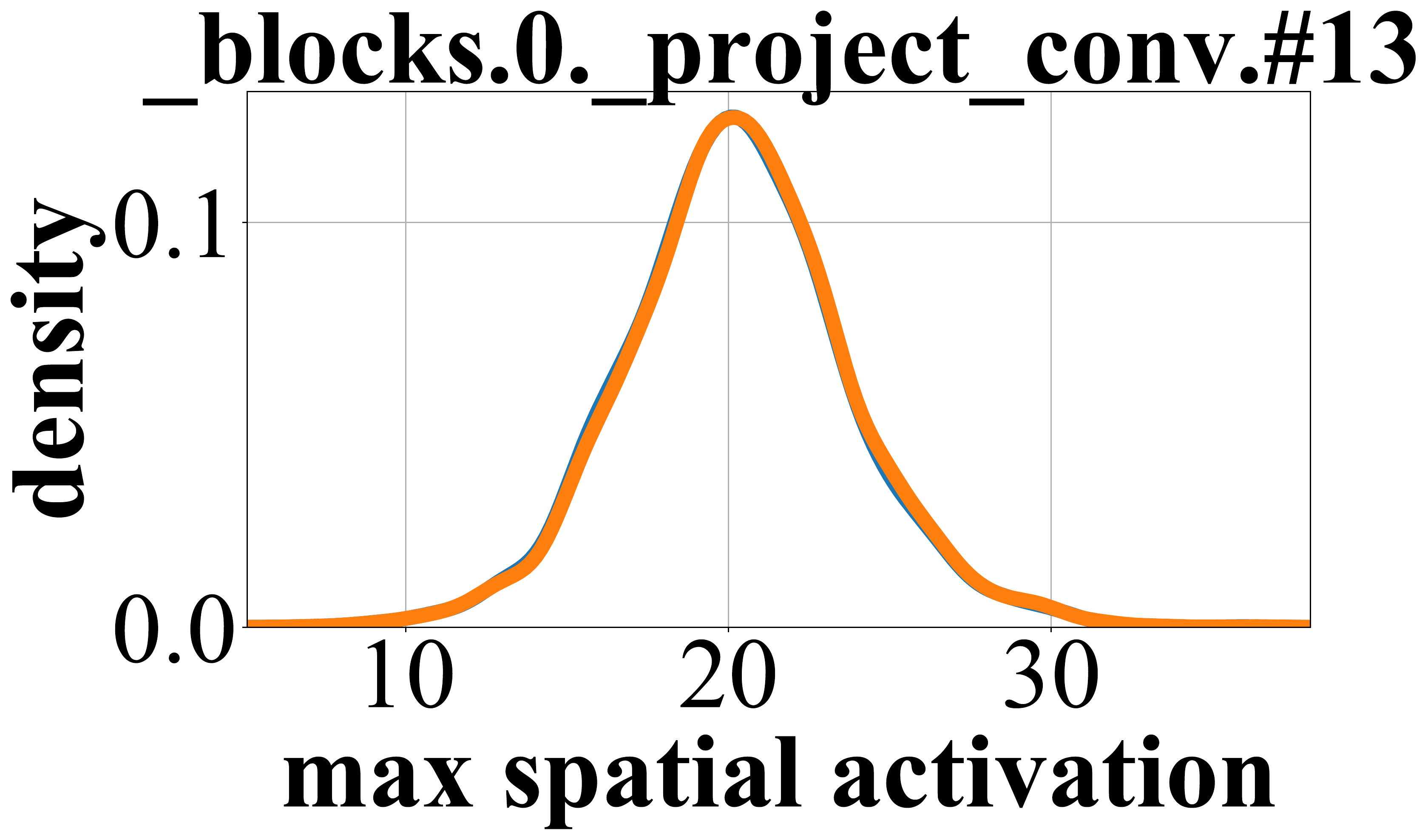} &
    \includegraphics[width=0.13\linewidth]{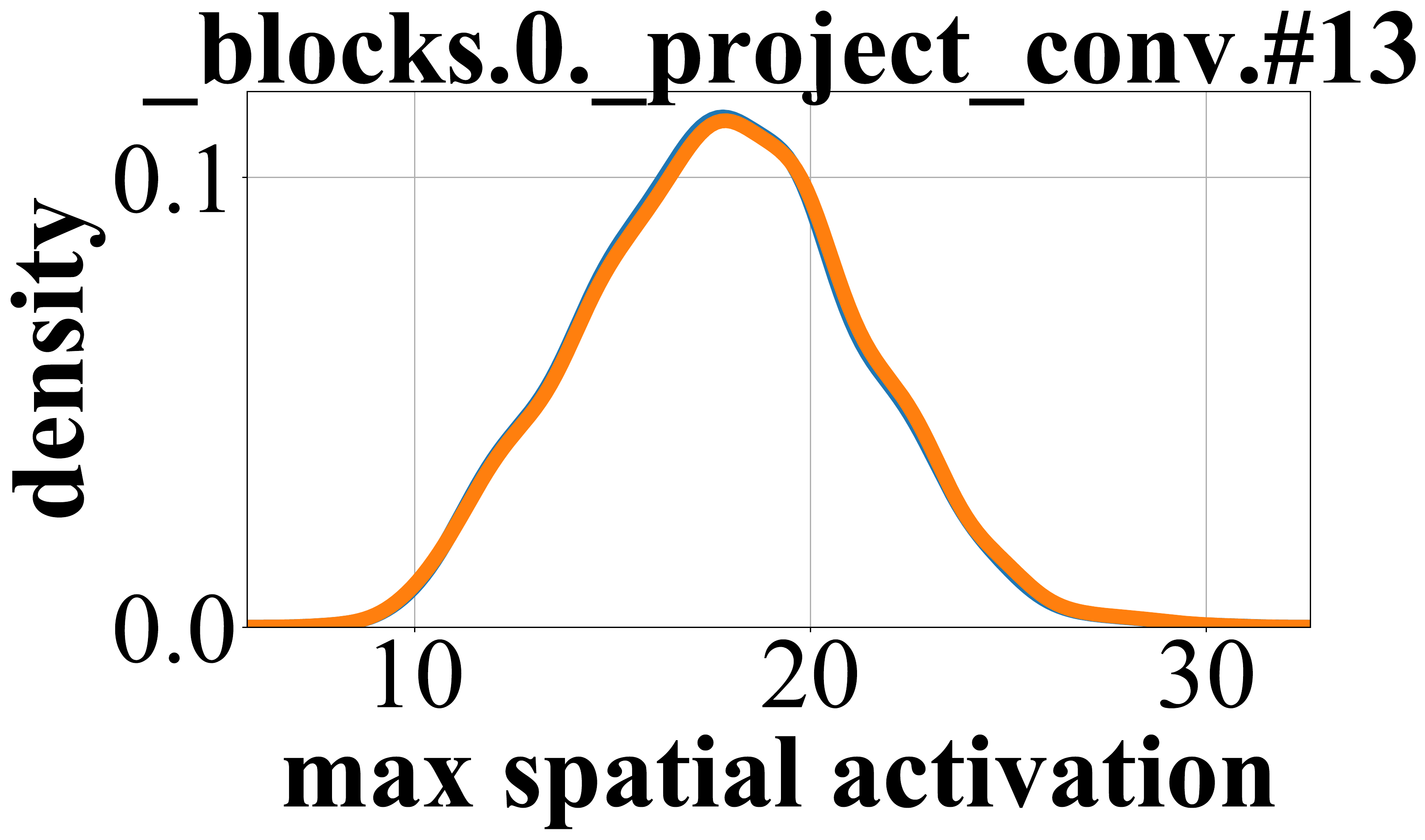} &
    \includegraphics[width=0.13\linewidth]{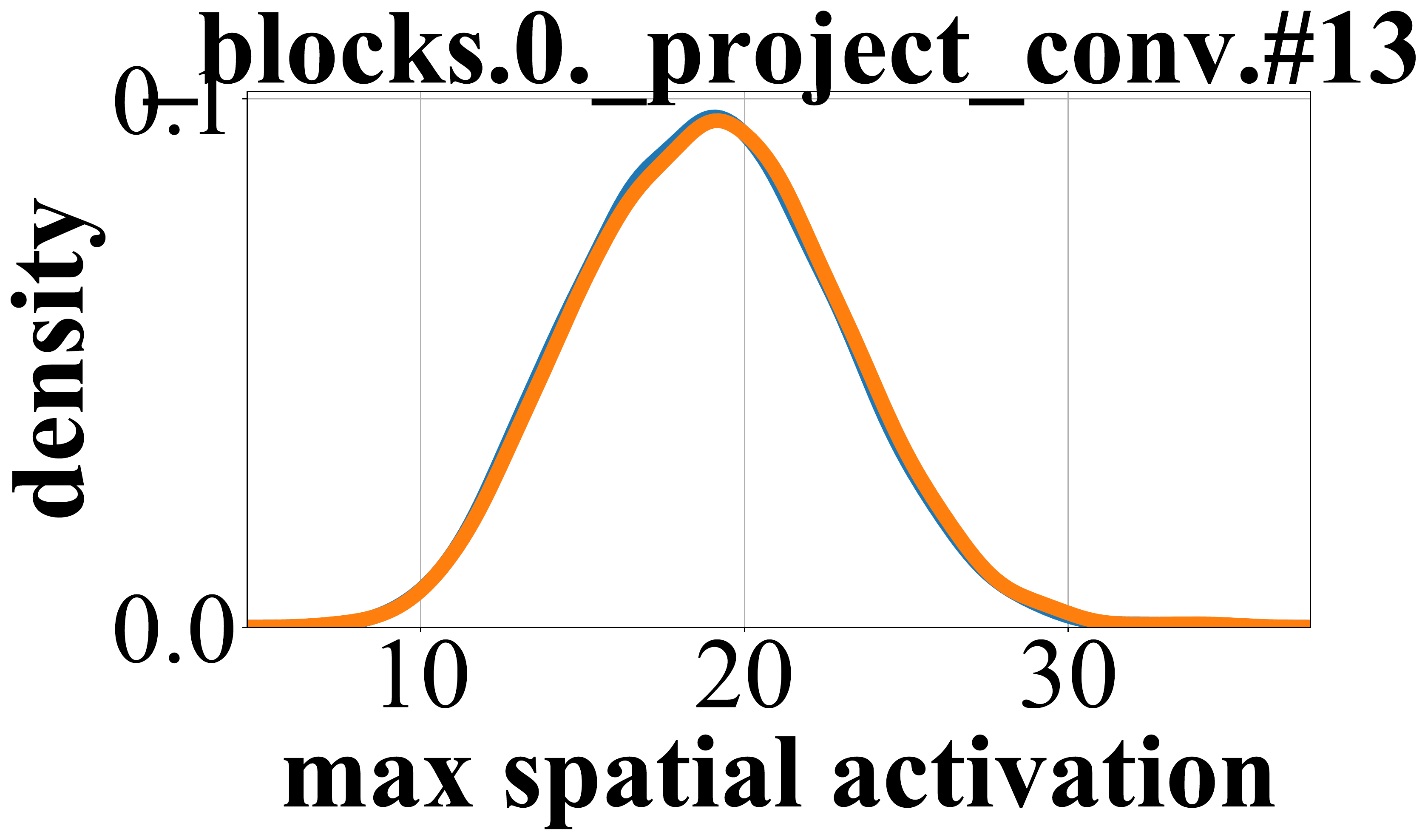} &
    \includegraphics[width=0.13\linewidth]{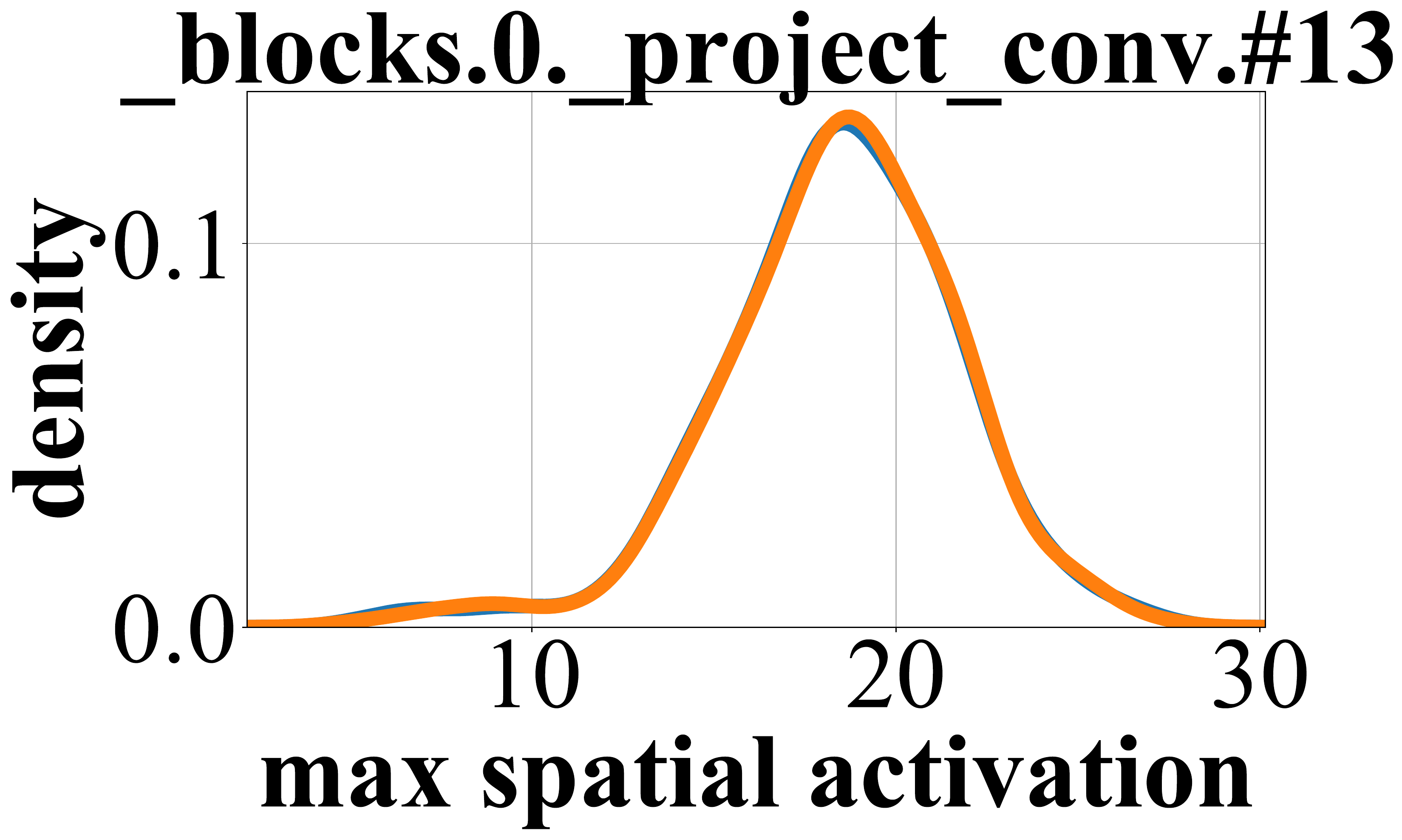} &
    \includegraphics[width=0.13\linewidth]{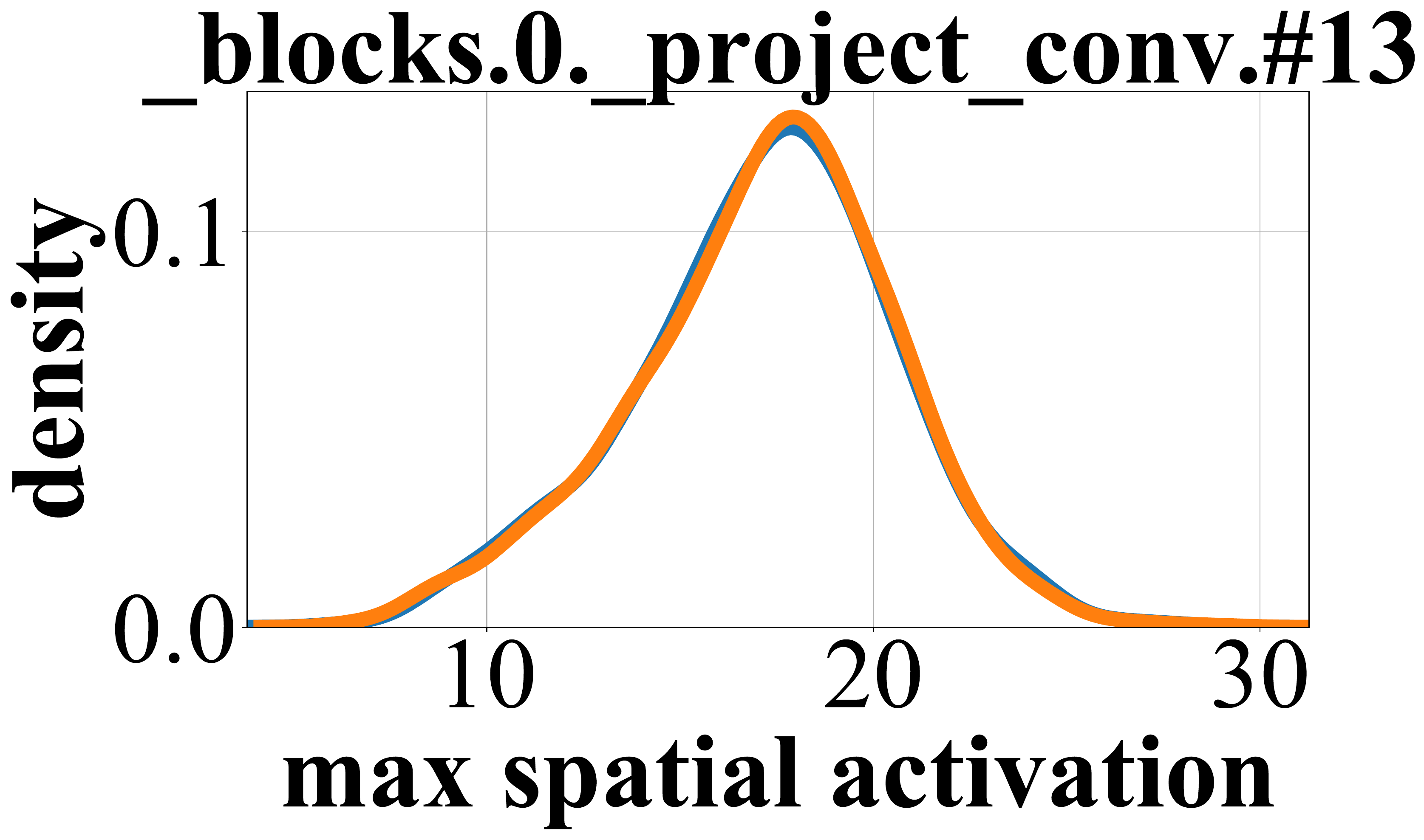} &
     \includegraphics[width=0.13\linewidth]{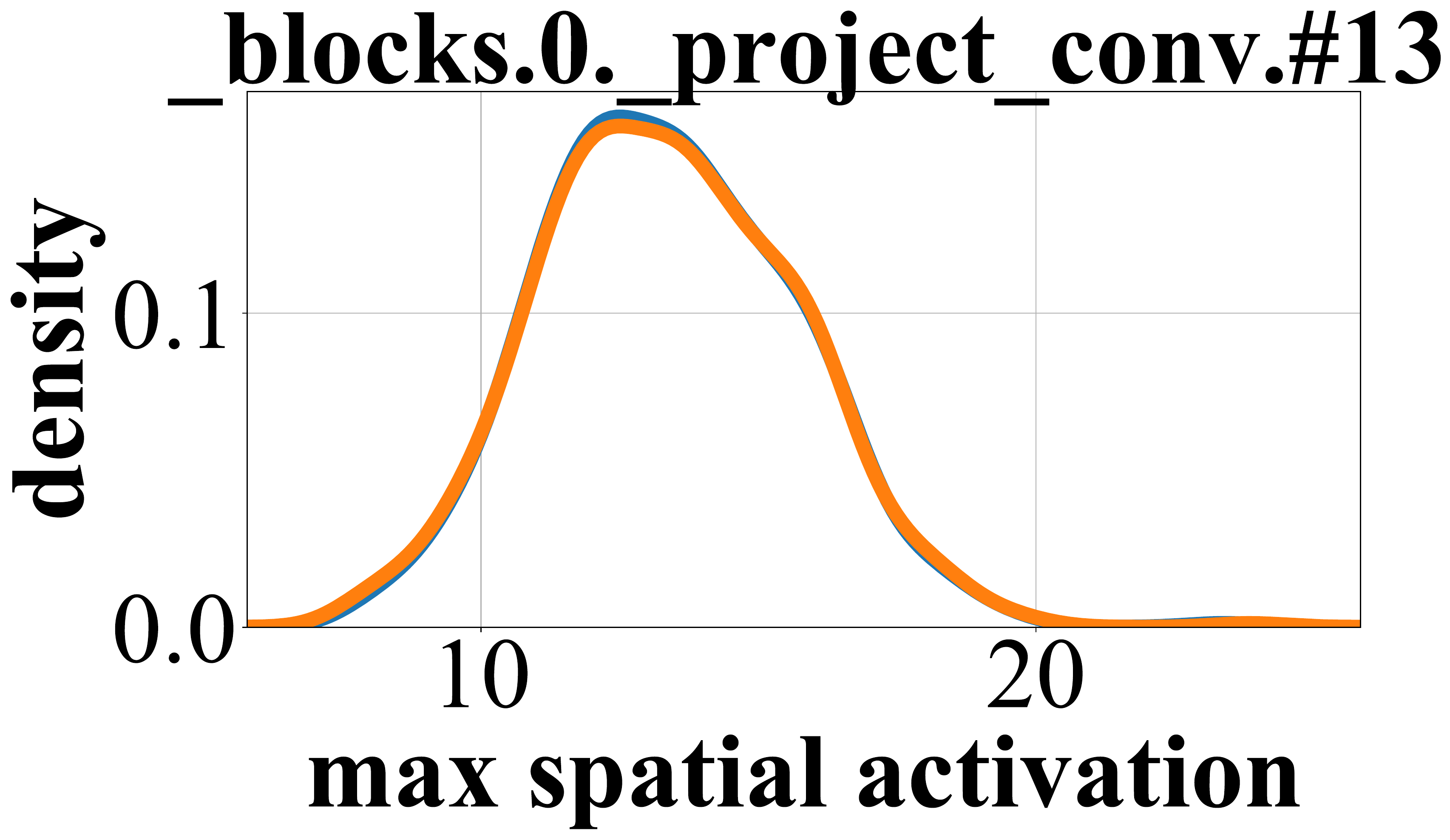} &
    \includegraphics[width=0.13\linewidth]{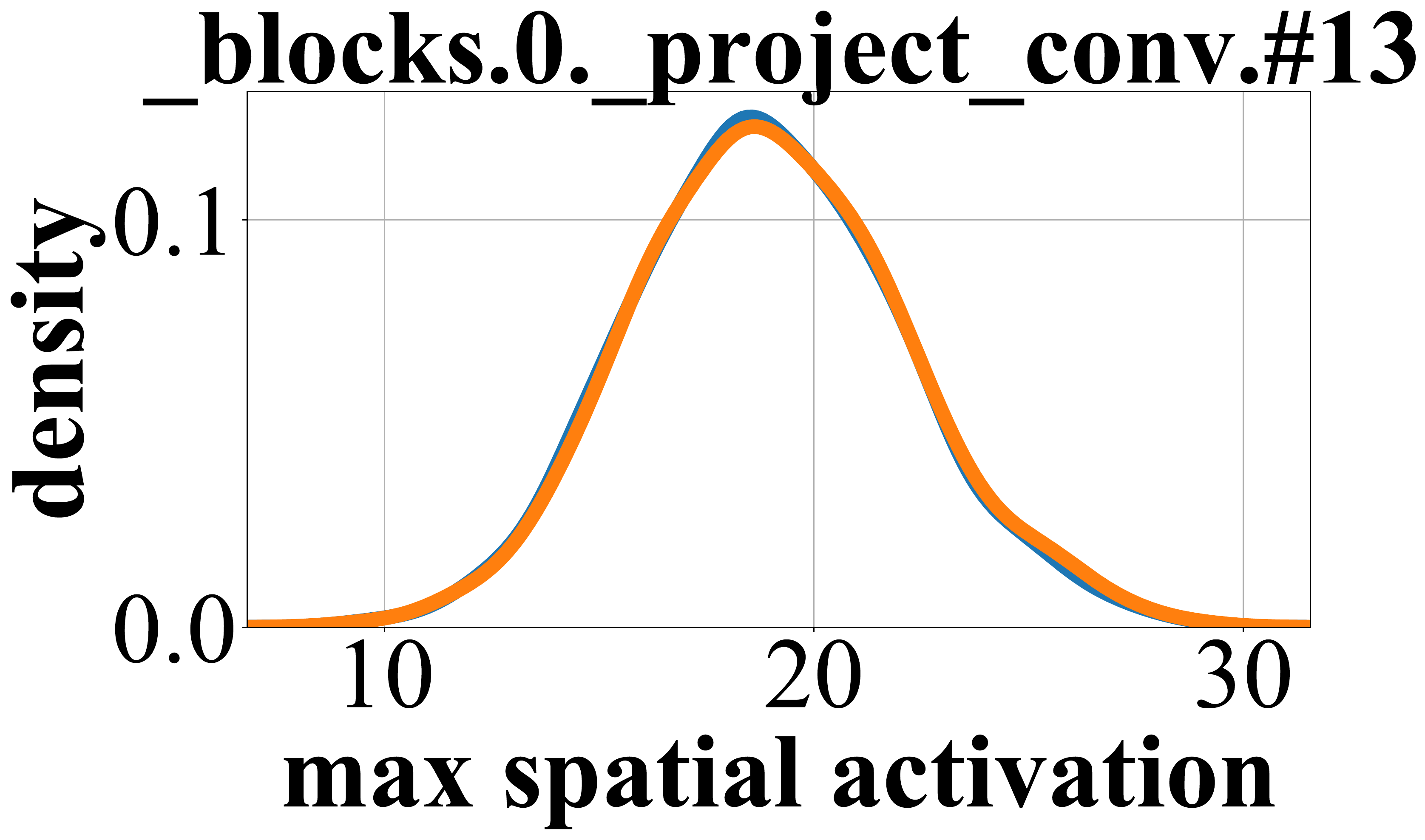}
    \\

\end{tabular}
\includegraphics[width=0.50\linewidth]{pics_supp/activation_histograms/r50_0.5/conv1-1/legend.pdf}
\caption{
\textit{Non Color-conditional T-FF in EfficientNet-B0:}
Each row represents a \textit{non} color-conditional \textit{T-FF} (exact same T-FF as shown in Fig. \ref{fig_supp:lrp_patches_efb0_non_color}), and 
we show the maximum spatial activation distributions
for ProGAN \cite{karras2018progressive}, StyleGAN2 \cite{Karras_2020_CVPR}, StyleGAN \cite{Karras_2019_CVPR}, BigGAN \cite{brock2018large}, CycleGAN \cite{zhu2017unpaired}, StarGAN \cite{choi2018stargan} and GauGAN \cite{park2019semantic} counterfeits
before (Baseline) and after color ablation (Grayscale).
We remark that for each counterfeit in the ForenSynths dataset \cite{Wang_2020_CVPR}, we apply global max pooling to the specific T-FF to obtain a {\em maximum spatial activation} value (scalar).
We can clearly observe that these \textit{T-FF} are producing identical spatial activations (max) for the same set of counterfeits after removing color information which demonstrates that these \textit{T-FF} do not respond to color information.
This clearly indicates that these \textit{T-FF} are \textit{not} color-conditional.
(Confirmed by our Mood's median test).
}
\label{fig_supp:activation_hist_efb0_non_color}
\end{figure}

There are a few \textit{T-FF} that are not color-conditional.
In this section, we show \textit{non} color-conditional T-FF.
We show LRP-max response image regions for ResNet-50 and EfficientNet-B0 in Fig. \ref{fig_supp:lrp_patches_r50_non_color} and \ref{fig_supp:lrp_patches_efb0_non_color} respectively.
We further show the maximum spatial activation distributions before and after color ablation for ResNet-50 and EfficientNet-B0 in Fig. \ref{fig_supp:activation_hist_r50_non_color} and \ref{fig_supp:activation_hist_efb0_non_color} respectively.
As one can observe using LRP-max response image regions, these \textit{non} color-conditional \textit{T-FF} contain frequency / texture artifacts. 
The maximum spatial activation distributions clearly show that these \textit{non} color-conditional \textit{T-FF} produce identical / similar distributions before and after color ablation.

\section{$k$ hyper-parameter in top-$k$ for \textit{T-FF}}
\label{sec_supp:k_hyper-parameter}

In this section, we include more discussion regarding the $k$ hyper-parameter in top-$k$. 
We show that as we increase $k$, AP and GAN detection accuracies drop across ProGAN \cite{karras2018progressive} and all unseen GANs \cite{Karras_2020_CVPR,Karras_2019_CVPR,brock2018large,zhu2017unpaired,choi2018stargan,park2019semantic}.
For our analysis, we identify the \textit{smallest k} with a substantial drop in cross-model forensic transfer as indicated by AP and GAN detection accuracies. The results for ResNet-50 and EfficientNet-B0 detectors are shown in Table \ref{table_supp:find_k}

\begin{table}[h]
\caption{
Sensitivity assessments for different $k$ values using feature map dropout of discovered \textit{T-FF}:
We show the results for the publicly released ResNet-50 universal detector 
\cite{Wang_2020_CVPR} 
(top) and 
our own version of EfficientNet-B0 \cite{tan2019efficientnet}
universal detector (bottom) following the exact training / test strategy proposed in
\cite{Wang_2020_CVPR}. 
We show the AP, real and GAN detection accuracies for baseline \cite{Wang_2020_CVPR} and different top-$k$ forensic feature dropout. 
Feature map dropout is performed by suppressing (zeroing out) the resulting activations of target feature maps (i.e.: top-$k$). 
We can clearly observe that feature map dropout of topk-$k$ corresponding to \textit{T-FF} results in substantial drop in AP and GAN detection accuracies across ProGAN and all 6 unseen GANs 
\cite{Karras_2020_CVPR,Karras_2019_CVPR,brock2018large,zhu2017unpaired,choi2018stargan,park2019semantic}
as we increase $k$.
Given that we aim to identify the $smallest$ $k$, we identify $k=114 $ and $k=27$ as the suitable $k$ for ResNet-50 and EfficientNet-B0 universal detectors.
}
\begin{center}
\begin{adjustbox}{width=1.0\columnwidth,center}
\begin{tabular}{c|ccc|ccc|ccc|ccc|ccc|ccc|ccc}
\multicolumn{22}{c}{\bf \Large ResNet-50} \\ \toprule
\textbf{} &\multicolumn{3}{c}{\textbf{ProGAN} \cite{karras2018progressive}} 
&\multicolumn{3}{c}{\textbf{StyleGAN2} \cite{Karras_2020_CVPR}} 
&\multicolumn{3}{c}{\textbf{StyleGAN} \cite{Karras_2019_CVPR}} &\multicolumn{3}{c}{\textbf{BigGAN} \cite{brock2018large}} &\multicolumn{3}{c}{\textbf{CycleGAN} \cite{zhu2017unpaired} } &\multicolumn{3}{c}{\textbf{StarGAN} \cite{choi2018stargan}} &\multicolumn{3}{c}{\textbf{GauGAN} \cite{park2019semantic}} \\

\cmidrule{2-22}

\textbf{} &\textbf{AP} &\textbf{Real} &\textbf{GAN} &\textbf{AP} &\textbf{Real} &\textbf{GAN} &\textbf{AP} &\textbf{Real} &\textbf{GAN} &\textbf{AP} &\textbf{Real} &\textbf{GAN} &\textbf{AP} &\textbf{Real} &\textbf{GAN} &\textbf{AP} &\textbf{Real} &\textbf{GAN} &\textbf{AP} &\textbf{Real} &\textbf{GAN} \\
\midrule

baseline \cite{Wang_2020_CVPR} &100.0 &100.0 &100.0 &99.3 &95.5 &95.0 &99.3 &96.0 &95.6 &90.4 &83.9 &85.1 &97.9 &93.4 &92.6 &97.5 &94.0 &89.3 &98.8 &93.9 &96.4 \\

top-29 &\textbf{98.6} &99.9 &\textbf{40.7} &\textbf{84.9} &89.2 &\textbf{62.3} &\textbf{84.9} &92.9 &\textbf{52.4} &\textbf{66.8} &85.1 &\textbf{35.4} &\textbf{76.9} &89.4 &\textbf{42.2} &\textbf{87.7} &98.2 &\textbf{30.4} &\textbf{85.6} &94.0 &\textbf{45.6} \\

top-57 &\textbf{96.8} &99.9 &\textbf{26.3} &\textbf{84.0} &91.1 &\textbf{54.9} &\textbf{84.0} &92.4 &\textbf{50.6} &\textbf{63.2} &83.3 &\textbf{30.9} &\textbf{71.4} &88.9 &\textbf{30.6} &\textbf{86.0} &98.1 &\textbf{29.0} &\textbf{82.4} &92.7 &\textbf{41.2} \\

top-114 &\textbf{69.8} &99.4 &\textbf{3.2} &\textbf{56.6} &89.4 &\textbf{11.3} &\textbf{56.6} &90.6 &\textbf{13.7} &\textbf{55.4} &86.3 &\textbf{18.3} &\textbf{61.2} &91.4 &\textbf{17.4} &\textbf{72.6} &89.4 &\textbf{35.9} &\textbf{71.0} &95.0 &\textbf{18.8} \\

top-228 &\textbf{58.6} &99.3 &\textbf{2.3} &49.2 &29.2 &76.6 &49.2 &24.5 &76.2 &51.6 &48.1 &50.6 &50.2 &83.0 &16.2 &59.3 &46.7 &66.4 &60.7 &65.5 &52.5 \\
\bottomrule
\end{tabular}
\end{adjustbox}
\end{center}

\begin{center}
\begin{adjustbox}{width=1.0\columnwidth,center}
\begin{tabular}{c|ccc|ccc|ccc|ccc|ccc|ccc|ccc}
\multicolumn{22}{c}{\bf \Large EfficientNet-B0} \\ \toprule
\textbf{} &\multicolumn{3}{c}{\textbf{ProGAN} \cite{karras2018progressive}} 
&\multicolumn{3}{c}{\textbf{StyleGAN2} \cite{Karras_2020_CVPR}} 
&\multicolumn{3}{c}{\textbf{StyleGAN} \cite{Karras_2019_CVPR}} &\multicolumn{3}{c}{\textbf{BigGAN} \cite{brock2018large}} &\multicolumn{3}{c}{\textbf{CycleGAN} \cite{zhu2017unpaired} } &\multicolumn{3}{c}{\textbf{StarGAN} \cite{choi2018stargan}} &\multicolumn{3}{c}{\textbf{GauGAN} \cite{park2019semantic}} \\

\cmidrule{2-22}

\textbf{} &\textbf{AP} &\textbf{Real} &\textbf{GAN} &\textbf{AP} &\textbf{Real} &\textbf{GAN} &\textbf{AP} &\textbf{Real} &\textbf{GAN} &\textbf{AP} &\textbf{Real} &\textbf{GAN} &\textbf{AP} &\textbf{Real} &\textbf{GAN} &\textbf{AP} &\textbf{Real} &\textbf{GAN} &\textbf{AP} &\textbf{Real} &\textbf{GAN} \\
\midrule

baseline \cite{Wang_2020_CVPR}&100. &100. &100. &99.0 &95.2 &85.4 &99.0 &96.1 &94.3 &84.4 &79.7 &75.9 &97.3 &89.6 &93.0 &96.0 &92.8 &85.5 &98.3 &94.1 &94.4 \\

top-5 &\textbf{91.8} &99.9 &\textbf{14.5} &\textbf{68.9} &75.1 &\textbf{53.7} &\textbf{68.9} &74.6 &\textbf{38.3} &\textbf{57.4} &74.6 &\textbf{38.3} &\textbf{78.9} &85.5 &\textbf{54.4} &\textbf{82.4} &94.2 &\textbf{40.8} &\textbf{70.7} &97.4 &\textbf{13.9} \\

top-27 &\textbf{50.0} &100. &\textbf{0.0} &\textbf{52.1} &94.3 &\textbf{7.0} &\textbf{52.1} &97.3 &\textbf{2.6} &\textbf{53.5} &97.4 &\textbf{3.8} &\textbf{47.5} &100.0 &\textbf{0.0} &\textbf{50.0} &100. &\textbf{0.0} &\textbf{46.2} &100. &\textbf{0.0} \\

top-49 &\textbf{50.0} &100. &\textbf{0.0} &\textbf{50.0} &100. &\textbf{0.0} &\textbf{50.0} &100. &\textbf{0.0} &\textbf{50.0} &100. &\textbf{0.0} &\textbf{50.0} &100. &\textbf{0.0} &\textbf{50.0} &100. &\textbf{0.0} &\textbf{50.0} &100. &\textbf{0.0} \\

\bottomrule
\end{tabular}
\end{adjustbox}
\end{center}

\label{table_supp:find_k}
\end{table}


\section{Cross-model forensic transfer using BigGAN \cite{brock2018large} pre-training dataset}
\label{sec_supp:biggan_transfer}
\setcounter{figure}{0} 
\setcounter{table}{0} 
In this section, we show that color is a critical \textit{T-FF} using an additional training dataset. 
We use BigGAN real / fake as second dataset with 1.04M images to train universal detectors following Wang \etal  \cite{Wang_2020_CVPR} and verify
our findings. We remark that ForenSynths \cite{Wang_2020_CVPR} uses ProGAN real / fake dataset. We perform large-scale experiments using EfficientNet-B0 universal detector. We report median counterfeit probability results for all 7 GANs \cite{Karras_2020_CVPR,Karras_2019_CVPR,brock2018large,zhu2017unpaired,choi2018stargan,park2019semantic} in Fig. \ref{fig_main:biggan_median_color_ablation}. 
Our results show on a second dataset that
color ablation causes counterfeit probability to drop by
$>50\%$ for all unseen GANs. These results on another
dataset further support that color is a critical \textit{T-FF} in universal detectors for counterfeit detection.

\begin{figure}[!t]
\centering
\begin{tabular}{ccccccc}
    {\tiny BigGAN \cite{brock2018large}} &
    {\tiny ProGAN \cite{karras2018progressive}} &
    {\tiny StyleGAN2 \cite{Karras_2020_CVPR}} &
    {\tiny StyleGAN \cite{Karras_2019_CVPR}} &
    {\tiny CycleGAN \cite{zhu2017unpaired}} &
    {\tiny StarGAN \cite{choi2018stargan}} &
    {\tiny GauGAN \cite{park2019semantic}} \\
    
    \includegraphics[width=0.13\linewidth]{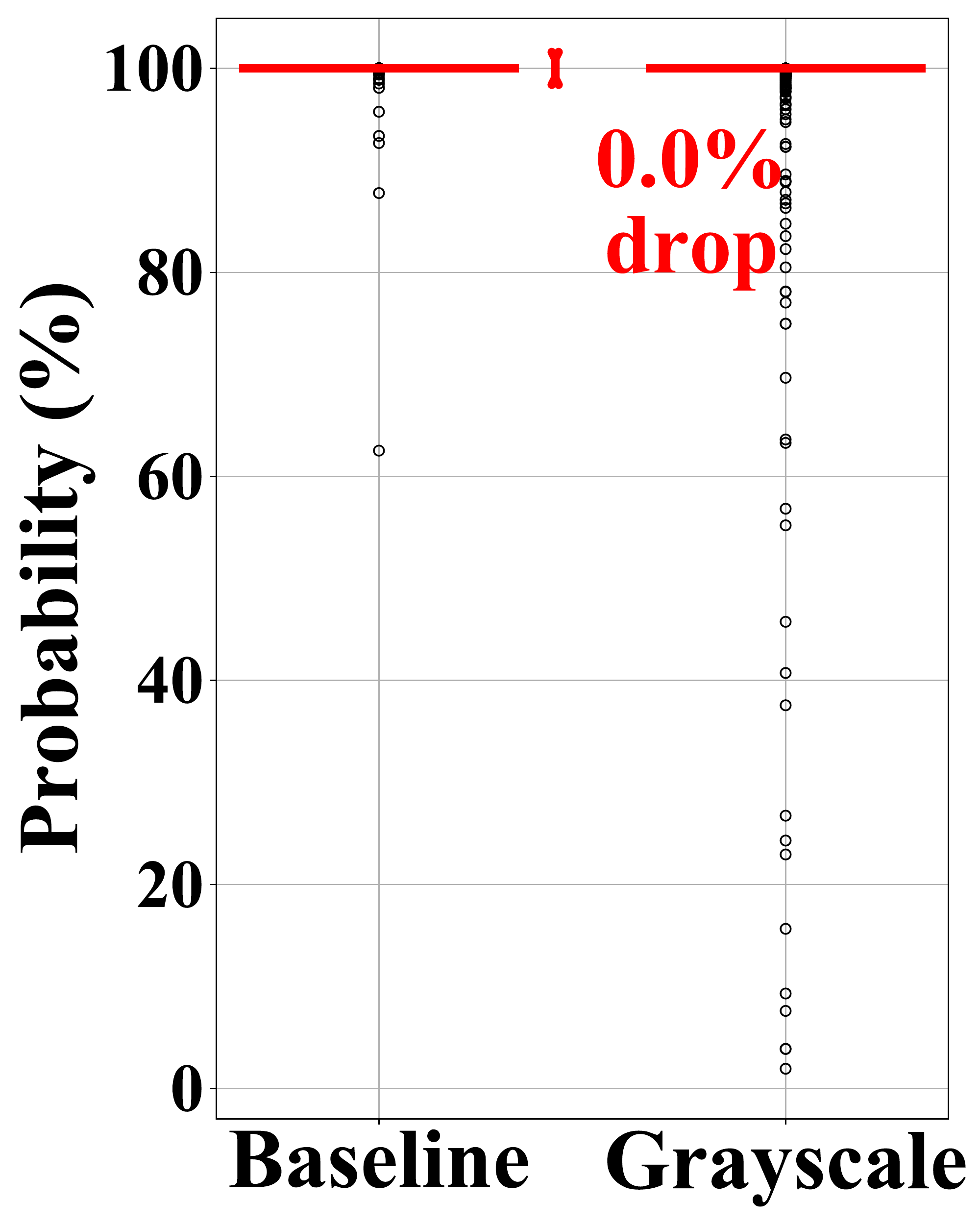} &
    \includegraphics[width=0.13\linewidth]{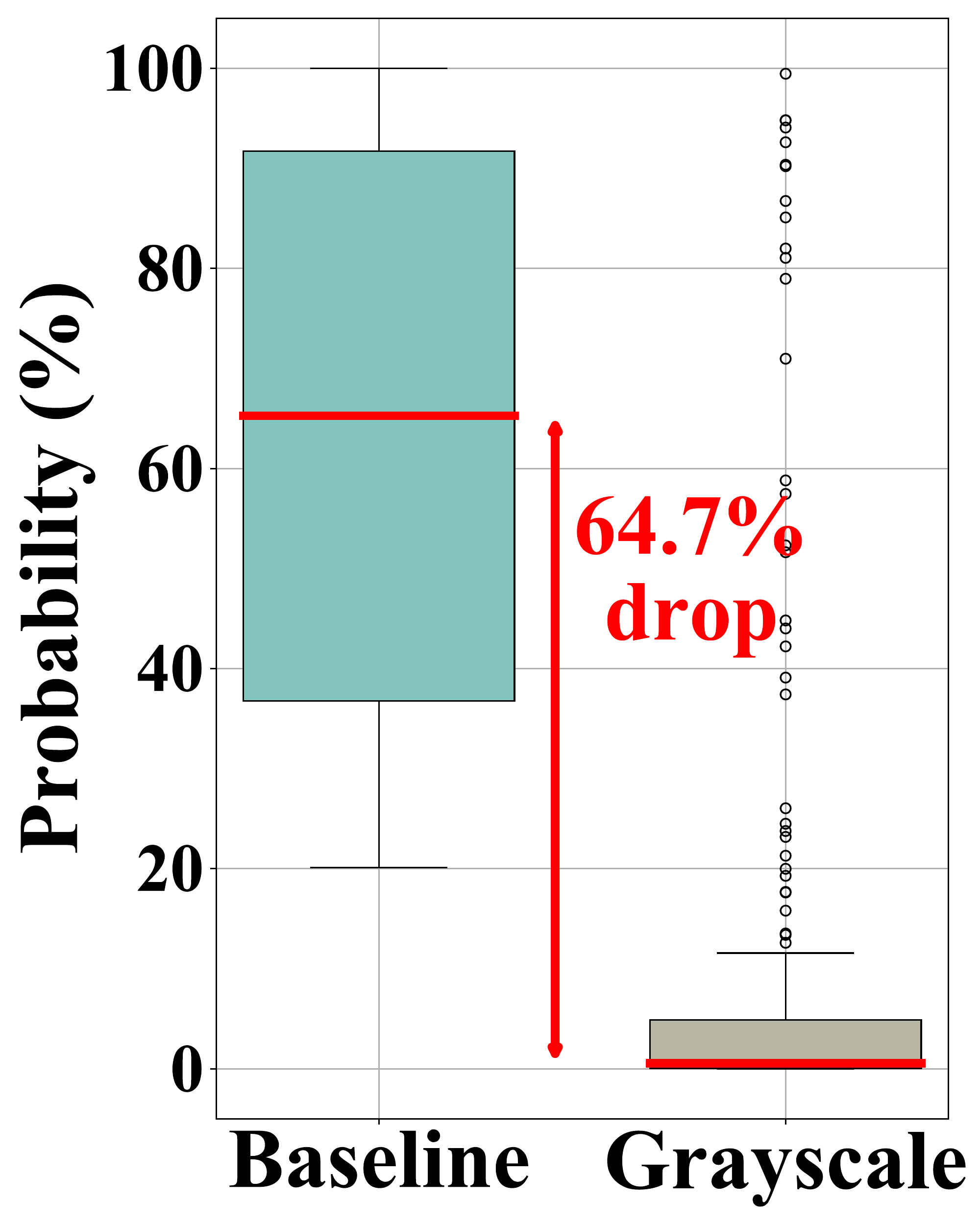} &
    \includegraphics[width=0.13\linewidth]{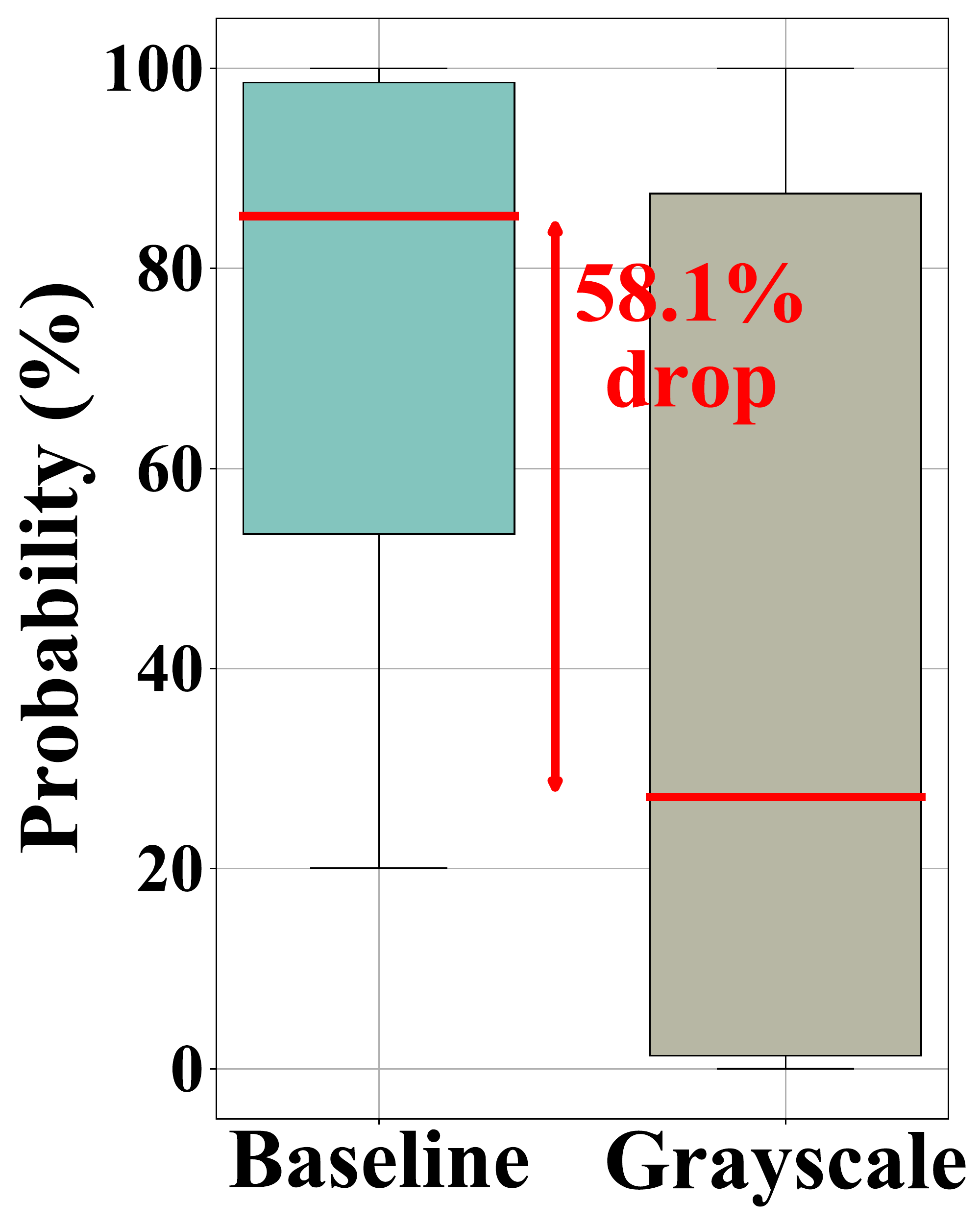} &
    \includegraphics[width=0.13\linewidth]{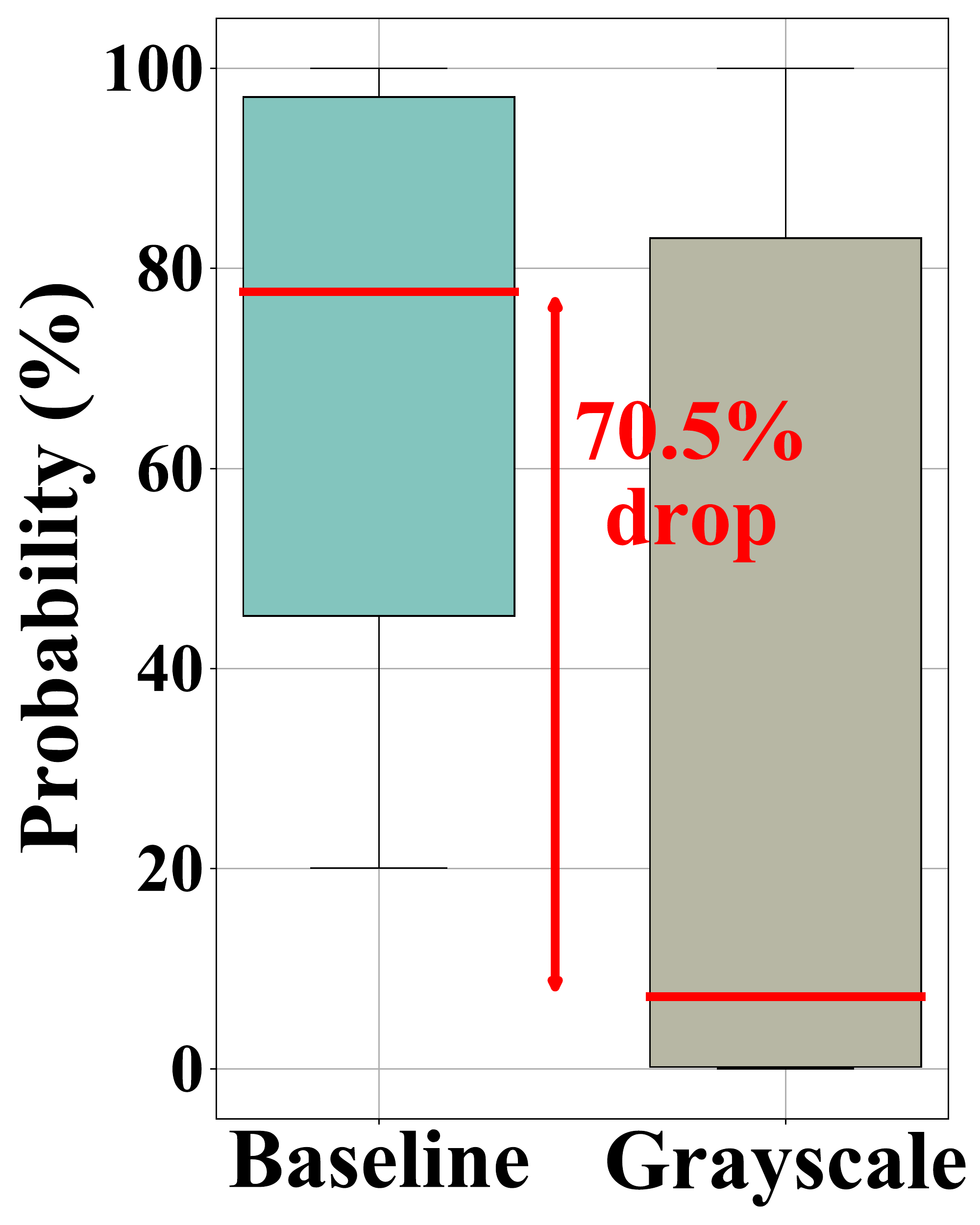} &
    \includegraphics[width=0.13\linewidth]{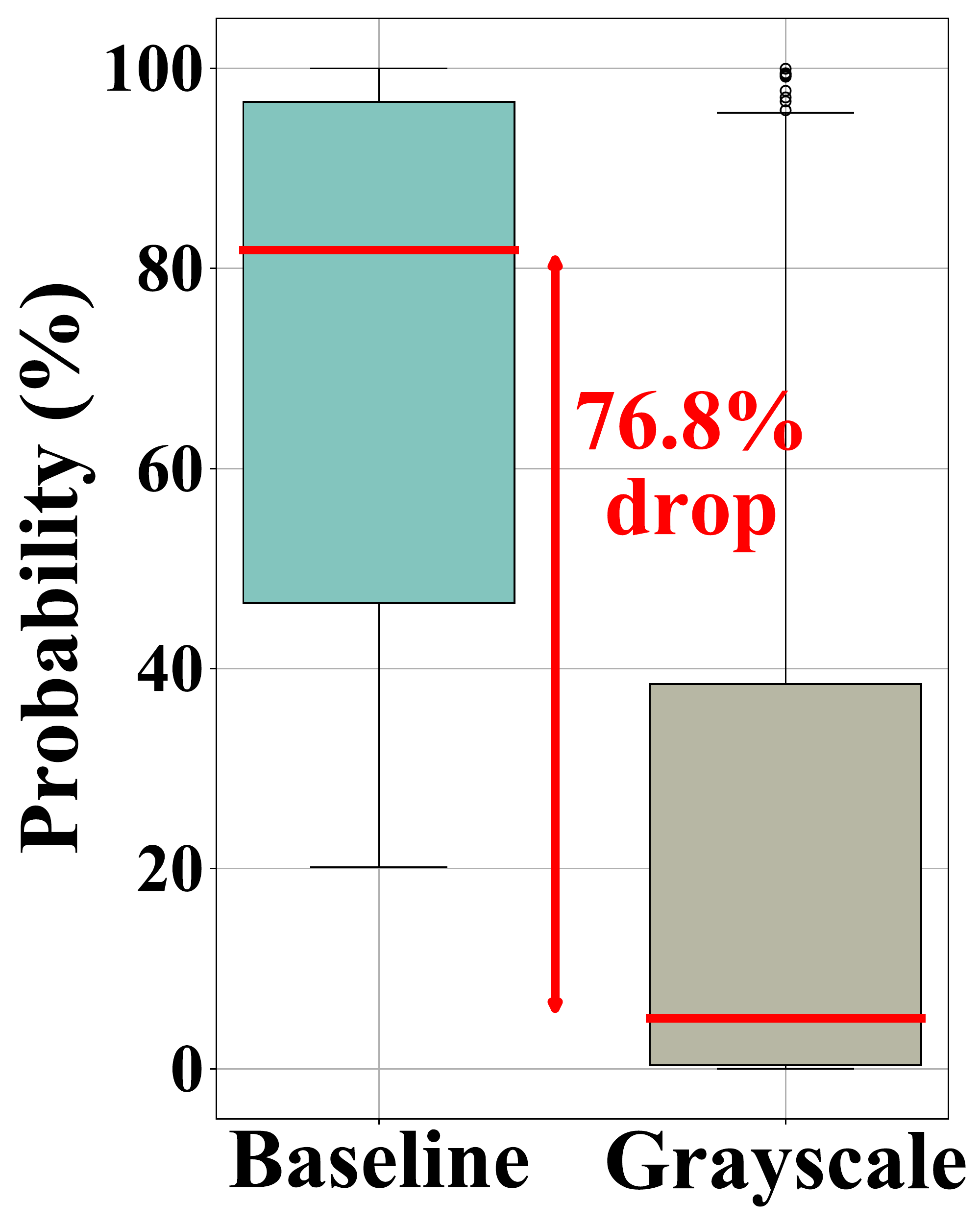} &
    \includegraphics[width=0.13\linewidth]{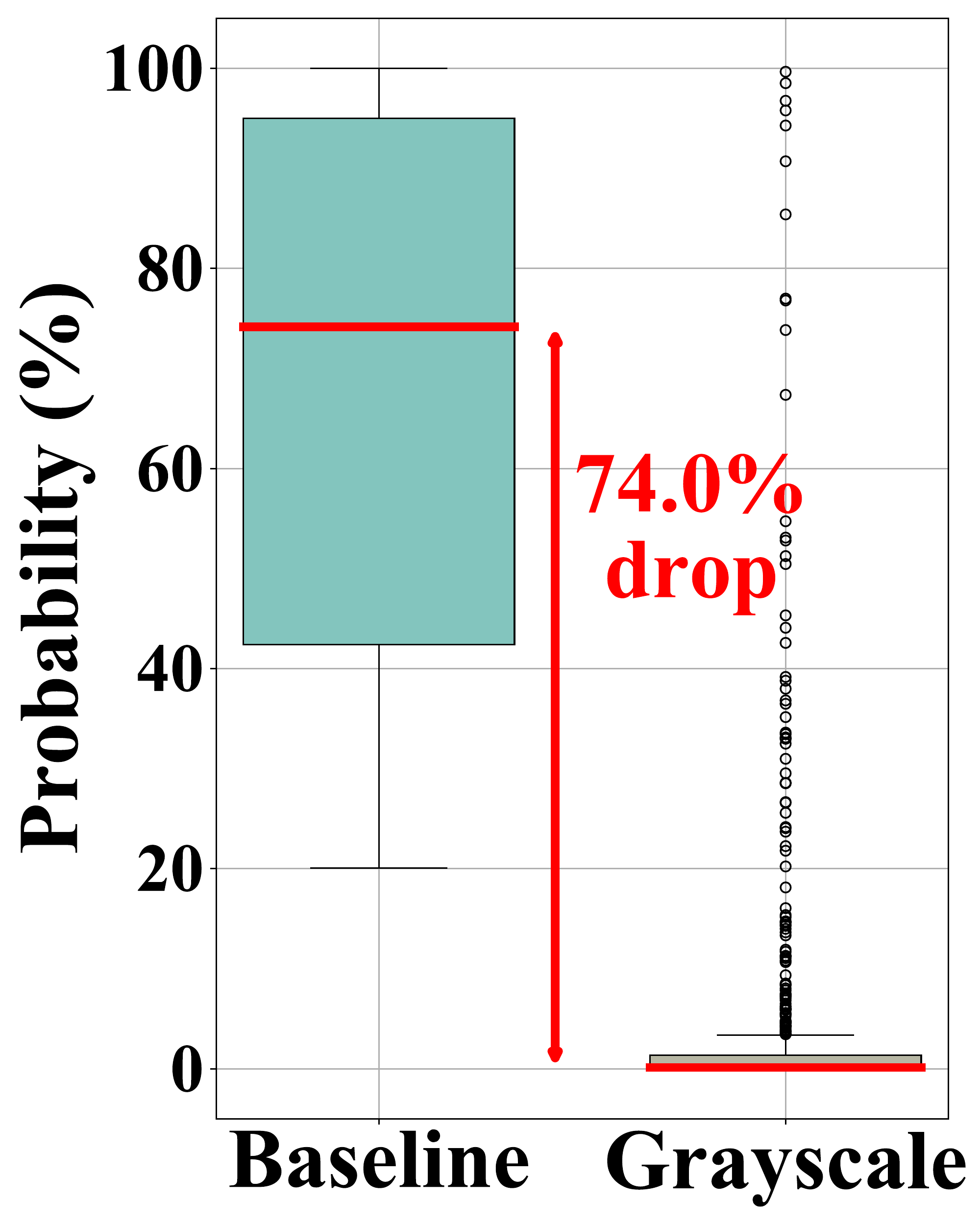} &
    \includegraphics[width=0.13\linewidth]{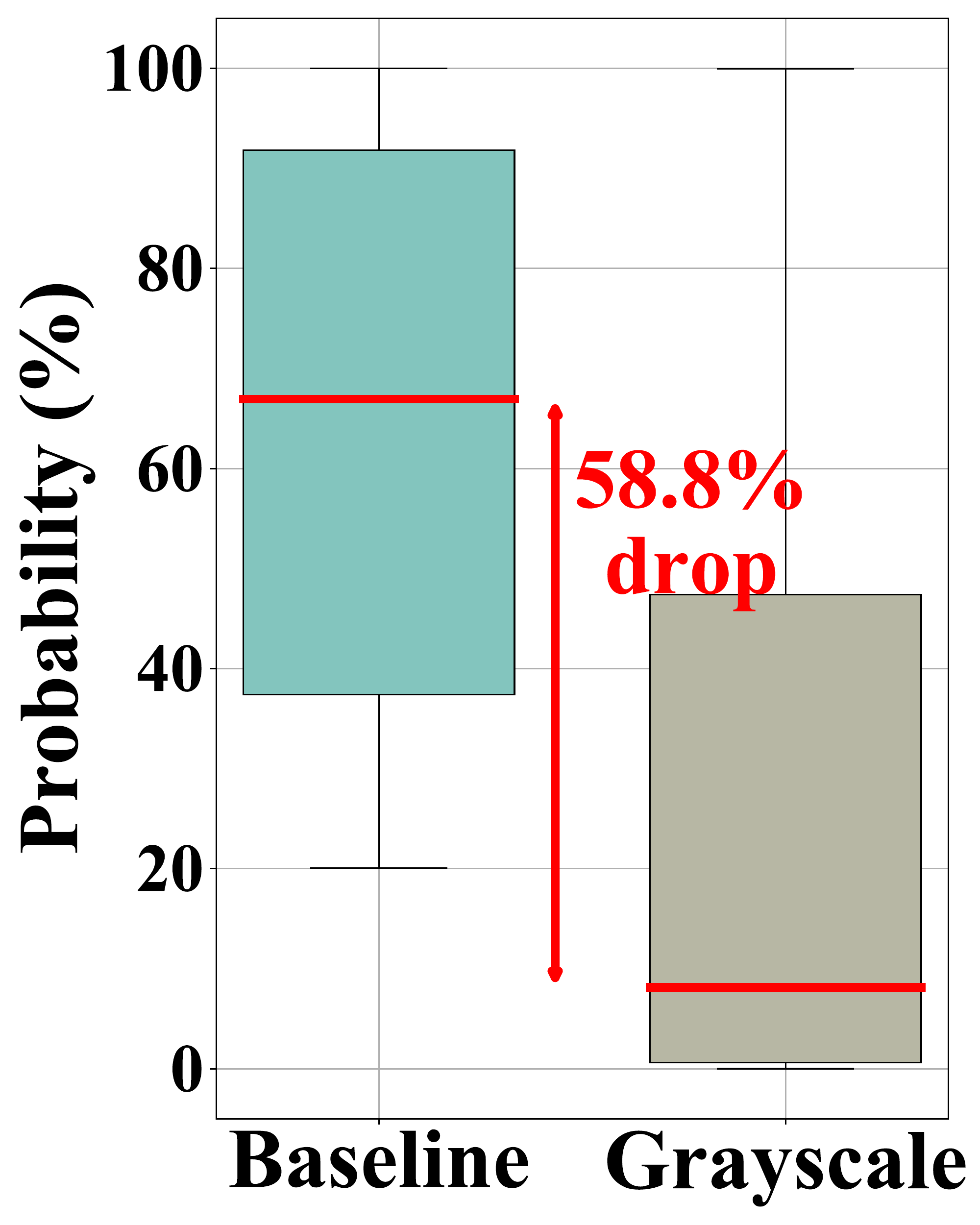}
    \\

\end{tabular}
\caption{
\textit{Color} is a critical \textit{T-FF} in \textit{universal detectors} (Shown using BigGAN \cite{brock2018large} pre-training dataset):
We show the box-whisker plots of probability (\%) predicted by the universal detector for counterfeits before (Baseline) and after \textit{color ablation} (Grayscale) for BigGAN \cite{brock2018large}, ProGAN \cite{karras2018progressive}, StyleGAN2 \cite{Karras_2020_CVPR}, StyleGAN \cite{Karras_2019_CVPR}, CycleGAN \cite{zhu2017unpaired}, StarGAN \cite{choi2018stargan} and GauGAN \cite{park2019semantic}. The red line in each box-plot shows the median probability. 
We show the results for the 
EfficientNet-B0
universal detector following the exact training / test strategy proposed in
\cite{Wang_2020_CVPR}.
Using BigGAN real / fake dataset we verify that Color is a critical \textit{T-FF} in Universal Detectors. We show that color ablation results in
median probability for counterfeits drop by $>$ 58\% across all unseen GANs. Do note that median probability does not drop significantly for BigGAN during color ablation showing the importance
of color for cross-model forensic transfer.
}
\label{fig_main:biggan_median_color_ablation}
\end{figure}

\section{Is the performance degradation in universal detectors due to unseen corruptions (OOD)?}
\label{sec_supp:performance_degrade}
We remark that some performance degrade is
due to CNNs’ poor generalization to unseen corruptions / OOD (grayscale), but here we show that significant amount of degradation is due to color being a critical transferable forensic feature (T-FF) in the universal detector, therefore
ablation of color (i.e., grayscale) leads to significant performance degrade. 
Specifically, we perform an experiment
using official EfficientNet-B0 ImageNet classifier (architecture identical to our universal detector) under Grayscale (OOD) setup. 
We measure the median probability of the
correct class before and after Grayscale (OOD) and observe only 17\% drop due to Grayscale. 
Comparing the within-model OOD setup with the cross-model setup, the median probability drop during cross- model forensic transfer is
much larger, i.e.: median probability drop during cross-model forensic transfer is $>$ 89\% (ProGAN pre-training, Fig. \ref{fig_main:median_color_ablation}) and $>$ 58\% (BigGAN pre-training, Fig. \ref{fig_main:biggan_median_color_ablation}) for EfficientNet-B0
universal detector. 
This shows that color is critical in forensic transfer compared to within-model OOD setups. 
See row 1, col 1 in Fig. \ref{fig_main:median_color_ablation} and Fig. \ref{fig_main:biggan_median_color_ablation}, col 1 to verify that the median probability
does not drop much for the GAN used to train universal detectors under Grayscale (OOD).

\section{Color-conditional \textit{T-FF} (Additional Results)}
\label{sec_supp:color_tff}
\setcounter{figure}{0} 
\setcounter{table}{0} 

\begin{figure}[!t]
\centering
\begin{tabular}{ccccccc}
    \multicolumn{1}{p{0.125\linewidth}}{\tiny \enskip ProGAN \cite{karras2018progressive}} &
    \multicolumn{1}{p{0.15\linewidth}}{\tiny  \enskip StyleGAN2 \cite{Karras_2020_CVPR}} &
    \multicolumn{1}{p{0.14\linewidth}}{\tiny StyleGAN \cite{Karras_2019_CVPR}} &
    \multicolumn{1}{p{0.125\linewidth}}{\tiny BigGAN \cite{brock2018large}} &
    \multicolumn{1}{p{0.132\linewidth}}{\tiny CycleGAN \cite{zhu2017unpaired}} &
    \multicolumn{1}{p{0.135\linewidth}}{\tiny \enskip StarGAN \cite{choi2018stargan}} &
    {\tiny GauGAN \cite{park2019semantic}} \\

    \multicolumn{7}{c}{\includegraphics[width=0.99\linewidth]{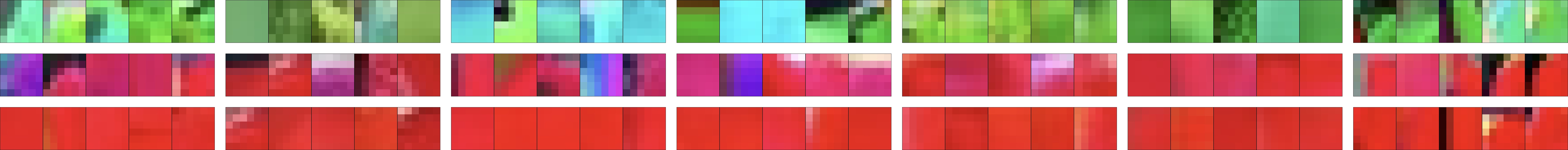}}
    
\end{tabular}
\caption{
Additional results demonstrating that color is a critical \textit{transferable forensic feature (T-FF)} in universal detectors (ResNet-50):
Large-scale study on visual interpretability of \textit{T-FF} discovered through our proposed \textit{forensic feature relevance statistic (FF-RS)
,} reveal that color information is critical for cross-model forensic transfer.
Each row represents a color-conditional \textit{T-FF} and
we show the LRP-max response regions for ProGAN \cite{karras2018progressive}, StyleGAN2 \cite{Karras_2020_CVPR}, StyleGAN \cite{Karras_2019_CVPR}, BigGAN \cite{brock2018large}, CycleGAN \cite{zhu2017unpaired}, StarGAN \cite{choi2018stargan} and GauGAN \cite{park2019semantic} counterfeits 
for the publicly released ResNet-50 universal detector by Wang \etal \cite{Wang_2020_CVPR}.
This detector is trained with ProGAN
\cite{karras2018progressive} 
counterfeits \cite{Wang_2020_CVPR} and cross-model forensic transfer is evaluated on other unseen GANs.
All counterfeits are obtained from the ForenSynths dataset 
\cite{Wang_2020_CVPR}.
The consistent color-conditional LRP-max response across all GANs for these \textit{T-FF} clearly indicate that \textit{color} is critical for cross-model forensic transfer in universal detectors.
}
\label{fig_supp:lrp_patches_r50}
\end{figure}

\begin{figure}
\centering
 
\begin{tabular}{ccccccc}
    {\tiny ProGAN \cite{karras2018progressive}} &
    {\tiny StyleGAN2 \cite{Karras_2020_CVPR}} &
    {\tiny StyleGAN \cite{Karras_2019_CVPR}} &
    {\tiny BigGAN \cite{brock2018large}} &
    {\tiny CycleGAN \cite{zhu2017unpaired}} &
    {\tiny StarGAN \cite{choi2018stargan}} &
    {\tiny GauGAN \cite{park2019semantic}} \\
    
    \includegraphics[width=0.13\linewidth]{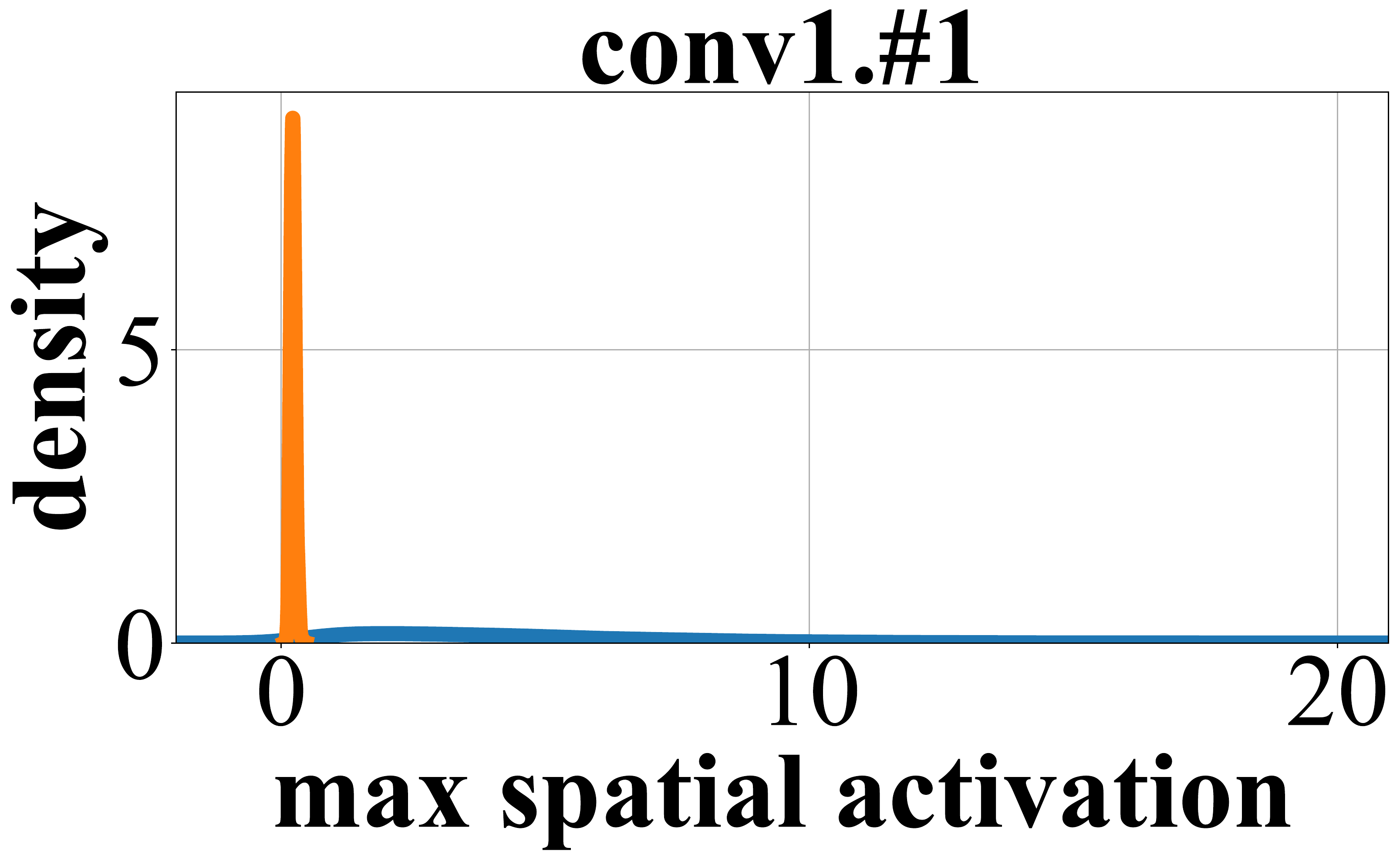} &
    \includegraphics[width=0.13\linewidth]{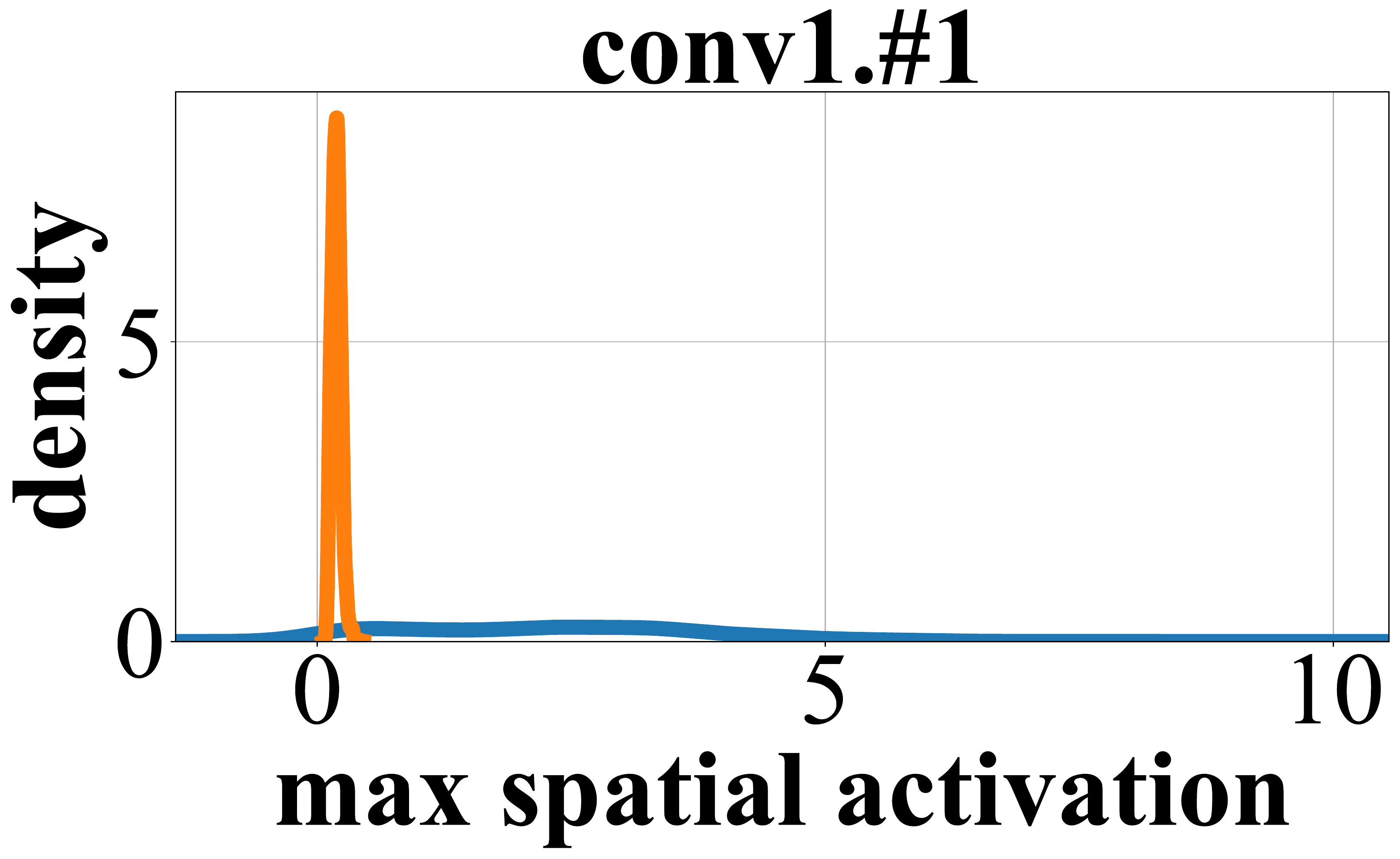} &
    \includegraphics[width=0.13\linewidth]{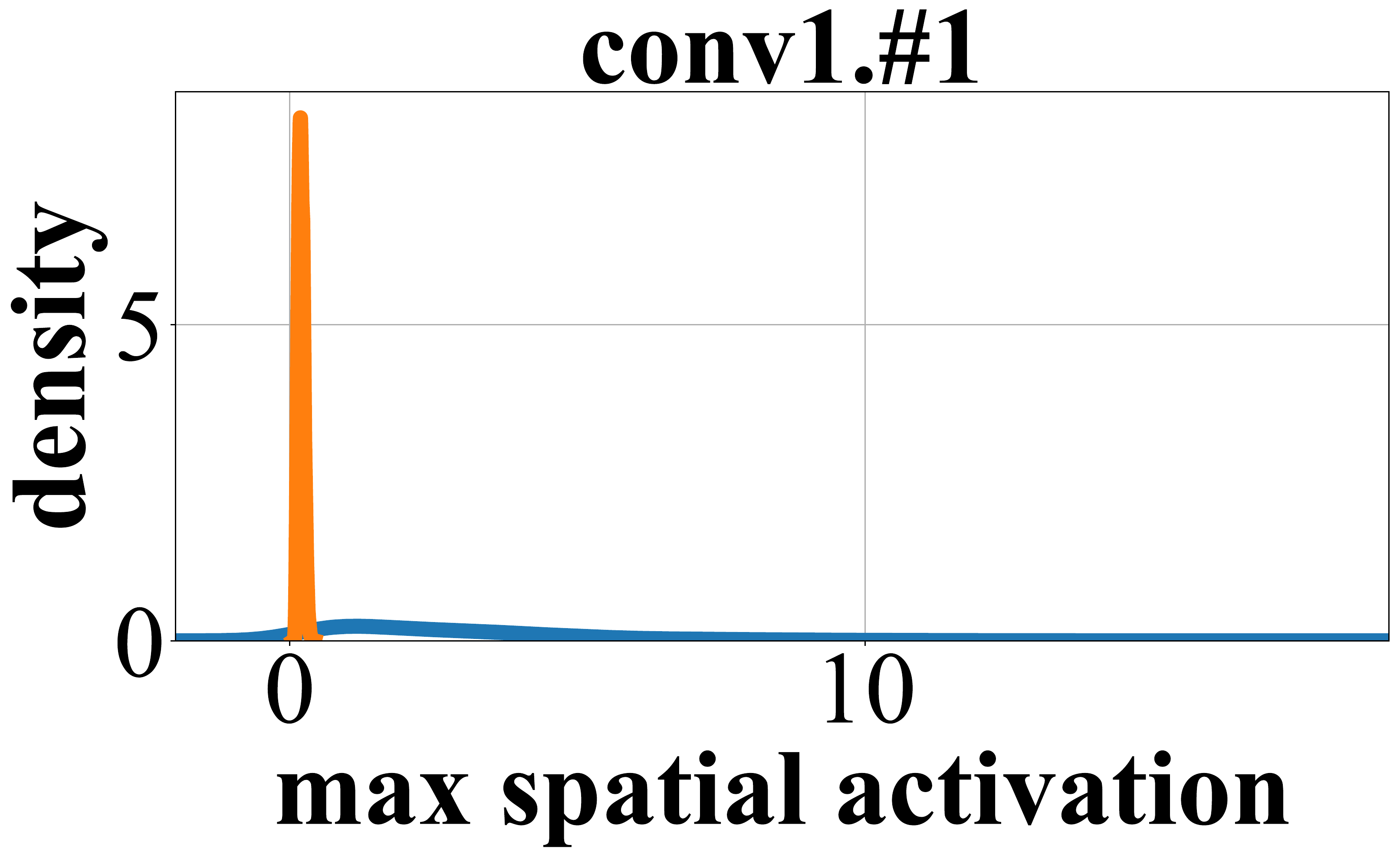} &
     \includegraphics[width=0.13\linewidth]{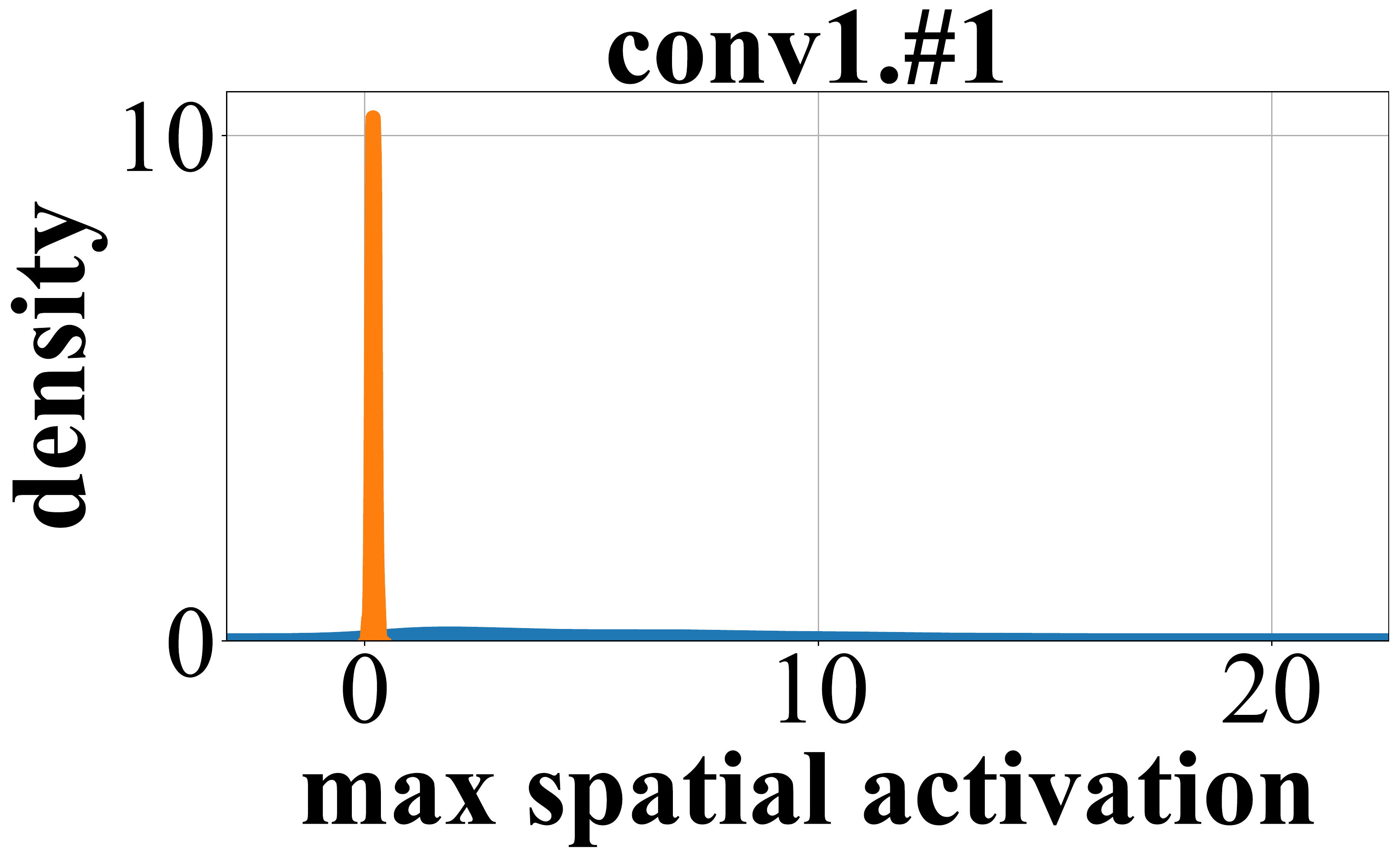} &
    \includegraphics[width=0.13\linewidth]{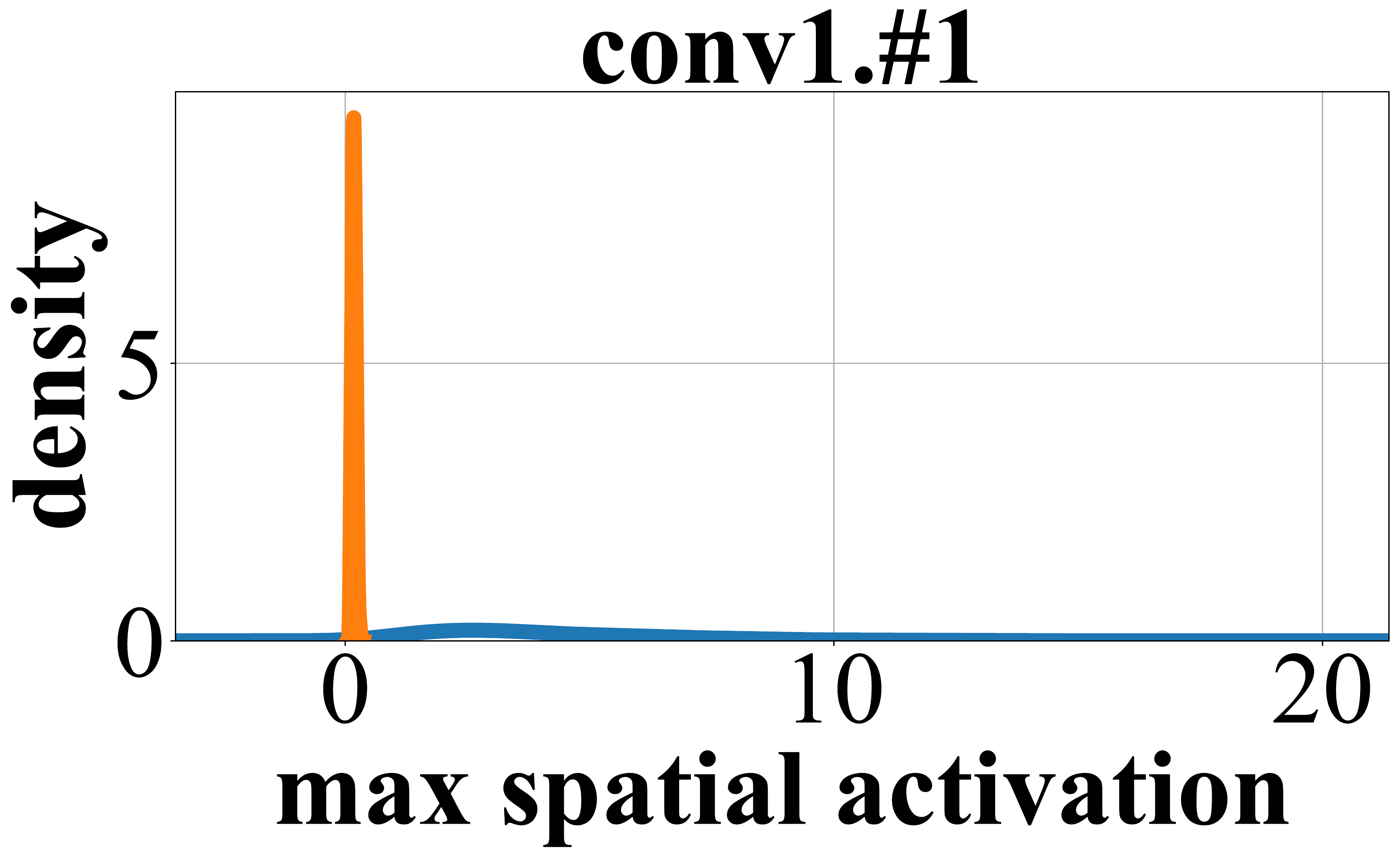} &
     \includegraphics[width=0.13\linewidth]{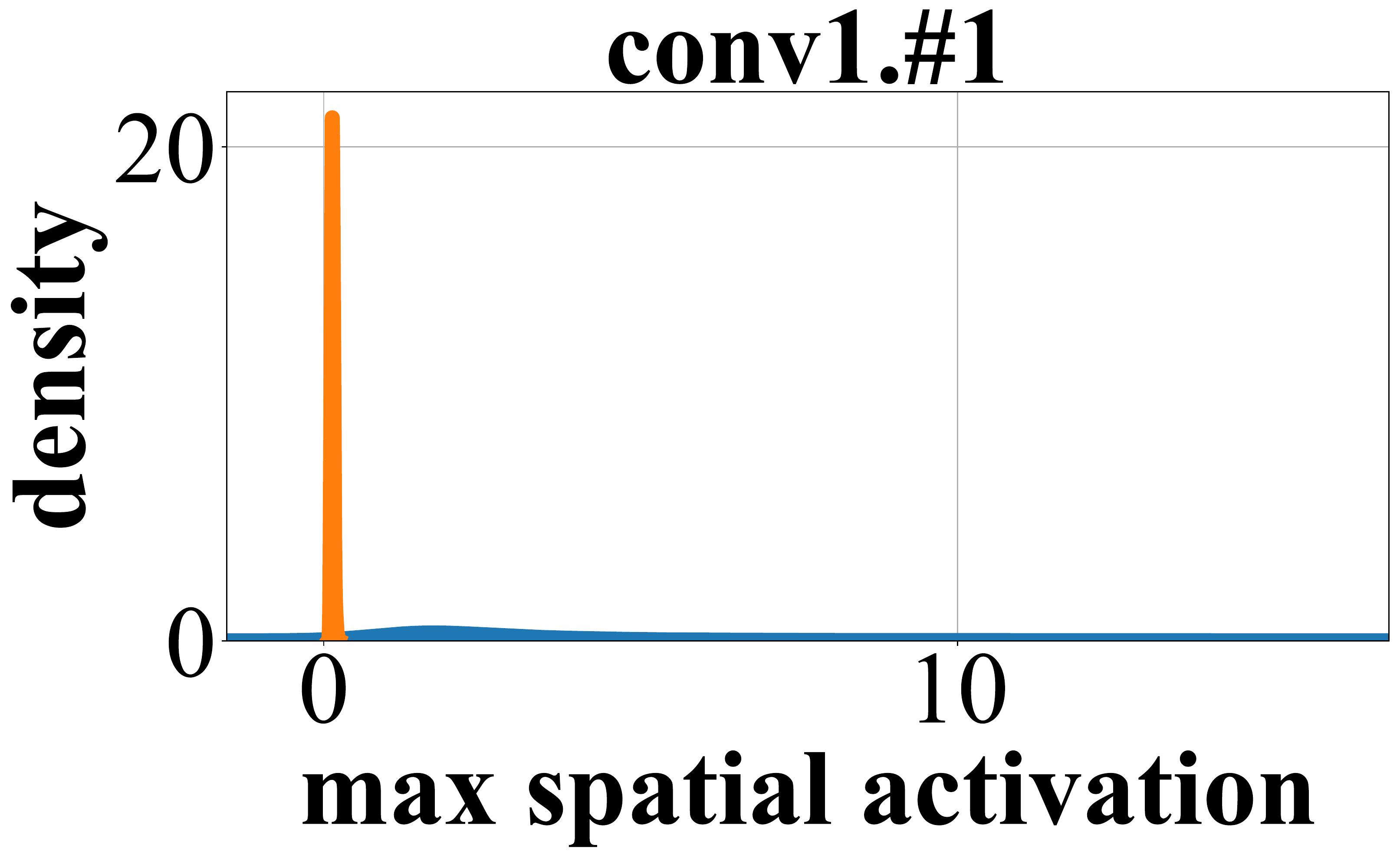} &
     \includegraphics[width=0.13\linewidth]{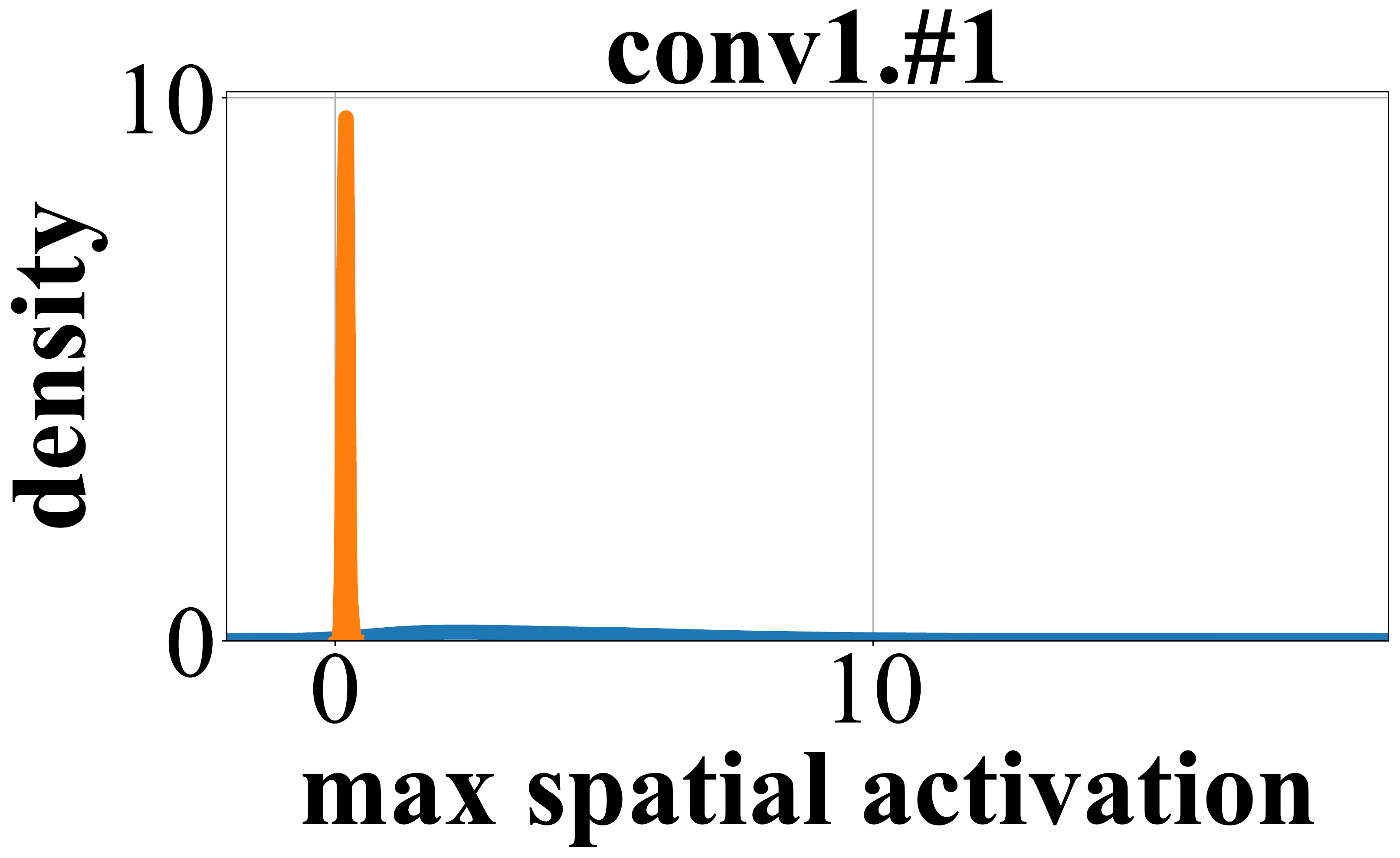}
    \\
    
    \includegraphics[width=0.13\linewidth]{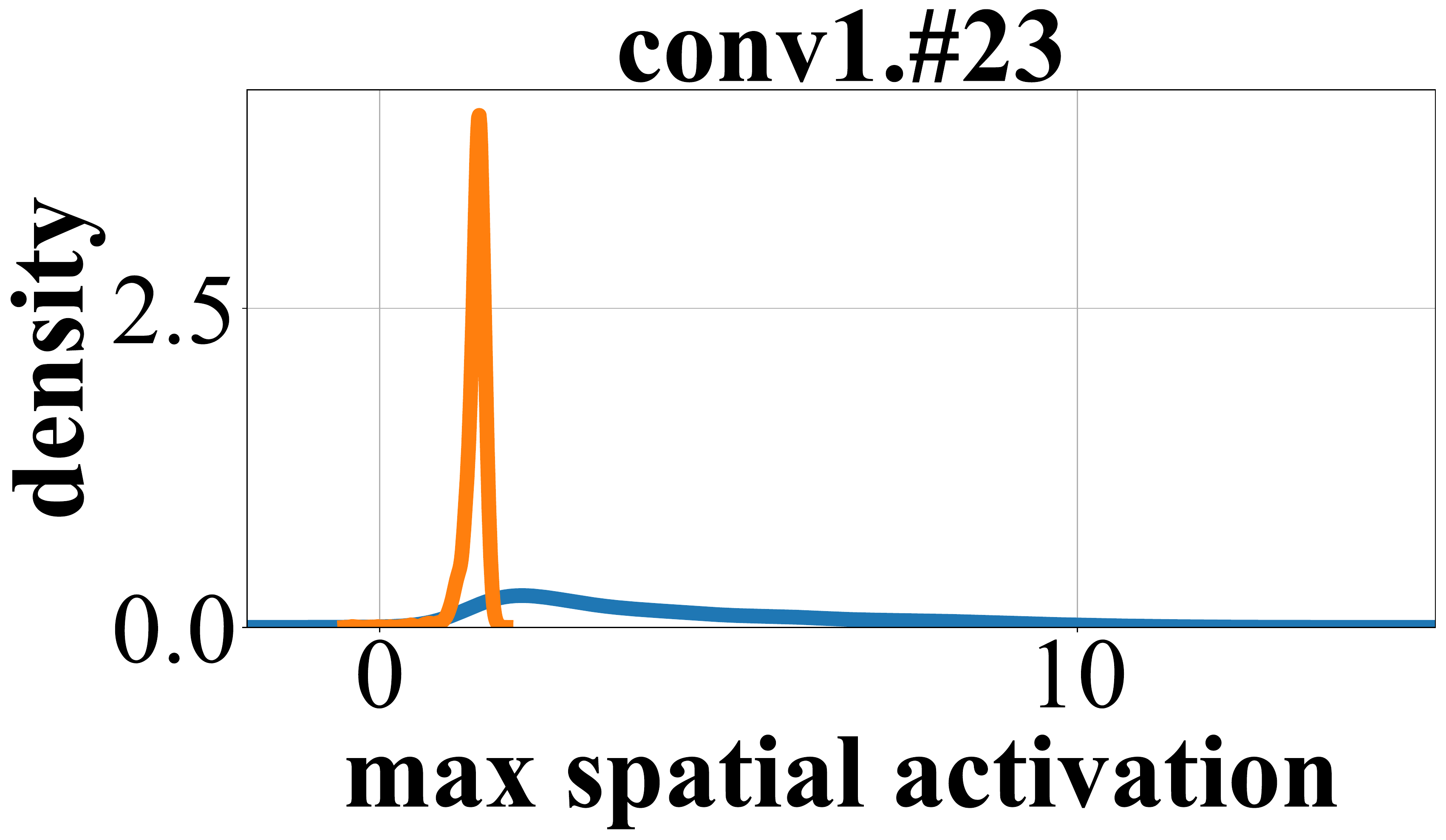} &
    \includegraphics[width=0.13\linewidth]{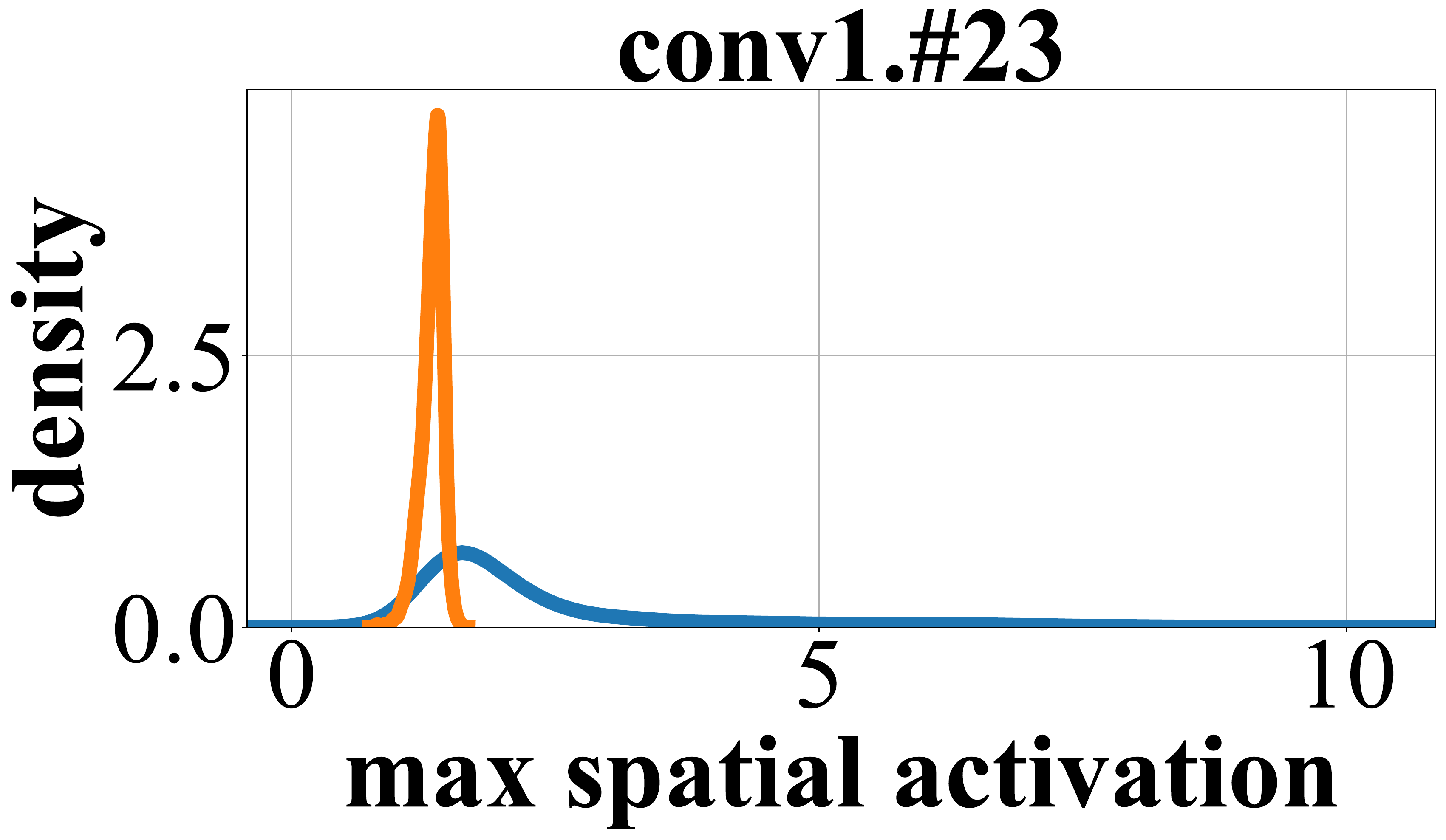} &
    \includegraphics[width=0.13\linewidth]{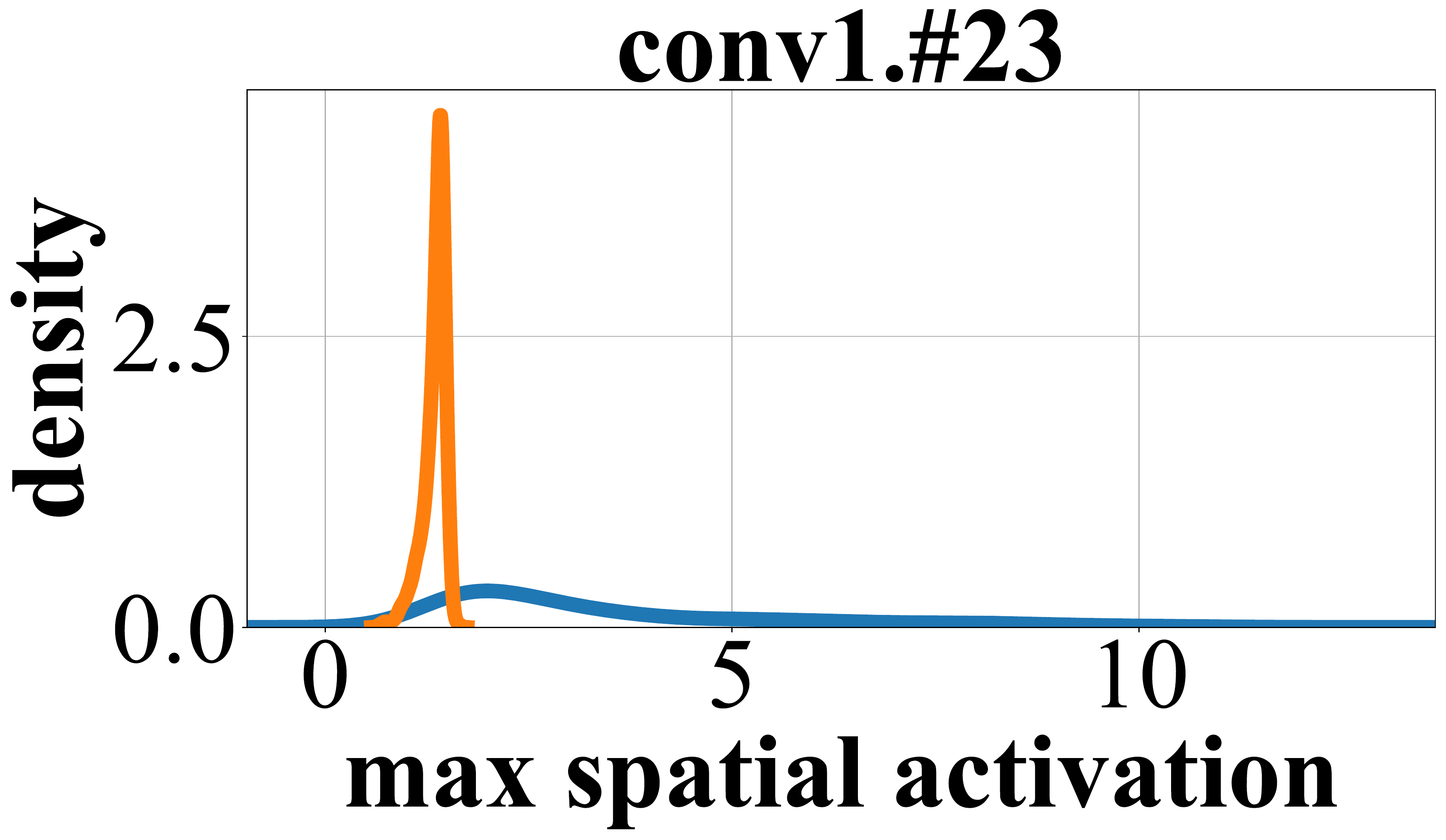} &
     \includegraphics[width=0.13\linewidth]{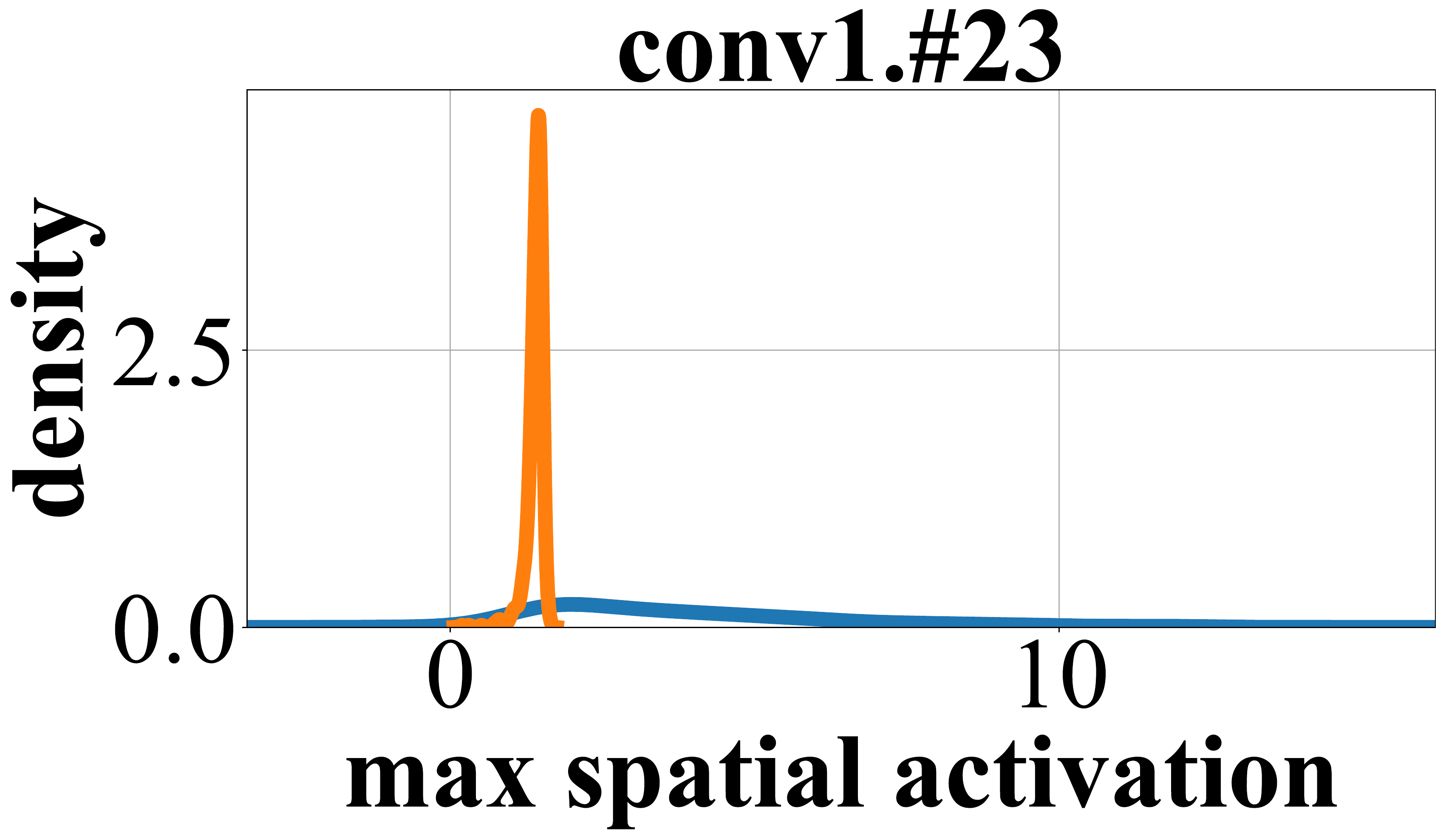} &
    \includegraphics[width=0.13\linewidth]{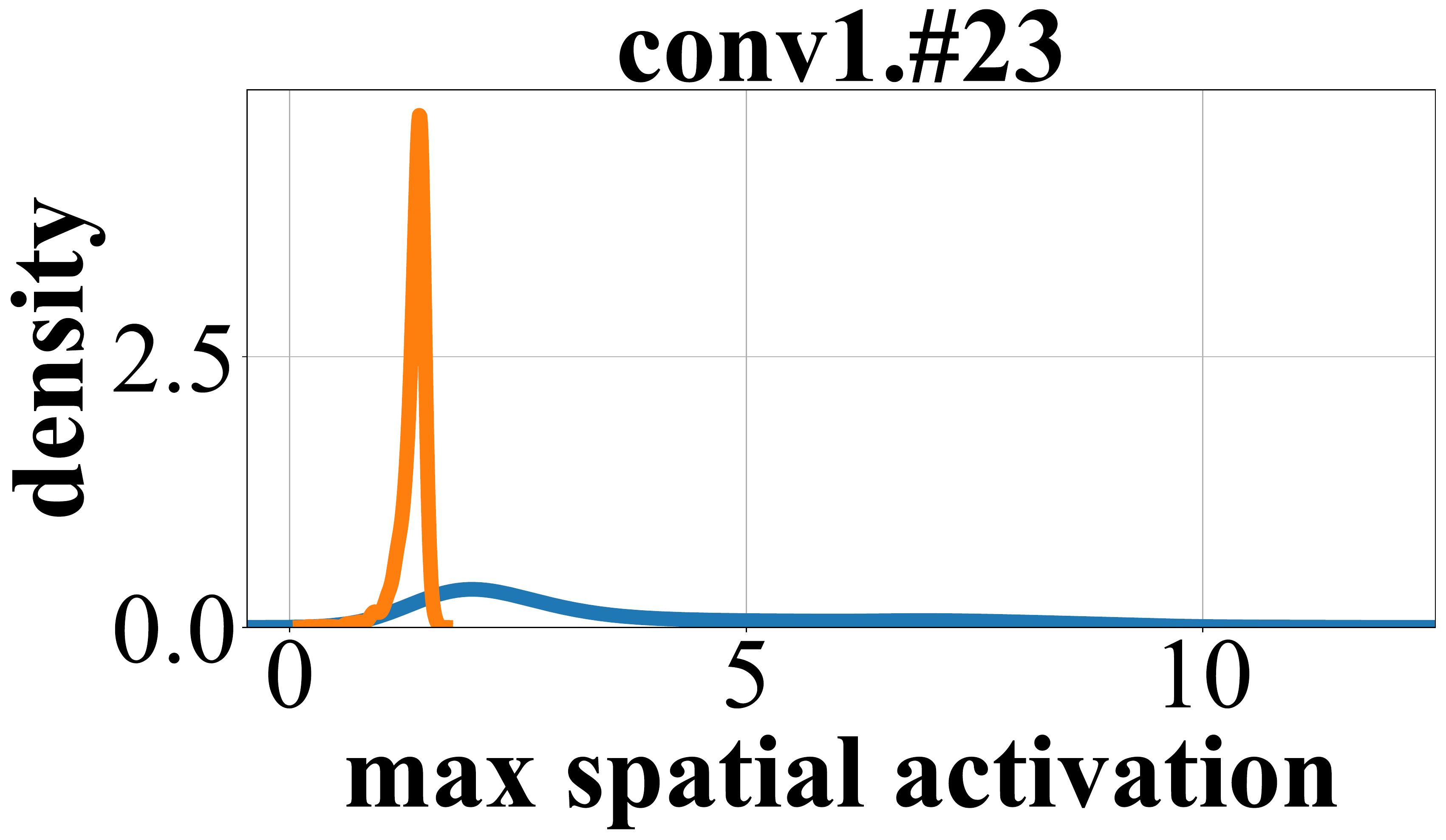} &
     \includegraphics[width=0.13\linewidth]{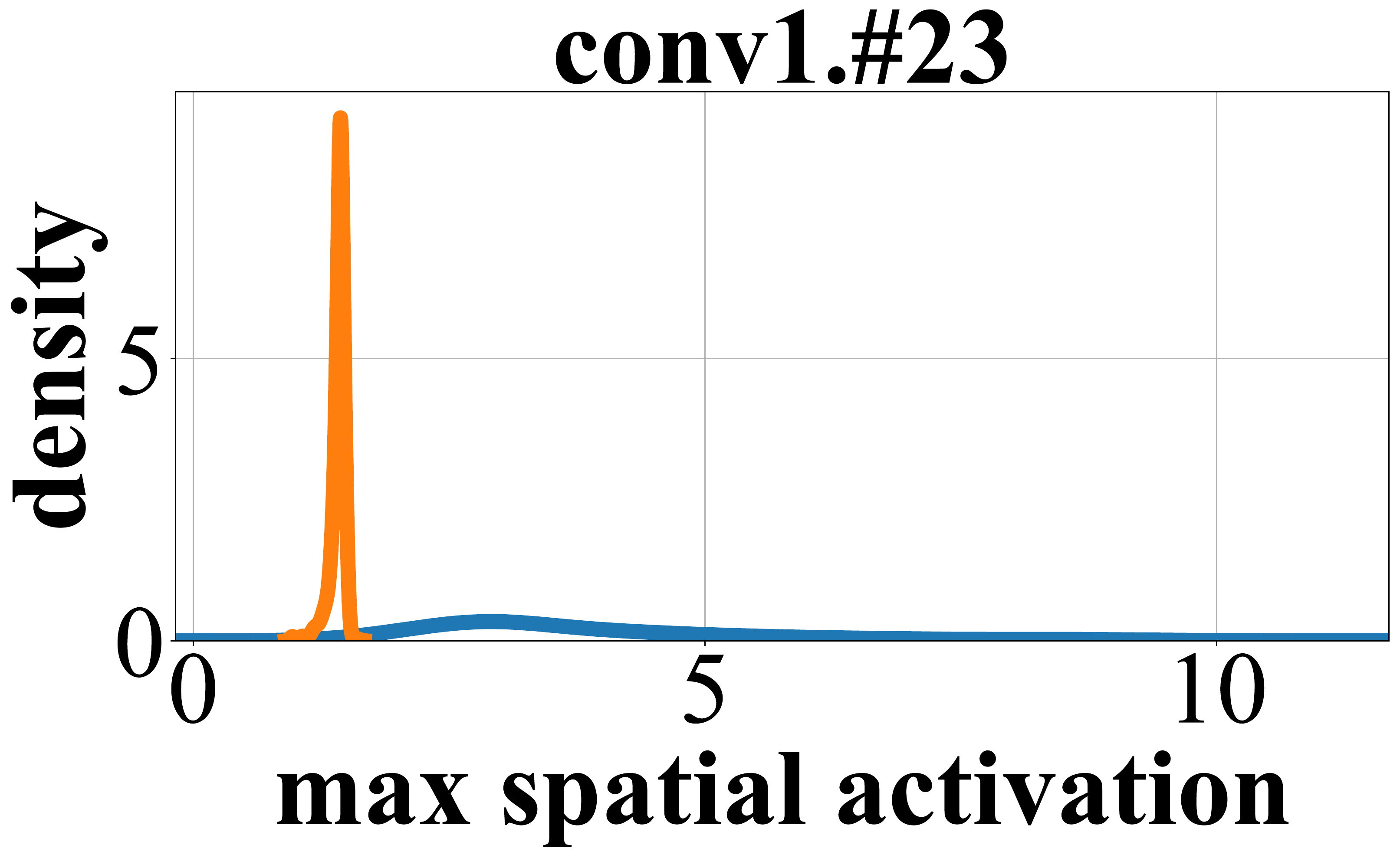} &
     \includegraphics[width=0.13\linewidth]{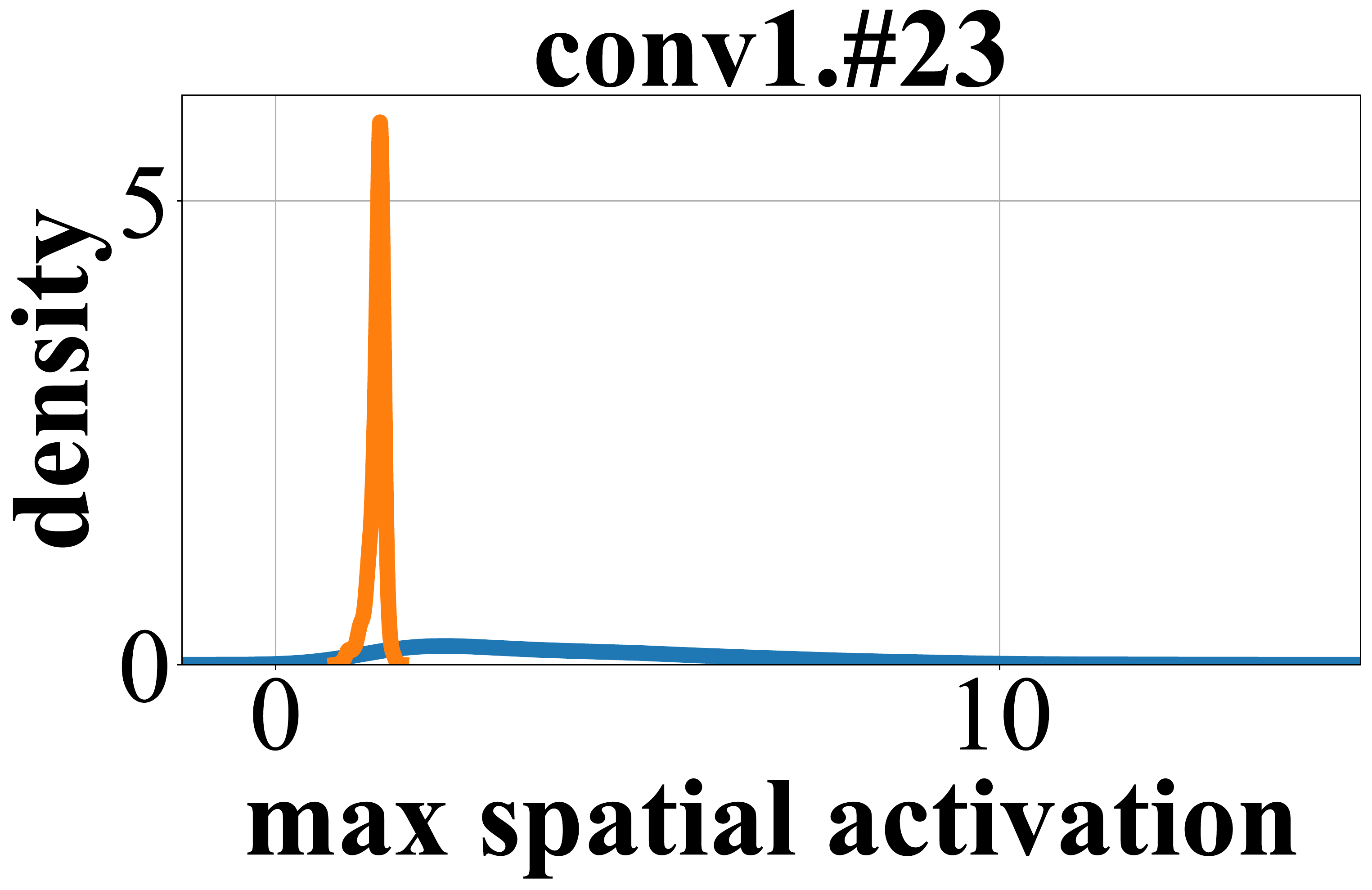}
    \\
    
     \includegraphics[width=0.13\linewidth]{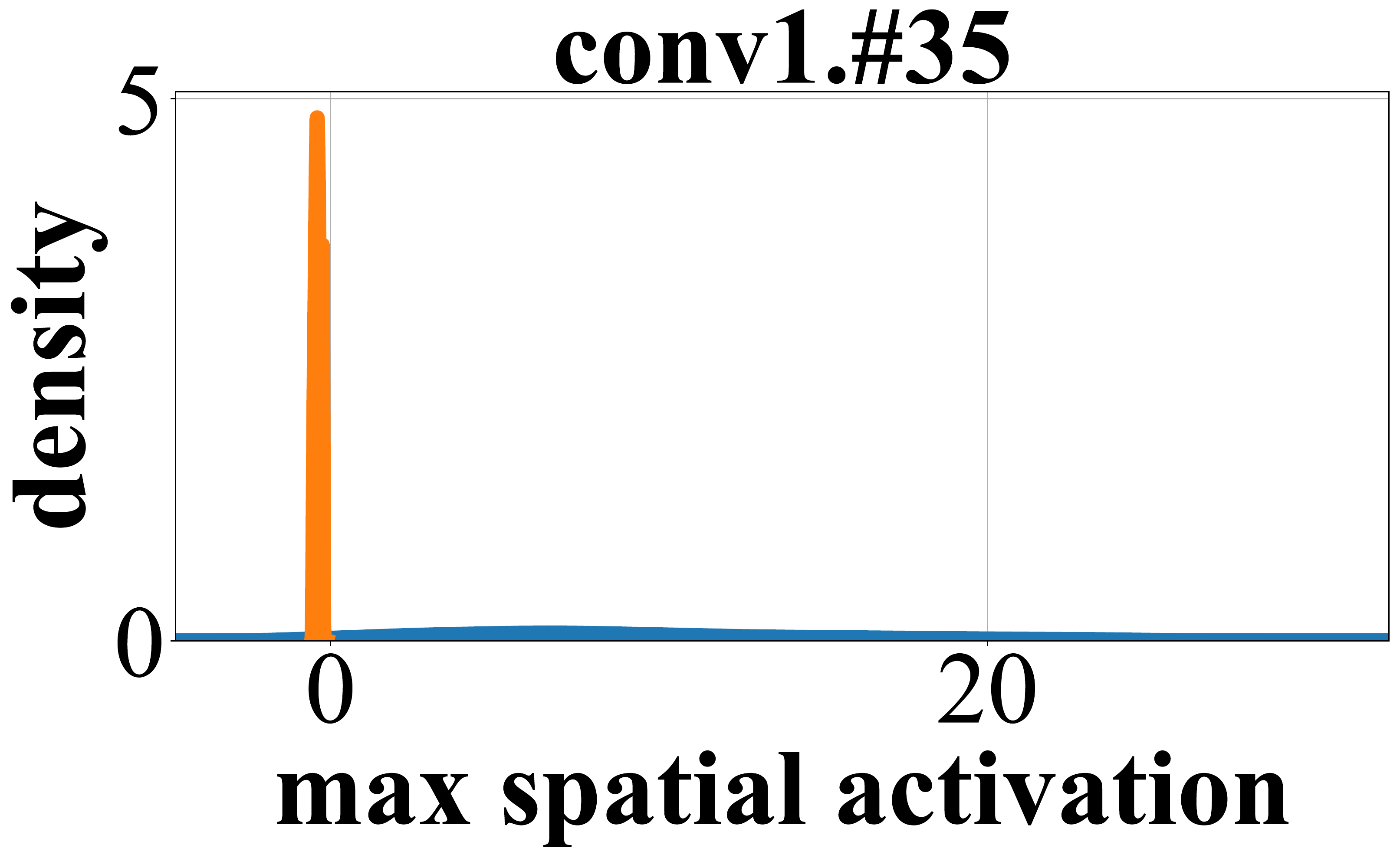} &
    \includegraphics[width=0.13\linewidth]{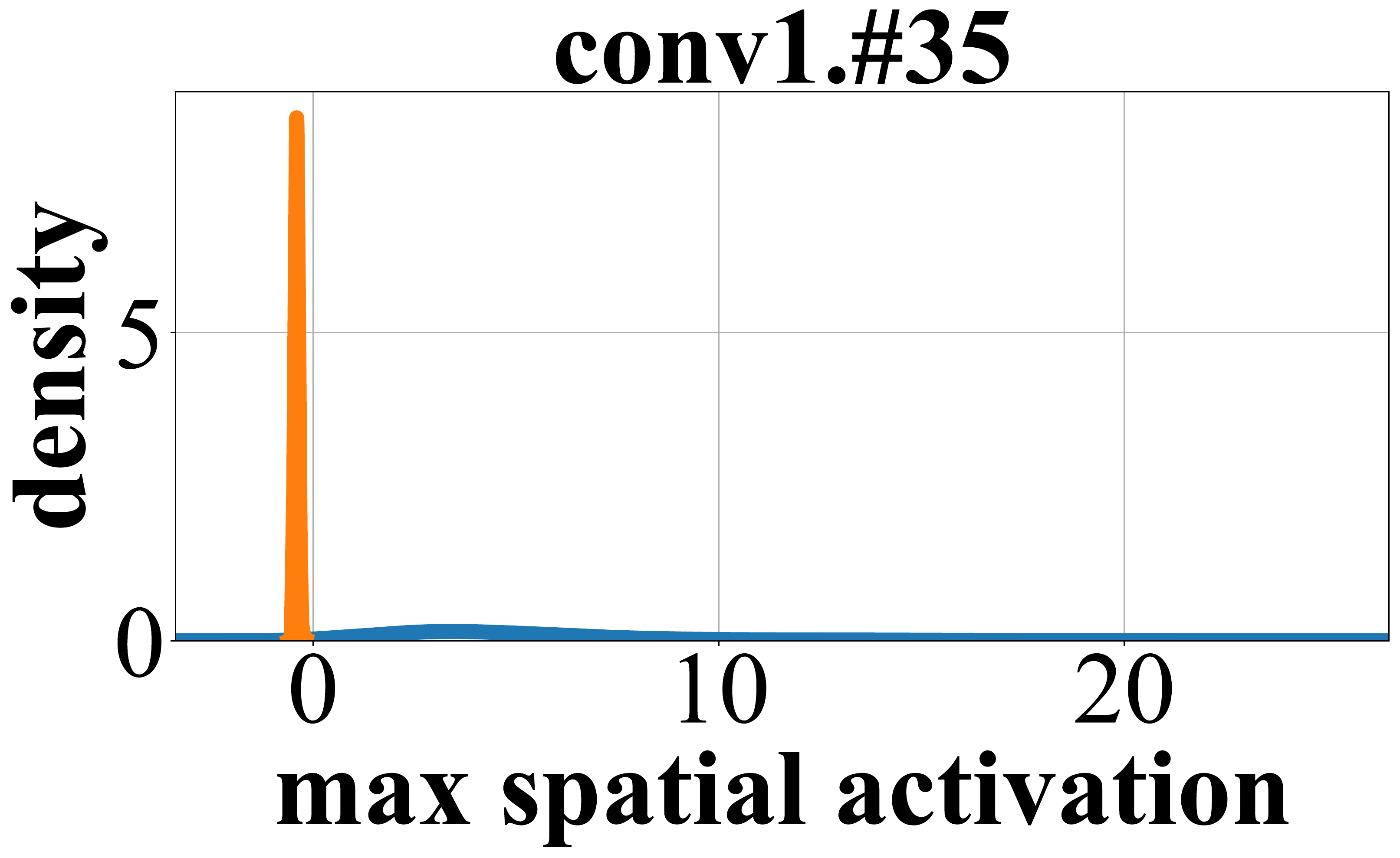} &
    \includegraphics[width=0.13\linewidth]{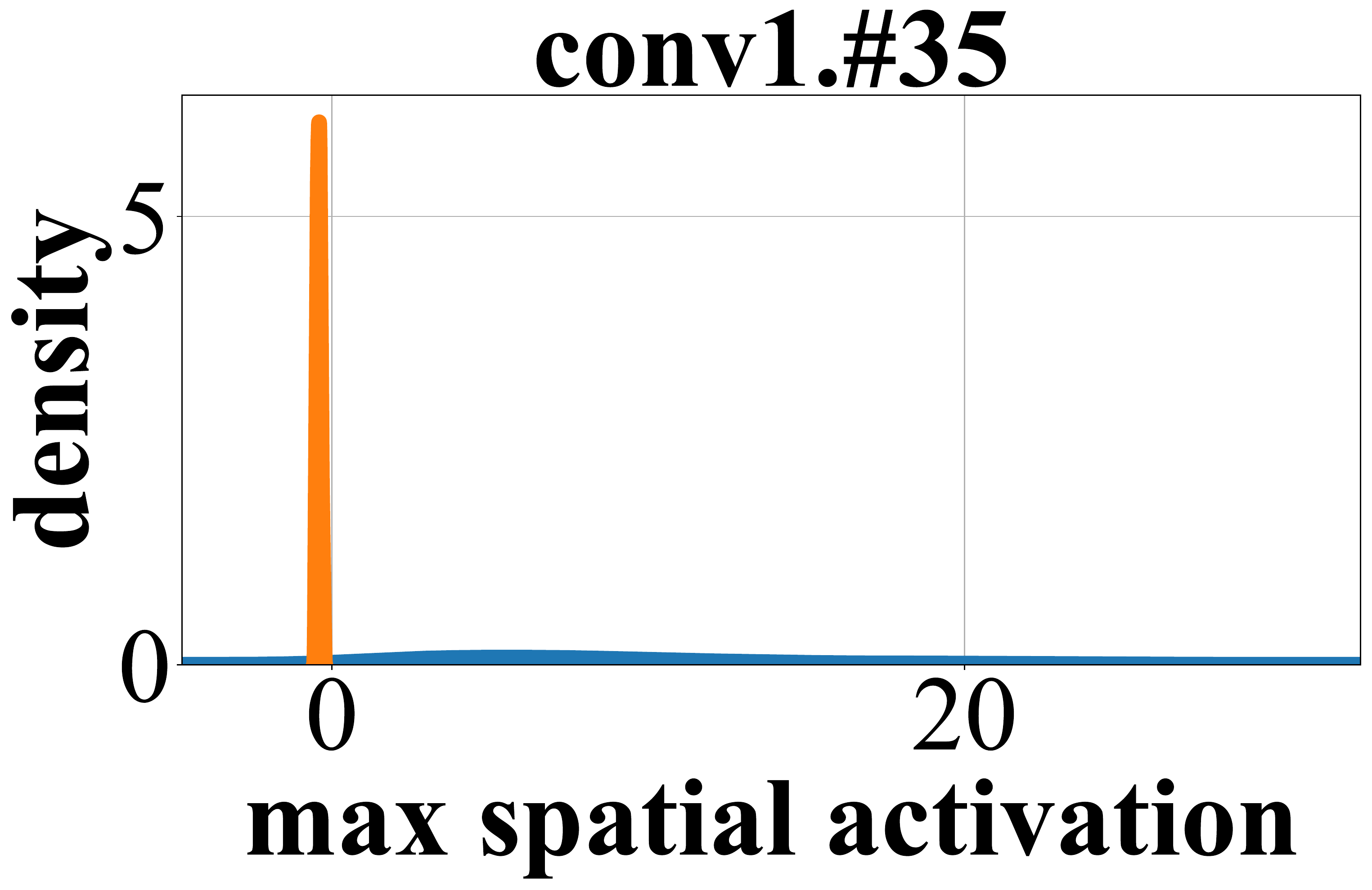} &
     \includegraphics[width=0.13\linewidth]{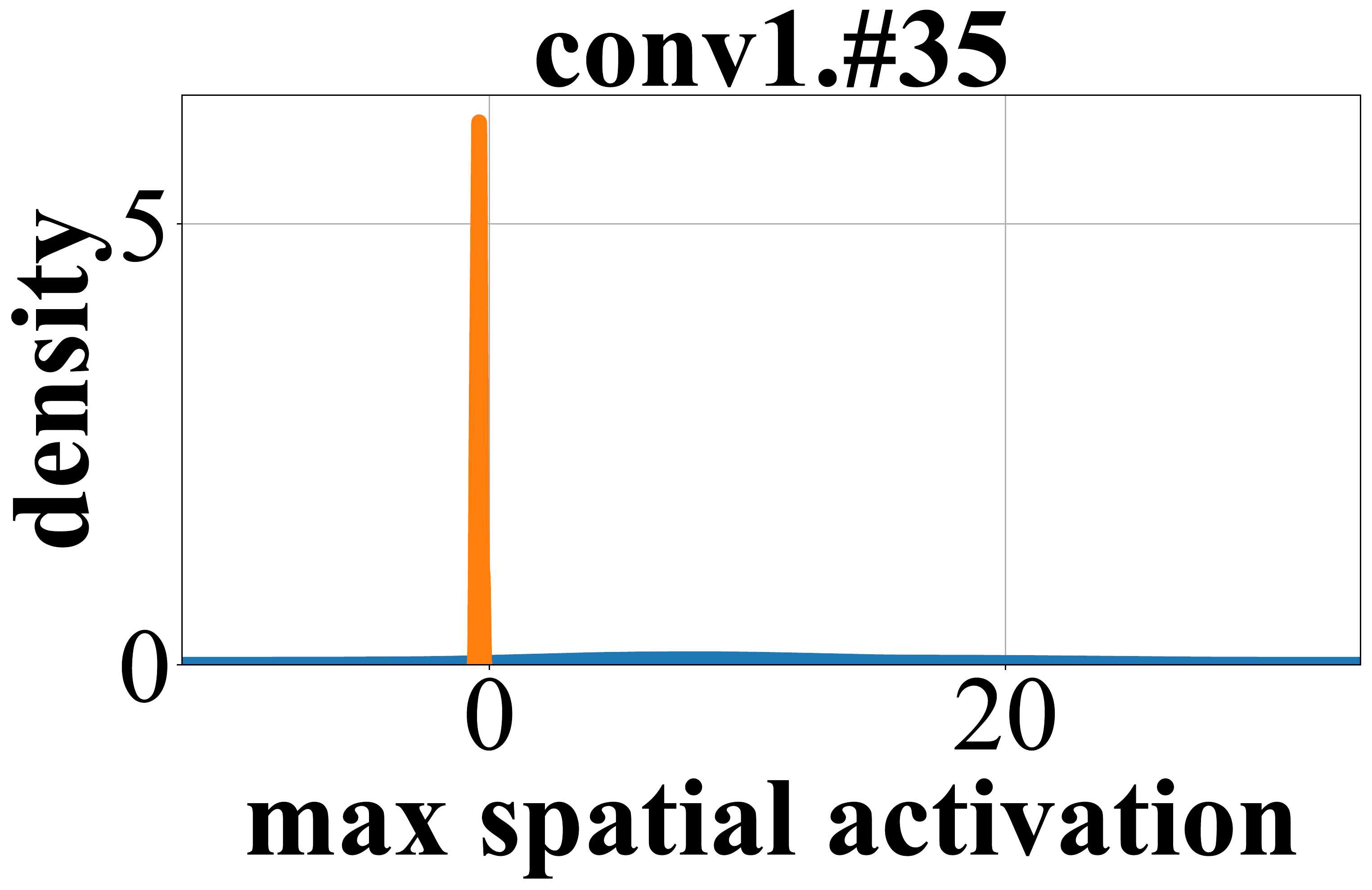} &
    \includegraphics[width=0.13\linewidth]{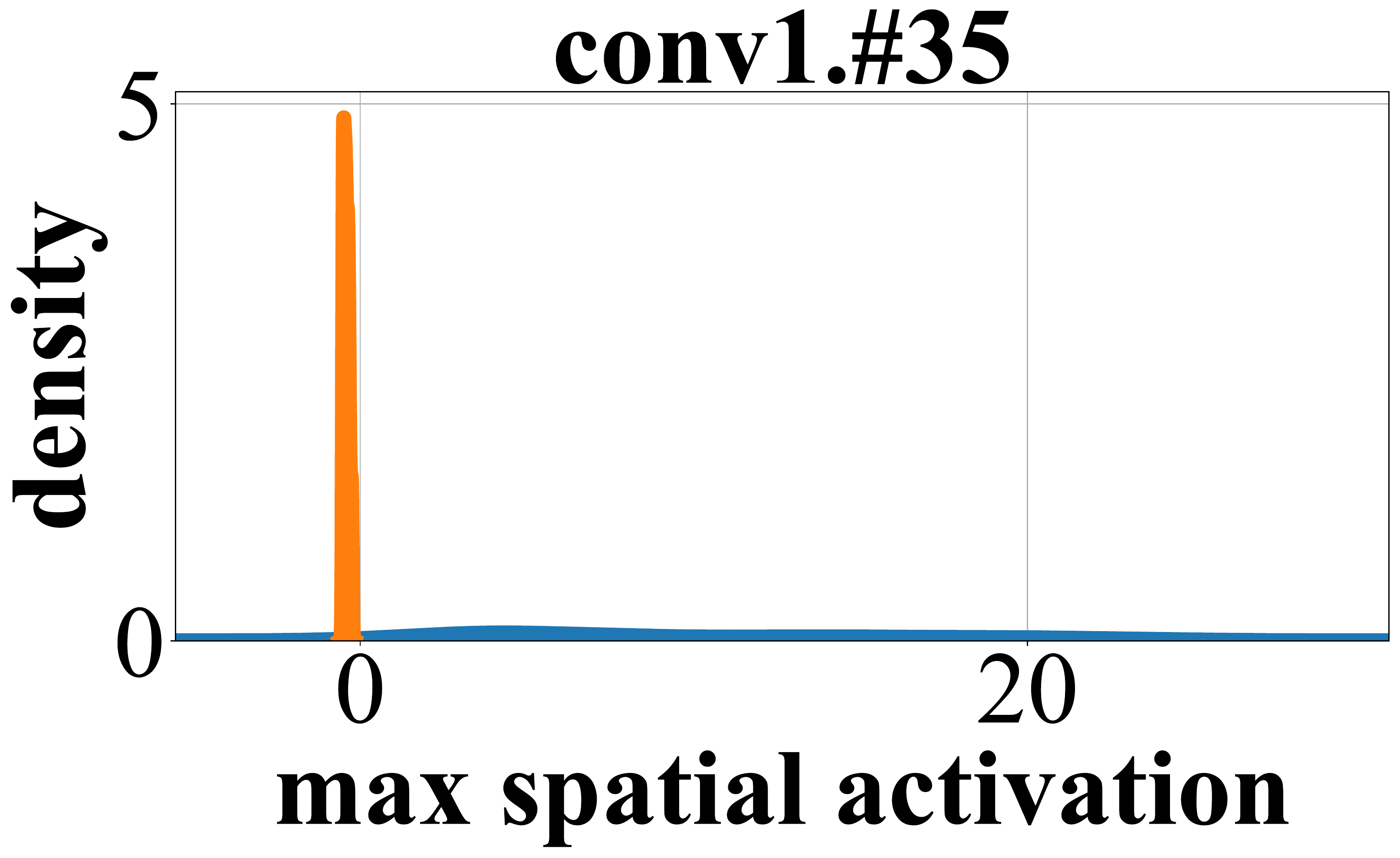} &
     \includegraphics[width=0.13\linewidth]{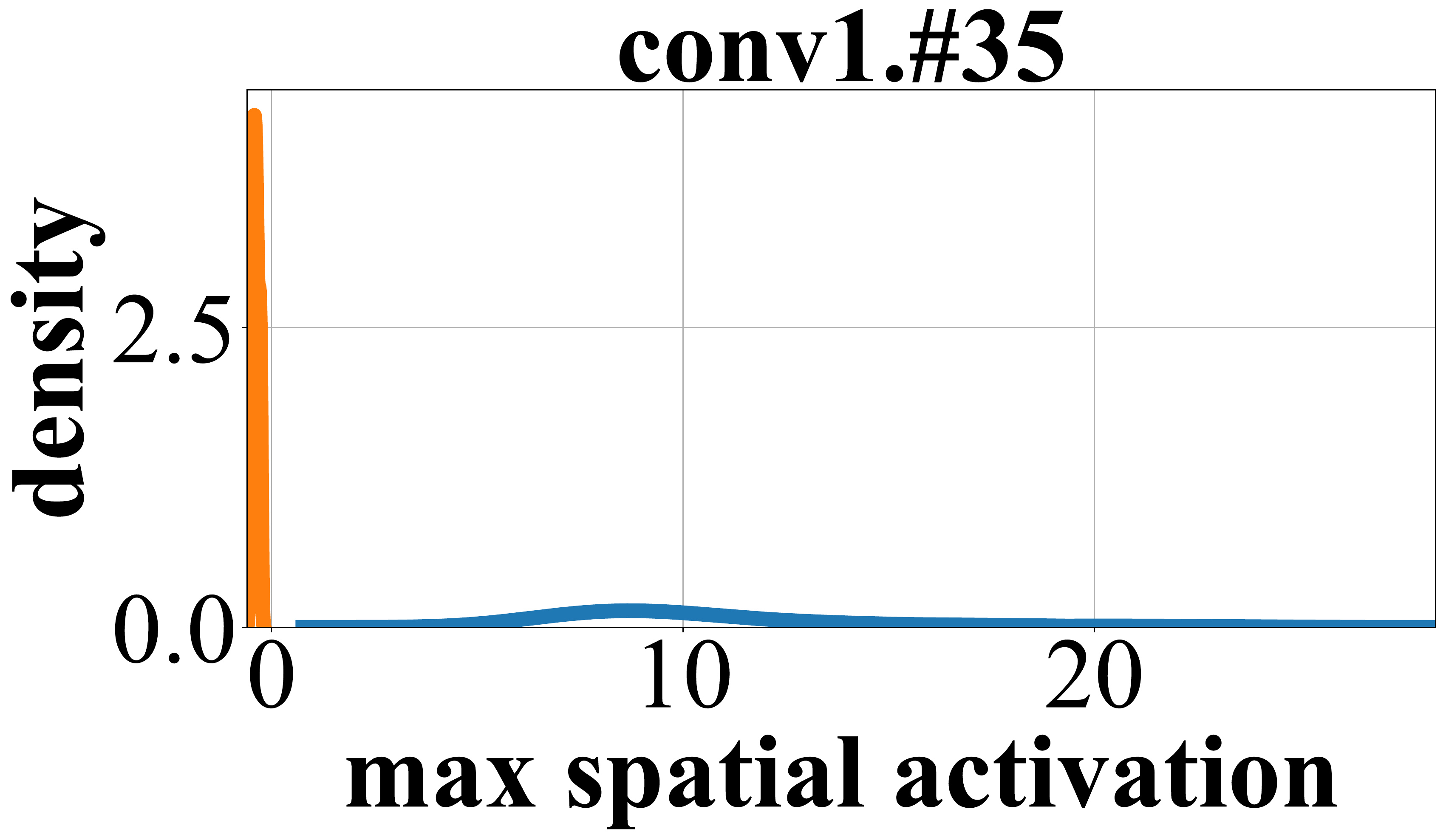} &
     \includegraphics[width=0.13\linewidth]{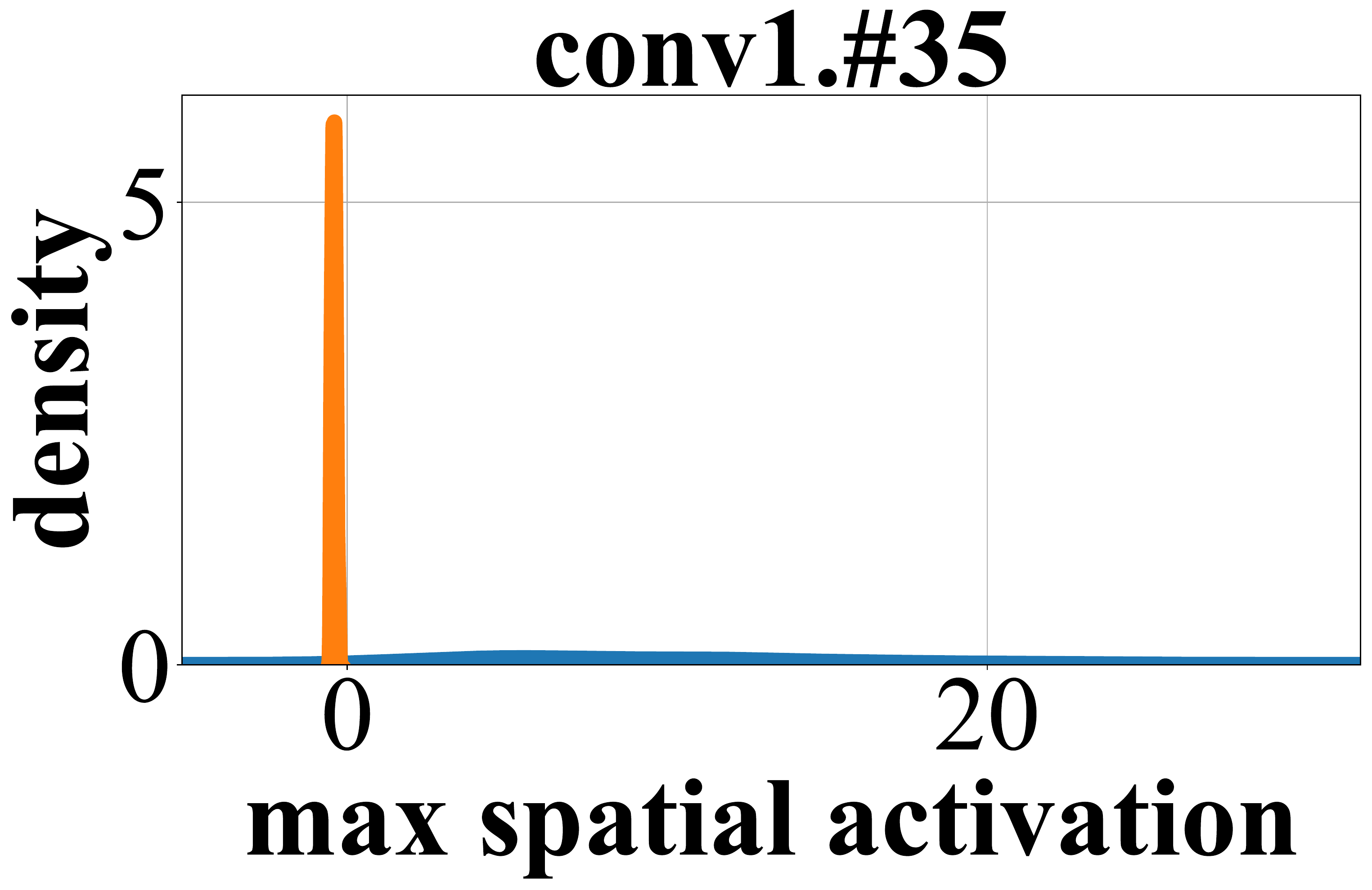}
    \\

\end{tabular}
\vspace{-0.1cm}
\includegraphics[width=0.50\linewidth]{pics_supp/activation_histograms/r50_0.5/conv1-1/legend.pdf}
\caption{
Additional results showing 
\textit{Color-conditional T-FF in ResNet-50:}
Each row represents a color-conditional \textit{T-FF} (exact same T-FF as shown in Fig. \ref{fig_supp:lrp_patches_r50}), and 
we show the maximum spatial activation distributions
for ProGAN \cite{karras2018progressive}, StyleGAN2 \cite{Karras_2020_CVPR}, StyleGAN \cite{Karras_2019_CVPR}, BigGAN \cite{brock2018large}, CycleGAN \cite{zhu2017unpaired}, StarGAN \cite{choi2018stargan} and GauGAN \cite{park2019semantic} counterfeits
before (Baseline) and after color ablation (Grayscale).
We remark that for each counterfeit in the ForenSynths dataset \cite{Wang_2020_CVPR}, we apply global max pooling to the specific T-FF to obtain a {\em maximum spatial activation} value (scalar).
We can clearly observe that these \textit{T-FF} are producing noticeably lower spatial activations (max) for the same set of counterfeits after removing color information. 
This clearly indicates that these \textit{T-FF} are color-conditional (Confirmed by Mood's median test).
}
\label{fig_supp:activation_hist_r50}
\end{figure}

\begin{figure}[!b]
\centering
\begin{tabular}{ccccccc}
    \multicolumn{1}{p{0.125\linewidth}}{\tiny \enskip ProGAN \cite{karras2018progressive}} &
    \multicolumn{1}{p{0.15\linewidth}}{\tiny  \enskip StyleGAN2 \cite{Karras_2020_CVPR}} &
    \multicolumn{1}{p{0.14\linewidth}}{\tiny StyleGAN \cite{Karras_2019_CVPR}} &
    \multicolumn{1}{p{0.125\linewidth}}{\tiny BigGAN \cite{brock2018large}} &
    \multicolumn{1}{p{0.132\linewidth}}{\tiny CycleGAN \cite{zhu2017unpaired}} &
    \multicolumn{1}{p{0.135\linewidth}}{\tiny \enskip StarGAN \cite{choi2018stargan}} &
    {\tiny GauGAN \cite{park2019semantic}} \\

    \multicolumn{7}{c}{\includegraphics[width=0.99\linewidth]{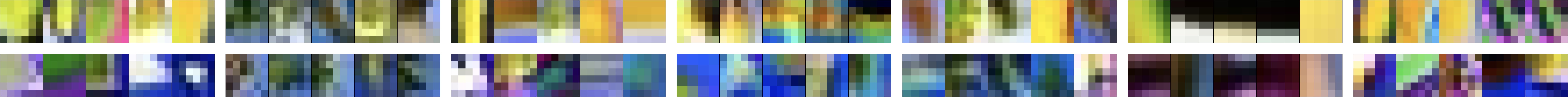}}

\end{tabular}
\caption{
Additional results demonstrating that color is a critical \textit{T-FF} in universal detectors (EfficientNet-B0):
Large-scale study on visual interpretability of \textit{T-FF} discovered through our proposed \textit{FF-RS ($\omega$)} reveal that color information is critical for cross-model forensic transfer.
Each row represents a color-based \textit{T-FF} and
we show the LRP-max response regions for ProGAN
\cite{karras2018progressive}, StyleGAN2 \cite{Karras_2020_CVPR}, StyleGAN \cite{Karras_2019_CVPR}, BigGAN \cite{brock2018large}, CycleGAN \cite{zhu2017unpaired}, StarGAN \cite{choi2018stargan} and GauGAN \cite{park2019semantic} 
counterfeits 
for our own version of EfficientNet-B0 \cite{tan2019efficientnet} universal detector following the exact training / test strategy proposed
by Wang \etal 
\cite{Wang_2020_CVPR}.
This detector is trained with ProGAN
\cite{karras2018progressive} 
counterfeits \cite{Wang_2020_CVPR} and cross-model forensic transfer is evaluated on other unseen GANs.
All counterfeits are obtained from the ForenSynths dataset
\cite{Wang_2020_CVPR}.
The consistent color-conditional LRP-max response across all GANs for these \textit{T-FF} clearly indicate that \textit{color} is critical for cross-model forensic transfer in universal detectors.
}
\label{fig_supp:lrp_patches_efb0}
\end{figure}

\begin{figure}[!t]
\centering
\begin{tabular}{ccccccc}
    {\tiny ProGAN \cite{karras2018progressive}} &
    {\tiny StyleGAN2 \cite{Karras_2020_CVPR}} &
    {\tiny StyleGAN \cite{Karras_2019_CVPR}} &
    {\tiny BigGAN \cite{brock2018large}} &
    {\tiny CycleGAN \cite{zhu2017unpaired}} &
    {\tiny StarGAN \cite{choi2018stargan}} &
    {\tiny GauGAN \cite{park2019semantic}} \\

 \includegraphics[width=0.13\linewidth]{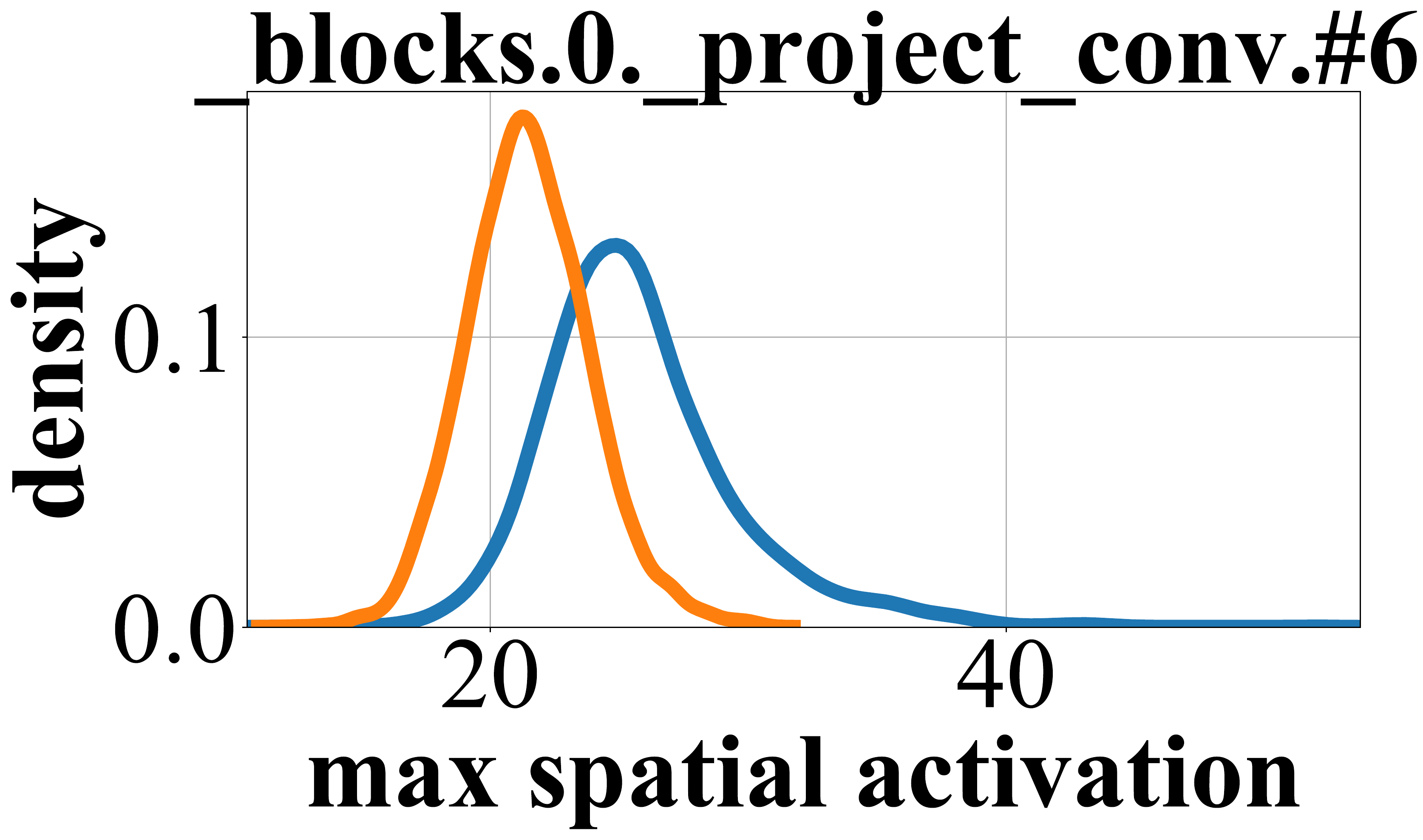} &
    \includegraphics[width=0.13\linewidth]{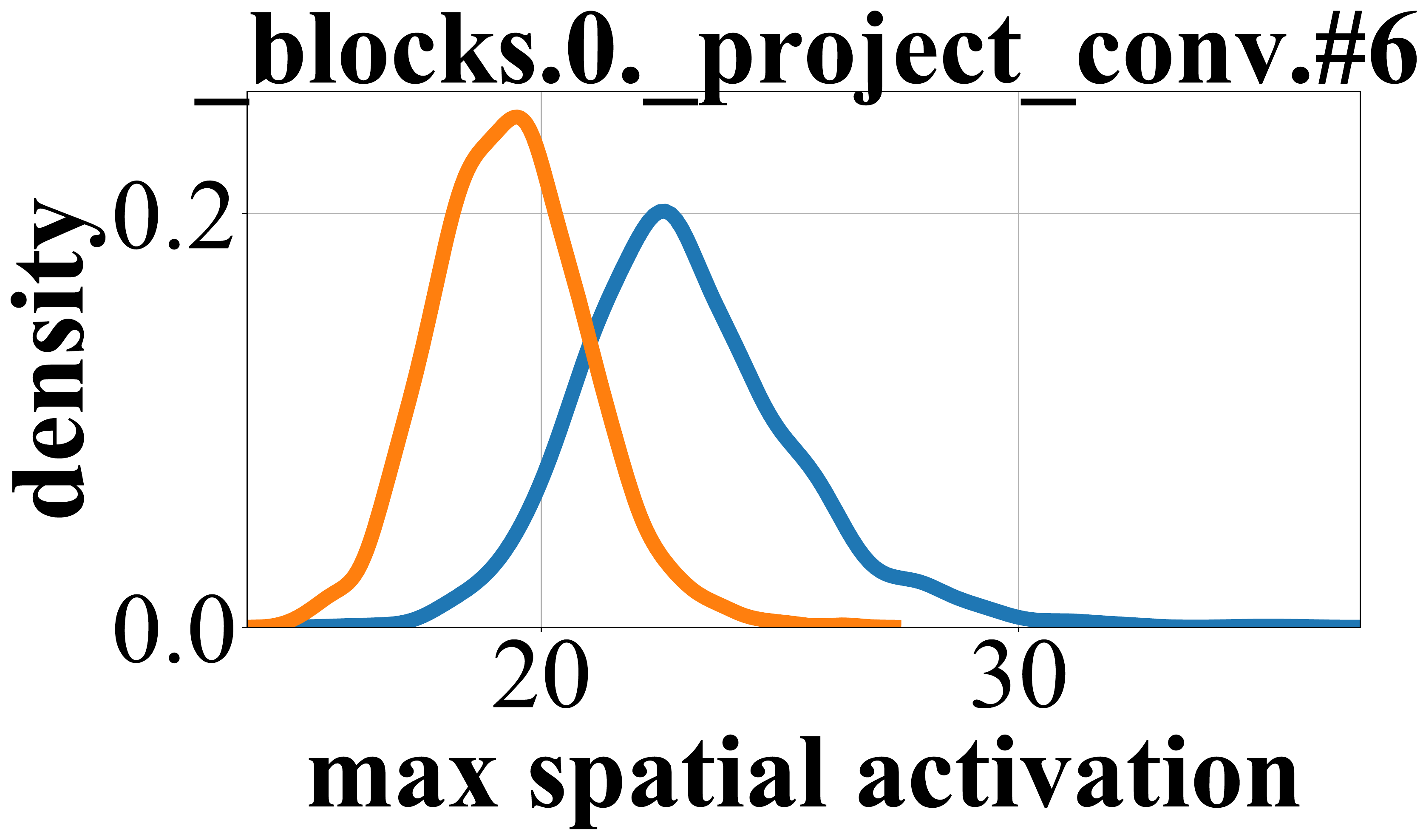} &
    \includegraphics[width=0.13\linewidth]{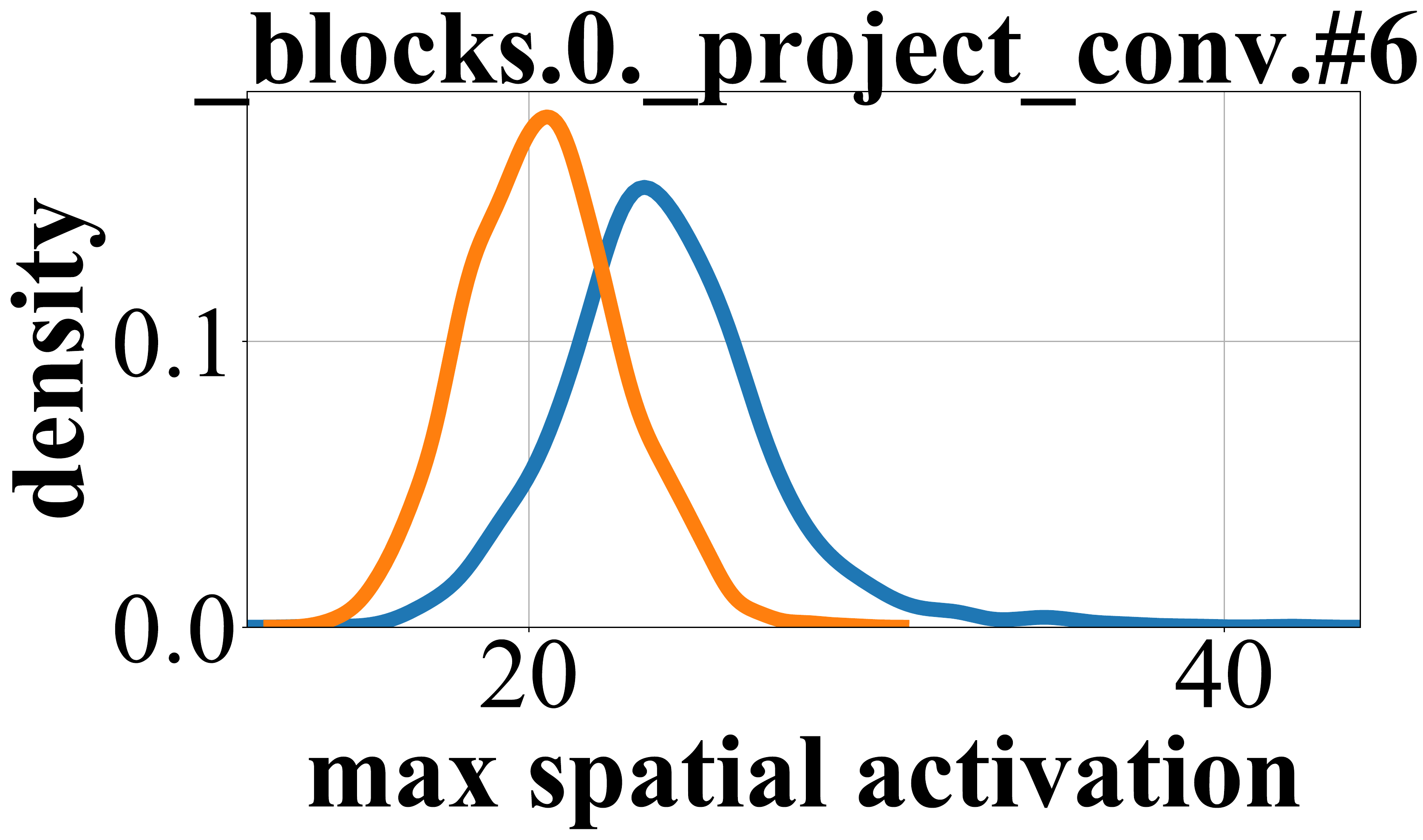} &
    \includegraphics[width=0.13\linewidth]{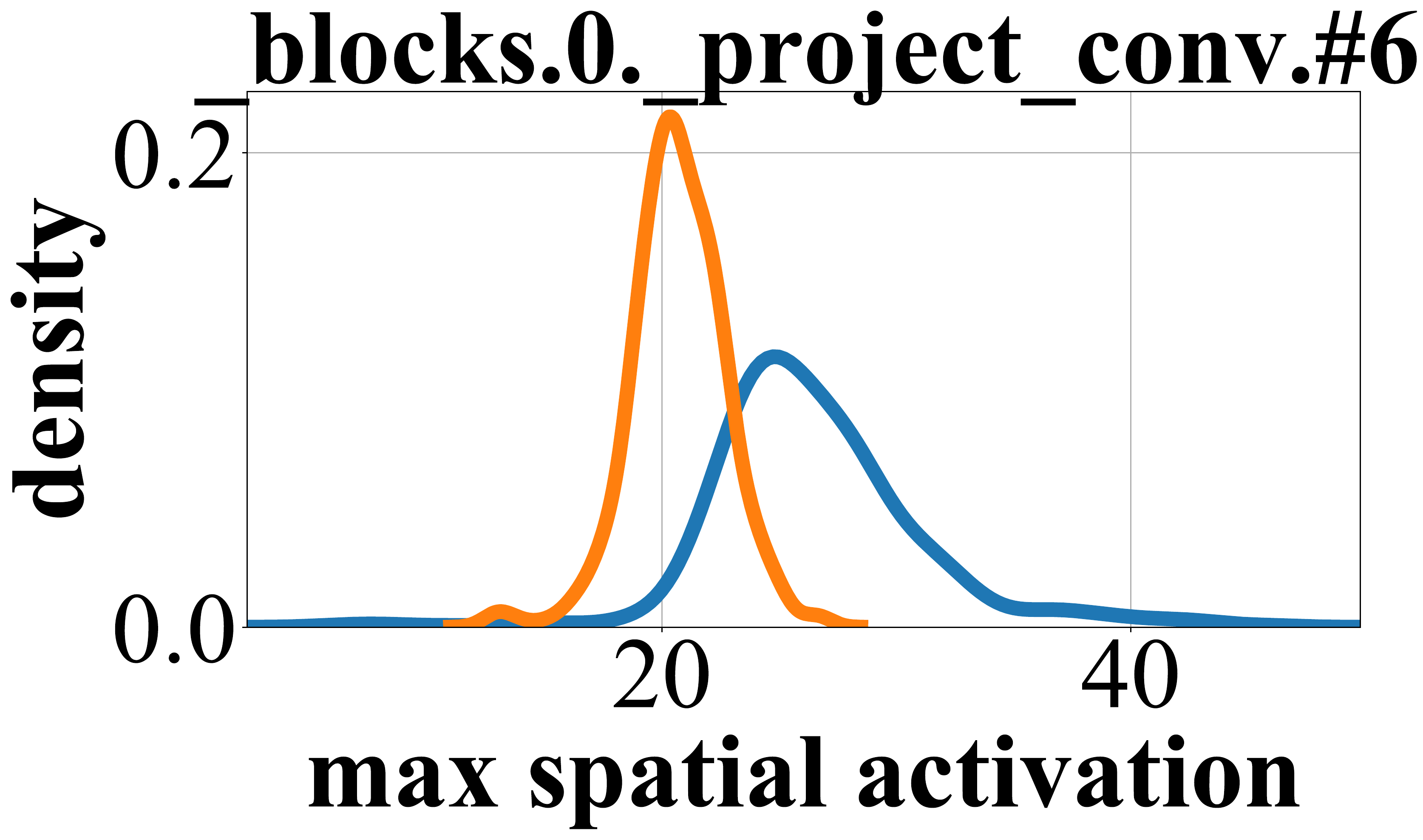} &
    \includegraphics[width=0.13\linewidth]{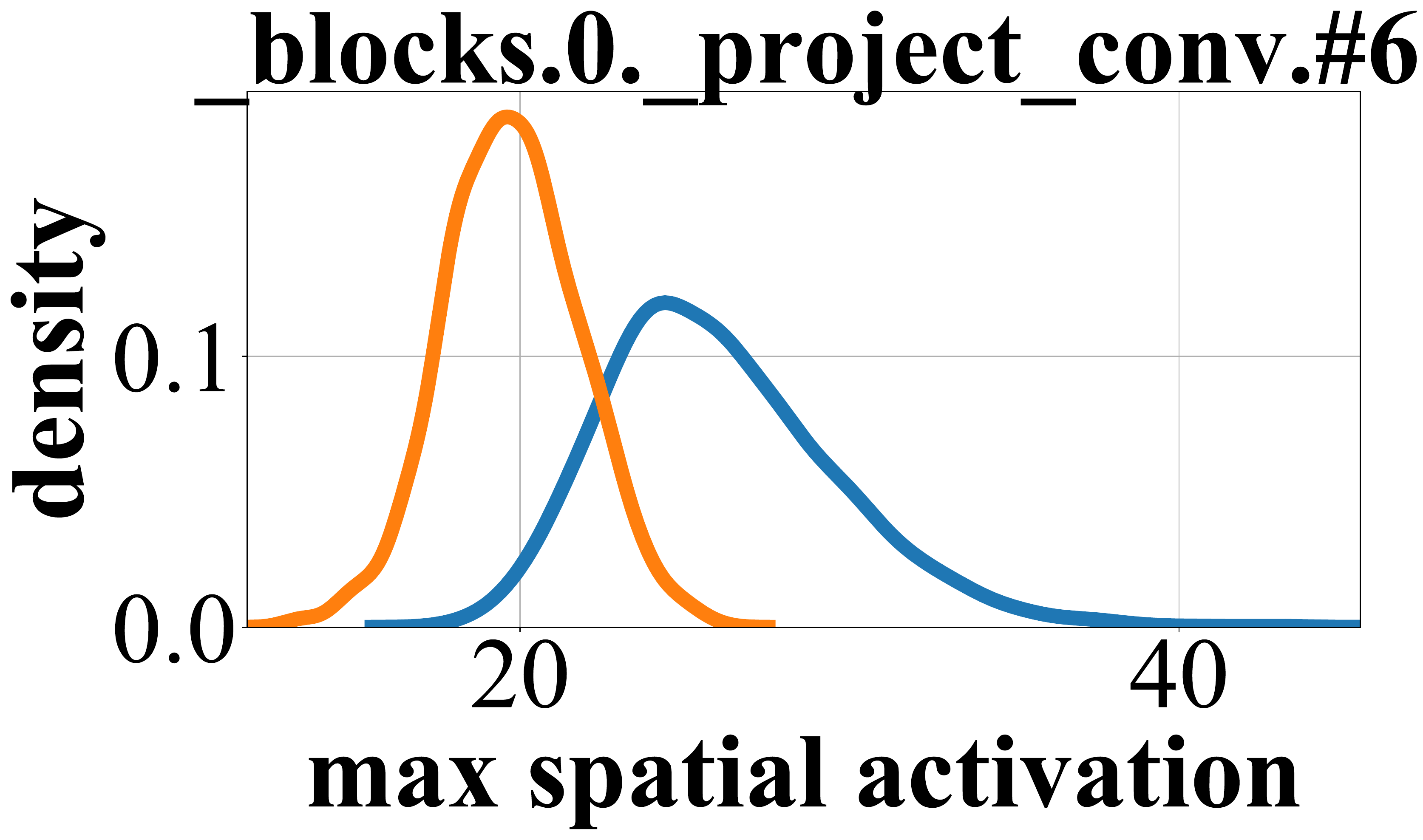} &
     \includegraphics[width=0.13\linewidth]{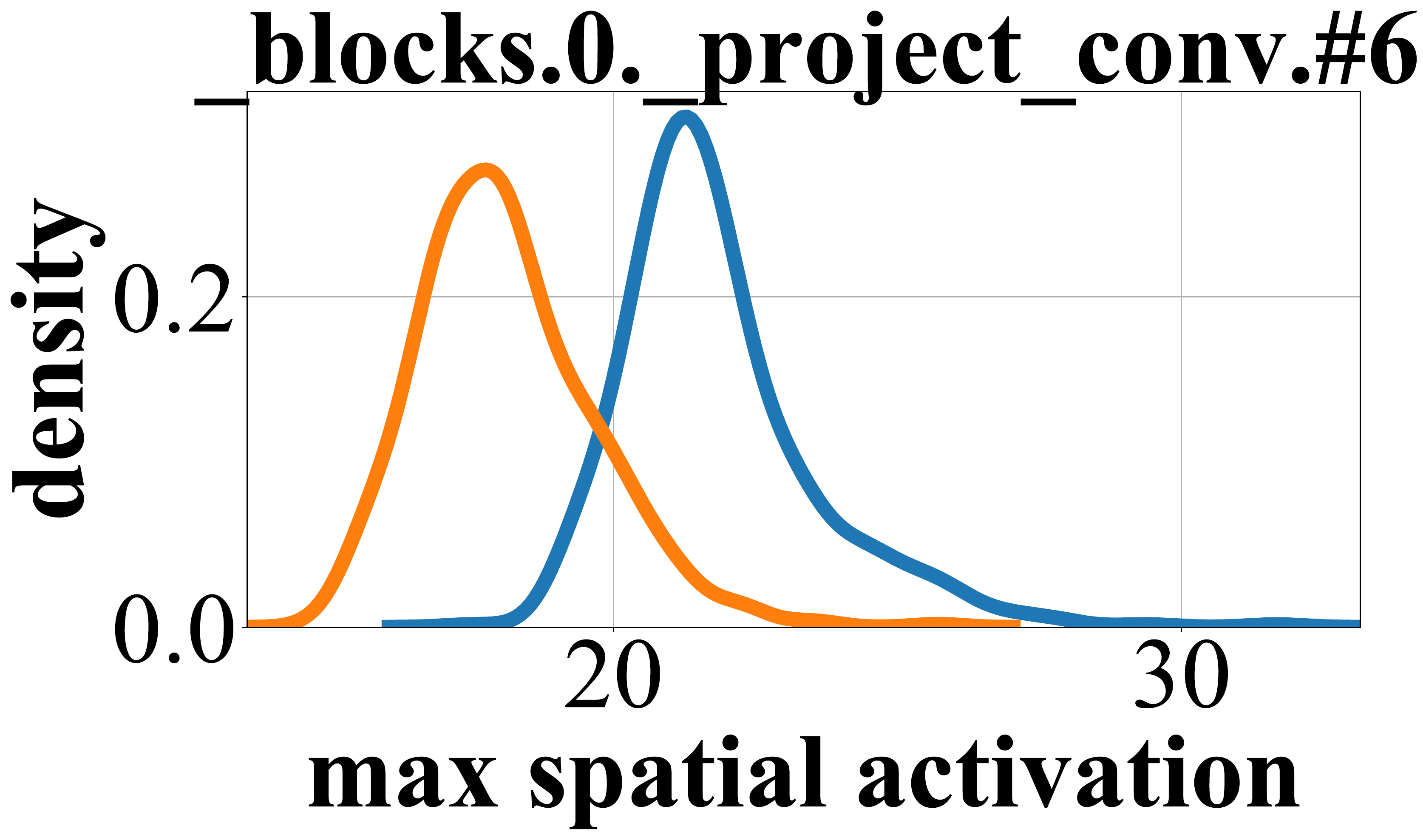} &
    \includegraphics[width=0.13\linewidth]{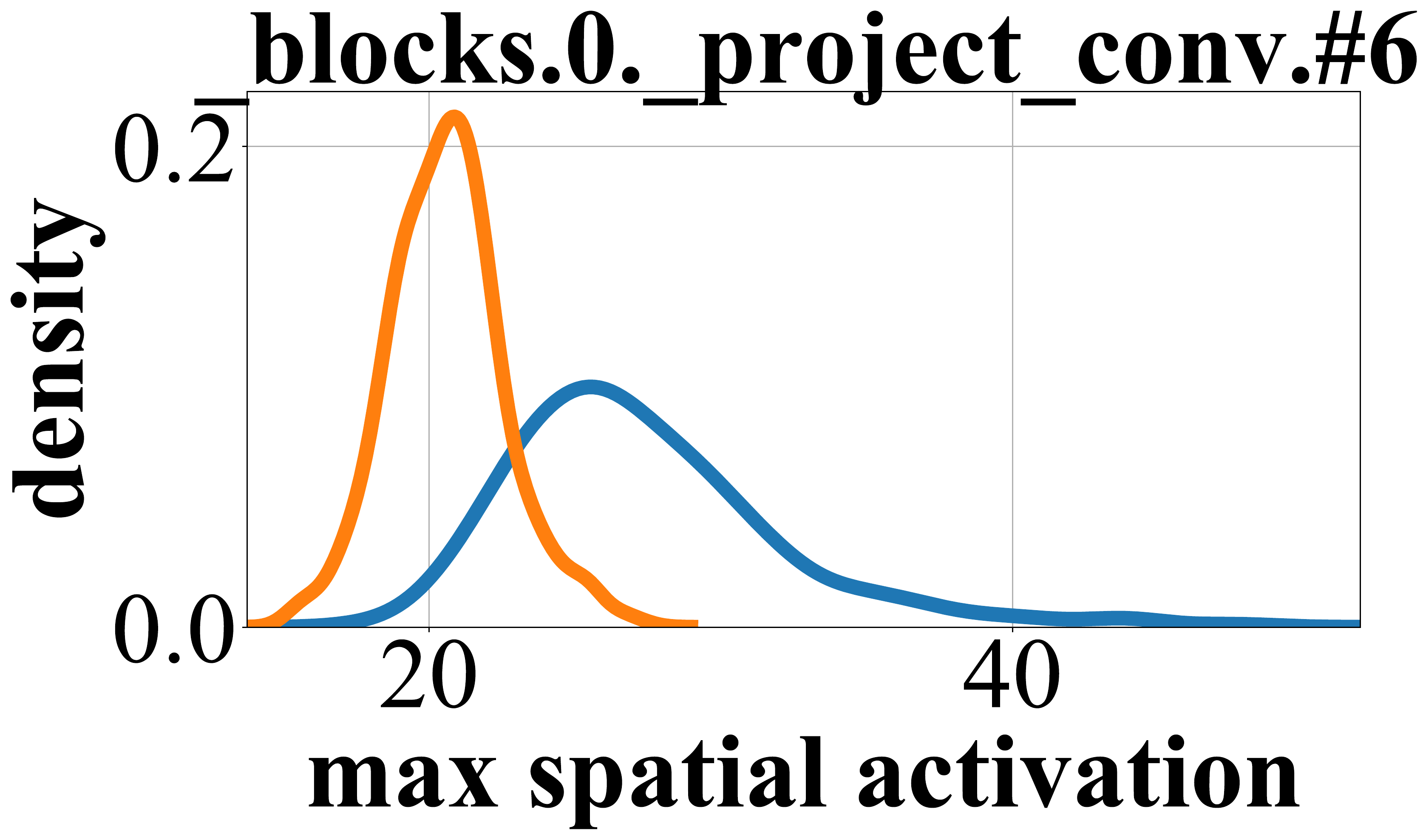}
    \\
    
     \includegraphics[width=0.13\linewidth]{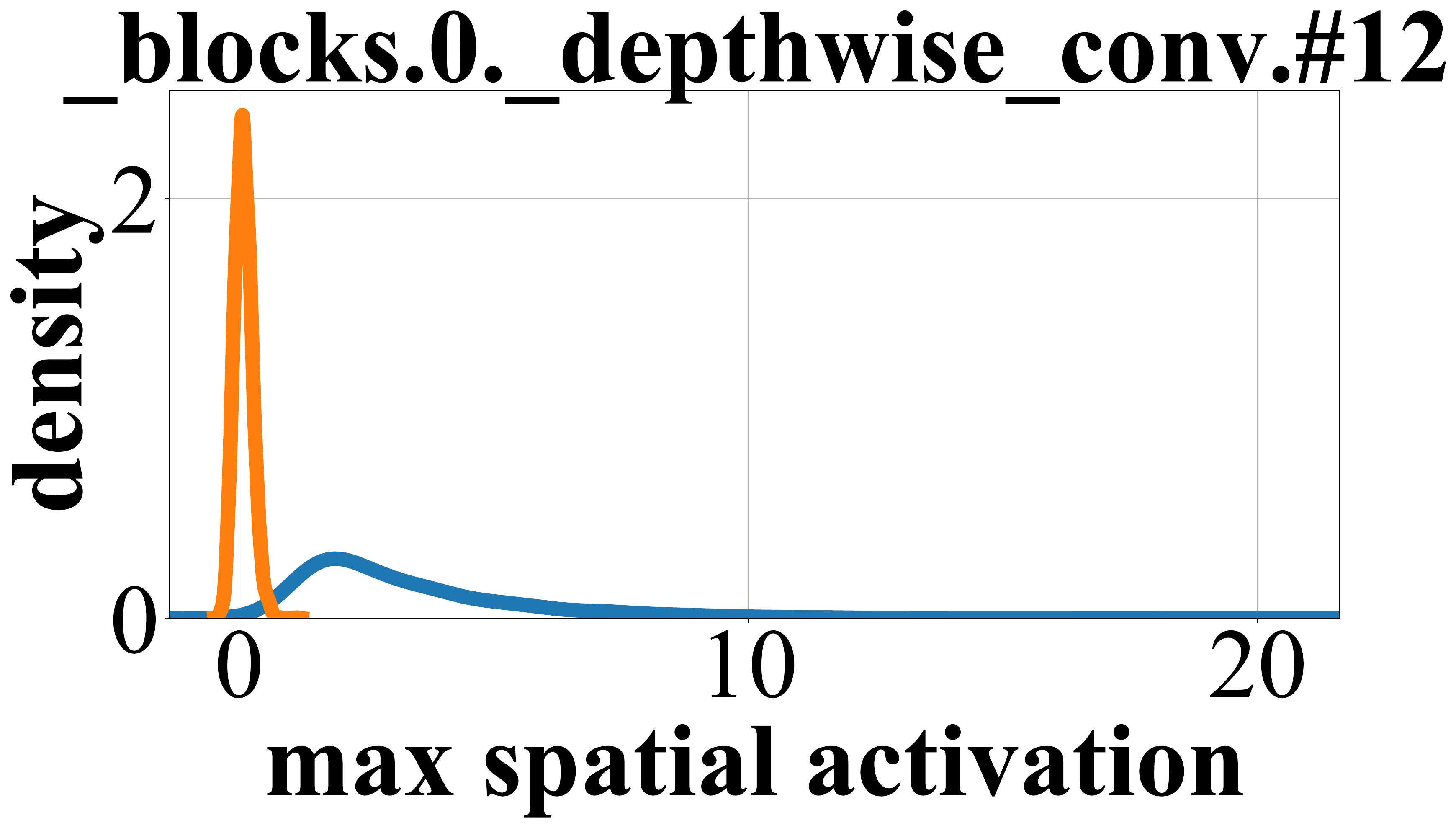} &
    \includegraphics[width=0.13\linewidth]{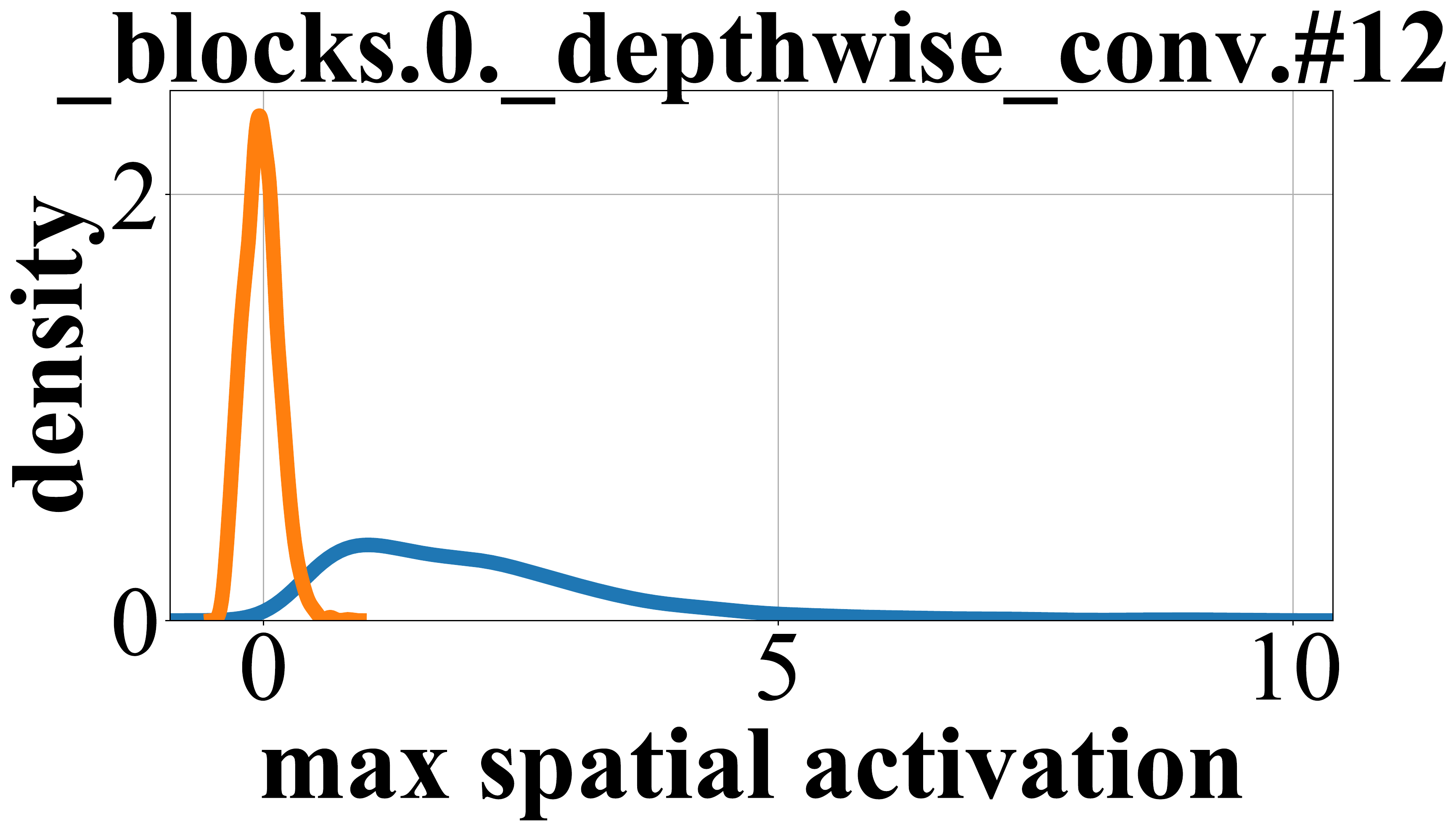} &
    \includegraphics[width=0.13\linewidth]{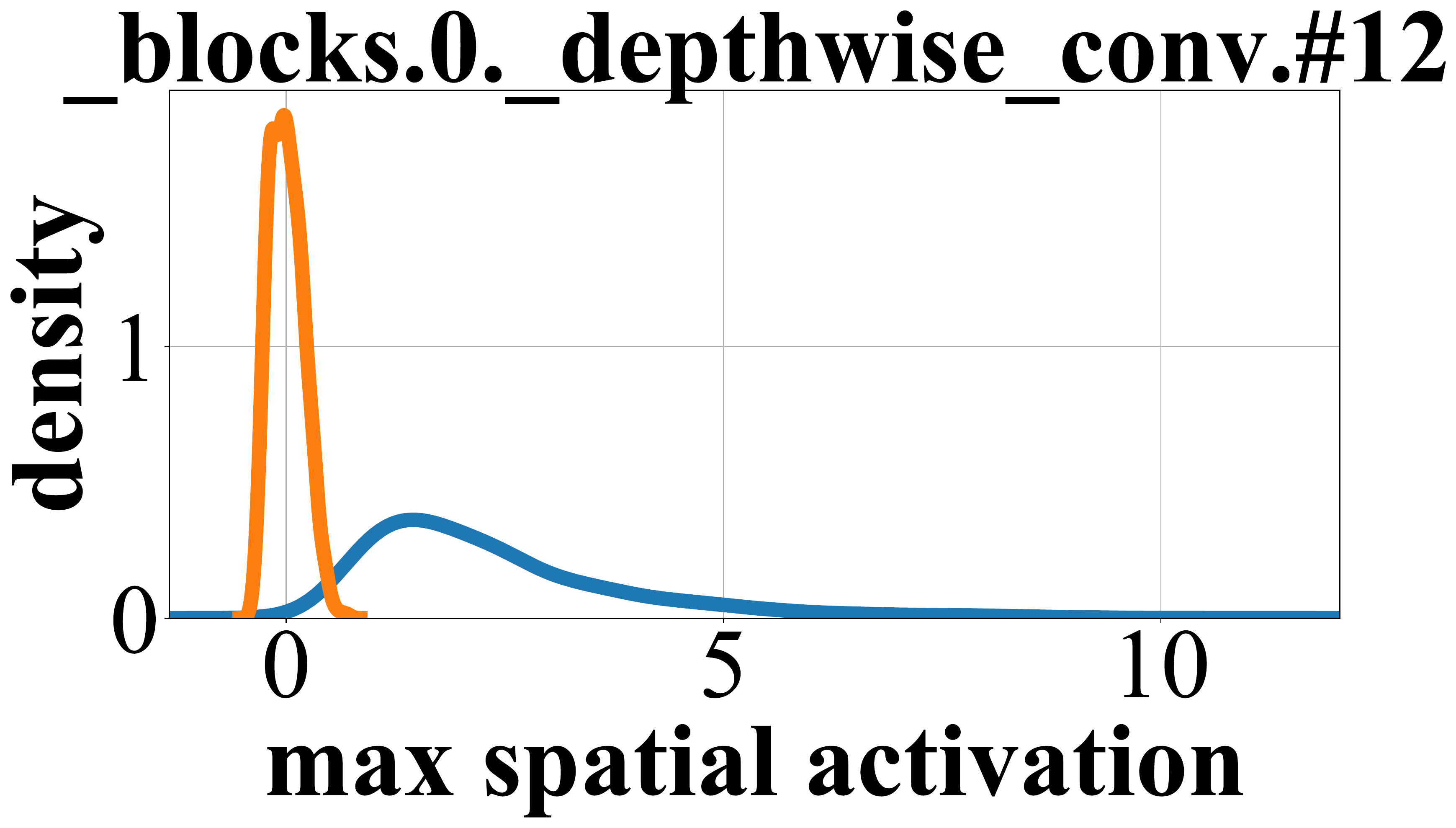} &
    \includegraphics[width=0.13\linewidth]{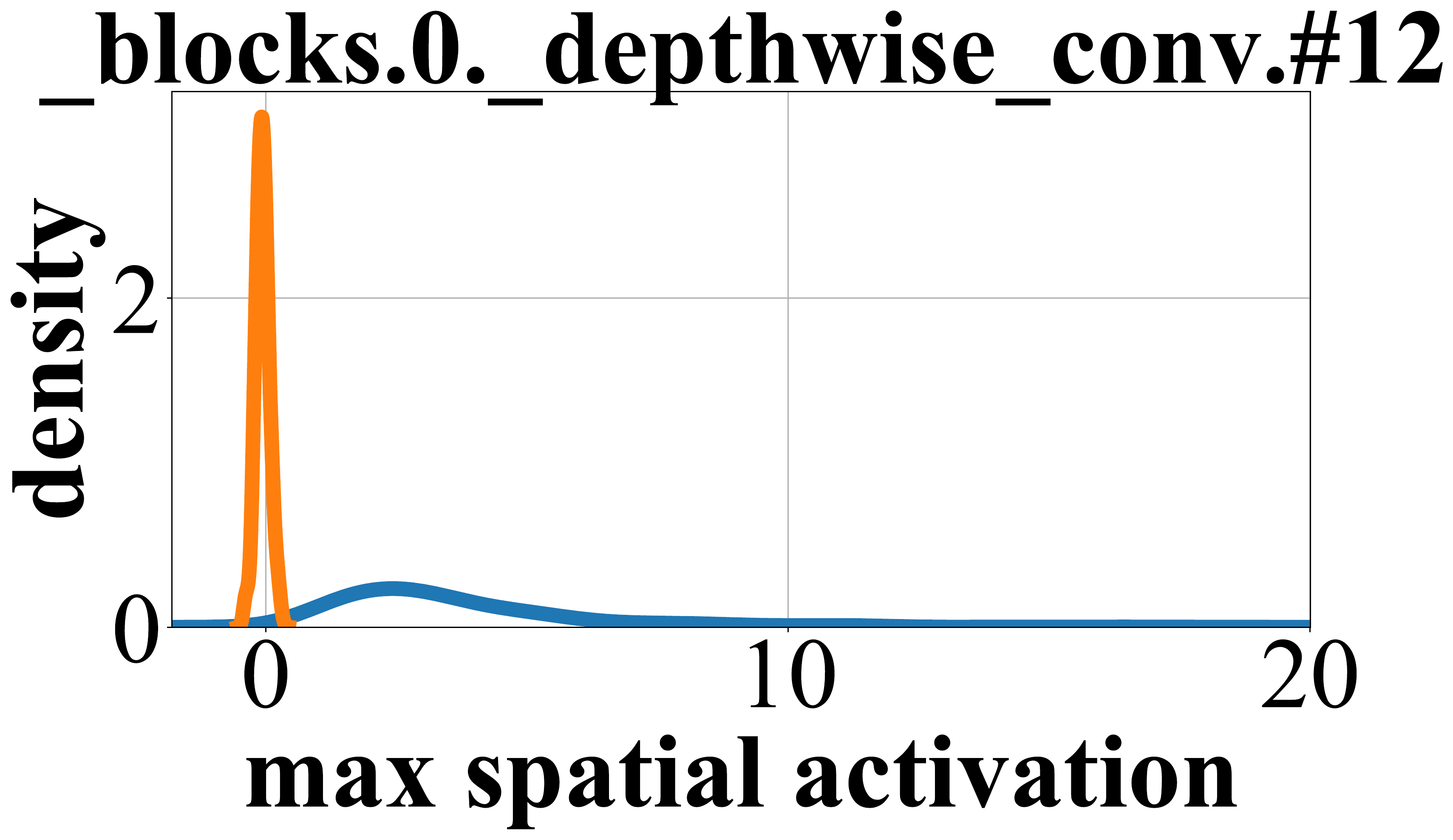} &
    \includegraphics[width=0.13\linewidth]{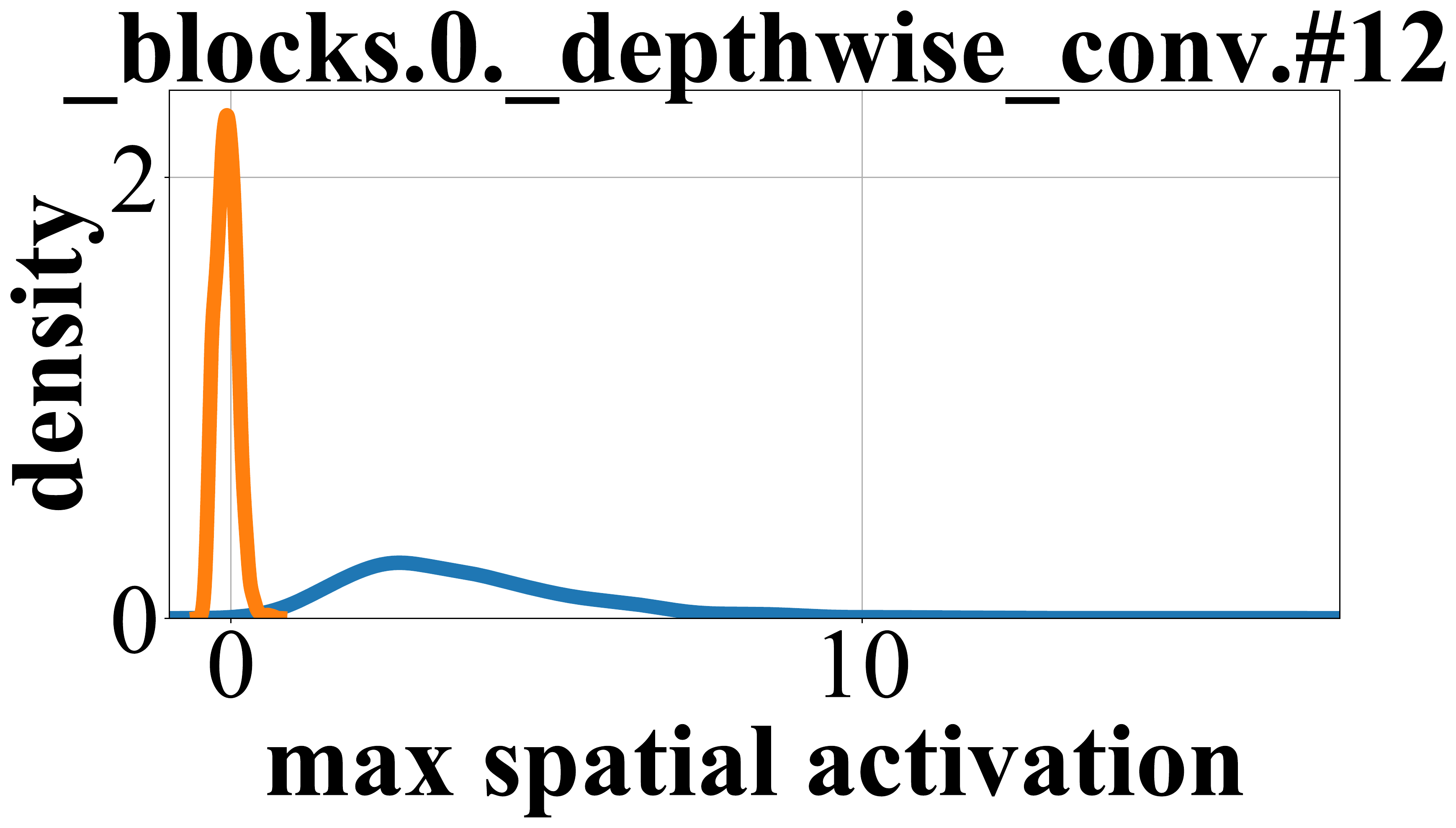} &
     \includegraphics[width=0.13\linewidth]{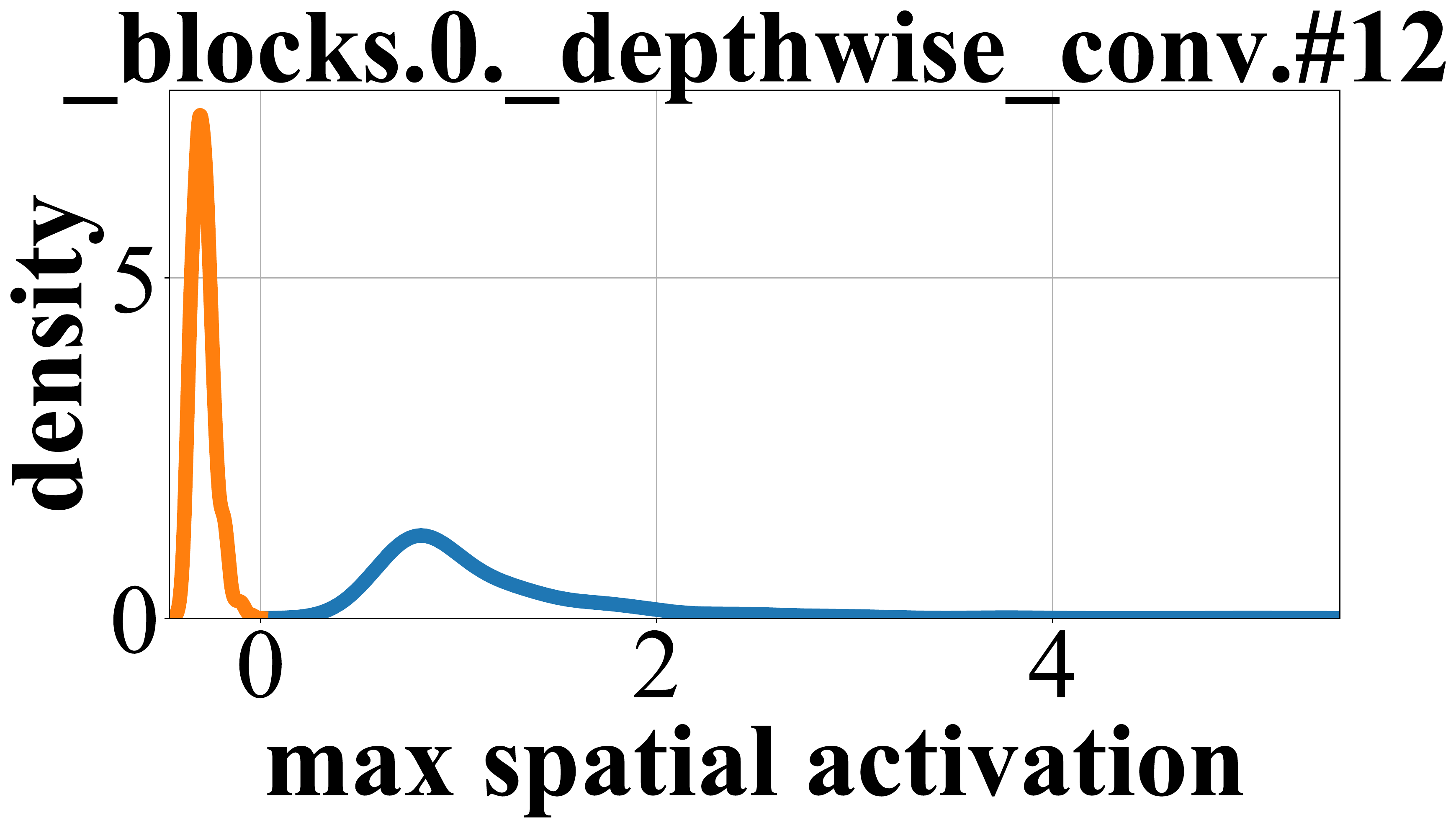} &
    \includegraphics[width=0.13\linewidth]{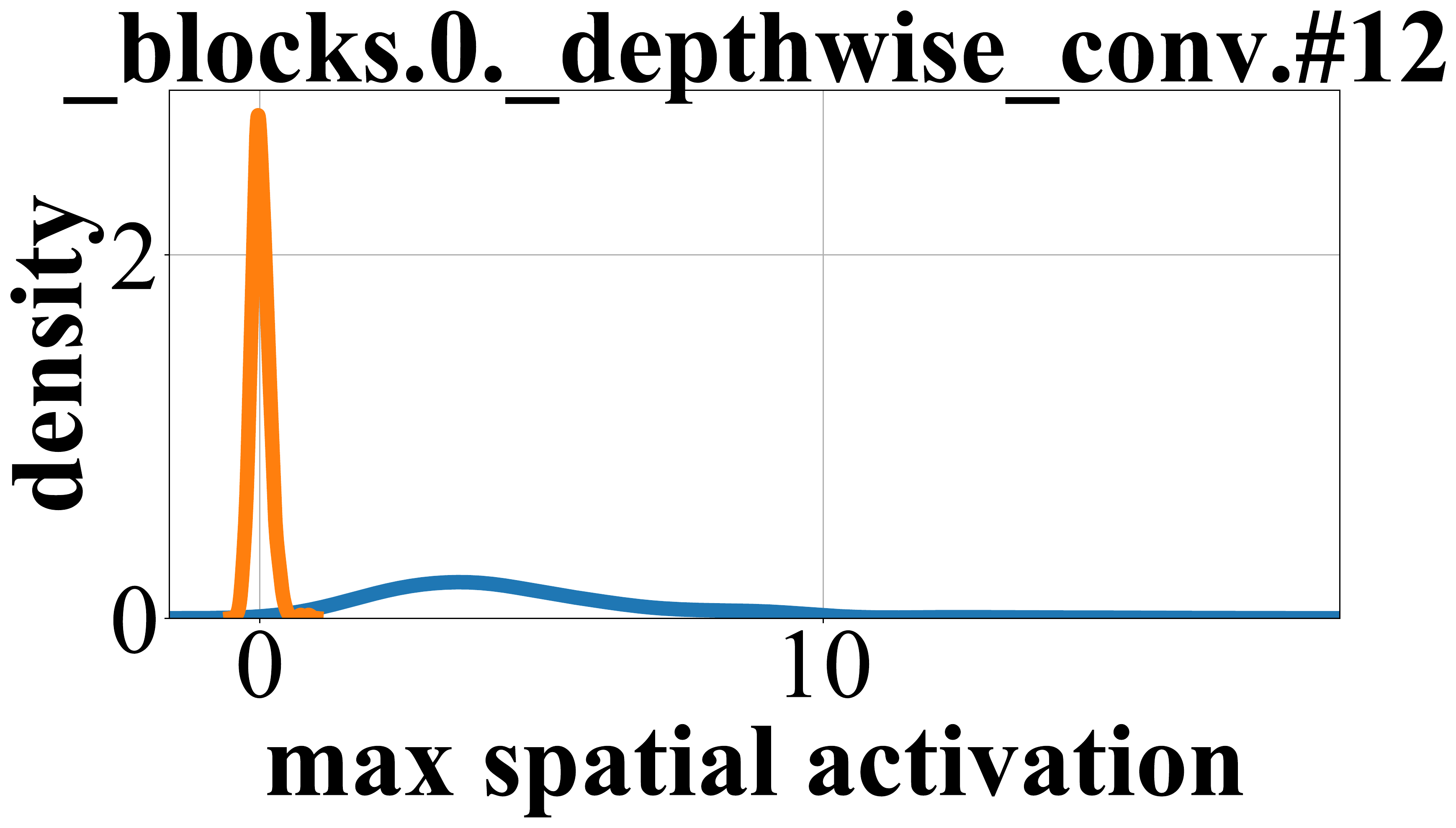}
    \\

\end{tabular}
\includegraphics[width=0.50\linewidth]{pics_supp/activation_histograms/r50_0.5/conv1-1/legend.pdf}
\caption{
Additional results showing 
\textit{Color-conditional T-FF in EfficientNet-B0:}
Each row represents a color-conditional \textit{T-FF} (exact same T-FF as shown in Fig. \ref{fig_supp:lrp_patches_efb0}), and 
we show the maximum spatial activation distributions
for ProGAN \cite{karras2018progressive}, StyleGAN2 \cite{Karras_2020_CVPR}, StyleGAN \cite{Karras_2019_CVPR}, BigGAN \cite{brock2018large}, CycleGAN \cite{zhu2017unpaired}, StarGAN \cite{choi2018stargan} and GauGAN \cite{park2019semantic} counterfeits
before (Baseline) and after color ablation (Grayscale).
We remark that for each counterfeit in the ForenSynths dataset \cite{Wang_2020_CVPR}, we apply global max pooling to the specific T-FF to obtain a {\em maximum spatial activation} value (scalar).
We can clearly observe that these \textit{T-FF} are producing noticeably lower spatial activations (max) for the same set of counterfeits after removing color information. 
This clearly indicates that these \textit{T-FF} are color-conditional
(Confirmed by Mood's median test).
}
\label{fig_supp:activation_hist_efb0}
\end{figure}

In this section, we show more color-conditional T-FF to support our finding that \textit{color} is a critical \textit{T-FF}. 
We show LRP-max response image regions for ResNet-50 and EfficientNet-B0 in Fig. \ref{fig_supp:lrp_patches_r50} and \ref{fig_supp:lrp_patches_efb0} respectively.
We further show the maximum spatial activation distributions before and after color ablation for these color-conditional \textit{T-FF} in Fig. \ref{fig_supp:activation_hist_r50}(ResNet-50) and Fig. \ref{fig_supp:activation_hist_efb0}(EfficientNet-B0) respectively.
\vspace{1cm}


\section{CR-Universal Detectors (Additional Results)}
\label{sec_supp:cr_universal_detectors}
\setcounter{figure}{0} 
\setcounter{table}{0} 

We show the AP, real and GAN detection accuracies for the universal Detectors in Table \ref{table_supp:original} and CR-Universal Detectors trained using our proposed data augmentation scheme in Table \ref{table_supp:cr}.
As one can observe, our proposed CR-universal detectors are more robust and can avoid attacks from color-ablated counterfeits compared to the original detectors proposed by Wang \etal \cite{Wang_2020_CVPR}.

\begin{table}[!h]
\caption{
Universal detectors are more susceptible to color ablated counterfeit attacks as color is a critical \textit{T-FF}:
We show the results for the publicly released ResNet-50 universal detector 
\cite{Wang_2020_CVPR} 
(top) and 
our own version of EfficientNet-B0 \cite{tan2019efficientnet}
universal detector (bottom) following the exact training and test strategy proposed in
\cite{Wang_2020_CVPR}. 
We show the AP, real and GAN image detection accuracies for Baseline and Grayscale (color ablated) images.
As one can observe, AP and GAN detection accuracies drop \textit{substantially} during cross-model transfer when removing color information from counterfeits.
}
\begin{center}
\begin{adjustbox}{width=1.0\columnwidth,center}
\begin{tabular}{c|ccc|ccc|ccc|ccc|ccc|ccc|ccc}
\multicolumn{22}{c}{\bf \Large ResNet-50} \\ \toprule
\textbf{} &\multicolumn{3}{c}{\textbf{ProGAN} \cite{karras2018progressive}} 
&\multicolumn{3}{c}{\textbf{StyleGAN2} \cite{Karras_2020_CVPR}} 
&\multicolumn{3}{c}{\textbf{StyleGAN} \cite{Karras_2019_CVPR}} &\multicolumn{3}{c}{\textbf{BigGAN} \cite{brock2018large}} &\multicolumn{3}{c}{\textbf{CycleGAN} \cite{zhu2017unpaired} } &\multicolumn{3}{c}{\textbf{StarGAN} \cite{choi2018stargan}} &\multicolumn{3}{c}{\textbf{GauGAN} \cite{park2019semantic}} \\

\cmidrule{2-22}

\textbf{} &\textbf{AP} &\textbf{Real} &\textbf{GAN} &\textbf{AP} &\textbf{Real} &\textbf{GAN} &\textbf{AP} &\textbf{Real} &\textbf{GAN} &\textbf{AP} &\textbf{Real} &\textbf{GAN} &\textbf{AP} &\textbf{Real} &\textbf{GAN} &\textbf{AP} &\textbf{Real} &\textbf{GAN} &\textbf{AP} &\textbf{Real} &\textbf{GAN} \\
\midrule

Baseline &100.0 &100.0 &100.0 &99.1 &95.5 &95.0 &99.3 &96.0 &95.6 &90.4 &83.9 &85.1 &97.9 &93.4 &92.6 &97.5 &94.0 &89.3 &98.8 &93.9 &96.4 \\

\midrule

Grayscale &99.9 &100.0 &81.5 &\textbf{89.1} &92.7 &\textbf{61.9} &\textbf{96.7} &94.6 &\textbf{84.8} &\textbf{75.2} &85.8 &\textbf{48.8} &\textbf{84.2} &94.5 &\textbf{41.0} &\textbf{89.2} &93.4 &\textbf{60.7} &\textbf{97.6} &97.7 &\textbf{78.8} \\
\bottomrule
\end{tabular}
\end{adjustbox}
\end{center}

\begin{center}
\begin{adjustbox}{width=1.0\columnwidth,center}
\begin{tabular}{c|ccc|ccc|ccc|ccc|ccc|ccc|ccc}
\multicolumn{22}{c}{\bf \Large EfficientNet-B0} \\ \toprule
\textbf{} &\multicolumn{3}{c}{\textbf{ProGAN} \cite{karras2018progressive}} 
&\multicolumn{3}{c}{\textbf{StyleGAN2} \cite{Karras_2020_CVPR}} 
&\multicolumn{3}{c}{\textbf{StyleGAN} \cite{Karras_2019_CVPR}} &\multicolumn{3}{c}{\textbf{BigGAN} \cite{brock2018large}} &\multicolumn{3}{c}{\textbf{CycleGAN} \cite{zhu2017unpaired} } &\multicolumn{3}{c}{\textbf{StarGAN} \cite{choi2018stargan}} &\multicolumn{3}{c}{\textbf{GauGAN} \cite{park2019semantic}} \\

\cmidrule{2-22}

\textbf{} &\textbf{AP} &\textbf{Real} &\textbf{GAN} &\textbf{AP} &\textbf{Real} &\textbf{GAN} &\textbf{AP} &\textbf{Real} &\textbf{GAN} &\textbf{AP} &\textbf{Real} &\textbf{GAN} &\textbf{AP} &\textbf{Real} &\textbf{GAN} &\textbf{AP} &\textbf{Real} &\textbf{GAN} &\textbf{AP} &\textbf{Real} &\textbf{GAN} \\
\midrule

Baseline &100.0 &100.0 &100.0 &99.0 &95.2 &85.4 &99.0 &96.1 &94.3 &84.4 &79.7 &75.9 &97.3 &89.6 &93.0 &96.0 &92.8 &85.5 &98.3 &94.1 &94.4 \\

\midrule

Grayscale &99.9 &100.0 &80.0 &\textbf{91.0} &95.2 &\textbf{26.6} &\textbf{91.0} &97.2 &56.0 &\textbf{68.4} &91.7 &\textbf{28.9} &\textbf{86.5} &96.4 &\textbf{40.0} &\textbf{91.8} &91.3 &\textbf{72.9} &\textbf{93.7} &99.7 &\textbf{48.2} \\

\bottomrule
\end{tabular}
\end{adjustbox}
\end{center}

\label{table_supp:original}
\end{table}

\begin{table}
\caption{
CR-Universal detectors trained using our proposed data augmentation scheme are more robust to color ablated counterfeits:
We show the results for the ResNet-50 universal detector 
\cite{Wang_2020_CVPR} 
(top) and 
our own version of EfficientNet-B0 \cite{tan2019efficientnet}
universal detector (bottom) following the exact training / test strategy proposed in
\cite{Wang_2020_CVPR}. 
We show the AP, real and GAN image detection accuracies for Baseline and Grayscale (color ablated) images.
As one can observe, AP and GAN detection accuracies remain similar during forensic transfer when removing color information from counterfeits.
}
\begin{center}
\begin{adjustbox}{width=1.0\columnwidth,center}
\begin{tabular}{c|ccc|ccc|ccc|ccc|ccc|ccc|ccc}
\multicolumn{22}{c}{\bf \Large CR-ResNet-50} \\ \toprule
\textbf{} &\multicolumn{3}{c}{\textbf{ProGAN} \cite{karras2018progressive}} 
&\multicolumn{3}{c}{\textbf{StyleGAN2} \cite{Karras_2020_CVPR}} 
&\multicolumn{3}{c}{\textbf{StyleGAN} \cite{Karras_2019_CVPR}} &\multicolumn{3}{c}{\textbf{BigGAN} \cite{brock2018large}} &\multicolumn{3}{c}{\textbf{CycleGAN} \cite{zhu2017unpaired} } &\multicolumn{3}{c}{\textbf{StarGAN} \cite{choi2018stargan}} &\multicolumn{3}{c}{\textbf{GauGAN} \cite{park2019semantic}} \\

\cmidrule{2-22}

\textbf{} &\textbf{AP} &\textbf{Real} &\textbf{GAN} &\textbf{AP} &\textbf{Real} &\textbf{GAN} &\textbf{AP} &\textbf{Real} &\textbf{GAN} &\textbf{AP} &\textbf{Real} &\textbf{GAN} &\textbf{AP} &\textbf{Real} &\textbf{GAN} &\textbf{AP} &\textbf{Real} &\textbf{GAN} &\textbf{AP} &\textbf{Real} &\textbf{GAN} \\
\midrule

Baseline &100.0 &100.0 &100.0 &98.5 &94.4 &92.8 &99.5 &97.4 &95.3 &89.9 &80.3 &86.8 &96.6 &90.2 &90.3 &96.2 &91.2 &88.8 &99.5 &96.5 &96.8 \\

\midrule

Grayscale &100.0 &100.0 &100.0 &\textbf{98.0} &90.0 &\textbf{95.0} &\textbf{99.6} &95.1 &\textbf{98.0} &\textbf{87.6} &72.7 &\textbf{88.8} &\textbf{91.1} &81.6 &\textbf{81.8} &\textbf{95.4} &87.0 &\textbf{89.5} &\textbf{99.4} &95.1 &\textbf{97.2} \\
\bottomrule

\end{tabular}
\end{adjustbox}
\end{center}

\begin{center}
\begin{adjustbox}{width=1.0\columnwidth,center}
\begin{tabular}{c|ccc|ccc|ccc|ccc|ccc|ccc|ccc}
\multicolumn{22}{c}{\bf \Large CR-EfficientNet-B0} \\ \toprule
\textbf{} &\multicolumn{3}{c}{\textbf{ProGAN} \cite{karras2018progressive}} 
&\multicolumn{3}{c}{\textbf{StyleGAN2} \cite{Karras_2020_CVPR}} 
&\multicolumn{3}{c}{\textbf{StyleGAN} \cite{Karras_2019_CVPR}} &\multicolumn{3}{c}{\textbf{BigGAN} \cite{brock2018large}} &\multicolumn{3}{c}{\textbf{CycleGAN} \cite{zhu2017unpaired} } &\multicolumn{3}{c}{\textbf{StarGAN} \cite{choi2018stargan}} &\multicolumn{3}{c}{\textbf{GauGAN} \cite{park2019semantic}} \\

\cmidrule{2-22}

\textbf{} &\textbf{AP} &\textbf{Real} &\textbf{GAN} &\textbf{AP} &\textbf{Real} &\textbf{GAN} &\textbf{AP} &\textbf{Real} &\textbf{GAN} &\textbf{AP} &\textbf{Real} &\textbf{GAN} &\textbf{AP} &\textbf{Real} &\textbf{GAN} &\textbf{AP} &\textbf{Real} &\textbf{GAN} &\textbf{AP} &\textbf{Real} &\textbf{GAN} \\
\midrule

Baseline &100.0 &100.0 &100.0 &98.1 &92.3 &74.5 &98.1 &97.2 &90.5 &82.3 &78.0 &70.3 &95.7 &89.0 &88.5 &95.9 &90.2 &87.3 &99.0 &96.4 &94.5 \\

\midrule

Grayscale &100.0 &100.0 &100.0 &\textbf{98.8} &91.4 &\textbf{77.9} &\textbf{98.8} &95.7 &\textbf{94.4} &\textbf{81.0} &76.5 &\textbf{71.3} &\textbf{91.3} &85.9 &\textbf{78.5} &\textbf{94.8} &90.5 &\textbf{84.0} &\textbf{98.8} &95.2 &\textbf{94.1} \\

\bottomrule

\end{tabular}
\end{adjustbox}
\end{center}

\label{table_supp:cr}
\end{table}

\begin{figure}
\centering
\begin{tabular}{ccccccc}
    \multicolumn{1}{p{0.125\linewidth}}{\tiny \enskip ProGAN \cite{karras2018progressive}} &
    \multicolumn{1}{p{0.15\linewidth}}{\tiny  \enskip StyleGAN2 \cite{Karras_2020_CVPR}} &
    \multicolumn{1}{p{0.14\linewidth}}{\tiny StyleGAN \cite{Karras_2019_CVPR}} &
    \multicolumn{1}{p{0.125\linewidth}}{\tiny BigGAN \cite{brock2018large}} &
    \multicolumn{1}{p{0.132\linewidth}}{\tiny CycleGAN \cite{zhu2017unpaired}} &
    \multicolumn{1}{p{0.135\linewidth}}{\tiny \enskip StarGAN \cite{choi2018stargan}} &
    {\tiny GauGAN \cite{park2019semantic}} \\
    
    \multicolumn{7}{c}{\includegraphics[width=0.99\linewidth]{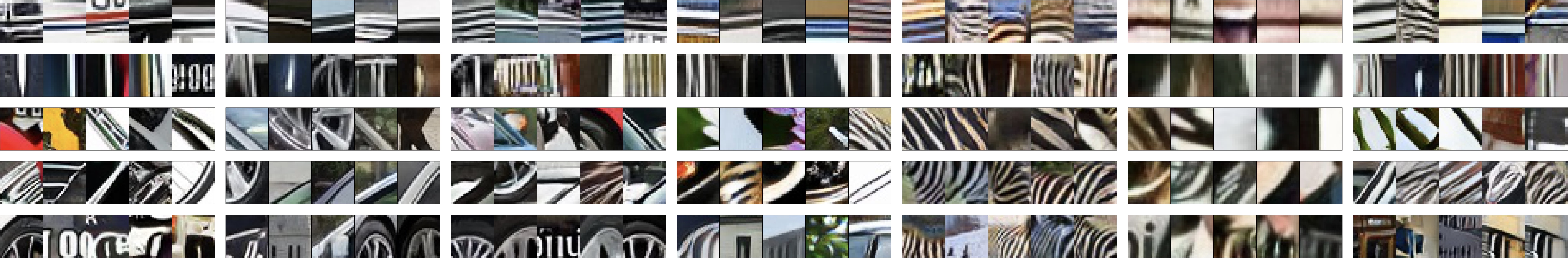}}

\end{tabular}
\caption{
\textit{T-FF in CR-ResNet-50:}
Each row represents a \textit{T-FF}. 
As visible, these T-FF are largely faintly colored. 
Notable patterns include wheels (row 5 in ProGAN \cite{karras2018progressive}, StyleGAN2 \cite{Karras_2020_CVPR}, StyleGAN \cite{Karras_2019_CVPR}) and stripes in Zebra (rows 1-5 in CycleGAN \cite{zhu2017unpaired}).
We remark that CR-Universal detectors also contain a few color-conditional T-FF.
}
\label{fig_supp:lrp_patches_cr_r50}
\end{figure}


\section{Pixel-wise explanations are not informative to discover \textit{T-FF} (Additional Results)}
\label{sec_supp:pixel-wise_explanations}
\setcounter{figure}{0} 
\setcounter{table}{0} 
In this section, we show additional results to demonstrate that direct pixel-wise explanations of universal detector decisions are not informative to discover \textit{T-FF}. 
Similar to main paper, we use 2 popular interpretation methods namely Guided-GradCAM \cite{selvaraju2017grad} and
LRP \cite{lrp} to analyse the pixel-wise explanations of universal detector decisions.
We show additional results for ResNet-50 detector in Fig. \ref{fig_supp:pixel_wise_explanations_r50}.
We also show results for EfficientNet-B0 in Fig. \ref{fig_supp:pixel_wise_explanations_efb0} and \ref{fig_supp:pixel_wise_explanations_efb0_set2}.
As one can observe from Fig. \ref{fig_supp:pixel_wise_explanations_r50}, \ref{fig_supp:pixel_wise_explanations_efb0} and \ref{fig_supp:pixel_wise_explanations_efb0_set2} pixel-wise explanations of universal detector decisions
are not informative to discover T-FF due to their focus on spatial localization.

\begin{figure}
\centering
\begin{tabular}{c @{\hskip 0.03in} 
                c @{\hskip 0.0in} c @{\hskip 0.02in}
                c @{\hskip 0.0in} c @{\hskip 0.02in}
                c @{\hskip 0.0in} c @{\hskip 0.02in} 
                c @{\hskip 0.0in} c @{\hskip 0.02in} 
                c @{\hskip 0.0in} c @{\hskip 0.0in}}
    
    \multicolumn{1}{c}{} &
    \multicolumn{2}{p{2cm}}{\tiny \quad \quad ProGAN \cite{karras2018progressive}} &
    \multicolumn{2}{p{2.2cm}}{\tiny \quad \quad StyleGAN2 \cite{Karras_2020_CVPR}} &
    \multicolumn{2}{p{2.2cm}}{\tiny \quad \quad StyleGAN \cite{Karras_2019_CVPR}} &
    \multicolumn{2}{p{2.2cm}}{\tiny \quad \quad  BigGAN \cite{brock2018large}} &
    \multicolumn{2}{c}{\tiny CycleGAN \cite{zhu2017unpaired}}
     \\
     
    \begin{turn}{90} 
    \textbf{\tiny Image}
    \end{turn} &
    
    \multicolumn{10}{c}{\includegraphics[width=0.92\linewidth]{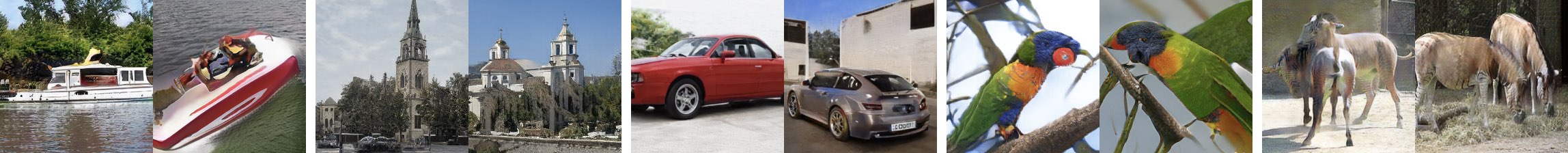}}

    \\
    
    \toprule
    \multicolumn{11}{c}{\tiny Pixel-wise explanations  of universal detector decisions \cite{Wang_2020_CVPR} using Guided-GradCAM (GGC) \cite{selvaraju2017grad} and LRP \cite{bach2015pixel} }\\
    
    \begin{turn}{90} 
    \textbf{\tiny GGC \cite{selvaraju2017grad}}
    \end{turn} &
    
    \multicolumn{10}{c}{\includegraphics[width=0.92\linewidth]{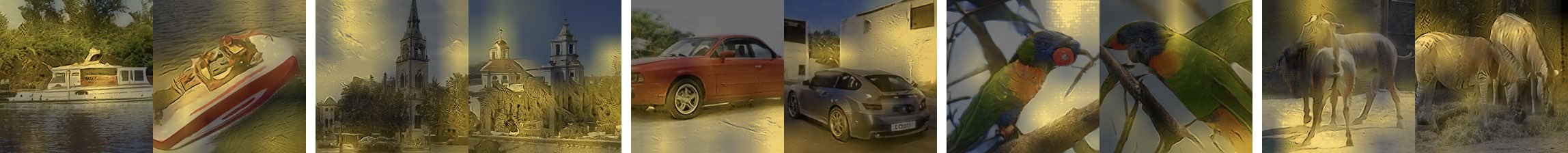}}
    
    \\
    
     \begin{turn}{90} 
    \textbf{\tiny LRP \cite{bach2015pixel}}
    \end{turn} &
    
    \multicolumn{10}{c}{\includegraphics[width=0.92\linewidth]{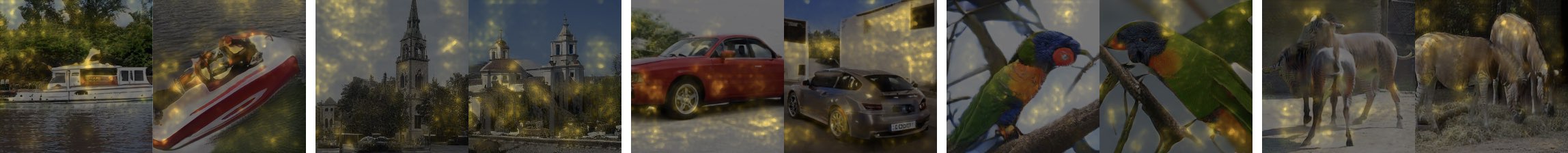}}
      
    \\
    
    \toprule
    \multicolumn{11}{c}{\tiny Pixel-wise explanations of ImageNet classifier decisions using Guided-GradCAM (GGC) \cite{selvaraju2017grad} and LRP \cite{bach2015pixel} }\\
    
    \begin{turn}{90} 
    \textbf{\tiny GGC \cite{selvaraju2017grad}}
    \end{turn} &

    \multicolumn{10}{c}{\includegraphics[width=0.92\linewidth]{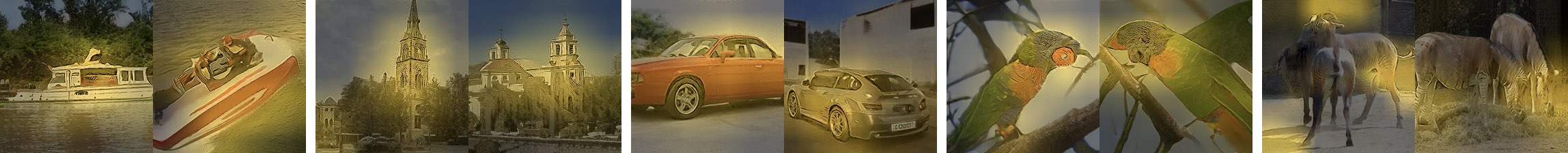}}
   
    \\
    
    \begin{turn}{90} 
    \textbf{\tiny LRP \cite{bach2015pixel}}
    \end{turn} &

    \multicolumn{10}{c}{\includegraphics[width=0.92\linewidth]{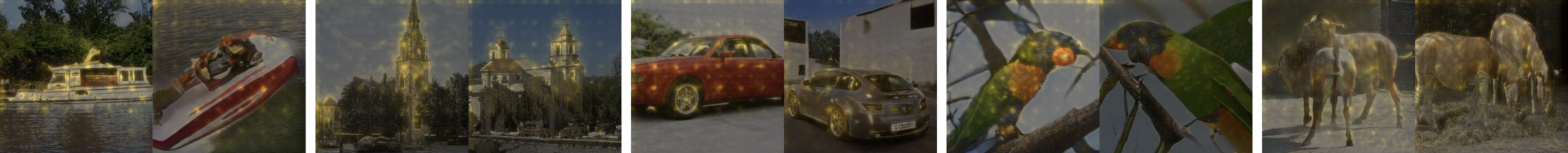}}
    
    \\

\end{tabular}
\caption{
Additional results showing that pixel-wise explanations of universal detector decisions are
not informative
to discover \textit{T-FF}:
We show pixel-wise explanations using Guided-GradCAM (GGC) (row 2) \cite{selvaraju2017grad} and LRP (row 3) \cite{lrp} for the 
ResNet-50 universal detector 
\cite{Wang_2020_CVPR} for ProGAN \cite{karras2018progressive}, CycleGAN \cite{zhu2017unpaired}, StarGAN \cite{choi2018stargan}, BigGAN \cite{brock2018large} and StyleGAN2 \cite{Karras_2020_CVPR}.
The universal detector predicts probability $p>=95\%$ for all counterfeit images shown above. 
All these counterfeits are obtained from the ForenSynths dataset 
\cite{Wang_2020_CVPR}.
For LRP \cite{lrp}, we only show the positive relevances.
We also show the pixel-wise explanations of ImageNet classifier decisions for the exact counterfeits using GGC (row 4) and LRP (row 5). This is shown as a control experiment to emphasize the significance of our observations.
As one can clearly observe, pixel-wise explanations of universal detector decisions are 
not informative
to discover \textit{T-FF} (row 2 and 3) as the explanations appear to be random and not reveal any meaningful visual features used for counterfeit detection.
Particularly, it remains unknown as to why the universal detector outputs high detection probability ($p>=95\%$) for these counterfeits.
On the other hand, pixel-wise explanations of ImageNet classifier decisions produce meaningful results. 
i.e.: The GGC (row 4) and LRP (row 5) explanation results for car samples (columns 5, 6) show that ImageNet uses features such as wheels/body to classify cars.
This clearly shows that interpretability techniques such as GGC and LRP are 
not informative
to discover \textit{T-FF} in universal detectors.
In other words, we are unable to discover any forensic footprints based on pixel-wise explanations of universal detectors.
}
\label{fig_supp:pixel_wise_explanations_r50}
\end{figure}



\begin{figure}[!h]
\centering
\begin{tabular}{c @{\hskip 0.03in} 
                c @{\hskip 0.0in} c @{\hskip 0.02in}
                c @{\hskip 0.0in} c @{\hskip 0.02in}
                c @{\hskip 0.0in} c @{\hskip 0.02in} 
                c @{\hskip 0.0in} c @{\hskip 0.02in} 
                c @{\hskip 0.0in} c @{\hskip 0.0in}}
    
    \multicolumn{1}{c}{} &
    \multicolumn{2}{p{2cm}}{\tiny \quad \quad ProGAN \cite{karras2018progressive}} &
    \multicolumn{2}{p{2.2cm}}{\tiny \quad \quad StyleGAN2 \cite{Karras_2020_CVPR}} &
    \multicolumn{2}{p{2.2cm}}{\tiny \quad \quad StyleGAN \cite{Karras_2019_CVPR}} &
    \multicolumn{2}{p{2.2cm}}{\tiny \quad \quad  BigGAN \cite{brock2018large}} &
    \multicolumn{2}{c}{\tiny CycleGAN \cite{zhu2017unpaired}}
     \\
     
    \begin{turn}{90} 
    \textbf{\tiny Image}
    \end{turn} &
    
    \multicolumn{10}{c}{\includegraphics[width=0.92\linewidth]{pics_supp/interpretability/r50_0.5/row1.001.jpeg}}

    \\
    
    \toprule
    \multicolumn{11}{c}{\tiny Pixel-wise explanations  of universal detector decisions \cite{Wang_2020_CVPR} using Guided-GradCAM (GGC) \cite{selvaraju2017grad} and LRP \cite{bach2015pixel} }\\
    
    \begin{turn}{90} 
    \textbf{\tiny GGC \cite{selvaraju2017grad}}
    \end{turn} &
    
    \multicolumn{10}{c}{\includegraphics[width=0.92\linewidth]{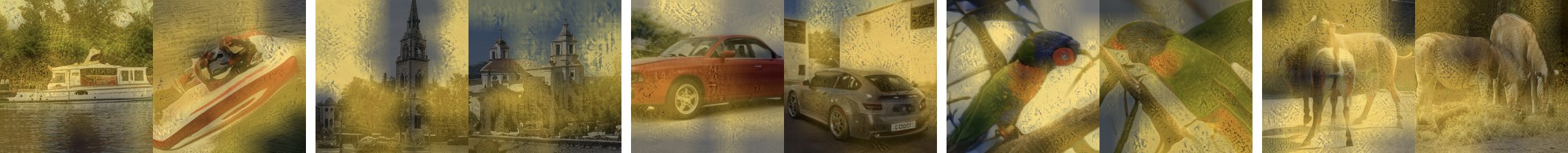}}
    
    \\
    
     \begin{turn}{90} 
    \textbf{\tiny LRP \cite{bach2015pixel}}
    \end{turn} &
    
    \multicolumn{10}{c}{\includegraphics[width=0.92\linewidth]{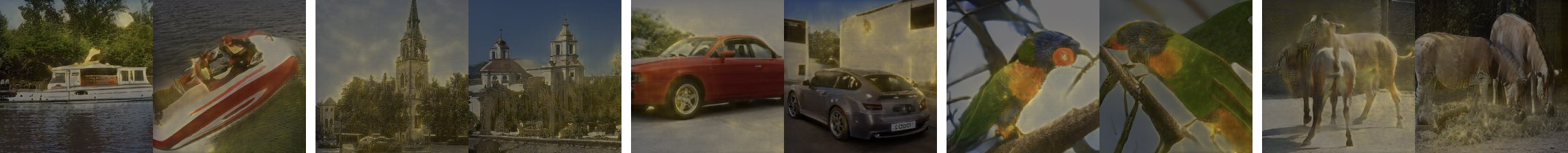}}
      
    \\
    
    \toprule
    \multicolumn{11}{c}{\tiny Pixel-wise explanations of ImageNet classifier decisions using Guided-GradCAM (GGC) \cite{selvaraju2017grad} and LRP \cite{bach2015pixel} }\\
    
    \begin{turn}{90} 
    \textbf{\tiny GGC \cite{selvaraju2017grad}}
    \end{turn} &

    \multicolumn{10}{c}{\includegraphics[width=0.92\linewidth]{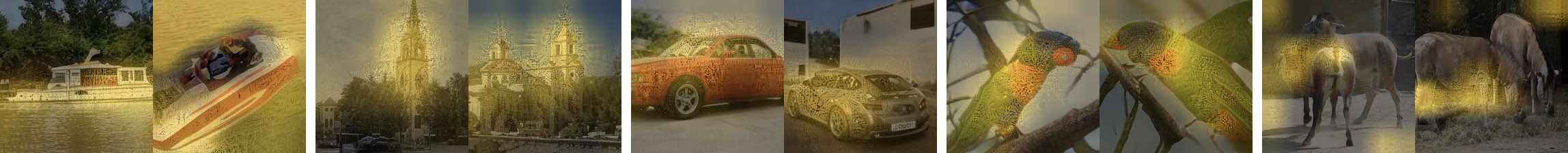}}
   
    \\
    
    \begin{turn}{90} 
    \textbf{\tiny LRP \cite{bach2015pixel}}
    \end{turn} &

    \multicolumn{10}{c}{\includegraphics[width=0.92\linewidth]{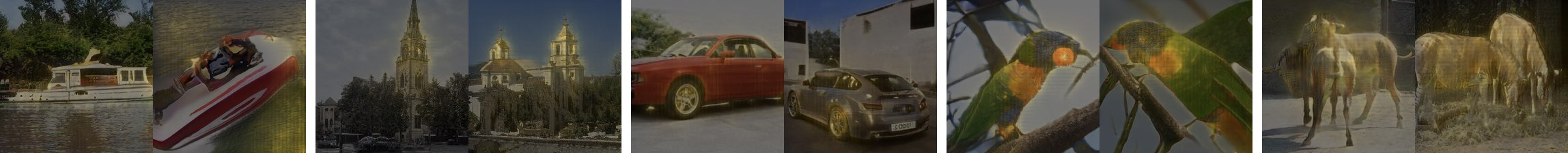}}
    
    \\

\end{tabular}
\caption{
Additional results showing that pixel-wise explanations of universal detector decisions are
not informative
to discover \textit{T-FF} (EfficientNet-B0):
We show pixel-wise explanations using Guided-GradCAM (GGC) (row 2) \cite{selvaraju2017grad} and LRP (row 3) \cite{lrp} for 
our version of EfficientNet-B0 universal Detector following the exact training / test strategy proposed in
\cite{Wang_2020_CVPR} for ProGAN \cite{karras2018progressive}, CycleGAN \cite{zhu2017unpaired}, StarGAN \cite{choi2018stargan}, BigGAN \cite{brock2018large} and StyleGAN2 \cite{Karras_2020_CVPR}.
The universal detector predicts probability $p>=95\%$ for all counterfeit images shown above. 
All these counterfeits are obtained from the ForenSynths dataset 
\cite{Wang_2020_CVPR}.
For LRP \cite{lrp}, we only show the positive relevances.
We also show the pixel-wise explanations of ImageNet classifier decisions for the exact counterfeits using GGC (row 4) and LRP (row 5). This is shown as a control experiment to emphasize the significance of our observations.
As one can clearly observe, pixel-wise explanations of universal detector decisions are 
not informative
to discover \textit{T-FF} (row 2 and 3) as the explanations appear to be random and not reveal any meaningful visual features used for counterfeit detection.
Particularly, it remains unknown as to why the universal detector outputs high detection probability ($p>=95\%$) for these counterfeits.
On the other hand, pixel-wise explanations of ImageNet classifier decisions produce meaningful results. 
i.e.: The GGC (row 4) and LRP (row 5) explanation results for car samples (columns 5, 6) show that ImageNet uses features such as wheels / body to classify cars.
This clearly shows that interpretability techniques such as GGC and LRP are 
not informative
to discover \textit{T-FF} in universal detectors.
In other words, we are unable to discover any forensic footprints based on pixel-wise explanations of universal detectors.
}
\label{fig_supp:pixel_wise_explanations_efb0}
\end{figure}

\begin{figure}[!h]
\centering
\begin{tabular}{c @{\hskip 0.03in} 
                c @{\hskip 0.0in} c @{\hskip 0.02in}
                c @{\hskip 0.0in} c @{\hskip 0.02in}
                c @{\hskip 0.0in} c @{\hskip 0.02in} 
                c @{\hskip 0.0in} c @{\hskip 0.02in} 
                c @{\hskip 0.0in} c @{\hskip 0.0in}}
    
    \multicolumn{1}{c}{} &
    \multicolumn{2}{p{2cm}}{\tiny \quad \quad ProGAN \cite{karras2018progressive}} &
    \multicolumn{2}{p{2.2cm}}{\tiny \quad \quad StyleGAN2 \cite{Karras_2020_CVPR}} &
    \multicolumn{2}{p{2.2cm}}{\tiny \quad \quad StyleGAN \cite{Karras_2019_CVPR}} &
    \multicolumn{2}{p{2.2cm}}{\tiny \quad \quad  BigGAN \cite{brock2018large}} &
    \multicolumn{2}{c}{\tiny CycleGAN \cite{zhu2017unpaired}}
     \\
     
    \begin{turn}{90} 
    \textbf{\tiny Image}
    \end{turn} &
    
    \multicolumn{10}{c}{\includegraphics[width=0.92\linewidth]{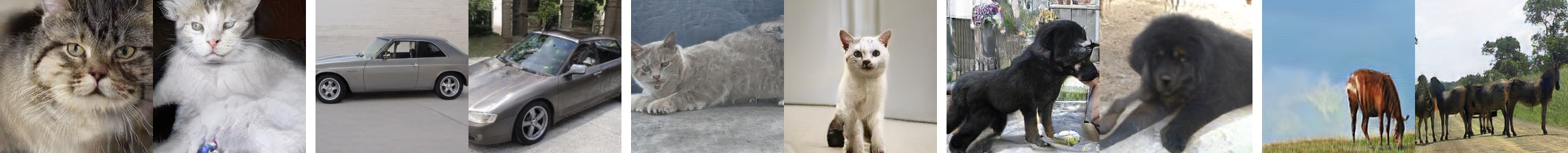}}

    \\
    
    \toprule
    \multicolumn{11}{c}{\tiny Pixel-wise explanations  of universal detector decisions \cite{Wang_2020_CVPR} using Guided-GradCAM (GGC) \cite{selvaraju2017grad} and LRP \cite{bach2015pixel} }\\
    
    \begin{turn}{90} 
    \textbf{\tiny GGC \cite{selvaraju2017grad}}
    \end{turn} &
    
    \multicolumn{10}{c}{\includegraphics[width=0.92\linewidth]{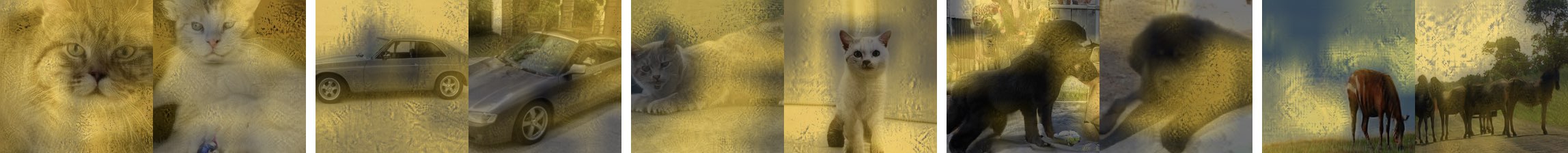}}
    
    \\
    
     \begin{turn}{90} 
    \textbf{\tiny LRP \cite{bach2015pixel}}
    \end{turn} &
    
    \multicolumn{10}{c}{\includegraphics[width=0.92\linewidth]{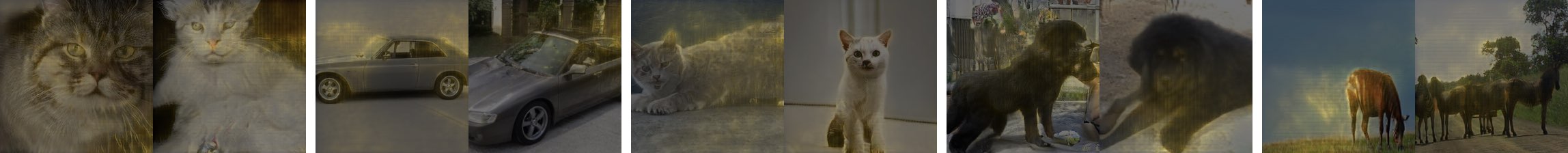}}
      
    \\
    
    \toprule
    \multicolumn{11}{c}{\tiny Pixel-wise explanations of ImageNet classifier decisions using Guided-GradCAM (GGC) \cite{selvaraju2017grad} and LRP \cite{bach2015pixel} }\\
    
    \begin{turn}{90} 
    \textbf{\tiny GGC \cite{selvaraju2017grad}}
    \end{turn} &

    \multicolumn{10}{c}{\includegraphics[width=0.92\linewidth]{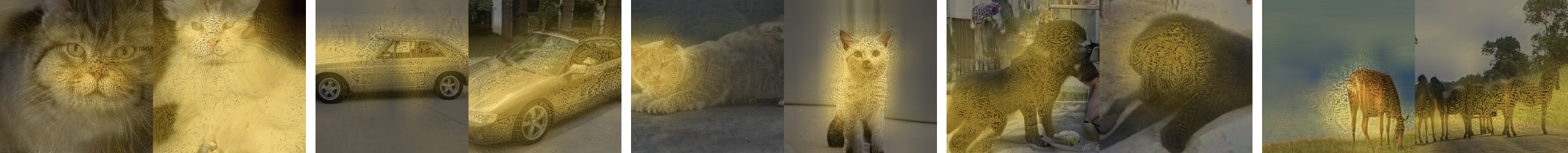}}
   
    \\
    
    \begin{turn}{90} 
    \textbf{\tiny LRP \cite{bach2015pixel}}
    \end{turn} &

    \multicolumn{10}{c}{\includegraphics[width=0.92\linewidth]{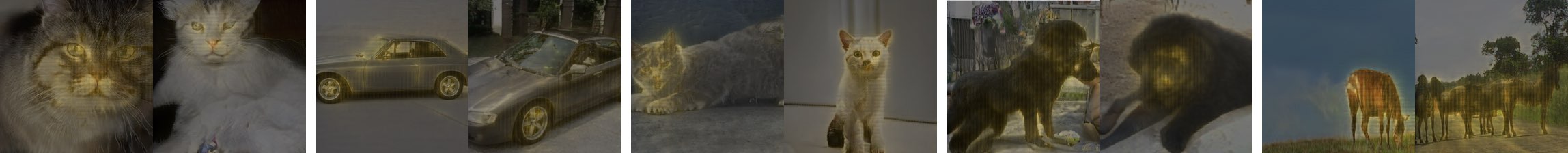}}
    
    \\

\end{tabular}
\caption{
Additional results showing that pixel-wise explanations of universal detector decisions are
not informative
to discover \textit{T-FF} (EfficientNet-B0):
We show pixel-wise explanations using Guided-GradCAM (GGC) (row 2) \cite{selvaraju2017grad} and LRP (row 3) \cite{lrp} for 
our version of EfficientNet-B0 universal Detector following the exact training / test strategy proposed in
\cite{Wang_2020_CVPR} for ProGAN \cite{karras2018progressive}, CycleGAN \cite{zhu2017unpaired}, StarGAN \cite{choi2018stargan}, BigGAN \cite{brock2018large} and StyleGAN2 \cite{Karras_2020_CVPR}.
The universal detector predicts probability $p>=95\%$ for all counterfeit images shown above. 
All these counterfeits are obtained from the ForenSynths dataset 
\cite{Wang_2020_CVPR}.
For LRP \cite{lrp}, we only show the positive relevances.
We also show the pixel-wise explanations of ImageNet classifier decisions for the exact counterfeits using GGC (row 4) and LRP (row 5). This is shown as a control experiment to emphasize the significance of our observations.
As one can clearly observe, pixel-wise explanations of universal detector decisions are 
not informative
to discover \textit{T-FF} (row 2 and 3) as the explanations appear to be random and not reveal any meaningful visual features used for counterfeit detection.
Particularly, it remains unknown as to why the universal detector outputs high detection probability ($p>=95\%$) for these counterfeits.
On the other hand, pixel-wise explanations of ImageNet classifier decisions produce meaningful results. 
i.e.: The GGC (row 4) and LRP (row 5) explanation results for cat samples (columns 1, 2, 5, 6) show that ImageNet uses features such as eyes and whiskers to classify cats.
This clearly shows that interpretability techniques such as GGC and LRP are 
not informative
to discover \textit{T-FF} in universal detectors.
In other words, we can not discover any forensic footprints based on pixel-wise explanations of universal detectors.
}
\label{fig_supp:pixel_wise_explanations_efb0_set2}
\end{figure}

\section{Research Reproducibility / Code Details}
\label{sec_supp:reproducibility}
\setcounter{figure}{0} 
\setcounter{table}{0} 

\textbf{Code:} Pytorch code is available at
\href{https://keshik6.github.io/transferable-forensic-features/}{here}.
Refer to README
for step-by-step instructions. 
The codebase is clearly documented. 
The code is structured as follows:

\begin{itemize}

    \item \textbf{lrp/:} Base Pytorch module containing LRP implementations for ResNet and EfficientNet architectures. This includes all Pytorch wrappers.
    
    \item \textbf{fmap\_ranking/:} Pytorch module to calculate \textit{FF-RS ($\omega$)} for counterfeit detection.
    
    \item \textbf{sensitivity\_assessment/:} Pytorch module to perform sensitivity assessments for \textit{T-FF} and color ablation.
    
    \item \textbf{patch\_extraction/:} Pytorch module to extract LRP-max response image regions for every \textit{T-FF}.
    
    \item \textbf{activation\_histograms/:} Pytorch module to calculate maximum spatial activation for images for every \textit{T-FF}.

    \item \textbf{utils/:} Contains all utilities, helper functions and plotting functions.
    
\end{itemize}

\textbf{Pre-trained models:} 
All pretrained models can be found at 
\href{https://keshik6.github.io/transferable-forensic-features/}{here}.
We provide both ResNet-50 and EfficientNet-B0 pretrained universal detectors. 
We also include CR-universal detector models.
All our claims reported in Main / Supplementary can be reproduced using these checkpoints.

\textbf{Docker information:} 
For training /analysis in containerised environments (HPC, Super-computing clusters), please use
nvcr.io/nvidia/pytorch:20.12-py3 container.

\textbf{Experiment details and hyper-parameters:}
For training universal detectors, we use the exact setup proposed in \cite{Wang_2020_CVPR} with Adam optimizer ($\beta_1 = 0.9, \beta_2 = 0.999$), batch size of 64 and initial learning rate of $1e^{-4}$.
For data augmentation, we use the exact setup proposed in \cite{Wang_2020_CVPR} that includes random cropping (224x224), random horizontal flip and 50\% JPEG + Blurring.
All experiments were repeated 3 times.
For LRP, we use $\beta=0$ rule. For statistical tests, we use Mood's median test with a significance level of $\alpha = 0.05$.

\clearpage
\section{Future Work: Can we identify globally relevant channels for counterfeit detection in a Generator?}
\label{sec_supp:future_work}
\setcounter{figure}{0} 
\setcounter{table}{0} 
\vspace{-1cm}
\begin{figure*}[!h]
\centering
\includegraphics[width=0.7\linewidth]{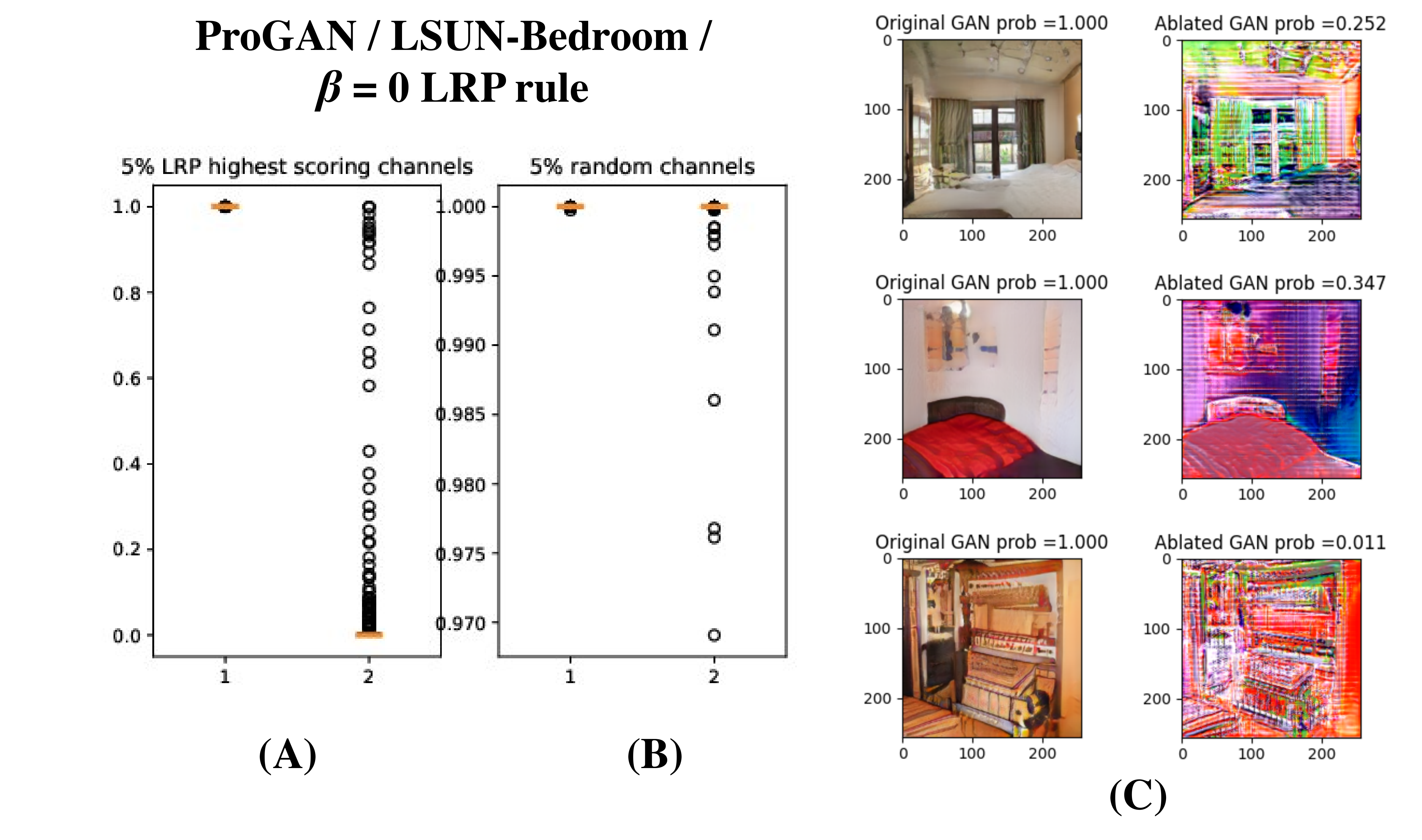}\\
\vspace{-0.68cm}
\caption{
\textbf{(A)} ResNet-50 detector \cite{Wang_2020_CVPR} scores before and after masking the 5\% channels in the generator according to highest LRP scores computed for the generator. 
\textbf{(B)} ResNet-50 detector \cite{Wang_2020_CVPR} scores before and after masking the 5\% channels selected randomly in the generator. The orange line depicts the median of the box plot. Higher difference between both box plots within a subplot is better. Computed over 500 generated images trained over the LSUN-Bedrooms \cite{yu15lsun} class using a ProGAN \cite{karras2018progressive}. 
One can see that masking 5\% channels found by LRP in the generator leads to a very strong drop in detector scores (A) compared to masking 5\% randomly selected channels results in a much smaller score decrease (B).
\textbf{(C)} Original and ablated GAN samples with corresponding detector probabilities.
}
\label{fig_supp:gan_ablation}
\end{figure*}

\vspace{-0.8cm}
This section serves to motivate future directions from an image synthesis perspective. 
Particularly, we ask the question as to whether it's possible to identify feature maps in \textit{GANs} that are responsible for generating forensic features that are detected by universal detectors.

In this section, we show preliminary results suggesting that it's possible to identify such globally relevant channels in a generator. 
Particularly, we perform LRP all the way into the Generator to identify the top highest scoring GAN channels that are responsible for counterfeit detection (i.e.: In the computational graph, the image is generated from a pre-trained ProGAN \cite{karras2018progressive} model).
We show that ablating these top-scoring GAN channels consequently results in large drop in probability predicted by the universal detector (We use the publicly released ResNet-50 in this experiment). This result is shown in Fig. \ref{fig_supp:gan_ablation} that propagating LRP into the generator is able to identify the globally top-5\% relevant channels for images. 
The box plot (A) shows a strong decrease after ablating these high-scoring GAN channels (though ablated GAN samples have poor visual quality). This can be compared to (B) where 5\% of randomly selected GAN channels are ablated, which results in a very small decrease in counterfeit detection scores. 
These results show promising directions for understanding image synthesis methods, and we hope to explore this area in future work.
We also hope to explore the properties of Fair Generative models \cite{xu2018fairgan,8869910fairnessgan,teo2021measuring,choi2020fair,tan2020improving}, GANs / detectors trained using different techniques (regularization, knowledge transfer, pruning, few-shot learning, self-supervised learning) \cite{pmlr-v162-chandrasegaran22a,abdollahzadeh2021revisit,YEOM2021107899,he2020momentum} and face-forgery detectors \cite{kim2021fretal,chen2022self,shiohara2022detecting,zhao2021multi,zhao2021learning,li2020face,haliassos2021lips}.

\clearpage

\end{document}